Thesis n° 11 559

# Studying Maps at Scale: A Digital Investigation of Cartography and the Evolution of Figuration

Presented on 31st October 2025

College of humanities
Digital Humanities Laboratory
Doctoral program in Digital humanities

for the award of the degree of Docteur ès Sciences (PhD)

by

**Rémi Guillaume PETITPIERRE**

Accepted on the jury's recommendation

Prof. J. Baudry, jury president
Prof. F. Kaplan, Dr I. Di Lenardo, thesis directors
Prof. L. Isaksen, examiner
Prof. S. Leyk, examiner
Dr M. Salzmann, examiner

# Table of Contents

# List of Figures

# List of Tables

# Abstract

This aim of this thesis is to develop methods and create datasets to investigate cartographic heritage on a large scale and from a cultural perspective. In recent years, heritage institutions worldwide have digitized more than one million maps. Specific techniques have been developed for the automated recognition and extraction of map image content, opening new avenues in historical geographic research. However, thus far, these methods have engaged little with the view that maps are cultural objects, or with the history of cartography itself. Recognizing maps as cultural objects implies that they are not objective, undistorted records of territory but purposeful representations affected by political, epistemic, and artistic expectations. Moreover, cartography functions as a semantic-symbolic system that relies on cultural conventions evolving over time.

This work leverages a large, diverse corpus comprising 771,561 map records and 99,715 digitized images, aggregated from 38 digital catalogs across 11 countries. After normalization, the dataset includes 236,925 distinct contributors—mapmakers or publishers—and spans six centuries, from 1492 to 1948. These extensive data make it possible to chart and visualize the geographic structures and global chronology of map publication. In addition, the spatial focus of cartography on particular regions is analyzed in relation to political dynamics, evidencing a conjunctural relationship between Atlantic maritime charting, the triangular trade, and colonial expansion. Further results document the historical progression of national, domestic focus and the impact of military conflicts on publication volumes.

The present research also introduces state-of-the-art semantic segmentation techniques and detection models for the generic recognition of land classes and cartographic signs. The training relies on dedicated sets of annotated data, augmented with synthetic images, for segmentation. The analysis of semantic land classes—such as buildings, water, and roads—shows that maps are designed images: framing and composition emphasize particular geographic features through centering and semantic symmetries.

The study of cartographic figuration relies on the encoding of 63 million signs—such as icons and symbols—and 25 million fragments—depicting texture, linework, and color—into a visual, latent space. The distribution of signs and fragments across periods reveals moments of rupture, denoting figurative shifts, like the replacement of relief hachures by terrain contours. Despite a trend toward diversification, signs tend to form locally consistent, univocal systems. The analysis of social transmission through collaboration underscores the influence of legitimacy and the role of larger actors in the emergence of figurative norms. Finally, models of diffusion across publication centers show that larger cities tend to develop distinctive, differentiated semiotic cultures and function as transmission hubs.

To conclude, this thesis presents large datasets and computational methodologies to study the influence of political power, technological change, semiotic constraints and cultural transmission on the evolution of cartography.

# Résumé

Cette thèse vise à créer des bases de données et à développer des méthodes pour étudier le patrimoine cartographique à grande échelle. Au cours des dernières années, les institutions patrimoniales à travers le monde ont numérisé plus d'un million de cartes. Des technologies spécifiques ont été mises au point permettant la reconnaissance automatique du contenu des images, avec des applications en géographie historique et en économie spatiale. Cependant, jusqu'à présent, ces approches n'ont que peu bénéficié à l'histoire de la cartographie. De plus, elles manquent à traiter les cartes comme des objets culturels, soit comme des représentations subjectives et imparfaites, influencées par les développements politiques, économiques, épistémiques, et artistiques. En outre, la cartographie fonctionne comme un système sémantique et symbolique reposant sur des conventions culturelles qui elles-mêmes varient au fil du temps.

Ce travail s'appuie sur un corpus vaste et diversifié comprenant 771,561 notices bibliographiques et 99,715 images de cartes numérisées. Ce corpus est issu de quelques 38 catalogues numériques situés dans 11 pays. Après leur normalisation, les données comptent 236,925 contributeurs distincts (cartographes ou éditeurs) et couvrent six siècles, de 1492 à 1948. Cette masse de données permet de cartographier et de visualiser les structures géographiques et chronologiques de l'histoire de la cartographie. L'attention portée par les cartes à certaines régions géographiques est étudiée en regard de certains développements politiques et historiques, soulignant par exemple la relation conjoncturelle entre cartographie de l'Atlantique, commerce triangulaire et expansion coloniale. Les résultats révèlent également l'importance croissante accordée à la cartographie nationale et régionale propre ainsi que l'impact des conflits militaires sur les volumes de publication.

Cette recherche introduit également des modèles de pointe pour la segmentation sémantique des classes de terrain et la détection des signes cartographiques. L'entraînement s'appuie sur des annotations spécifiques, enrichies d'images synthétiques. L'analyse des classes sémantiques—telles que les bâtiments, l'eau et les routes—montre que les cartes sont des images conçues : le cadrage et la composition tendent à souligner certaines classes d'objets par des effets de centrage ou via des symétries sémantiques.

L'étude de la figuration cartographique s'appuie sur quelques 63 millions de signes, icônes ou symboles, et 25 millions de fragments représentant la texture, le tracé et la couleur. Ces éléments sont encodés dans un espace visuel latent et leur distribution temporelle est utilisée pour identifier des moments de rupture, indiquant des évolutions figuratives, comme le remplacement progressif des hachures par des courbes de niveaux, pour la représentation du dénivelé. En dépit d'une tendance à la diversification, les signes forment généralement des systèmes localement cohérents et univoques. L'analyse des transmissions sociales souligne l'importance de la posture de légitimité et le rôle des acteurs plus importants dans l'émergence des conventions figuratives. Finalement, la modélisation des diffusions entre centres de publication montre que les grandes villes ont tendance à développer des cultures sémiotiques distinctives et reconnaissables, et à fonctionner comme des hubs de transmission.

Pour résumer, cette thèse introduit des méthodologies computationnelles qui, appliquées à de grands ensembles de données, permettent d'étudier l'influence du pouvoir, des innovations technologiques, des contraintes sémiotiques et de la transmission culturelle sur l'évolution de la cartographie.

# Acknowledgements

First, I would like to sincerely thank the examination committee for their interest, commitment, and diligence: the experts, Stefan Leyk, Leif Isaksen, Mathieu Salzmann, and the president, Jérôme Baudry. I am also deeply grateful to my supervisors, Frédéric Kaplan and Isabella di Lenardo, for giving me the opportunity to explore a subject I am passionate about, for the trust they placed in me in agreeing to embark on a research project that long seemed risky, for their boundless enthusiasm, and their unwavering encouragement. I would also like to thank Béla Kapossy for accompanying and supporting me during the first steps of this rewarding journey.

Next, I wish to thank my students, especially Damien Donoso-Gomez, Ben Kriesel, Jiaming Jiang, and Danyang Wang, for their help in annotating some of the data used in this thesis. I would also like to acknowledge and thank my colleagues at the Time Machine Unit, who supported me at various stages along the way: Lucas Rappo, Nicolas Mermoud-Ghraichy, Irene Bianchi, Cédric Viaccoz, Marion Kramer, and Isabelle Hügli, as well as my colleagues at the EPFL Digital Humanities Institute, particularly Beatrice Vaienti, Paul Guhennec, Raimund Schnürer, Didier Dupertuis, Raphaël Barman, Sven Najem-Meyer, Maud Ehrmann, Manuel Ehrenfeld, Gaël Paccard, and Alicia Foucart. I also extend my thanks to my colleagues from dhelta, the Digital Humanities Association of EPFL and the University of Lausanne, particularly Clémence Danesi, Thalie Schmid, Nicolas Bovet, Ravinithesh Annapureddy, Théo Rochat, and Miguel Almeida Henriques, for their lasting commitment to the vivid community.

In addition, I want to acknowledge a few colleagues outside EPFL, co-authors, conference friends, or researchers whose work has been especially valuable in shaping my thinking on digital scholarship and historical maps. Meeting you made this journey more exciting, motivating, and rewarding: Johannes Uhl, Yizi Chen, Gabriel Loumeau, Bertrand Duménieu, Nathalie Abadie, Solenn Tual, Lorenz Hurni, Sidi Wu, Magnus Heitzler, Xue Xia, Chengjing Jiao, Katherine McDonough, Polly Hudson, Hendrik Herold, Yanos Zylberberg, Valentine Bernasconi, and Ludovic Pollet.

Finally, I wish to thank, from the bottom of my heart, my family, who patiently accompanied me through these four years of doctoral work and the long months of writing, and who carefully proofread several chapters. I truly owe you everything. Warm thanks as well to my friends outside academia, who supported me and encouraged me unconditionally.

**Funding.** This thesis was entirely funded by public funds. Half of the funding came from a two-year fully funded PhD scholarship (129,850 CHF) awarded by the College of Humanities at EPFL on the basis of the quality of the doctoral project proposal.

**Declaration of competing interests.** The author declares that he has no conflicts of interest.

**Note on the use of Artificial Intelligence.** Artificial intelligence (AI) agent technologies and language models were occasionally employed in the preparation of this thesis, in particular DeepL's translation models and OpenAI's GPT-4o. These models were utilized to translate texts from French into English, to propose more idiomatic alternative formulations, to ensure consistency in references, for text correction, and proofreading. They were also applied to generate code for standard and readily verifiable tasks, such as data visualization, always under human supervision. Conversely, AI was not employed for the production of scientific content, such as the description of the methodology, the analysis or discussion of results, or the formulation of conclusions.

# Introduction

## I.1 Context

*The* History of Cartography *series is generally regarded as one of the most significant scholarly undertakings in the history of maps. The project was conceived in 1977, with the first volume published in 1987. The second, third, sixth, and fourth volumes appeared in 1995, 2007, 2015, and 2020, respectively. Presently, the work remains in progress, with the fifth volume expected in 2027, almost fifty years after David Woodward and John B. Harley initiated the project. The third, fourth, and sixth volumes are of particular relevance to this dissertation as they examine the history of cartography during the European Renaissance, the Enlightenment, and the 20$^{th}$ century. Together, these three volumes comprise more than 6,100 pages and 3,160 images. They furthermore include contributions from over 400 authors.*

The *History of Cartography* comes as close as possible to a comprehensive history of maps. A few years ago, the *New York Times* described it as "the most ambitious overview of map making ever undertaken" (Rothstein, 1999). Beyond its scale, the series possesses several noteworthy qualities that speak for the entire field. First, it presents a diversity of viewpoints by giving voice to authors from multiple generations and from both Western and non-Western contexts. Second, it adopts a broad disciplinary perspective. Although centered on the history of cartography, the volumes incorporate the interdisciplinary perspectives of cultural history, the history of ideas, the history of science and technology, the history of printing, art history, historical geography, economic history, postcolonial studies, and the sociology of power and space. Third, the series ambitions to relate the particular contexts in which maps have been produced to a bigger picture. It also engages with an extensive chronology, adopts a cross-cultural focus, and addresses a wide range of topics with an inclusive understanding of cartography. Finally, its commitment to open-access format and its full search functionality are examples of the most commendable academic practices.

Despite its depth, the *History of Cartography* does not constitute the endpoint of research on the history of maps. To the reader, it might appear that this PhD dissertation begins at the top of the preceding page; it does not. In fact, the dissertation truly starts on page 8. While you might have passed it without paying much attention, page 8 contains an eloquent visualization. Indeed, if we were to consider every map ever created on the one hand, and every map displayed in the *History of Cartography* on the other, and if we were to draw a single black pixel for each of the latter on a page, we would obtain an image close to that shown on page 8. Indeed, despite its considerable size and the vast endeavor it represents, a conservative estimate would conclude that the *History of Cartography* contains less than 0.02% of all maps ever created[1].

### *Digitizing map archives*

Over the past few decades, most large libraries, geographic societies, mapping agencies, and heritage institutions, undertook the sheer endeavor to digitize their cartographic archives. However, as maps are difficult objects to store—given the variety of dimensions and conservation formats, from rolls to atlases and oversized maps mounted on canvas—digitizing cartographic documents has been an intensive and lengthy task. The debate on how to document maps in digital notices is not straightforward either. For instance, does georeferencing historical maps and presenting them as raster geographic layers make sense? How should roles and map authorship be described? What date should be associated with the document: the date of survey or the date of publication? How to handle graphical scales? These questions have resulted in as many different answers and ways of documenting and disseminating maps in digital portals.

As will be shown in Chapter 1, over a million map records are already available in the main digital library catalogs worldwide, corresponding effectively to several million maps[2]. This represents only a small fraction of the historical map production that remains to be documented and digitized. Thus, the number of digitized map documents continues to grow as digitization efforts proceed.

### *Maps as data*

The extent of digitized cartographic collections and the density of information they contain increasingly led scholars to consider maps as minable data (McDonough, 2024). Some of the earliest works attempted to extract simple shapes, lines, and polygons (Cofer & Tou, 1972; Kasturi & Alemany, 1988). Maps presented interesting challenges for computer vision, like the superimposition of multiple graphical layers of information, color drift and fading, discontinuous lines, and unclosed polygons, as well as the tension between digitization quality, image size, and large elements that sometimes span the entire document. This led engineers to develop a variety of specialized techniques, leveraging texture, color, and morphology, for the recognition of cartographic features (Brügelmann, 1996; Khotanzad & Zink, 2003; Yamada et al., 1993). These efforts really took off after the arrival of convolutional neural networks and their first applications to historical maps (Heitzler & Hurni, 2020; Oliveira et al., 2019; Petitpierre, 2020; Uhl et al., 2020). Although research is still ongoing, neural models can now segment and semanticize map series with reasonable accuracy and automate the creation of vector geographic layers from cartographic archives. Beyond geographic objects, researchers have developed a growing interest in extracting map text (Chiang, 2017; McDonough et al., 2024). Text detection augments the searchability of processed map data as demonstrated by the large–scale extraction of place names from the David Rumsey Map Collection, or more recently by the extraction of London 1890s plan published by the Ordnance Survey (McDonough et al., 2023; Olson et al., 2024; Zou et al., 2025b). Often, textual data constitutes the missing link between maps and other archival sources. Through street names, house numbers, owner names or plot identifiers, information can be linked to censuses, trade registers, almanacs, and directories, creating dense and spatialized historical databases on socio–economic activities. Map recognition technologies have enabled to reconstitute the historical territory of past cities with extensive planar detail, and even to create three–dimensional models thereof (Chen, 2023; Petitpierre et al., 2024a; Vaienti et al., 2023). They made it possible to trace the development of cities, road networks or railways at the scale of entire nations, over decades (Heitzler & Hurni, 2020; Jiao et al., 2022; López-Rauhut et al., 2025; Uhl et al., 2021, 2025; Xia et al., 2023). At a time when glaciers are retreating and wetlands are disappearing, these techniques provide an opportunity to monitor environmental changes and, perhaps, address them (Levin et al., 2025; Mazier et al., 2015). One of the final frontiers of map recognition, that of map signs, icons and symbols, is now being crossed as well, opening the possibility of remapping trees that stood a century ago, or water mills that turned during the French Enlightenment (Smith et al., 2025; Sutty & Duménieu, 2024).

[1] Indeed, the History of Cartography series currently contains a little over 4,900 images. To provide an order of magnitude, the 10 largest map collections worldwide already report over 21 million documents—albeit not all of them are listed in online catalogs. While some of these maps may be duplicates, many documents are also stored in local or national collections.

[2] A record may correspond to several maps, particularly in the case of atlases, map series kept in the same binder, or double-sided documents. To provide a general order of magnitude, each record is documenting *de facto* an average 3–4 maps.

While research continues, data mined from maps is already attracting interest for their potential applications in historical geography, economic geography, urban economics, sociology and social history, urban history, and environmental sciences (Chiang et al., 2020; Combes et al., 2022; di Lenardo et al., 2021; Lane et al., 2022). In these fields, the prospect of studying large-scale spatial or spatialized phenomena over several centuries appears both exciting and promising. Among these possibilities, however, one field has so far derived surprisingly little benefit from map recognition technologies: the history of cartography. One possible explanation lies in the fact that historians of cartography tend to view maps as *cultural objects* rather than minable geohistorical data (McDonough, 2024).

## *Maps as cultural objects*

Recognizing maps as cultural objects implies the realization that maps are not objective, undistorted and unambiguous records of the Earth (Harley, 1989); mapping is, on the contrary, an *intentional* process that necessarily reflects the cartographer's perspective. For instance, maps never aim to represent *everything*; they are purposeful, selective, representations. First, cartographers must choose what will be depicted on the map: buildings? bridges? borders? What about blueberry bushes? benches? boats? secret bunkers? Second, they must determine what territory the map will describe and therefore define a bounded space. As with features, establishing the map frame involves a subjective evaluation of what is "important" and what is not—what should lie at the center of the image and which features can be interrupted at its border. Assuming the space depicted is not fictional but exists in the physical world, cartographers must also decide how to *project* that three-dimensional space onto a two-dimensional document. Maps do not always result from mathematical transformations, and even when they do, projection inevitably introduces some distortion. Finally, mapmakers must decide *how* to depict the selected features, which ones to differentiate, and which ones to emphasize with graphical means. This process generally entails the definition of a set of icons, symbols, and colors that together form a visually consistent representation system.

For each of these choices, mapmakers are influenced by the maps they have seen, the people they work with, the community in which they live, and their own conception of the space being represented. Consequently, mapmakers often tend to replicate preexisting symbols. This practice fosters the emergence of shared cultural conventions within which symbols such as ♁, ☼, or ◬ acquire particular *meanings*. Even when mapmakers intentionally seek to distinguish themselves, their selection of symbols remains tied to a specific cultural frame. For example, a mapmaker might design a new icon to represent a vine, such as the sign $. Yet, why choose to differentiate vines graphically rather than blueberry bushes? Every decision, deliberate or otherwise, occurs within a particular cultural context.

In this perspective, cartography is not a closed system. Social, cultural, and political contexts influence the work of mapmakers. For instance, for whom is the map intended? Who is sponsoring its production? How might the map appear more attractive to this audience? How might it look more truthful, rational, and reliable? Moreover, because map production is often a particularly lengthy and costly undertaking, choices are also shaped by available resources (Sponberg Pedley, 2005a; Woodward, 1996). Certain pigments may be especially expensive, and engraving certain textures can require several weeks of additional labor. Some techniques promise faster production, whereas others ensure that the printing plate endures longer. New materials may become available through trade and innovations in chemistry. New surveying methods or instruments to measure elevation might affect modes of representation. New printing techniques may alter the linework, facilitate the reproduction of certain textures, or change overall costs. The construction of previously nonexistent infrastructures, like railroads, levees, or dams might drive the invention of new signs. Historical developments such as war, colonization, or urbanization may increase the demand for certain kinds of map (Hale, 2007; Lois, 2019; Picon, 2003; *The Mapmaker's Craft: A History of Cartography at CIA*). They might also sustain the occurrence of certain cartographic features, such terrain contours to support ballistics calculation, water depths to facilitate ship navigation, or grid plans to advertise expansion projects (Glover, 1996; Sutton, 2015; Zentai, 2018). More abstractly, the graphical form of maps or the elements they depict may be affected by ideas and ideals, like territorial states or scientific cartography. In turn, maps may contribute to the conceptualization of these ideas. Last, cartography and art continually influence each other. Aesthetic considerations are not limited to iconographic or pictorial maps; they pervade cartography in its entirety (Cartwright et al., 2009; Cosgrove, 2005; Rees, 1980; Woodward, 1987).

Thus, cartography is related to political, artistic, economic, epistemic, and scientific developments. Together, these dimensions contribute to what one might call *culture*. In this sense, culture is not restricted to knowledge of the humanities and fine arts. Rather, one could understand *culture* as the set of ideas, forms of knowledge, beliefs, conventions, and practices that are socially learned and transmitted. Maps are *cultural objects* insofar as they constitute a semantic–symbolic system through which culture is transmitted and constructed.

## *Problem statement*

On the one hand, the history of cartography analyzes maps with a wide perspective of the information they might hold, treating them as dense cultural objects; yet it engages with only a fraction of the overall historical production. On the other hand, heritage institutions house large collections of cartographic documents, and sustained efforts have already enabled the digitization of more than one million map records and images. Furthermore, map recognition techniques have been developed to extract geographic content from these digitized map images.

Nevertheless, digitized cartographic collections have not yet been integrated for large-scale analysis, and map recognition techniques have not been adapted to investigate maps as cultural objects. Consequently, the potential contribution of digitization and map recognition approaches to the cultural history of cartography remains unprobed. This leaves open a large research space for the exploration of these possibilities.

## I.2 Aim & objectives

In view of the above, the overarching aim of this dissertation is to:

*Create datasets and develop methodologies to investigate cartographic heritage on a large scale from a cultural perspective.*

This aim entails three distinct avenues for advancing research. First, the creation of datasets for the history of cartography. Second, the development of specific approaches for the analysis of cartographic heritage. Third, a proof-of-concept for the two preceding objectives—that is, the demonstration of their discovery potential. The latter aspect is perhaps the most ambiguous. Here, discovery potential is understood in a broad sense. It can assume several forms, including the creation of reference statistics on historical cartographic production, the formulation of hypotheses based on analyses, and the empirical validation of existing theoretical hypotheses. The notion of *validation* itself can likewise be ambiguous. In the present case, it involves (1) tracing links to the literature on the history of cartography, (2) implementing strategies for the statistical validation of quantitative results, and (3) qualitatively assessing the plausibility of the results and their explainability.

Please allow me to insist on the notion that, despite being based on a large data set, this work does not claim nor intend to rewrite or supersede the history of cartography. The history of cartography constitutes an entire field, built over decades through the direct examination of sources, discussion, and critical reflection. This work humbly aims to consolidate and enrich the field by developing tools that may complement, complete, confirm, or inform the way we study maps.

Within the three dimensions specified above, i.e., data, methods, and analysis, the overarching aim can be broken down into more manageable, clearly identified research objectives (ROs):

**RO1.** Collect accessible digital map records and digitized map images representative of multiple cartographic heritage collections across the world. Aggregate the collected entries into an operable database with informative metadata.

**RO2.** Create datasets and develop generic methods and models for (i) the recognition of map semantics and (ii) the detection of cartographic icons and symbols.

**RO3.** Develop approaches to extract and analyze elements of map form beyond icons and symbols, such as texture, linework, and color.

**RO4.** Chart and visualize how map publication is distributed spatially, and chronologically.

**RO5.** Chart and visualize the geographic foci of historical maps.

**RO6.** Investigate the quantitative relationship between map publication, geographic foci, and historical political developments, such as colonization, wars, and the construction of modern nation-states.

**RO7.** Visualize and investigate the social structure of map making.

**RO8.** Examine whether cartography may be categorized into distinct semantic modes.

**RO9.** Assess how the concepts of framing and image composition can be operationalized to study selective semantic emphasis and intentional design choices.

**RO10.** Visualize, describe, and examine how (i) icons and symbols and (ii) other elements of map form, such as texture, linework, and color, are distributed across geographic contexts and historical periods.

**RO11.** Investigate the historical manifestations of map scale and their relationship to cartographic form.

**RO12.** Investigate the relationship between elements of map form and geographic semantics.

**RO13.** Visualize and investigate the dynamics of transmission of the elements of map form.

These research objectives address the overarching aim of this dissertation by creating datasets (RO1–2) and by approaching cartographic heritage from three main perspectives: first, through a quantitative analysis of map publication volumes (RO4–7); second, through the study of image composition and the map frame (RO8–9); and third, through the analysis of map signs (RO2–3, RO10–13).

## I.3 Interdisciplinary positioning

### *Cultural analytics*

A first disciplinary—or rather interdisciplinary—perspective in which this dissertation is situated is that of cultural analytics. Cultural analytics is a field that relies on computational methods to study and explore cultural artifacts at scale, particularly through statistical modeling and data visualization approaches. Although its founder, Lev Manovich, conceived this field to approach and analyze the big data generated by contemporary digital culture (Manovich, 2020), the latter has expanded to include historical cultural big data as well, with applications for example in computational literature and musicology (e.g. Barkwell et al., 2018; Erlin, 2017). One of the foundations of cultural analytics is the database (Manovich, 1999). A typical first step, therefore, consists in applying data-extraction techniques such as named entity recognition, tokenization, or object detection in images. A central methodological concept here is *distant reading* (Moretti, 2013). Complementary to close reading, distant reading aims to uncover macroscopic patterns that are often difficult to detect or grasp through individual, qualitative interpretation of texts. In this prospect, the field draws on approaches like topic modeling or network analysis (e.g. Suárez et al., 2015). Arnold and Tilton (2019) later theorized the concept of *distant viewing*, which has since become popular within cultural analytics and more broadly within the digital humanities. This development took place in what has been called the *visual turn*, following which digital humanities research increasingly engaged with computer vision technologies and visual embedding methods (Wevers & Smits, 2020). The visual turn has also opened the way to the emergence of new disciplines, like digital art history (Drucker, 2013; Impett & Offert, 2022) and digital visual studies, enabling scholars to study large-scale contagion of artistic patterns (Joyeux-Prunel, 2019). Cultural analytics is closely connected to other fields, especially the computational humanities, from which it is only partly distinct, notably through its stronger emphasis on cultural theory, and data visualization.

### *Digital cartography*

Over the last few years, several projects in digital cartography have developed new ways of enhancing and exploring maps. For instance, OldMapsOnline is a platform that brings together more than half a million digitized maps from institutions such as the David Rumsey Map Collection, ETH Library, or Charles University in Prague (Klokan, n.d.). The platform relies on crowdsourcing to georeference historical cartographic collections. Other interfaces have also been designed to visualize maps *in situ*, such as MapScholar and Digital Mappa (Edelson & Ferster, 2013; Hodel et al., 2022). In addition, several institutions have undertaken significant campaigns to georeference their map series (Allord et al., 2014; Fleet, 2019; Fleet et al., 2012; Lallemant et al., 2017; Timár et al., 2010). A notable project is Recogito, an initiative of the Pelagios network to develop digital tools for annotating digitized maps and reference geographic places mentioned in archival sources (Simon et al., 2017). These new methods of annotating and enriching cartographic sources treat maps as linked data that are both spatially and historically situated (Görz et al., 2021; Liceras-Garrido et al., 2019). Visualizing and reprojecting historical spatial information, or historical maps, into modern geographic frameworks has nurtured a more critical sensitivity to ancient cosmographies, enriching the qualitative interpretation of these documents (Isaksen, 2011; Pereda et al., 2024; Reckziegel et al., 2021).

### *On the emergence of a computational history of cartography*

At present, and within this work, these technologies are progressively being integrated with computational analytical approaches, opening a way to the computational exploration of historical maps at scale, and from a cultural perspective. Pioneering issues includes the quantitative analysis of distortions, and computational approaches to cartographic stemmatology (Vaienti et al., 2025b), and the use of map recognition methods to monitor the diffusion of geographic knowledge and surveying techniques (Pala, 2023). The potential of computer vision for offering new ways of investigating digitized map images and the related printing and figurative processes (Hosseini et al., 2022; Petitpierre, 2020; Petitpierre, Uhl, et al., 2024). These works developed along with the present dissertation, tracing the contours of a new digital scholarship in the history of cartography. On the one hand, it draws on computational methods like computer vision, visual embedding, and statistical modelling to model cultural transmission and the evolution of figuration at scale. On the other hand, it develops visualization strategies to provide interpretable results and adopts a critical perspective. These interactions are summarized graphically in Figure 1.

In many respects, this research is admittedly exploratory. To meet its aim, it intentionally addresses a broad set of interrogations, using diverse methodologies, which I hope will not unsettle informed readers, regardless of their background. It also had to leave aside several perspectives that pervade the history of cartography, like the processual perspective, the circulation of geographic knowledge and cosmological models, and the interpretation of paracartographic textual sources. Nevertheless, I hope it fulfills its purpose by contributing to build a scholarly curiosity for digital and visual approaches to cartography and the digital investigation of maps as cultural objects.

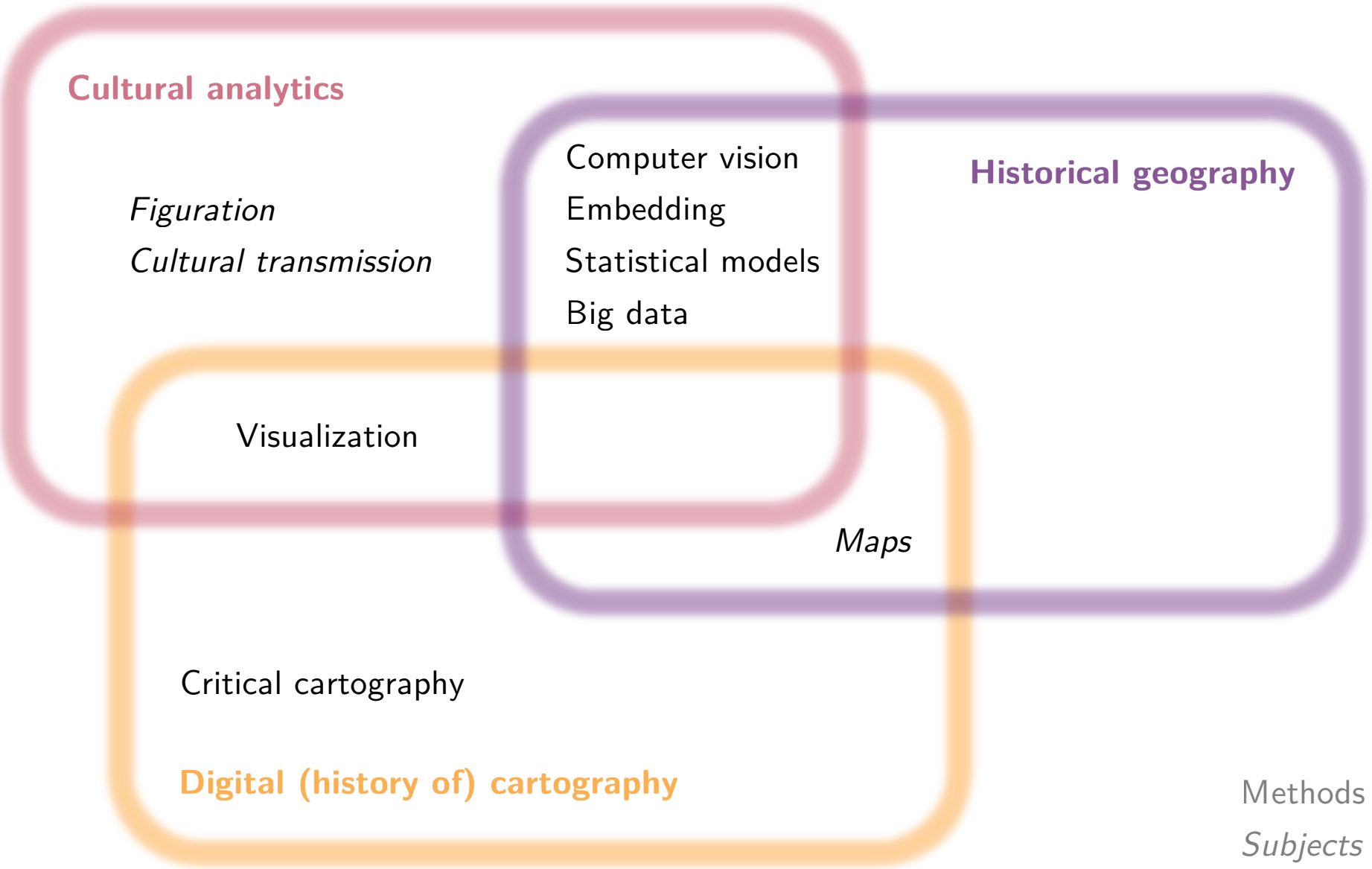

**Figure 1 | Venn diagram of the related research landscape.** The diagram visualizes the nine main families of Methods and *Subjects* found in the thesis and their inclusion within the research landscape, and three related fields.

## I.4 Structure of the Dissertation

The remainder of the dissertation is divided into two parts comprising seven chapters, followed by a general conclusion. Figure 2 summarizes the organization of chapters in relation to one another. The first part of the monograph, which corresponds to the first three chapters, addresses data collection, aggregation, and the investigation of map records' metadata. The second part of the dissertation pertains to the analysis of map images. It employs map recognition techniques, visualization, and statistical approaches to study cartographic figuration.

In the first chapter, *Harvesting the Big Data of Cartography*, I compile the Aggregated Database for the History of Cartography (ADHOC), a large corpus derived from 38 collection catalogs of major cartographic collections distributed across 11 countries and representing 771,561 map records. I also assemble a representative sample of 99,715 digitized map images published in six countries between 1492 and 1947. To my knowledge, this is the most extensive database yet assembled for studying the history of cartography. The primary challenge in this procedure lies in the reconciliation of inconsistent metadata through natural language processing, textual heuristics, and the alignment to reference geographic databases.

In the second chapter *The Making of Cartography*, I present statistics on the spatial dynamics of map publication, by charting historical centers of map making, along with their publication volumes and periods of activity. I visualize the chronology of map publication across different countries and discuss its relationship to the existing historiography. In the final section, I examine the social structure of map making by inferring a social graph of work collaborations, based on metadata extracted from map records. I assess the way in which the graph structure reflects publication periods, places, and maker typologies.

The third chapter, *Map and its Power(s)*, explores the geographic foci of historical cartography, treating them as manifestations of selective, resource-limited *spatial attention*. It also investigates the quantitative relationship between map publication and historical political developments. For instance, I discuss the progressive rise of national self-attention that accompanied the political construction of nation-states and the attendant drive to enforce control over national territories. I also highlight the marked surge in map publication during the two World Wars and consider its implications for cartography as a strategic military technology. Finally, I investigate the interaction between Atlantic charting and the triangular trade, showing how maps may have facilitated maritime navigation and reinforced the ideal of the "New World" and colonization, ultimately boosting the demographic and economic growth of the colonies.

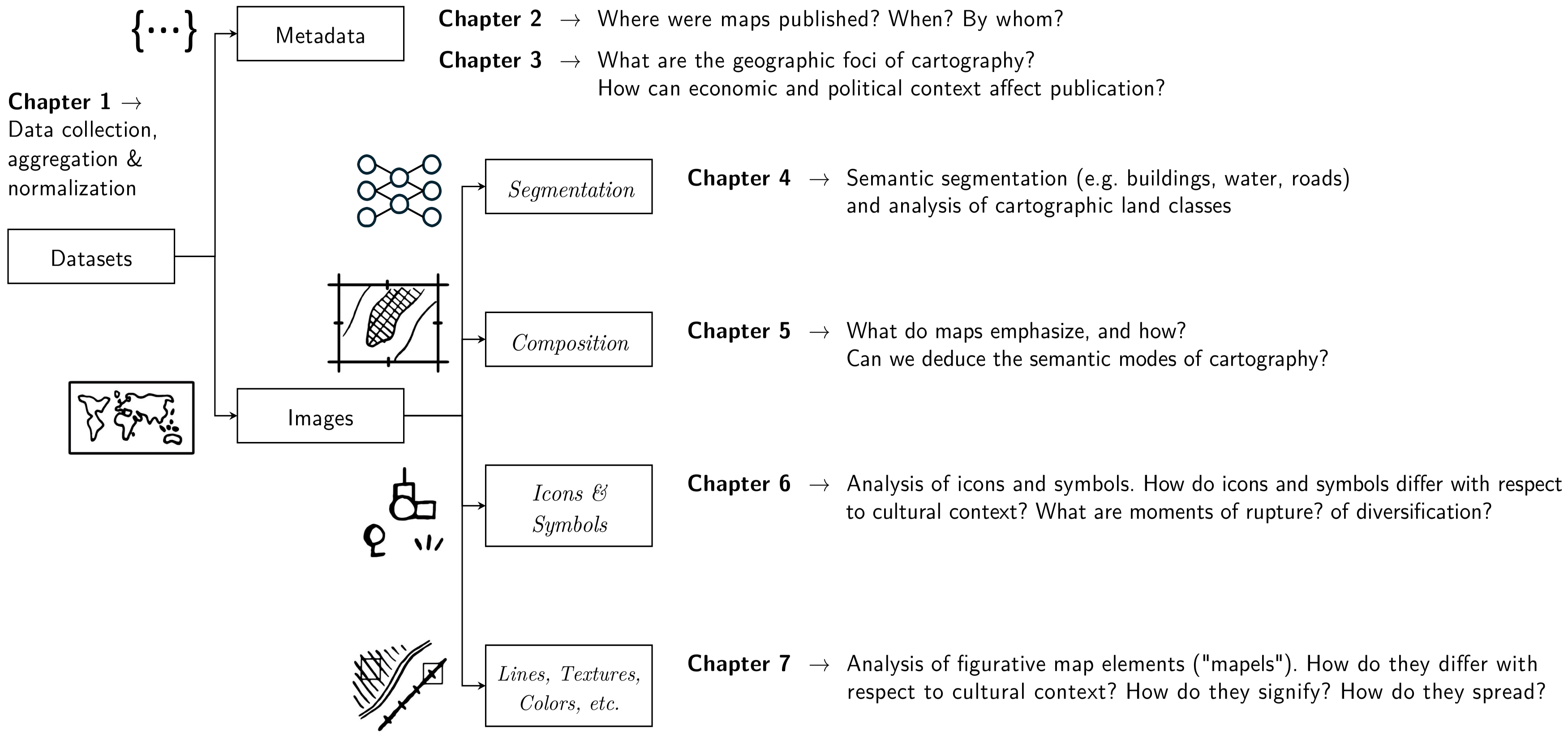

**Figure 2 | Graphical summary of the structure of the dissertation.** The dissertation is divided into 2 parts and 7 chapters, followed by a conclusion. The first part includes the chapters 1–3, while the second part corresponds to chapters 4–7.

The fourth chapter, *On Semantic Segmentation, or (Un)drawing Geography*, marks the beginning of the second part of this dissertation, which focuses primarily on images. Specifically, this chapter addresses the recognition, segmentation, and quantitative analysis of geographic semantic classes such as buildings, water, or roads. Semantic segmentation is a pixel-level classification task in which each pixel in an image is assigned a semantic class. In this case, the neural model was trained with a combination of manually annotated images and synthetic data. To achieve genericity, a new manually dataset, Semap, was manually annotated, comprising 1,439 map samples addressing a wide variety of cartographic cases. The resulting model, based on Mask2Former architecture, significantly improves segmentation performance relative to the former state of the art. It served to segment the 99,715 maps contained in the ADHOC Images database. The extracted data makes it possible to analyze the relative share of geographic classes as a function of publication period, map scale, and place of production.

The fifth chapter, *Maps as Pictures: Framing & Composition*, considers maps as designed images marked by intentionality. Specifically, it investigates the idea that mapping processes, like the choice of the map frame, or the enforcement of spatial relationships like centering, indicate a hierarchization of map content, whereby certain geographic features are emphasized at the expense of others. In addition, the occurrence of semantic symmetries suggests that mapping is also an artistic composition process in which the preservation of visual balance matters. In the last section of this chapter, I study the way cartography may be divided into distinct semantic modes that correspond to particular map types, such as maritime charts, chorographic maps, or city plans. I then visualize how the relative frequencies of these cartographic modes vary as a function of publication period, map scale, and geographic context.

The sixth chapter, *The Map Code: Investigating the Evolution of Cartographic Signs*, focuses on cartographic signs such as icons and symbols. The beginning of the chapter describes the method used to extract 63.2 million signs from the ADHOC Images dataset. To enable a quantitative and comparative analysis, signs were first encoded into a latent space of representation, where they were then grouped into small clusters, each corresponding to a fraction of the entire space of figuration. This space is structured hierarchically, reflecting the phylogeny of cartographic signs. The distribution of map signs varies as a function of historical period and cultural context. Map signs may also vary according to one another, forming sign complexes whose components tend to appear jointly. Comparative analysis of these distributions makes it possible to identify moments of rapid change, or rupture, and to measure variations in diversity over time. The last part of the chapter approaches regional cultural distinctions, which are also qualitatively discussed based on the visualization of characteristic exemplars. Finally, I attempt to model the way semiotic differences across publication centers are related to geographic distance, period of activity, and city size.

The seventh chapter, *The Semiotic Microscope*, approaches map figuration at an even more granular scale by focusing on small image fragments, or map elements ("*mapels*"), that represent a variety of colors, textures, and lines. The analysis of mapels is coupled with the results of semantic segmentation to investigate how visual forms may convey distinct semantics according to their publication period or map scale. The chapter also discusses quantitative changes in the relative frequencies of map forms and graphical variables. I discuss the way mapels seem to form locally consistent semiotic systems, within which each sign tends to take a univocal meaning. The last part of the chapter extends and complements the model introduced in Chapter 6 by including the analysis of semiotic diffusion across the social graph of mapmakers generated as part of Chapter 2. Finally, it visualizes diachronic transmission across publication centers, highlighting how certain cities develop as cultural *hubs* while others fail to sustain autonomous and distinctive semiotic systems.

The structure of the dissertation follows a gradual decrease in the level of observation. The second and third chapters analyze cartography primarily through the prism of publication volumes, studying how the general distribution of maps informs the spatial structure and chronology of map production. They also discuss the relationships between geopolitical historical developments and publication volumes. In Chapters 4 and 5, the unit of observation is a single map: one document. There, the analysis focuses on semantic content, frame choice, centering, and spatial composition within the map image, which reflect intentional emphasis. Finally, the last two chapters adopt an increasingly granular level of observation, focusing on signs, icons, or small visual fragments. Fragmentation enables the computational analysis of cartographic figuration by allowing the comparison and assimilation of simple figurative elements. At this level of observation, it is possible to discuss details of the image as well as cultural transmissions on a macroscopic scale. Thus, while the dissertation approaches maps at progressively finer levels of observation, it continually oscillates between a macroscopic level of analysis and the discussion of specific examples.

At last, the argument progressively introduces a multifaceted interpretive framework. It begins by presenting the political perspective that maps are instruments of power and territorial control. Later, it considers the idea that mapmaking is also shaped by broad and partly self–organizing economic and artistic dynamics. Finally, it approaches cartography as a continuously evolving semantic–symbolic system, marked by processes of cultural transmission and replication. In the conclusion, I discuss the way these distinct perspectives complement each other to form a balanced and descriptive analytical framework for the computational analysis of maps and the digital history of cartography.

# Part I

# Records.

# Chapter 1

## *Harvesting the Big Data of Cartography*

In this first chapter, I present the methodology used to construct and structure the Aggregated Database on the History of Cartography (ADHOC Records). This composite corpus, comprising 771,561 map records compiled from the digital catalogs of 38 digital libraries in 11 countries, constitutes the dataset on which chapters 2 and 3 of this thesis will rely. It is complemented by a sample of 99,715 digitized map images (ADHOC Images), the analysis of which is the focus of chapters 4 through 8.

### 1.1 Temporal scope

The temporal boundaries of the corpus, which extend from the end of the $15^{th}$ century to the middle of the $20^{th}$ century, are primarily defined by the density of documents located in digital map collections. As such, and despite regular progress in digitization, the availability of sources is still a limiting factor.

The determination of the corpus initial year also relates to the history of map printing and the first recognized printed map, in 1472. The interval between 1472 and 1500 is often considered a transitional phase for mapmaking (T. Campbell, 1987a). Printing technologies facilitated the production of maps in greater quantities and, consequently, the gradual development of a market hitherto constrained by artisanal reproduction (Robinson, 1975). The *Age of Reconnaissance* (Parry, 1981) also contributed to fueling the desire for maps in Europe (see Chap. 3) and the progressive, yet steady, expansion of mapmaking.

Christopher Columbus's (1451–1506) first voyage across the Atlantic, in 1492, marked a significant shift in cosmographic and geographical paradigms at the close of the 15th century[1]. The narrative that ensued from the "Discovery of the New World" also had a profound impact on cartography over the subsequent centuries. Together with the rounding of the Cape of Good Hope—an unforeseen southern passage between the Atlantic and Indian Oceans—Columbus's discoveries have called into question Claudius Ptolemy's *Geography* (Isaksen, 2011), profoundly impacting the world model. Coupled with the development of cartographic printing which happened progressively during the same period, the year 1492 thus constitutes a turning point and offers a suitable starting date for the corpus.

Counterintuitively, perhaps, the corpus is constrained not only by the scarcity of Renaissance maps but also by the limited representation of more recent documents in digital libraries. This situation largely stems from copyright policies: most maps produced in the second half of the 20th century have not yet entered the public domain and therefore cannot be disseminated online by heritage institutions, which generally do not hold the rights to distribute them. In most Western countries, copyright expires 70 years after the author's death. Consequently, digital collections theoretically comprise only maps created by authors who died before 1955, or whose creators, e.g. public agencies, have voluntarily renounced copyright. Additionally, determining the precise year of copyright expiry is not always straightforward, as maps can be edited by multiple individuals and entities. Therefore, as a precaution, libraries may wait 100 or 120 years after publication before considering a document part of the public domain. Consequently, online cartographic documents represent only a fraction of the maps issued after 1905, with a bias toward the overrepresentation of materials produced by cartographic agencies or other public institutions; this bias is expected to intensify as the publication year approaches 1955.

[1] Indeed cosmographic theories appear to have played a significant part in the decision to finance Columbus's first voyage (Haase & Meyer, 1994). Columbus's proposal was twice rejected before it ultimately received approval. In Salamanca in 1487, the committee of scholars charged with evaluating the project dismissed it because of a cosmographic and cartographic argument advanced by the theologian Paul of Burgos (c. 1351–1435). According to Burgos, the Earth consisted of not one but two spherical bodies, one composed entirely of water and the other of earth, which God separated on the Third Day so that their centers moved apart and the terrestrial sphere emerged upon the surface of the larger aqueous sphere (Fig. A1, in Appendix). Under this model, navigation was feasible only along continental margins—as it had been hitherto, except on inland seas—and the global ocean was vast, empty, and arguably impassable. The Portuguese discovery of the Cape of Good Hope in 1488, together with knowledge of the Azores, apparently persuaded the Spaniards assembled again in Santa Fe in 1491 of the inadequacy of Burgos's interpretation, opening an avenue for alternative cosmographical models.

Consequently, 1948 was selected as the final year included in the corpus. The later end of the study period thereby encompasses both World Wars, which were periods of sustained cartographic production, particularly by national and military agencies, as will be highlighted in Chapter 3. 1948 witnessed the enactment of the Marshall Plan, marking the beginning of Europe's reconstruction and the onset of the Cold War.

## 1.2 Spatial scope

One of the principal risks inherent to any quantitative approaches is sampling bias. More extensive and inclusive corpora may mitigate the impact of minor biases, whereas more substantial distortions typically originate in the collection methodology and may be detected through data analysis[2]. The present collection methodology concentrates on current national boundaries in order to model the cartographic heritage produced and amassed within each country under examination. The relevance of employing national political boundaries as historically meaningful delimiters is assessed in Chapter 2. The primary reason for this choice is that, in many cases, cartographic heritage is collected at the national level. The main institutional actors are national libraries, national cartographic institutes, and universities. They are complemented by private collections, regional archives, and—in major cities—municipal libraries. The national scale, therefore, constitutes a relevant scale to gauge archiving, collection, and cataloging practices. In ideal cases, national systems implement homogeneous collection policies. However, technical discrepancies and imbalances due historical collection flaws may arise. In the present approaches, whenever these biases are judged significant, the entire country is excluded from the corpus. This decision is justified by the perspective that, from a research point of view, an admittedly partial corpus is more desirable than an imbalanced one.

Decisions of inclusion, or exclusion, are based on the following criteria:

i. The country should maintain a comprehensive policy for the collection, preservation, and digitization of its cartographic heritage. At least one institution should be identifiable in which more than 1,000 map records have been digitized. Additionally, at least 4,000 digital records should be available at the national level, which should correspond to the work of multiple creators and publishers, covering an extensive range of the study period.

ii. Records include informative metadata, particularly the year and place of publication of each document.

[2] An emblematic example of large collection bias is that of Google Books, where scientific literature was found to be overrepresented compared to average text productions (Pechenick et al., 2015).

iii. Digital platforms provide search systems capable of identifying documents relevant to this study. An explicit, publicly accessible digital path to metadata records should also be available, for example a unique identifier or url.
iv. Automated collection of metadata for each individual cartographic record is enabled, for instance, by an API, a transparent web implementation, or a bulk-download functionality.
v. Digital platforms did not implement any mechanisms intended to prevent automated data collection.
vi. No major map collection should be amiss for a given country.

After evaluation, 11 countries that satisfied these criteria were identified, corresponding to 38 map collections, heritage institutions, and digital catalogs. Listed in descending order of catalog count, these countries are the United States (11 catalogs), Germany (8), Spain (4), Switzerland (3), France (3), the Netherlands (3), Italy (2), Denmark (1), Australia (1), Portugal (1), and Japan (1).

One of the countries most represented in the historiography yet absent from this dataset is the United Kingdom (cf. Chapter 2, Fig. 1). In this specific case, the decisive criterion was the inability to collect data from the British Library (point vi) owing to the prolonged and significant disruption of services following a cyberattack in 2023 (*Cyber Incident Update*, 2025).

## 1.3 Fields reconciliation

The first step in building a reconciled corpus from disparate data sources is to regroup the metadata fields into overarching categories. Depending on the catalog, information concerning the *location depicted on a map* appear under different field names:

ambito_geografico, area_geografica, centre_del_full, city, classement_geographique, conca_hidrografica, contents, coordenades_de_la_imatge, coordinate_geografiche, coordinates, coordonnees_geographiques, country, county, coverage, geographic_coverage, koordinaten, location, luogo_normalizzato, ortsbezug, place, place_name, place_names, places, region, secondary_location, sessional_paper_maps___geography_covered, spatial coverage, state province, subject_geographic, sujets_geographiques, swisstopo_coordinates, world_area.

Because these fields refer to similar information, related to geographic coverage, they can be aggregated into a single meta-column while documenting the original field names. This procedure helps reduce the initial 493 catalog fields to 19 meta-columns:

1) record identifier; 2) library specifics (e.g., storage location, documental series); 3) item URL; 4) digitization paradata (e.g., resolution); 5) document title; 6) complement to the title; 7) general publication information; 8) publication year; 9) publication place; 10) publisher; 11) author/contributor; 12) geographic coverage; 13) cartographic data (e.g., scale); 14) description of the physical object; 15) type of document; 16) subject/topic; 17) language of the document; 18) additional notes and comments; 19) note on usage and copyright.

## 1.4 Named Entity Recognition (NER)

The second step toward creating a reconciled dataset involves normalizing the metadata themselves. For instance, the following is an example entry from the general publication-information meta-column:

'pub_title': 'Historia mundi: or Mercator's atlas : Containing his cosmographicall description of the fabricke and figure of the world. Lately rectified in divers places, as also beautified and enlarged with new mappes and tables; by the studious industry of Iudocus Hondy. Englished by W.S. generosus, & Coll. Regin. Oxoniae. London Printed for Michaell Sparke, and are to be sowld in Greene Arboiure, 1637. Second edytion.'

As in the example, a significant part of the information stored in digital catalog metadata remains unstructured. Named Entity Recognition (NER) makes it possible to extract relevant phrases from such texts and classify them to produce structured representation of the information.

AUTHOR/CONTRIBUTOR: ['Mercator'; 'Iudocus Hondy'; 'W.S. generosus, & Coll. Regin. Oxoniae']
PUBLICATION_PLACE: ['London']
DATE: ['1637']

Beyond the presence of obsolete language, the main challenge in this example is to discriminate target entities—e.g., the authors and contributors—from other individuals, such as the sponsor, Mr. Michaell Sparke. Similarly, the place of publication (London) should be distinguished from the location where the document was intended to be sold (Greene Arboiure). Language models can learn to make such distinctions based on context and supervised training. A further challenge is handling multilingual texts. The metadata are provided in more than a dozen languages, including some for which NER models are comparatively well trained, such as English and French, but also less common ones, like Latin and Japanese.

In order to train NER models to extract structured entity information from the metadata, a random sample containing 2,581 field entries from columns 5–13 was annotated manually. The sample comprised 5,261 named entities in total. The meta-columns were further regrouped into four informational classes: publication (columns 7–10, sample size n = 1,701), authorship (column 11, n = 855), geographic coverage or cartographic scale (columns 12–13, n = 849), and title (columns 5–6, n = 1,856). Each subset was divided into a training set and a validation set (80%/20%). Four specialist NER models were then trained, one for each informational class. Each model was initialized with the pretrained xx_ent_wiki_sm model, a generic multilingual model trained on the WikiNER corpus (Nothman et al., 2013). xx_ent_wiki_sm is part of the spaCy library and is based on Tok2Vec and a convolutional architecture (*spaCy NLP*, 2022). The models were trained in two stages: first jointly on the four tasks, then fine-tuned separately on each informational class.

Table 1 reports the performance of each model, calculated on the 1,049 entries from the validation sets. The fine-tuned models outperform the baseline xx_ent_wiki_sm model, corroborating the benefit of training task-specific models.

**Table 1 | Recall (R) and precision (P) of the named entity recognition models**, reported for each informational class and named-entity specialist model with a tolerance of ±1 character. Values in brackets represent the performance of the spaCy *xx_ent_wiki_sm* multilingual NER model, provided for baseline comparison. In the subset column, N denotes the total number of entities annotated in the corresponding named-entity class. All metrics were computed on the validation set. Mean estimates were calculated as the average recall and precision, weighted by the total number of entities retrieved at inference time and adjusted for recall and precision. A cross (×) indicates that the metrics are based on only n = 4 samples.

| Subset | Location | Author/Publisher | Year | Scale |
|---|---|---|---|---|
| **publication** | R 0.93 (0.69) | R 0.95 (0.53) | R 0.90 | R 1.00× |
| N = 356 | P 0.97 (0.53) | P 0.94 (0.54) | P 0.98 | P 1.00× |
| **authorship** | R 0.90 (0.90) | R 0.93 (0.26) | R 0.53 | – |
| N = 165 | P 0.90 (0.24) | P 0.88 (0.19) | P 1.00 | |
| **coverage, scale** | R 0.87 (0.48) | – | – | R 0.98 |
| N = 122 | P 0.84 (0.37) | | | P 0.96 |
| **title** | R 0.96 (0.44) | R 0.74 (0.51) | R 0.96 | R 0.80 |
| N = 406 | P 0.98 (0.51) | P 0.63 (0.26) | P 0.98 | P 0.92 |
| *weighted mean* | *R 0.93* | *R 0.90* | *R 0.81* | *R 0.80* |
| *estimate* | *P 0.94* | *P 0.85* | *P 0.99* | *P 0.92* |

On average, 80% to 93% of the named entities were identified, depending on entity type. Recall was higher for location and for authorship, perhaps because these entities fall within the generic types already recognized by the pretrained xx_ent_wiki_sm model. Mean recall was slightly lower for year and cartographic scale. For scale, the decrease may also reflect the comparatively lower sample count in the training set. Precision ranged from 85% to 99%. The precision with which authors and publishers were recognized was marginally lower than that observed for other entity types, possibly because the model must discriminate mapmakers, publishers, and creators from other individuals or organizations mentioned in the texts. For reference, the original xx_ent_wiki_sm exhibits a recall of 83% and a precision of 84% on its own validation dataset. Thus, the overall performance of the present models is considered satisfactory.

At inference time, 8.1 million named entities were extracted from the metadata, corresponding to 3.2 million *locations*, 2.5 million *authors and publishers*, 2.1 million *year* entities, and 220,000 *scale* mentions. The most informative columns are those related to publication (38% of all entities extracted; aggregated columns 7–10), the title (30%; columns 5–6), authorship (20%; column 11), and, finally, location or scale information (12%; columns 12–13). It is worth noting that a substantial portion of the information is contained in the title, an initially unstructured field. This result indicates the potential for this approach to enrich collection metadata.

## 1.5 Normalizing named entities and geocoding place names

### *Years*

The final step toward a reconciled dataset consists of normalizing the retrieved named entities and realigning them to a common, canonical form. This operation appears straightforward for some informational classes. Normalizing years, for example, simply requires converting uncertainty symbols into a time range, based on string heuristics. The field "185?" should thus be interpreted as the range 1850–1859. The dataset also contains century indications, usually recorded in Arabic numerals (e.g., "18. Jahrhundert" → "18" → 1701–1800). Although the corpus comprises Japanese maps, the original year and era indications are also Westernized in the metadata (e.g., "安政2 [1855]," transcribed "Ansei 2 [1885]").

### *Names*

Disambiguating less-structured data—such as author names and publishers—can prove complex. First, abbreviations can generate uncertainty. For instance, "J. H. Colton" is likely the same individual as "Joseph Hutchins Colton", but what about "Stone, W. J."? Is he the same person as "William Stone"? Second, names are frequently acculturated. For example, Joan Blaeu signed his French maps "Jean Blaeu", and Andrea Pograbski is sometimes cited under the Latinized form "Andreae Pograbius". Finally, an ambiguity often exists between individual mapmakers and their workshops. Thus, “A. M. Dulhunty” is arguably equivalent to “A. M. Dulhunty & Co”, whereas “Johann Homann” should be distinguished from “Studio Homannianorum Heredum” (workshop of Homann's heirs), even though the former is syntactically included in the latter.

Given the difficulty of the task, optimal results may only be achieved through manual interpretation. In the present case, however, the raw dataset comprises 352,720 unique name variants—that is, approximately 124 billion potential word pairs—which is well beyond the potential reach of manual review. Moreover, string-distance heuristics—based on Levenshtein distance or longest common substring—would perform poorly when disambiguating even simple variants such as “Cassini, Giovanni Domenico” and “Jean Dominique de Cassini.” The approach that was ultimately implemented is thus grounded in foundational language models and text-embedding vectors.

More precisely, each unique variant name was embedded in a 1,536-dimensional space using the GPT text-embedding-3-small model. For every word vector $\vec{v_i}$ its nearest neighbor $\vec{v_j}$ was retrieved, as measured by cosine distance. Five hundred word pairs were then sampled and annotated manually. The annotation was binary, comprising two classes: $\vec{v_i} \approx \vec{v_j}$ signifying that $\vec{v_i}$ and $\vec{v_j}$ are variants of the same normal entity, and $\vec{v_i} \not\approx \vec{v_j}$, indicating that they correspond to distinct entities. 304 neighboring word pairs were annotated as cognate and 142 as distinct; 54 samples were excluded as they contained NER errors (see Tab. 1).

Based on the annotated data, a threshold $t = 0.17$ was imposed on the distance $d$. Cognate word pairs were matched based on this threshold, which yields a match recall of 57 % and a precision of 94 % (see Fig. A2). The one-to-one matching policy was therefore relatively conservative. A graph $G$ was then constructed, that contained 352,720 nodes—one per word variant—and 171,479 edges, each representing a cognate relationship, i.e. a valid word pair. Here, only the three nearest neighbors to each word were considered potential candidates[3]. In the end, 194,298 nodes formed no edges, whereas 171,479 were linked to at least one other node, yielding 42,627 connected components. Connected components were grouped together; the node with the most mentions in each component was designated as the canonical form. Consequently, the number of unique contributors was reduced from 352,720 initially to 236,925.

[3] More distant relationships are modeled by common connections.

### *Places*

Place name geocoding is an operation that can facilitate normalization. Geocoding involves matching each place name with the geographic location it designates, corresponding to a point, a line, or polygon geometry represented in a geographic coordinate system. Because geocoding encounters issues similar to those involved by name normalization, including multilingual context and heterogeneous structures, specialized reference geodatabases and algorithms have been developed for this task. Here, two distinct application programming interfaces (APIs) were considered. First, the Google V3 Geocoding API (*Geocoding API*, n.d.), which was found to be more precise (see Tab. 2), was used to geocode the 40,000[4] most frequent place names, accounting for 79 % of the dataset. The remaining 273,969 entities were geocoded with the open-source Nominatim[5] Geocoding API (*Nominatim*, n.d.). Georeferencing accuracy was evaluated manually on a common, random sample of 100 entries. The results are reported in Table 2. Although Nominatim appeared slightly more sensitive, Google V3 seemed substantially more precise and generated fewer false positives. Based on those figures, one can estimate that 75.9 % of the unique place names were geocoded, with a precision of nearly 95.7 %.

**Table 2 | Evaluation of the two geocoding APIs** based on a random sample of 100 entries. Positive precision is defined as the number of correctly geocoded locations divided by the API's positive response count. Overall recall equals the number of correctly geocoded locations divided by the total number of true locations in the sample (n = 92).

| Geocoder | True positive (correct) | True negative (not a place) | False negative (missing) | False positive (mistake) | Positive Precision | Recall |
|---|---|---|---|---|---|---|
| GoogleV3 | 69 | 8 | 22 | 1 | 98.6% | 75.0% |
| Nominatim | 73 | 4 | 10 | 13 | 84.9% | 79.3% |

[4] There is a monthly free tier of 40,000 requests (USD 200) for the GoogleV3 geocoding API.
[5] Nominatim is the geocoding solution developed by the OpenStreetMap community.

## 1.6 Overview of the aggregated dataset

Chapters 2 and 3 will present detailed and quantitative historical analyses of the ADHOC Records database. In this respect, the publication period constitutes an indispensable piece of information. The dataset was thus restricted to records for which the publication year, or at least an approximate period, such as a century, was available. This restriction reduced the size of the database by 16%, from 999,750 to 832,880 records.

Then, because the database was aggregated from distinct digital libraries, it contained duplicate records, which were removed. Moreover, several map series—such as national topographic maps, fire insurance plans, and cadastres—were overrepresented, sometimes comprising tens of thousands of nearly identical entries; a subsampling procedure was thus applied. After addressing these redundancies, the final ADHOC Records database contains 771,561 document records. Each document record corresponds to one or several maps—e.g. in the case of atlases or cartographic series. As such, the database effectively describes more than 1.4 million individual maps (see Section 1.7).

Figure 1 presents the completeness of the dataset, for each field.

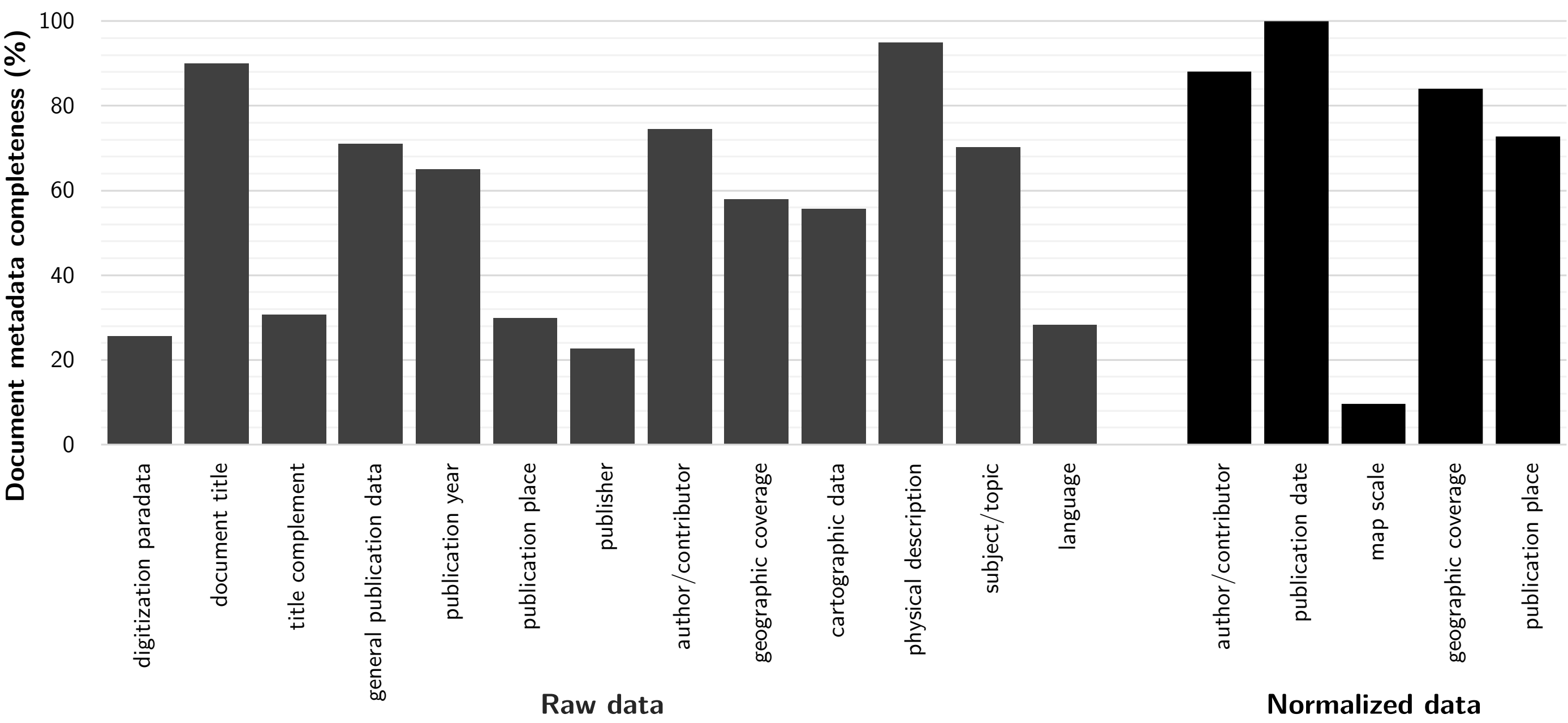

**Figure 1 | Completeness of the fields in the aggregated dataset.** The left entries correspond to the raw data, i.e., the data as they were initially harvested, after field reconciliation. The right entries correspond to the normalized data, obtained after named-entity recognition, normalization, and georeferencing.

Notably, the completion rate of normalized fields can exceed that of raw fields, in spite of the additional NER and georeferencing processing steps. This is because normalized fields aggregate information extracted from raw unstructured fields. For example, geographic coverage information may appear not only in the reconciled field `geographic coverage` but also in the map title. For authorship, place of publication, and geographic coverage, metadata completeness ranges from 73 % to 88 %. Information on map scale, however, is retrieved in only 10 % of cases. Two reasons may explain this low representation. First, the modern conceptualization of scale is a relatively recent historical development (Edney, 2019). Second, in most documents, scale is indicated graphically—e.g., as a graduated bar at the bottom of the map—accompanied by historical distance units, for example "two leagues of Picardy." Converting these units into modern scalar ratios (e.g., 1:37,500) is not always straightforward for catalogers. Consequently, the field is often omitted.

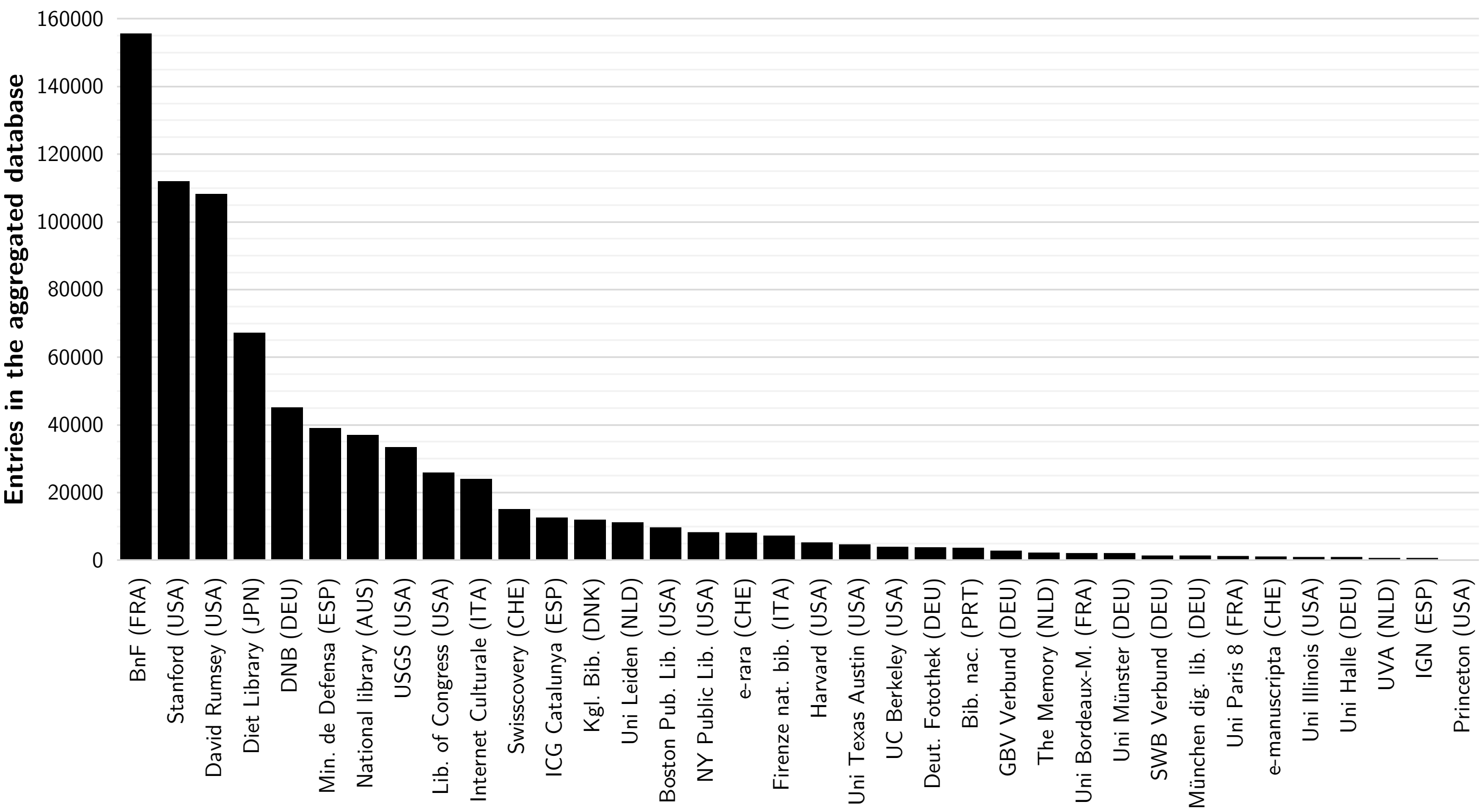

**Figure 2 | Composition of the aggregated database by digital map catalog.** The country acronyms on the horizontal-axis labels correspond to the ISO-3 country codes. Exact map counts and additional information are provided in Table A1 in the Appendix.

Figures 2 and 3 illustrate the composition of the dataset by collection or catalog. Generally, the digital catalog corresponds to both the heritage institution and the collection. Catalogs may be linked to national libraries (e.g., the Bibliothèque nationale de France, BnF), city libraries (e.g., the New York Public Library), or university libraries (e.g., Harvard Library). They may also be associated with a government library (e.g., the library of the Spanish Ministry of Defense), a parliamentary library (e.g., the Library of Congress or the library of the Japanese Diet), or a mapping agency (e.g., the United States Geological Survey, USGS). There are also private collections, such as the David Rumsey Map Collection. Finally, common portals have sometimes been established by library federations (e.g., SWB, Südwestdeutscher Bibliotheksverbund, or Swisscovery), thus the ambiguity between collections, institutions, and catalogs. Extensive information on provenance and catalogs is provided in Table A1, in the Appendix.

Figure 2 ranks all catalogs from the most represented—BnF, Stanford Library, and David Rumsey, each comprising more than 100,000 entries—to the least represented, namely the Princeton Library. Some catalogs, such as Princeton's, ultimately contain fewer than 1,000 records even though selection criterion (i) in Section 1.2 required at least 1,000. Indeed, the steps of normalization, duplicate deletion, and map-series subsampling successively reduced the number of retained records. The map in Figure 3 shows the spatial distribution of all catalogs and associated institutions. The institutional framework for map collection and archiving varies by country. While the conservation structure is centralized in France, Japan, Australia, and Portugal, it is polycentric in Switzerland and Germany. There, federations tend to aggregate documents held by separate local institutions. In Spain, state agencies such as the Biblioteca Virtual de Defensa, linked to the Ministry of Defense, the National Geographic Institute (IGN), and the Institut Cartogràfic i Geològic de Catalunya (ICGC) predominate.

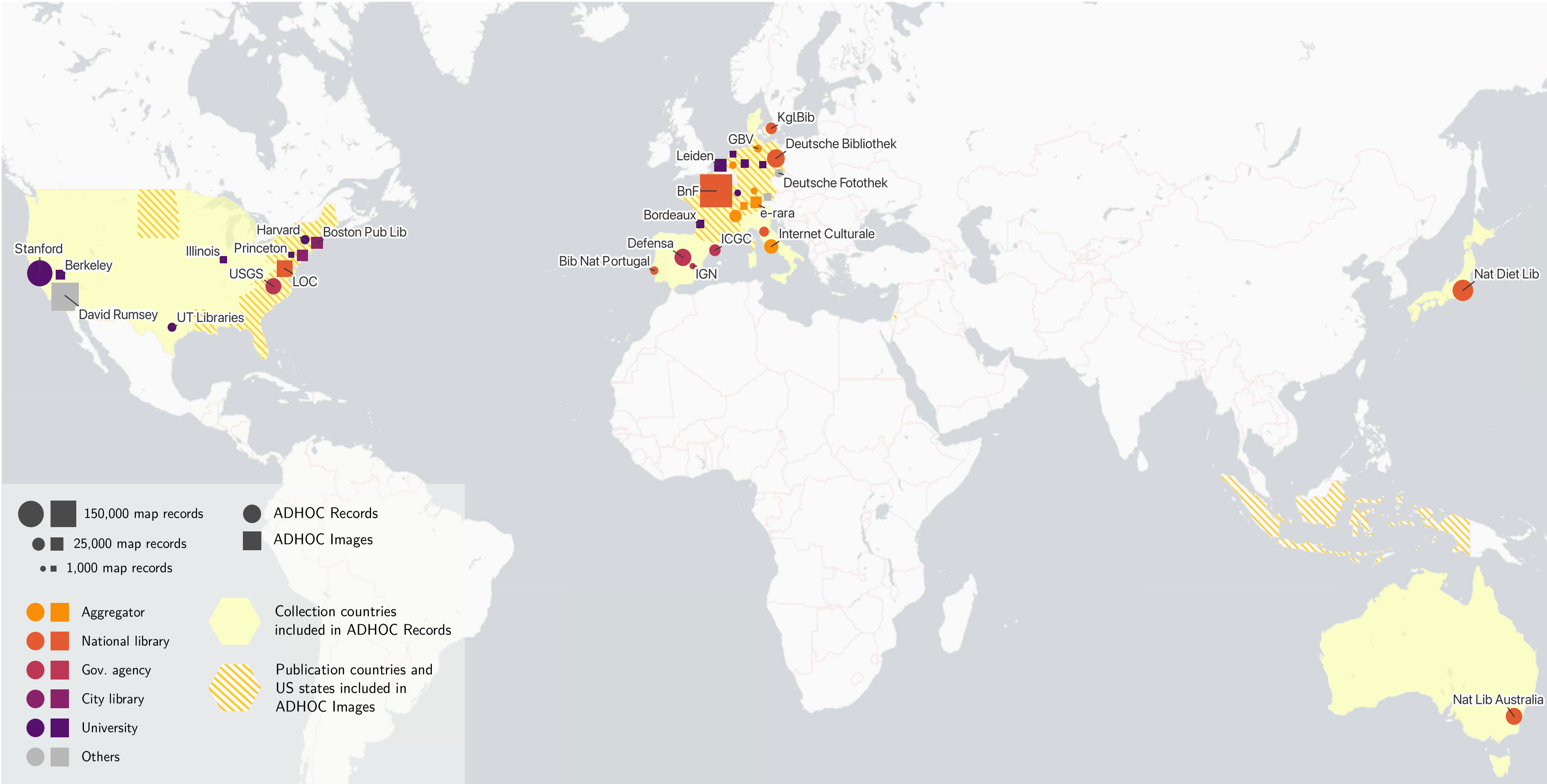

**Figure 3 | Map of the catalogs included in ADHOC Records and ADHOC Images datasets.** Circle markers indicate that the catalog is included only in ADHOC Records, while square markers that the catalog is included in both ADHOC and ADHOC Images. The size of the markers depends on the number of entries in ADHOC Records. The marker color indicates the category to which the catalog belongs (see also Tab. A1). The countries included in ADHOC Records are highlighted in bright yellow. The publication countries and U.S. states included in ADHOC Images are hatched.

## 1.7 ADHOC Images

*Assembling the image database*

In addition to collecting and analyzing metadata, one of the objectives of this work is the digital investigation of map images. The image corpus, ADHOC Images, was derived using a methodology similar to the one described in previous pages. The criteria applied for the selection of image catalogs correspond to those described in Section 1.2, with the addition of a seventh criterion:

vii. Images are digitized in a high resolution, ideally 300 ppi or more, but at least 180 ppi.

Images must also have a publicly accessible digital path (iii), and automated downloading must be enabled (iv, v), for example via the International Image Interoperability Framework (IIIF). Overall, 16 digital image platforms were retained[6].

The image corpus had to be further subsampled due to technical constraints. Indeed, the institutions considered in five collecting countries—United States, France, Germany, the Netherlands, and Switzerland—already comprised approximately 537,730 records, corresponding to 1,201,848 digitized images, or nearly 16.5 TB of data. Processing such a corpus at a rate of one minute per image would require roughly 835 days of computation, a duration that was utterly impractical in the context of this thesis. Thus, a series of sampling steps were implemented. First, the corpus was limited to maps published in six countries and nineteen U.S. states (Fig. 3). The countries retained are the United States, France, Germany, the Netherlands, Switzerland, and Indonesia[7]. The American subset focuses more specifically on maps issued in the Northeast (ME, NH, VT, MA, CT, RI, NY, PA, NJ), the South Atlantic Coast states (DE, MD, VA, NC, SC, GA, FL), Louisiana, and North and South Dakota. Two criteria guided the choice of these countries: the existence of at least three catalogs per country and the presence of significant cultural ties among selected countries and regions. Notably, the five European countries are contiguous. The territories of the American Atlantic coast were colonized comparatively earlier by European powers, among which England, but initially also the Netherlands and France.

---

[6] Due to long download times, and in order to avoid overloading hosting servers, ADHOC Images was collected over an extended period of time, spanning from summer 2022 to fall 2023. The collection of ADHOC Images thus precedes that of ADHOC Records, which occurred between spring 2024 and fall 2024. For these reasons, the exact content of the fields may differ, as catalog data were updated and digital platforms underwent various changes. The normalization output might also vary marginally as NER models and geocoding APIs were upgraded during that period.

[7] The inclusion of Indonesia as a publication country was decided in view of its high representation in Dutch cartographic collections, thus constituting a special case compared to Western cartographic collections.

Louisiana and the Dakotas were included because of their significant French and German communities, respectively[8].

An additional verification step was then implemented to ensure that the retrieved images are indeed maps. The collections sometimes catalog as “maps” materials such as chronologies, diagrams, land profiles, celestial charts, maps of the moon or other planets, photographed globes, street indexes, and technical drawings of facades. The digitized images may also contain blank versos, cover pages, and texts. In the present case, any document in which the map occupies less than 10 % of the image is considered inadequate. These items, harvested during the automatic collection procedure, must be identified and discarded, preferably automatically.

This step was addressed using a ResNet50 classifier (He et al., 2015). First, each image was resized so that its longest side did not exceed 1,000 pixels. A sample of 5,403 images was manually labeled: 4,460 images (82.5 %) were maps, whereas 943 (17.5 %) were not. The classifier was trained to convergence using a binary cross-entropy loss function. Then, the model was applied to the 212,082 remaining images. For 3,010 images whose logit output—interpretable as a confidence index—fell between 5 % and 95 %, an additional manual review and correction step was performed. After this step, the overall classification accuracy can be estimated at 99.97 %. In total, 173,826 images were identified as maps (78 % of the corpus), whereas 38,256 harvested images were identified as non-maps.

In addition to refining the corpus, an inadvertent application of the classifier’s confidence index is to measure the typicality of map images: among the 3,010 manually inspected images, several atypical maps were indeed observed, generally associated with low confidence. Representative examples appear in Figures A3–A7 in the Appendix.

At the end, the corpus was further reduced by privileging entries with complete metadata for both place of publication and publication year, a criterion already applied to ADHOC Records. To this end, the location entries were georeferenced using Google V3 API[9]. Finally, when a single data entry corresponded to multiple images, a sample of 1 to 30 images was selected[10]. The resulting ADHOC Images corpus contains 99,715 digitized map images, corresponding to 83,481 records.

[8] Between 30% and 45% of the population of North and South Dakota is of German descent. Louisiana belonged to France until 1803, as did most of the Midwest. The state of Louisiana, specifically, retained a significant French-speaking minority until the middle of the 20th century.

[9] Which exhibited improved precision over Nominatim, as documented in Table 2. The location metadata in ADHOC Images are thus slightly more reliable compared to ADHOC Records.

[10] For each record, the image sample size is logarithmically proportional to the initial number of images.

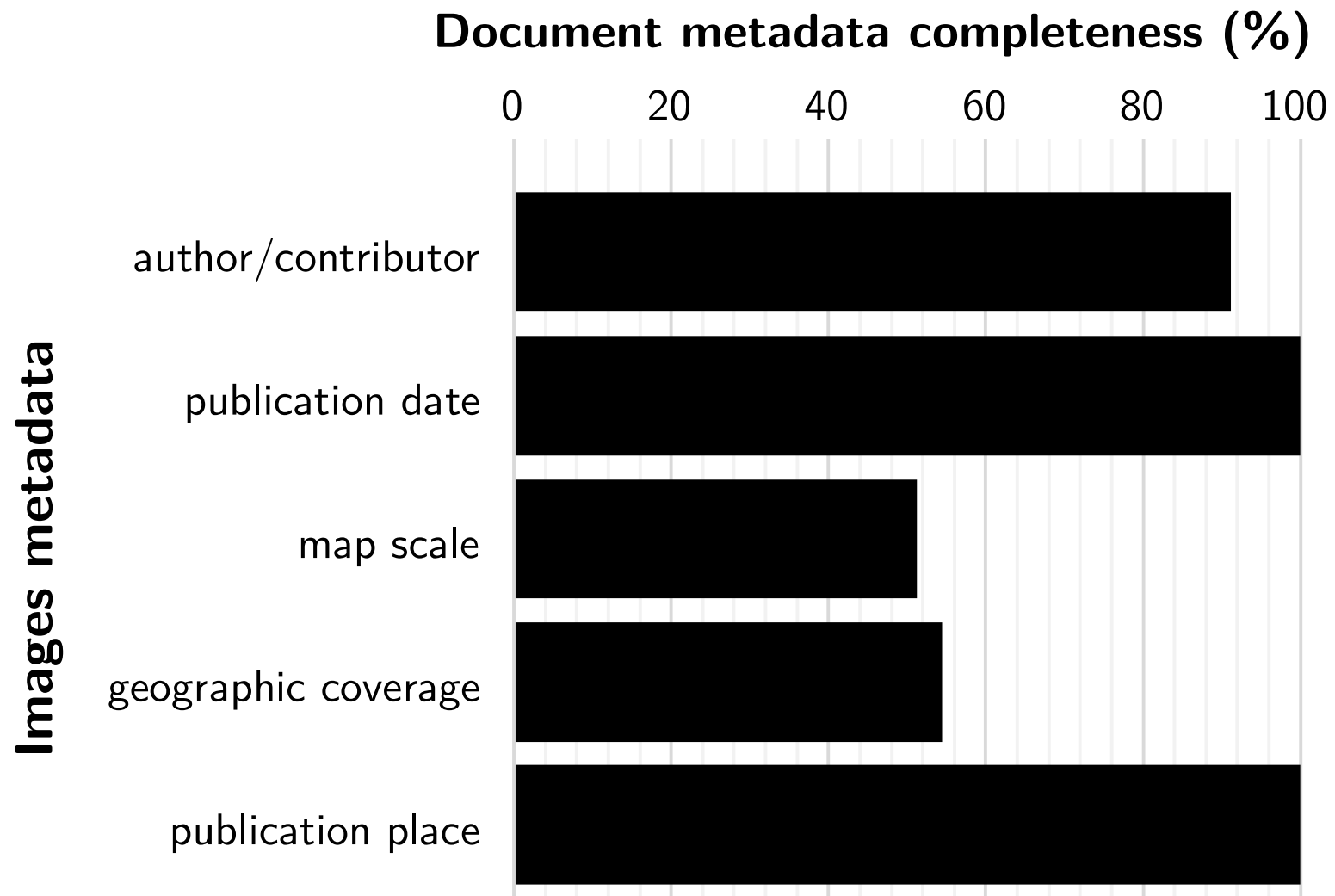

**Figure 4 | Completeness of the fields in the ADHOC Images dataset.** *The availability of date and place of publication are inclusion criteria; map scale is documented in ca. 51% of cases.*

### *Assessment of ADHOC Images and comparison with ADHOC Records*

As indicated in Figure 4, the metadata in ADHOC Images is comprehensive with respect to the date and place of publication. In more than 90 % of cases, the author and other contributors, such as the publisher, are also identified. Over half of the entries include an indication of scale, a proportion considerably higher than that observed in ADHOC Records. By contrast, the geographical coverage is less often documented. This is not very limiting as the second part of the dissertation will extract content information directly from map images (c.f. Chapters 4 and 5).

Figure 5 compares the metadata distributions of ADHOC Records and ADHOC Images and evaluates the delta between the two corpora. The Kolmogorov–Smirnov (KS) statistic quantifies the dissimilarity between the two distributions based on their empirical cumulative distribution functions, noted $F(x)$ and $G(x)$, respectively. The KS statistic is defined as:

$$KS = sup_x|F(x) - G(x)| \tag{1}$$

Simply put, the statistic measures the maximum deviation between the two cumulative functions. The Kolmogorov–Smirnov–type statistic $\widehat{KS}$ is also defined for cumulative-sum paths.

$$\widehat{KS} = sup_x\left|\hat{F}(x) - \hat{G}(x)\right| \tag{2}$$

$\widehat{KS}$ is calculated identically to the traditional KS statistic. However, it takes two cumulative-sum processes $\hat{F}(x) = \int_0^x f(x)$ as inputs instead of two cumulative distribution functions.

Figure 5 indicates that the chronological distributions of the two datasets are broadly similar ($\widehat{KS} = 14.3\%$). Most short-term fluctuations also align until the late 18[th] century. Later, ADHOC

Records continues to rise, peaking during World War II, whereas ADHOC Images exhibits a downward trajectory from the 1890s onward, with only a local peak during World War II. This divergence is most plausibly attributable to copyright constraints. As noted in Section 1.1, copyright generally expires seventy years after the death of all authors, although in certain jurisdictions the term may extend to one hundred years. Accordingly, many maps published after the late 19$^{th}$ century remain outside the public domain and therefore cannot yet be disseminated online by heritage institutions[11].

The distribution of map scales differs more markedly between the two datasets ($\widehat{KS} = 46.2\%$). Large-scale maps (e.g., city maps) and 1:25,000, 1:50,000, and 1:100,000 maps are much less frequent in ADHOC Images. In contrast, country maps or world maps (1:1,000,000–1:50,000,000) occur comparatively more often in ADHOC Images. Harmonic scale peaks are observed for ADHOC Records, which suggests a higher representation of map series, compared to ADHOC Images where these scales are less represented. The cause of remaining scale distribution discrepancy is unclear. A possible explanation could be the lower documentation of graphic scales for non-digitized maps (Fig. 1, Fig. 4).

The final statistic presented in Figure is the distribution of the number of maps published per creator. In the figure, creators are ordered from the most to the least productive according to the number of maps they published. The cumulative statistic shows that the two distributions are very similar ($KS = 11.6\%$). ADHOC Images appears slightly more distributed, whereas the relative share of large creators is marginally higher for ADHOC Records. ADHOC Records also contains more unique authors overall, which is expected because the database is larger. These findings seem consistent with scale effects and attest to the wide diversity of authors and map types represented in ADHOC Images.

[11] Let us take the example of a map published in 1895, where at least one of the authors was 20 years old at the time of publication. If this author lived another 60 years, until 1955, when he was 80 years old, the map did not enter the public domain before 2025, i.e., 130 years after its publication.

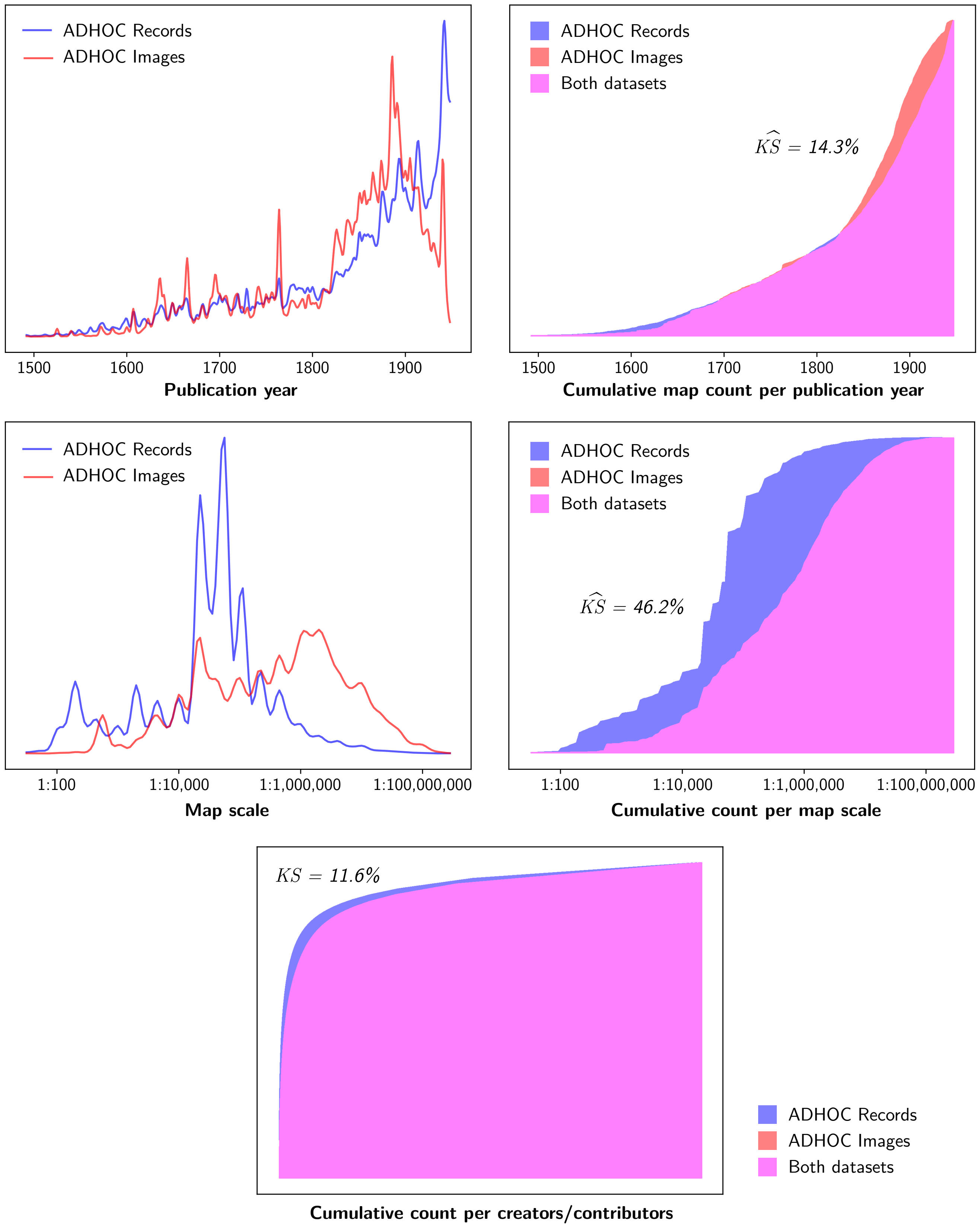

**Figure 5 | Comparison of metadata distributions between ADHOC Records and Images datasets.** Normalized curves (left) and cumulative counts (right). *KS*, respectively $\widehat{KS}$, report the regular Kolmogorov–Smirnov statistic, respectively KS–type statistic for cumulative-sum paths, expressing distribution dissimilarity. *The two distributions are similar regarding publication year and diversity of creators but differ according to map scale.*

The maps in Figures 6 and 7 enable a qualitative comparison of publication places, and locations depicted, for both databases. The decision of restricting the Images corpus to five countries and nineteen U.S. states is reflected in Figure 6. The selected countries (France, Switzerland, Germany, the Netherlands, and Indonesia) appear in similar proportions across the two datasets. In the United States, the principal publication centers—New York, Philadelphia, and Boston—are well represented, whereas Washington, Chicago, and San Francisco occur less frequently in the Images database. As their mauve shading indicates, these centers are not absent though, since the corpus includes numerous cases of co-publication. Within Europe, publication centers that are less represented in the Images corpus comprise London, Madrid, Florence, Rome, Venice, Vienna, St. Petersburg, and Moscow. More generally, the British Isles, the Iberian Peninsula, and Italy are comparatively less represented in the latter. Japan—particularly Tokyo—as well as Australia (especially the south-east) and Cuba are likewise comparatively infrequent. Indonesia, by contrast, is slightly overrepresented in the image database, as indicated by its fuchsia hue. A more systematic analysis of publication centers represented in ADHOC Records will be presented in Chapter 2.

The two databases appear quite congruent with respect to depicted locations. Geographic coverage will be the focus of Chapter 3. Thus, only the differences between the two datasets will be highlighted here. Specifically, certain cities seem to be comparatively less frequently depicted in the image corpus, including Los Angeles, London, Berlin, Gibraltar, Rome, and Tokyo. Several countries or regions are likewise underrepresented, including Cuba, Japan, Iceland, and southeastern Australia. The remainder of the world appears equally covered by both datasets, suggesting that, in this respect, the image corpus is quite similar of the larger Records corpus.

Overall, although the ADHOC Images corpus is almost ten times smaller than ADHOC Records, it remains representative of the latter, except for publication locations—which are limited expressly—and map scale. Regarding scale, ADHOC Images primarily subsamples ranges that are overrepresented in ADHOC Records, and thus arguably presents an improved distribution balance. Map images from the early 20$^{th}$ century are also slightly underrepresented because of copyright restrictions. Noting these differences will facilitate the discussion of the results from Chapter 4 and following, in light of the broader database composition.

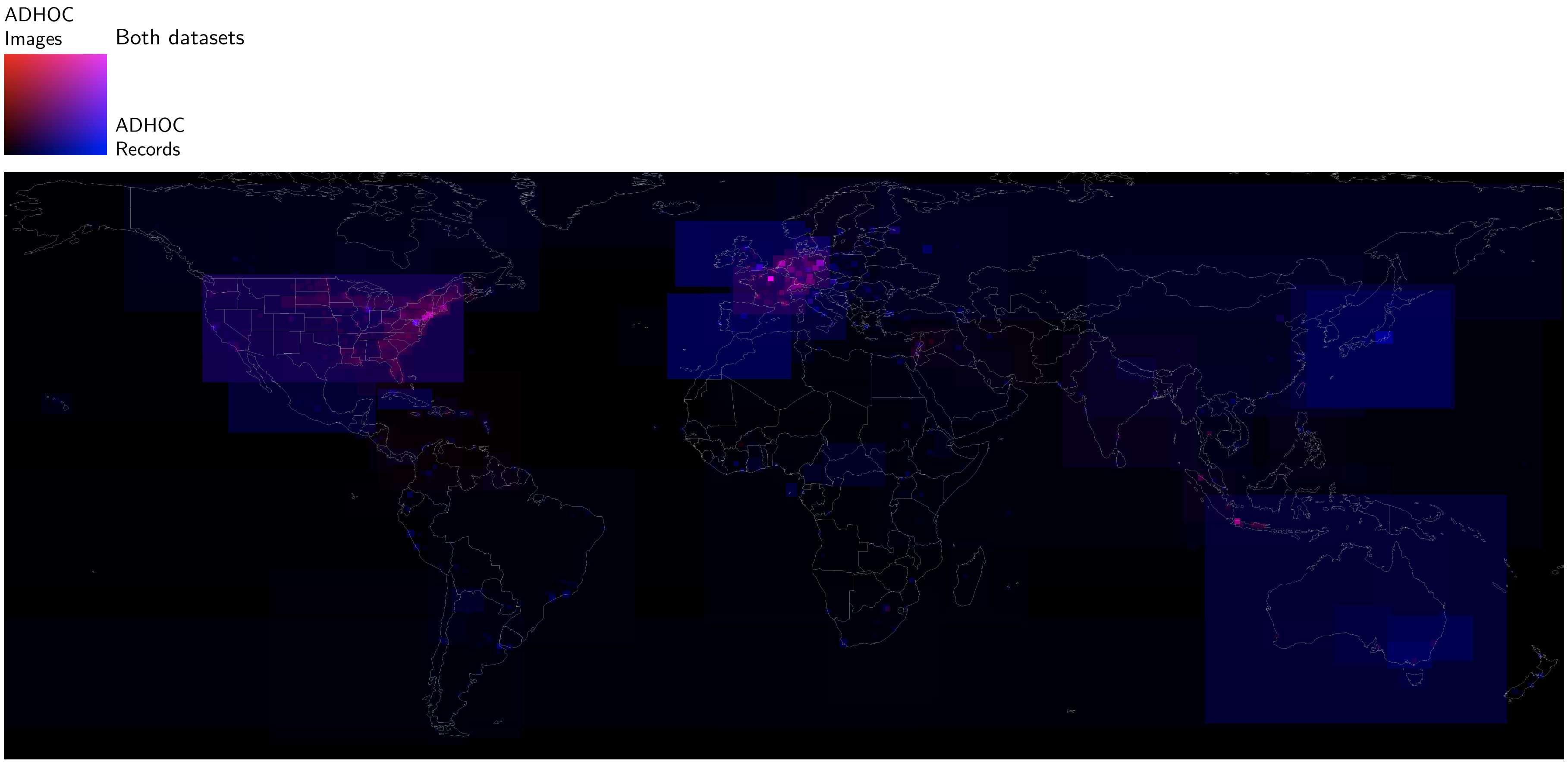

**Figure 6 | Publication places documented in ADHOC Records and ADHOC Images datasets.** Color intensity is relative and denotes the number of map records published in a particular area. Blue hues indicate that the area is a frequent place of publication in ADHOC Records, while red hues indicate that the area is a frequent place of publication in ADHOC Images. Pink areas are common to both datasets. Colors are not categorical, but express a gradient. Besides entry counts, color intensity is degressive with respect to the extent of the area. *New York and Paris appear as the main publication centers in both datasets, whereas Chicago, Washington D.C. and London are more represented in ADHOC Records.*

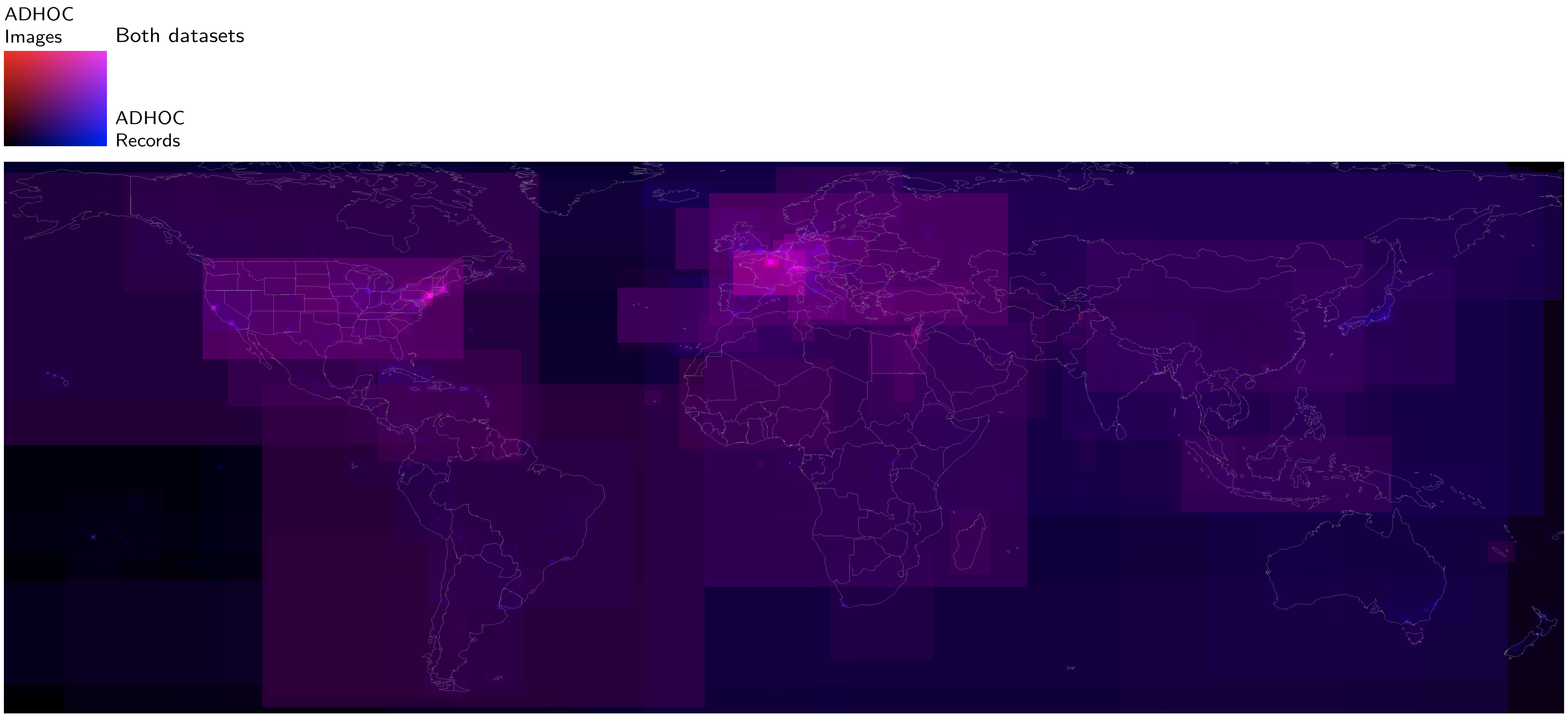

**Figure 7 | Locations depicted in the ADHOC and ADHOC Images datasets.** Color intensity is relative and reflects the number of map records depicting a particular area. Blue hues indicate that the area is frequently depicted in ADHOC Records, while red hues indicate that the area is frequently depicted in ADHOC Images. Pink areas are common to both datasets. Colors are not categorical, but express a gradient. Besides entry counts, color intensity is degressive with respect to the extent of the area. *New York, Paris, Zurich and France are among the areas most depicted in both datasets.*

### *A brief overview of the diversity of maps included in ADHOC Images*

The diversity of ADHOC Images is illustrated in Figure 8, which displays a random sample of map thumbnails from the image corpus. This figure showcases the range of scales, themes, and map styles. Figures A3–A7 in the Appendix present atypical examples. For instance, Figure A3 is a mnemonic map conceived as a way to learn the geography of the world and the French departments by mentally associating specific parts of the hand with geographic locations. Figure A4 is a burlesque jigsaw puzzle of Europe, intended for children, in which countries are represented by caricatured physiognomies. Figure A5 is a completely empty canvas, except for the cartouche "Pacific Ocean. Drawn by …" and the graticule of an equal-area projection. This map was originally included in an exercise book for geography students; it was to be completed with the contours of the continents and Pacific islands. Figure A6 is an allegorical map of the Canton of Bern, Switzerland, in which the canton and its dependencies (Vaud, Fribourg, and Jura) are depicted as a menacing bear, poised to defend the Swiss Confederacy against its enemies. Finally, Figure A7 is a 19$^{th}$ century embossed paper atlas map of the USA for visually impaired people. These cases represent only a fraction of the diversity of cartographic forms and the rich scope of cartographic heritage.

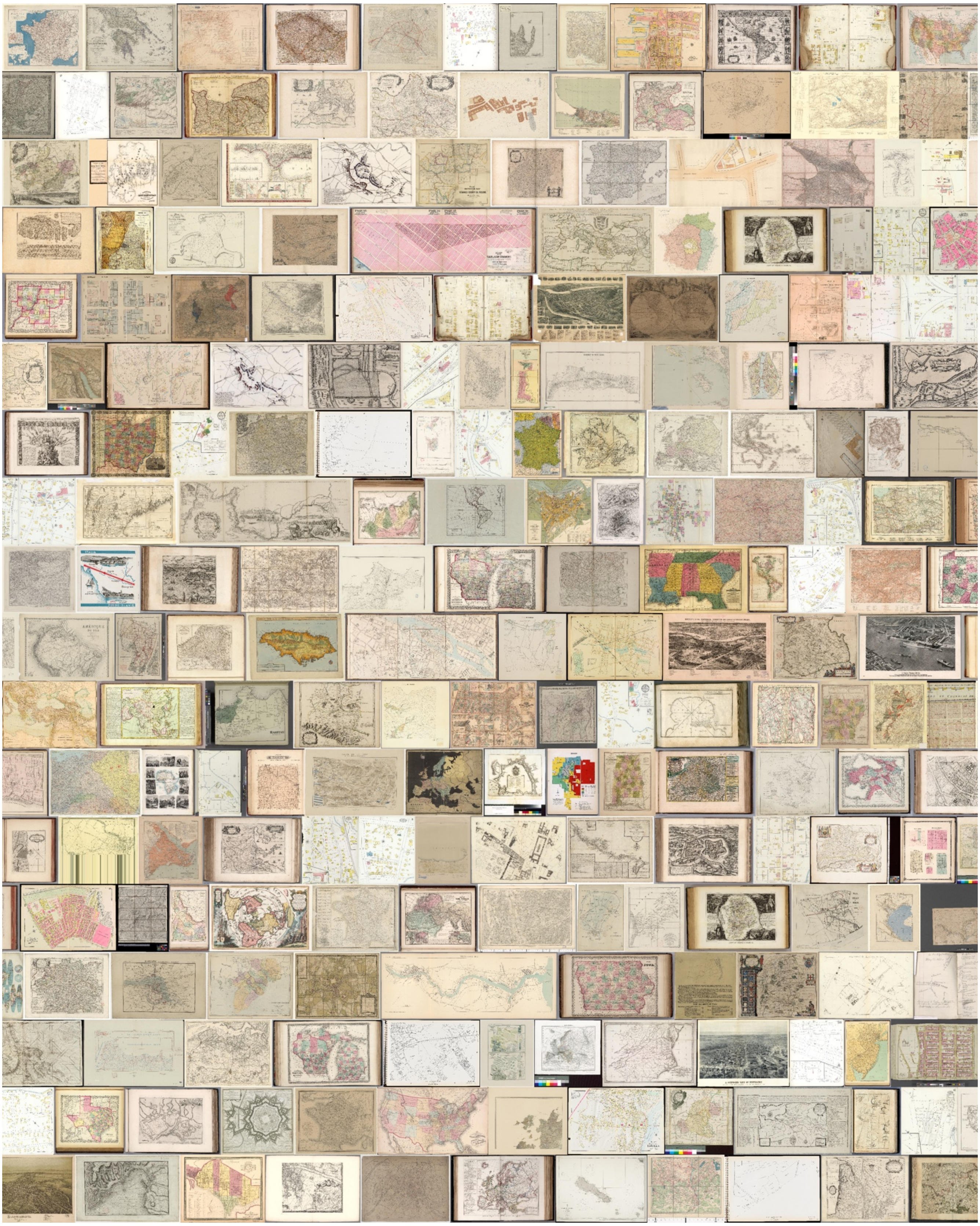

**Figure 8 | Mosaic of maps randomly sampled from ADHOC Images,** showing the diversity of maps in the corpus.

## 1.8 Conclusion

This chapter presented the scale and scope of the corpus, as well as the methodology employed to constitute the Aggregated Database on the History of Cartography. Composed of two related datasets, ADHOC Records and ADHOC Images, the database paves the way for a digital history of cartography.

Containing 771,561 records aggregated from 38 library catalogs across 11 countries, ADHOC Records is arguably the largest and most diverse dataset on the history of cartography. Several steps were implemented to enhance metadata completeness, disambiguate entries, normalize them, and eliminate redundant records. With 99,715 digitized images, ADHOC Images also constitutes a unique corpus for the study of cartographic figuration and geographic representation. The previous pages further documented the scope and limitations of both related datasets, facilitating future updates and improvements. Both datasets are to be released with this doctoral thesis (Petitpierre, 2025a), facilitating downstream studies, complement, and updates.

The database will serve as the basis for the next chapters, which will focus on developing approaches and methods for the computational study of historical map records and images.

# Appendix A – Supplementary materials

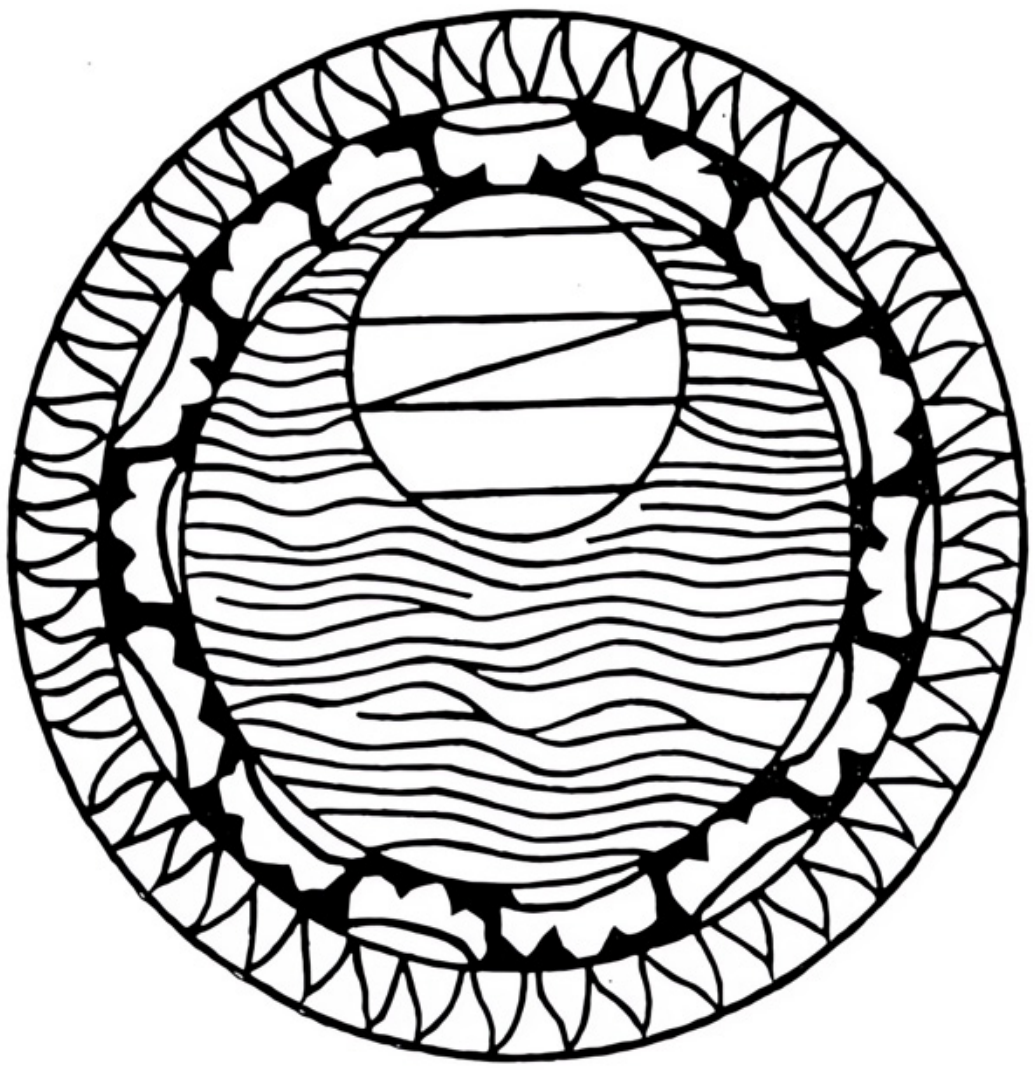

**Figure A1 | Cosmographical model of Paul Burgos**, depicted in Johannes de Sacrobosco's Sphaera between 1485 and 1490. The terrestrial orb emerges from the ocean. Water and earth are, in turn, surrounded by air and fire, which together constitute the four Aristotelian elements.

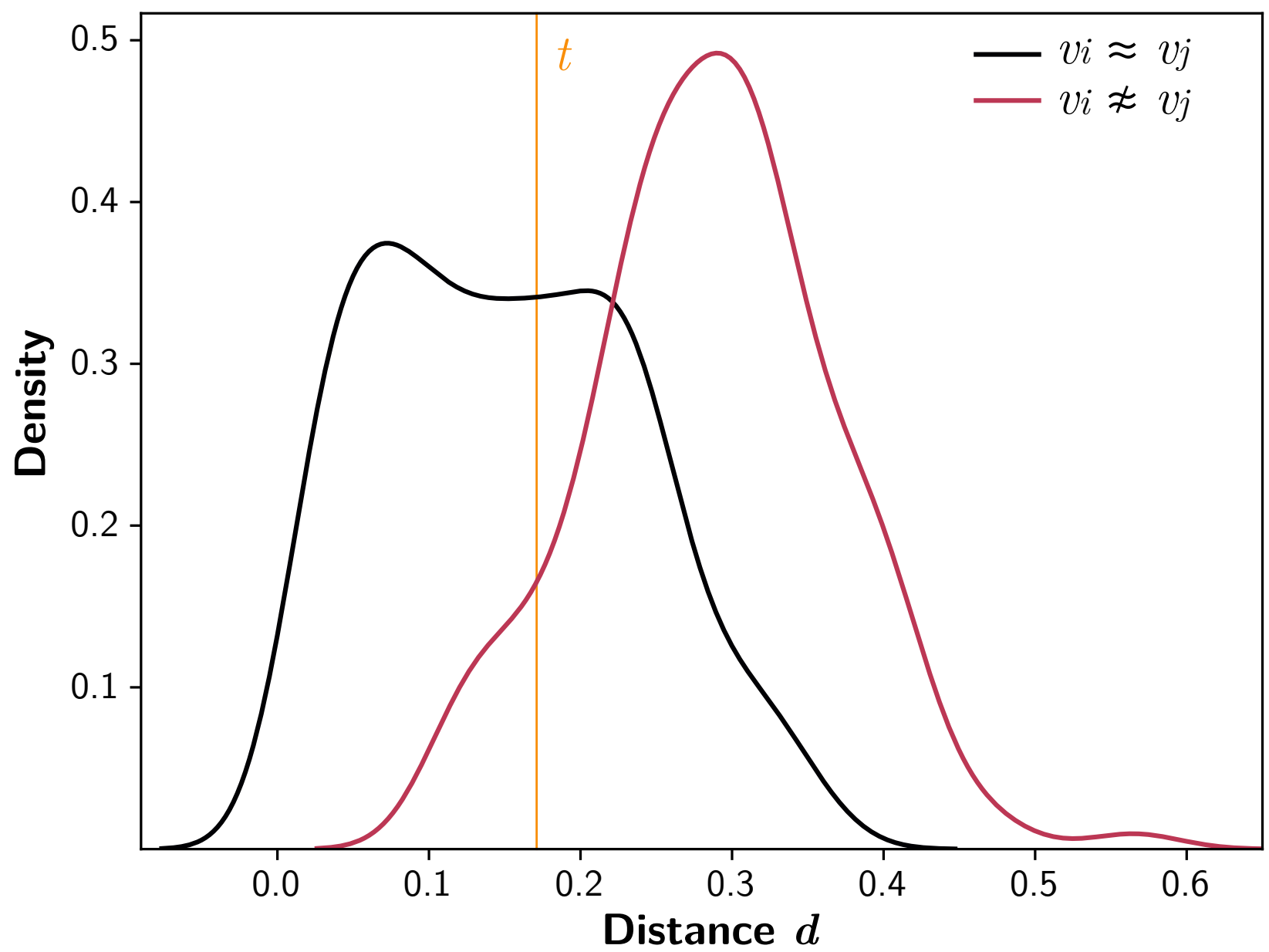

**Figure A2 | Kernel density estimation** of the cosine distance d between cognate ($\vec{v_i} \approx \vec{v_j}$) and distinct ($\vec{v_i} \not\approx \vec{v_j}$) variant name pairs. A vertical yellow bar indicates t, the decision threshold used for classification.

**Table A1 | Detail of the provenance of the ADHOC map records.** The columns # ADHOC Records and Images report the number of records, per catalog/institution. The country code corresponds to the ISO-2 code. FR = France, US = United States, JP = Japan, DE = Germany, ES = Spain, AU = Australia, IT = Italy, CH = Switzerland, DK = Denmark, PT = Portugal, NL = Netherlands.

| Catalog/Institution | # ADHOC Records | # ADHOC Images | Website | Country | Aggregator | National lib. | Gov. agency | City library | University | Others |
|---|---|---|---|---|---|---|---|---|---|---|
| Bibliothèque nationale de France | 155,630 | 21,571 | gallica.bnf.fr | FR | | X | | | | |
| Stanford Library | 111,977 | · | library.stanford.edu | US | | | | | X | |
| David Rumsey | 108,148 | 29,272 | www.davidrumsey.com | US | | | | | | X |
| National Diet Library | 67,094 | · | www.ndl.go.jp | JP | | X | | | | |
| Deutsche Nationalbibliothek | 45,149 | · | www.dnb.de | DE | | X | | | | |
| Biblioteca Virtual de Defensa | 39,078 | · | bibliotecavirtual.defensa.gob.es | ES | | | X | | | |
| National Library of Australia | 36,984 | · | www.nla.gov.au | AU | | X | | | | |
| United States Geological Service | 33,390 | · | www.usgs.gov | US | | | X | | | |
| Library of Congress | 25,894 | 10,962 | www.loc.gov | US | | X | | | | |
| Internet Culturale | 23,919 | · | www.internetculturale.it | IT | X | | | | | |
| swisscovery | 15,097 | · | epfl.swisscovery.slsp.ch | CH | X | | | | | |
| Institut Cartogràfic i Geològic de Catalunya | 12,690 | · | www.icgc.cat | ES | | | X | | | |
| Det Kgl. Bibliotek | 11,930 | · | www.kb.dk | DK | | X | | | | |
| Leiden University Libraries | 11,185 | 6,672 | digitalcollections.universiteitleiden.nl | NL | | | | | X | |
| Boston Public Library | 9,746 | 2,930 | www.bpl.org | US | | | | X | | |
| New York Public Library | 8,330 | 2,137 | www.nypl.org | US | | | | X | | |
| e-rara | 8,205 | 4,428 | www.e-rara.ch | CH | X | | | | | |
| Biblioteca Nazionale Centrale di Firenze | 7,266 | · | www.bncf.firenze.sbn.it | IT | | X | | | | |
| Harvard Library | 5,272 | · | library.harvard.edu | US | | | | | X | |

| Catalog/Institution | # ADHOC Records | # ADHOC Images | Website | Country | Aggregator | National lib. | Gov. agency | City library | University | Others |
|---|---|---|---|---|---|---|---|---|---|---|
| UT Libraries | 4,738 | · | www.lib.utexas.edu | US | | | | | X | |
| UC Berkeley Library | 4,030 | 208 | www.lib.berkeley.edu | US | | | | | X | |
| Deutsche Fotothek | 3,822 | · | www.deutschefotothek.de | DE | | | | | | X |
| Biblioteca Nacional de Portugal | 3,674 | · | www.bnportugal.gov.pt | PT | | X | | | | |
| Gemeinsamen Bibliotheksverbundes | 2,884 | · | www.gbv.de | DE | X | | | | | |
| The Memory | 2,299 | · | geheugen.delpher.nl | NL | X | | | | | |
| Université Bordeaux Montaigne | 2,161 | 991 | www.u-bordeaux-montaigne.fr | FR | | | | | X | |
| Universität Münster | 2,139 | 640 | www.ulb.uni-muenster.de | DE | | | | | X | |
| Südwestdeutsche Bibliotheksverbund | 1,481 | · | www.bsz-bw.de | DE | X | | | | | |
| München Digitization Center | 1,380 | 391 | www.digitale-sammlungen.de | DE | | | | | | X |
| Université Paris 8 | 1,339 | · | www.univ-paris8.fr | FR | | | | | X | |
| e-manuscripta | 1,159 | 42 | www.e-manuscripta.ch | CH | X | | | | | |
| University of Illinois Library | 976 | 106 | www.library.illinois.edu | US | | | | | X | |
| Universitäts- und Landesbibliothek Sachsen-Anhalt | 975 | 536 | bibliothek.uni-halle.de | DE | | | | | X | |
| Universiteit van Amsterdam | 683 | 372 | uba.uva.nl | NL | | | | | X | |
| Instituto Geográfico Nacional | 670 | · | www.ign.es | ES | | | X | | | |
| Princeton University Library | 167 | 2,223 | library.princeton.edu | US | | | | | X | |

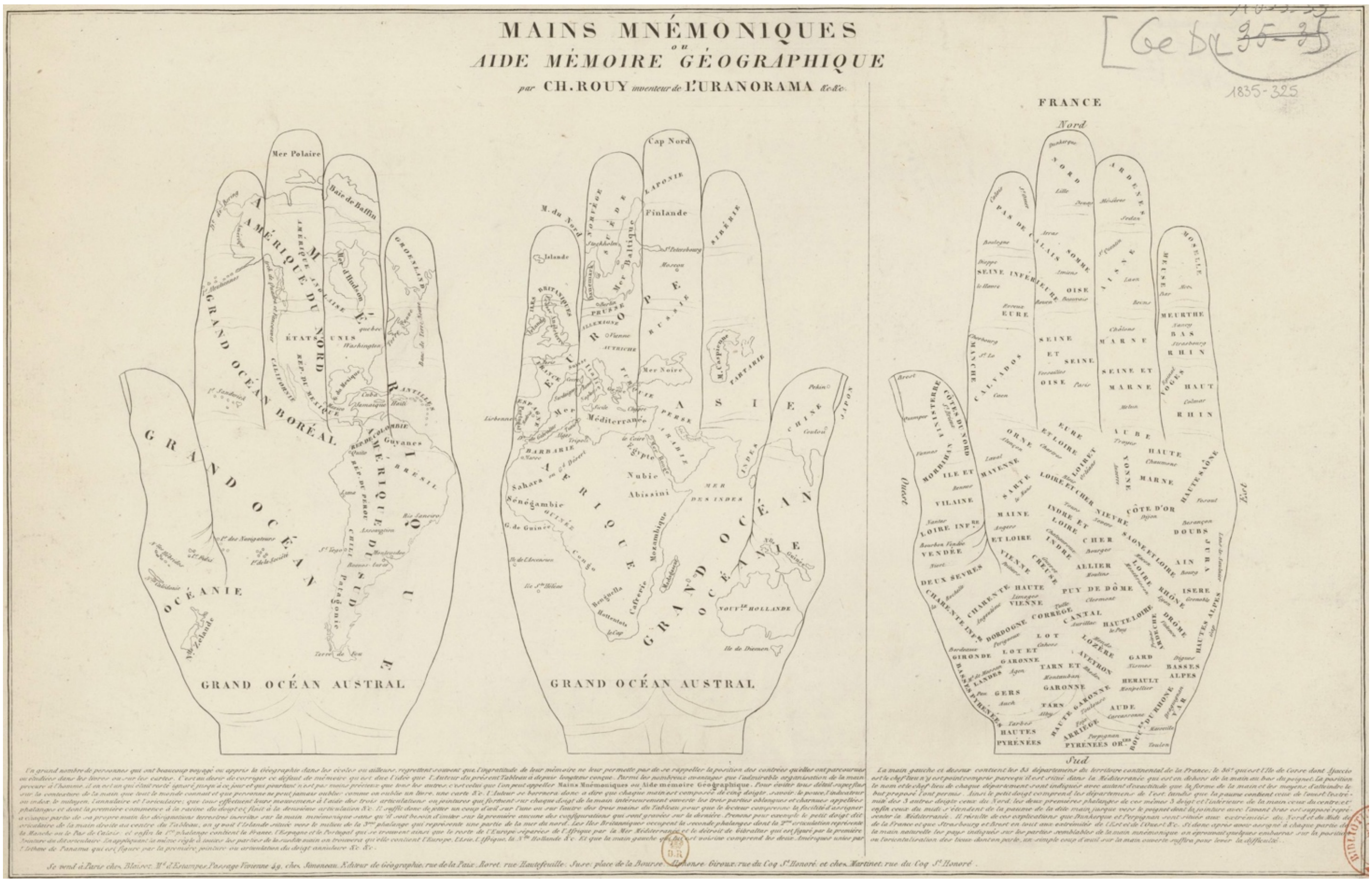

**Figure A3 | Mnemonic hands, geographical memory aid, 1835.** Charles Rouy. *Mains mnémoniques, Aide mémoire géographique*, 1835. Published by Blaisot, Paris. 28.5 x 44.5 cm. BnF, Ge DL 1835–325. URL: gallica.bnf.fr/ark:/12148/btv1b53035378g

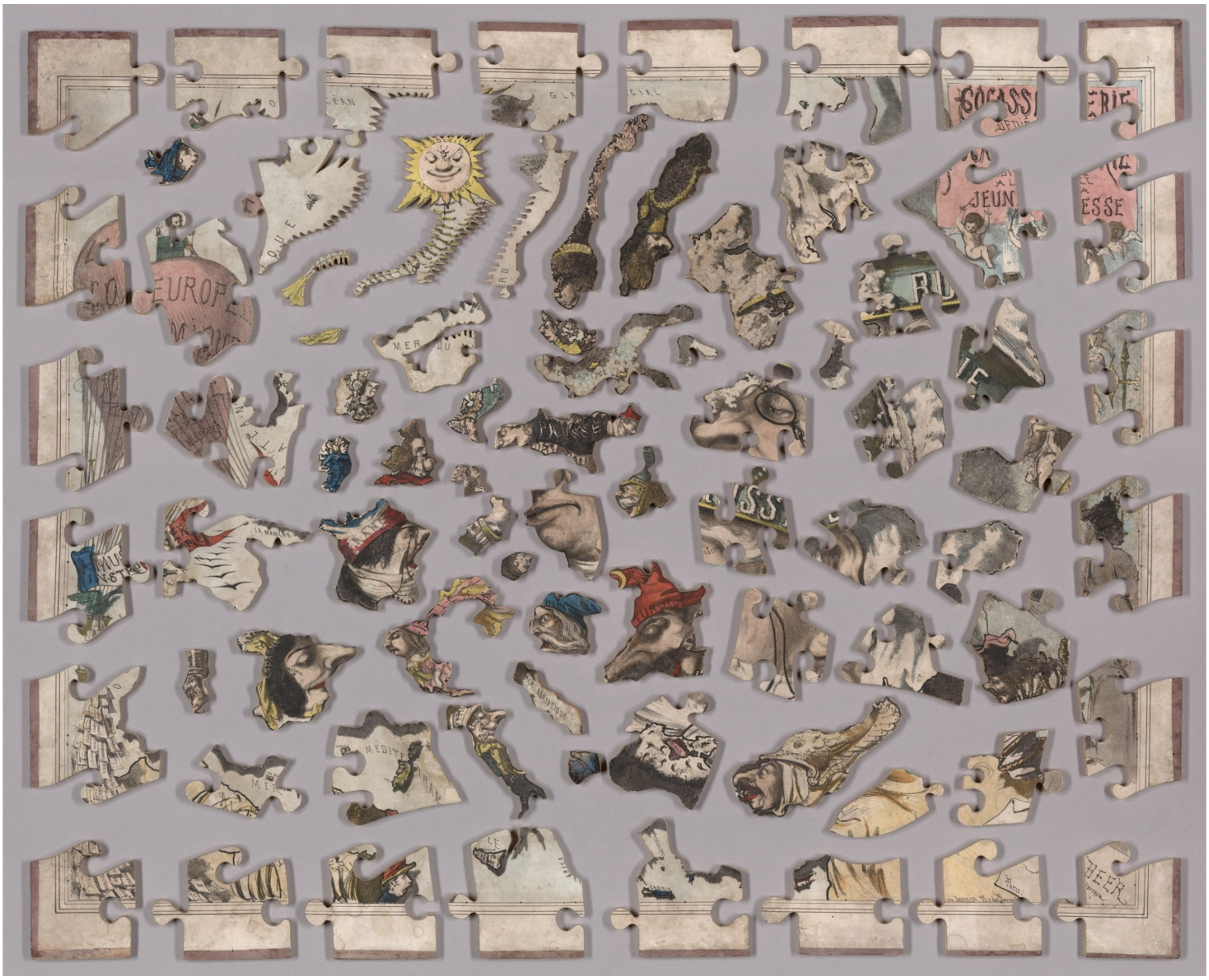

**Figure A4 | Comic jigsaw puzzle of Europe, 1867.** André Belloguet, H. Jannin. *L'Europe Comique, Cocasserie dediée à la Jeunesse*, 1867. Published by Beer. Lithography. 45 x 62 cm. David Rumsey Collection, 10147.002. URL: davidrumsey.com/luna/servlet/detail/RUMSEY~8~1~305137~90075642

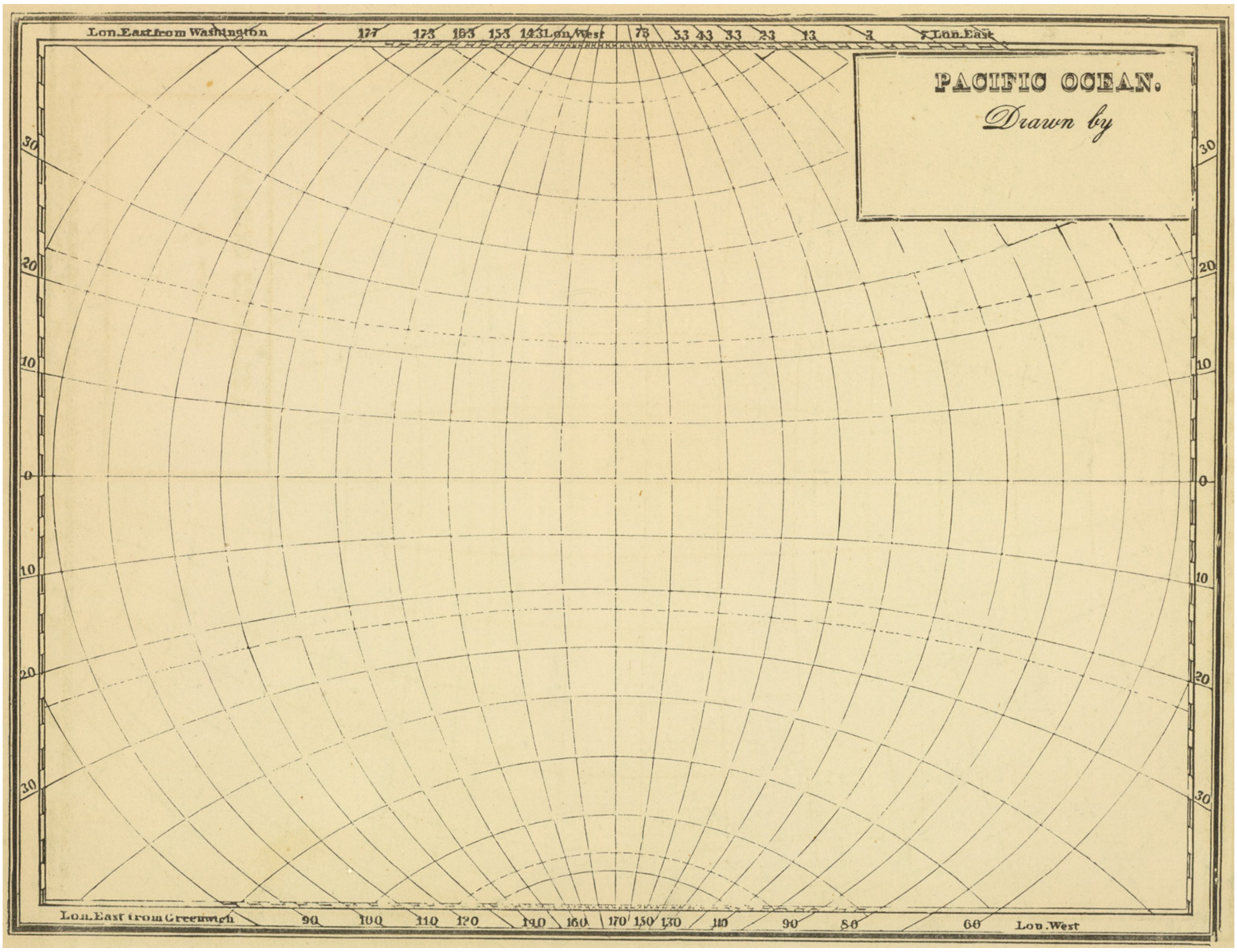

**Figure A5 | Empty map of the Pacific Ocean in a school atlas, 1850.** George W. Fitch. *Pacific Ocean*, 1850. Published by Blakeman, Sheldon & Co, New York. 27 x 21 cm. David Rumsey Collection, 0289.000. URL: davidrumsey.com/luna/servlet/detail/RUMSEY~8~1~28079~1120214

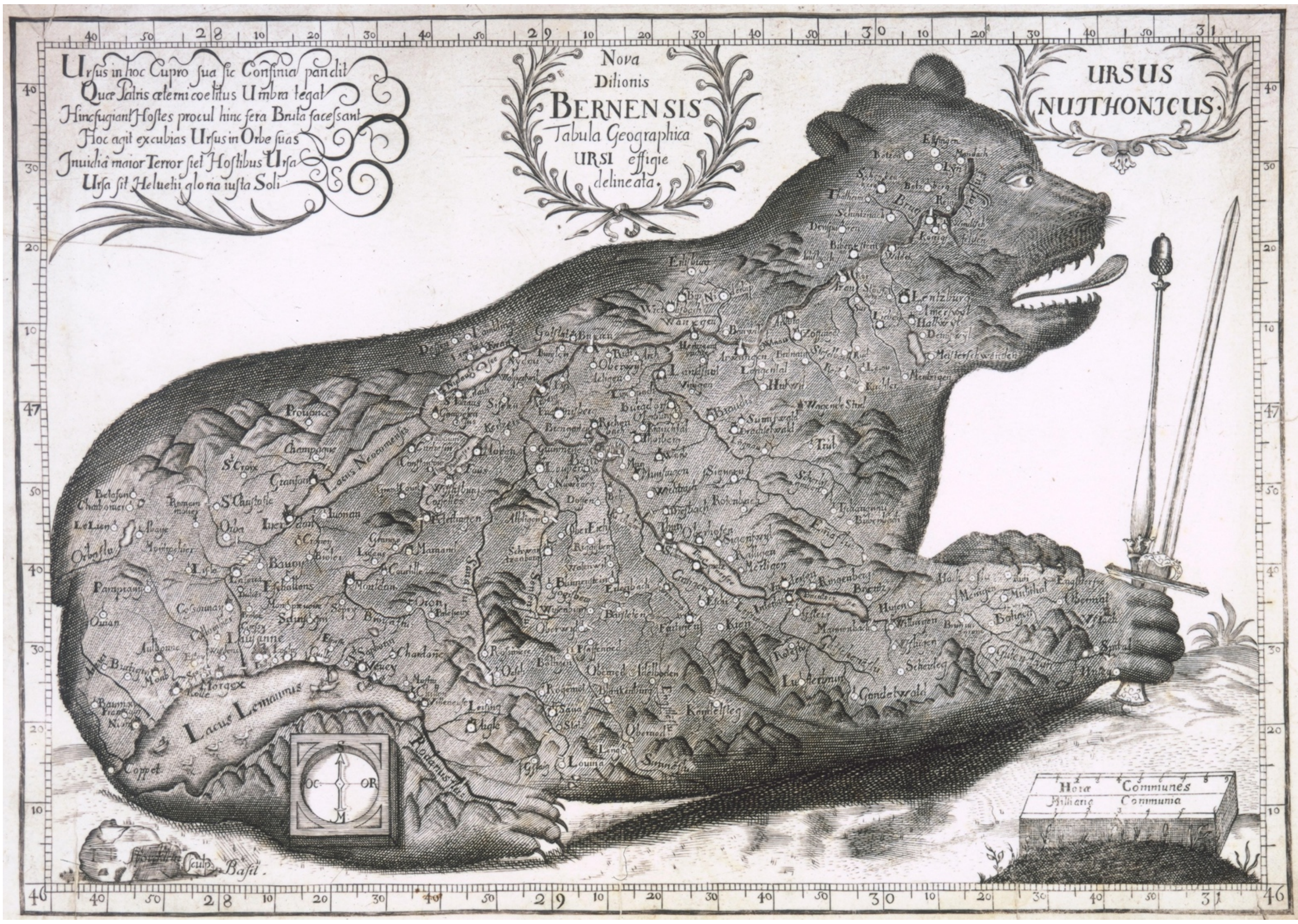

**Figure A6 | Allegoric map of the Canton of Bern, represented as a bear, around 1700.** Jakob W. Störcklein. *Ursus nuithonicus: Nova ditionis Bernensis tabula geographica ursi effigie delineata*, 1879. Basel. Copperplate. 23 x 33 cm. University Library Bern, MUE Ryh 3211:25A. doi: 10.3931/e-rara-96513

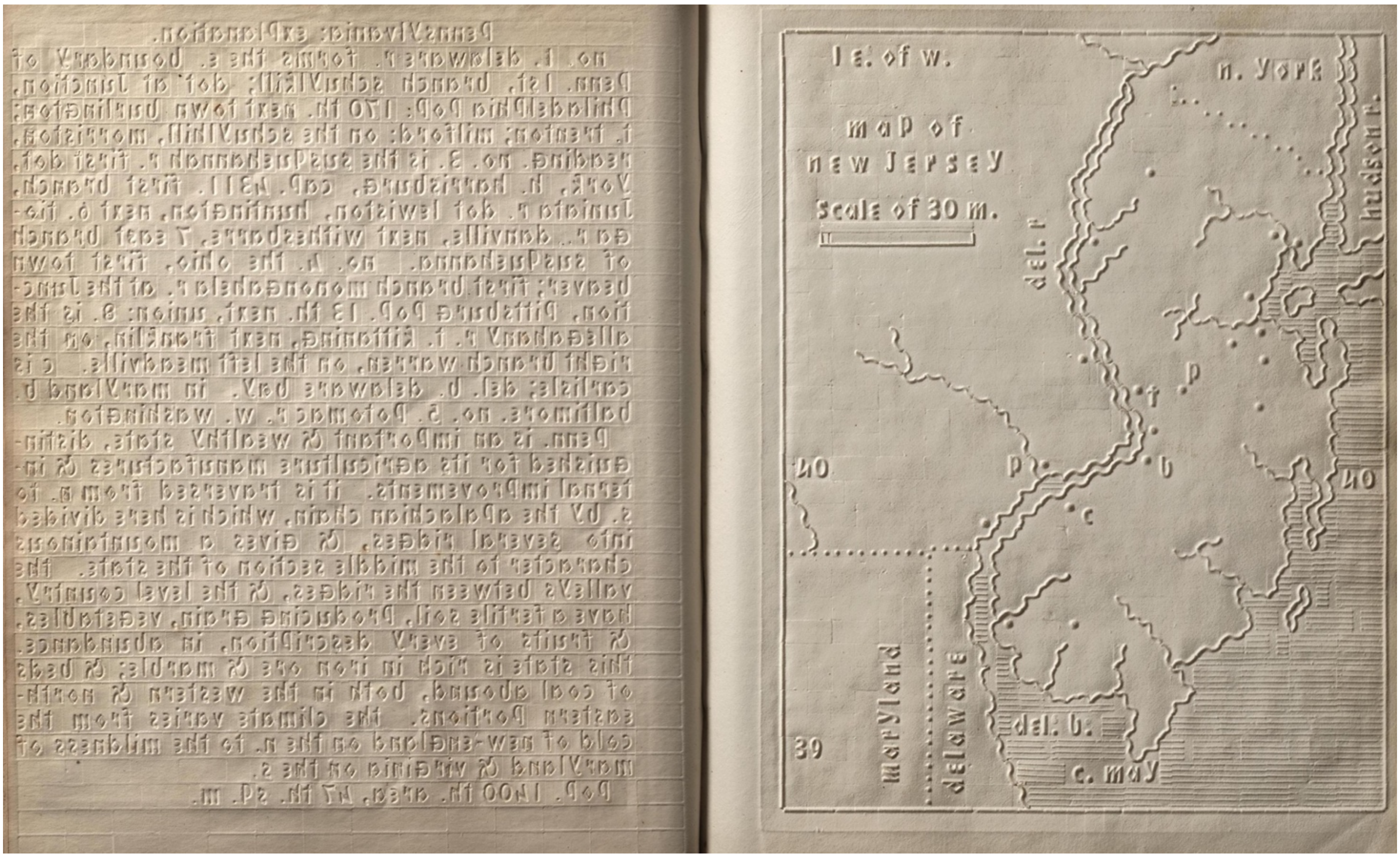

**Figure A7 | Embossed atlas of the United States for the visually impaired people, 1837.** Samuel G. Howe, Stephen P. Ruggles. *New Jersey*, 1837. Published by the New England Institution for the Education of the Blind, Boston. Embossed paper. 29 x 24 cm. David Rumsey Collection, 5956.000. doi: URL: davidrumsey.com/luna/servlet/detail/RUMSEY~8~1~226468~5506802

# Chapter 2

## *The Making of Cartography*

This chapter investigates ADHOC Records, the Aggregated Database on the History of Cartography, assembled in Chapter 1. Specifically, it focuses on the visualization and quantitative data analysis of the spatial, temporal, and social dimensions of map publication. This systematic approach, enabled by the extent of collected data, underscores cultural, economic, and political structures that have profoundly marked the history of cartography.

First, the representativeness of the corpus will be investigated from the perspective of national publishing figures. This step also provides information on the relative and historical volumes of cartographic production. Second, the chapter will address the basic geographical unit of map production: the city. European, American, Japanese, and Australian urban production centers are mapped at a granular level, revealing large-scale geographical structures of map publication and distinct national production strategies. The third part of this chapter examines the chronology of map production, with particular attention to national trajectories; this perspective highlights periods of sustained activity, which I discuss in relation to the existing historiography. Finally, I approach production from the perspective of its actors by reconstructing the social graph of map makers and cartographic agencies, based on collaboration relationships. The network analysis exposes social structures that underlie spatial and historical patterns. It also reveals the progressive diffusion of cartographic practices through the network of collaboration, thereby highlighting several pioneers and early adopters. The final section examines the different roles of map makers within the network, using a deep semantic approach.

## 2.1 National publishing figures

Focusing first on the national scale (Fig. 1a), one can observe that the country most represented by the total number of publications is the United States, with almost 160,000 maps, followed at a distance by France, Germany, Great Britain, and Japan. Together, these five countries account for 78 percent of all publications. The presence of Great Britain in the top five is surprising, since no British collection was included in the ADHOC dataset. One can even better gauge the importance of British cartography by canceling out the collection effect, which can be achieved by excluding publications from the country in which the heritage institution is located. In that case (Fig. 1b), Great Britain ranks first, followed by Germany, France, and the Netherlands. The United States appears only in fifth position, almost ex aequo with Italy, and is followed by Austria. This suggests a probable over-representation of American maps in the corpus, as well as of other comparatively smaller producers, such as Spain and Switzerland, due to collection biases.

Figure 1d provides an estimate of the most frequently depicted geographic locations. In this instance, the most common country is the United States[1], followed by France, and Germany. However, when excluding domestic collections (Fig. 1e), Great Britain once again ranks first, followed by France, Germany, the United States, and Italy.

Figure 1c features the countries predominantly focused on depicting foreign locations. In this respect, Germany appears to be the largest producer of non-domestic maps, followed by France, Great Britain, and the United States, with the Netherlands and Japan close behind. Conversely, the country that attracts the most attention from other countries (Fig. 1f) is the United States, followed by France. It is also noteworthy that when collection and production biases are excluded (Figs. 1e, 1f), several countries outside the historical West rank among the most depicted[2]—notably China, Russia, India, and Indonesia in Asia; Brazil, Cuba, Mexico, and Argentina in Latin America; Tanzania and Uganda in Africa. Several of these territories were major theaters of Western colonization. The magnitude of the phenomenon, reflected in the publication volumes, warrants the future investigation of quantitative relationships between colonization and map publication. This issue will be addressed in Chapter 3.

[1] Independently of map scales and levels of detail.

[2] In this Chapter, the geographic coverage by country does not refer to country maps per se. The country is merely used as the higher hierarchical level of place locations. For instance, a map focusing on the city of Amsterdam is counted as a map covering the Netherlands.

While this preliminary data analysis reveals the major historical role of England—which was excluded from the ADHOC dataset due to the existing risks of representation bias—it nevertheless reassuringly indicates that other leading countries are properly represented. The findings also provide a first quantitative view of the historical balance of cartographic publication volumes.

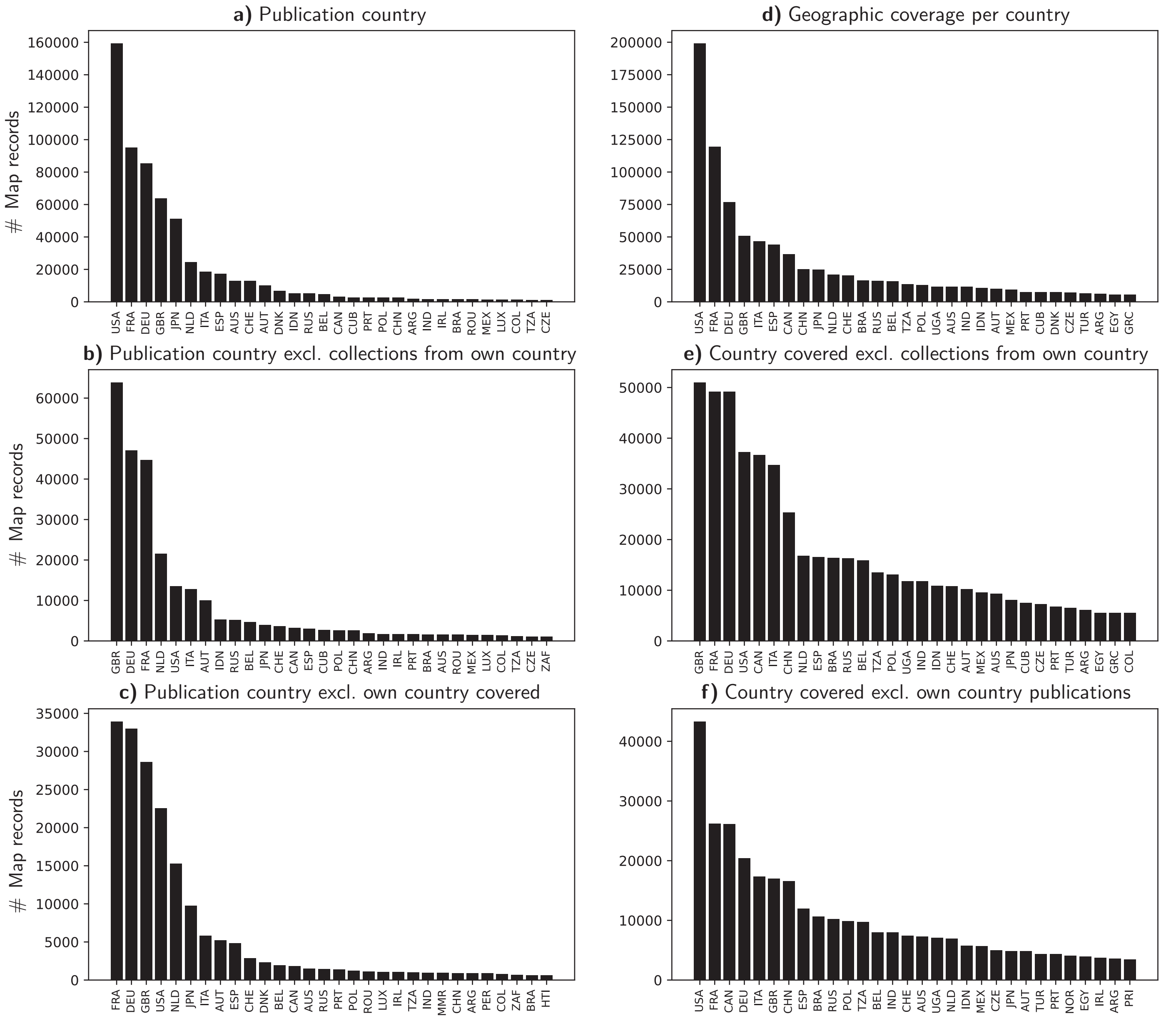

**Figure 1 | Number of map records by publication country (a–c) and by geographic coverage country (d–f).** For each bar plot, only the top 30 countries are shown. The acronyms on the horizontal axis correspond to ISO-3 country codes, whose correspondences are reported in the Appendix. *The United States appear the most common publication country in the corpus. However, it is superseded by the United Kingdom after correcting for collection biases.*

## 2.2 Map production centers

Regional and national statistics provide an overview of cartographic publication volumes and coverage. However, mapmaking was often localized in a few urban centers with specialized craftsmen who, together, formed production centers. These centers can be located and visualized, providing a closer look at the geographic and historical dynamics of map publication.

Figures 3–6 depict publication centers across the United States, Western Europe, and the Pacific region. Here, cartographic representation relies on publication orbs, a visual device whose operation is explained in Figure 2. The area of each circle is proportional to the total number of maps published in the corresponding publication center. The surrounding chronological arcs function like a clock, documenting the temporal extent of publication activity. The inner, thinner arc marks the years of the earliest and latest publications in the city, whereas the outer, thicker arc identifies the most active publication period—during which 90 percent of local publications occurred. In addition, I introduce the concepts of *domestic* and *non-domestic* map production. A domestic map is one whose geographic subject lies within the country where it was produced; a non-domestic map covers a geographic location located abroad.

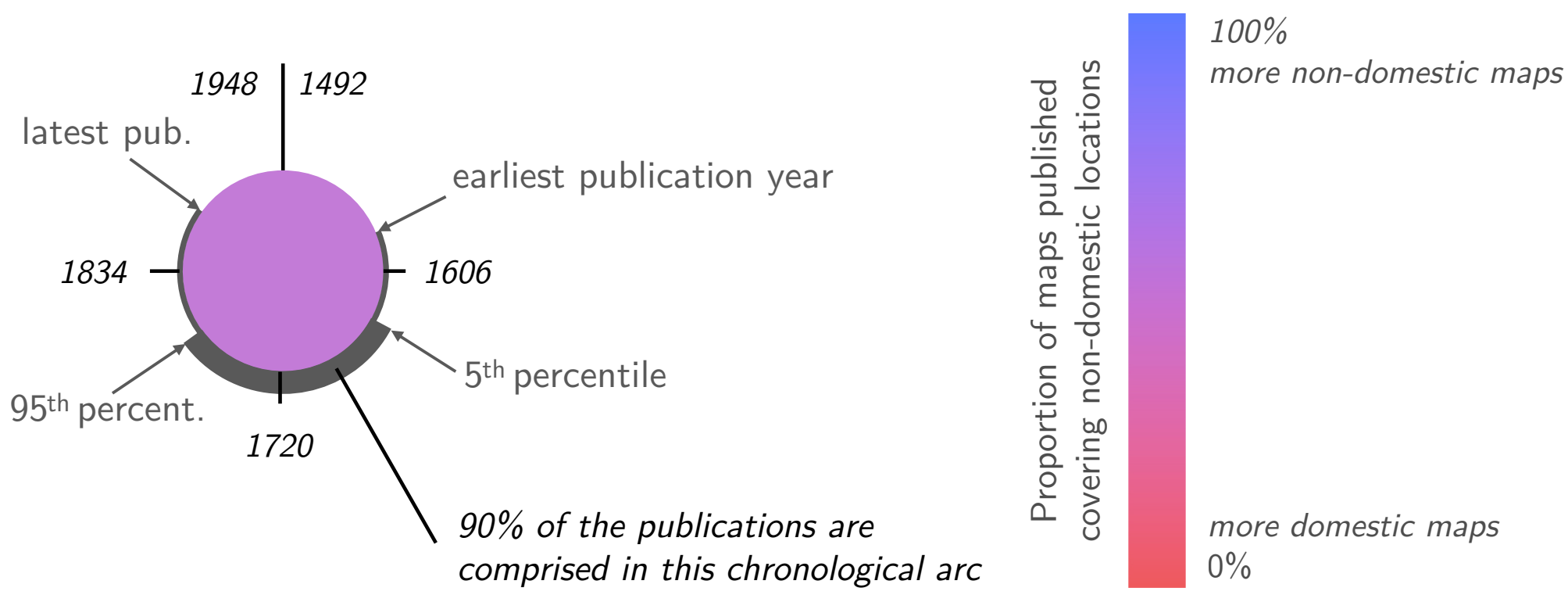

**Figure 2 | Explanation of the publication orb.** The area of the circle is proportional to the number of maps published. The surrounding chronological arcs document the temporal distribution of local production between 1492 and 1948 (to be read clockwise). The color hue indicates the proportion of maps that, for a particular publication center, depict non-domestic geographic locations (red indicates a higher share of domestic maps).

Figure 3 suggests that the United States benefited from strong and diversified cartographic publication activity, concentrated primarily in New England, the historic industrial belt of the northeastern United States (colloquially called the “Rust Belt”), and California. Several major publication centers can be identified, most notably Washington D.C. (50,532 map records), whose cartographic output was visibly driven by a demand for domestic map, related to the presence of government agencies, particularly the US Army Map Service and the Department of Agriculture.

New York (20,205 records) and Philadelphia (16,088) were also important publication centers. Their output was driven mainly by private mapping, with actors such as George W. Bromley (1857–1929), Joseph H. Colton (1800–1893), Henry S. Tanner (ca. 1786–1858), and Samuel A. Mitchell (1790–1868). Other major mapmaking cities were Chicago (11,873), Boston (4,969), and San Francisco (4,783). Whereas Boston and Philadelphia were already active publication centers in the early 19$^{th}$ century, the chronological arcs of Washington and San Francisco indicate that these cities did not begin publishing significant volumes of maps until the late 19$^{th}$ century.

An overwhelming share of Japanese map production (Fig. 4) was concentrated in Tokyo (39,416 records). Most of these maps were published in the early 20$^{th}$ century, toward the end of the Meiji era. Production there was also largely driven by national or military mapping agencies, such as the Land Surveying Department and the Rikugun (Imperial Japanese Army). Kyoto and Osaka account for only a few hundred publications, from which earlier works by the military mapmaker Hayashi Shihei (1738–1793) and by Sekisui Nagakubo (1717–1801; see Fig. A1).

American cartographic output appears polycentric. Production was driven by several dozen medium-sized centers, which together accounted for 55% of all publications. This proportion contrasts with the patterns observed in Japan, France, and Spain (Figs. 4–6), where Tokyo, Paris, and Madrid concentrated the majority of production. Paris, specifically, is the largest single publication center in the corpus, with 69,964 records.

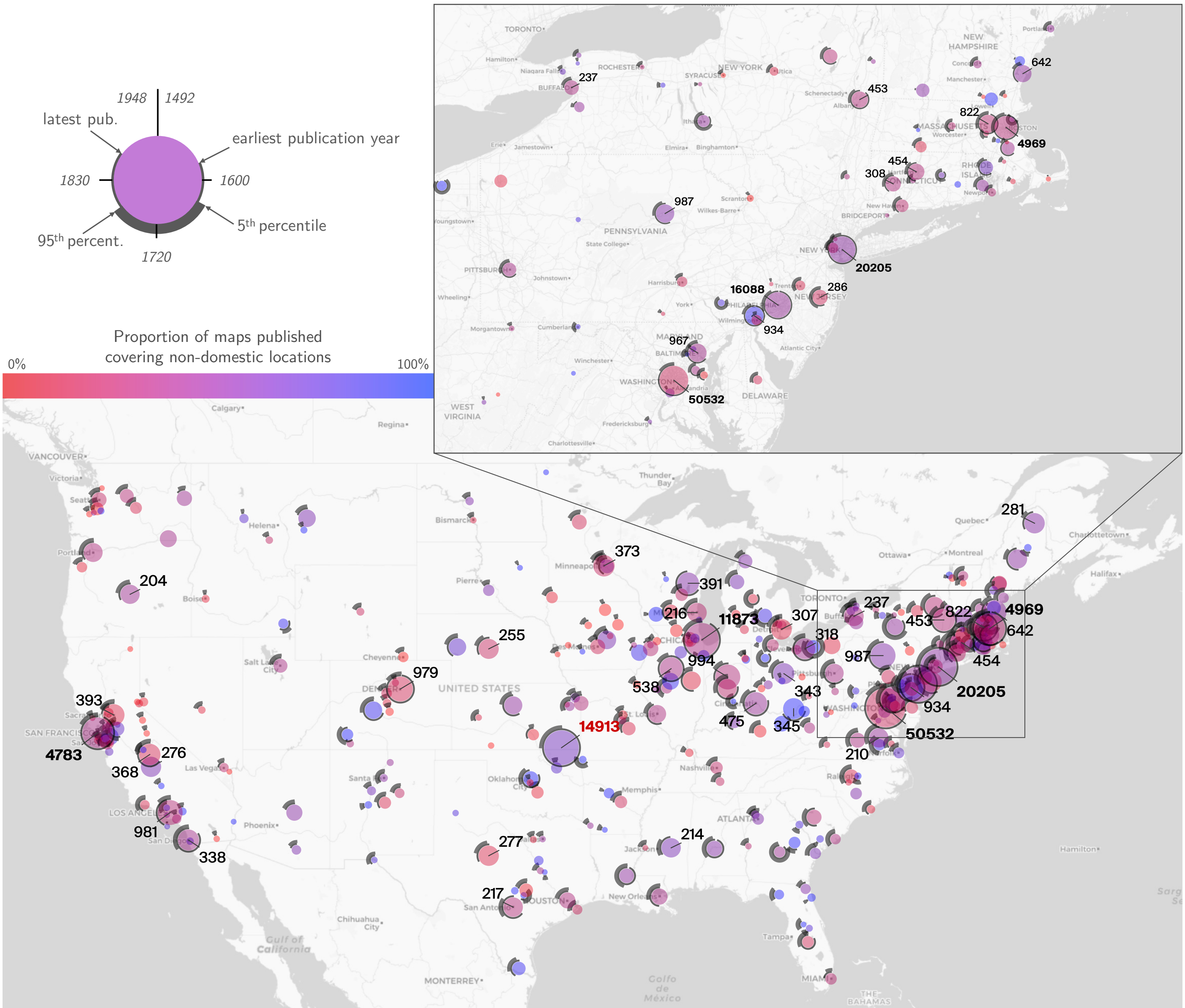

**Figure 3 | Historical map publication centers in the United States.** The key is provided in Figure 2. The number of maps published is explicitly indicated on the map when it exceeds 200 documents. Only publication centers with more than 5 published documents are represented. The count is printed in red when the geographic location is approximate (region-level geolocation). *American map publication appears broadly polycentric, with a concentration of large publication centers along the Northeast Corridor.*

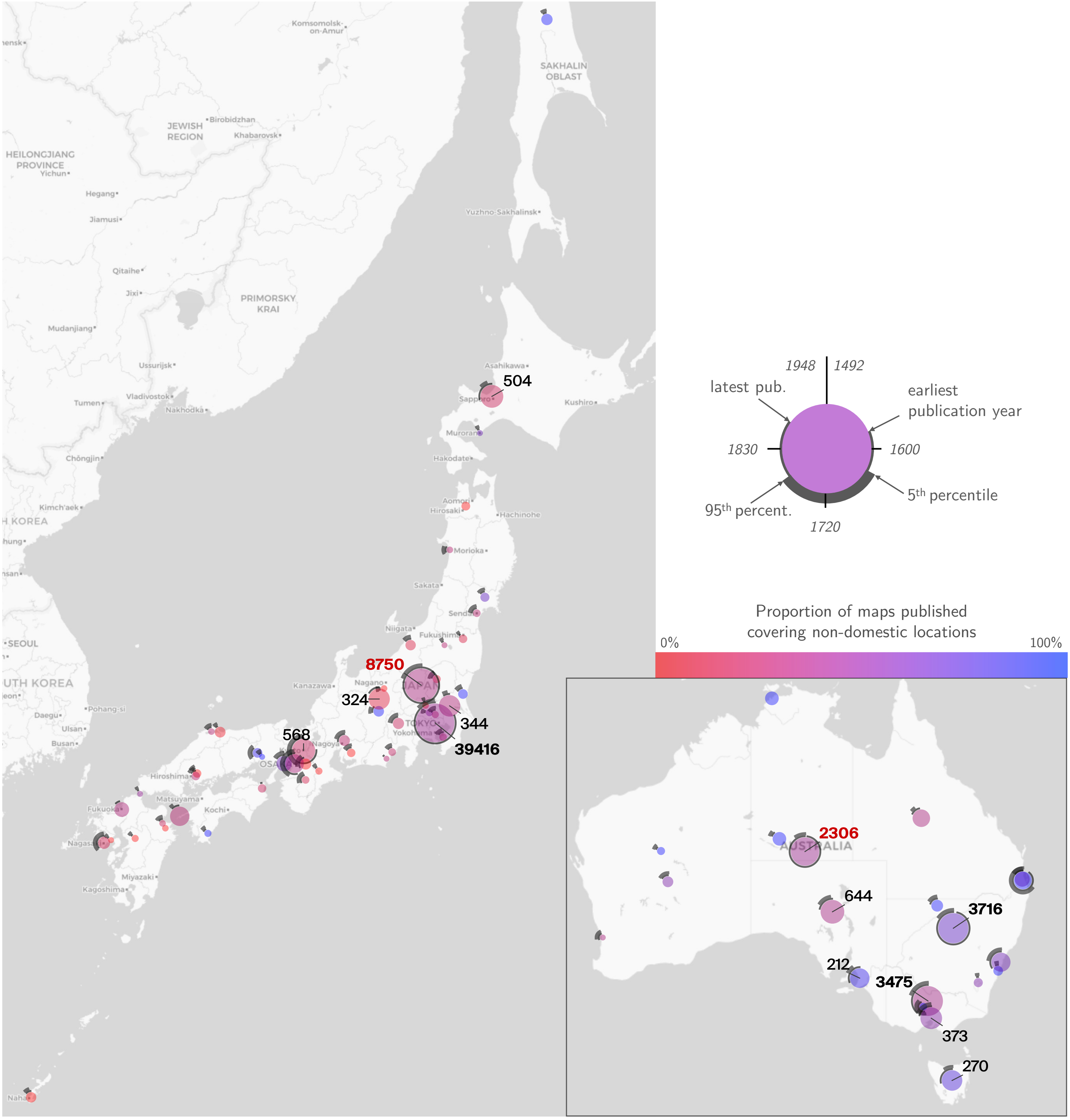

**Figure 4 | Historical map publication centers in Japan and Australia.** The area of the circles is proportional to the number of maps published. See the key in Figure 2. *Japanese publication was concentrated in Tokyo. Australian map trade originated in the southeastern urban centers.*

A polycentric publication pattern can be observed in Australia as well (Fig. 4), where the cities of Sydney (3897 records), Melbourne (3848), and, to a lesser extent, Canberra (856), along with the three corresponding States, represent a majority of publications. In Italy (Fig. 6), Rome (3954), Florence (3909), and Venice (3284) were all three significant map publication centers, with Venice being particularly active from the first half of the $16^{th}$ century. The chronological arc of Rome shows that Roman map trade started around the same period but remained significant into the $20^{th}$ century.

Another $16^{th}$-century, European map-publication center (Fig. 5) was Köln (2,289 records) with mapmakers such as Georg Braun (ca. 1541–1622) and Franz Hogenberg (ca. 1535–1590), authors of *Civitates Orbis Terrarum* (see Fig. A2), and printers like Peter von Brachel (active ca. 1575–1645). Nuremberg, Franconia, and Bavaria account for 5,626 records, beginning in the mid-$17^{th}$ century. This is where Johann B. Homann (1664–1724) and his heirs were established, for instance. In the Netherlands, at the same time, both Amsterdam (16,693) and Leiden (1,301) were major publication centers. They supported a prolific map-engraving and publishing trade, sustained by nascent capitalism and the creation of chartered companies. Emblematic Dutch creators include as Jodocus Hondius (1567–1612), Joan Blaeu (1596–1673), Gerhard Mercator (1512–1594), Frederick de Wit (1629–1706), and Nicolaes Visscher (1618–1709). In France, Lyon (612) and Orléans were significant publication centers from the beginning of the $16^{th}$ century, with printers like Jean d'Ogerolles (deceased ca. 1580). Two notable cases are Strasbourg (1,734) and Basel (1,147). Having been at the heart of numerous border conflicts between France and the German states—alternating sovereignty on several occasions in the case of Strasbourg—the two towns maintained an early and then continuous cartographic output from the beginning of the $16^{th}$ century onward. Both Strasbourg and Basel are also located in the Rhine Valley, which seems to concentrate a large part of historical map publication in Northern Europe, including centers like Amsterdam (16,693 map records), Leiden (1,301), Köln (2,289), and Frankfurt (1,887), with ramifications toward Stuttgart (1,933), Bern (5,357), and Zürich (3,318). Another area with a high output is the Venice–Milan axis. Both Venice (3,284) and Milan (1,273) constitute two of the largest European map centers. However, what defines the unique production structure of both the Rhine Valley and the Venice–Milan axis is also the concentration of multiple medium-sized and smaller centers—such as The Hague, Essen, Düsseldorf, Bonn, Darmstadt, Freiburg im Breisgau, and Aarau in the Rhine Valley, and Novara, Bergamo, Bologna, and Padua in Northern Italy.

This geography overlaps with the European Dorsal (Brunet, 1989), a corridor of urbanization stretching from the Po Plain to the West of England, via the Rhine Valley, which is considered Europe's main geo-economic structure, both demographically (over 100 million inhabitants today) and economically (Eurostat, 2025) since the $19^{th}$ century and the Industrial Revolution. As a long-term structure, the European Dorsal changes, shifting and widening over time (Faludi, 2015; Hospers, 2003). The present results suggest the existence of a prior, related structure where publishing trades were more prevalent. In this instance, a concentration of major and medium publication centers seems to have existed as early as the 17th century, with the cities of Amsterdam, Essen, Köln, Strasbourg, Basel and Venice forming the main trunk. The extent to which this "Cartographic Dorsal" may be interpreted as the manifestation of a knowledge diffusion path, or a geo-economic structure older than theorized, remains an open question. Braun and Hogenberg's chorographic atlas of 543 world cities, *Civitates Orbis Terrarum* (Fig. A2), published between 1572 and 1617, similarly focused on highly urbanized areas, localized in the Netherlands, along the Rhine Valley, and on the Swiss Plateau, aside other regions, including Andalusia, Schleswig-Holstein, and Latium (van der Krogt, 2008).

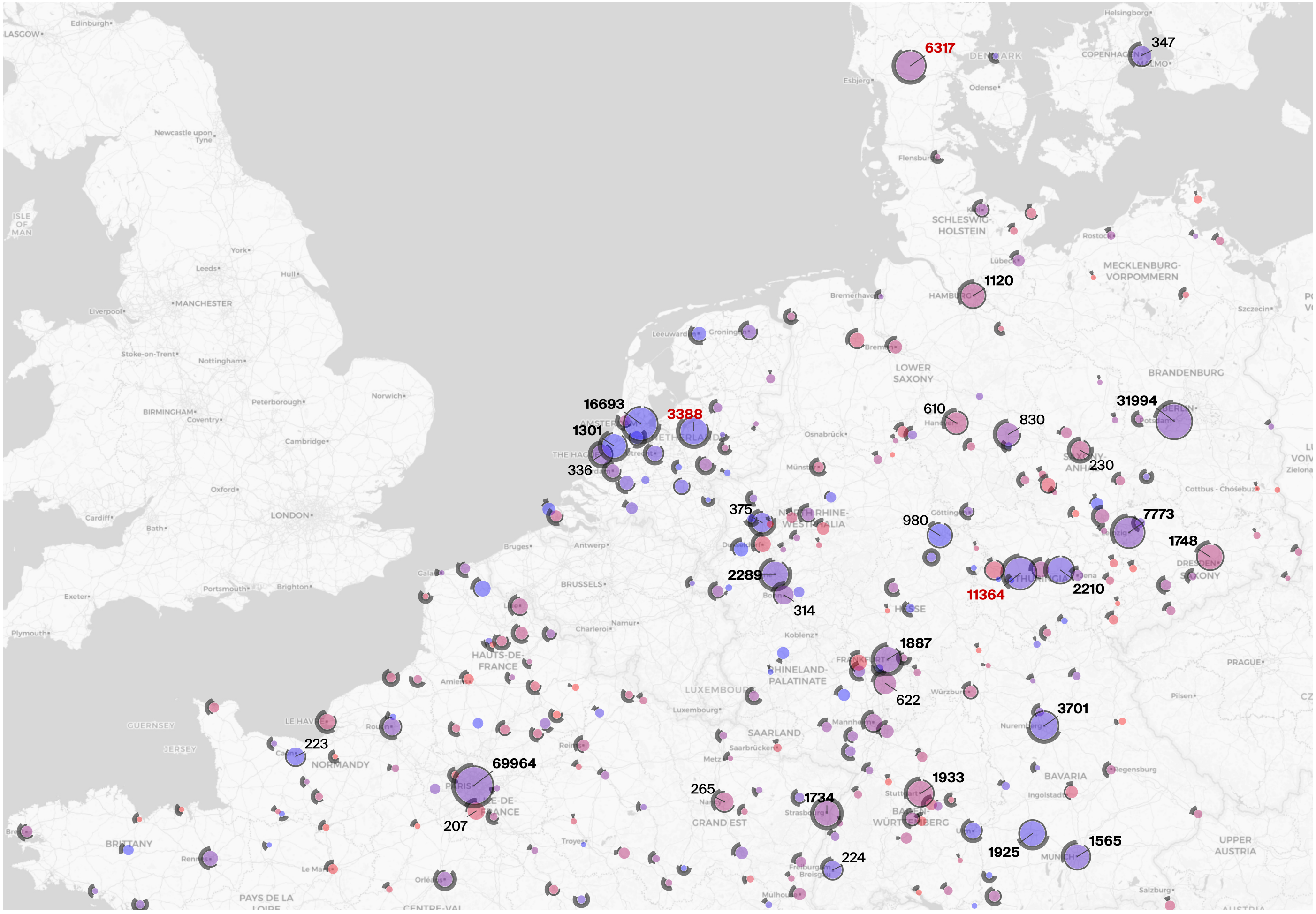

**Figure 5 | Historical map publication centers in northwestern Europe.** The area of each circle is proportional to the number of maps published. See the key in Figure 2. Only German, French, Dutch, Danish, and Swiss publication centers are depicted. *Paris emerges as the largest publication center; Dutch cartographic output is mainly foreign-focused, indicated by the blue hue of publication orbs.*

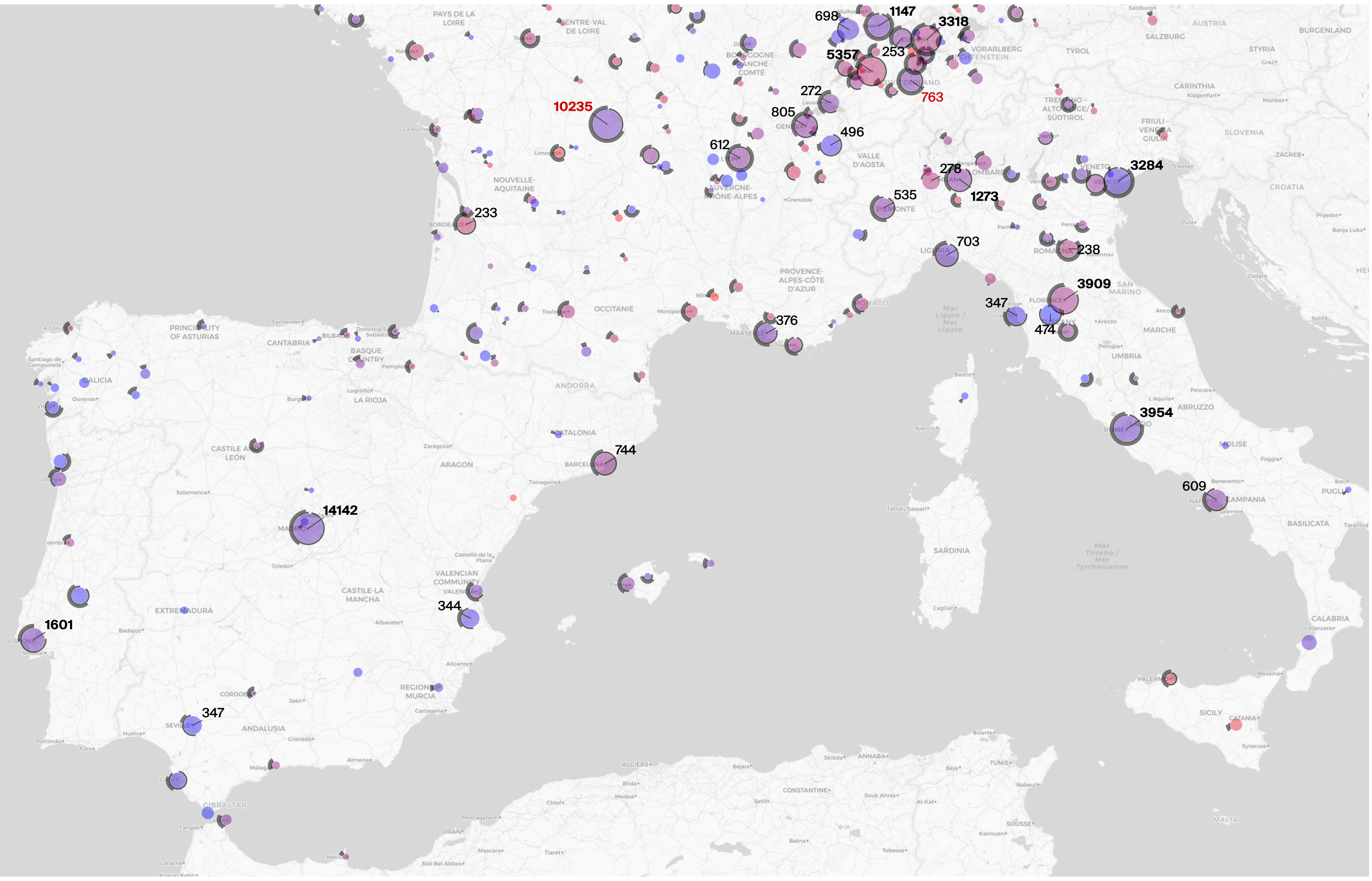

**Figure 6 | Historical map publication centers in southwestern Europe.** The area of the circles is proportional to the number of maps published. See the key in Figure 2. Only French, Italian, Spanish, Portuguese, and Swiss publication centers are depicted. *The Swiss Plateau and the Venice–Milan axis were dense publication regions.*

At any rate, the evidence unambiguously points to the essentially urban character of mapmaking. The majority of publication was concentrated in two urban corridors—the Cartographic Dorsal and the Rust Belt—and in a few capital cities including Paris, Tokyo, and Madrid. The tendency for large cities to concentrate an important share of publication reflects the fact that map making was not a craft like any other: it often required the collaboration of multiple skilled and educated artisans. Moreover, the map trade depended on a viable market, formed by a sufficiently large bourgeois population (Sutton, 2015), as well as the presence investors or patrons. Consequently, map production was generally located near centers of political power.

In this respect, I should also comment on the expanding role of several capital cities in cartographic publication from the beginning of the 20th century. At that time, Washington (50,532), Tokyo (39,416), and Berlin (31,994) ranked among the largest publication centers in the corpus. In these cities, however, publication only took off in the second half of the 19th century and became even more prominent in the first half of the 20th century. This shift was partly due to the establishment of national mapping agencies, usually linked to administrative or military authorities, within the capitals. In this context, maps contributed to the administrative institution of nation-states and to the political construction of national identity. Maps can be viewed as objects of "power–knowledge", in that they construct intersubjective "truths" as "knowledge" altering the way territories are seen or administered (Crampton, 2001). In this prospect, capital cities and cartography tended to legitimate each other: whereas capital-based agencies granted an enhanced legitimacy to cartography, maps in turn contributed to reinforce the legitimacy of capitals, through representation, thereby reproducing the relations of power within which they were created.

## 2.3 Temporal dynamics of map production

The diachronic examination of publication volumes, at the country level, informs the relationship between map publication and political dynamics. Whereas global map publication is dynamic and shifts over time, exhibiting a general upward trend, it is regionally driven by distinct and complex regional and historical dynamics.

Figure 7 depicts the chronology of publication by country[3]. Overall, the number of maps published increases consistently over time, and at a particularly fast rate beginning in the 19th century. Although some apparently aleatoric peaks appear throughout the entire period, significant conjunctural variations are also visible, mainly from the second half of the 19th century and particularly around World War I and World War II. This period will be examined more specifically as part of Chapter 3.

[3] Please keep in mind that the data refer to current national borders.

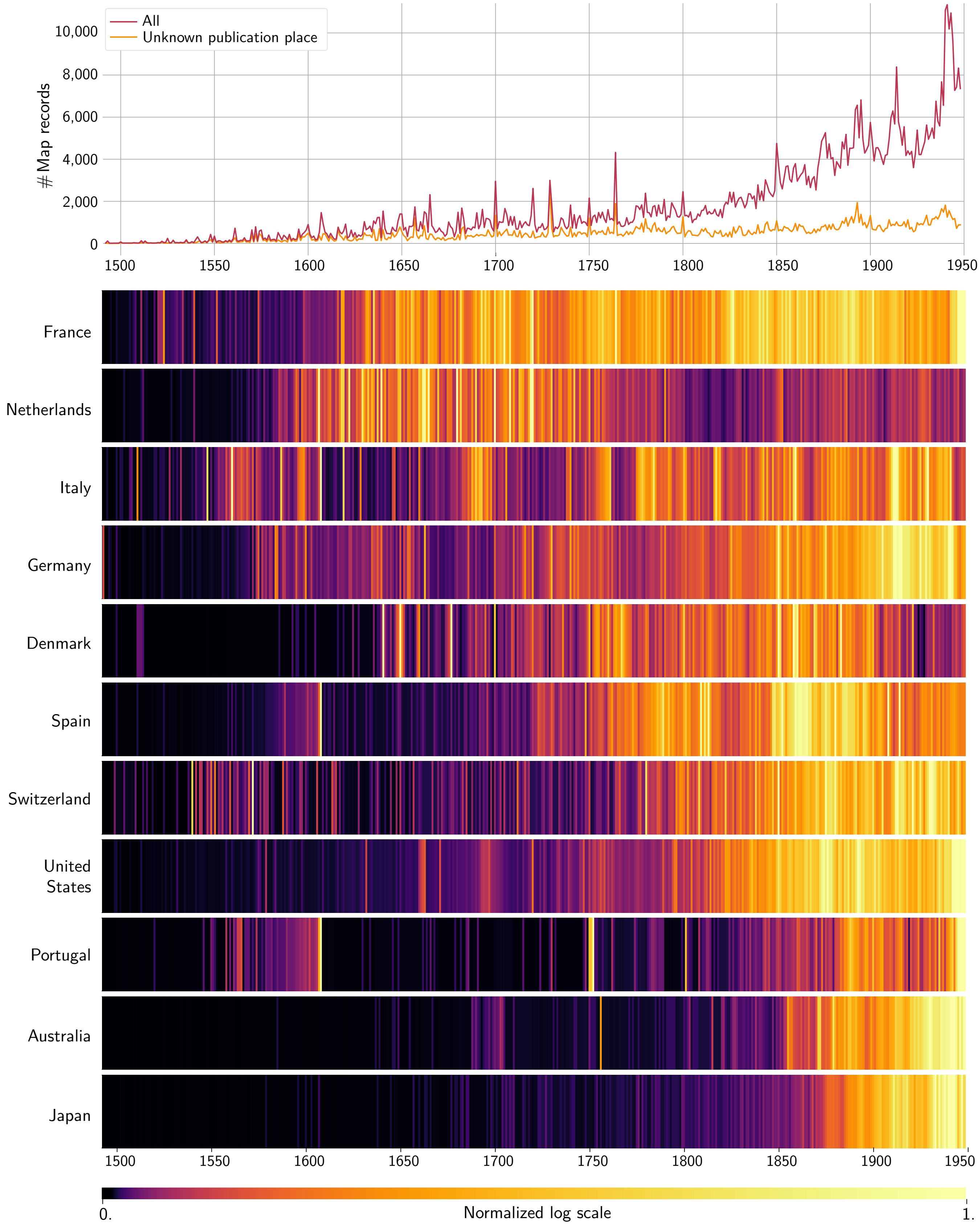

**Figure 7 | Distribution of map records by year of publication**. Top: Aggregated distribution of map records by year of publication, together with the chronological distribution of map records for which the country of publication is unknown. Bottom: Chronological distribution of map records by country of publication. Distributions are normalized for each country. *French publication activity extends primarily from the first half of the $17^{th}$ century onward.*

In the earlier period (ca. 1491–1560), one of the areas whose production is remarkably low in the data compared with the literature is the German States. With the first woodcuts, the German states were the historical cradle of map printing[4]. The data, however, exhibit a relatively low level of German production during this period. There are at least two reasons for this observation. First, German maps were predominantly engraved on wood blocks, a relatively inexpensive method. As such, they were often printed on low-quality paper, making them less likely to endure (Woodward, 1996, p. 33). Second, many of these maps were printed in books, for example atlases. At its beginnings, map-related trades were undifferentiated from book printing (Woodward, 1996, p. 38). Thus, because each atlas corresponds to only one map record, even though it might contain several map sheets, these maps are less visible in the data; only a few strips can be seen for Germany and Switzerland.

By contrast, the predominance of Italian cartographic production in the 16th century is clearly visible. Woodward, drawing on a list of 614 atlases maps compiled by Tooley, identifies Venice, and Rome, as the two main centers of map production during this period, peaking between 1555 and 1575 (Tooley, 1939; Woodward, 1996, p. 4,41). The Italian masters distinguished themselves through their mastery of intaglio, a printing technique in which ink is deposited in narrow grooves incised into a printing plate, usually made of copper. Intaglio engraving produced particularly sharp impressions, and enduring printing plates, which contributed to the wider number of reproductions, and overall availability of maps (Woodward, 1996, p. 39,49-52). This technique was generally employed by printmakers, who often relied on a rolling press that permitted the reproduction of single, large, map sheets. Woodward also describes the marked decline of Italian cartographic output caused by the plague of 1575–1577. After the epidemic, Italian engravers were outcompeted by their Dutch counterparts (Woodward, 1996, p. 5), as is also clearly visible in Figure 7.

Indeed, in the first half of the 17th century, the Netherlands, and more specifically the city of Amsterdam, became the main center of map production. There, cartography developed alongside early capitalism, colonies, and overseas trade (Sutton, 2015). Two chartered companies contributed more significantly to the expansion of cartography in the Netherlands: The West India Company (WIC, 1621–1792) and the East India Company (VOC, 1602–1799). The publication volumes calculated from ADHOC and presented in Figure 7 suggest that the Dutch cartographic Golden Age lasted from ca. 1585 to ca. 1755, which closely reflects the chronological extent of the two chartered companies. For the WIC and the VOC, cartography served three main functions: (1) legitimize overseas territorial claims, (2) strengthen the command of the Dutch navy against rival Spain, and (3) create an enticing image of the Dutch colonies that would stimulate capital investments. This third point was most clearly reflected in the depiction of regular grid plan cities, which conveyed the idea of organized, civilized, and reliable implantation projects (Sutton, 2015, p. 5). In turn, these maps reinforced the functional ideal of grid plans (Nijenhuis, 2011).

[4] Historiography generally refers to the world map in the encyclopedic incunabulum *Etymologiæ* by Isidore of Seville (560-636 BC), printed in Augsburg in 1472, as the first printed map (Norman, n.d., a).

France emerged as another cartographic superpower in the 17th century. At that time, the patronage of the reigning Bourbon dynasty (1589–1792) fostered the development of map making. Numerous cartographers, including Nicolas Sanson (1600–1667), Nicolas De Fer (1646–1720), Jean-Baptiste Nolin (1657–1708) and Alexis-Hubert Jaillot (1632–1712), were linked to the French royal power (Petto, 2007, pp. 21–35). Foreigners like Vincenzo Coronelli (1650–1718) and Jean-Dominique Cassini (1625–1712) also contributed to the development of French cartography. Johannes Blaeu (1596–1673) offers an eloquent example of the efforts deployed by certain map makers to attract the patronage of the French monarch through lavishly flattering dedications[5]. In France, at that time, cartography was regarded as an issue of scientific and technical supremacy. Mapping the provinces of France was, for instance, one of the first missions assigned to the Académie des Sciences, established in 1666. The French navy also used to commission maps of foreign territories in order to safeguard the colonial interests of the French Crown in the Americas and to prepare for potential conflicts on the European continent, and with England (Petto, 2007, p. 57,72).

One of the most significant patterns visible in Figure 7 is the association between strong, centralized, and sovereign regimes and publication activity. This is evident in the case of France, where production expanded under the absolute monarchy of Louis XIV (1638–1715), but also in Spain, for instance, with the advent of absolute monarchy in 1700. Conversely, the end of the Danish absolute monarchy in 1901 coincided with a decline in cartographic activity, a trend that apparently continued into World War I—during which Denmark remained neutral—and afterward. By the same token, autonomy and sovereignty also appear to have influenced cartographic publication. American publication output increased at the end of the 18th century, following independence, and Australian output grew after the acquisition of legislative autonomy in 1850. In Japan, it was during the Meiji era (1868–1912), which saw Japan strengthen its technological position against the West and pursue its own expansion policy in Eastern Asia and the Pacific, that cartographic production flourished. By contrast, historically fragmented regions, such as the

[5] The French edition of Blaeu's cosmography, for instance, starts with the following dedication: *To the most Christian King Louis XIV, Sire, I take the boldness to offer to Your Majesty, admired by the whole universe, not the feigned Atlas of the poets, which, according to them, supports the weight of it on the shoulders; but, under a famous name, the maps and descriptions of all the provinces of the world. I believed that such a considerable work, which together with the History makes the most beautiful book of kings, now that my care and labor have increased and enriched it with all the ornaments it could expect from printing, would not be altogether unworthy, when moreover put into French, to appear in the eyes of the most powerful monarch of Europe.* Read the original excerpt in the Appendix, Text A1.

German States and Italy, seem to have developed a stable, regular, and comparatively strong publication output only at a later stage, relative to their European neighbors. These distinct macroscopic trends support the view that cartography was generally produced in contexts where economic, political, military, or administrative power was concentrated.

## 2.4 Actors of map production

Up to now, the social dimension of map making was not directly discussed. However, social transmission is more directly related with the way cultural features are replicated and diffused. The analysis of chronologies also suggested that individuals and collective entities may have played a key role in macroscopic dynamics. This section will specifically examine map publication from the angle of work collaborations.

One challenge of studying maps as editorial objects is that the notion of authorship is not without ambiguities. Not only is this information frequently missing from catalog records, but mapmaking also involves many trades: surveying, geometric calculations, engraving, typography, printing, and coloring. These tasks are often carried out by distinct individuals, whose individual roles and contributions are not always described in bibliographic notices, and rarely in a standardized manner. For instance, creators are often not explicitly distinguished from patrons or publishers. Consequently, the respective roles of each contributor can hardly be studied quantitatively. The collaborative nature of mapmaking, however, can serve an alternative perspective that is particularly relevant to the study of cultural circulations: that of the network or, more specifically, a social graph.

### *Graph construction*

Reconstituting the social graph of work collaborations across map makers involves building a network in which each creator (map makers or collective entity) is a node. Two creators are connected by an edge if they have published at least one map together. The edge weight is proportional to the number of maps on which they collaborated. Figures 8, 9, 11, and 12 visualize the social graph of creators from three distinct perspectives. First, from a geographic perspective (Figs. 8 and 9), by considering the relationship between the structure of the social graph and geographic divisions: countries, and urban centers. Second, from a chronological perspective (Fig. 11), by considering the period of activity of each creator. Third, from the perspective of creator typologies, e.g. military, scientific, or administrative entities. The latter perspective, based on the semantic analysis of creator names, adopts an analytical viewpoint popular in the field of the history of cartography, which consists in studying corpora defined by the communities that produced them, for instance learned societies, or national cartographic agencies (Broc, 1974; Herbert, 2018; Monmonier, 1981). When discussing power, relationships are also commonly employed to highlight and qualitatively discuss the role of certain rulers as patrons (Petto, 2007; Rees, 1980).

### *Semantic encoding creator typologies*

To provide breadth to the analysis, linguistic markers are used to inform creator typologies. Indeed, many entries, such as ‘D. Appleton & company’ or ‘Oberleut Woerner,’ include linguistic cues that can identify the community to which a particular creator belongs. In this example, the tag ‘& company’ indicates a private company, whereas the title ‘Oberleut’ suggests a military officer. A baseline approach to this problem would involve compiling extensive vocabularies and categorizing entries through pattern matching. However, it appears very long and difficult to list every possible linguistic marker and its variants without ending up classifying all entries manually. This task would be even more challenging in the present multilingual context. Moreover, linguistic indices can be subtle. For instance, a phrase like ‘Mussard et Lavalier’ may already hint at private mapmaking, even though the French conjunction ‘et’ is not, alone, a reliable marker of that semantic domain.

Thus, the present work opts for a more indirect approach to the problem, based on latent semantic spaces. Several studies have demonstrated the pertinence of using the latent spaces of large foundational language models as semantic spaces (Gurnee & Tegmark, 2024; Hewitt & Manning, 2019; Roberts et al., 2023). Here, regions of the semantic space, called *semantic domains*, are used to differentiate typologies of creators, e.g. scientific, touristic, or nobility-related entities. Each domain is delineated by a set of pointer-keywords. Let us name $P_i = \left\{\overrightarrow{p_i^1}, \overrightarrow{p_i^2}, \ldots, \overrightarrow{p_i^N}\right\} \in \Phi_i$ the set of pointer-keywords defining the semantic domain $\Phi_i \in \mathbb{R}^{1536}$. We state that $\vec{e}$ is included in $\Phi_i$ if and only if $s_i^{\vec{e}} > t,\ t = 2.02$, where $s_i^{\vec{e}}$ is the *salient similarity score* between $\vec{e}$ and $P_i$, and $t$ is a *saliency threshold* defined empirically. This salient similarity score is obtained by computing the smallest cosine distance between $\vec{e}$ and the set of pointers $P_i$:

$$\sigma_i^e = \min_{\forall n}\left(1 - \frac{\langle e, p_i^n \rangle}{\|e\|\|p_i^n\|}\right)$$

$$s_i^e = \frac{\sigma_i^e}{\sum_{\forall i} \sigma_i^e + \varepsilon}, \qquad \varepsilon = 10^{-2}$$

The salient similarity score indicates whether the embedding $\vec{e}$, corresponding to a specific creator, is close to a particular semantic domain. As such, the strategy employed to assess specificity is similar to Lowe’s ratio (Lowe, 1999). Specifically, we measure the fit between a source name—represented by $\vec{e}$—and a target semantic domain, based on the saliency of this domain *compared* to other domains. This is a way to ensure the quality of the match and avoid classifying ambiguous terms.

The complete list of pointer-keywords used to delineate semantic domains is reported in the Appendix (Tab. A2). For example, keywords like 'defense', 'admiral' and 'armed forces' point to the *military* domain, while 'municipality', 'commission' and 'department' point to the *administrative* domain.

Eight domains were defined in total: (1) military, (2) land planning, agriculture, construction (3) royalty and nobility, (4) administration and bureaucracy, (5) science and learned societies, (6) travel, transportation, and touring, (7) private companies (including charter, insurance, electricity, and gas), (8) engraving, printing, and edition. Each domain corresponds to a typology of creator. Is it possible for an entity to belong to more than one domain, provided that both are sufficiently salient compared to other domains. For instance, the *Ministère de la Reconstruction et de l'Urbanisme* (Ministry of Reconstruction and Urban Planning) is assigned to both (2) and (4). A sample of results is provided in the Appendix (Tab. A3).

### *Measuring determinants of the structure of the social graph*

Another analytical approach relevant to the study of the social network of map makers is to determine which variables best explain the way in which creators are grouped into communities (or clusters). The sociological concept of *homophily* predicts that actors who share common characteristics (e.g., a *habitus*) have a greater interaction likelihood (McPherson et al., 2001). Here, belonging to the same domain (e.g., scientific) could be considered such a characteristic (a socio-professional proxy). The influence of location and period of activity—which, for individuals, can serve as a proxy for generation—can also be considered. The potential of a particular variable to explain the observed structure of the social graph is quantified by the *modularity* coefficient. The modularity coefficient evaluates the quality of a graph partition into communities by comparing, on one hand, the average number and weight of edges within those communities with the expected occurrence of edges if links were randomly set, on the other hand. If we denote $m$ the total number of edges, $k_i$ the degree of entity $i$, and if the function $c(i)$ returns the community to which entity $i$ belongs, the modularity $Q$ is given by the following equation:

$$Q = \frac{1}{2m} \sum_{\forall i,j} \left( A_{ij} - \frac{k_i k_j}{2m} \right) \mathbf{1}_{\{c(i)=c(j)\}}$$

Here, $A_{ij}$ denotes the adjacency matrix, which determines the weight of all edges, and $\mathbf{1}_{\{c(i)=c(j)\}}$ is a function that returns 1 when $i$ and $j$ belong to the same community and 0 otherwise. A graph partition is generally considered significant if $Q > 0.3$. Higher values indicate greater descriptive power.

Figures 8, 9, 11, and 12 display, as nodes, the 236,013 unique creators extracted from the corpus and normalized according to the methodology described in Chapter 1. Each of the 1,590,093 edges represents a collaboration between two creators. The edges are weighted, and, although edge widths

are not quite visible due to image size, they influence the relative positions of the nodes. The relative location of the nodes was computed using ForceAtlas2, a spatialization algorithm which clusters highly connected nodes near the center, while independent entities appear in the two elongated lobes at the periphery of the main circle (Jacomy et al., 2014). The size of each node reflects the number of maps published by the corresponding creator.

### *Multiscale geographic structures*

Figure 8 displays the principal country of production, for each map creator. The main countries (see also Fig. 1) are color-coded. Countries correspond to clearly distinct regions of the graph, a finding confirmed by a high modularity ($Q = 0.595$). These results partly justify the relevance of considering contemporary national borders an approximation of historical political divisions[6]. Figure 8 presents a central cluster, corresponding to the Western group, with Japan on the periphery. The Western group is further divided into two lobes. The United States and Great Britain are strongly integrated and constitute the upper lobe. Australia, which is not colored because of its comparatively smaller size, also lies within this lobe. The lower Western lobe is further subdivided: France, Italy, the Netherlands, Spain, and Portugal (not colored) form the first subcluster, whereas Germany and Switzerland form the second. The relative position of countries is significant in this visualization, thanks to the spatialization algorithm. Hence, spatial relations—like when the United Kingdom is surrounded by France and the United States—are indicative of the country's relative socio-cultural position. Let us also note the proximity of Italy and the Netherlands. Figure 11, which will be discussed later, suggests that this closeness may be related to the historical role of the two countries as crucibles of mapmaking. German map makers also seem to be present in that region of the social graph, which is coherent with the cradle interpretation. Finally, the position of Switzerland within Figure 8, near the center of the graph, may be interpreted as the reflection of the relative intermediacy of Swiss cartography, from a geographic and cultural point of view, compared to its European neighbors—France, Germany, and Italy—although it appears more closely related to the German States.

[6] While current borders do not precisely reflect historical political and administrative divisions, they remain related to earlier cultural, geographic, and political boundaries from which they descent.

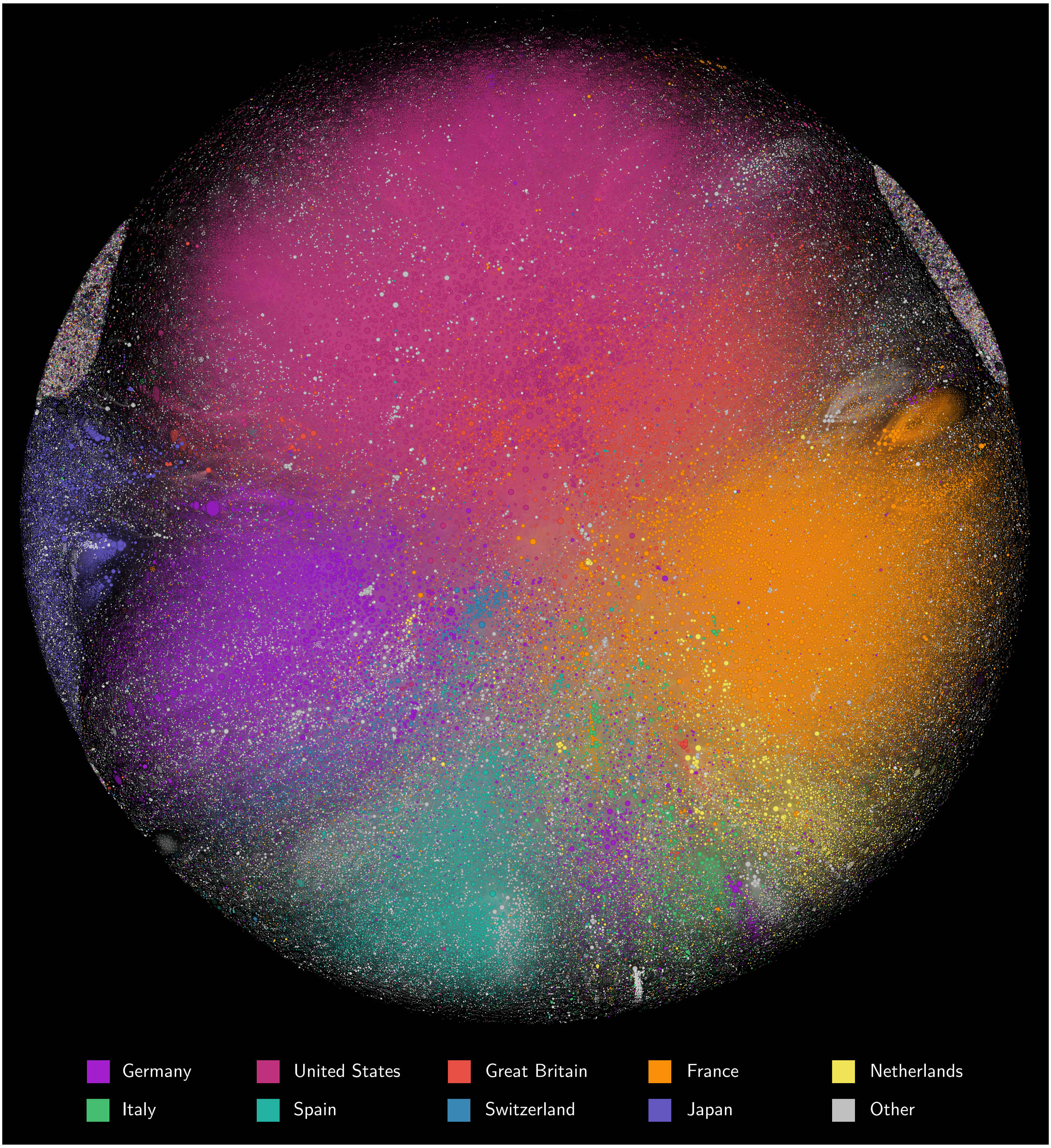

**Figure 8 | Network of map creators by country of publication.** Each node corresponds to a mapmaker, a mapping agency, or a company. Edges indicate collaboration between two creators in the publication of a map. The size of each node is logarithmically proportional to the number of maps published, and the width of each edge is proportional to the number of collaborations. For each node, color indicates the most frequent country of publication for each node. *Map makers are grouped into distinct social clusters by country.*

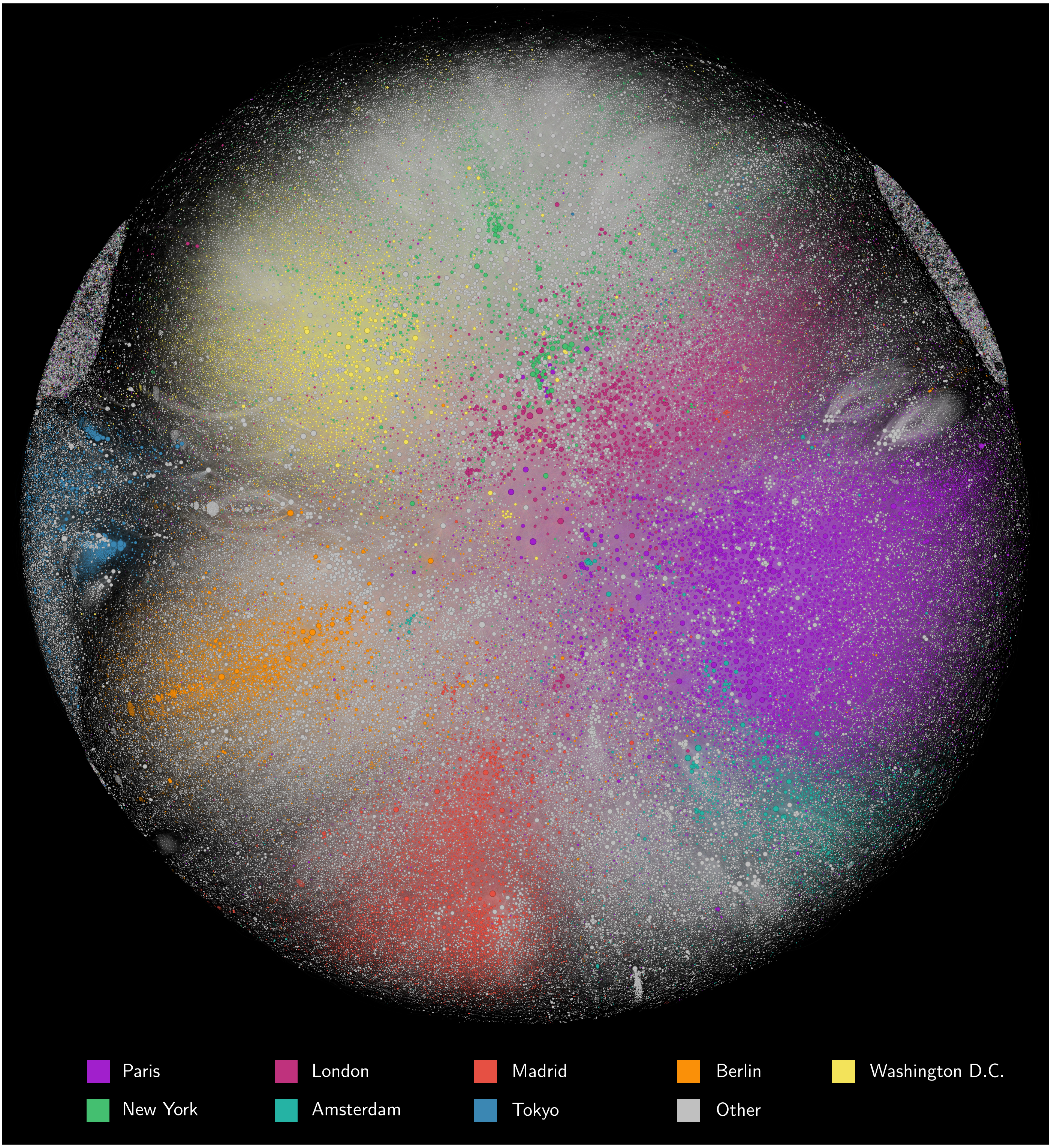

**Figure 9 | Network of map creators by city of publication.** Each node corresponds to a mapmaker, a mapping agency, or a company. Edges indicate collaboration between two creators in the publication of a map. The size of each node is logarithmically proportional to the number of maps published, and the width of each edge is proportional to the number of collaborations. For each node, color indicates the most frequent city of publication. Only the 8 most frequent cities are colored. *Map makers are grouped into distinct social clusters by city.*

As the modularity analysis shows, national subdivisions are the strongest explanatory variable of the graph structure. However, the impact of space is not restricted to these political boundaries. Instead, Figure 9 suggests that it is hierarchical. The figure highlights the largest[7] map publication centers: Paris, London, Madrid, Berlin, Washington D.C., New York, Amsterdam, and Tokyo. Modularity—computed not only on the main publication centers but on all the centers identified—is again high ($Q = 0.563$). Yet, one might have expected that being based in the same publication center would have a greater effect on the probability of collaboration, compared to the country. The result highlights the significance of national networks. Beyond modularity, the relative occupation of the social graph is also an indicator of geographic publication structures described earlier in this chapter. For instance, Parisian creators accounts for the large majority of French cluster, illustrating the centralization of the French model. A similar structure appears for London, Amsterdam, and Madrid, as well as Tokyo, to a certain extent. In contrast, production in the United States is more polycentric, with the large hubs of New York and Washington D.C., but also smaller centers such as Philadelphia, Chicago, Boston, and San Francisco (not color-coded). The German model likewise shows that, while Berlin is central, cartography is distributed across multiple centers (such as Leipzig and Nuremberg).

### *Social transmission*

The temporal dimension is a structuring feature of the social graph. This variable is represented in Figure 10, where the study period 1492–1948 is divided into widening temporal bins. Specifically, the plot evaluates the modularity $Q$ as a function of the partition of the graph into distinct temporal bins. Here, significant modularity indexes ($Q > 0.3$) indicate that the probability of collaboration is greater if two creators were active within the same time frame, corresponding to a fixed time stratum. For instance, Figure 10 shows that modularity is significant for any temporality between a decade and a century. The peak of modularity ($Q > 0.4$) is observed between 25 and 76 years, reaching its maximum for a timespan of 51 years. This result can be interpreted as an indicator of the duration of a generation of map makers: if two creators were active within the same period of about 50 years, the probability of observed collaboration increases. The significance of longer time periods, up to a century, reflects the presence in the social graph of legal entities, such as cartographic and military agencies, companies and administrative entities, which can have longer “lifespans” than individuals.

[7] Here, the ranking is based not on the number of maps published but on the number of authors active in each center.

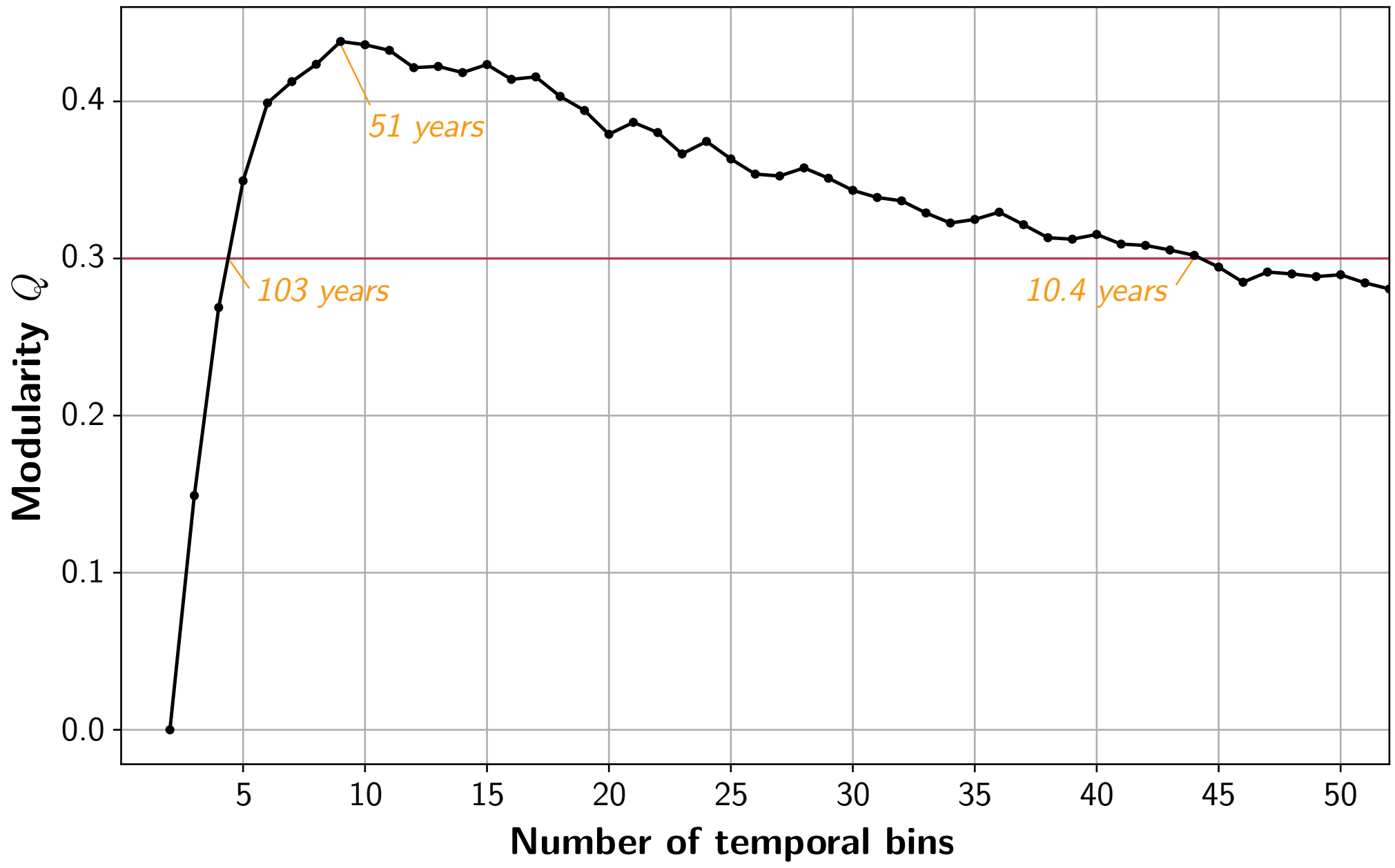

**Figure 10 | Influence of time granularity on modularity.** The plot shows how modularity $Q$ changes as the graph covering 1492–1948 is partitioned into fixed temporal bins. The red horizontal bar at $Q = 0.3$ marks the threshold of significance. Labels indicate the boundaries of significance and peak value. *Map makers active within the same period of ca. 50 years have an enhanced probability of collaboration.*

Figure 11 presents the social graph from a chronological perspective, treating time as a continuous variable. The most conspicuous phenomenon is the emergence of a diffusion pattern, with cartography originating in the bottom-left area and progressively expanding across the graph. Transmission between people and institutions indeed constitute the primary vector for the spread of innovations (Rogers, 1995). Cartography, however, does not consist of a single innovation but of a set of parallel and related practices and techniques. Figure 11 thus shows the spread of mapping practices from an individual to another through the network of collaborations, and other non-accounted for transmission channels. Propagation seems to follow a geographic–political coherence. The historical crucible of cartography, mainly formed by Italian and Dutch map makers, is clearly visible at the bottom right, followed by the propagation of map making trades to Paris and London, located on the right hemisphere of the visualization. From there, cartography seems to have rapidly spread to throughout the West, while Japan, on the left, followed only later.

By showing contrasts, Figure 11 also highlights forerunners, or early adopters, corresponding to lighter–colored nodes. Early adopters are individuals who adopted mapping technologies and practices ahead of their network (Rogers, 1995). As such, they helped spreading cartography to new regions and social circles. Seven such early adopters are highlighted in Figure 11. The next paragraphs will investigate their individual career paths, qualitatively and one-by-one, and attempt to understand what characterizes them.

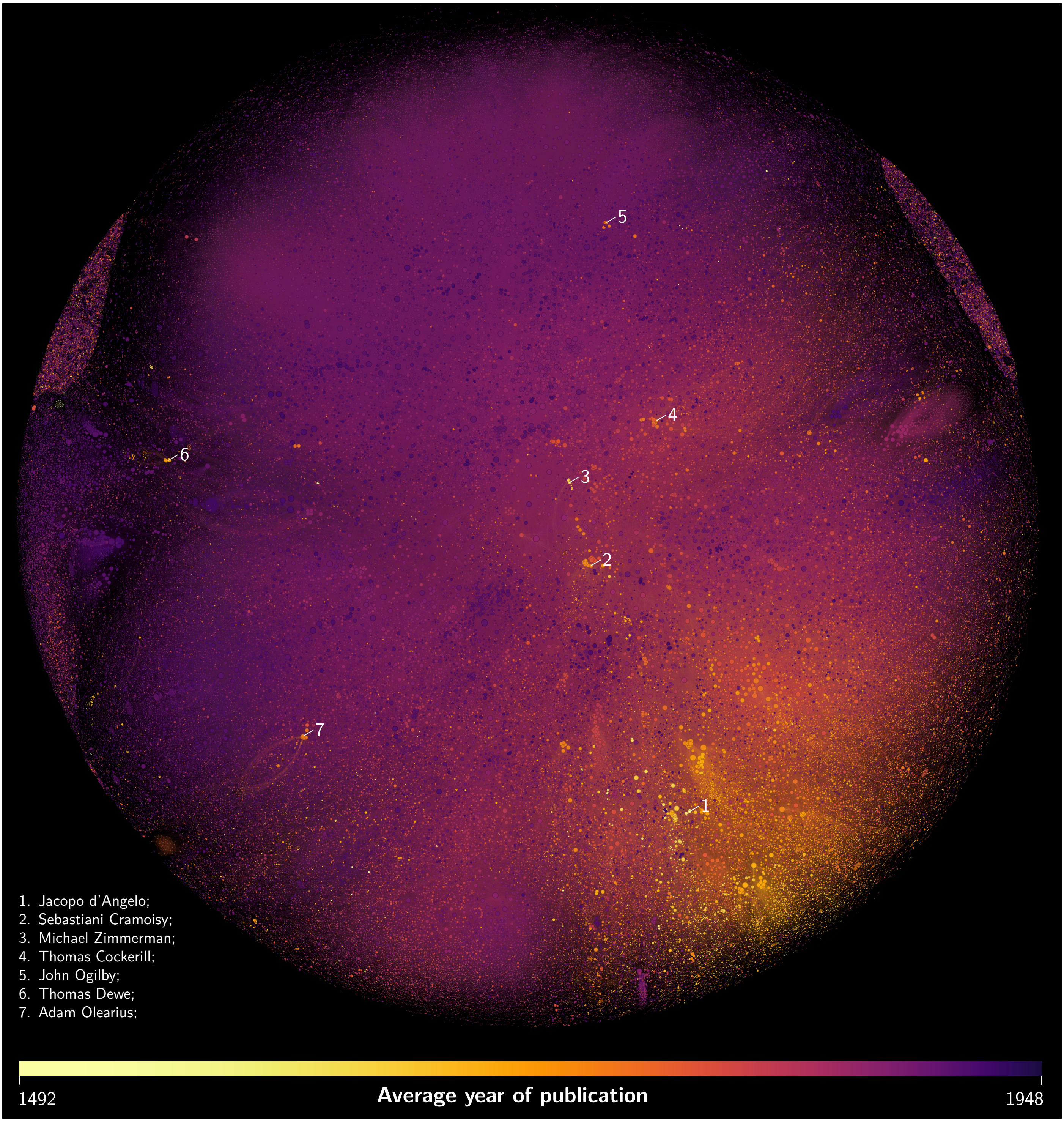

**Figure 11 | Network of map creators by average year of publication.** Each node corresponds to a mapmaker, a mapping agency, or a company. Edges indicate collaboration between two creators in the publication of a map. The size of each node is logarithmically proportional to the number of maps published, and the width of each edge is proportional to the number of collaborations. For each node, color denotes the average year of publication activity. *The gradual transmission of cartographic processes and techniques is visible, starting in the lower-right of the social graph.*

The first, Jacopo d'Angelo (ca. 1360–1411), actually predates the period under study. His presence in the corpus therefore corresponds to later reprints or republications of his work. An archetypal Florentine scholar of the Renaissance, Jacopo d'Angelo is best known for his translation of Claudius Ptolemy's Geography (ca. 100–168). Although his translation was later criticized (Dalché, 2007, p. 341) his prominence in the graph attests to the enduring influence of the Latin and Greek heritage on the first printed maps and cartographic incunabula.

The second early adopter is Sébastien Cramoisy (1584–1669). Cramoisy was one of five French booksellers and printsellers who, between 1633 and 1635, through Christophe Tassin (deceased 1660), engineer and geographer to the king, obtained the privilege of publishing a series of seven atlases of France. These atlases enjoyed notable public success, and are regarded as the first instances of independent printed cartography in France (Hofmann, 2007, p. 1580). Between 1648 and 1649, Sébastien and Gabriel Cramoisy also participated in the publication of a world atlas by the Jesuit monk Philippe Briet (1601–1668). During this period, the Jesuits contributed significantly to the popularization of cartography, using maps such as Briet's atlas as teaching material (Febvre, 1950). This, in turn, contributed to the emergence of a market for maps (Hofmann, 2007, p. 1579).

The third observed early adopter is Michael Zimmerman (deceased 1565), a Viennese printer associated with Sigmund Freiherr von Herberstein (1486–1566). Von Herberstein was Austrian ambassador to Moscow. He compiled a map of the Russian principalities, in 1526, and a chorography of Muscovy, in 1546. These publications had a lasting impact; as late as 1706, Guillaume Delisle (1675–1726) used the 1526 map as the basis for his own cartography of Russia (Goldenberg, 2007, p. 1857).

Thomas Cockerill (active 1674–1702) was the co-producer of a large part of Robert Morden's work (ca. 1650–1703). Their collaboration yielded a curated atlas of English counties, republished maps of Europe and the world, and original maps of the British colonies in the Americas and of Ceylon. Morden's business strategy has been noted for his outstanding abilities to market maps as authoritative goods and for occupying a specific market segment, that of atlases (York, 2013).

John Ogilby (1600–1676) was the first royal cosmographer of the British monarchy. Prior to this appointment, Ogilby had been a dancer and poet. His designation by Charles II marked a turning point in the king's desire to take control of map production, while his predecessors had foremost remained consumers (Kagan & Schmidt, 2007, p. 668). He published atlases of the Americas and Asia—largely copied from Dutch sources—yet failed in his objective to produce a road atlas of England and Wales.

Thomas Dewe (active 1613–1622) was the printer of Poly-Olbion (Fig. A3, in Appendix), which David Rumsey considers the first work of pictorial maps (Rumsey, n.d.). This chorograph, created by Michael Drayton (1563-1631) illustrates a poem about Britain, its rivers, mountains and forests.

The last early adopter highlighted in Figure 11 is Adam Olearius (1599-1671). Olearius was a diplomat from the Duchy of Holstein (1474–1864, Northern Germany) who was sent to Russia and Persia. He published extensive observations on his voyages, including a cartography of the Volga, and a chorography of the cities he visited (Brancaforte, 2004).

These seven early adopters were selected among others in Figure 11, based on visual salience. They were often related to small communities, which again reflects the collaborative nature of mapping practices. It appears from the results that early adopters are not necessarily the most influential or well-known figures in the network, nor were they the most prolific producers of their time. They were not recognized as the best cartographers of their generations, and they were not necessarily original *innovators*. For instance, Jacopo d'Angelo's translations were considered mediocre by his successors. John Ogilby's atlases of Asia and America were widely plagiarized, and he failed to complete his atlas of English itineraries. The position of these early adopters within the network, which places them upstream of the general pattern of adoption, merely suggests the comparatively long temporal reach of their work.

Specific features common to several early adopters can be outlined. Many pertain primarily to the narrative impact of their cartographic work, that is, how compelling their message was. The first recurring feature, as such, is the adoption of a legitimacy stance. Demonstrating one's legitimacy is key to successful map marketing and an integral dimension of narrative construction (Sponberg Pedley, 2005a). Appointment to a diplomatic or scientific advising position is one of the conditions that may help legitimize a map maker, through the authority derived from the State or institutions. Another strategy is the mobilization of knowledge ideals. Morden, for instance, built his image of legitimacy on a rationalist ideal of cartography. Conversely, by focusing on the re-edition of Ptolemy's work, d'Angelo drew on the classical ideal of the Renaissance.

The second feature common to several early adopters is the ability to draw on the imaginary. Compelling narratives can be based on the description of places considered exotic or mysterious. Mythologic references and artistic quality may also contribute to the construction of appealing narratives. An archetypal example is Drayton's Poly-Olbion, which is characterized as much by its pictorial originality as it is by the mythological imagination it conjured. Other cartographers, like d'Angelo, Briet, von Herberstein, Morden, and Olearius, are notable for having mapped remote territories, which might have appeared exotic and therefore appealing to their audience.

The third feature is the reliance on dependable distribution networks. While the circulation of Briet's maps was facilitated by a network of Jesuit monks, other cartographers, including Cramoisy, Drayton, and Ogilby, may have benefited their local focus to reach an audience.

Finally, the last feature can simply rather be described as chance, or opportunity. For example, von Herberstein, Morden, and Olearius seized the opportunity afforded by travel. Cramoisy, Tassin, and Ogilby, as well as von Herberstein and Olearius in their diplomatic capacity, benefited from privileges granted by the sovereign. This apparently enabled them to undertake cartographic projects that would otherwise have been out of reach and thus set them apart.

### *Mixed communities*

The last approach to the analysis of the social graph focuses on semantic domains. Figure 12 shows the attribution of nodes to one of the eight semantic domains. When a creator is matched to more than one domain, only the most salient is shown. Note that many nodes were not attributed, since a threshold was imposed on the salient similarity score, as explained earlier. A few nodes are highlighted, such as the U.S. Army (est. 1775) and the *Cuerpo de Ingenieros del Ejército* (Army Corps of Engineers) for the military domain, the *Ministère de la Reconstruction et de l'Urbanisme* (Ministry of Reconstruction and Urban Planning, 1944–1953) and the *Eidgenössische Stabsbureau* (Federal Staff Bureau) for the administrative domain. There are also three examples from commercial mapmakers: Rand McNally & Co. (est. 1856), Eleanor Foster (active 1931–1936), and E. R. Mitford (1811–1869). While Rand McNally was one of the largest private map producers in history, as stated in the next chapter, Eleanor Foster, located at the periphery of the social graph, is completely absent from the literature. From the data, we only know that she drew a pictorial map of Florida dated 1935 (see Fig. A4). The map contains a detailed chronology of the United States, as well as illustrations of local agricultural products and leisure activities. Eustace Revely Mitford (1825–1905) is another example of an understudied small producer located in the periphery of the social graph. He drew several cadastral maps, especially in the mining areas of the Yorke Peninsula near Adelaide, Australia, as a complement to the official cadastral surveys.

Further qualitative examples of semantic attributions can be found in Table A1.

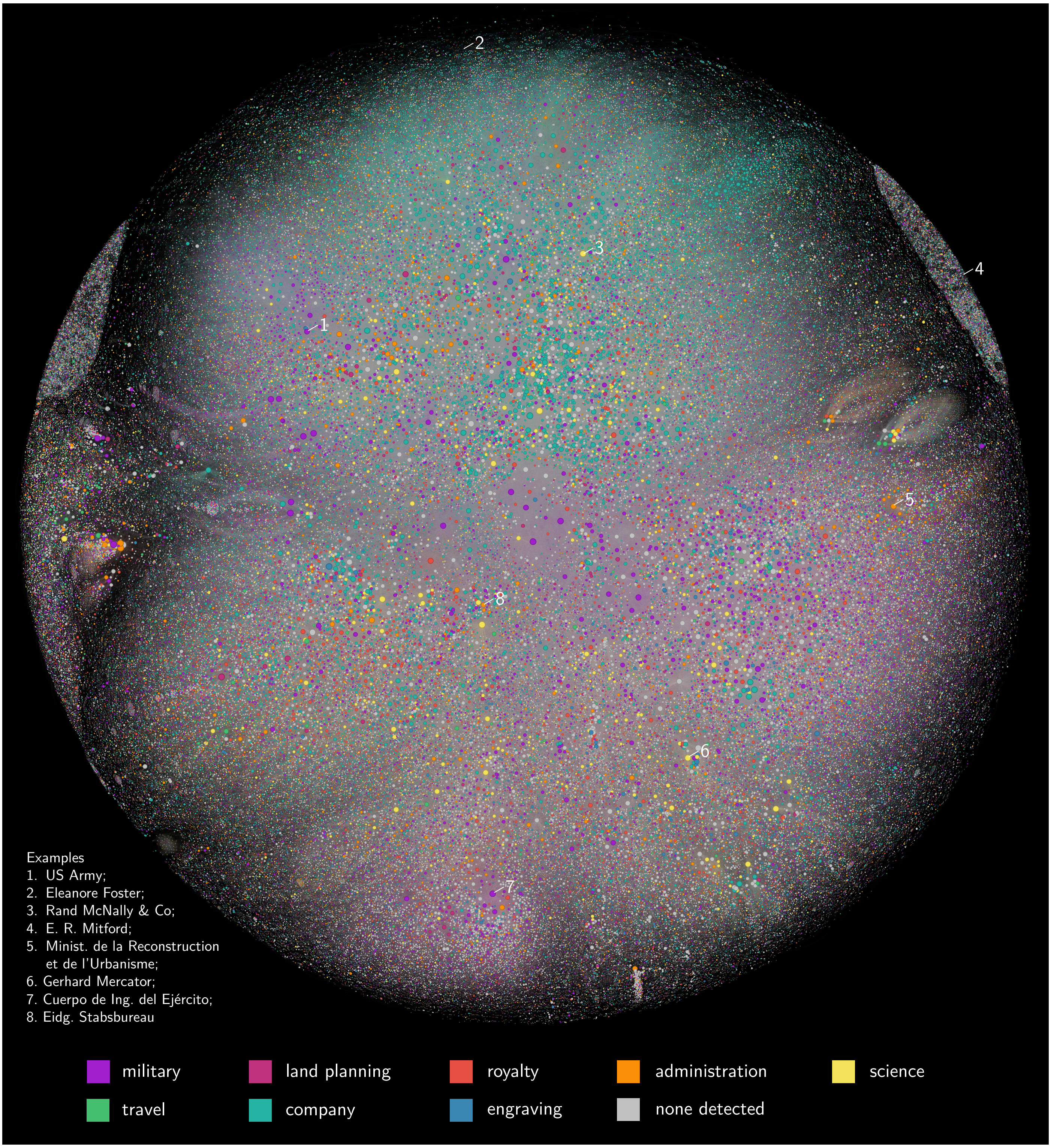

**Figure 12 | Network of map creators by typology of activity.** For each node, color denotes the semantic proximity between a creator's name and one of 8 semantic domains, corresponding to distinct typologies of creators. *Typology of activity is not a strong determinant of work collaboration.*

Figure 12 shows that semantic domains are highly intermingled. From a qualitative point of view, American mapmakers, up top, seem to be more often attached to the commercial domain. The portion of the graph corresponding to Washington D.C. also contains more creators from the military and administrative domains compared to the New York area. French map producers, on the right-hand side of the visualization, appear to be closer to the military domain as well. Overall, however, in contrast to spatial and temporal variables, semantic domains do not correspond to clearly distinct areas of the social graph. This is reflected in the rather insignificant modularity ($Q = 0.09 < 0.3$), whereby the domains seem to be densely interconnected rather than forming separate work clusters. This finding challenges the normal practice of studying maps by topic (e.g., military maps, commercial maps), advocating instead for a holistic perspective, or the study of corpora focused on specific regions or time periods.

Figure 13a illustrates the relative importance of semantic domains over time, in the corpus under study. Specifically, it presents the share of map creators associated with each domain, denoted as cumulative average likelihoods rather than absolute counts. Figure 13b, on the right, indicates the relative production count for each type of creators, corresponding to the node sizes in Figure 12. Probabilistic estimates of overall map publication volumes could thus be calculated by multiplying the values in Figure 13a by the average counts in Figure 13b.

In the 16th century, mapmakers connected to the printing and engraving industry, as well as creators patronized by royal power or associated with the nobility, seem to dominate. These two domains appear to progressively loose importance, although the latter remains highly influential until at least the 18th century. By contrast, the involvement of military-related producers increases gradually, plateaus in the late 18th century, decreases slightly at the end of the 19th century, and rises again at the beginning of the 20th century. Semantic markers linked to government and administration expand slowly throughout the period. Although they never become the most influential domain, they rank third at the end of the period under study. Creators associated with travel and touring follow a similar trajectory: they peak in the early 20th century and then decline slightly as the military domain expands. It is worth noting, though, that the travel and touring domain corresponds to comparatively low publication volumes, as visible in the upper right histogram. The semantic domain showing the most pronounced growth—peaking in the late 20th century well ahead of all others—is that of commercial and corporate cartography. Along with military and administrative entities, private commercial mapmakers also tend to produce *more maps* than average. Overall, the largest producers, however, are learned societies and academic cartographers, which partially counterbalance the small number of distinct creators in that category, reflected in the persistently low salient similarity score.

The pronounced peak in corporate mapping suggests a proliferation of private producers in the 19th century. At that time, new printing technologies and the appearance of color printing enabled maps to become mass-market products (Ristow, 1975). Maps were also increasingly produced and used by private insurances, gas, electricity, oil, and mining companies. For instance, the Sanborn Map Company, established in 1866, produced hundreds of thousands of detailed plans of U.S. cities, including building footprints, for fire insurance purposes (Grim, 2015). This popularization was reflected, at the turn of the 20th century, in the growth of touring and travel cartography with the entry into the market of companies such as Michelin in France (est. 1889), which offered itineraries, travel guides, and road maps (Olson, 2010; Pannetier & Houdoy, 2015). Although the average number of publications in the travel and touring domain is low, the number of maps sold was not necessarily low since a single map could be printed and sold in many exemplars. This nuance applies to other domains as well. Administrative or scientific documents, however, were less likely to be disseminated in large quantities.

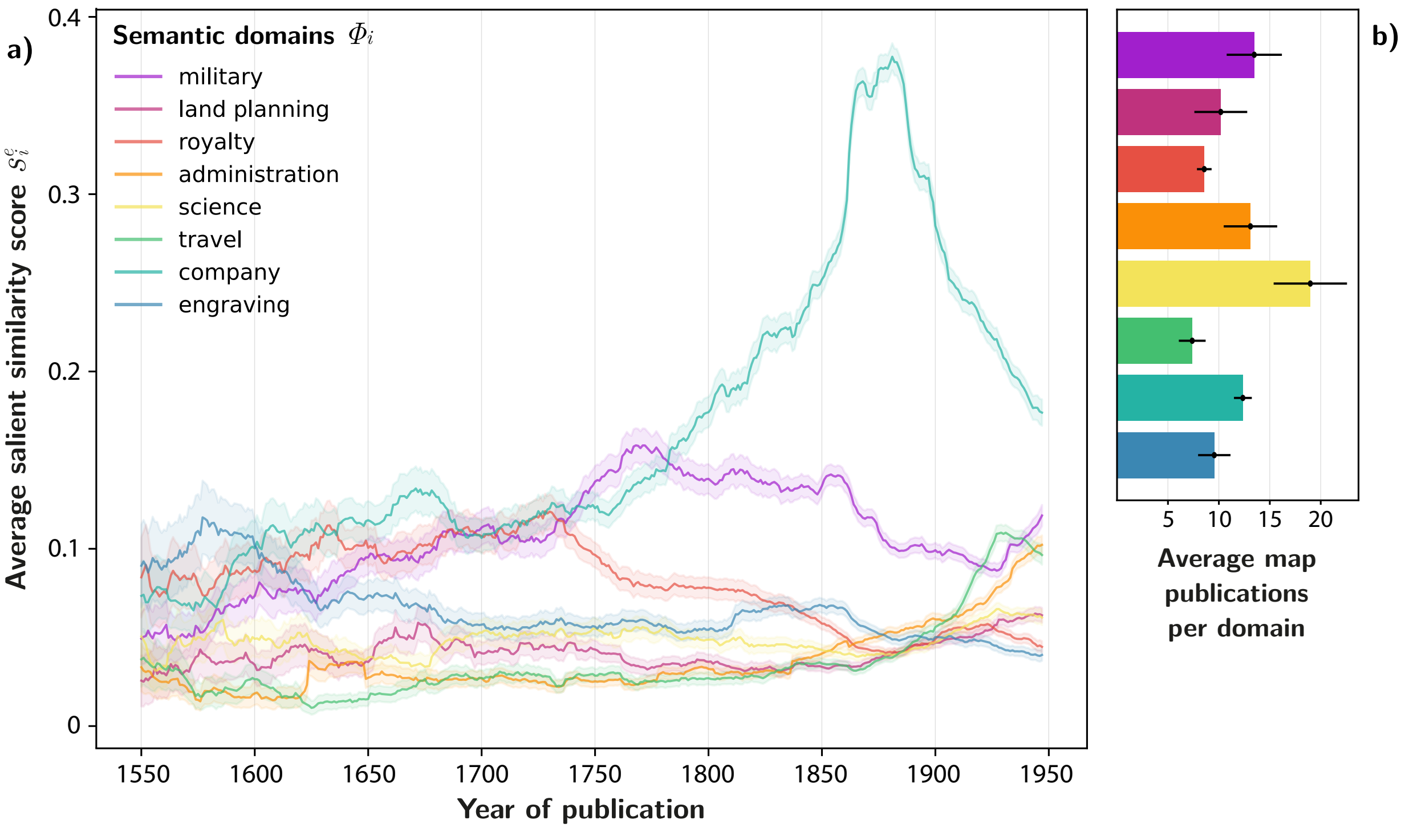

**Figure 13 | (a) Semantic domain of map creators by year of publication.** Mean cosine similarity between map creators' names and 8 semantic domains, based on GPT text-embedding-3-small. Similarity values are normalized and clipped between 0.25 (threshold) and 1.0. Each semantic domain is represented by a set of embedded keywords (see Tab. A3 in Appendix). Means are computed with a 25-year moving window. The colored area indicates the 95% confidence interval of the mean. Values prior to 1550 are not shown because of low statistical significance. **(b) Average number of publications per creator and semantic domain.** Colors match the legend of subfigure (a). The horizontal bar denotes the 95 % confidence interval of the mean. *Mapping companies are by far the most common type of map creator in the latter half of the 19th century.*

Two hypotheses might explain the rise and subsequent decline of corporate mapping between 1800 and 1950. The first concerns competition. As new technologies, such as lithography, were developed at the beginning of the 19th century and facilitated map printing, additional producers entered the market. Increased competition arguably allowed some mapmakers to distinguish themselves at the expense of others, who may have abandoned the market. The second hypothesis is lexical, and data related. From this perspective, the observed surge could be attributable to a popularization of linguistic markers, such as ‘& co’[8], which may have elevated the average salient similarity score for this particular domain.

One question raised by these results—which chapters 6 and 7 of the dissertation will modestly attempt to address—is the extent to which these evolutions in the typology of creators also affected the form of cartographic figuration, and how.

## 2.5 Conclusion

This chapter engaged with the data contained in ADHOC Records database through quantitative methods and the discussion of selected examples. It laid initial foundations for a digital analysis of cartographic history, highlighting the geographical, social, and temporal structures of mapmaking, through the analysis of publication volumes. In doing so, it mentioned in multiple occasions that mapmaking is linked to issues of power and knowledge creation. Chapter 3, that follows, will build on this premise by examining how cartographic output can be viewed as the manifestation of *spatial attention*.

[8] This hypothesis is supported by a baseline investigation of the top 100 unigrams and bigrams, from which we identified five n-grams arguably related to the *company* domain: ‘ co.’, ‘ co$_{\langle end \rangle}$’, ‘comp’, ‘& co’, and ‘Verlag von’. Together, they totaled 111,903 records. Pearson’s correlation between the n-gram time series and $s^{e}_{company}$ reached a value of 0.84 ($p_{value} = 5 \cdot 10^{-55} \ll 0.05$), indicating a very strong dependency. This dependency, however, seemed to be characterised by a temporal offset, where the increase of $s^{e}_{company}$ preceded the n-grams by 15 to 20 years, when it reached a maximum correlation of 0.98 ($p_{value} = 5 \cdot 10^{-138} \ll 0.05$). This result suggests that both economic competition and semantic shift could be valid explanations. In this regard, the semantic shift might even be a marker of the stabilization of the market, with the consolidation of the most successful mapmakers as an outcome.

# Appendix A – Supplementary Materials

**Text A1 | Joan Blaeu's dedication of the French version of the Cosmography to the King of France Louis XIV.** Original French transcript. Joan Blaeu. *Le Grand atlas, ou Cosmographie blaviane, en laquelle est exactement descritte la terre, la mer et le ciel.* 1663. Amsterdam. Bibliothèque de Besançon, 313 T.1. URL: https://memoirevive.besancon.fr/ark:/48565/016w3gdzv7b4.

Au Roy tres chrestien Louys XIV,

Sire, Je prends la hardiesse d'offrir à Vostre Majesté, admirée de tout l'Univers, non le fein Atlas des Poëtes, qui, selon leur dire, en soustient le faix sur les espaules ; mais sous un nom fameux les Cartes, & la Description de toutes les Provinces du Monde. J'ay creu qu'un ouvrage si considerable, qui joint avec l'Histoire, fait le plus beau livre des Roys, à present que mes soins, & mes travaux l'ont augmenté, & enrichy de tous les ornemens, qu'il pouvoit attendre de l'impression, ne seroit pas tout à fait indigne, quand on l'auroit outre cela mis en François, de paroistre aux yeux du plus puissant Monarque de l'Europe.

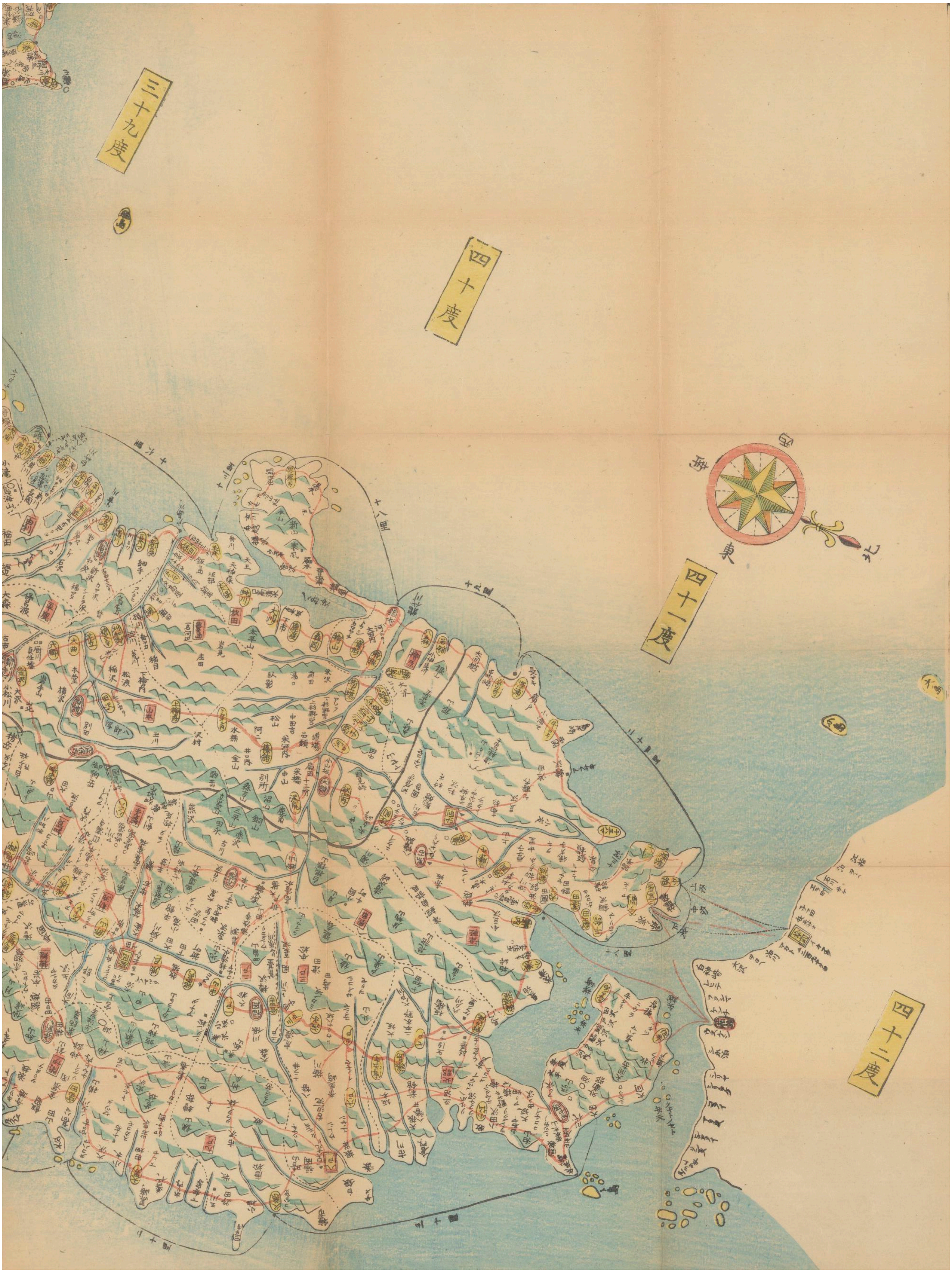

**Figure A1 | Map of Mutsu Province, Japan, 1775**. By Sekisui Nagakubo. *Map of Japan*. Osaka. 1775. BnF, GE FF-420. 48 x 36 cm. URL: gallica.bnf.fr/ark:/12148/btv1b53165324t/f2.

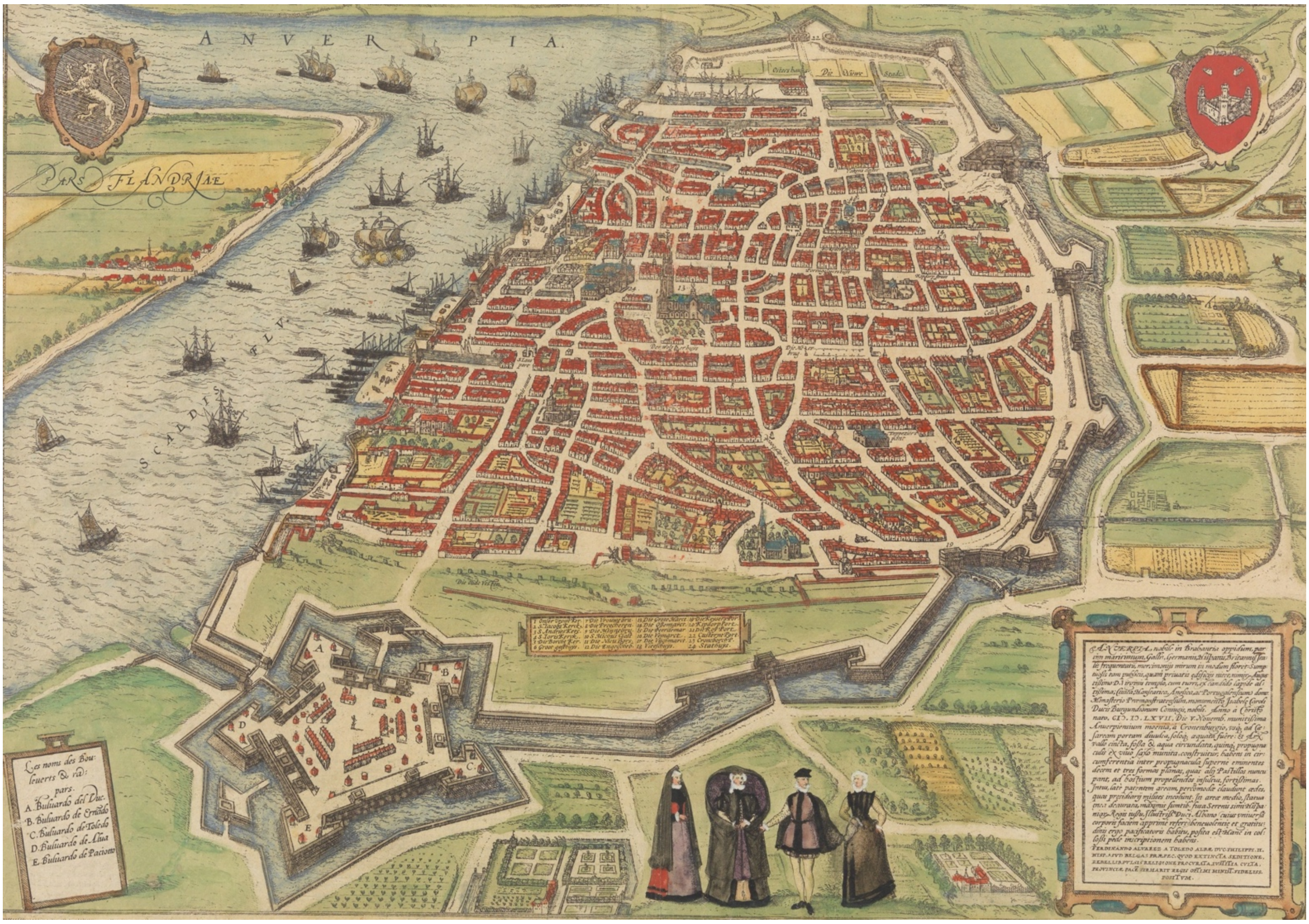

**Figure A2 | Chorographic map of Antwerp in oblique perspective, between 1572 and 1594**. By Joris Hoefnagel, Cornelius Caymox, Franz Hogenberg, and Simon Novellanus. *Anverpia, Civitates Orbis Terrarum*. Edited by Georg Braun, Cologne. Between 1572 and 1594. Metropolitan Museum of Art, 61.650.37. 38 x 53 cm. Hand-colored etching. URL: www.metmuseum.org/art/collection/search/366878.

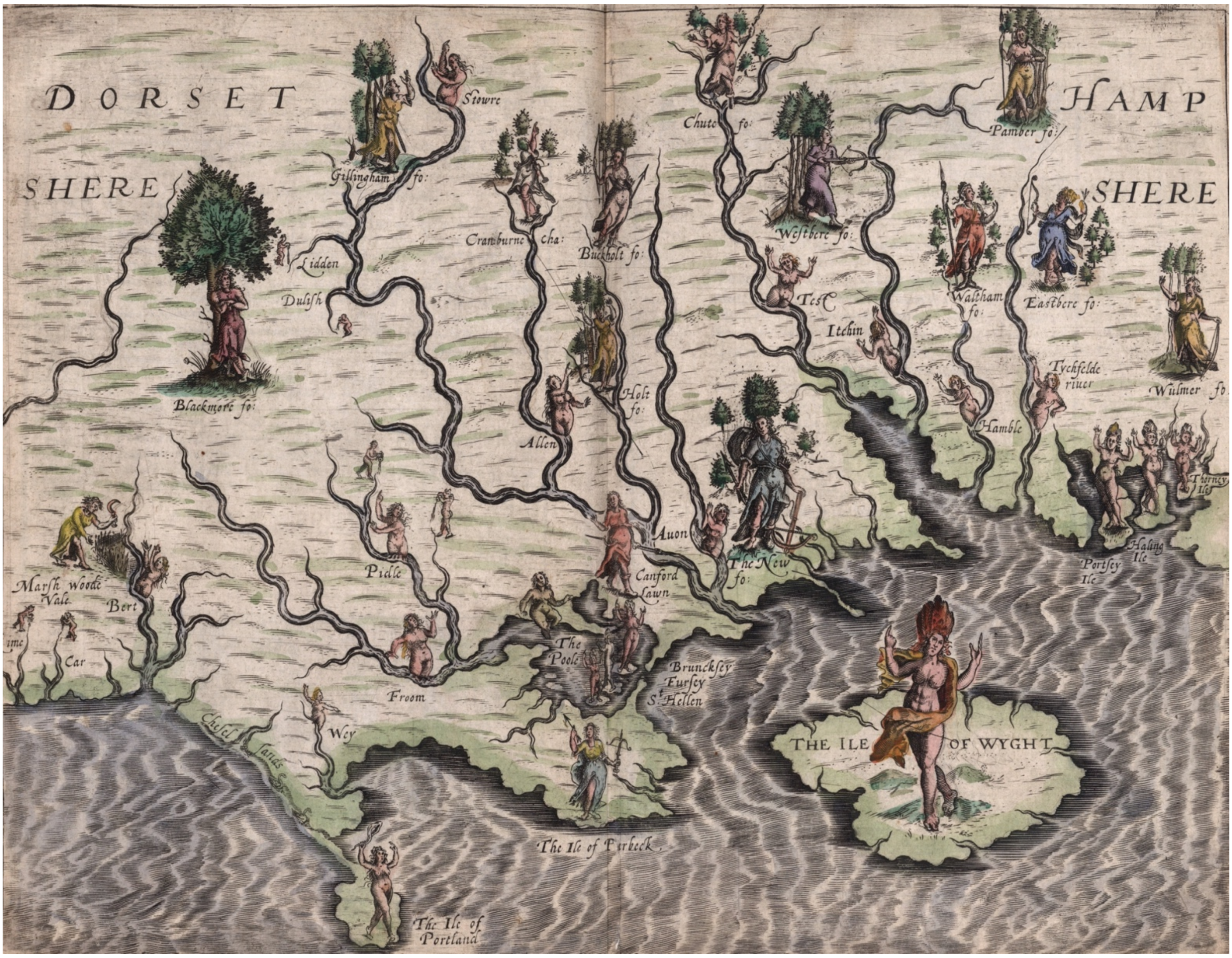

**Figure A3** | **Pictorial allegoric map of Dorsetsh[ire], Hampsh[ire], Poly-Olbion, 1622**. By Michael Drayton. *A Chorographicall Description of All the Tracts, Rivers, Mountains, Forest, and other Parts of this Renowned Isle of Great Britain*. Printed for Augustine Mathewes, John Marriott, John Grismand, and Thomas Dewe, London. 1622. David Rumsey Map Collection, 12180.000. 28 x 21 cm. URL: www.davidrumsey.com/luna/servlet/s/xueoro.

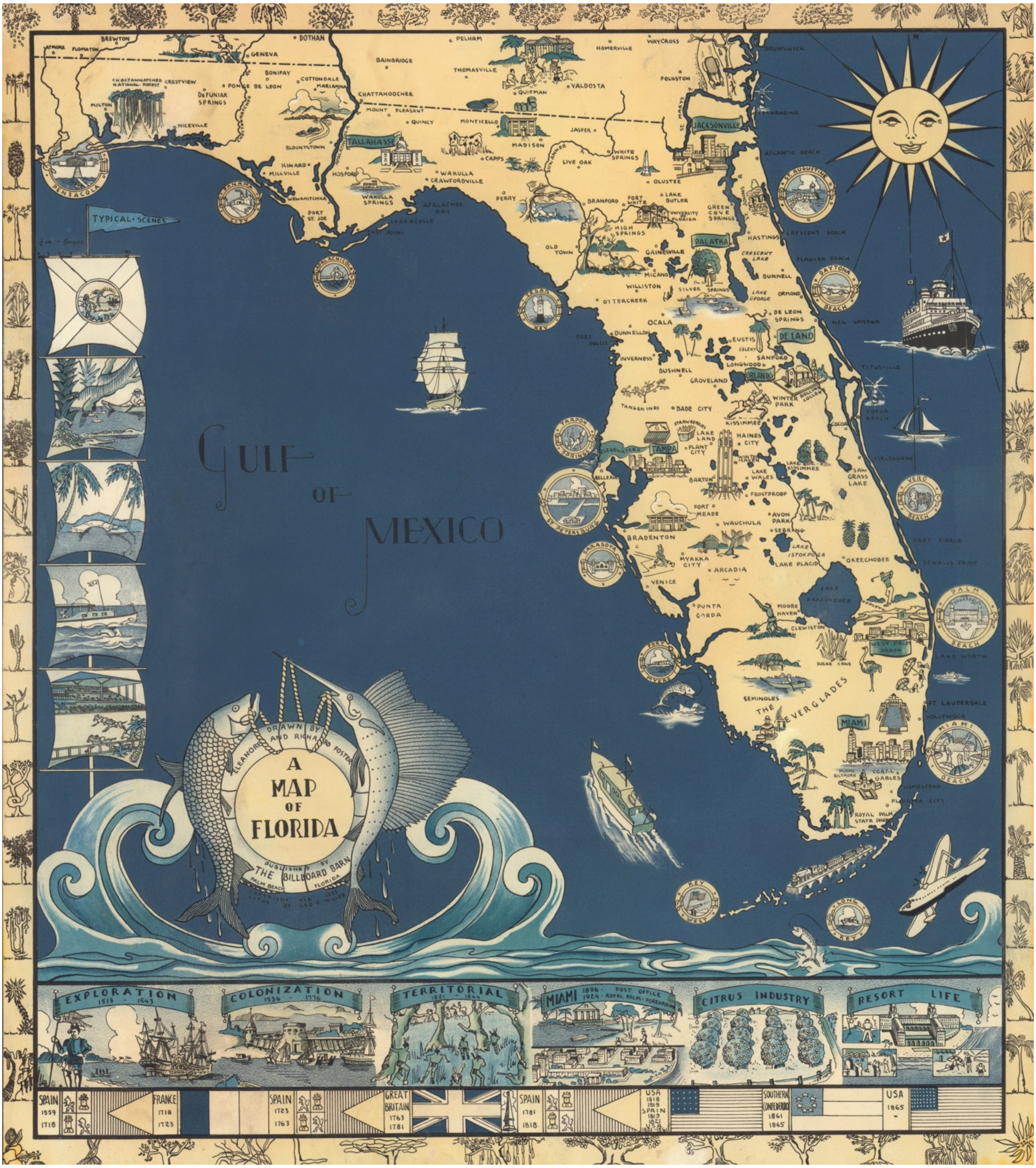

**Figure A4 | Pictorial map of Florida, 1935**. By Eleanor Foster, Richard Foster, and George C. Miller. *A Map of Florida*. Published by The Billboard Barn. 1935. David Rumsey Map Collection, 10558.000. 66 x 59 cm. Lithogravure. URL: www.davidrumsey.com/luna/servlet/s/5iqj94.

**Table A1 | List of the ISO-3 country codes** found in the manuscript.

| | | | |
|---|---|---|---|
| ARG = Argentina | CUB = Cuba | ITA = Italy | PRT = Portugal |
| AUT = Austria | DEU = Germany | JPN = Japan | ROU = Romania |
| AUS = Australia | DNK = Denmark | LUX = Luxembourg | RUS = Russia |
| BEL = Belgium | ESP = Spain | MMR = Myanmar | STP = São Tomé and Príncipe |
| BRA = Brasil | FRA = France | MEX = Mexico | TUR = Turkey |
| CAN = Canada | GBR = United Kingdom | NLD = Netherlands | UGA = Uganda |
| CHE = Switzerland | GRC = Greece | NOR = Norway | USA = United States |
| CHL = Chile | HTI = Haiti | PER = Peru | ZAF = South Africa |
| CHN = China | IDN = Indonesia | PHL = Philippines | |
| COL = Colombia | IND = India | POL = Poland | |

**Table A2 | List of keywords used to represent the semantic domains of map creators**

| Military | Company | Royalty | Science |
|---|---|---|---|
| admiral | & co. | emperor | academy of sciences |
| air force | & company | empire | cartographic institute |
| armed forces | & sons | imperial | geographical society |
| army | business | king | geography |
| army corps | chartered company | prince | knowledge |
| battle | commercial associate | queen | measure |
| captain | corporate | real | professor |
| commandant | electricity | reich | research |
| defense | gaz company | royal | scientific society |
| leutnant | insurance | sovereign | study |
| marine | registered firm | | university |
| military | | | |
| military officer | | | |
| military survey | | | |
| navy | | | |
| strategic | | | |
| troops | | | |
| war | | | |

| Administration | Engravers | Company | Travel |
|---|---|---|---|
| administrative | colored | agriculture | city guide |
| agency | copperplate | architecture | guidebook |
| bureau | engraver | construction | itinerary |
| commission | etching | forestry | road map |
| committee | iconographic | land | touring club |
| department | lithographer | landscape | tourism |
| government service | printing | planning | transport map |
| ministry | typographer | public works | travel |
| municipality | | regional planning | |
| public office | | soil | |
| secretariat | | territory | |
| state office | | urban | |

**Table A3 | Sample of 35 creators and corresponding salient similarity scores $s_i^{\vec{e}}$.** Drawn at random. From left to right, the columns correspond to the semantic domains $\Phi_i$: Military, Land planning, Royalty, Administration, Science, Travel, Company, Engraving

| Creator | Milit. | Land | Royal. | Admin | Sci. | Travel | Comp. | Engraving |
|---|---|---|---|---|---|---|---|---|
| Schropp, Simon | · | · | · | · | · | · | · | · |
| Corsi, Francesco S. Zuccagni-Orlandi | · | · | · | · | · | · | · | · |
| Pennacca, J. | · | · | · | · | · | · | · | · |
| 満州国林野局 (Manchuria National | · | 0.5 | · | **1.3** | · | · | · | · |
| Forestry Bureau) | · | · | · | · | · | · | · | · |
| Grinderslev Sogn, Nörre Herred | · | · | · | · | · | · | · | · |
| Lanée et Bazin | · | · | · | · | · | · | · | · |
| Ubiña | · | **0.9** | · | · | · | · | · | · |
| Vinal, J | · | · | · | · | · | · | · | · |
| Chevellier, Henry | · | · | · | · | · | · | **0.3** | · |
| K. v. Spruner Hand | · | · | · | · | · | · | **1.9** | · |
| Howell, Reading | · | · | · | · | · | · | **0.4** | · |
| Apud P. Mariette | · | · | · | · | · | · | · | · |
| Boermayer Officier | **0.8** | · | · | 0.1 | · | · | · | · |
| Bat° | · | · | · | · | · | · | · | · |
| Richard Boeckh und Heinrich Kiepert | · | · | · | · | · | · | **2.2** | · |
| Five Principal Races | · | · | · | · | · | · | · | · |
| S. J. Anville | · | · | · | · | · | · | **0.8** | · |
| Kajiya, Keitarō | · | · | · | · | · | · | · | · |
| Brull y | · | · | · | · | · | · | **0.6** | · |
| R.O. de 18 de Octbre | · | · | · | · | · | · | · | · |
| Eniwetok Atoll | · | · | · | · | · | · | · | · |
| Nakanishi, Kyūichi | · | · | · | · | · | · | · | · |
| von E. | · | · | · | · | · | · | · | · |
| Kesler | · | · | · | · | · | · | · | · |
| F. Restelli | · | · | **0.9** | · | · | · | · | · |
| Communes d'Epinal, Golbey | · | · | · | **0.1** | · | · | · | · |
| Consolidates Southern Railway | · | · | · | · | · | · | **1.8** | · |
| dos Engenheiros Civis Portugueses | · | **0.3** | · | · | · | · | · | · |
| Stansbury, Howard | · | · | · | · | · | · | **0.6** | · |
| Changemens Politiques | · | · | · | · | · | · | · | · |
| J.M. Cunningham | · | · | · | · | · | · | **1.3** | · |
| Warren, Hubert | **0.6** | · | · | · | · | · | · | · |
| Étallon, Auguste | **0.1** | · | · | · | · | · | · | · |
| S. Mahon ... New York | · | · | · | · | · | · | **1.6** | · |
| Bernecker | · | · | · | · | · | · | · | · |
| | Milit. | Land | Royal. | Admin | Sci. | Travel | Comp. | Engraving |

# Chapter 3

## *The Map and its Power(s)*

The previous chapter examined how maps were produced—who produced them, where, and when. This chapter will turn to their content, by addressing the geographical subjects of maps or, more precisely, the *spatial attention* of cartography. It will also prolong the interpretative framework of the preceding chapter by investigating relations of power in cartography. In particular, it will engage with political, economic, and military dimensions.

### 3.1 Defining spatial attention

One of the most salient components of map content is what one might call spatial coverage, or the geographic focus of the map. In this chapter, however, I suggest a new perspective of the underlying phenomenon, and thus propose an alternative construct, which I term *spatial attention.*

Cartography does not aim to create a uniform, agnostic layer of information, nor to represent the territory in a complete and invariable manner. As Umberto Eco humorously demonstrated, there can be no such thing as an exhaustive map (Eco, 1994). Mapmaking is a resource intensive process, constrained by funding, time, and production capabilities (Sponberg Pedley, 2005b). Therefore, it must be a partial process, focusing selectively on specific geographic areas and environmental attributes. I refer to this specific emphasis as *spatial attention.*

In contrast to a perspective where the world would be conceptualized as a grid whose quadrangles can be filled with information until, eventually, the entire surface of the Earth is covered[1], speaking about spatial attention acknowledges that mapping demands active and sustained engagement. Attention is never random. It may be directed deliberately, for instance toward a territory deemed strategically or commercially significant. Attention may also be triggered, by change or a by perceived threat. Spatial attention can also emerge from cultural attraction, a fascination with the exotic, or the romantic imagery of places such as Venice, Jerusalem, or the Rocky Mountains. The outcome of attention is not necessarily truthful information. Rather, it is, to extend the cognitive metaphor, a *percept* that results from both external cues (e.g., the territory) and internal representations (e.g., cultural assumptions). Ultimately, spatial attention may arise from different *intentions*.

---

[1] In this perspective, it would suffice to extend the grid concept inherent in the notion of spatial coverage to the time dimension.

This idea naturally prompts to ask why certain areas receive heightened detail whereas others remain obscured. It further leads to questioning the role of small-scale maps—those that depict entire countries or continents—which offer limited detail on the experienced environment but are found to be so frequent. Figures 1–5 attempt to visualize the notion of spatial attention. Each figure or subfigure maps the cumulative spatial attention of a specific publication country over the entire study period. Color intensity corresponds to the number of map records depicting a specific area. Accordingly, lighter areas are frequently depicted in cartographic production, whereas darker areas are scarcely represented. The intensity diminishes progressively as the scale decreases, denoting differences in spatial granularity.

## 3.2 Investigating spatial attention

### *American cartography*

Figure 1 portrays American spatial attention. Its distribution peaks in New York; in other words, New York constitutes, according to the ADHOC database, the most common geographic subject in American cartography. Other major American cities, including Boston, Philadelphia, Washington, Chicago and San Francisco are also commonly depicted. Outside the United States, the American attention is widely distributed, with Europe, and more prominently Western Europe, attracting comparatively higher attention. American maps of Europe included low-scale, detailed maps, which indicates *close attention*, as well as *distant*, larger-scale maps. The United States similarly covered other regions with a relatively large-scale cartographic production, particularly Latin American and the Caribbean, the Coast of Guinea in Africa, and, to a lesser extent, Japan, China, and the Philippines. Finally, sparks of attention are localized on prominent cities, like of Hong Kong, Syndey, Melbourne, Auckland, as well as Jerusalem.

The representation of spatial attention is particularly dense, which is why discussion mainly focuses on its salient aspects. First, the pre-eminence of New York reflects the city's early and enduring importance, particularly in economic terms. Furthermore, although New York's cartographic output (20,205 map records) starts earlier, it is far surpassed by Washington (50,532) in absolute terms. The city's heightened attention thus underscores its cultural symbolism within the Western imaginary, notably as the gateway to America for many immigrants. Accordingly, the proportion of maps published in the United States that were dedicated to Europe is notable. This pattern may be attributable, at least partly, to the historical role of Europe as a cultural reference or, perhaps, to a certain interest for the homeland—evidenced by the frequent appearance of cities like London, Amsterdam, Paris, and Rome. In addition, section 3.5 will demonstrate that the two World Wars prompted a substantial cartographic output, which may also explain for some part the American focus on northeastern France, the Benelux countries, and the Rhine Valley.

Overall, and particularly when compared with other countries, the results reflect the United States' global ambitions. Although Latin America and the coast of Guinea did not attract the same detailed attention as Europe, both regions were economically important for the supply of metals, rubber, coffee and, historically, slave trade. From 1823 onward, cartography also served to legitimize American influence in the Americas under the Monroe Doctrine, when the United States positioned itself against European colonialism on the continent.

Cartography of Asia also reflects the economic and political interests of American imperialism. One of the main historical milestones in this policy was the Treaty of Kanagawa (1854), through which the United States imposed economic integration on Japan, an event which prompted Japan to modernize rapidly to confront Western powers.

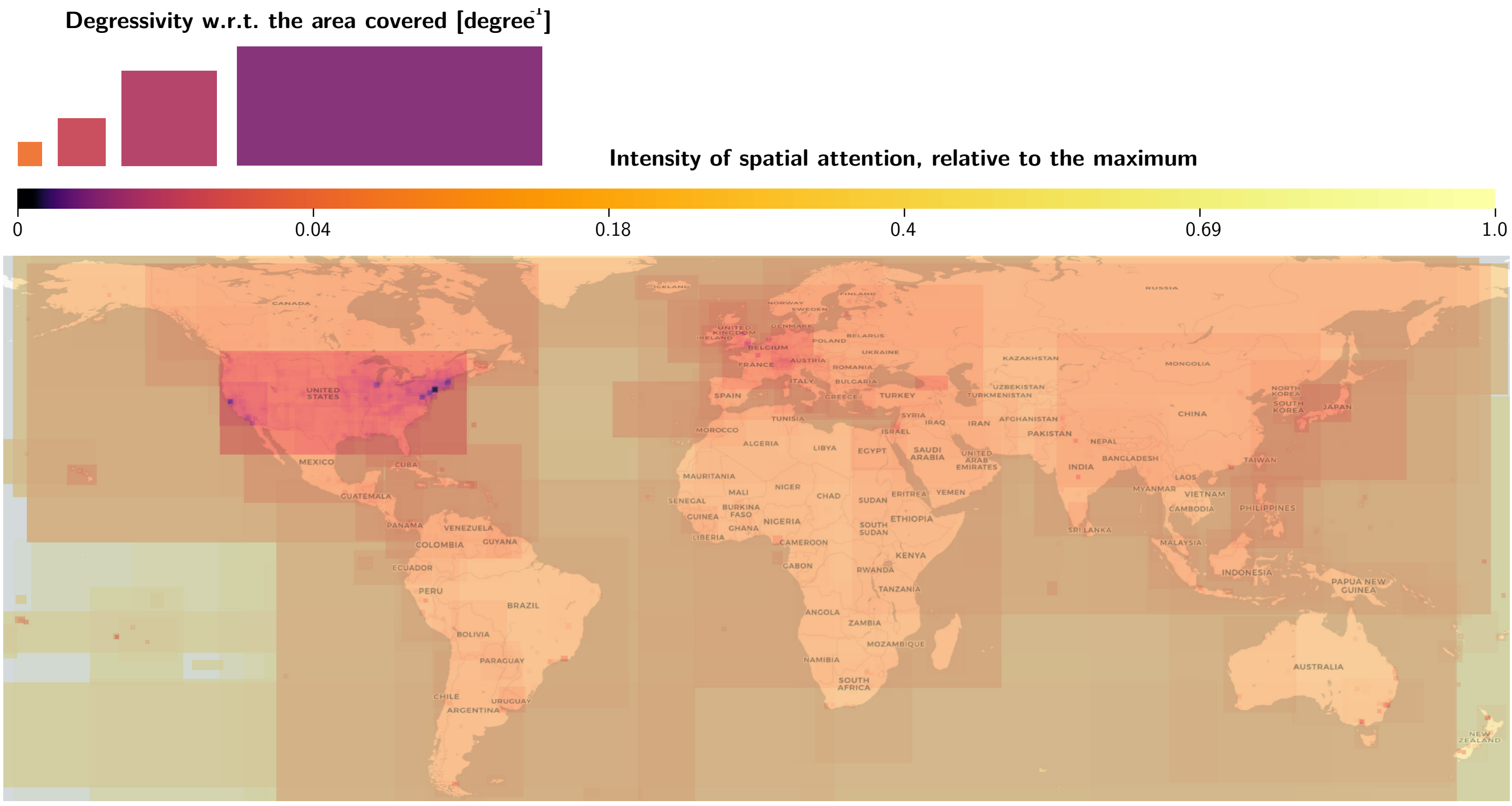

**Figure 1 | Spatial attention map of American cartography.** The color scale intensity is relative, reflecting the number of map records depicting a given area. Lighter areas are extensively depicted, whereas the darkest areas are scarcely covered by cartography. The impact of smaller scale maps is distributed over the depicted area, i.e., the intensity decreases with the area covered (see legend on top). Large-scale maps are enlarged to a minimum of 30' for visualization purposes. WGS84 projection. *American cartography is mainly focused on American cities and western Europe.*

### *European focus*

French spatial attention is comparatively less widespread than American attention; the focus on Paris largely dominates. The French output seems lower than the American one. However, the color intensity in the figure is normalized. Thus, the comparatively tenuous intensity of French attention primarily reflects the overrepresentation of Paris maps. Close attention was nevertheless also paid to the rest of France. The Mediterranean region was commonly depicted, especially Northern Africa. Likewise, numerous areas of the Caribbean were regularly mapped (Haiti, Guadeloupe), as were the shores of the St. Lawrence River in Quebec. In the United States, Louisiana, San Franscisco and the Northeastern cities were the most frequent cartographic topics.

Dutch cartography exhibits a pattern similar to the French. In addition to granular mapping of the Low Countries (including present-day Belgium), it maintained a close focus on neighboring regions, notably France, with a particularly pronounced concentration on Paris, as well the rest of Western Europe. Outside Europe, Dutch attention turns to New Holland and the Caribbean. The South American coasts were depicted as well, particularly Guiana. Dutch cartographic production was most sustained, however, in representations of Indonesia, and the Island of Java, the Cape, in South Africa, and Ceylon. Charting of the Atlantic coast and islands also appear remarkable common.

As with its production pattern (see Chapter 2), Italy's domestic cartographic focus is multipolar. Rome, Florence, Milan, Venice and Naples are all often depicted. Paris, London, and the Ardennes region in France are also salient. In smaller-scale maps, Italy and its European neighbors were common subjects, but also Eritrea, Libya, and Egypt in the South. Northeastern American States, Cuba, and Uruguay likewise received a certain attention. San Francisco, Los Angeles and Cape Town were also recurrent topics. In the East, Palestine and Lebanon are well represented.

Spanish spatial attention was concentrated on the Iberian Peninsula, with the regions of Madrid, Barcelona and Cadix being the most represented, according to the corpus. The northern coast of Morocco, the Baleares, and the Canary Islands likewise received heightened attention. Outside Europe, maps tend to focus on historical Spanish colonies, notably Cuba, Puerto Rico and the island of Hispaniola, in the Caribbean, as well as California, Mexico, and Panama in Central and Northern America, more prominently on settlements, like Havana, Los Angeles, San Francisco, Veracruz and Mexico. In the south, Spanish cartography addressed both the coast and the interior, with an emphasis on major cities, the Bay of Buenos Aires, Uruguay, and, farther east, southern Brazil. In East Asia, the focus centers on the Philippines.

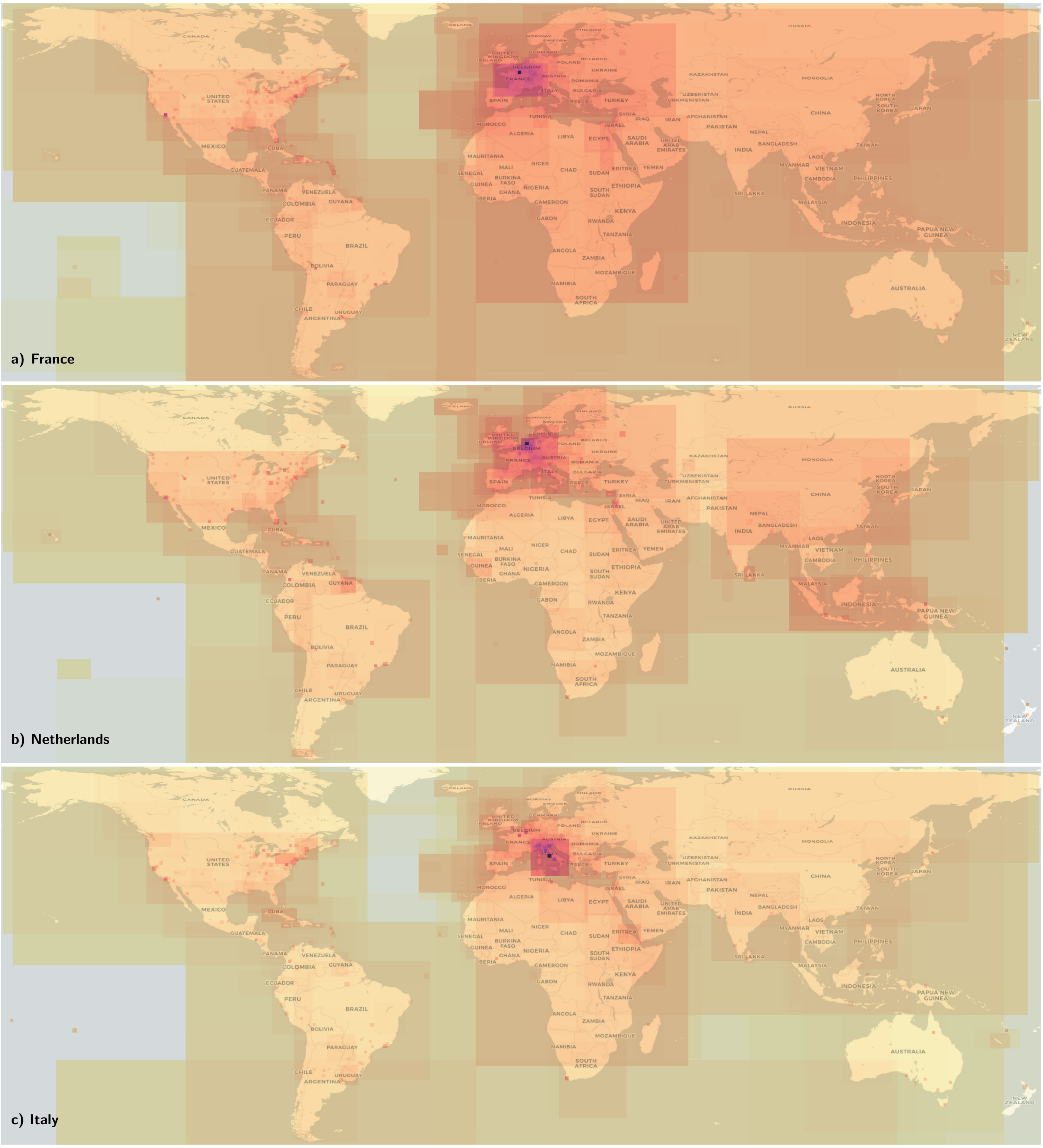

**Figure 2 | Spatial attention map of (a) French, (b) Dutch, and (c) Italian cartography.** See Fig. 1 for legend details. *French and Dutch cartographies were mainly focused on their respective national capitals, their European territories, and historical overseas.*

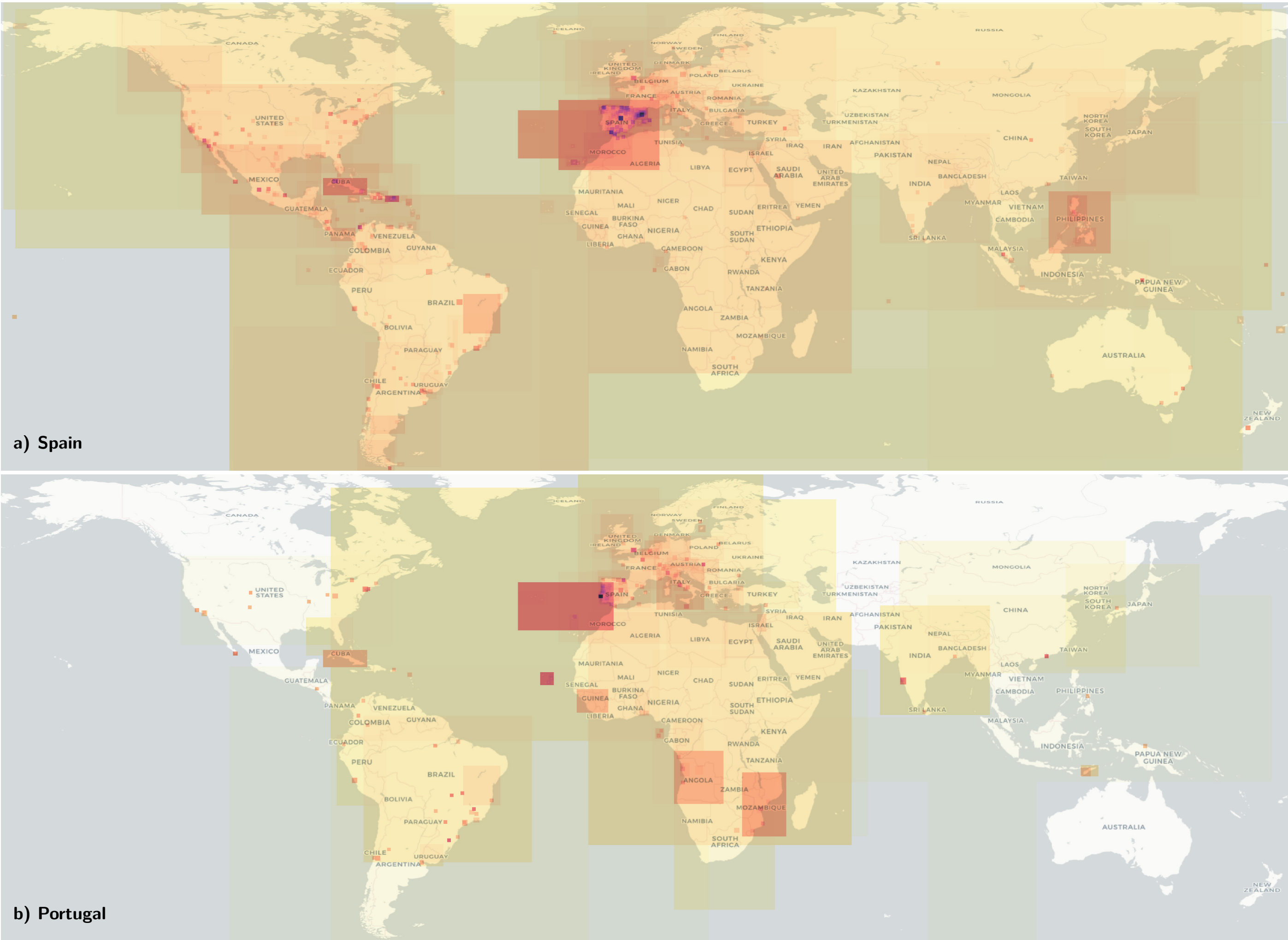

**Figure 3 | Spatial attention map of (a) Spanish, and (b) Portuguese cartography.** See Fig. 1 for legend details. *Spanish and Portuguese cartographies were mainly focused on their European territories, and American historical dependencies.*

Portuguese cartographic production was more modest yet highly focused. It devoted salient, close attention to Portugal, Western Europe, and neighboring Spain, especially the port city of Bilbao. Madrid, Valencia, Milano, Rome, Malta, London, Paris, and Budapest are also represented. In North America, New York stands out. Again, the Spanish and Portuguese islands in the eastern Atlantic were recurring subjects. On the other continents, Brazil appeared to receive the most focused Portuguese attention, with coastal cities such as Porto Alegre as well as inland settlements such as São Paulo. Other territories colonized by Portugal, notably Guinea, Angola, São Tomé and Mozambique in Africa, Timor in Oceania, the city of Goa in western India, and that of Macao in China, also stand out, suggesting pronounced cartographic focus.

Danish mapping exhibits a similar pattern, featuring a pronounced concentration on Copenhagen, followed by relatively strong representation of Jutland, Bornholm, and Schleswig, to the West. Within Europe, the Low Countries, and the Rhine Valley, were also often represented, including with large-scale maps. Sweden, Norway, the Faroe Islands, Iceland, and Greenland were depicted as well, albeit without granular detail.

Concerning Switzerland, self-attention largely dominates. Zurich, the economic capital and largest city in the country, is the most common geographic topic, followed by other cities like Basel and Bern. In addition, Swiss attention is focused on the Rhine Valley in the north. The proportion of maps representing areas outside Europe is low, addressing foremost American cities.

The spatial attention of the German states is largely concentrated on Europe at a distant scale, with a marked production of maps on the neighboring countries, England, and the continent itself. At a finer scale, this attention shifts east of present-day Germany toward Bohemia and Prussia. The United States, particularly Northeastern cities, were also portrayed, while the Caribbean received modest coverage compared to other European States. Beyond these areas, attention to the remainder of the world is diffuse and its repartition parallels French cartography (Fig. 2).

European states deployed dense, granular attention to their own territories and to adjacent European regions. A pronounced self-attention is observed, often centered on political or economic capitals such as Paris, Amsterdam, or Zurich. The spatial attention of the principal European colonial empires (France, Spain, the Netherlands, and Portugal) generally targeted their historical colonial dependencies, although rarely with the intensity accorded to their European territories, except in the Caribbean and other Atlantic and Pacific islands. Portugal, whose European territory was relatively small compared to other colonial power, constitutes a partial exception; it devoted a substantial share of its cartographic production to its historical overseas possessions.

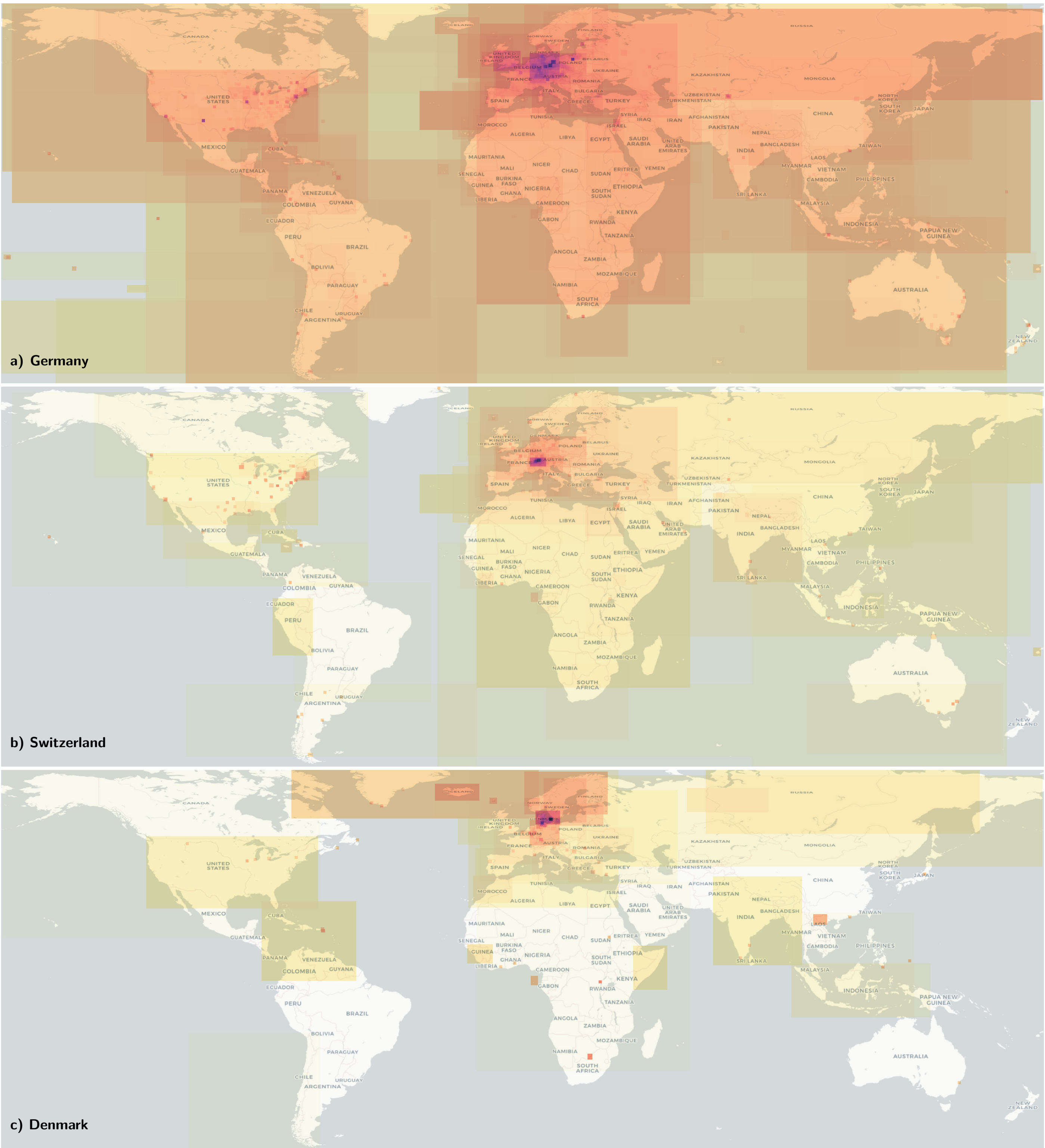

**Figure 4 | Spatial attention map of (a) German, (b) Swiss, and (c) Danish cartography.** See Fig. 1 for legend details. *German cartography was focused on Europe, often with large-scale maps; Swiss and Danish cartographies were mainly concentrated on their national territories.*

I employed the terms *close attention*, or granular attention, and *distant attention* to characterize the results. These two terms articulate a recurring distinction in the structure of attention, which the visualization highlights. Close and distant attention correspond not only to different map scales but may also reflect distinct uses and purposes. Close attention pertains to relatively detailed maps, produced at large- or medium-scale, such as cadastral plans, city maps, and topographic maps.

Most of the time, large-scale maps served as administrative devices for enforcing land ownership, supporting urban planning, land planning, and the organization of the territory. Medium-scale topographic maps, by contrast, were typically used in a military context. The particularly dense and close attention to Europe reflects the considerable financial and scientific resources mobilized by European states, especially from the late 18th century onward, to produce detailed and geometrically accurate cartographic documentation of their national territories, notably through the establishment of national mapping agencies and cadastral offices (Clergeot, 2007). Such cartographic endeavors typically constitute processes of power-knowledge (Crampton, 2001; Crampton & Elden, 2007), as maps strengthened the capacity of states to administer territory, develop infrastructure, and levy taxes. A comparable pattern occurred in the overseas possessions, where cartography facilitated extractivist policies, and legitimized colonial authorities to claim lands portrayed either as uninhabited or, conversely, as populated by cruel or immoral "savages" (Genevois et al., 2024; Kivelson, 2009, pp. 50–51). In this respect, the most thoroughly mapped areas—often islands—were also frequently colonies that remained longer in the European sphere of influence, sometimes until the mid–20th century. The influence of cartography on that situation must, however, be moderated, since islands are also often easier to defend; the continued European presence therefore impacted cartographic production at least as much as the reverse. Moreover, close cartographic attention seems to be indicative of the depth of colonial implantation, population and culture, even in territories that had obtained *de jure* independence. This pattern is evident in the case of the United States, South Africa, Québec, and much of Latin America, all now frequently considered Western offshoots. A corollary of the probable relationship between close attention and colonial implantation is the differential mapping of interiors: whereas in North and South America, maps abundantly depict inland settlements and regions, in Asia they tend to concentrate on port cities, such as Goa, Kolkata, and Hong Kong.

Both similar and dissimilar to close attention, *distant* attention is characterized by the production of smaller-scale maps that portray large regions, countries, or even continents. Even though these maps afford less direct administrative control over the physical environment they depict, they are no less powerful or historically consequential. By shaping socio-cultural constructions, such as the Italian *nation*, the *New World*, or *Pan-Germanist Europe* (Figs. 1, 2, and 4), they effectively enforce borders, spheres of influence, and claims of sovereign legitimacy over territory.

The structure of spatial attention also reflects distinct geopolitical stances. The United States and, to a lesser extent, France, the Netherlands, and Italy, articulate global ambitions characterized by widely distributed—albeit non-homogenous—cartographic attention. The German states, although maintaining a global outlook, concentrate primarily on Europe. Whereas the German Empire pursued overseas expansion after the Berlin Conference of 1884, particularly in Togo, but also in Cameroon, Namibia, Tanzania, and New Guinea, these regions seem scarcely represented in German map production. The relatively brief duration of the German colonial empire, which ended after World War I, provides a plausible explanation: mapping a territory is a lengthy process that often requires several decades, whereas the colonial German Empire persisted for around thirty years only. This situation is not confined to the German presence in Africa. The continent in its entirety remains comparatively underrepresented in European cartography—aside from territories bordering the Gulf of Guinea, South Africa, the Canary Islands, and the island of São Tomé—notwithstanding its status as a major target of colonial expansion during a historical period of sustained cartographic production.

The results also highlight increased selectivity among states with lower map production, underscoring constrained production capabilities and the resource-intensive character of mapmaking. This pattern characterizes the Spanish Empire, for instance, whose historical map production was probably five to ten times lower than that of France or Germany (cf. Chapter 2, Fig. 1). In such circumstances, a majority of the production was concentrated on the colonial empire. The effect is even more apparent in the Portuguese case, where colonies received heightened attention, comparable to the homeland. For the Swiss Confederation, attention is overwhelmingly centered on the economic capital, Zurich, the national territory, and adjacent regions. The Kingdom of Denmark exhibits a similar tendency, although it constitutes an intermediate case in view of its historical expansion in Greenland and the Caribbean.

### *Another point of view: the Asia–Pacific region*

Two distinct study cases may provide comparative context for the discussion of European and American cartographic outputs. On the one hand, Japan, and on the other, Australia. Japanese cartographic attention was concentrated on Tokyo. The country's four principal islands—Honshu, Hokkaido, Kyushu, and Shikoku to the southwest—were mapped densely and with granular detail. The Kuril Islands, Sakhalin, and Taiwan were also represented, although with progressively less intensity. Beyond Japan, attention extended mainly southward and westward toward China, Russia, Burma, Malaysia and, to a lesser extent, Indonesia and the Philippines. The coasts of Korea and China, in particular, were the subject of numerous large-scale maps. Western Europe and the United States were also mapped, but only peripherally, with detailed production focusing often on the coasts rather than the capitals.

Australian cartography provides a particularly instructive contrast to Japanese and American practices. Domestically, it concentrated on the temperate, urbanized regions, notably Victoria, Melbourne, Sydney, Canberra, Adelaide, Brisbane, and Perth in the west. However, unlike its American counterpart, most of Australia's interior received little cartographic coverage. Despite the continent's great distance from Europe and the United States, and in contrast to Japan, Australian mapmakers devoted close and sustained attention to European cities, especially in the United Kingdom, as well as the American East Coast, and the Midwest. This outward focus is directed particularly to culturally or politically prominent cities, notably New York, Los Angeles, London, Edinburgh, and Dublin, encompassing also inland capitals like Paris and Berlin.

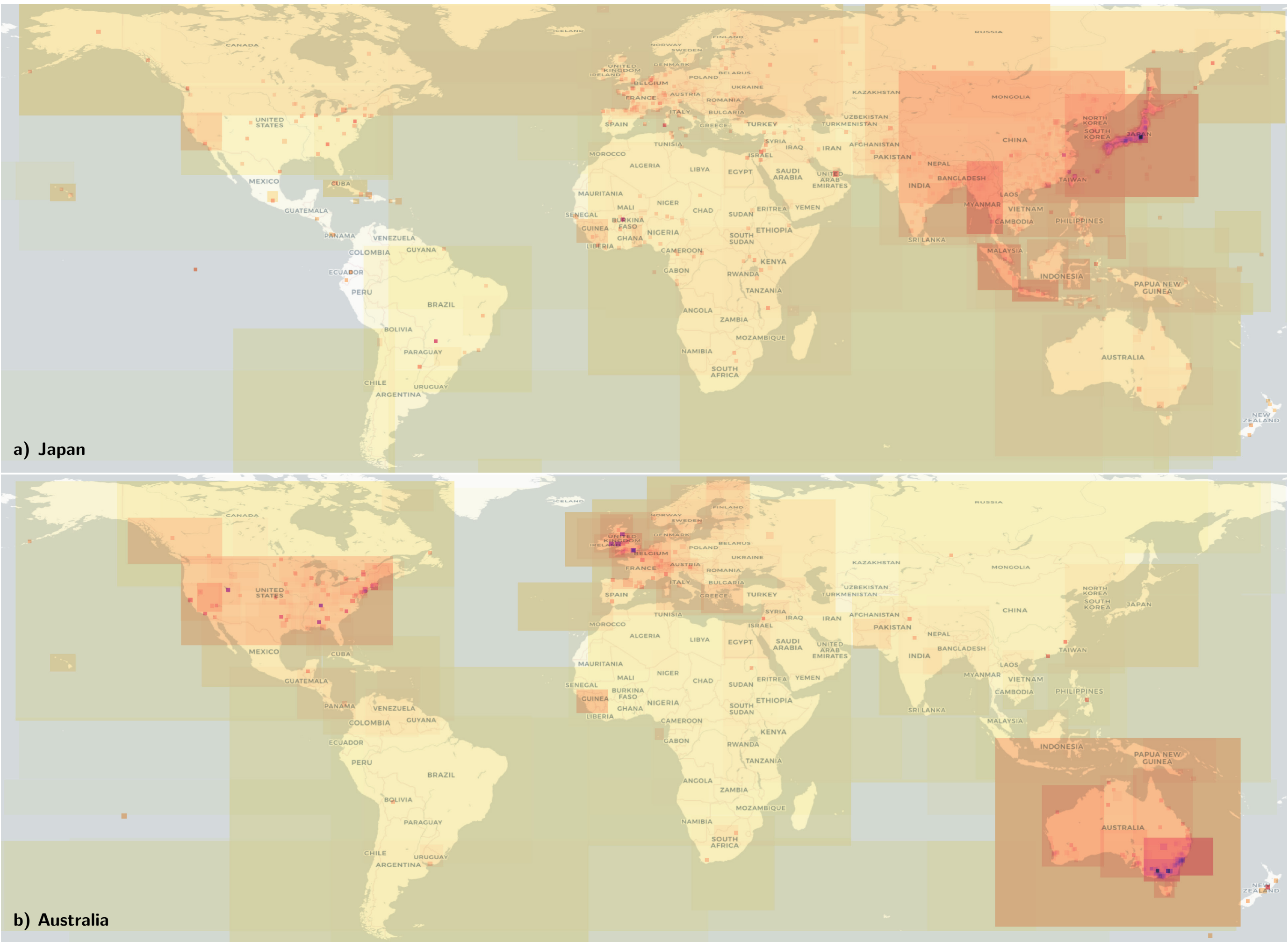

**Figure 5 | Spatial attention map of (a) Japanese, and (b) Australian cartography.** See Fig. 1 for legend details. *Japanese cartography is concentrated on their national territory, and small-scale maps of neighboring countries.*

Domestically, Australian cartographic attention seems to be focused primarily on economically significant areas, corresponding to highly organized, productive territory. Abroad, Australian production illustrates the role cartography plays in providing cultural references. This role is reflected in the production of maps of cities that symbolize the Western urban model or have played a historical role in the narrative of Western civilization. This cultural dimension, and the ability of maps to convey a compelling and vivid story about the world, is particularly significant for Australians, for whom animating the Western narrative was a means to construct their own cultural identity, in spite of their relative geographic isolation.

The Japanese case is striking in its structural similarity to European empires, although its geographical focus is shifted eastward. The similar pattern of dense, granular depiction of the national territory, centered on the capital, is visible. Cartography also extends to peripheral regions, such as Sakhalin, and the Kuril Islands. The latter were mapped as early as 1644, following their claim by Japanese aristocrats (Unno, 1995). Like in the Western case, a distinctive, distant, kind of focus is also visible, supporting the interpretative framework articulated around two complementary attention mechanisms: close and distant. Both forms of attention are apparent, for instance, in the case of China, a historical rival of Japan which was partially occupied by Japanese imperial forces in the 1930s, as were Burma, Indonesia, and Malaysia. The existing focus on Russia is unsurprising as well, given the longstanding animosity between the two powers that culminated in the Russo-Japanese War (1904–1905). Japan's victory in that conflict significantly advanced its expansionist ambitions. Finally, the close cartographic attention to European coastal regions, probably bolstered by the expansion of trade with the West in the second half of the 19th century, mirrors, in many respects, Western attention to South and East Asia, likewise centered on coastal cities.

The beginning of this chapter introduced the concept of spatial attention and examined its historical and socio-cultural dimensions, particularly its relationship to territorial control, colonization, and military expansion. Cartography can also reflect and impact economic interests through nautical charts and the mapping of coastal cities. Furthermore, the findings suggested that cartography could play a role in the construction of cultural references and national narratives. The remainder of this chapter will examine three major historical developments that shed light on three effective facets of the power of maps.

## 3.3 Maps to build nations

Several scholars have highlighted the importance of cartography in the construct of modern and territorial nation-state. J. B. Harley in his seminal essay *Maps, knowledge, and power* (Harley, 1988) argues that both are “inextricably linked”. Susan Schulten, examining U.S. cartography, highlights the pioneering role of mapmakers like Emma Willard (1787–1870), and educational maps in the early history of the American nation (Schulten, 2012). Through graphical features minor in appearance, such as lines, maps also played an instrumental role in conceptualizing borders and ideating modern territorial states (Branch, 2013). In France, this process is manifest in the great national series, initiated with the Cassini map, completed in 1790, and continued with the carte d'État-major (Arnaud, 2022b). Through triangulation, these maps achieved the remarkable feat of establishing the precise geographic extent of the state. Simultaneously, planimetric city maps became increasingly common. Although the impact of such maps may appear less direct, city plans shaped the urban ideas of the 19th century, from hygienism to the regular grid plan, thereby enabling capital cities and economic centers to grow beyond previous infrastructural limits. An emblematic example is Cerdà’s plan for Barcelona (Aibar & Bijker, 1997). A third development at the beginning of the 19th century in Europe was the boom of cadastral campaigns. The primary purpose of cadastres was to map land ownership (Vivier, 2007). By facilitating property taxation, cadastres reinforced private ownership and supported the enlargement of state administration (Clergeot, 2007). At every scale, then, maps and plans proved instrumental to the emergence of the modern state throughout the long 19th century.

This effect is manifest in the increasingly nation-centered orientation of map production shown in Figure 6. Although it may appear counterintuitive, as geographical knowledge of the world—the *theatrum mundi*—expanded with early globalization, intensified navigation, and growing international trade, cartographic activity became increasingly inward-looking. Between 1787 and 1876, the share of maps depicting domestic territories[2] rose significantly[3], from about 40% to nearly 70%.

This reinforces the idea that the map—much more than a *cosmographiæ*, a representation of the known world—is a socially constructed representation. The way mapmakers conceptualize the world is shaped by existing maps, laws, social norms, and other authoritative resources. When they draw the world, mapmakers tend to replicate those conceptions. Through their work, they generate new authoritative resources, effectively contributing the reproduction of existing cultural and social conceptions about space.

[2] Remember that here we are using contemporary national borders. Hence, the increase is not caused by the production of maps of the colonies, for example.

[3] This was verified using a Mann-Kendall statistical trend test, resulting in a highly significant and strong increasing trend ($p \approx 0.0$, $Tau = 0.99$).

Even ancient maps depict lines and boundaries, be it natural ones, such as rivers or mountain ranges. Thus, maps visually convey the notion of the border. Toward the end of the 18th century, spatial accuracy increased through triangulation campaigns, facilitated by incremental innovations in geometry and surveying techniques. Within the prevailing rationalist episteme, these advances further legitimized maps as authoritative documents and rightful repositories of geographic knowledge. Simultaneously, map production surged following the invention of lithography in the early 19th century. These developments in cartography, alongside other political events such as the French Revolution, fostered the emergence of a new intersubjective reality: the modern nation-state.

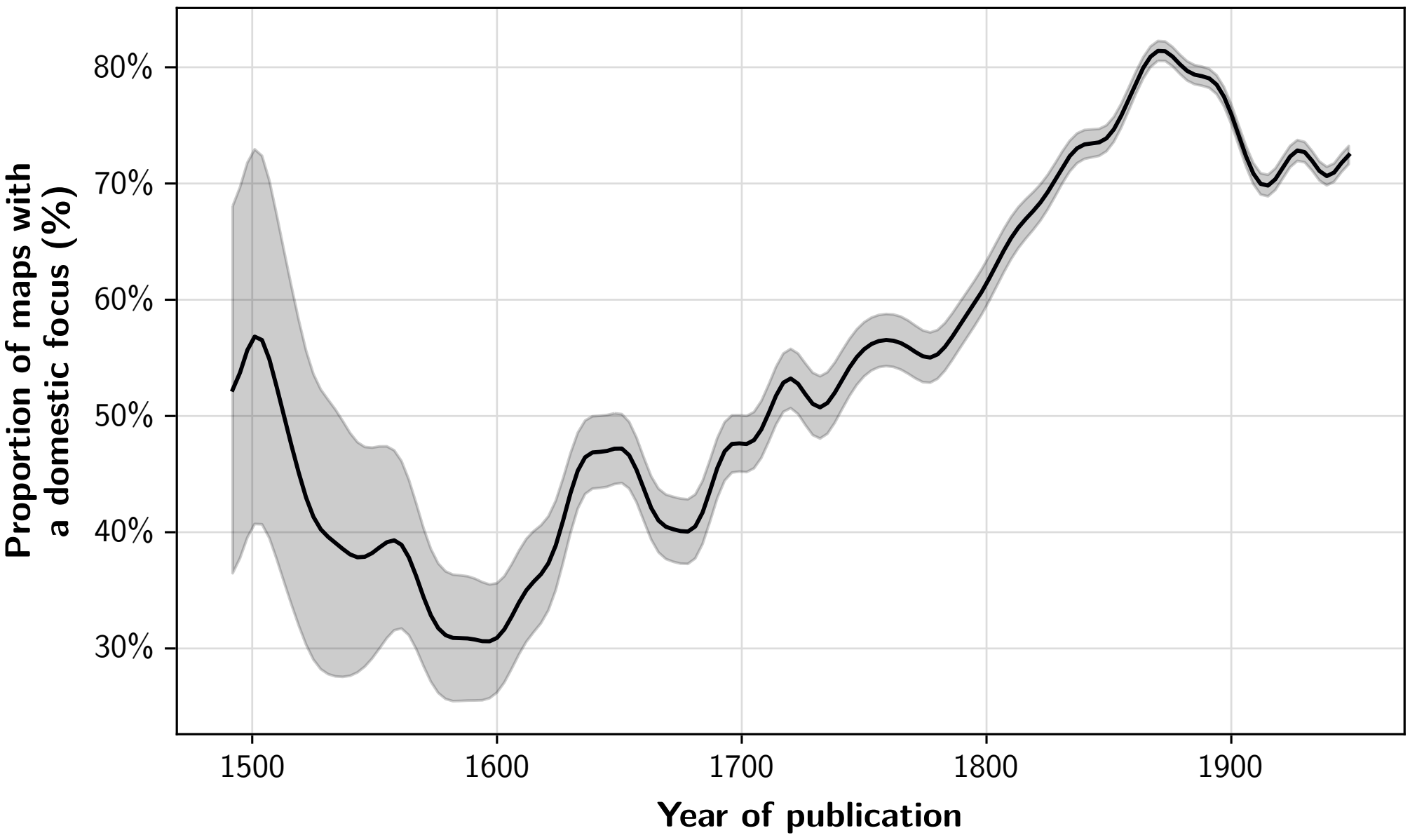

**Figure 6 | Proportion of maps with a domestic focus by year of publication.** Values are averaged over a 3-year window. The curve was smoothed using a Gaussian filter (σ = 2.5). The shaded area delineates the 95% confidence interval of the mean. *From 1600 to ca. 1880 the domestic focus of cartography increased.*

## 3.4 Maps to colonize and exploit

In the second example, I examine the relationship between Atlantic charting, navigation, and the triangular trade. It will not come as a surprise that cartography—particularly maritime charting—was instrumental in identifying and adopting secure, rapid navigation routes across the Atlantic Ocean. Map production was often the primary objective of exploration missions (Whitfield, 2015). Near the coasts, survey and reconnaissance were essential for identifying mooring points and ensuring that ships did not run aground in shallow waters. Consequently, many historical maritime charts include depth indications that were essential for facilitating approach maneuvers to commercial settlements in the Americas. Lower-scale Atlantic charts generally included detailed azimuth information, which enabled navigators to adjust a vessel's itinerary, even mid-ocean, using celestial navigation techniques. Such charts could shorten voyage times and reduce the likelihood of a ship becoming lost.

In addition, decorative cartography was widely used to advertise colonial projects to investors during the early stages of capitalism. As anticipated in the previous Chapter, the Dutch West India Company (WIC, 1621–1792) emerged as a major producer of maps. It employed cartographers such as Willem Blaeu (1571–1638) to create maps of Africa and the New World, one example of which appears in Figure 7. These maps were designed to extol the wealth and purported urban order of American cities, depicted here in the banner. Simultaneously, they portrayed autochthonous peoples as primitive undressed savages (e.g., *freti magellanici accolæ*, bottom right of Fig. 7) or even cannibals (*brasiliani*, middle right), thereby presenting Western colonial projects as morally defensible.

Thus, colonization, maritime trade, and cartography were closely intertwined; they reinforced and reproduced one another in a continuous feedback loop. The development of cartography facilitates seafaring and trade, and it encouraged investment as well as the establishment of colonial settlements. In turn, chartered companies, European sovereigns, and other stakeholders funded surveys and patronized mapmakers.

The existence and strength of this feedback mechanism are not merely hypothetical. The following paragraphs will demonstrate that they are supported by robust empirical evidence. First, based on ADHOC Records, the volumes of map publication addressing the Atlantic Ocean and its coastal territories was computed for each year. Then, these figures were cross-referenced with the Trans-Atlantic Slave Trade Database (2019) compiled by the Emory Center for Digital Scholarship and the University of California (Irvine, Santa Cruz). The period considered, 1501–1848, spans from the onset of the triangular trade to the abolition of the slave trade by the principal European colonial powers. After applying a light Gaussian filter to suppress aleatory fluctuations and correlating the two time series, the correlation coefficient exceeds 0.91, which indicates a very strong conjunctural dependency (Fig. 8)[4]. The magnitude of this relationship suggests that cartography is a key indicator in the rise of the Trans-Atlantic slave trade.

[4] Gaussian filtering results in an unbiased estimator.

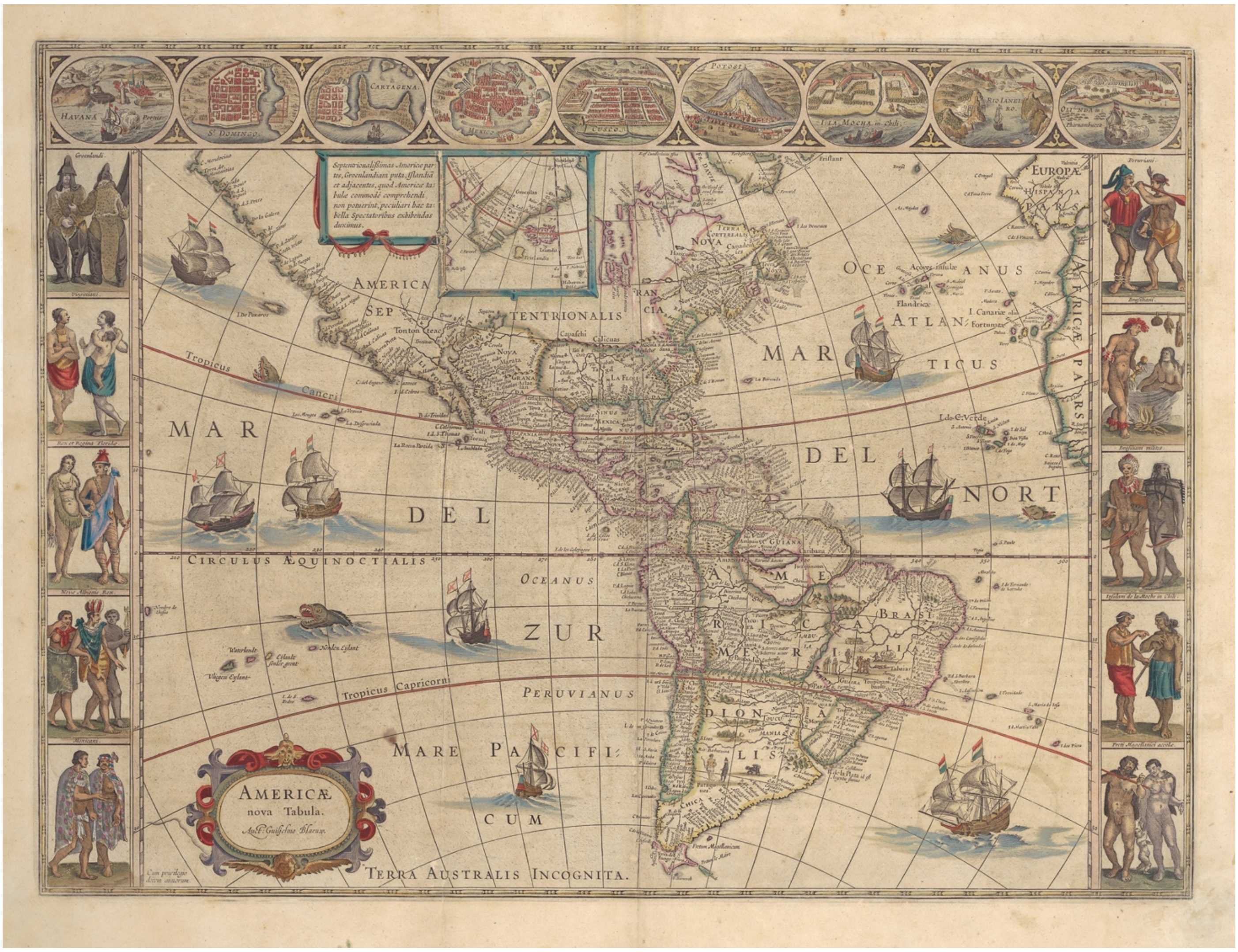

**Figure 7** | ***Americæ nova Tabula* by Willem Blaeu, 1614.** Copperplate engraving, hand colored. UTA Libraries / Wikimedia Foundation. *The side vignettes depict indigenous people in a stereotypical manner.*

The temporal relationship between the two processes can be estimated by shifting the map publication curve temporally. Figure 8 shows the result of this operation; the correlation peaks at t–8 (r = 0.93), implying a *delay* of approximately eight years between Atlantic charting and its effect on Trans–Atlantic slave trade. More specifically the temporal dependency between the two processes exhibits a tidal pattern, composed of three successive waves[5], suggesting that the impact of map publication on the slave trade culminates roughly 8 and 20–25 years after publication. Conversely, increases or decreases in the number of deported people are translated in almost identical proportional variations in maritime charting approximately five years later. This tidal pattern likely suggests the existence of a feedback mechanism. Moreover, the prolonged reciprocal influence between Atlantic charting and the triangular trade, evidenced by the very high levels of correlation maintained across substantial temporal offsets, suggest the *conjunctural* nature of this dependency.

[5] All values are statistically significant. Moreover, the last two waves are also statistically distinct ($p_{val} < .025$) from the trough. The first highpoint is tendentially distinct.

These findings illuminate another facet of the power of maps, supporting a plausible relationship between cartography and major historical developments, like colonial expansion, transatlantic trade, and slavery. They also provide empirical evidence that substantiates the presence of *feedback mechanisms*.

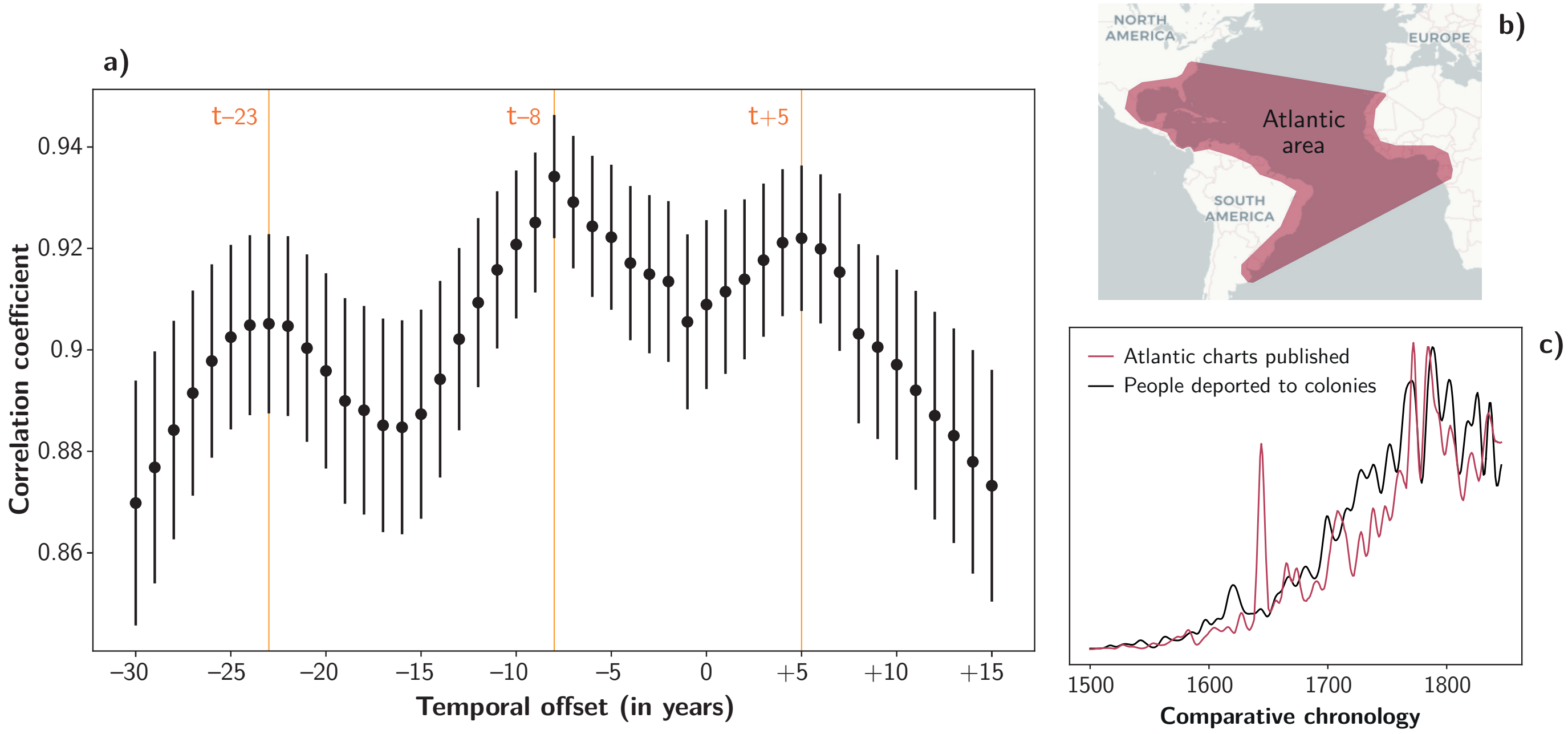

**Figure 8 | Conjunctural dependency between Atlantic charting and Trans–Atlantic slave trade.** The plot (a) displays the temporal evolution of Pearson's correlation coefficient between the annual production of maps depicting the Atlantic area and the number of enslaved captives embarked each year from 1501 to 1848. The short horizontal bars denote the 95% confidence interval of the coefficient, whereas the long, yellow bars indicate local maxima. To attenuate the impact of yearly aleatory fluctuations and ensure the measurement of medium-term conjunctural dependencies rather that shorter term relationships, both time series are treated with a light Gaussian filter ($\sigma = 2.5$, radius $= 5$). The resulting curves are shown in (c). The Atlantic area considered, represented in (b), is bounded by the coast of Mauritania and the Gulf of Guinea to the east, the Bay of Buenos Aires to the south, by the Caribbean Sea and the Gulf of Mexico to the west, and the coast of Virginia to the north. The Trans-Atlantic slave trade database is provided by the Slave Voyages Project. *Atlantic charting and Trans–Atlantic slave trade are very strongly correlated, reaching a maximal for a temporal offset of 8 years.*

## 3.5 Maps to win wars

The third dimension of power this Chapter will address pertains to the military domain. Tracing the military use of maps is difficult, since their practical utility, e.g. to compute ballistic trajectories, or plan troop movements largely depended on the gradual improvement in the geometric accuracy and surveying techniques. Moreover, armies and navies were historically intertwined with commercial and political interests, often governed by related social elites. By the late 18th century, maps had become the dominant medium through which strategy was conceived, and cartography was widely taught in military schools (Edney, 1994). Numerous maps of the period, intended for the instruction of aspiring officers, depicted famous battles and the relative positions of cannons and armies on the battlefield (see Chapter 6 on icons). The imperative to plan troop logistics and movements more effectively also motivated terrain surveying campaigns, first at strategic locations, e.g. along the border, and soon across entire national territories (Svenningsen, 2016).

A century later, at the beginning of the 20th century, maps had become indispensable to modern warfare. As shown in Figure 9, nine of the twenty largest map producers at the time were directly affiliated with the army or the navy. Nine were governmental survey agencies, such as the Ordnance Survey and the *Landesaufnahme*, which produced topographic maps intended for both civilian and military applications. Only two, by contrast, were primarily civilian organizations (Rand McNally and G. W. Bromley). The results suggest that map publication was not only related to defense policies; bellicist German and Japanese armies appear to have published more maps that their American or French counterparts, for instance. Planning and coordinating the logistics and movement of millions of troops would virtually have been impossible without detailed cartographic support. For example, the central fixture in Churchill government's war headquarters was a wall-size map of Europe and the world on which thousands of colored pins, representing Allied and enemy units, were repositioned each day to monitor the conflict. Other aspects of modern warfare, notably long-range artillery and air force arguably could not have been developed without cartography.

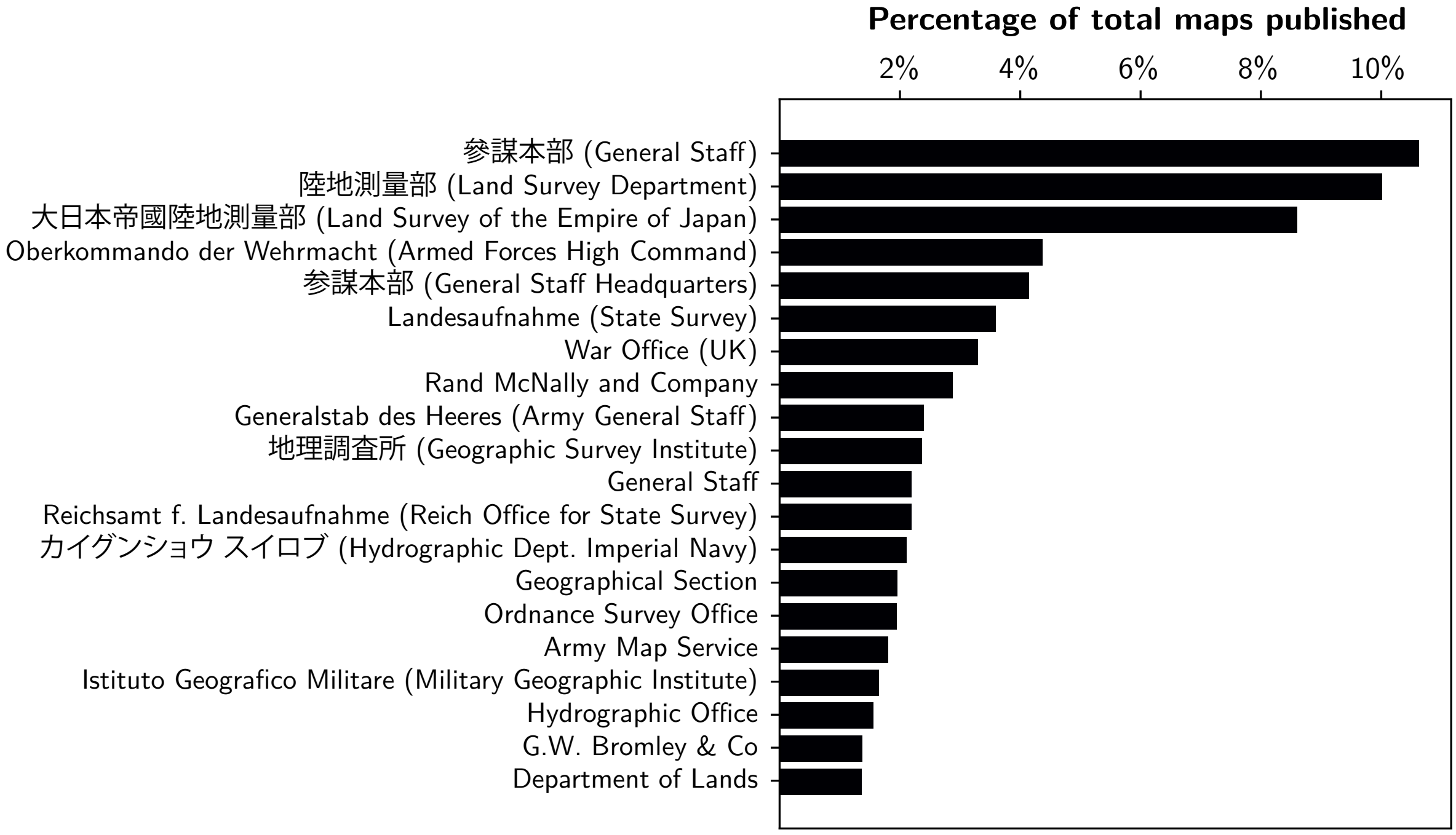

**Figure 9 | Top 20 map creators worldwide between 1897 and 1948,** ranked by their share of maps published. Percentages may exceed 100% as some maps have more than one creator. *Armed forces and related agencies are among the largest creators during the period.*

Figure 10 shows the substantial impact of large military conflicts on 20$^{th}$ century map publication volumes. As evidenced in Figure 10b, map publication not only surged during the two world wars (1914–1918, 1939–1945) but also closely reflected the intensity of each conflict, expressed as the number of casualties (Lyall, 2022). For instance, despite significant German advances, the number of casualties during the first two years of WWII (approximately 415,000–725,000 deaths per year) was about ten times lower than during the last four years (4,240,000–6,580,000). Map publication figures reflect this difference. In the first two years of the war, publication volumes merely pursued the increasing interwar trend. The rupture only occurred in 1941, after the extension of the conflict to all of Europe and North Africa and the involvement of Italy, the United States, and Japan.

Unlike the relationship between Atlantic charting and the triangular trade, the analysis of temporal dependencies between map production and war casualties in the first half of the 20$^{th}$ century does not reveal a tidal pattern; instead, it indicates an occurrent conjunction of limited temporal extent (Fig. 8a). Although the surge in map production tends to precede conflict amplification by a few months to a year—at least in the case of WWI—this dependence does not appear to be conjunctural. The statistical dependency diminishes rapidly, remaining significant for only two to three years and appearing negligible over periods exceeding eight years.

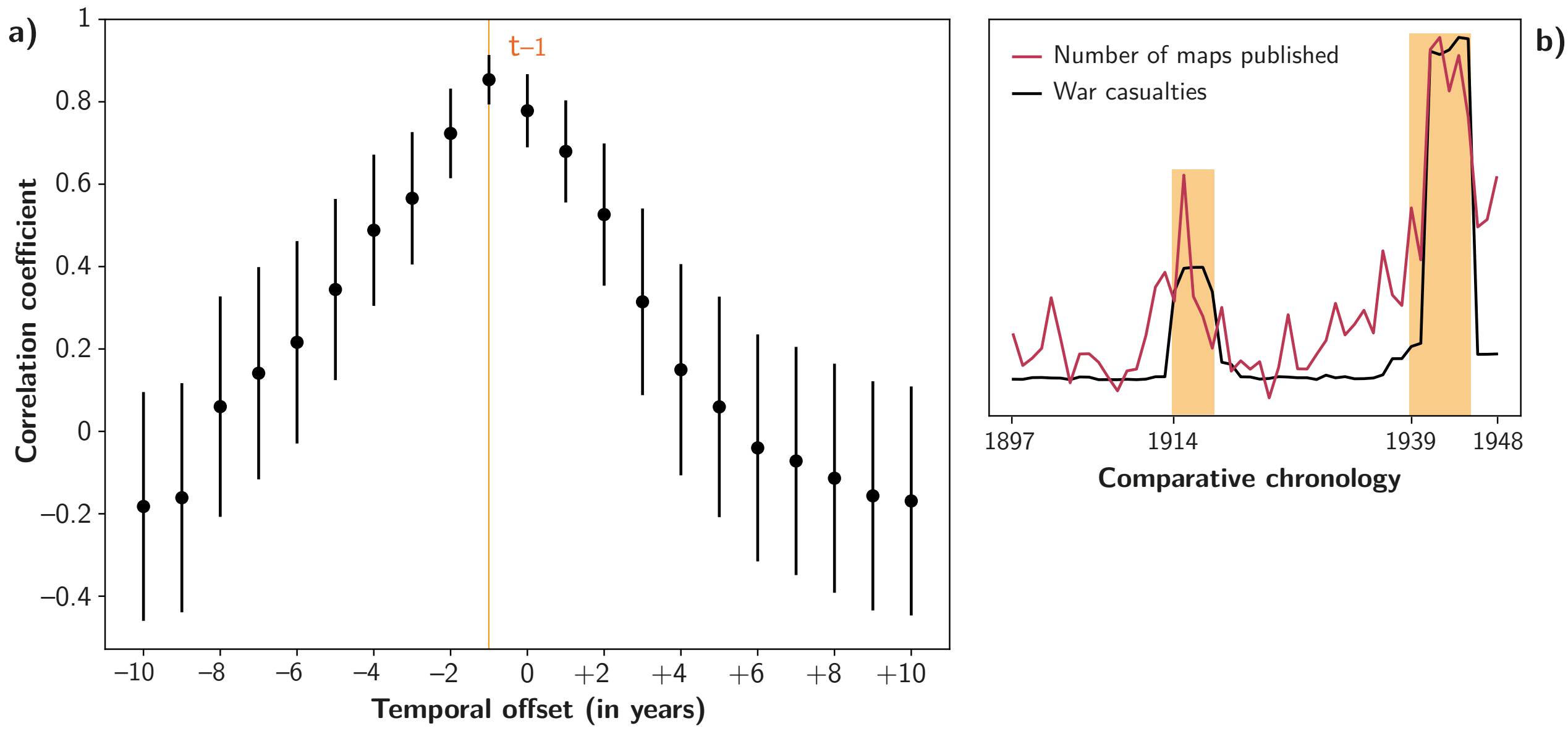

**Figure 10 | Occurrent dependency between map production and the intensity of 20th century global military conflicts.** Plot (a) depicts the temporal of Pearson's correlation between total map production and conflict intensity during the period 1897–1948. The number of war casualties worldwide serves as a proxy for conflict intensity. Horizontal bars represent the 95% confidence interval of the correlation coefficient, whereas the long yellow bar marks its maximum value. Statistics on war casualties (average estimates) are sources from Jason Lyall's Project Mars (Lyall, 2022). Both time series are also displayed in subfigure (b). There, highlighted areas denote the years of active conflict during WW1 (1914–1918) and WW2 (1939–1945). *Map publication volumes are highly correlated to the intensity of military conflicts during the period.*

## 3.6 Conclusion

This chapter engaged with the manifestations and relations of power in cartography. First, it introduced a new concept, that of spatial attention, which views cartographic coverage as a directed and purposeful process, that demands active, sustained, and effortful engagement. This notion adds an additional dimension to the analysis of the geographical focus of maps, enabling the conceptualization of qualities such as self-attention and domestic map production. The second part of the chapter suggested the existence of a feedback mechanism between Atlantic charting and the triangular trade, while the final experiment highlighted the relationship between cartography and military conflicts. Through these distinct perspectives, the chapter advanced an understanding of the power of maps and underscored the relevance, for empirical research, to consider mapping a process of power-knowledge. This conception is already acknowledged by scholars in map studies and the history of cartography, where it is often employed as a framework for interpreting historical maps, although it has rarely been explored with statistical tools.

This third chapter concludes the first part of the dissertation. Chapter 1 focused on the creation of two large, aggregated databases on the history of cartography: ADHOC Records and ADHOC Images. Chapter 2 analyzed the geographical and chronological dimensions of map publication, from the 15th to the 20th century and examined the social structure of map making through the social graph of mapmakers' collaborations. Accordingly, the analyses in this first part of the thesis rested primarily on publication volumes and metadata, although specific examples were also presented to illustrate the discourse.

The second part of this dissertation will focus on images rather than records metadata. This shift will enable a more direct discussion of maps as documents and designed representations, and the analysis of map figuration. Although they address image data instead of records, the next two chapters will extend the perspective advanced here, particularly the idea that maps tend to reproduce cultural conventions, through the examination of map content and map makers' choices. Chapter 4 will extract and study the classes of geographic objects represented, whereas Chapter 5 will investigate the composition of cartographic representations, and the way it is used to emphasize certain geographic objects at the expense of others.

# Part II

# Images.

# Chapter 4

## *On Semantic segmentation, or (Un)drawing Geography*

This chapter will mainly focus on recognizing geographic classes, such as built-up areas, water and roads, from maps in the ADHOC Images dataset. The last section will also investigate the historical distribution of these classes, by scale, place, and year of publication. Semantic segmentation is a machine learning task that aims to assign a semantic label to each pixel in an image. The process can also be described as a pixel classification issue. In the present study, semantic segmentation is employed to partition the geographic content of the map into four elementary geographic classes. This step is a prerequisite that will enable the computational investigation of map content in the following chapters. Specifically, Chapter 5, will rely on semantic segmentation masks to study how maps frame the territory, and identify several semantic modes (or cartographic "themes"). Chapter 7 will rely on semantic masks to model map semiotics.

### 4.1 Related work

The last decade has witnessed rapid progress in historical map recognition, driven by convolutional and Transformer-based neural networks. In a seminal work, Uhl et al. (2017) introduced a patch-based classifier. However, the first application of semantic segmentation to maps was presented by Oliveira et al. (2019), who demonstrated the efficacy of UNet (Ronneberger et al., 2015) architectures and ResNet (He et al., 2015) encoders to segment historical cadastres. Petitpierre (2020) extended the approach to medium-scale city maps and investigated its applicability to diverse map collections. In parallel, Heitzler and Hurni (2019, 2020) evidenced the scalability of CNN-based methods, as well as their relevance to process even smaller-scale maps, by extracting thousands of buildings and city blocks from the Siegfried Swiss topographic map series.

At the same time, Chiang et al. (2020) addressed the extraction of railroads from the USGS map series. Because the reliable and uninterrupted detection of linear features poses a significant challenge, roads quickly emerged as a primary extraction target (Can et al., 2021; Ekim et al., 2021; Uhl et al., 2022). Jiao et al. (2021) presented a comprehensive survey of road extraction methods. While Hosseini et al. (2022) recommended reverting to coarser, yet more reliable, patch-based classifiers, Jiao et al. introduced pioneering approaches that rely on synthesized training samples for segmenting roads in the Siegfried maps (Jiao et al., 2022a, 2022b). To address the issue of line discontinuity, Xia et al. (2024b) proposed an end-to-end vector framework that implements Transformer-based line segment detection (Xu et al., 2021).

In urban historical context, segmentation has proven to be useful for extracting building footprints and identifying urban extents (Uhl et al., 2021). Chazalon et al. (2021), and Chen et al. (2021) concentrated on closed shape extraction from the Paris atlases. By contrast, Petitpierre et al. (2023, 2024a) focused on the extraction of distinct land classes from Swiss Napoleonic cadastral plots, using HRNet, a high-resolution CNN-based model for edge filtering (J. Wang et al., 2021), and OCRNet, a Transformer-based model for contextual segmentation (Yuan et al., 2020). Other researches similarly attempted to extract building footprints from Franciscean cadastres with CNNs (Göderle et al., 2023; Göderle et al., 2024; Rampetsreiter et al., 2023). Vaienti et al. (2023) demonstrated the possibility to use semantically segmented footprints for 4D historical city reconstruction. These technologies have also attracted interest in environmental history, for the study of land use change (Levin et al., 2025; Mäyrä et al., 2023), wetland monitoring (Vynikal et al., 2024), and surface mine mapping (Maxwell et al., 2020).

Several recent studies on map recognition still rely on CNN-based frameworks (Martinez et al., 2023). In a benchmark paper, Chen et al. (2024) advanced that Vision Transformers (Dosovitskiy et al., 2021) are unable to produce reliable segmentation masks. Similarly, S. Wu et al. (2023) have recognized the inferior performance of Segformer (Xie et al., 2021) relative to UNet. However, other context-aware Transformer-based architectures, like OCRNet (Jan, 2022; Petitpierre et al., 2023), have achieved performance gains compared to convolutional UNets. Xia et al. (2023) retained a UNet backbone, but demonstrate the potential of Swin encoders (Liu et al., 2021) to address the complex task of railways segmentation. S. Wu et al. (2023) proposed a multi-head cross-attention module that leverages spatio-temporal cues to segment aligned maps, treating the extraction of map series as a joint segmentation task. In a similar attempt to exploit spatial context, Arzoumanidis et al. (2023, 2025) introduced a self-constructing graph convolutional networks (SCGCN), reporting encouraging results. Other promising approaches include contrastive learning (Xia et al., 2023) and the adaptation of large vision models such as Segment Anything (Kirillov et al., 2023; Xia et al., 2024a).

The performance of semantic segmentation models is largely contingent on the quality of the training data (Petitpierre & Guhennec, 2023). To alleviate this time-consuming task, some scholars proposed to propagate semantic labels across time and throughout map series (Uhl et al., 2020; Yuan et al., 2025). By contrast, several studies have highlighted the potential of synthetic data pretraining, with the aim to focus the model's robustness in the face of diverse geographic configuration, or graphic noise. This research avenue has yielded promising results, for instance, in road segmentation (Jiao et al., 2022a; Mühlematter et al., 2024; Zhao et al., 2024), and map text detection (Z. Li, 2019; Z. Li et al., 2021; Zou et al., 2025b). Similar research has shown that adversarial generative models can enrich the diversity of map samples through style transfer (Arzoumanidis et al., 2024; Christophe et al., 2022). Nevertheless, they also highlighted the difficulty of generating credible synthetic historical data directly from GIS-derived information or segmentation masks (Kang et al., 2019).

## 4.2 Training and validation data

Thus, one of the principal obstacles to the development of semantic segmentation technologies for historical maps is the scarcity of ground-truth labels with which to train models. Whereas manually *vectorized* historical maps are relatively common, the process of annotating for *semantic segmentation* is distinct. The purpose differs substantially. Semantic segmentation (Long et al., 2015) aims to assign each pixel in an image a defined semantic label (e.g., water). Vectorization, by contrast, entails redrawing the document's geometries in a digital format, typically compatible with geographic information systems (GIS). Although geometries may subsequently be semanticized by assigning them specific labels, vectorization normalizes the geometries by simplifying them into assumed shapes (e.g., rectangles) or by interpolating them from a set of representative points. The fine correspondence between pixels and labels is therefore lost, making it difficult to repurpose these data for training segmentation models. Consequently, research generally relies on dedicated datasets.

### *Semap Dataset, a new annotated dataset for generic map recognition*

At present, one of the main open-source dataset for semantic segmentation is the Historical City Maps Semantic Segmentation Dataset (HCMSSD, Petitpierre, 2021). This dataset, which exhibits a variety of content and style, was compiled as part of my master's thesis (Petitpierre, 2020). It contains 330 annotated map patches extracted from a corpus of maps of Paris curated by the French National Library (BnF) and the Historical Library of the City of Paris (BHVP). It also comprises 305 city map extracts, depicting 182 distinct cities in 90 countries, collected from 32 heritage institutions and digital libraries, many of which were also considered for the creation of ADHOC Images. Each patch measures $1{,}000 \times 1{,}000$ pixels. The annotations encompass five semantic classes; the first four pertain to geographic content: *built* (e.g., buildings, walls, porticoes), *road network* (e.g., streets, squares, walkways, bridges), *water* (e.g., rivers, seas, lakes, canals), and

*non-built*—which covers agricultural and natural land as well as green spaces (e.g., parks, gardens). The fifth class, *background*, was defined explicitly; it captures map components that are not directly geographic, such as document borders, legends, ornaments (e.g., cartouches), and peripheral layout components (e.g., compasses, graphical scales).

The present research addresses an even greater diversity of maps, both in terms of content and in scale. More recent, larger neural models further require larger training samples. Thus, a new dataset, the Semantic Segmentation Map Dataset (Semap), consisting of 1,439 map samples, was compiled. Semap is a composite dataset balanced to match ADHOC Images while capitalizing on existing annotated data. Aside a curated subset of 356 patches from the HCMSSD, it also includes 78 samples extracted from Napoleonic or renovated cadastres (di Lenardo et al., 2021; S. Li et al., 2024; Petitpierre, 2023) and three from Paris city atlases (Chazalon et al., 2021). These data are augmented with an additional 1,002 newly annotated samples of dimension $768 \times 768$ pixels[1], drawn from ADHOC Images. The overall composition of Semap is detailed in Table 1.

The addition of the contour class enhances the descriptive power of semantic labels by integrating the notion of object, or instance. With this class, one can differentiate, for example, a plan depicting building footprints from one showing undivided urban blocks. The contour class also encodes the number of objects in the image and, hence, the semantic density of the map. Introducing this fifth class required updating all samples from HCMSSD and related datasets to include the supplementary label. In total, the annotation process demanded approximately 400 hours of manual work[2]. Sample annotations from the resulting Semap dataset are shown in Figure 1.

[1] The choice of this lower resolution is determined by the architecture of the segmentation model, and memory constraints. Besides, it enables a slightly shorter annotation time per sample, thereby allowing more samples to be annotated and a greater variety of maps to be included in the training set.

[2] From which 325 h performed by me, and 75 h carried out by paid master-level student assistants enrolled at EPFL.

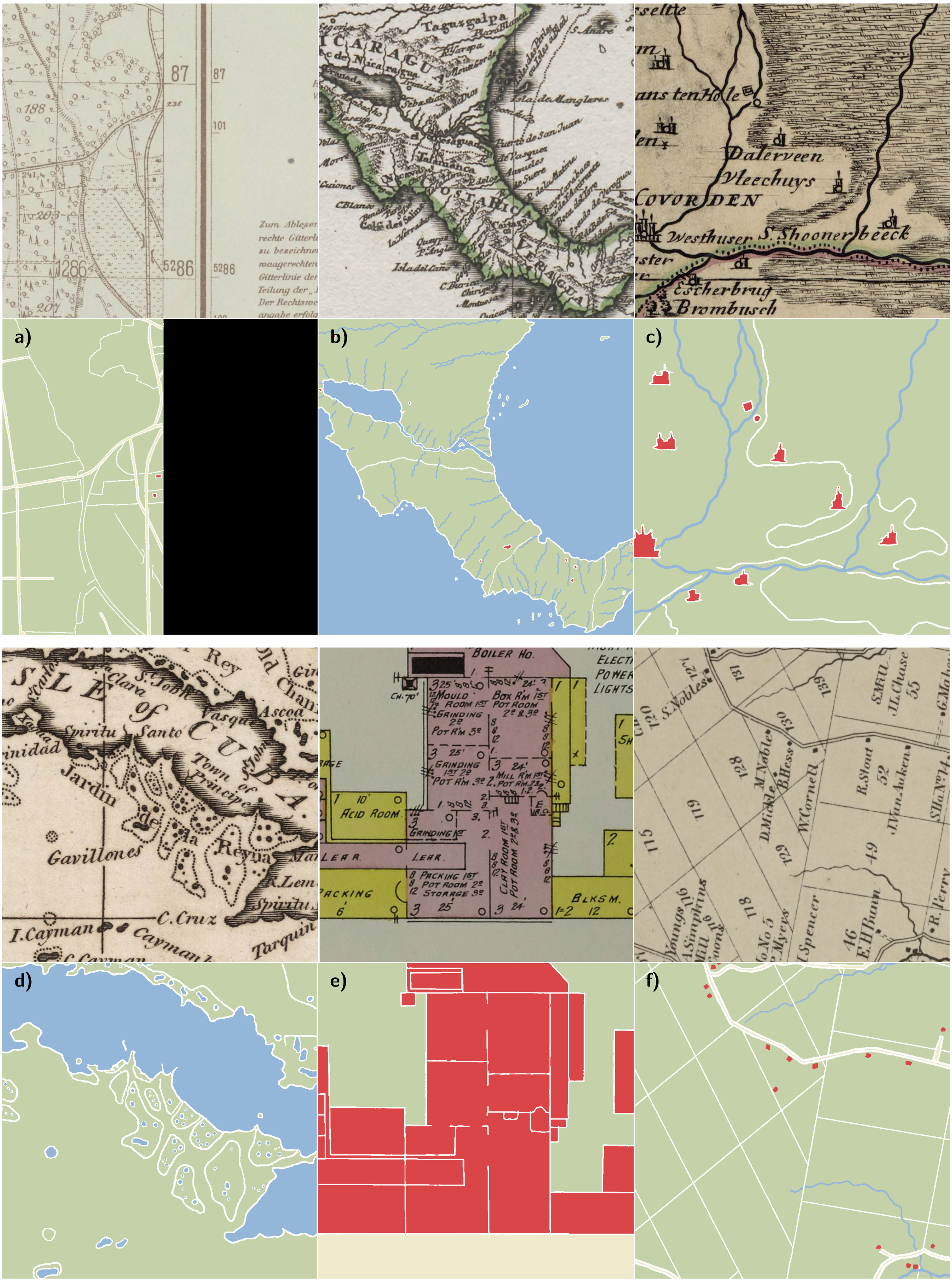

**Figure 1 | Manually annotated training samples.** Real map crops (top) with corresponding labels (bottom). The label color code follows the legend of Fig. 2.

**Table 1 | Composition of Semap dataset.**

| Institution(s) | Part of | Landscape | Scale | Period | # samples |
|---|---|---|---|---|---|
| Vaud Cantonal Archives | Lausanne cadastre | Urban, periurban | very high | 1722–1882 | 42 |
| Neuchatel Cadastre Office | Neuchatel cadastre | Urban, periurban | very high | 1886–1888 | 10 |
| Geneva Cadastre Office | Geneva cadastre | Urban, periurban | very high | 1845–1856 | 8 |
| French Dept. Archives & others | Napoleonic cadastre | Urban, rural | very high | 1808–1888 | 18 |
| BHVP | Paris Atlas | Urban | very high | 1898–1926 | 3 |
| BHVP | HCMSSD – Paris | Urban | very high | 1785–1950 | 70 |
| BnF | HCMSSD – Paris | Urban | high | 1765–1949 | 189 |
| (32 institutions) | HCMSSD – World | Urban, periurban | high | 1701–1949 | 97 |
| Berkeley library | Semap | diverse | low–high | 1657–1946 | 8 |
| BnF | Semap | diverse | low–high | 1494–1947 | 262 |
| U Bordeaux-Montaigne | Semap | diverse | low–high | 1676–1946 | 11 |
| Boston Leventhal Map Centre | Semap | diverse | low–very high | 1493–1947 | 17 |
| David Rumsey | Semap | diverse | low–high | 1525–1947 | 355 |
| e-rara | Semap | diverse | low–high | 1513–1947 | 47 |
| Library of Congress | Semap | diverse | low–very high | 1507–1947 | 165 |
| NY Public Library | Semap | diverse | low–very high | 1542–1947 | 44 |
| Princeton Library | Semap | diverse | low–high | 1509–1947 | 28 |
| U Leiden | Semap | diverse | low–high | 1545–1947 | 40 |
| WWU Munster | Semap | diverse | low–high | 1550–1938 | 12 |
| (5 other institutions) | Semap | diverse | low–high | 1492–1947 | 13 |
| **(50+ institutions)** | **Semap** | **diverse** | **low–very high** | **1492–1950** | **1439** |

### *Synthetic data generation*

In addition to manually labeled data, the training set includes synthetically generated data. Here, synthetically generated data are derived from contemporary reference geodata. In that prospect, the spatial distribution—geographic coordinates of centers—and the scale distribution of ADHOC Images are used to constrain contemporary geodata sampling. The map scale $s$, expressed as a fraction (e.g., 1:20,000), is converted to an approximate zoom level $z$ by means of the following formula, calibrated to reproduce the visual rendering of the archival maps[3]:

$$z = \alpha - \log_2(1/s) \quad z \in \mathbb{N}, \qquad \alpha = 28.1$$

The geographic bounding box is calculated as a function of the zoom level and geographic coordinates, using WGS84 pseudo-Mercator projection (EPSG:3857) and a resolution of 300 dpi, resulting in a final target dimension of 768 × 768 pixels.

[3] This implies not only accounting for average scanning resolution, but also for instance compensate for the smaller area of historical buildings, or the lesser width of historical roads, compared to modern ones.

Then, the corresponding vector tiles are queried from MapTiler Planet (*MapTiler Planet*, n.d.) through the public application programming interface (API). The complete set of retrieved objects is listed in Table A3, in Appendix. With respect to object sizes, the objective is not to reproduce historical data with factual precision, but rather to approximate their visual appearance. For instance, airport names constitute anachronistic data points; nonetheless, they may be useful for labeling larger cities with a distinctive typography. In a similar perspective, motorways can be employed to differentiate major roads. Few motorways existed before 1947 and even fewer historical roads resembled motorways in terms of both layout and trajectory. However, on smaller-scale maps, motorways may serve as a plausible proxy for historical national road networks. Using this approach, the selection of contemporary features was refined iteratively, based on qualitative assessments, informed by the specific expertise of content and stylistic conventions observed in ADHOC corpus of historical maps.

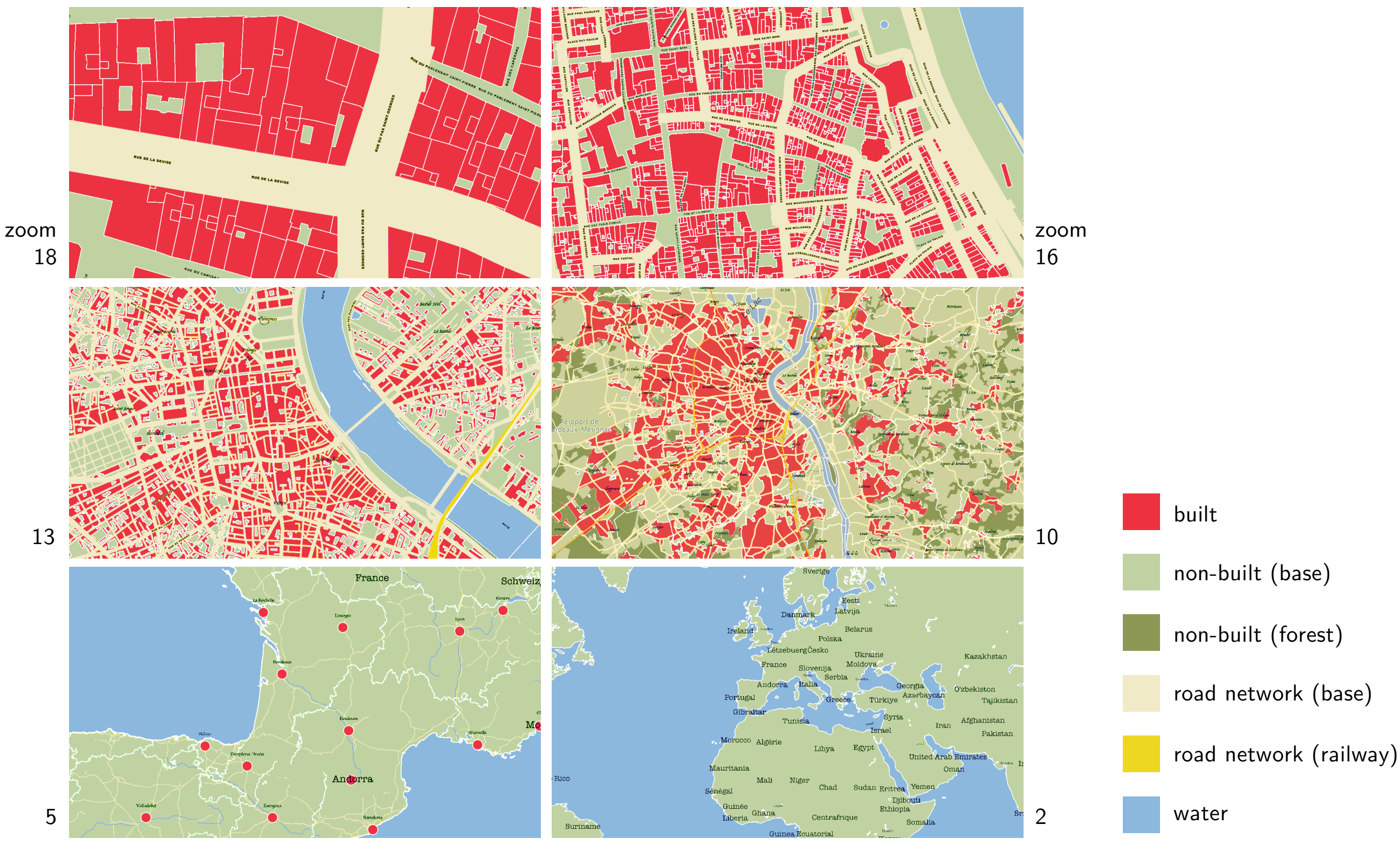

**Figure 2 | Objects queried from MapTiler API displayed as a function of the zoom level.**

In addition to restricting the retrieval of certain features to specific zoom levels, some components, like forests, were randomly hidden to increase the diversity of landscapes. Indeed, the set of classes retrieved from MapTiler database exceeds the number of semantic classes defined in the segmentation ontology. It includes, for instance, forests, and railroads. The objective is thus to augment diversity and train the network to expect semantic classes (such as non-built) to be subdivided into several subclasses (e.g., agricultural land and forest), each adopting a distinct graphical form. Figure 2 portrays examples of retrieved data, as a function of the zoom level.

A total of 12,122 samples were collected from contemporary data. These template samples were then converted into raster format, in which each pixel corresponds to a class or subclass. A stylization process is subsequently employed to generate visually plausible synthetic maps (Figs. 3 and 4). Stylization entails the probabilistic application of graphical processes to the template samples. Here, a wide variety of generative drawing functions contributed to the production of diverse synthetic maps. Parameters of the stylization algorithm—e.g., color distributions and the relative frequency of graphical processes—were initially derived from Semap data, and iteratively adjusted, based on qualitative assessments, until a visually satisfactory result was attained.

Graphical processes included:

- *plain color* (e.g., Fig. 3a, 4e)
- regular or random *dotted patterns*, with various size and density (e.g., Fig. 3e, 4f)
- horizonal or oblique *hatchings*, with random orientation, width, and spacing (e.g., Fig. 4b, 4c)
- *waterlines*, with various width and spacing function (e.g., Fig. 3d)
- application of *texture masks* from a library of 99 patterns extracted from Semap, extended through random scaling and flipping.
- application of *icon sprites*, introduced in Chapter 5 (e.g., Fig. 4f)

Most graphical processes involved the application of color (e.g. ink) to the image. Colors—including those of ink and paper—are sampled, on a class-by-class basis, from a Gaussian mixture derived from Semap images and class labels. To increase color variability, darker spots were also applied to the canvas. Rescaling and anti-aliasing are employed to reproduce natural color interpolation. In 15% of the cases, the image was converted to grayscale (e.g., Fig. 3b, 3f). In addition, images were saved in JPEG format to simulate compression artifacts that are recurring defects of authentic digitized map images.

The stylization algorithm also approximates the representation of relief. It is desirable for the semantic segmentation algorithm to learn such information, as relief is a recurrent contextual cue which might help recognizing certain geographical features—for instance, rivers in hilly landscapes. The stylization algorithm approximates three common orographic processes: hatching (e.g., Fig.

4b), hillshading (e.g., Fig. 3c), and *isolines*[4] (e.g., Fig. 4f). Elevation data were retrieved through the Mapbox TerrainV2 API (*Mapbox Terrain v2 | Tilesets | Mapbox Docs*, n.d.). In addition, the image was sometimes purposely cropped along one edge or at a corner to imitate the map frame (e.g., Fig. 3b). The thicknesses of linear features (e.g., watercourses or the road network) and contour lines was partly randomized to increase representational diversity. To prevent the segmentation model from overfitting to the relative positions of text labels and to enhance typographic diversity, additional text labels sampled from the repertoire of place names in the ADHOC database were sometimes randomly inked (e.g., Fig. 3c–3e, 4e). Finally, an artificial graticule was occasionally imprinted on the image, in the form of horizontal or vertical lines that intersect to imitate a map grid (e.g., Fig. 3f).

Table 2 below presents the resulting class–area distribution of Semap annotated and synthetically generated datasets. The built and road network classes seem visibly more represented in annotated map data compared to GIS–derived synthetic data. By contrast, non-built and water classes are under-represented in real historical map samples. These ratios will be commented later on, and related to the distributions inferred from the ADHOC Images.

**Table 2 | Relative class–area distribution in the Semap and synthetic training datasets**, expressed as a percentage of total pixel area.

| | background | contours | built | non-built | water | road network |
|---|---|---|---|---|---|---|
| Semap | 36.6% | 3.8% | 11.7% | 31.8% | 7.9% | 8.2% |
| Synthetic training data | 7.9% | 4.3% | 2.8% | 72.9% | 9.8% | 2.3% |

[4] More commonly called *contour lines*, not to be confused with the semantic class we called *contours*.

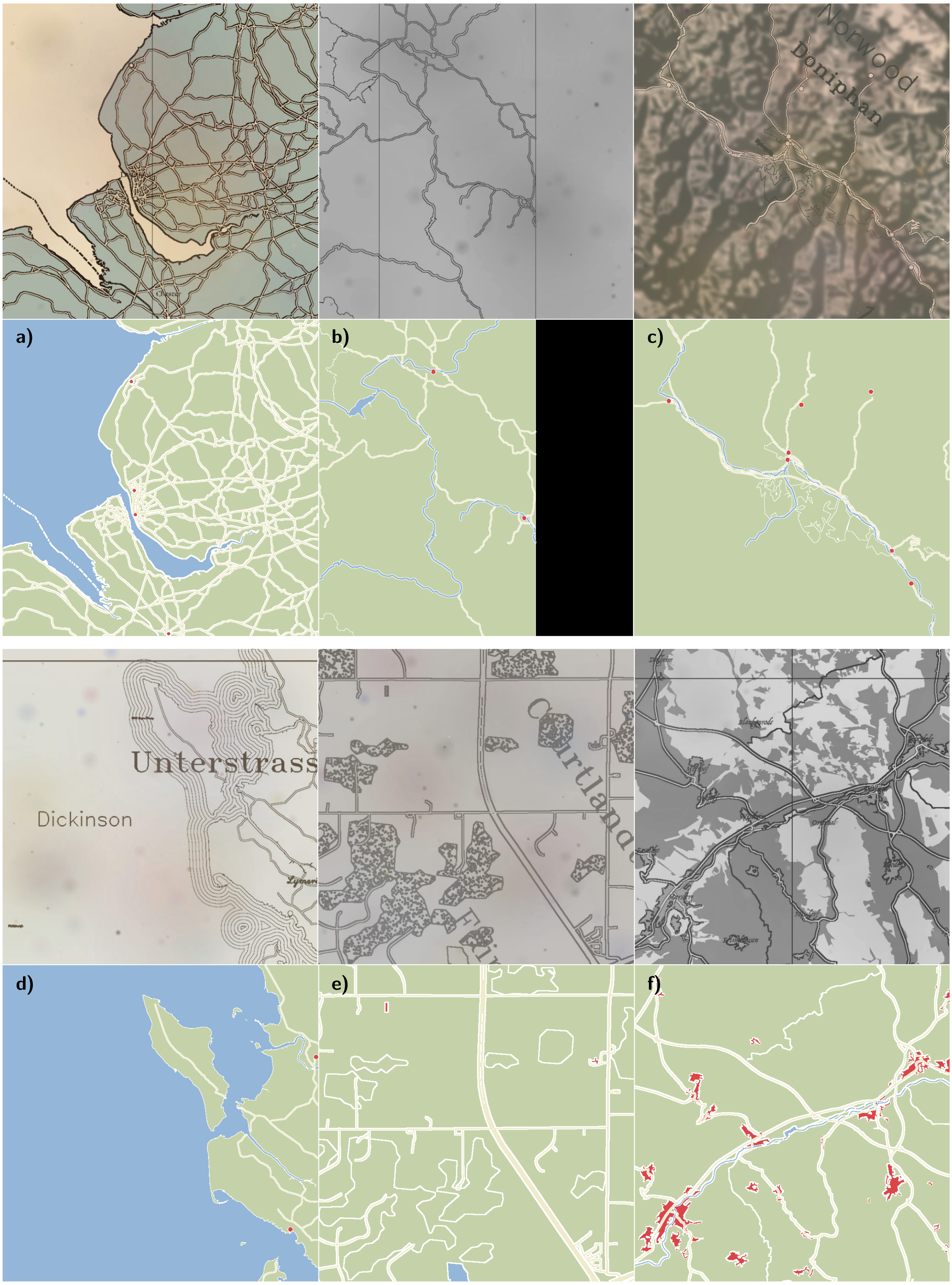

**Figure 3 | Synthetically generated training samples.** Synthetic map images (top) and corresponding labels (bottom). The label color code follows the legend of Fig. 2.

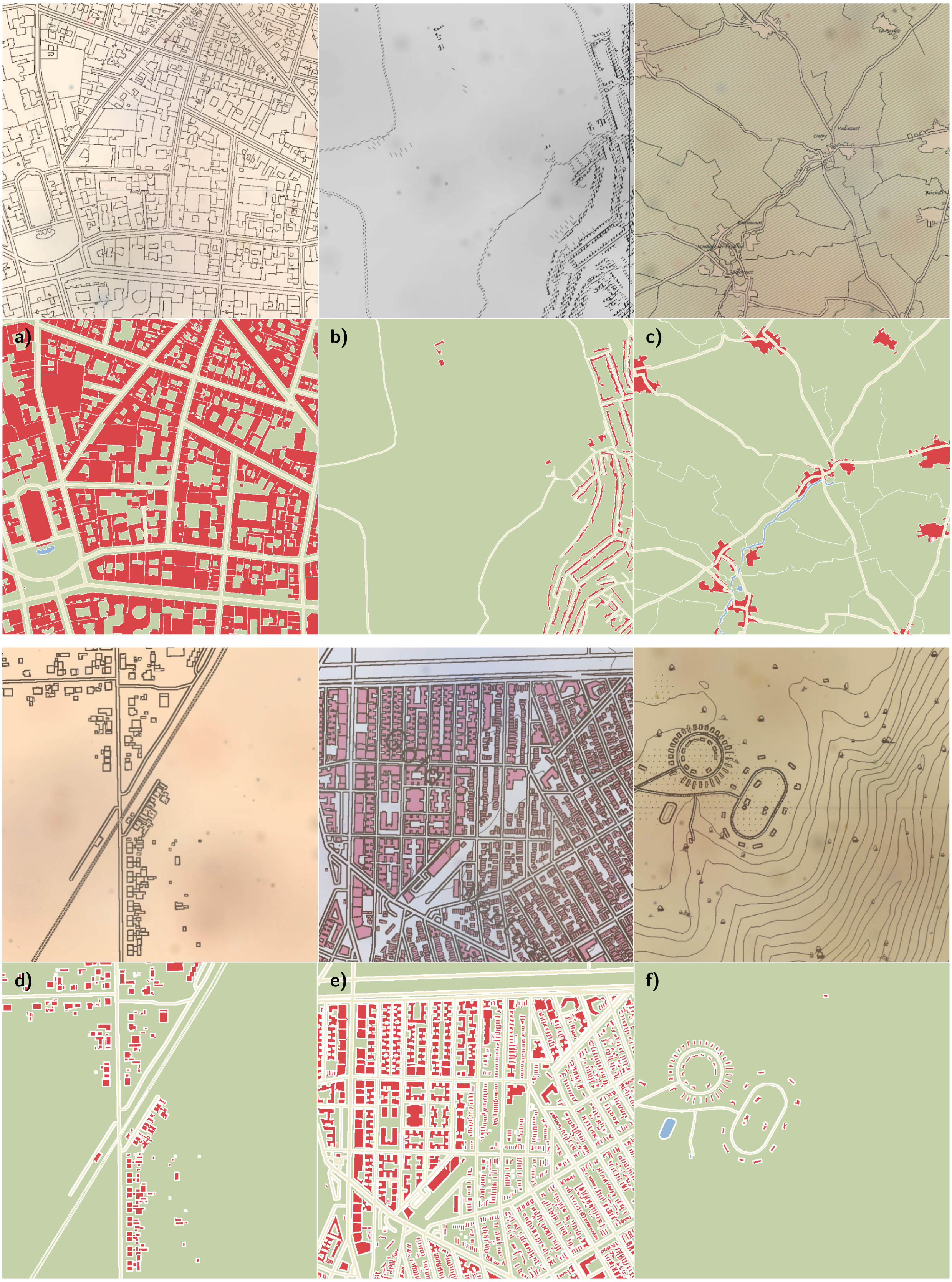

**Figure 4 | Synthetically generated training samples (continued).** Synthetic map images (top) and corresponding labels (bottom). The label color code follows the legend of Fig. 2.

## 4.3 Segmentation approach

### *Choice of model architecture*

We rely on Mask2Former, a masked-attention Mask Transformer (Cheng et al., 2022) with a Swin-L backbone (Liu et al., 2021) for semantic segmentation. Swin is a Transformer-based encoder (Vaswani et al., 2017) specifically designed to generate hierarchical feature representations. The ability of Swin-Transformers to accommodate multiscale image objects is crucial in the context of map recognition. At each downscaling step, the Swin encoder concentrates on increasingly local and granular windows of the input image. These windows do not overlap. Yet, each subsequent layer shifts the grid to mitigate border effect between patches. In addition, Swin implements a relative positional embedding mechanism. Unlike original Transformers, Swin Transformers rely on local self-attention, computed within each window. This strategy facilitates high-resolution feature representations on large images[5].

Mask Transformer models output a list of binary masks, one per object, together with their corresponding semantic labels. To predict the masks, the pixel decoder progressively upsamples the encoder output and injects the resulting feature maps into the Transformer decoder after each stage. This procedure somehow mirrors the Swin logic, enabling the integration of both highly contextual information from the latent representation and high-resolution cues from pixel-level features. Masked attention, introduced in Mask2Former (Cheng et al., 2022), accelerates convergence by restricting the attention operation to the foreground region of each detected object.

### *Training approach*

The Semap dataset was split into three partitions (70/20/10 %) for training, validation, and testing, respectively. The synthetical data were employed exclusively for training (80 %) and validation (20 %). Under this configuration, synthetic samples constitute 90.9 % of the training set. Because synthetic data already provide diverse inputs, extensive data augmentation is deemed unnecessary. Samples are thus merely augmented through random horizontal flipping.

The present study relies on the implementation of Mask2Former by MMsegmentation (*MMSegmentation*, 2022). The model was trained for 345,000 iterations with a batch size of 1 and a multiscale composite loss. During the first 205,000 iterations, the model was trained on both Semap and synthetic data. It was then tuned on Semap only for the remaining 140,000 iterations.

[5] Global self-attention computes the relationships between every pair of image tokens. Local self-attention, in the contrary, is restricted to the image tokens located in the same local window. Long-range dependencies are still accounted for by the hierarchical design of Swin Transformers, since in the first block, the local window effectively encompasses the entire image, albeit with a lower granularity.

At each stage, only the model that achieved the highest validation performance was retained (early stopping policy).

The loss function combines binary cross-entropy (BCE) for mask prediction, cross-entropy[6] (CE) for class attribution, and Dice loss (Milletari et al., 2016) to mitigate the issue of class imbalance. A lower weight (0.6) is assigned to the *contours* class as its relative difficulty could otherwise hinder training at the expense of the other classes, which are given higher priority in this research. The training relies on the AdamW optimizer (Loshchilov & Hutter, 2018) with a polynomial decay learning rate schedule.

## *Inference strategy & hierarchical integration*

Since physical map documents are often quite large, the corresponding digital images can exceed 10,000 × 10,000, or even 20,000 × 20,000 pixels. Consequently, each image had to be partitioned into smaller 768 × 768 pixel patches—the dimension on which the model was trained. Here, 64-pixel overlapping windows were employed to mitigate border effects arising from the disruption of context at patch boundaries.

Moreover, while the selected architecture already favors hierarchical image modeling, map images are so fundamentally multiscale that inference strategy is also designed to address this aspect. Specifically, the model is applied at two image scales and resolutions: first at the original resolution and then at half that resolution. The principal objective is to improve the recognition of large objects spanning several patches. Intuitively, this strategy may also enhance segmentation quality by enabling consensus predictions at a reduced computational cost, owing to the decreased spatial resolution. In the end, the raw logits from overlapping and multiscale prediction maps are averaged, and each pixel is assigned to the semantic class associated with the highest logit.

## *Performance evaluation*

Let $N$ be the number of test images and $K = 6$ the number of semantic classes. We write $Y_{i,k}$ the set of pixels $p$ in image $i$ that belongs to class $k$, and $\hat{Y}_{i,k}$ the set of pixels in image $i$ whose predicted class is $k$. We also define $\cup_{i,k} = Y_{i,k} \cup \hat{Y}_{i,k}$, the union of $Y_{i,k}$ and $\hat{Y}_{i,k}$, and $\cap_{i,k} = Y_{i,k} \cap \hat{Y}_{i,k}$ the intersection.

For every class and image, we compute the precision, recall, and the intersection over union (IoU):

$$\mathrm{Prec}_{i,k} = \frac{\sum_{\forall p} \cap_{i,k}}{\sum_{\forall p} \hat{Y}_{i,k}} \qquad \mathrm{Rec}_{i,k} = \frac{\sum_{\forall p} \cap_{i,k}}{\sum_{\forall p} Y_{i,k}} \qquad \mathrm{IoU}_{i,k} = \frac{\sum_{\forall p} \cap_{i,k}}{\sum_{\forall p} \cup_{i,k}}$$

[6] Cross-entropy $H$ is defined as $H = -\log p(c)$, where $p(c)$ is, for a given pixel, the predicted probability for the correct class $c$.

F1 scores are computed as the average between precision and recall. The contribution of each sample image $i$ to the class averages is then normalized with respect to the ground-truth count $\sum_{\forall p} Y_{i,k} Y_{i,k} \sum_{\forall p} Y_{i,k}$. This procedure ensures, for example, that an image containing only a small house does not influence the *built* class average as much as a map predominantly depicting an urban environment. The quantitative impact of this choice is detailed in Table A2 in the Appendix. As expected, normalization helps stabilize the results for heterogeneous datasets, such as Semap and HCMSSD–World. For each metric $M_k \in \{Prec_k, Rec_k, IoU_k\}$, we thus have:

$$M_k = \frac{\sum_{i=1}^{N} M_{i,k} \sum_{\forall p} Y_{i,k}}{\sum_{i=1}^{N} \sum_{\forall p} Y_{i,k}}$$

Global performance metrics are calculated as the mean of the four geographic classes—*built*, *non-built*, *water*, and *road network*. For the performance benchmark on the HCMSSD–Paris and HCMSSD–World datasets (Tabs. 4 and 5), the standard macro-average (per class) is used, to ensure comparability (see formula in Tab. A1, Appendix).

## 4.4 Results of segmentation

### *Overall performance*

**Performance per class.** Table 3 reports the semantic segmentation performance on Semap test set. The model performs best on the built and non-built classes, with IoU values of 79.8% and 81.8%, respectively. Water is also well segmented, achieving 72.2% IoU. The least accurately recognized geographic class is the road network, at 62.9% IoU. Contours, which were assigned a lower weight (0.4) during training, exhibit only 40.7% IoU, whereas the map background is well segmented, at 76.8%. Overall, the model shows high precision, ranging from 80.7% to 92.4% according to the geographic classes, with recall values between 72.7% and 86.0%. The mean performance is strong, with a mean IoU (mIoU) of 74.2%, a mean recall of 79.4%, and a mean precision of 85.4% across geographic classes.

**Table 3 | Performance of semantic segmentation on Semap dataset** ($n_{test}$=144). IoU denotes the Intersection over Union. The mean is computed across the four geographic classes (*).

| | **Test set (base)** | | | **Test set (no multiscale)** | | | **Test set (no synth. pretrain)** | | |
|---|---|---|---|---|---|---|---|---|---|
| class | IoU | Recall | Precision | IoU | Recall | Precision | IoU | Recall | Precision |
| background | 76.8 | 78.2 | 90.5 | 78.7 | 81.0 | 87.9 | 86.3 | 89.4 | 91.1 |
| contours | 40.7 | 59.0 | 56.2 | 41.9 | 55.9 | 60.6 | 39.7 | 58.2 | 55.4 |
| built* | 79.8 | 84.4 | 92.4 | 70.8 | 75.5 | 89.4 | 74.6 | 84.1 | 87.7 |
| non-built* | 81.8 | 86.0 | 87.2 | 79.3 | 84.4 | 85.1 | 78.4 | 83.0 | 87.0 |
| water* | 72.2 | 74.4 | 81.2 | 68.4 | 70.6 | 73.6 | 67.0 | 68.6 | 80.3 |
| road network* | 62.9 | 72.7 | 80.7 | 62.0 | 71.4 | 77.8 | 56.3 | 65.7 | 81.7 |
| ***mean** | **74.2** | **79.4** | **85.4** | **70.1** | **75.5** | **81.5** | **69.1** | **75.4** | **84.2** |

**Table 4 | Performance benchmark on HCMSSD–Paris dataset (validation set)**. mIoU denotes the average intersection over union of the four geographic classes (built, non-built, water, road network). mR = mean recall, mP = mean precision, F1 = (mP+mR)/2. Note that (Polák, 2024) aggregates both HCMSS–Paris and HCMSSD–World datasets. The multiscale strategy is not employed here.

| Reference | Model (*training*) | mIoU | mR | mP | F1 |
|---|---|---|---|---|---|
| (Polák, 2024) | UNet–Transformer | 28.1 | 43.9 | 45.2 | 44.6 |
| (Petitpierre et al., 2021) | UNet–ResNet101 | 54.3 | 77.3 | 66.5 | 71.9 |
| (Arzoumanidis et al., 2025) | SCGCN–ResNet101 | 63.5 | 82.7 | 74.2 | 74.2 |
| (Jan, 2022) | HRNet–OCRNet | 66.2 | – | – | – |
| Ours | Mask2Former–Swin-L (*"few-shots"*) | 71.5 | 77.9 | **89.1** | 83.5 |
| Ours | Mask2Former–Swin-L (*transfer*) | **76.0** | 83.8 | 88.2 | 86.0 |
| Ours | Mask2Former–Swin-L (*retrained*) | **76.0** | **84.2** | 88.6 | **86.4** |

**Table 5 | Performance benchmark on HCMSSD–World dataset (validation set)**. mIoU denotes the average intersection over union of the four geographic classes (built, non-built, water, road network). mR = mean recall, mP = mean precision, F1 = (mP+mR)/2. Note that (Polák, 2024) aggregates both HCMSS–Paris and HCMSSD–World datasets. The multiscale strategy is not employed here.

| Reference | Model (*training*) | mIoU | mR | mP | F1 |
|---|---|---|---|---|---|
| (Polák, 2024) | UNet–Transformer | 28.1 | 43.9 | 45.2 | 44.6 |
| (Petitpierre et al., 2021) | UNet–ResNet101 | 45.2 | 64.9 | 62.6 | 63.8 |
| Ours | Mask2Former–Swin-L (*"few-shots"*) | 74.4 | 81.7 | 88.6 | 85.2 |
| Ours | Mask2Former–Swin-L (*transfer*) | **76.3** | **88.4** | **92.6** | **90.5** |
| Ours | Mask2Former–Swin-L (*retrained*) | 74.2 | 86.8 | 84.1 | 85.4 |

**Comparison benchmark.** Tables 4 and 5 show that the Mask2Former–Swin-L architecture surpasses previous state-of-the art by a clear margin on both HCMSSD–Paris and HCMSSD–World benchmarks. It achieves markedly higher scores than UNet-based architectures which are still commonly used in map recognition tasks. For instance, on the Paris benchmark, the mIoU improves by 22 percentage points (pp) relative to the UNet-ResNet101 employed by (Petitpierre et al., 2021). On the World benchmark, the gain reaches 31 pp. Mask2Former with Swin encoder also outperforms recent graph neural network approaches, such as SCGCN (Arzoumanidis et al., 2025), raising the overall precision by 14 pp, while maintaining a comparable recall (+ 1 pp). It also exceeds the performance of the HRNet architecture with OCRNet backbone (Jan, 2022), improving mIoU by more than 12 pp. The also results illustrate the genericity of the model, evidenced by a strong "few-shots" performance when the base model is trained on only 196 samples from the HCMSSD–Paris subset of the Semap training set. They also highlight the model's potential for transfer learning, especially on diverse datasets like HCMSSD–World.

**Model capacity.** Table 6 presents the performance of the model performance on the training, validation, and test partitions of Semap and reports the ablation study. The model attains an mIoU of 94.4 % on the training set, 75.5 % on the validation set, and 74.2 % on the test set (Table 3). This disparity indicates that the model overperforms on the training data, signaling moderate overfitting. Such behavior is expected and not necessarily negative. For one, it proves that the model size and capacity are sufficient for the task at hand. It also validates the choice of architecture and suggests that additional training data could lead to further performance improvements. The actual harm caused by overfitting is limited by the early-stopping policy. Naturally, early stopping affects in turn the gap between validation and test performance, yet this difference remains limited. Before real-data fine-tuning, the validation mIoU is 70.9 % (F1 = 82.7 %), whereas the test mIoU is 69.1 % (F1 = 81.4 %), a marginal difference. After fine-tuning, the validation mIoU increases to 75.5 % (F1 = 84.5 %), and the test mIoU reaches 74.2 % (F1 = 80.2 %). These results confirm that the real-data fine-tuning step positively influences overall performance, including on unseen test samples.

**Impact of multiscale integration and synthetic data pretraining.** The effects of the two remaining training strategies, i.e., multiscale integration and synthetic data pretraining, are reported for the test set (Tab. 6). Both strategies enhance performance and removing either one decreases the mIoU by 4 to 5 percentage points. Synthetic data pretraining appears to benefit recall, whereas multiscale integration improves both recall and precision in comparable extents. Their influence on overall pixel-level accuracy, however, seems less equivocal. As detailed in Table 3, this ambiguity is mostly attributable to the background class, which is better recognized without multiscale integration or synthetic data pretraining. Conversely, Table 3 suggests that multiscale integration and synthetic data pretraining clearly improve the recognition of all four geographic classes, which constitutes the primary objective of the model. Multiscale integration seems to have a particularly positive effect on the recognition of the built (+9.0 IoU, +8.9 Recall, +3.0 Precision), and water classes (+3.8 IoU, +3.8 R, +7.6 P). The impact of synthetic data pretraining seems to be more distributed, with significant impact on the recognition of the road network (+6.6 IoU, +7.0 R, –1.0 P), water (+5.2 IoU, +5.8 R, –0.9 P), built (+5.2 IoU, +0.3 R, +4.7 P), and non-built (+3.4 IoU, +3.0 R, +0.2 P) classes.

**Table 6 | Ablation study** and comparison of the results on test, validation, and training sets ($n_{test}$=144, $n_{val}$=289, $n_{train}$=1,006). Removing one feature at a time, hereafter specified as *treatment*. mIoU = mean Intersection over Union, mR = mean recall, mP = mean precision, Acc = accuracy. Means are computed over the four geographic classes (built, non-built, water, road network), while Acc is computed as the overall pixel accuracy. F1 = (mP+mR)/2

| Set/*treatment* | mIoU | mR | mP | F1 | Acc |
|---|---|---|---|---|---|
| Test set (base) | **74.2** | **79.4** | **85.4** | **82.4** | 80.2 |
| – *multiscale* | 70.0 | 75.5 | 81.5 | 78.5 | 79.6 |
| – *synth. pretrain.* | 69.1 | 75.4 | 84.2 | 79.8 | **81.5** |
| – *finetuning* | 69.1 | 73.3 | 83.7 | 78.5 | 81.4 |
| Validation set | **75.5** | **80.5** | **87.8** | **84.1** | **84.5** |
| – *finetuning* | 70.9 | 76.3 | 86.7 | 81.5 | 82.7 |
| Train set | 94.4 | 96.9 | 97.1 | 97.0 | 97.8 |

**Confusion rates.** Figure 5 presents the confusion rates in the form of a confusion matrix. The diagonal of Figure 5a corresponds to precision values[7]. The results indicate that the contours, built, and road network classes appear to be overweighted, mainly at the expense of the non-built class. This imbalance does not substantially affect the non-built class, however, since its area is comparatively larger (see Table 1). Some regions that should be classified as background are instead attributed to the non-built, water, or road network classes. Finally, portions of the water and road network classes tend to be misclassified as contours. This may correspond to linear objects, like waterways and thin roads.

**Calculation time and efficiency.** The model was trained on an older-generation (2014) Nvidia GeForce Titan X GPU equipped with 12 GB of VRAM. On this hardware, the network required 150 hours to converge. At inference time, the computation took 0.66 seconds per 768 × 768 image sample, which corresponds, on average, to 101 seconds per map, including multiscale integration and image reading/writing time.

[7] You may notice that these values, as well as result of recall, differ from those in Table 3. Here the more standard micro average value—that is the average confusion per test sample—is used (cf. Tab. A1, and Tab. A2). This makes it possible to weight each confused pixel equally, regardless of the class to which it belongs.

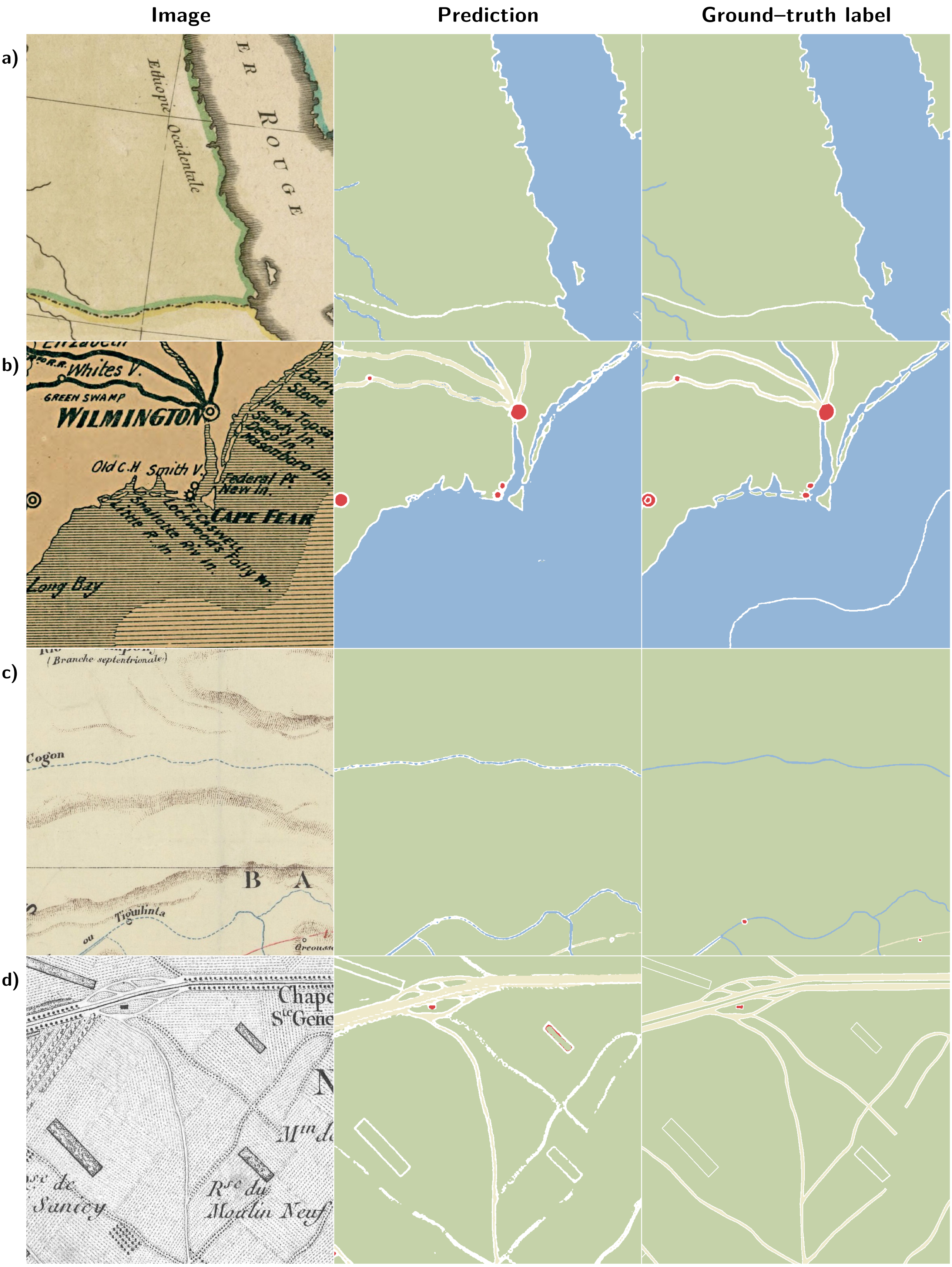

**Figure 6** | **Results of inference on the test set** for a random subset of eight samples (continued in Fig. 7).

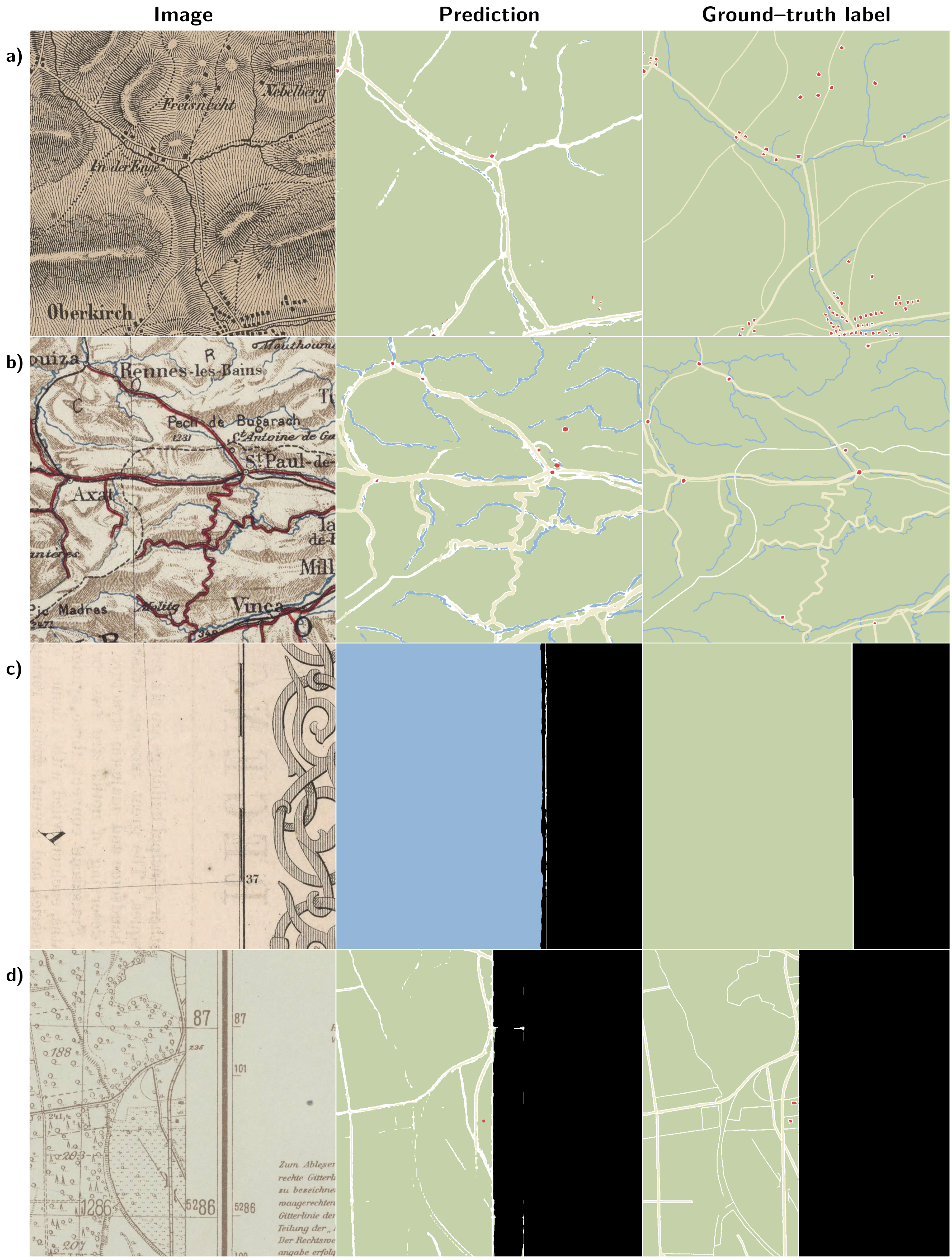

**Figure 7** | **Results of inference on the test set** for a random subset of eight samples (continued from Fig. 6).

*Qualitative analysis*

Figures 6 and 7 present inference details for a random subset of eight test samples. Figures 8 through 12 portray full-image semantic segmentations for a selected set of five difficult maps. Four additional examples, corresponding to different cartographic cases, are provided in Figures A1–A4, in the Appendix. The qualitative results show that the detection of the contours class remains unsatisfactory. This lower recognition performance also seems to marginally impact the recognition of other linear features, such as linear roads and riverways. By contrast, surfaces are generally well recognized.

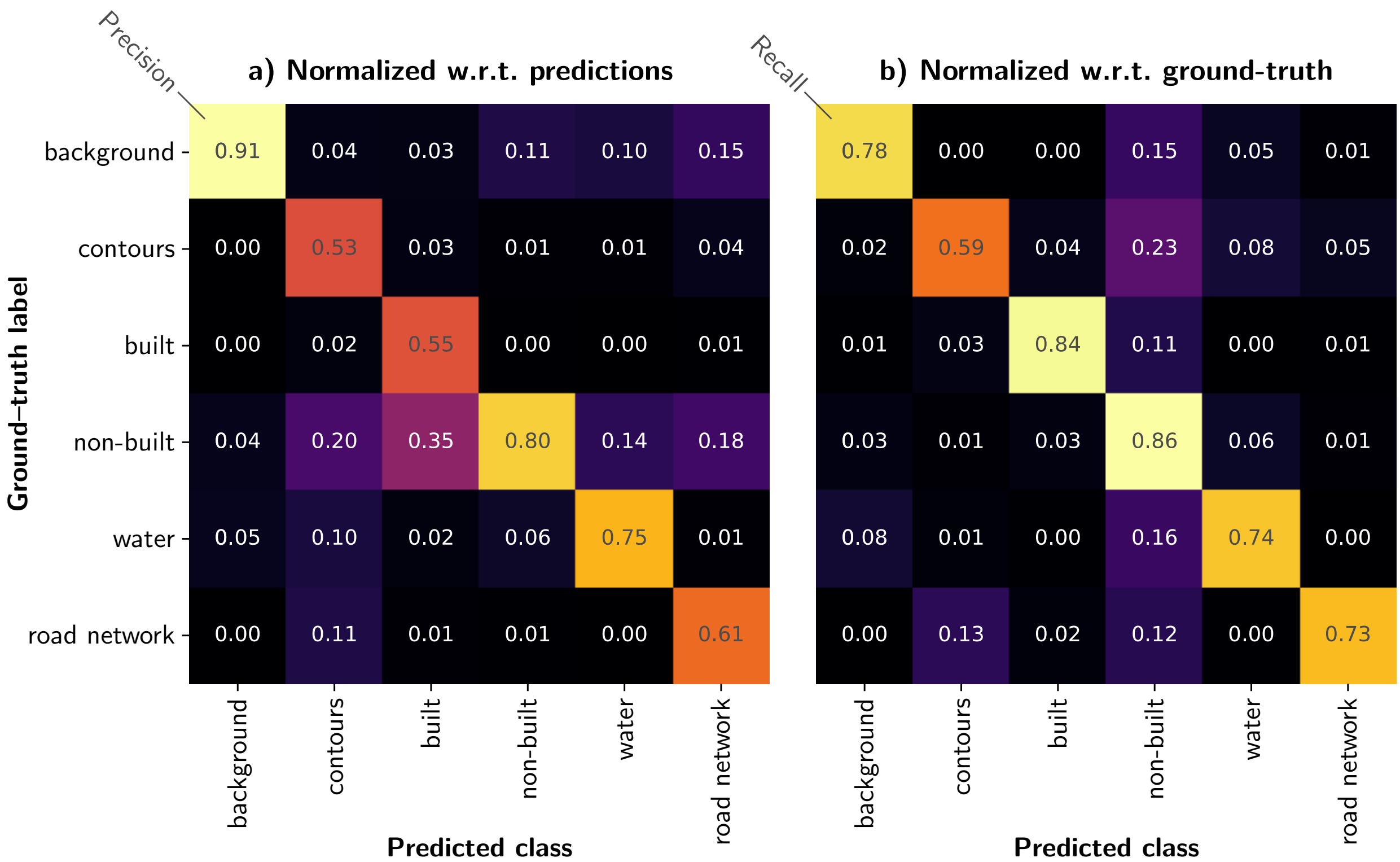

| | background | contours | built | non-built | water | road network |
|---|---|---|---|---|---|---|
| **a)** background | 0.91 | 0.04 | 0.03 | 0.11 | 0.10 | 0.15 |
| contours | 0.00 | 0.53 | 0.03 | 0.01 | 0.01 | 0.04 |
| built | 0.00 | 0.02 | 0.55 | 0.00 | 0.00 | 0.01 |
| non-built | 0.04 | 0.20 | 0.35 | 0.80 | 0.14 | 0.18 |
| water | 0.05 | 0.10 | 0.02 | 0.06 | 0.75 | 0.01 |
| road network | 0.00 | 0.11 | 0.01 | 0.01 | 0.00 | 0.61 |

| | background | contours | built | non-built | water | road network |
|---|---|---|---|---|---|---|
| **b)** background | 0.78 | 0.00 | 0.00 | 0.15 | 0.05 | 0.01 |
| contours | 0.02 | 0.59 | 0.04 | 0.23 | 0.08 | 0.05 |
| built | 0.01 | 0.03 | 0.84 | 0.11 | 0.00 | 0.01 |
| non-built | 0.03 | 0.01 | 0.03 | 0.86 | 0.06 | 0.01 |
| water | 0.08 | 0.01 | 0.00 | 0.16 | 0.74 | 0.00 |
| road network | 0.00 | 0.13 | 0.02 | 0.12 | 0.00 | 0.73 |

**Figure 5** | **Confusion matrix** based on Semap test set predictions. The matrix is computed as the average over all test samples (micro-average), normalized per class, **(a)** w.r.t. predictions or **(b)** w.r.t to the ground-truth. The diagonal values correspond to **(a)** precision, and **(b)** recall.

One notable exception is Figure 7c. In that case, no distinctive cues help discriminate the empty surface between non-built/land or of water/sea. When morphological or graphical information is available, like in Figures 6a and 8, land can usually be distinguished from seas. The model appears to rely on color for the detection of water when the map is colored; this reliance explains, for instance, the inversion of water and landmasses in Figure 9, an ethnographic map from the 19th century in which some land regions are tinted blue, leading to inversion of sea and landmasses, in the semantic segmentation mask. By contrast, iconographic content and allegories, like the marine creatures and ships depicted in Figure 10, are not always usefully considered by the network. One counterexample is the view of Thomaston shown in Figure 11, where the map is well segmented despite its pictoriality.

Distinguishing linear features from one another can be challenging. One example is Fig. 6b, where roads and rivers tend to be confused. In this case, the distinction relies mainly on context and on the graphical coherence between the sea and the river, both represented by horizontal hatchings. In Figure 6c, the *Cogon* River, drawn with a hatched blue line, is confused with a boundary line, a mistake that occurs quite frequently according to the confusion matrix (Fig. 5). In Figure 7b, however, the distinction between roads and waterways is generally accurate, probably owing to clear color differentiation and—possibly—hillshading, which might help the model recognize the rivers carving the landscape. In Figure 6d, the smaller paths are delimited only by a dotted line traversing a field that is itself depicted with a dotted texture. Although the model detects these paths, it does not classify them correctly, assigning them instead to the contours class. Heavy texturing also impairs the recognition of other thin or small features, as in Figure 7a, where hachures significantly complicate the detection of dotted paths, rivers, and even houses. Features below a minimal area of a few pixels are generally hard to recognize, as observed with the cities in Figure 7b and the houses in Figure 7d. In the latter case, low contrast and high graphical density might also be partly responsible.

Figure 8 constitutes an instructive case study. In this cartogrammatic map of railway usage, statistical data are overlaid on a map of Europe, yet the model manages to overlook these strong graphical cues in several locations, notably along the German–Russian border, in French Normandy, and, to some extent, in the northwest of Ireland and the northeast of England. Figure 12, a Sanborn fire insurance map, illustrates a converse situation in which graphical density is too low; although the image resolution is high, graphical cues remain sparse, making roads and non-built areas hard to recognize.

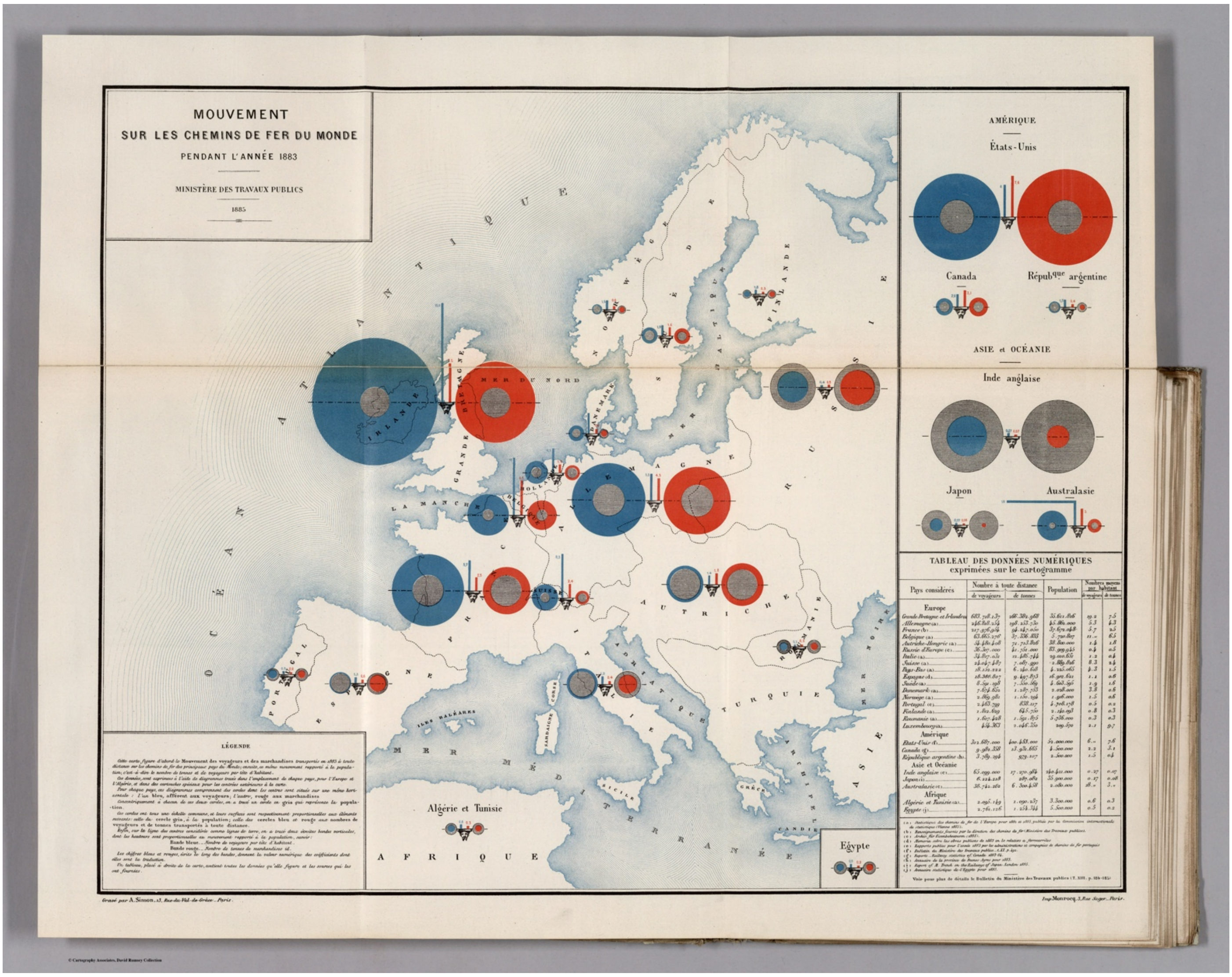

**Figure 8 | (a) Map of railway trade volumes, 1883.** Emile Cheysson. *Mouvement sur les chemins de fer du monde*, 1833. Ministère des Travaux Publics, Paris. 49 x 60 cm. David Rumsey Collection, 12516.014. URL: davidrumsey.com/luna/servlet/detail/RUMSEY~8~1~309240~90079149

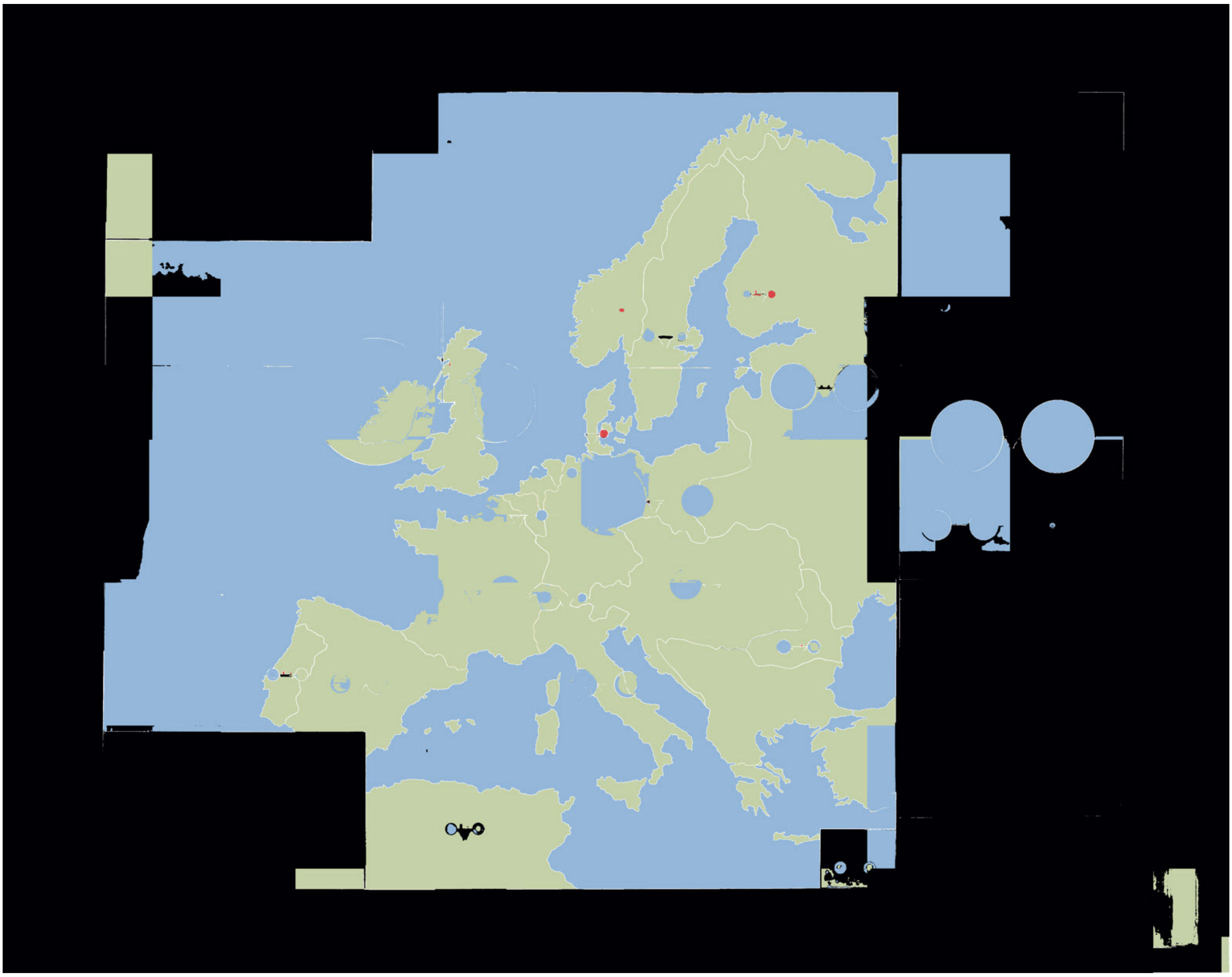

**Figure 8 | (b) Result of the semantic segmentation of Figure 8a.** *The recognition of coastlines is impacted, although not entirely hindered by the presence of overlapping circles. Dark blue areas tend to be mistaken for water.*

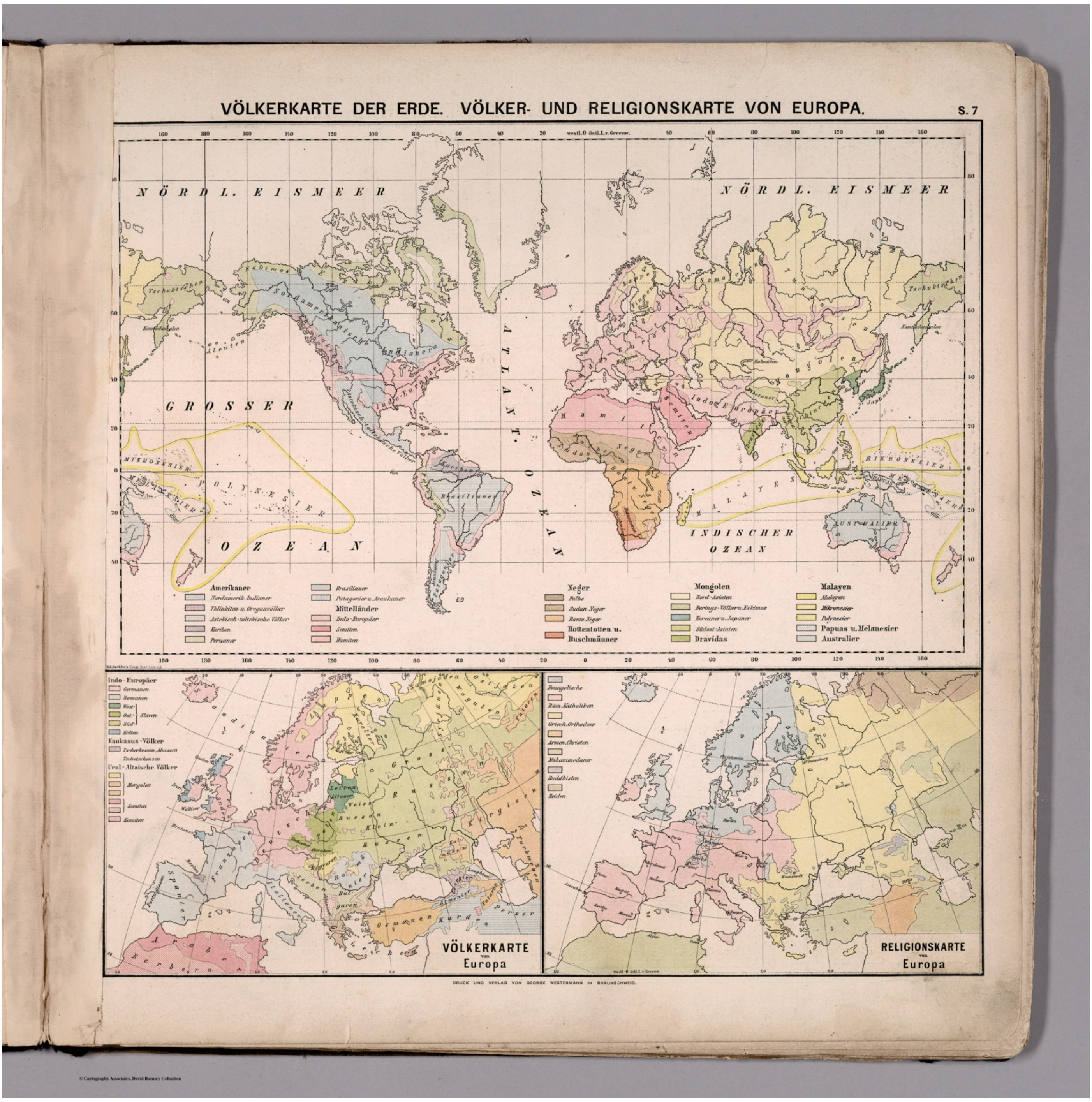

**Figure 9 | (a) Ethnographic and religious map, 1883.** Carl Diercke, Eduard Gaebler. *Völkerkarte der Erde. Völker- und Religionskarte von Europa*, 1883. Published by George Westermann, Braunschweig. 32 x 31 cm. David Rumsey Collection, 12198.011. URL: davidrumsey.com/luna/servlet/detail/RUMSEY~8~1~311889~90081579

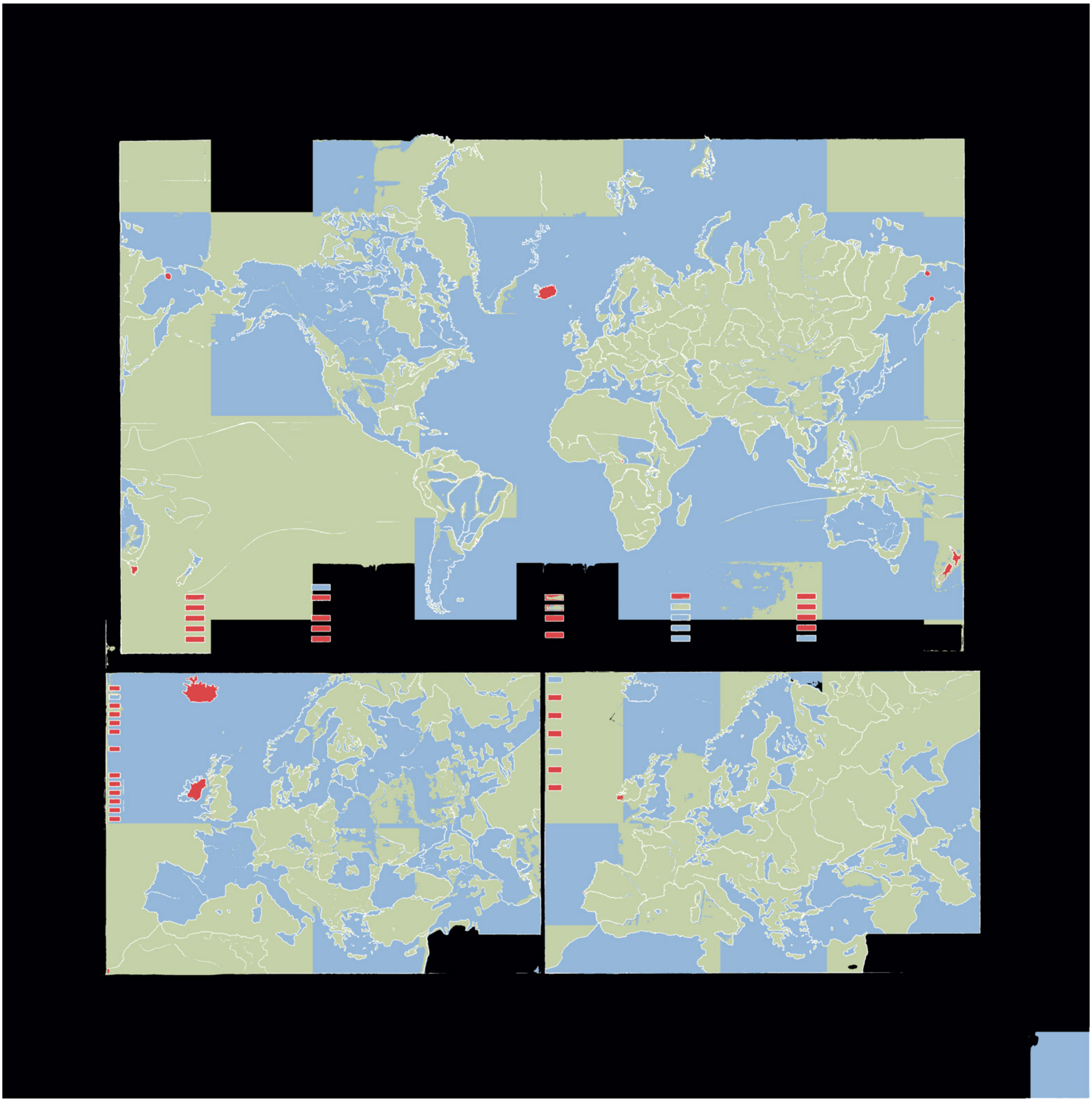

**Figure 9 | (b) Result of the semantic segmentation of Figure 9a.** *The distinction of land masses from seas seems affected by the use of a color code to represent thematic categories.*

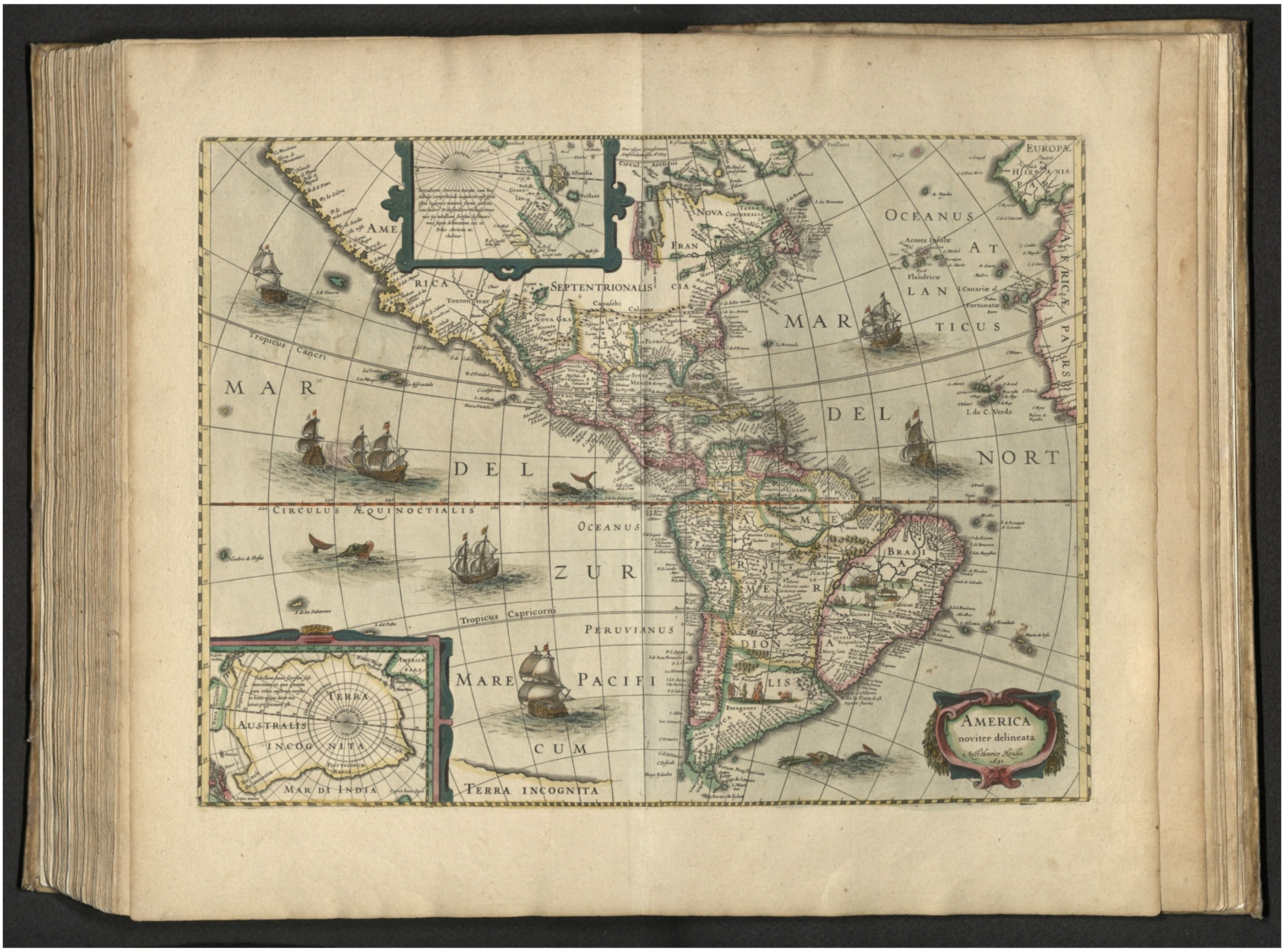

**Figure 10 | (a) New map of America, part of the Atlas of Gerard Mercator and Jodocus Hondius, 1633.** Hendrik Hondius. *America noviter delineata*, 1633. Published in Amsterdam. Copperplate, hand colored. 38 x 50 cm. David Rumsey Collection, 10621.195. URL: davidrumsey.com/luna/servlet/detail/RUMSEY~8~1~345039~90107477

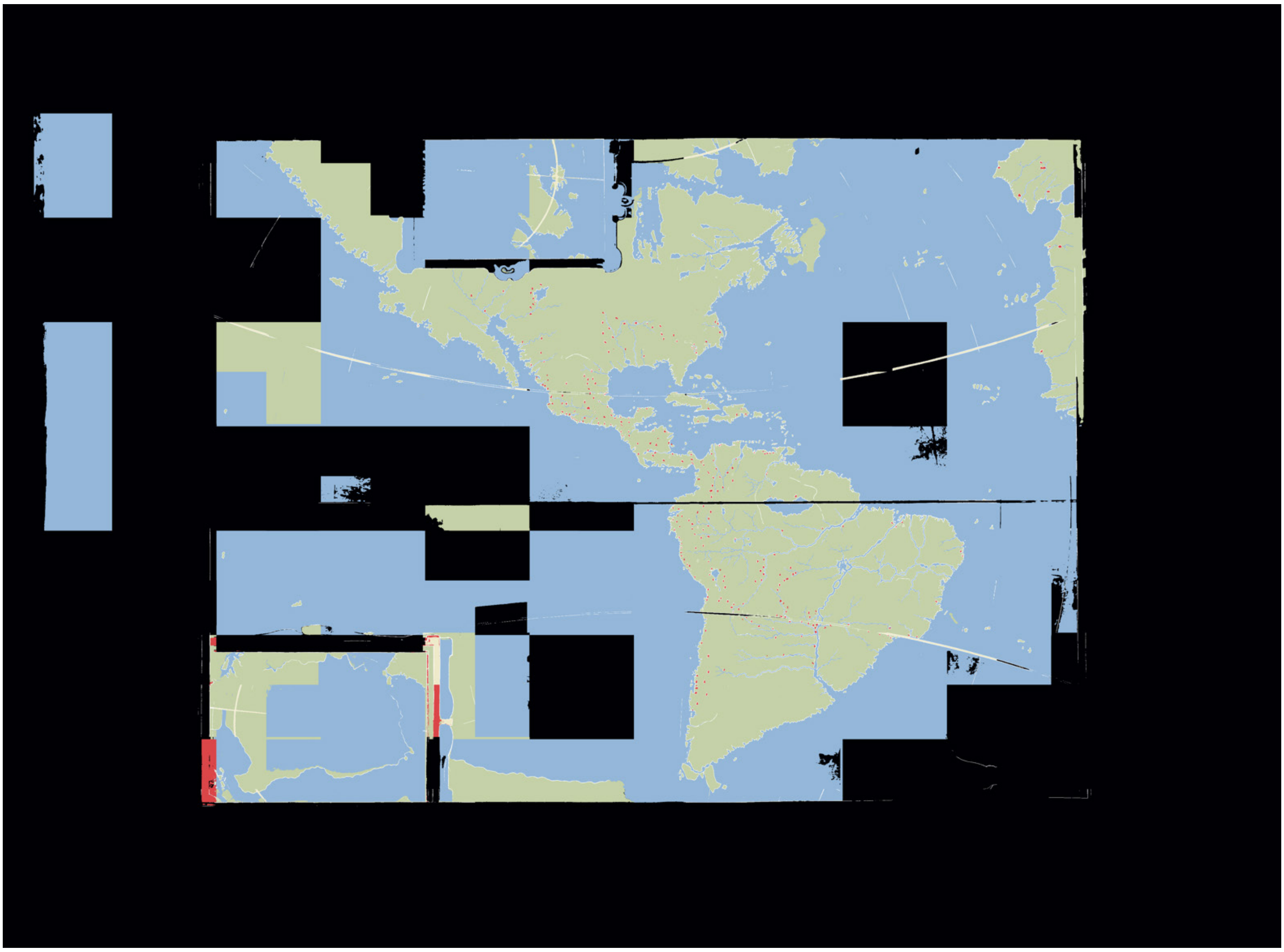

**Figure 10 | (b) Result of the semantic segmentation of Figure 10a.** *Landmasses are overall well segmented, whereas the recognition of seas seems impacted by the figuration of ship icons.*

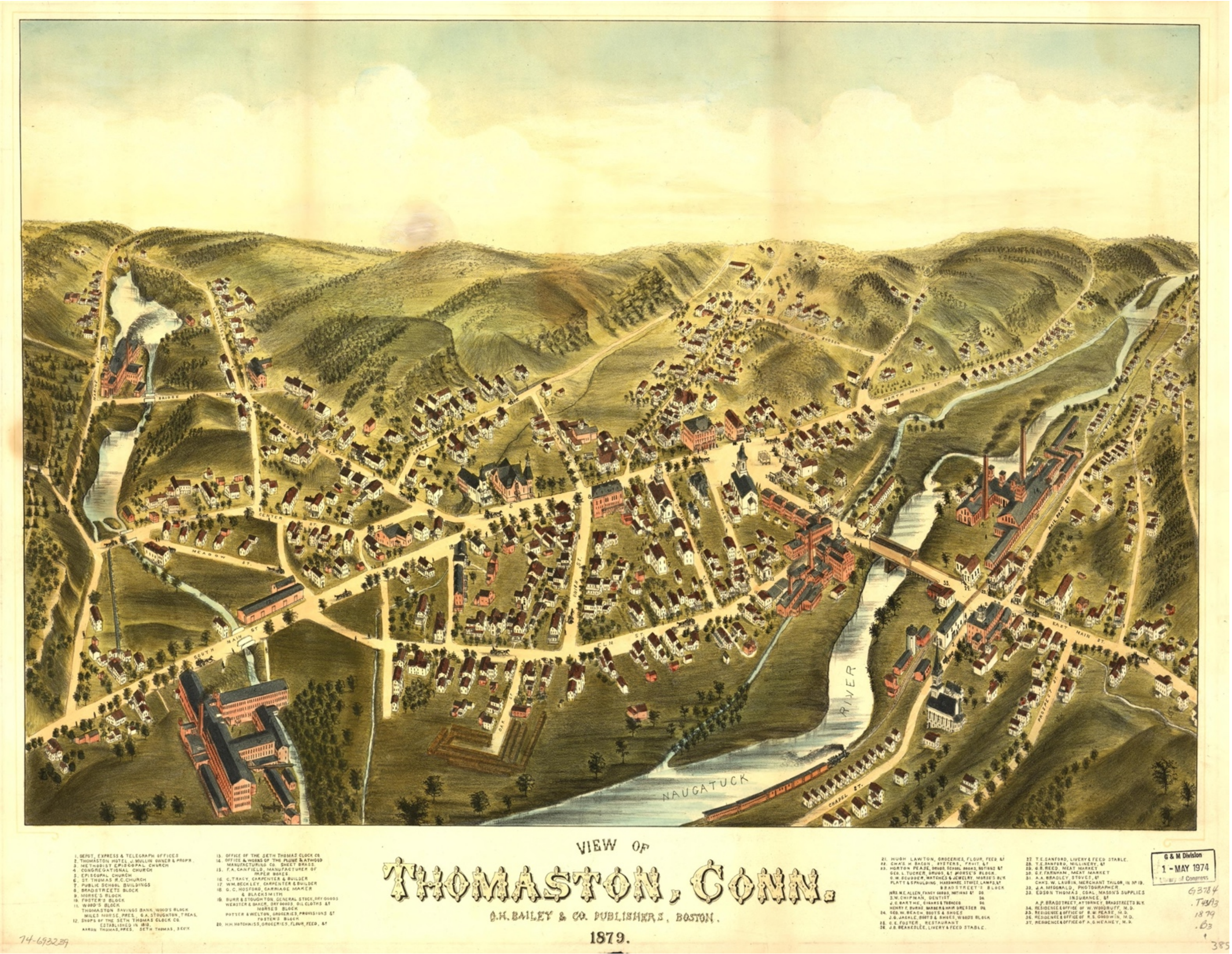

**Figure 11** | **(a) View of Thomaston, Connecticut, 1879.** O.H. Bailey. *View of Thomaston, Conn.*, 1879. O.H. Bailey & Co, Boston. Hand colored. 43 x 63 cm. Library of Congress, G3784.T43A3 1879.B3. URL: hdl.loc.gov/loc.gmd/g3784t.pm000974

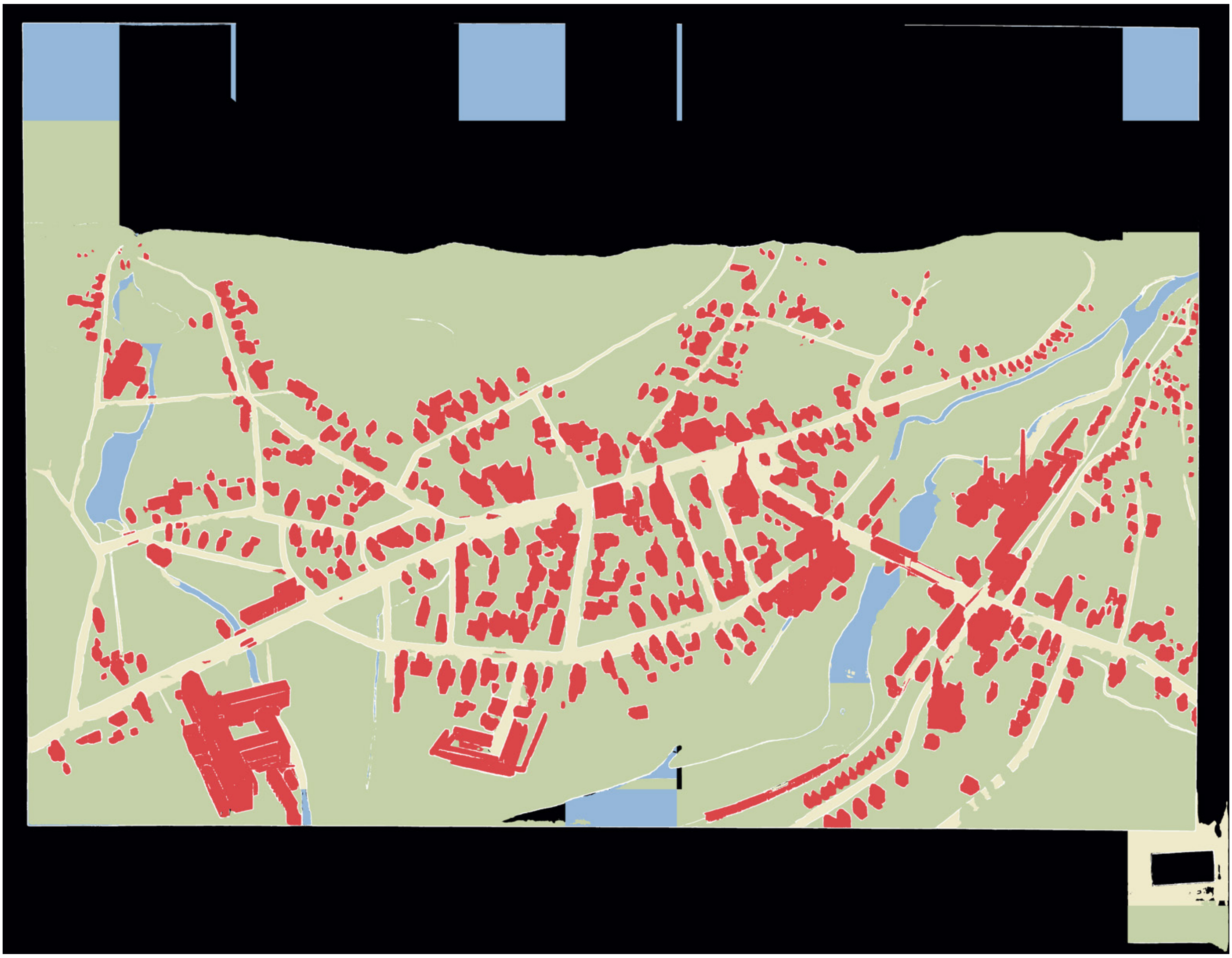

**Figure 11 | (b) Result of the semantic segmentation of Figure 11a.** *Dwellings are well recognized despite the strong iconographic character of the representation.*

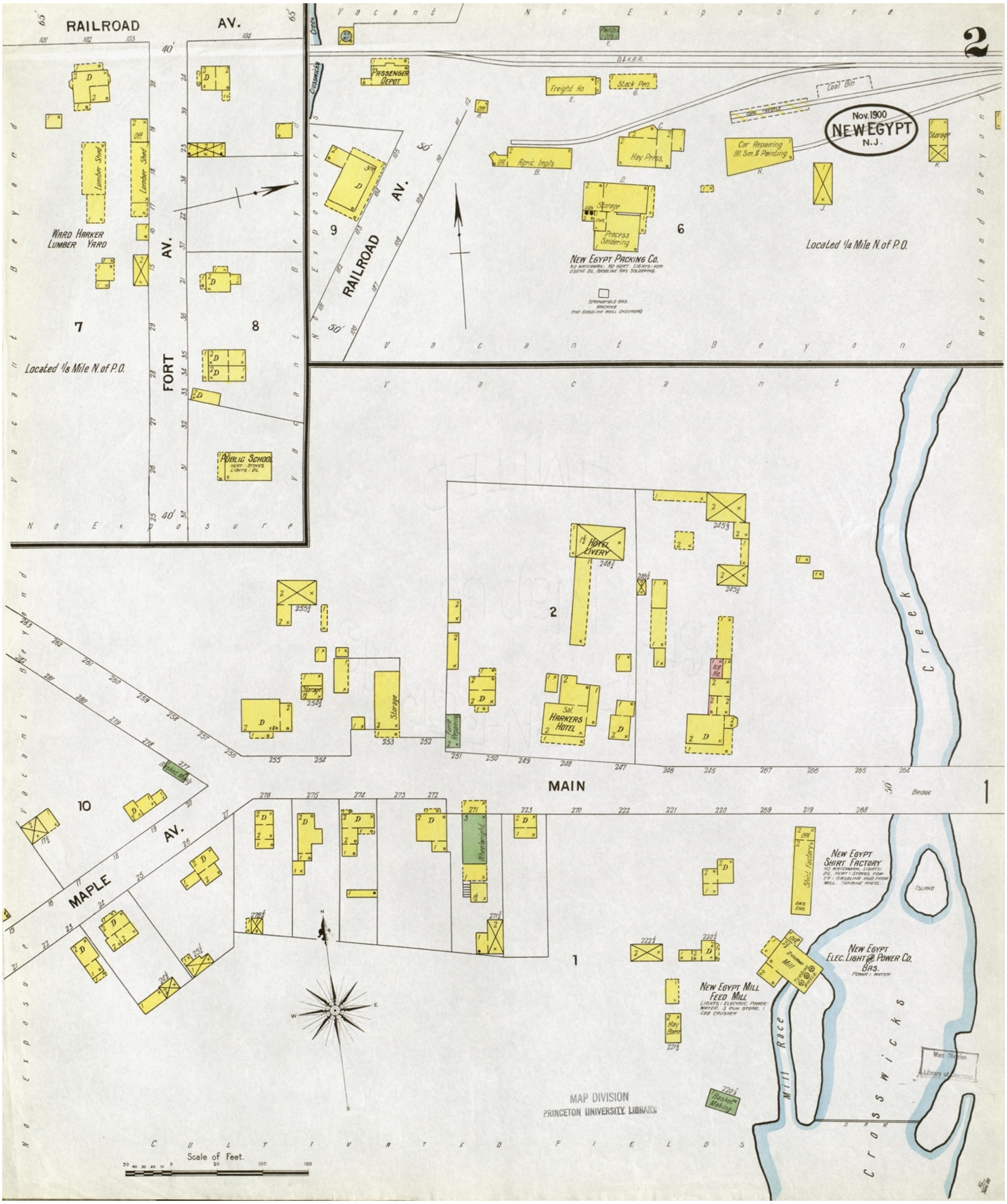

**Figure 12 | (a) Fire insurance plan of New Egypt, Ocean, New Jersey, 1900.** *New Egypt N.J. 2*, 1900. Sanborn Map Company, New York. 64 x 54 cm. Princeton library, HMC04. URL: catalog.princeton.edu/catalog/9955740603506421

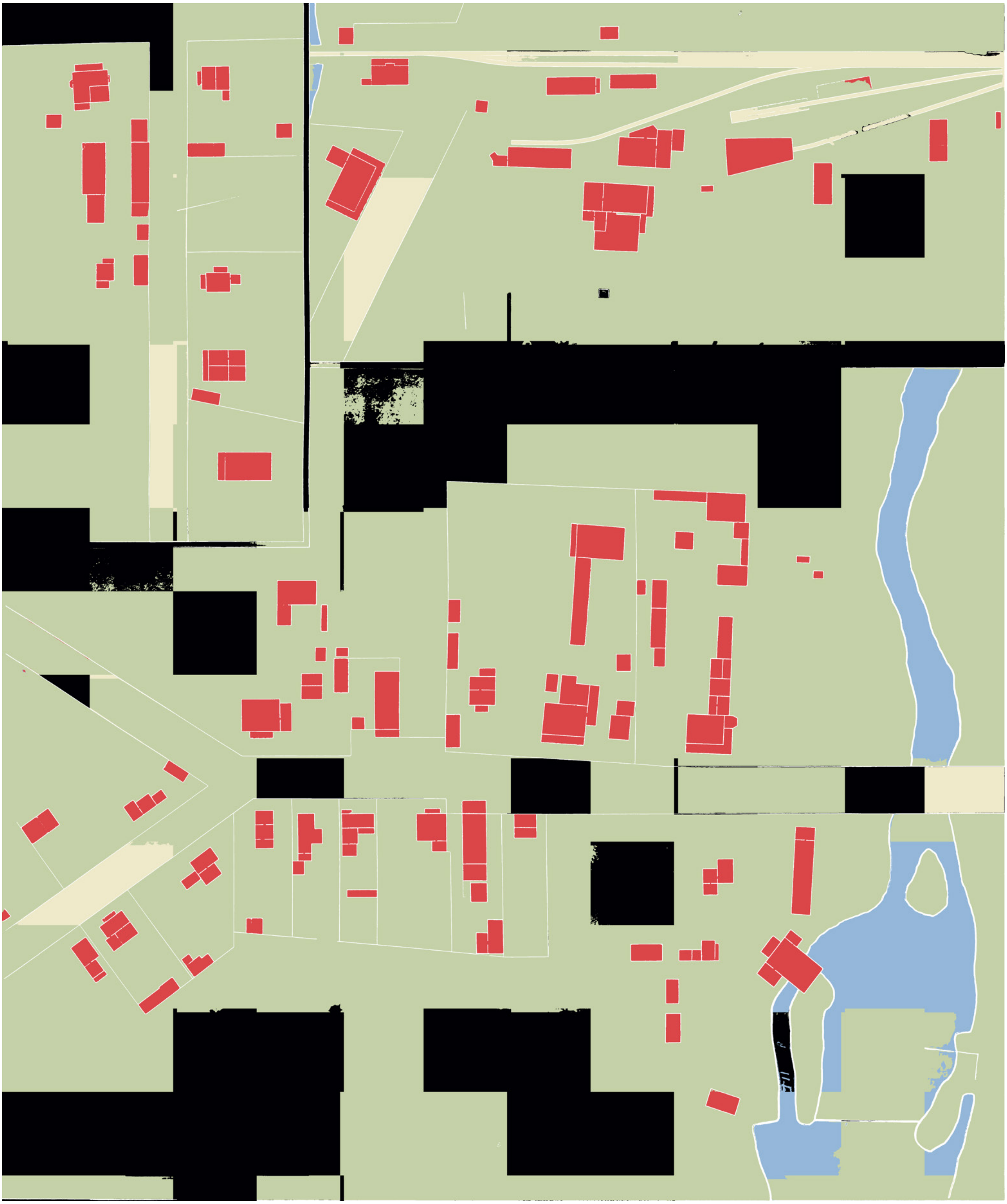

**Figure 12 | (b) Result of the semantic segmentation of Figure 12a.** *Building segmentation is accurate, whereas a lack of spatial context impairs the recognition of the road network and non-built surfaces.*

## 4.5 General discussion of segmentation results

The proposed method delivers strong performance across diverse map contexts. It significantly improves semantic segmentation results on both HCMSSD–Paris and HCMSSD–World compared with previous approaches. One of its main limitations remains the detection of contours and thin linear features, which, however, was not the principal focus of this research. Specialized architectures can be employed for precise deep edge filtering, including convolutional ones (Chen et al., 2021; Petitpierre et al., 2023). The training of a separate, specialist model would therefore be advisable if the primary objective were to extract closed shapes, i.e., to vectorize the geometries. The present study can accommodate this imprecision, since subsequent analysis primarily require an accurate recognition of areal classes, the investigation of image composition, or the semantic composition of map elements. In this respect, the Swin-based model adequately segments map areas by leveraging visual context and relying on hierarchical integration at inference time. In cases where graphical cues are sparse, as in Figs. A4 and A6, it may be beneficial to add further hierarchical integration stages; here, only two-stage integration was implemented. The additional computational cost would be minimal, since each extra stage requires only one quarter of the time needed for the preceding one. Beyond thin linear features, small objects, such as icons, may also be missed. This limitation is not critical in the present work, since Chapter 6 introduces a sign-detection model that specifically addresses this issue.

This research distinguishes itself from the way map processing tasks are usually conducted. While research on map recognition tends to focus on well-defined cartographic series, several recent works in the last few years have questioned the pertinence of this approach (Kim et al., 2023; Petitpierre, 2020). The present results indicate that the model performs similarly when trained on highly heterogenous datasets—in terms of map scale, content, period of publication, and style—compared with more homogenous ones, such as Paris HCMSSD. The versatility of the model is supported by synthetically generated data, used in complement to real data, which appears to drive the model to learn morphological cues for segmenting geographic classes, as indicated by the results from Figure 6a, for instance. Here, in contrast to what was observed with UNet (Petitpierre et al., 2021), color seems to exert a significant influence on segmentation accuracy, as exemplified in Figure 7b or, conversely, Figure 8. This shift toward more comprehensive map recognition models is illustrated as well in Table 5. Whereas convolutional models performed poorly when trained on diverse data (e.g. HCMSSD-World), compared to more homogeneous sets (e.g. HCMSSD-Paris), the proposed approach reverses this situation: high diversity seems to reinforce the model's robustness rather than confuse it. This translates in a substantial increase of 26.7 percentage points in the F1 score on the diverse HCMSSD–World dataset, up to 90.5 percent, compared with the more homogenous HCMSSD–Paris dataset ($F1_{score}$ = 86.4, +14.5 pp).

One conscious, noteworthy implementation choice is relying on procedural generation to produce synthetic training data where approaches employing generative models are generally preferred. This choice offered greater control on the relative frequency of distinct textures or visual features. It also helped enriching maps with additional layers, such as relief and place names. Procedural generation also permits the creation of visually diverse synthetic samples, whereas generative models may overfit on most common map figurations. This feat, however, may prove a disadvantage when algorithmic choices are not sufficiently informed by the expertise of real data.

The performance improvement should also be understood as a consequence of the model's augmented capacity: whereas ResNet101 contained 43 million parameters, Swin-L comprises 197 million. Although the Swin-L encoder may thus appear large, its size is only 18% of a contrastive vision transformer models like DINOv2 (Oquab et al., 2024), or 0.2% of the size of large language-vision models like Llama 3.2 (Grattafiori et al., 2024; *Llama 3.2*, n.d.), making it quite cost-efficient, by comparison. The results in Table 6 also indicated that the present model tended to overfit on training set, suggesting that it remains under-utilized; additional training data could further enhance performance in the future.

Up to now, this chapter focused primarily on the technical aspects of semantic segmentation, the remaining pages, however, will introduce a first analysis of map semantics. More specifically, they will describe the relative representation of semantic classes in historical cartography, leveraging the 99,715 segmented maps from ADHOC Images dataset.

## 4.6 Initial analysis of map semantics

### *What mapmakers were (not) drawing*

The overall distribution of classes observed in ADHOC Images is as follows[8]: 56.5% non-built, 15.9% water, 4.6% built or urban areas, 4.1% road network. The background occupies over 15.1% of the image areas, while contours represent 3.8%. These proportion lie close to the values observed in the training data (Tab. 2). While built and road network classes are more represented in Semap annotations (11.7% and 7.2%, respectively), non-built is more frequent in synthetic training data (72.9%). The water class appears underrepresented in both annotated (7.9%) and synthetic training data (9.8%) compared to the proportion observed in ADHOC Images predictions (15.9%), which is still much below the actual area covered by water on Earth (ca. 71%). This result underscores the idea that maps are not just snapshots of the Earth's geography, nor even are they representative thereof. Maps are designed images and are thus foremost representative of the cultural attention paid to certain classes of geographical objects. This focus can change over time, in function of the cultural and historical context.

[8] Mean value, normalized for each map image.

Figure 13 depicts the relative share of each of the five semantic classes as a function of scale, publication year, country, and coverage. The share of each class appears quite stable over time (Fig. 13a). One can note, however, two surges in the relative share of urban areas (built and road network classes), first between 1550 and 1650, and again between 1850 and ca. 1930. Water is also comparatively more prevalent between 1650 and 1850, which is consistent with the results discussed in Chapter 3, Fig. 8, regarding Atlantic charting. The early 17$^{th}$–century surge in urban maps may be discussed in view of the numerous city atlases issued at the time, most notably Braun's *Civitates Orbis Terrarum* (cf. Chapter 2). The second surge, 1850–1930, coincides with the onset of urban planning with, for instance, the Haussmannian transformations in Paris (1853–1870), Castro's plan for the enlargement of Madrid (1857), or the decree for the construction of Vienna's Ringstrasse (1857). These interventions were accompanied throughout the West by the development of urban infrastructure networks. This period also corresponds to the onset of urban geography, exemplified for instance by John Snow's (1813–1858) emblematic Broad Street cholera map, in London (1854).

The distinction among semantic classes is more marked when observed through the prism of map scale (Fig. 13b). Urban maps (built class) constitute a substantial share of maps up to a scale of 1:20,000. They vanish abruptly and almost disappear at smaller scales, although the built class appears to maintain a minimal residual level until very small–scale maps. Road networks exhibit a comparable pattern, although their disappearance is more gradual. This result will be discussed in greater detail in Chapter 5, section 5.3. While a lower relative representation of urban areas in small-scale maps is expected from a cartographic perspective, the abruptness of the trend is nonetheless surprising. It suggests a clear dichotomy in the way cities are represented according to map scale. The drop might be caused by a change in representation, such as the iconization of cities, which could explain the contrastively stable residual share of built class at smaller scales. Conversely, the share of water increases steadily as scale decreases, reaching nearly 38 % of map content at a scale of 1:20,000,000. This percentage is still only about half the proportion one would expect on a world map, reflecting the tendency of frames and map projections to exaggerate the surface of continents compared to seas.

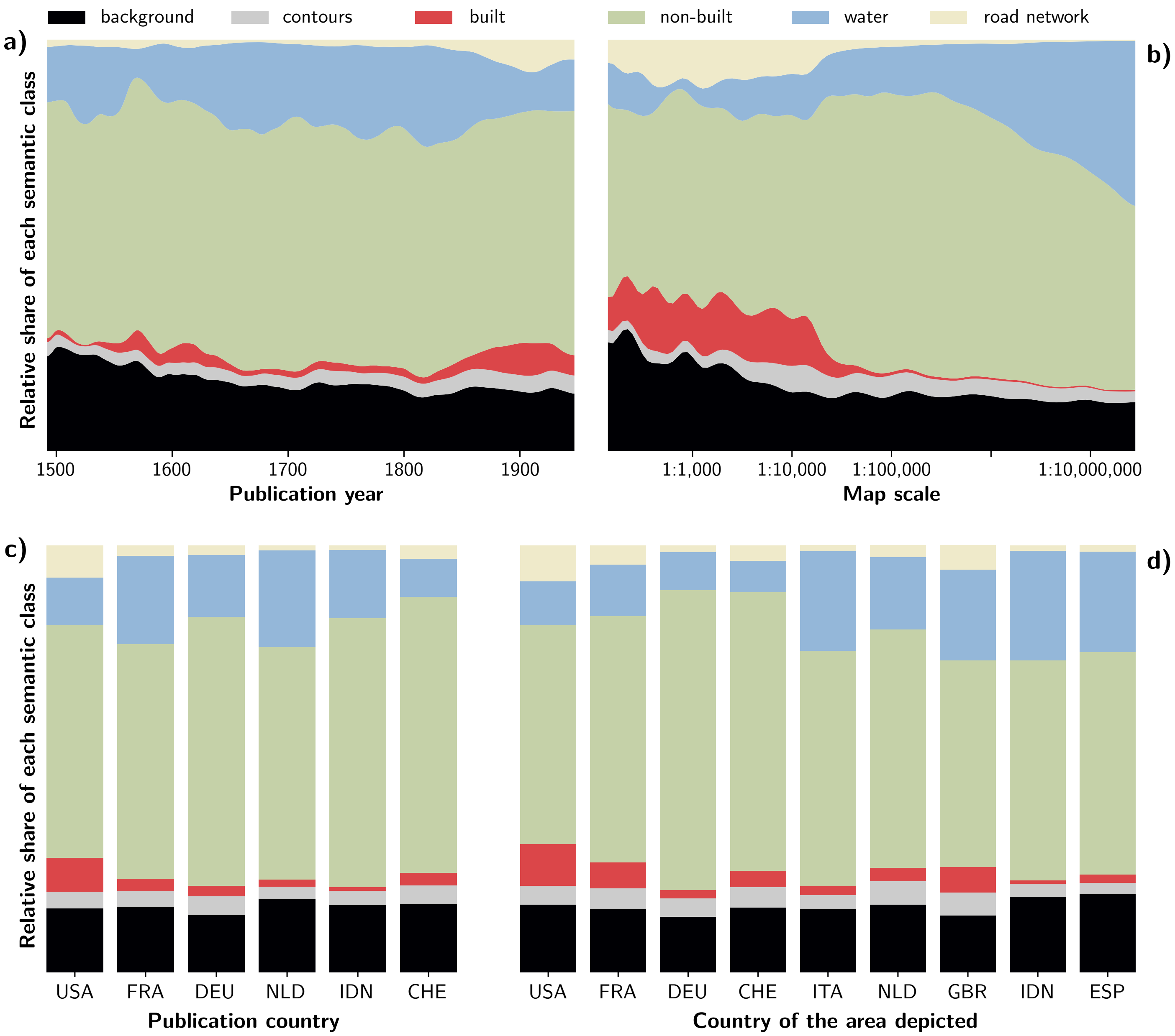

**Figure 13 | Relative share of each semantic class by publication year, map scale, publication country, and country of the area depicted.** For publication year **(a)**, values are averaged over 20 years, and a Gaussian filter ($\sigma = 4$) is applied to mitigate aleatoric noise. For map scale **(b)**, the scale is logarithmic. Values are averaged across a range of $\pm 10^{0.1}$, and a Gaussian filter ($\sigma = 1.5$) is applied to mitigate aleatoric noise. For publication countries **(c)**, and countries of the area depicted **(d)**, the legends correspond to ISO-3 country codes (USA = United States, FRA = France, DEU = Germany, NLD = Netherlands, IDN = Indonesia, CHE = Switzerland, ITA = Italy, GBR = United Kingdom, ESP = Spain).

Differences also emerge according to the publication country (Fig. 13c). Dutch maps, for example, tend to contain more water, whereas American maps tend to be more urban. A similar pattern appears when the coverage location is considered (Fig. 13d): maps of Switzerland exhibit the least water, while maps of Indonesia—typically produced by Dutch cartographers—display the most. Maps of American and, to a lesser extent, English and French maps, are comparatively more urban. At first sight, these proportions may appear to simply reflect the historical geography of these regions. However, this interpretation would be reductive and overestimate the geographic representativity of historical cartography. For instance, the population density of the United States in the second half of the $19^{th}$ century—when most maps in the corpus were published—was comprised between 3 to 10 people/$km^2$. Contrastively, the historical population density of Switzerland at the same time was tenfold: 50–85 people/$km^2$. Thus, even considering the distinct urban planning strategies, historical built-up area in the U.S. couldn't plausibly have been twice that of Switzerland. Therefore, a more plausible interpretation of the discrepancy is rather that maps of the United States had a greater tendency to depict cities compared to those of Switzerland[9].

## 4.7 Conclusion

This chapter described the methods employed for the semantic segmentation of 99,715 maps images comprised in the ADHOC Images database. The presented approach employed distinct strategies, like synthetic data pretraining and multiscale integration at inference time, to achieve state-of-the-art segmentation performance. This study also introduced a new training dataset, Semap, consisting of 1,439 manually annotated map samples. Synthetic and manually annotated training data are published in open access along with the dissertation (Petitpierre et al., 2025).

Segmenting large sets of historical maps opens new research avenues for the quantitative analysis of historical cartography. One simple preliminary example, introduced at the end of this chapter, is the investigation of the relative share of distinct classes of geographic objects. However, segmented masks enclose even more detailed cultural information, visible for instance on the spatial relationships between those objects, or their location on the image. In the resource–constrained context of cartography, such information can illuminate the way certain classes of geographic objects tend to be overrepresented and emphasized at the expense of others. This issue will be the focus of the following chapter.

[9] One could argue that the discrepancy is caused by Swiss maps being published earlier on average. However, even when setting publication year as a fixed effect, U.S. maps still portray significantly more built-up area.

# Appendix A – Supplementary Materials

**Table A1 | Performance assessment strategies**. Variant formulae for computing the mIoU. For mR, the term $\cup_{i,k}$ is replaced by $Y_{i,k}$. For mR, it is replaced by $\widehat{Y}_{i,k}$.

| | |
|---|---|
| Sample-normalized macro average | $mIoU=\frac{1}{K}\sum_{k=3}^{6}\frac{\sum_{i=1}^{N}\sum_{\forall p}\cap_{i,k}Y_{i,k}}{\sum_{i=1}^{N}\sum_{\forall p}\cup_{i,k}Y_{i,k}}$ |
| Micro average | $mIoU=\frac{1}{N}\sum_{i=1}^{N}\frac{\sum_{k=3}^{6}\sum_{\forall p}\cap_{i,k}Y_{i,k}}{\sum_{k=3}^{6}\sum_{\forall p}\cup_{i,k}Y_{i,k}}$ |
| Macro average | $mIoU=\frac{\sum_{k=3}^{6}\sum_{i=1}^{N}\sum_{\forall p}\cap_{i,k}}{\sum_{k=3}^{6}\sum_{i=1}^{N}\sum_{\forall p}\cup_{i,k}}$ |
| Per class macro average | $mIoU=\frac{1}{K}\sum_{k=3}^{6}\frac{\sum_{i=1}^{N}\sum_{\forall p}\cap_{i,k}}{\sum_{i=1}^{N}\sum_{\forall p}\cup_{i,k}}$ |

**Table A2 | Comparison of performance assessment strategies** on Semap, HCMSSD–Paris, and –World datasets (*retrained* models). Based on the formulae provided in Tab. A1. F1 = (mP+mR)/2.

| Set | Performance assessment strategy | mIoU | mR | mP | F1 |
|---|---|---|---|---|---|
| Semap | Sample-normalized macro average (base) | 74.2 | 79.4 | **85.4** | **82.4** |
| | Micro average | **74.4** | **80.3** | 80.3 | 80.3 |
| | Macro average | 67.0 | **80.3** | 80.3 | 80.3 |
| | Per class macro average | 57.4 | 79.4 | 67.8 | 73.6 |
| ACMSSD Paris | Sample-normalized macro average | 79.3 | 84.2 | 87.7 | 85.9 |
| | Micro average | **84.6** | **90.6** | **90.6** | **90.6** |
| | Macro average | 82.8 | **90.6** | **90.6** | **90.6** |
| | Per class macro average (base) | 76.0 | 84.2 | 88.6 | 86.4 |
| ACMSSD World | Sample-normalized class macro average | **81.4** | 86.8 | **92.0** | **89.4** |
| | Micro average | 80.7 | **87.8** | 87.8 | 87.8 |
| | Macro average | 78.2 | **87.8** | 87.8 | 87.8 |
| | Per class macro average (base) | 74.2 | 86.8 | 84.1 | 85.4 |

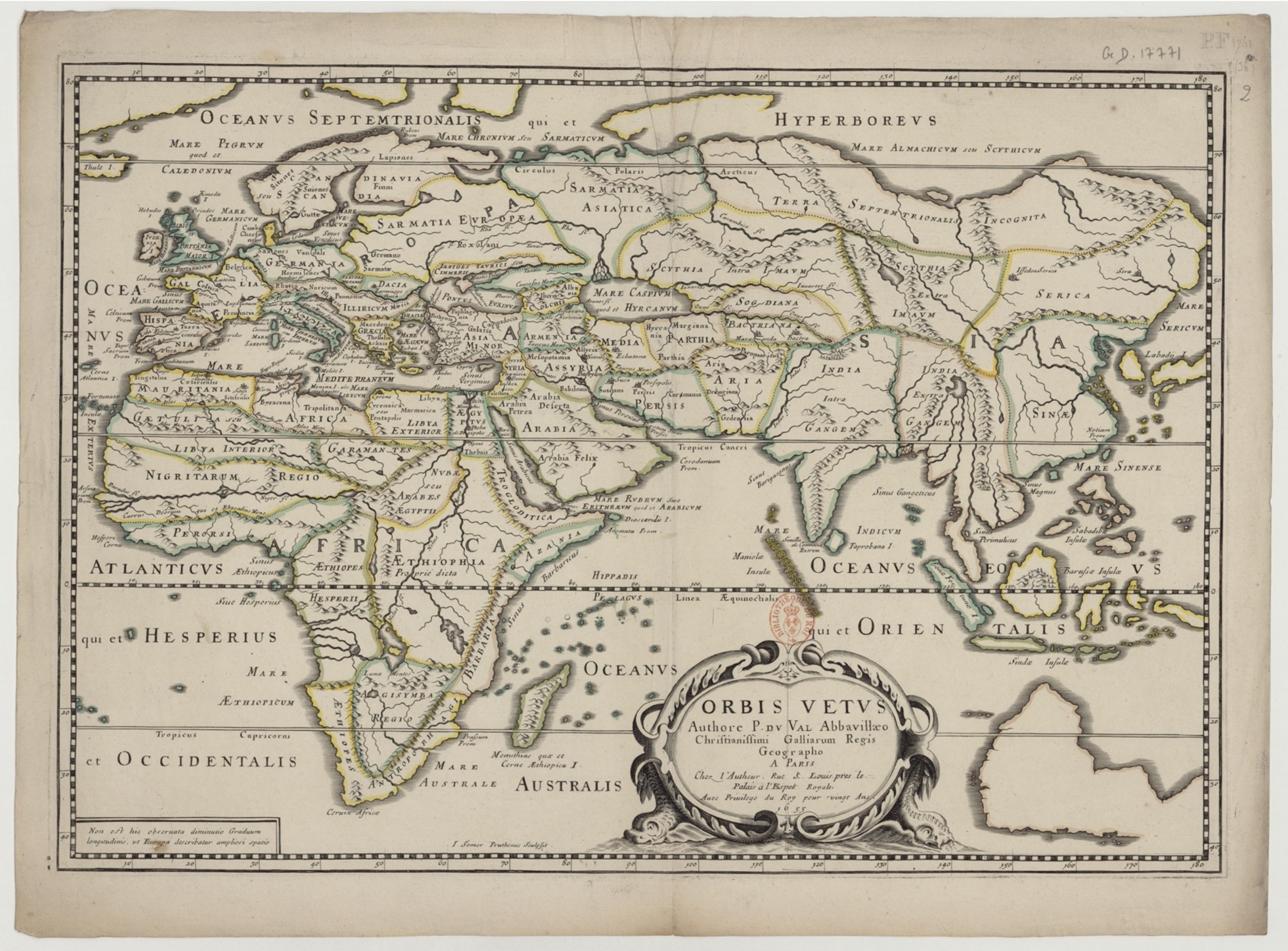

**Figure A1 | (a) Map of the Old World, 1655.** Pierre Duval. *Orbis Vetus*, 1655. Published in Paris. Copperplate, hand colored. 38.5 x 54 cm. BnF, GE D-17771. URL: gallica.bnf.fr/ark:/12148/btv1b84953694

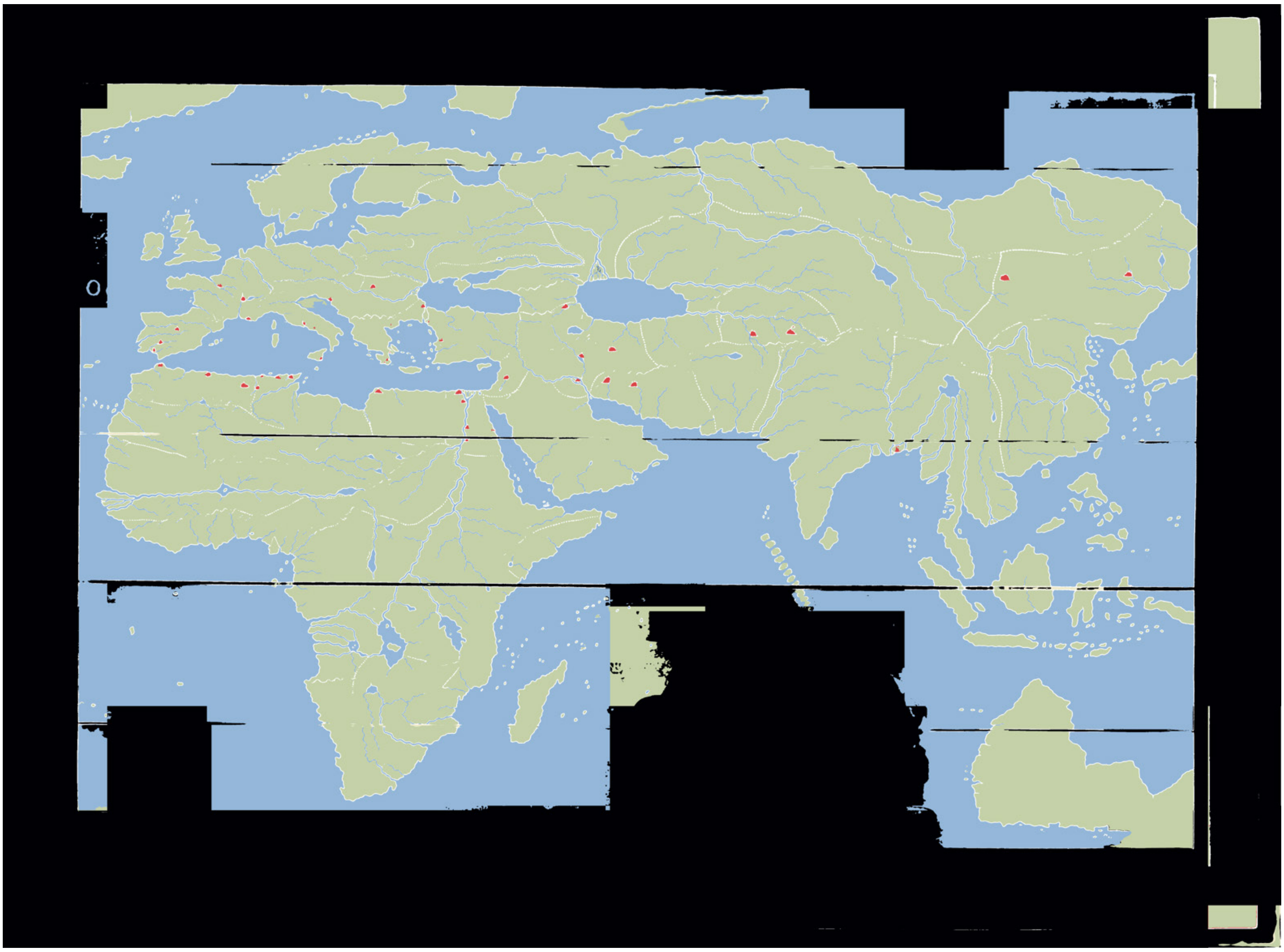

**Figure A1 | (b) Result of the semantic segmentation of Figure A1a.** *Seas and continents are correctly segmented despite their unusual shape.*

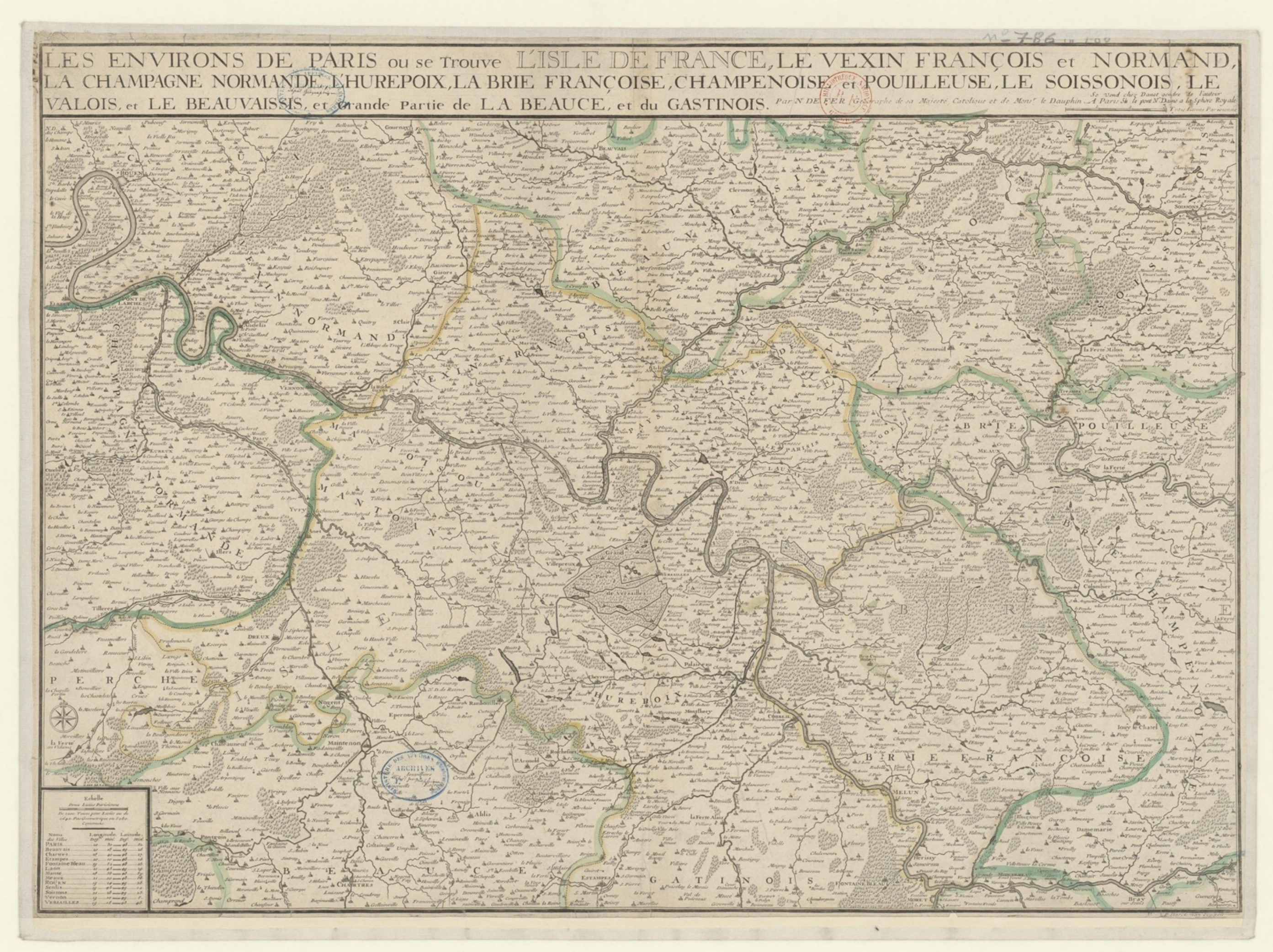

**Figure A2 | (a) Map of Ile-de-France region, 18th century.** Nicolas de Fer, P. Starckman. *Les environs de Paris [...]*, 18th century. Danet, Paris. Copperplate, hand colored. 48.5 x 67.5 cm. BnF, GE DD-2987 (786B). URL: gallica.bnf.fr/ark:/12148/btv1b530532084

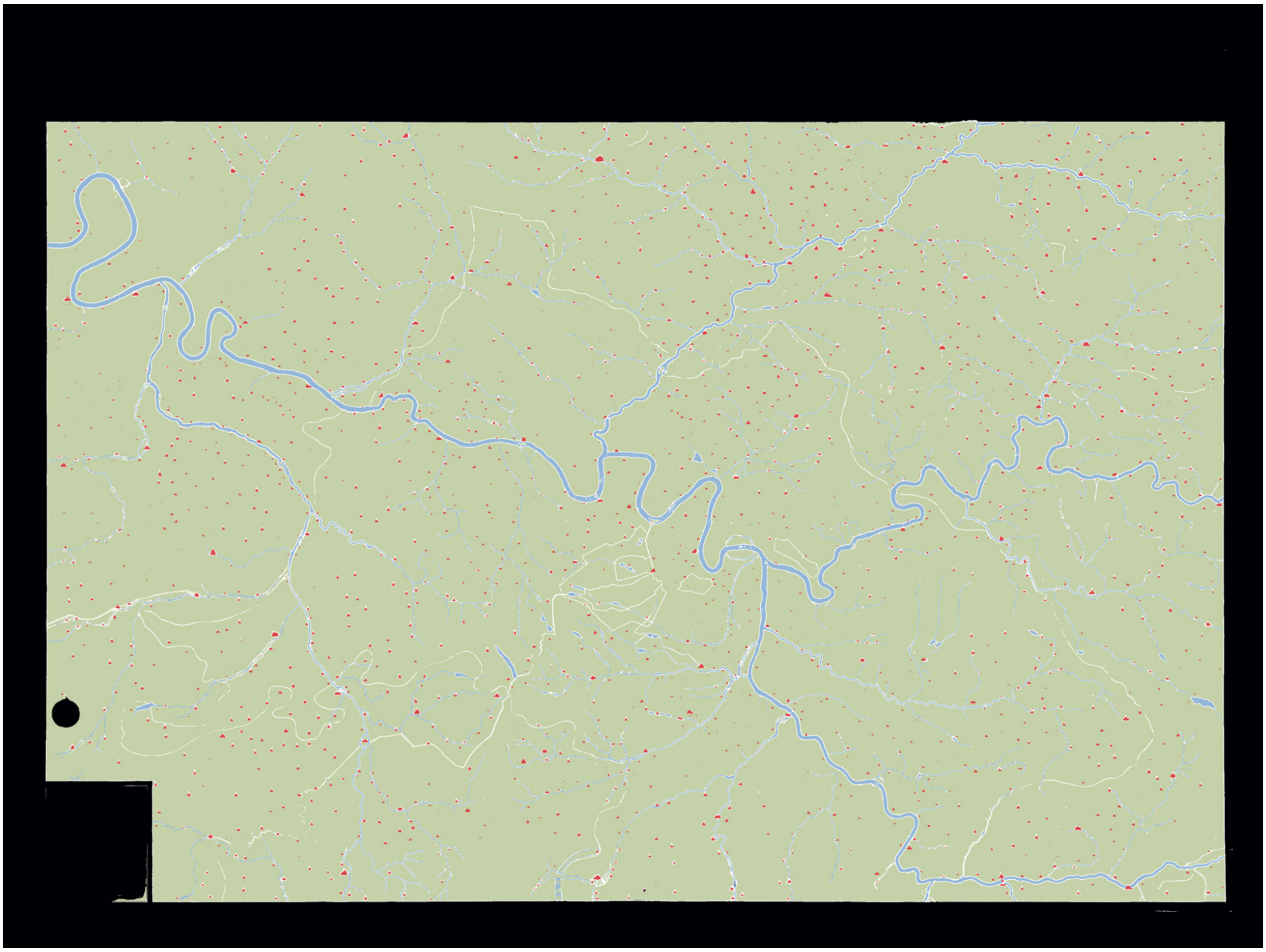

**Figure A2 | (b) Result of the semantic segmentation of Figure A2a.** *Riverways are well segmented; City icons are correctly classified as built.*

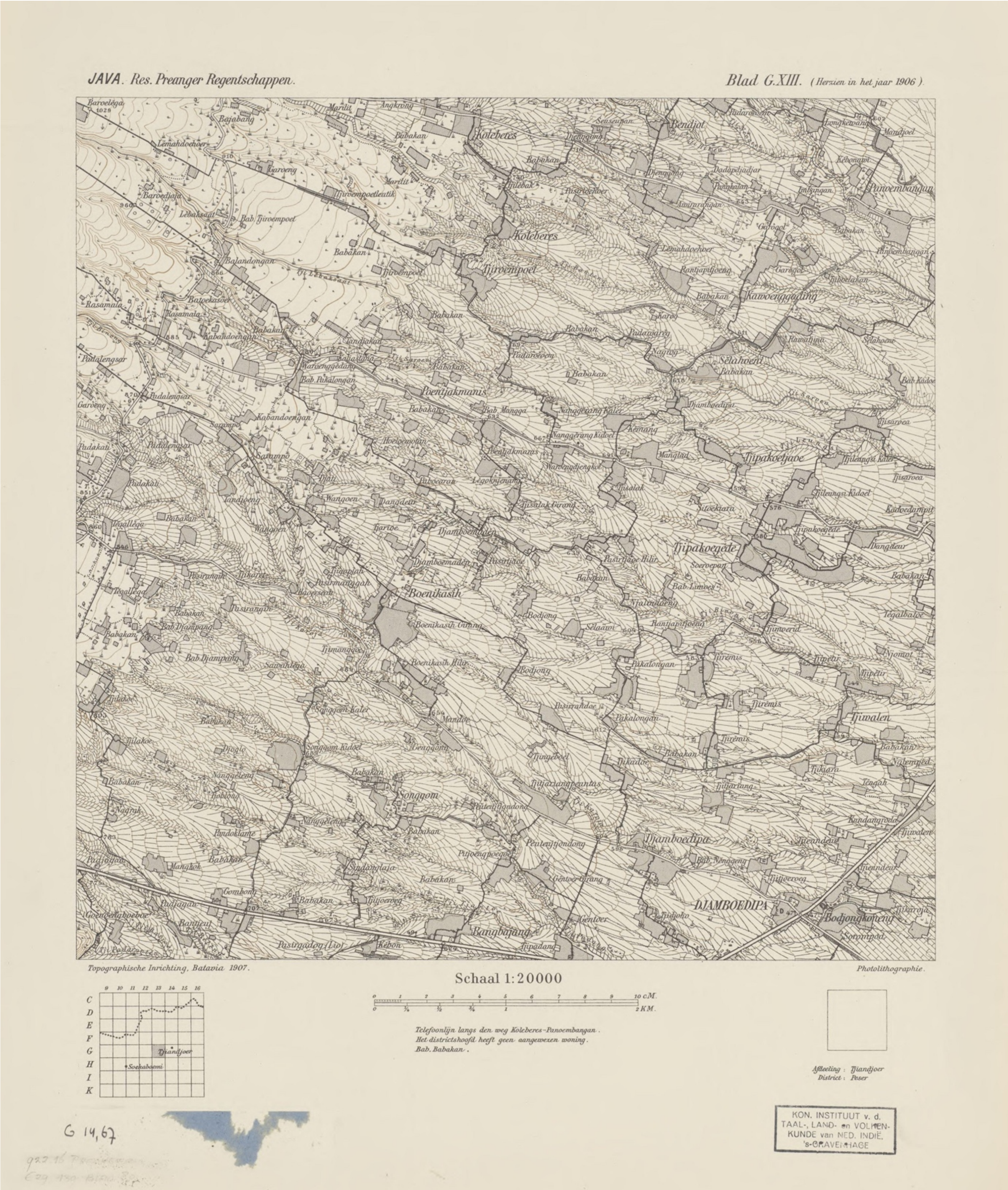

**Figure A3 | (a) Jambudipa in the topographic map of Java at the scale of 1:20,000, 1907.** *Java Res Preanger Regentschappen, Blad G.XIII*, 1907. Dutch Topographic Bureau, Batavia. Photoengraving. Leiden university library, DG 14,67. URL: hdl.handle.net/1887.1/item:815459

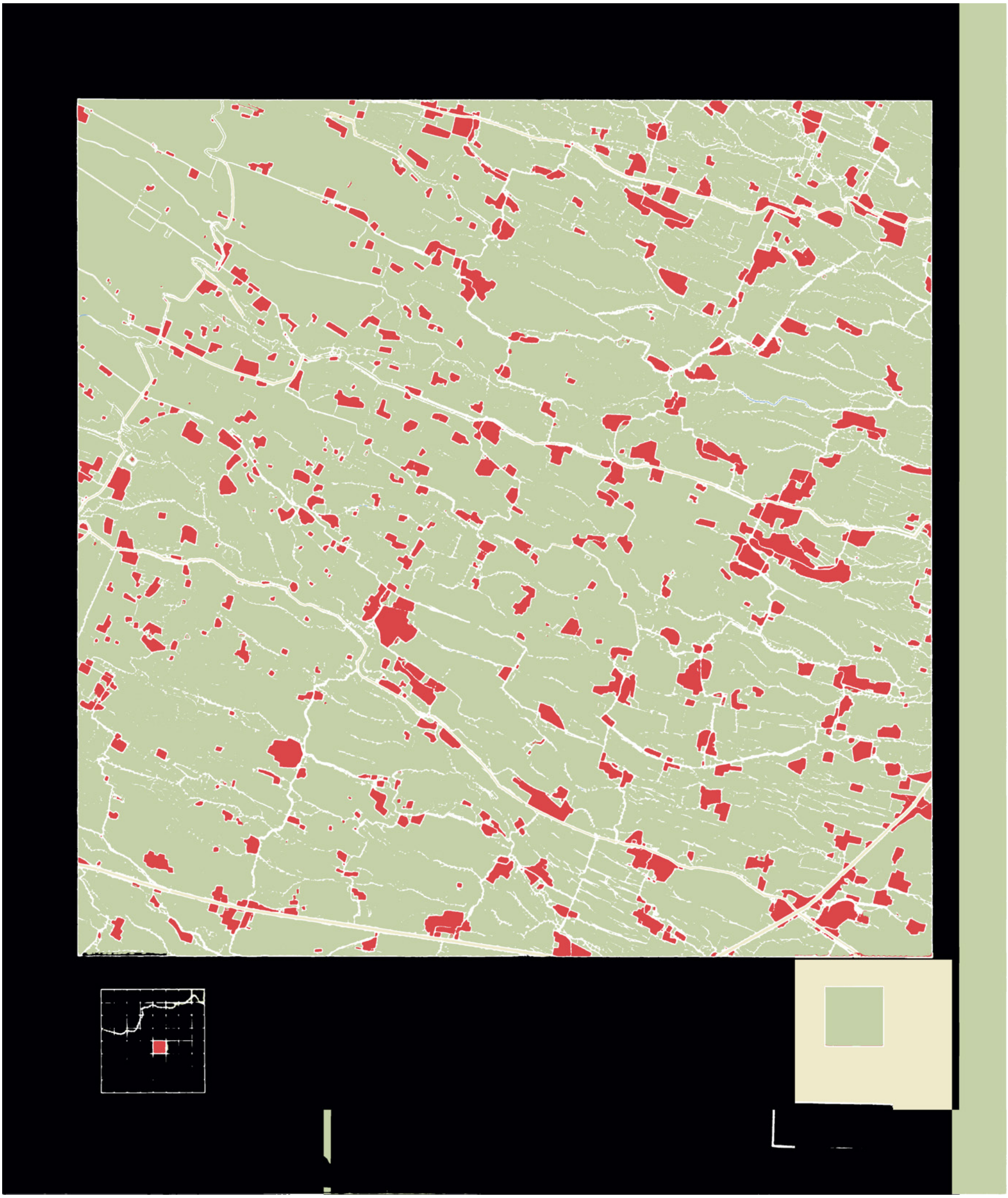

**Figure A3 | (b) Result of the semantic segmentation of Figure A5a.** *Urban areas are well segmented in spite of their unusual morphology; the recognition of contours seems impaired by the dense representation of relief.*

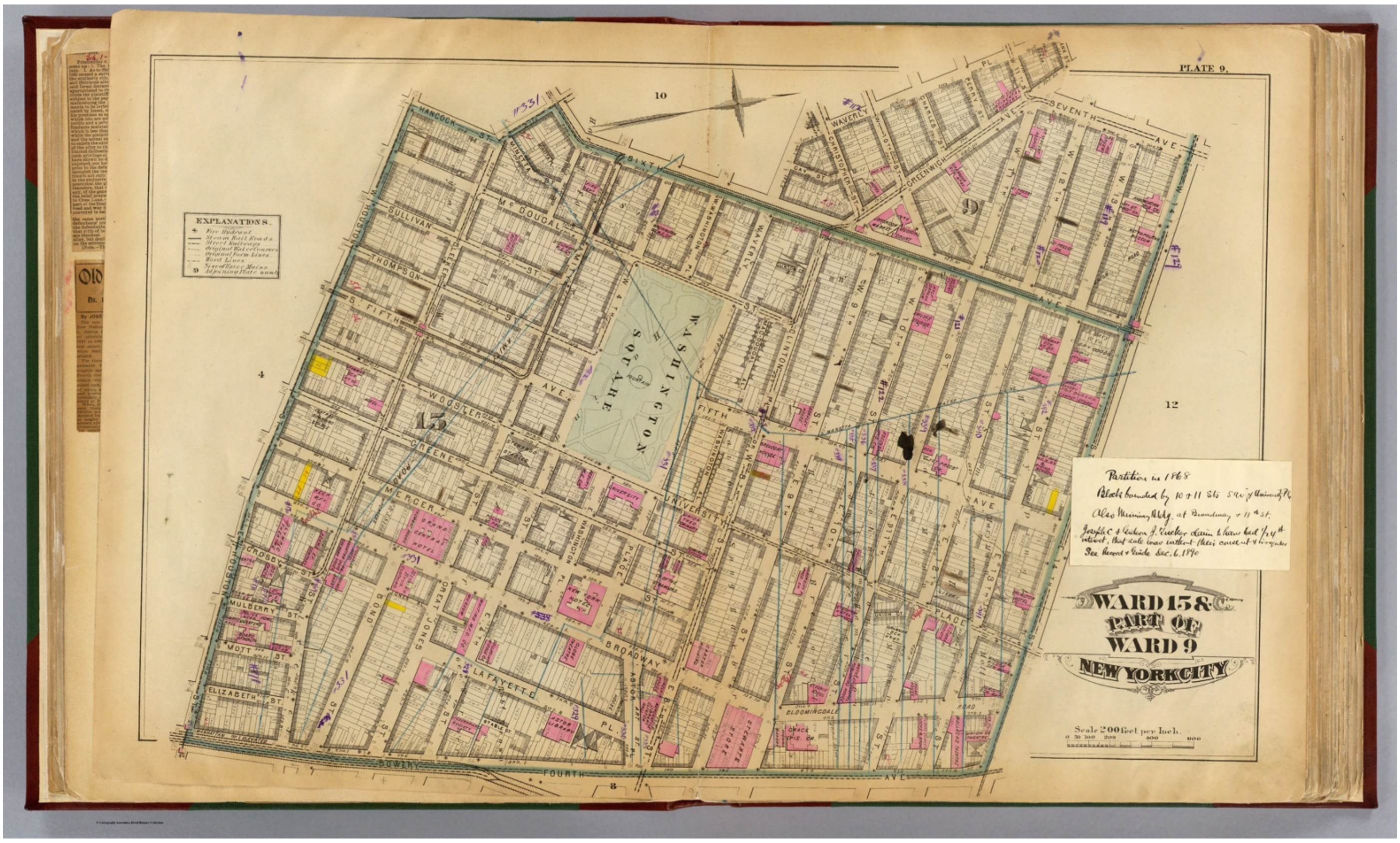

**Figure A4** | **(a) New York City Atlas, 1879.** George W. Bromley, Edward Robinson, August H. Mueller. *New York City, Ward 15 & Part of Ward 9*, 1879. Printed by F. Bourquin, Philadelphia. Published by G.W. Bromley & Co, and E. Robinson, New York. Lithography, hand colored. 46 x 67 cm. David Rumsey Collection, 2597.010. URL: davidrumsey.com/luna/servlet/detail/RUMSEY~8~1~30628~1150150

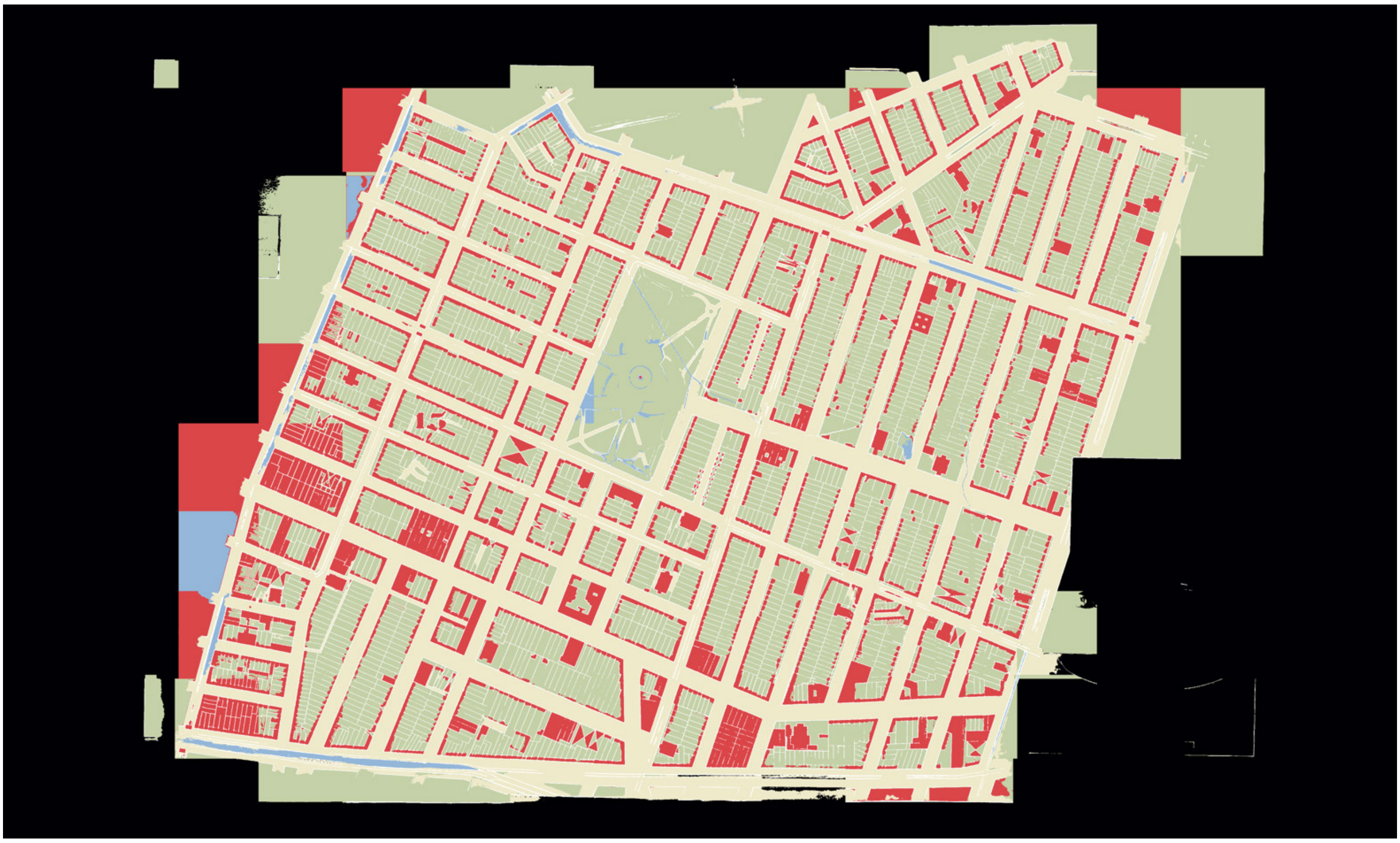

**Figure A4 | (b) Result of the semantic segmentation of Figure A9a.** *The street grid and land plots tend to be well segmented, although disruptions in the urban form, or the limited context in directly adjacent blank areas generates confusion.*

**Table A3 | List of vector objects queried from MapTiler Planet API**. The *feature* and *category* fields correspond to MapTiler's data structure. The *filter* field is adapted for legibility.

| Class* | Feature | Category | Zoom min | Zoom max | Query filter |
|---|---|---|---|---|---|
| built | landuse_residential | landuse | 10 | 11 | (class IN ('residential','suburb','neighborhood')) |
| nblt1 | landcover_grass | landcover | | | (class IS 'grass') |
| nblt2 | landcover_wood | landcover | | | (class IS 'wood') |
| nblt1 | landcover_sand | landcover | | | (class IS 'sand') |
| nblt1 | landcover_glacier | landcover | | | (subclass IN ('glacier','ice_shelf')) |
| water | water | water | | | NOT (intermittent) AND (brunnel IS NOT 'tunnel') |
| water | water_intermittent | water | | | (intermittent) |
| water | waterway | waterway | | | (brunnel IS null) OR (brunnel NOT IN ('tunnel','bridge') AND NOT (intermittent)) |
| built | building | building | 13 | | (all) |
| roadn1 | road_area_pier | transportation | | | (_geom_type IS Polygon) AND (class IS 'pier') |
| roadn1 | road_area_bridge | transportation | | | (_geom_type IS Polygon) AND (brunnel IS 'bridge') |
| roadn1 | road_pier | transportation | 14 | | (class IN ('pier')) |
| roadn1 | road_minor | transportation | 13 | | (class IN ('minor','service')) |
| roadn1 | road_major | transportation | | | (class IS 'motorway') |
| roadn1 | road_motorway | transportation | 4 | | (class IS 'motorway') |
| roadn2 | railway | transportation | 11 | | (class IS 'rail') |
| roadn1 | bridge | transportation | | | (brunnel IS 'bridge') AND (class IN ('primary','secondary','tertiary')) |
| cnt | admin_sub | boundary | 3 | | ('admin_level' IN (4,6,8)) |
| built | (all) | place | 3 | 9 | (class IN ('city','town')) |
| text | label_airport | aerodrome_label | 10 | | ('iata' IS NOT null) |
| text | label_road | transportation_name | 13 | | (_geom_type IS LineString) AND (subclass IS NOT 'ferry') |
| text | label_place_other | place | 8 | | (_geom_type IS Point) AND (class IS null OR class NOT IN ('city','state','country','continent')) |
| text | label_place_city | place | | 16 | (_geom_type IS Point) AND (class IS 'city') |
| text | label_country_other | place | | 12 | (_geom_type IS Point) AND (class IS 'country') |
| text | label_water | water_name | 10 | | (_geom_type IN (Polygon,LineString)) |

* nblt = non-built (1 base, 2 forest), roadn = road network (1 base, 2 railway), cnt = contours

# Chapter 5

## *Maps as Pictures: Framing & Composition*

The previous chapter presented the method for segmenting digitized map images into a simple set of semantic classes. The trained semantic segmentation model was then applied to the 99,715 maps of the ADHOC Images corpus. The result is a vast set of information documenting the history of map semantics, understood as broad classes of geographic objects. The present chapter aims to study which objects mapmakers tend to emphasize and which, by contrast, they tend to set aside through processes of framing and image composition. In this prospect, it conceives and implements approaches for the investigation of spatial relationships, such as co-occurrences, and centering. The research will also discuss the tension between rational expectations derived from map scale, graphic choices dictated by the imperatives of cartographic representation, and the prioritization of certain classes of geographic objects. Moreover, this chapter will examine the extent to which maps can be categorized into differentiated semantic types, characterized by distinctive compositional patterns. Ultimately, it will discuss the tension between figurative choices and the geographic referential function of maps.

By reverse engineering the focus of mapmakers on certain classes of geographic objects, this research prolongs the study of *spatial attention*, in Chapter 3. Here, the analysis concentrates not on the geographic locations depicted, but on the objects represented. It also adopts a closer level of observation, by studying maps as individual images. While recognizing maps as *intentional* cultural products that convey the point of view of their creator, this chapter also acknowledges the role of cultural norms and the replication of cultural conventions in the evolution of cartographic representation.

## 5.1 On the relationships between art and cartography

Although portolan maps are often described as a specific early type of nautical charts, many of them functioned, first and foremost, as objects of prestige and artistry (T. Campbell, 1987b, p. 440). Usually drawn on precious *vellum* parchment, they could be richly decorated with gold leaf and colored with precious pigments. Consequently, many were too valuable to be embarked on potentially perilous sea voyages and saw the deck of a ship only in the adjacent paintings that adorned the *palazzi* where they were exhibited. Early geographical prints were often displayed in the entrance of private mansions, or formed the decorative theme of a reception room, thereby signaling the learned and cosmopolite spirit of the host (Woodward, 1996, pp. 79–87). One can hardly consider the history of Renaissance cartography without also taking into account the history of landscape iconography. With the rise of copperplate engraving, in the 16$^{th}$ century, maps became even more closely related to iconographic prints, as both were generally produced by printmakers closely connected to the art and goldsmith trades (Woodward, 1996, p. 38). Several of the most influential artists of the Renaissance, including Albrecht Dürer (1471–1528), and Leonardo da Vinci (1452–1519), were also cartographers. Conversely, several mapmakers like Abraham Ortelius (1527–1598) and Gerhard Mercator (1512–1594) began their careers as pictorial artists (Rees, 1980).

The proximity between cartography and art, however, is not confined to distant historical times; map making has consistently evolved alongside artistic practice. Modern pictorial maps illustrate this continuity as, for instance, the map of Florida by Eleanor Foster (active 1931–1936), shown in Chapter 2, Figure A4, the *Picture Book of the States* by Berta and Elmer Hader (1890–1976, 1889–1973) or, more recently, the panoramic mountain maps of Pierre Novat (1928–2007), which you almost certainly saw if you have ever been to snow holidays in the Alps (Novat et al., 2019). Like art, cartography has been influenced by the development of printing technologies. The advent of color printing and chromolithography, in the mid–19$^{th}$ century, significantly expanded the presence of maps in popular culture (Petitpierre et al., 2024; Ristow, 1975). Throughout the 20$^{th}$ century, cartographers have engaged with artistic and sociocultural movements, such as Surrealism, Situationism, and Conceptual art. Contemporary cartography continues to be influenced by theories from the arts and design (Cartwright et al., 2009; Cosgrove, 2005; Ribeiro & Caquard, 2018).

The primary purpose of highlighting the similarities between art and cartography is to demonstrate that both are cultural *products*. From this perspective, the history of cartography, like art history, can be studied in terms of themes, genres, and movements. Common themes in cartographic scholarship include nautical charts (Cook, 2006; Randles, 1988), road mapping and railroad maps (Akerman, 2006; Delano-Smith, 2006; Musich, 2006), and city maps (Brown & Hunt, 2019). Drawing on sociocultural critique, Matthew Edney proposed the broader concept of *cartographic* *mode*, defined as "the combination of cartographic form and cartographic function" (Edney, 1993, p. 58).

While *function* is manifested in specific components of cartography—such as the classes of objects depicted, map scale, theme, and geographic coverage—*form* can be described in terms of "medium, line, colour and symbolization, [...] composition, framing, and perspective" (Cosgrove, 2005, p. 1) as well as other aspects, such as "light"—or rather shadow—for the depiction of relief. The components of form in art and cartography are therefore kindred. While digital art history has emerged as a set of methodologies to reveal facets of art history that only appear at a macroscopic scale (di Lenardo et al., 2016; Drucker, 2013; Impett & Moretti, 2017; Manovich, 2015), the concepts of cartographic frame and composition have never been explored quantitatively.

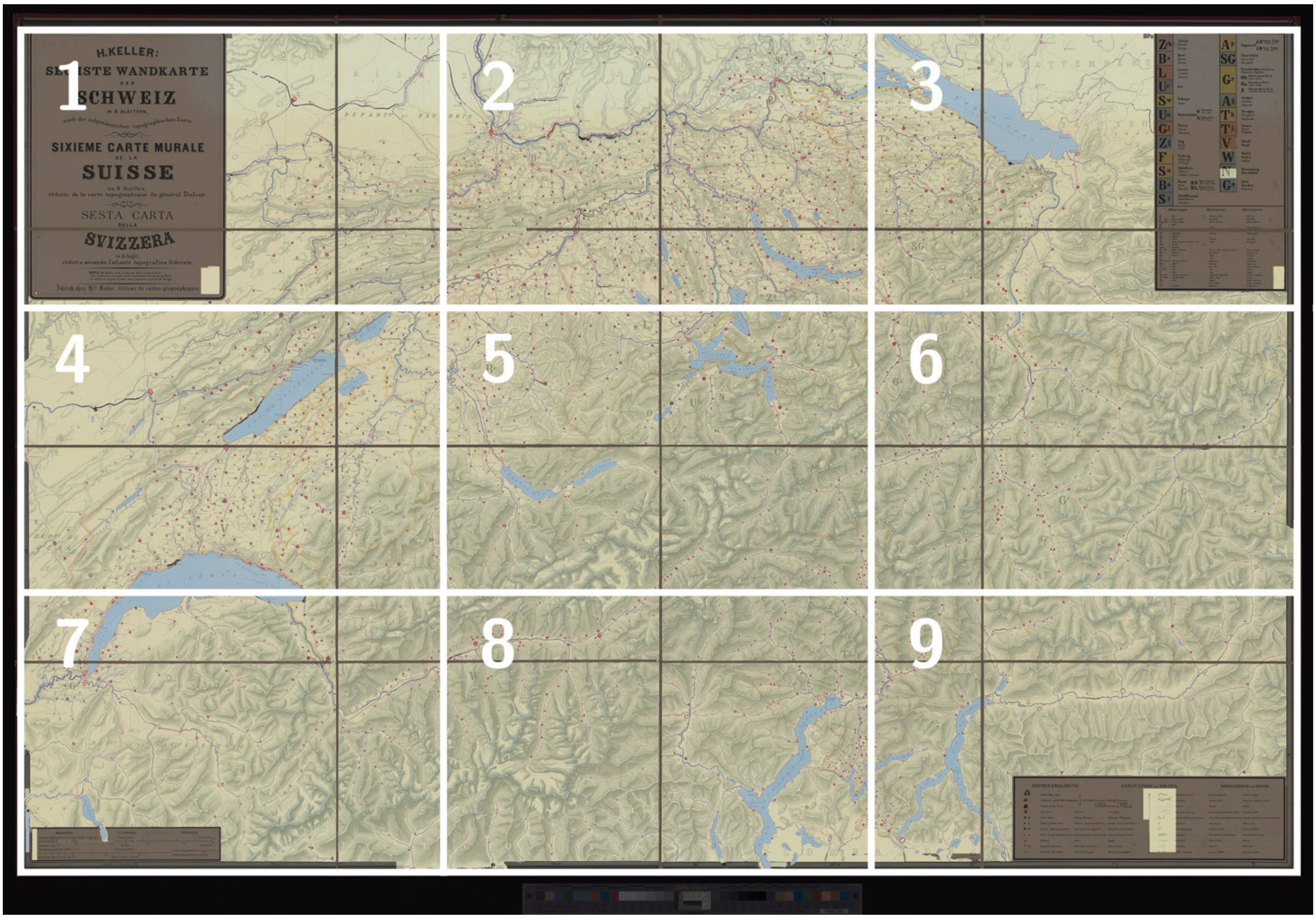

**Figure 1 | Division of a map into 9 quadrants.** The original map and the predicted segmentation mask are superimposed in this Figure. The map is partition into 9 identical quadrants (or "nonants"), whose outer boundaries are defined from to the projection profile of the geographic content, after exclusion of the background class. The original map is an exemplar of the semantic type #2 (chorographic). Source of the original map: Heinrich Keller and Rudoft Leuzinger. *Sechste Wandkarte der Schweiz*. 1875. H. Keller, Zürich. Zentralbibliothek Zürich, 5 Hb 06: 8. 114 × 175 cm. doi: 10.3931/e-rara-35673.

## 5.2 Composition, or maps as pictures

### *The quadrant framework*

The preliminary statistical investigation of semantic classes, in Chapter 4, revealed that maps do not represent the world uniformly. Urban environments, for example, are overrepresented, whereas oceans are underrepresented. The remainder of this chapter will study the discrepancy between classes of geographic objects by investigating how, within a map image, certain features are *emphasized*. The concept of *emphasis* implies the existence of hierarchical relationships between the objects depicted. Hierarchy could presumably be enforced by graphical means, but also through *spatial relationships* within the image, i.e. through image *composition*. The following paragraphs will adopt the latter angle of investigation. Composition implies that geographic objects on maps may be organized in a way that is intentional. This does not necessary imply that objects are displaced; there are many ways to enforce composition, one of which is *the choice of map frame*. For instance, when mapmakers define the map frame, they tend to do so in a way that the principal geographic subject is *centered*. Furthermore, they tend to place obstructing layout elements, such as the cartouche, the legend, or the graphical scale, so that the objects considered most important are not occluded. Consequently, the analysis of map composition, the relative position of the objects of the map and their spatial relationships to one another reflects the *intentionality* of the mapmaker, and design choices. The larger part of this chapter will be dedicated to "reverse–engineering" these operations in order to study the historical evolution of emphasis at scale. Furthermore, I hypothesize that the map frame, and potentially its orientation, are also set as to maintain an aesthetic and semantic equilibrium across the image. This typically implies the manifestation of axial relationships, perhaps even symmetries.

The analytical approach is based on a simple framework that entails dividing each map image into nine evenly shaped "quadrants"[1], as shown in Figure 1. Each quadrant results from partitioning the map's geographic extent into three equal segments, both vertically and horizontally. The resulting quadrants are not necessarily square; they follow the proportions of the map's rectangular shape. This simple framework operationalizes by design the notions of centering, axial symmetry, and map frame. It also permits the analysis of spatial co-locations among semantic classes.

[1] The etymologically correct word would be "nonant". However, because "nonant" is very unusual, the more common term of quadrant is used here.

### *Spatial co-locations*

Figure 2 reports co-locations among semantic classes, computed as the spatial cross-correlations within map quadrants. The four geographical classes—built, non-built, water and road network—form two distinct groups. First, built and road-network classes—both markers of urban environments—are rather strongly correlated (r = 0.67). Water and non-built are independent of the two former and negatively correlated with one another (r = –0.56). The non-built class is also negatively correlated with the built class (r = –0.34), whereas the negative relationship between water and built is weaker (r = –0.21). Unsurprisingly, the background is negatively correlated to all other classes. Contours, however, are often associated with built (r = 0.33) and road-network (r = 0.30) classes but seldom with water (r = –0.23). The latter observation is particularly noteworthy as the contours class arguably signals informational density. Indeed, an image containing numerous discrete geometric objects necessarily displays more contours. For instance, a city map that delineates individual building footprints will present more contours on average than an identical map of the same area limited to city blocks. Similarly, a plan *par masse de culture*[2] that distinguishes individual fields will include more contours than one depicting undivided farmland. Accordingly, the association between contours and urban environments indicates a higher level of detail, and, by extension, the comparatively greater importance ascribed to these territories. Conversely, water areas are rarely subdivided.

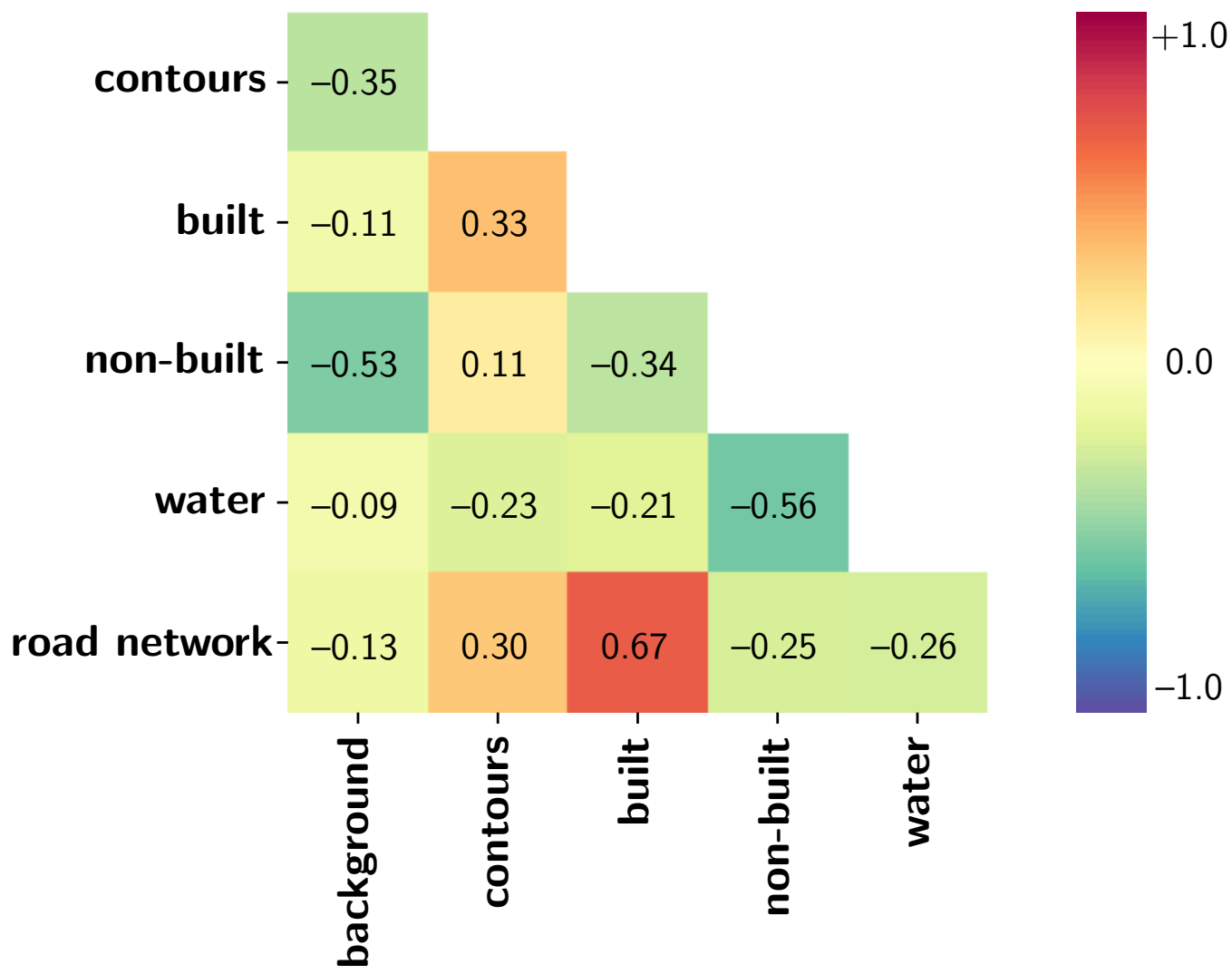

**Figure 2 | Matrix of co-location among the semantic classes** (n = 897,435), where each statistical sample is a map quadrant (see Fig. 1). All p-values are statistically significant ($p_{val} \ll .05$). *The built and road-network classes are often co-located.*

[2] Forerunners of cadastral plans.

### *Spatial semantic relationships*

The average semantic content of each map quadrant is shown in Figure 3. As anticipated, several classes are frequently located at the center of the image (quadrant nr. 5), chiefly the built class, the contours, and the road network; the non-built class is also regularly centered. By contrast, the background class is usually found in the map periphery (quadrants nr. 1, 3, 7, and 9). The water is seldom centered. This observation is a first cue supporting the interpretative framework that the map is a *designed image*, where centering may denote emphasis. Besides intention, non-randomness indicates the presence of *conventions of representation*. For instance, the emphasis on urban environment (built and road network) appears a largely shared and reproduced convention. Each time a map maker routinely replicates this cultural trait, *it structurally reinforces its normative character* and the conveyed viewpoint, for instance the greater cultural importance accorded to urban spaces compared to rural ones, or the emphasis on landmasses at the expense of marine environments.

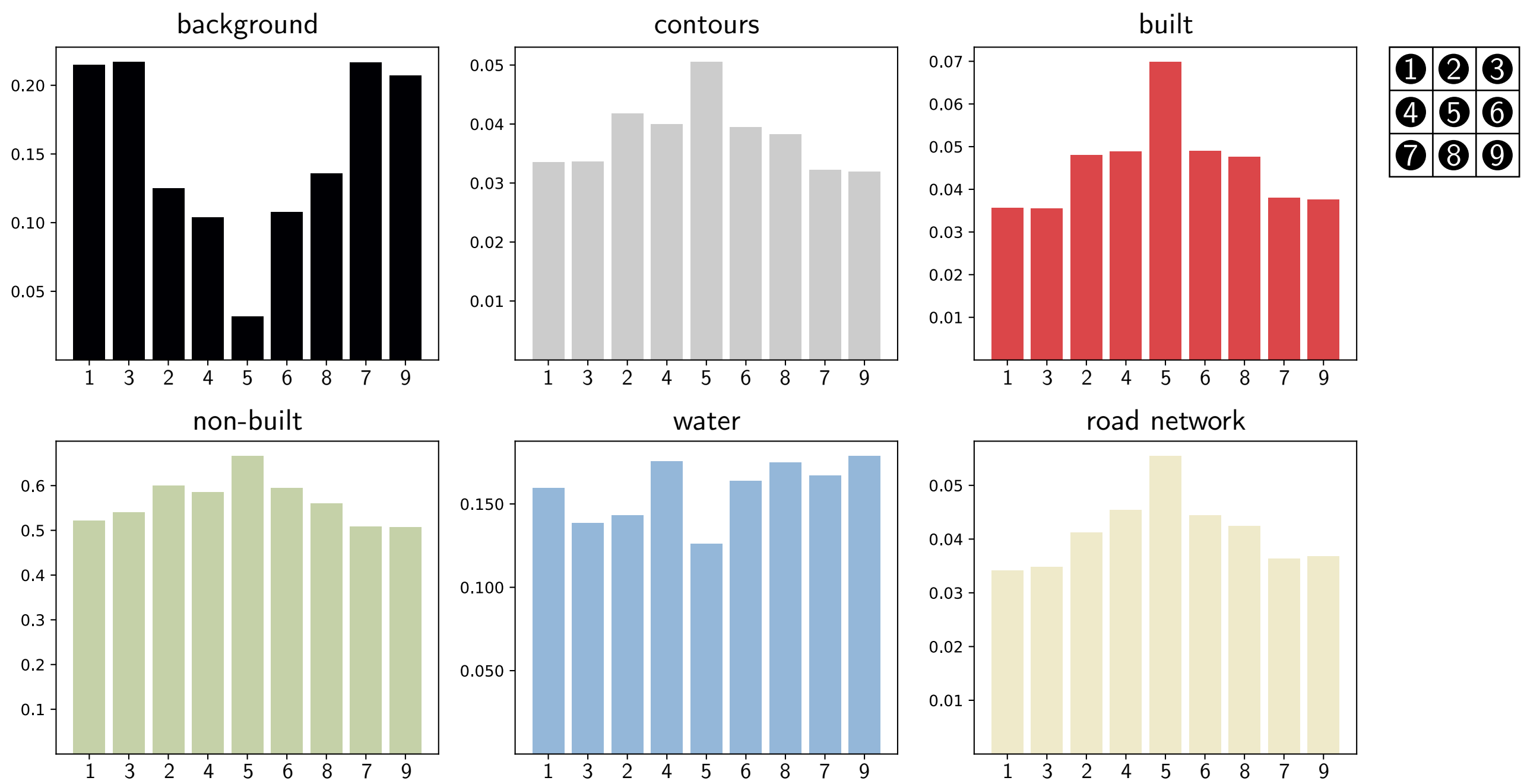

**Figure 3 | Relative share of each semantic class per image quadrant** in the entire ADHOC Images dataset. The relative positions of image quadrants 1–9 are depicted in the upper–right diagram. Quadrants are arranged linearly along the horizontal axis of the bar diagram, so that peripheral quadrants are positioned at the beginning or the end, and central quadrants are located in the middle. Confidence intervals of the means are negligible. *The built, contours, and road network semantic classes are often centered; the background is generally located in the corners.*

In Figure 3, classes that are tendentially centered, particularly the built class, also appear more prevalent in the quadrants nr. 2, 4, 6, and 8. This suggests that, in addition to centering, the map subject may be horizontally or vertically aligned, indicating that the composition is articulated around a *central cross* pattern.

The examination of pairwise correlations across image quadrants can help inform spatial relationships. These relationships can be modeled as a weighted graph with nine nodes and 36 edges, each node representing a quadrant and each edge the statistical correlation between two quadrants. In this perspective, strong horizontal axial relationships would be translated in the weight of horizontal edges such as (4, 5), (5, 6), (4, 6), etc. Figure 5a provides a graphical summary of this weighted graph, computed from the 99,715 maps in the ADHOC Images corpus. Inspection of the figure shows the prominence of the central cross, consistent with the relative frequencies reported in Figure 3. Pronounced circumjacent diagonals and the outer square are also evident. Figure 4 highlights these new structures.

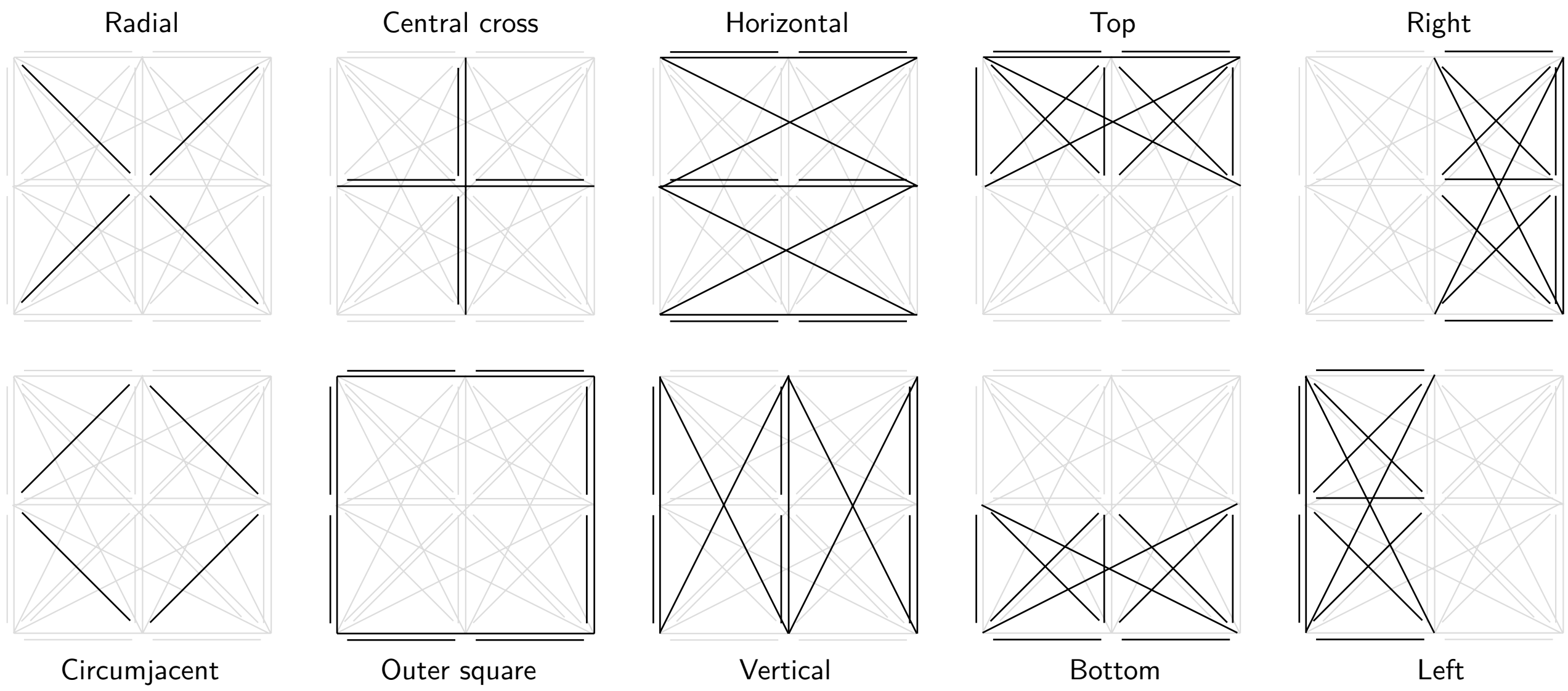

**Figure 4 | Spatial relationship hypotheses.** Graph edges correspond to the various conditions mentioned in the analysis of composition and in Table 1, below. Edges associated with the labeled condition are depicted in a darker shade

Figure 5b reports the relationship between spatial correlation and the distance between quadrant centers. In the absence of any composition effect, or if composition only entailed centering, the correlation between the semantic content of two quadrants would depend only on the distance between them. However, in the present case, 17.3% of the variation cannot be explained by distance. This outcome suggests that, in addition to centering, other spatial relationships participate in map composition. Here, five such relationships are considered: centrality, radiality (or circumjacency), horizontality (or verticality), laterality (left vs. right), and superoinferiority (top vs. bottom).

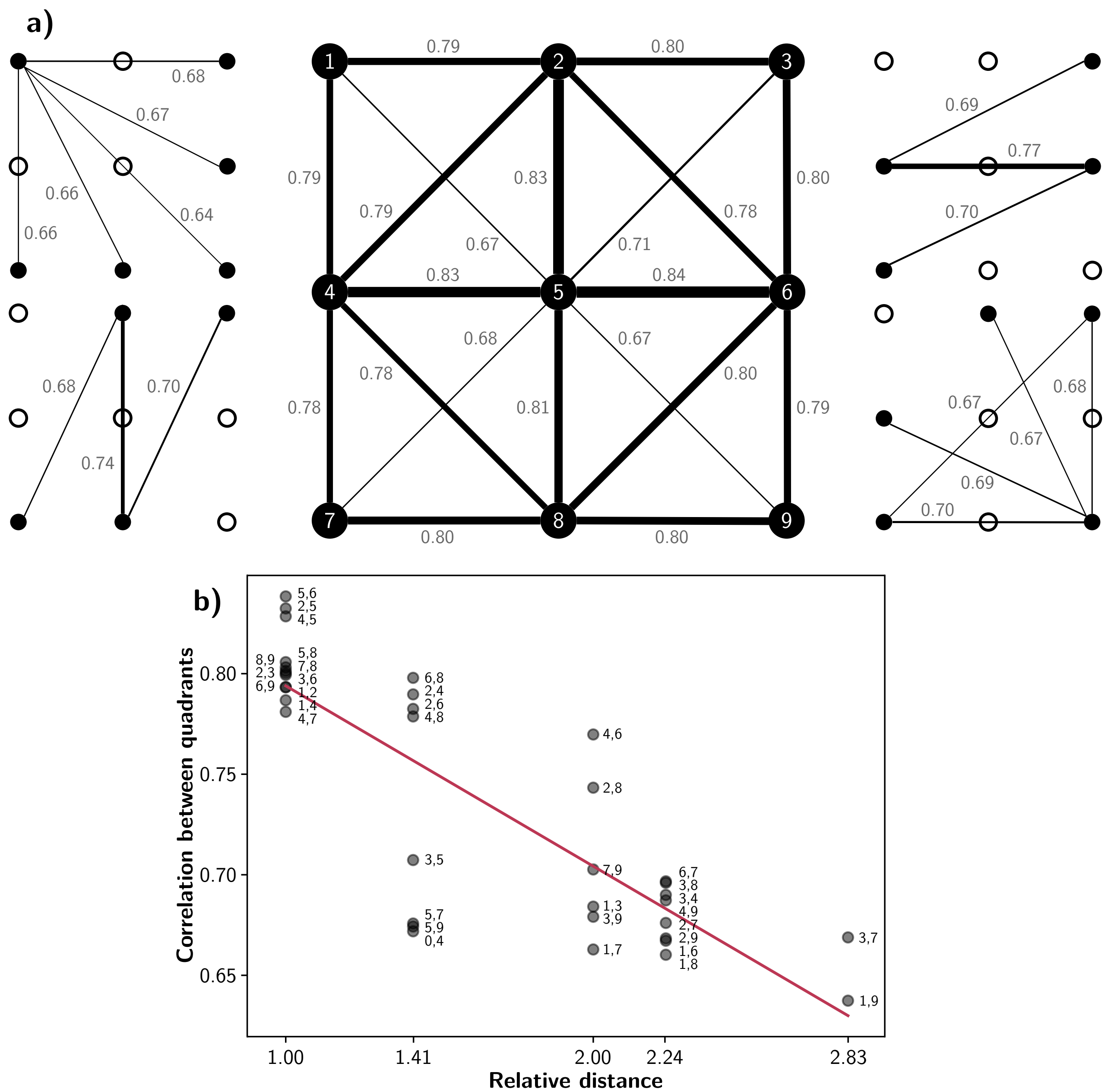

**Figure 5 | Composition. Semantic relationships between image quadrants. (a)** Visual representation of semantic relationships between image quadrants. Each node in the graph corresponds to one quadrant, whereas the edges represent correlations between those quadrants. The values reported are Pearson's correlation coefficients (n=99,715). All correlations are statistically significant. The subgraphs to the left and right of the central graph are provided for readability and display long-distance relationships. Line width varies according to the strength of the spatial semantic relationships and is not downscaled in the subgraphs. **(b)** Dependency of spatial semantic relationships to distance. The center of each quadrant is considered for distance computation; the image width and height are set to 2.5 distance units. Number pairs (e.g., 1,2) correspond to graph edges as defined in subfigure (a). The central red line was obtained by linear regression. *The central cross (5–6, 2–5, 4–5, 5–8) appears the strongest spatial relationship.*

Table 1 presents the results of the statistical tests, ordered by significance. The most robust outcome is the prominence of circumjacent relationships over radial ones ($\Delta r_{pearson} = .105$). This finding suggests that features are arranged regularly around the map center. Conversely, radial diagonal relationships appear weak. The test also confirms the salience of the central cross ($\Delta r_{pearson} = .046$). Within this cross, horizontal axiality seems notable ($\Delta r_{pearson} = .018$). Consistently, long-range horizontal relationships are also observed across the entire image ($\Delta r_{pearson} = .016$), implying an asymmetry around the vertical axis.

**Table 1 | Results of the statistical tests on composition**. The spatial relationship hypotheses are described in Fig. 4. Student's paired, and unpaired t-tests are used alternatively, depending on the hypothesis and the related graph edge structure. $\Delta\, r_{pearson}$ refers to the mean absolute difference between correlation values. Results are ranked by statistical significance; significant outcomes (for $\alpha = .01$) are marked with an asterisk (*).

| Hypothesis | n | test | statistic | $\Delta\, r_{pearson}$ | p-value |
|---|---|---|---|---|---|
| *Circumjacent > Radial | 80 | unpaired | 38.8 | .105 | $5 \cdot 10^{-53} < .01$ |
| *(Horizontal > Vertical) \| long-range | 140 | paired | 4.5 | .016 | $5 \cdot 10^{-22} < .01$ |
| *Central cross > Outer square | 180 | unpaired | 5.9 | .046 | $9 \cdot 10^{-9} < .01$ |
| *(Horizontal > Vertical) \| central cross | 60 | paired | 6.1 | .018 | $6 \cdot 10^{-7} < .01$ |
| *(Bottom > Top) \| water | 240 | paired | 4.0 | .048 | $4 \cdot 10^{-5} < .01$ |
| Horizontal > Vertical | 260 | paired | 1.8 | .015 | $.039 < .05$ |
| Right > Left | 240 | paired | 1.8 | .011 | $.036 < .05$ |
| Top > Bottom | 240 | paired | – | .001 | .47 |

The hypothesis of a more general tendency to horizontality, defined by the edges highlighted in Figure 4 and tested against a hypothesis of verticality, appeared significant only at a .05 significance level. Similarly, the existence of laterality, where spatial relationships would be stronger on the right side of the map image compared to the left, is not verified. The matching hypothesis, following which relationships in the upper half of the image would be stronger than those in the lower half, is clearly rejected. When the hypothesis is restricted to the water class, however, the inverse relationship appears to be confirmed, with a relatively high effect ($\Delta r_{pearson} = .048$). This finding is consistent with the results of Figure 3, which indicated the higher representation of the water class in quadrants nr. 7, 8, and 9, located in the lower third of the map image, compared to the upper or middle quadrants.

To summarize, we observe that semantic features tend to be organized following a central cross pattern, with the horizontal alignment being particularly pronounced. Denser urban content, or landmasses, are generally found at the center of the map, whereas other features are arranged circularly around the map center. Both results are also coherent with the occurrence of symmetric relationships on either side of the vertical axis. Water typically occupies the periphery of the image, with a modest prevalence in the lower third.

### *Two case studies: Lausanne and Paris*

Most of these compositional effects do not arise from the distortion of geographic information. For instance, the frequent presence of water at the bottom of the image cannot be attributed to a systematic attempt by mapmakers to relocate water features downward. Beside the placement of layout elements on the image, and the determination of the map's confines, mapping also involves the choice of map orientation. Whereas most modern maps are conventionally oriented with the north positioned at the top, ancient maps were commonly oriented toward the Orient, which could correspond to the east or the southeast. The change in the map's orientation to face North is a gradual development, based on mapping choices influenced by the progressive evolution of conventions of representation, around the turn of the 17th century (Favier, 2003). Most importantly, despite the existence of cartographic convention, it is not uncommon to tilt a map slightly to preserve visual balance. For example, the map of Lausanne shown in Figure 6, dating from 1900, is tilted approximately 21° from the north axis. This slight inclination is sufficient to center the city and mitigate the untidy impression produced by its irregular urban form, whose organic morphology is inherited from the medieval period, and influenced by the strong slope. In addition, it reinforces the regularity of the representation by aligning its horizontal orientation to the lake side and the parcel grid inherited from the Roman period (di Lenardo, 2025).

It is also worth mentioning from this example that the map frame is not constrained by municipal boundaries. Instead, both its extent and orientation appear to have been defined purposefully to maintain visual balance around the centered settlement. The resulting composition enforces horizontal equilibrium, and underlines the vertical succession of geographic objects, with the forest to the North, the main settlement in the center, and the village of Ouchy bordering the lake to the south. The graphical rendering of roads, and riverways traces vertical links between the three stages. In addition to the lake, the horizontal visual equilibrium is maintained near the center of the image by the marked rendering of the railways, and the darker beige parcels, on either sides of the city, and to the south, corresponding to vineyards (Petitpierre, 2025b). None of these geographic elements would have appeared to create a horizontal visual balance if the map was oriented straight north. A last observation is the gradient of semantic density: the center of the image is occupied by denser plot fabric, while larger plots—corresponding to lower intensity agricultural lands like meadows and pastures—extend southward and northward. In this respect, it is noteworthy that the cartouche bearing the map title should be deliberately placed at the top of the image, where geographic information is relatively sparse and "dispensable". Likewise, the graphic scale is positioned at the bottom left, in the lake, where it minimally interferes with urban-related geographic information.

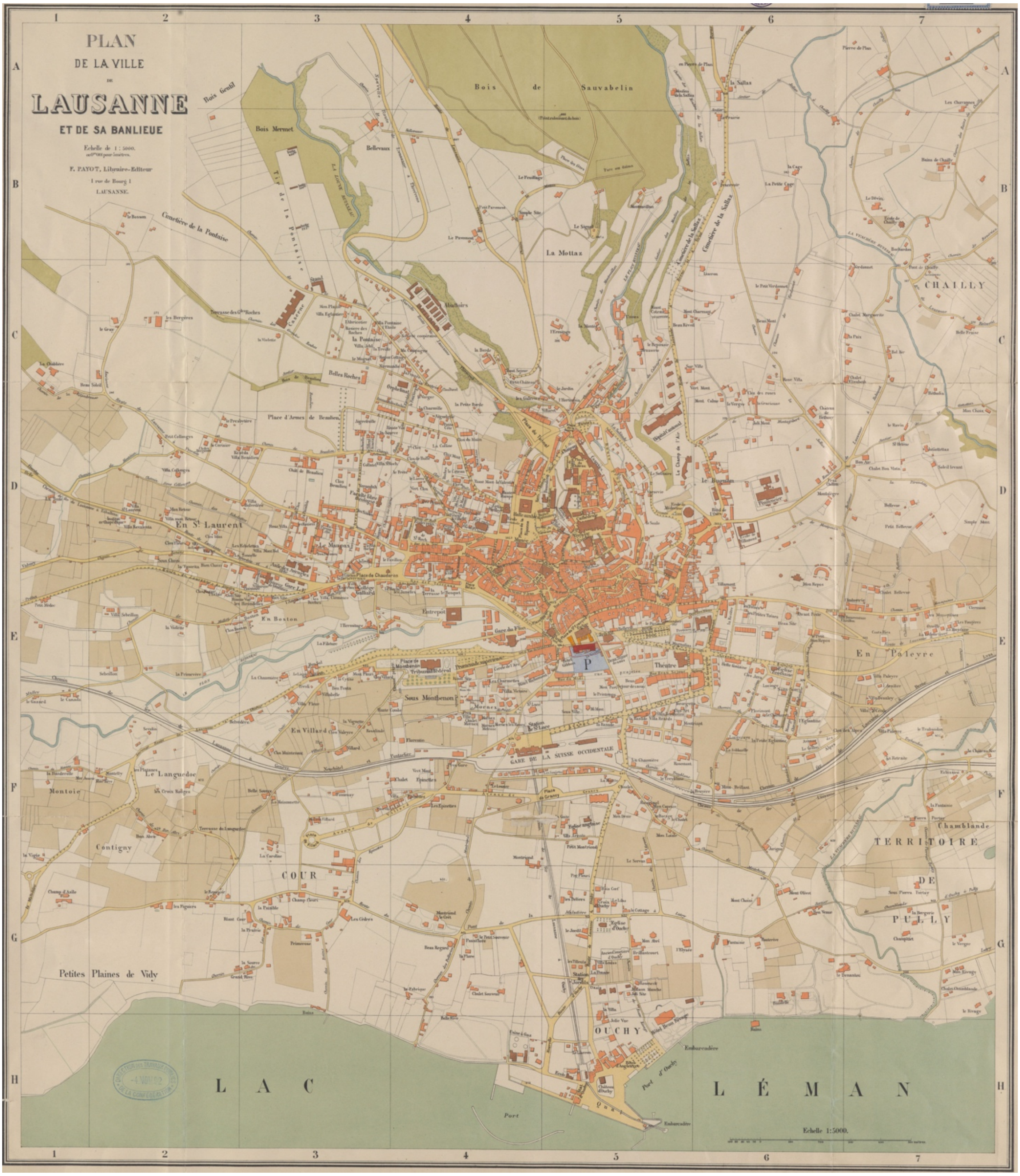

**Figure 6** | **Plan of Lausanne, 1900**, edited by F. Payot. *Plan de la Ville de Lausanne et de sa banlieue*, 1900. Wurster, Randegger & Co., Winterthur. Bibliothèque de Genève, 3 H8/Lau. *The plan is tilted 21° from the north axis, the city centering and framing maintain the visual balance between the left and right parts of the image.*

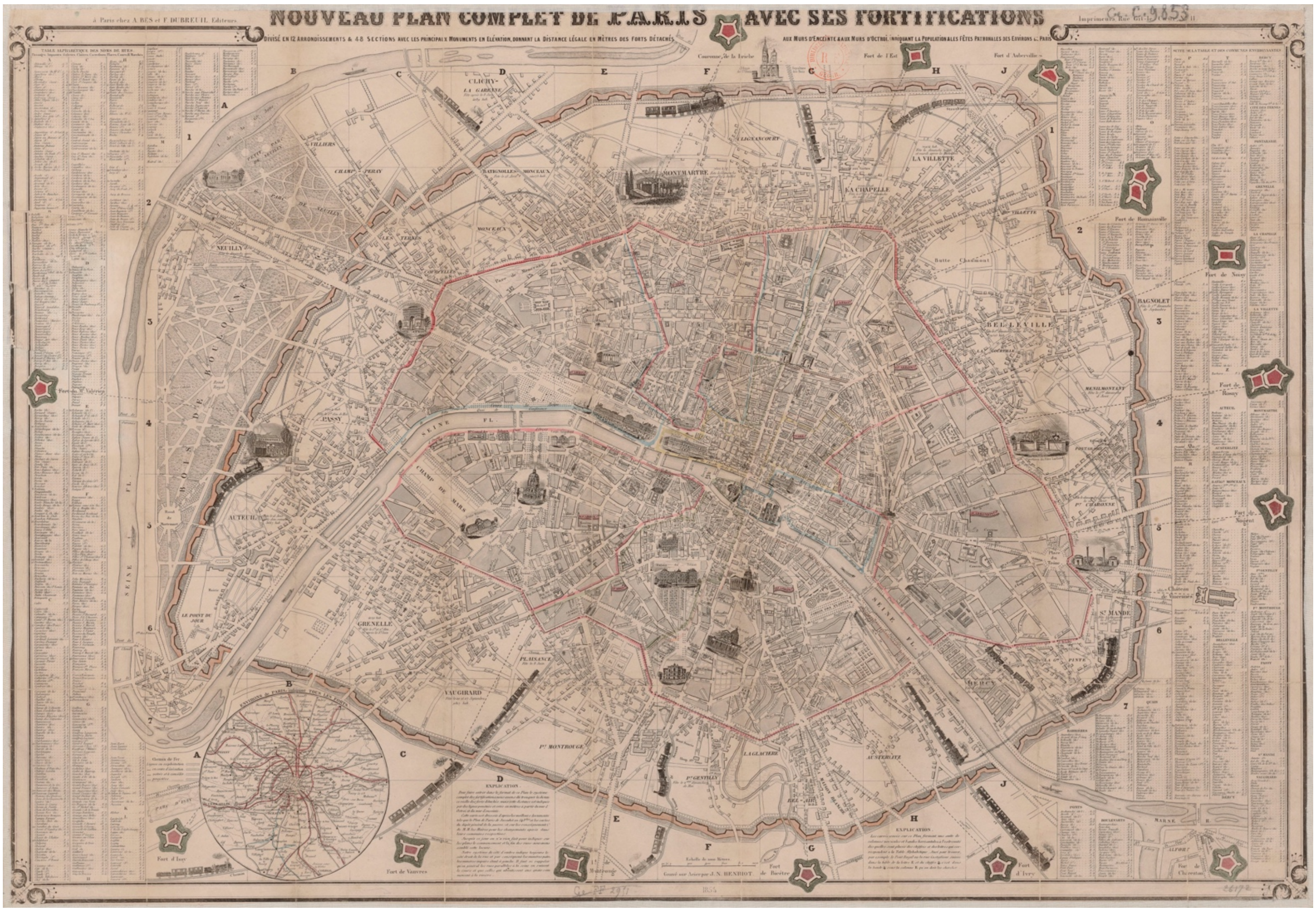

**Figure 7 | Plan of Paris with its fortifications, 1854.** J.–N. Henriot. *Nouveau plan complet de Paris avec ses fortifications*, 1854. A. Bès et F. Dubreuil, Paris. Steel engraving, hand colored. 85 × 58.5 cm. BnF, GE C-9853. URL: gallica.bnf.fr/ark:/12148/btv1b530997787. *The map is distorted to maintain the visual symmetry of the city's enclosure and outer fortifications.*

Along with framing and orientation, the placement of cartographic elements—such as the cartouche, legend, graphic scale, and compass rose—constitutes a key lever for maintaining compositional equilibrium and spatial relationships. An telling, second example is provided by the Plan of Paris and its fortifications (Fig. 7). This oft-reproduced representation of Paris emphasizes the city's symmetrical aesthetic by visually reinforcing the Tiers enclosure (outer fortification) and the *Mur des Fermiers Généraux* (rendered in red), both circumjacent to the city center. A third visual enclosure is formed by the arm of the Seine to the northwest, whose shape is deliberately modified from its actual geographic course to better encircle the city. A fourth concentric ring is formed by the forts depicted outside the city. The position of these forts is adjusted by arranging them close outside the city's enclosure to make them visible and by disposing them in a circumjacent pattern. Finally, the text legend is arranged so as to maintain the circularity and symmetry of the whole, enveloping the map while preserving the continuity of salient geographic features. Other graphic choices—notably the choice of textures and the figuration of monuments—also contribute to the visual balance of the ensemble.

## *On the horizontal orientation of early maps*

The central horizontal axis emerges as the dominant axial relationship in many maps, including the map of Lausanne and, arguably, that of Paris. This preponderance is paradoxical, given that nearly two-thirds of historical maps are horizontally oriented (Fig. 8a). When a map is wider than it is tall, the on-page distance between quadrants 4 and 6 generally exceeds that between quadrants 2 and 8; the corresponding geographical distance is therefore greater, and one would expect a weaker semantic correlation between these quadrants—that is, a weaker horizontal axiality. The paradox arises because the opposite pattern is observed, indicating that the horizontal extension of maps is entirely compensated by composition, framing choices, and, possibly, orientation adjustments.

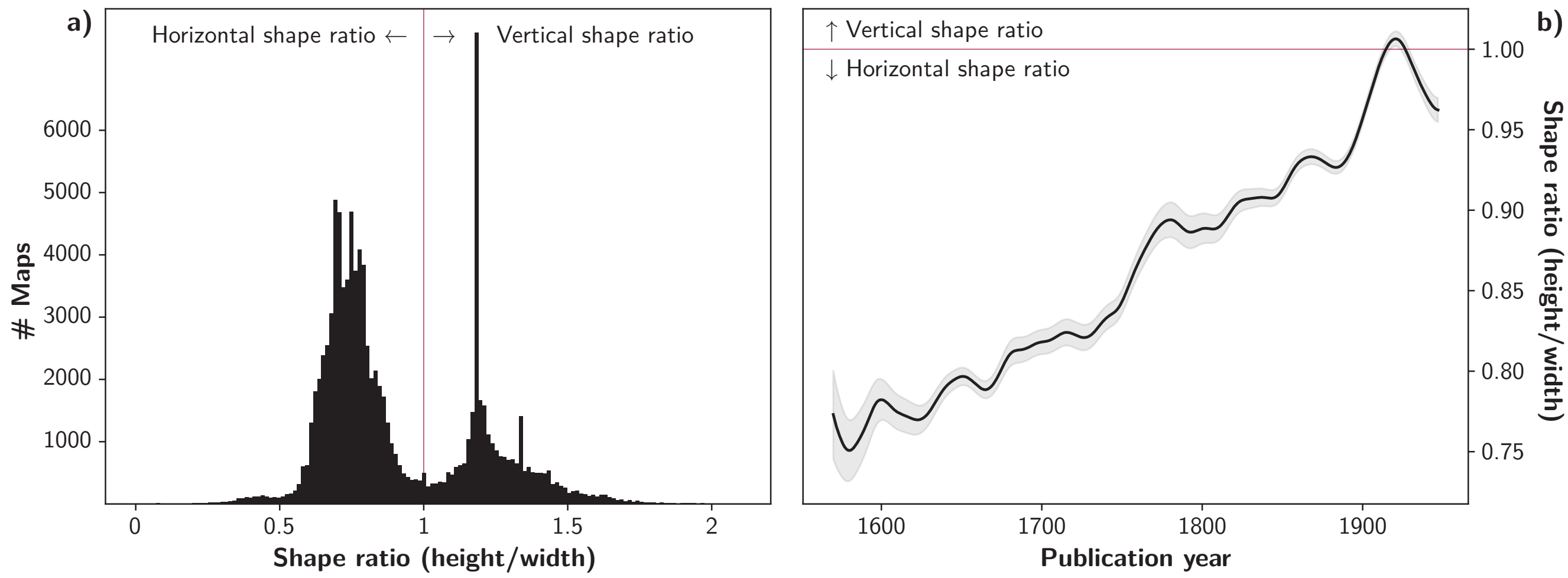

**Figure 8 | Shape ratio (height/width) of the maps.** The image area considered corresponds to the projection profile of the geographic content, after exclusion of the background class. **(a)** Histogram of shape ratios. **(b)** Mean shape ratio by publication date. The values are averaged over a 20–year sliding window and smoothed with a Gaussian ($\sigma = 2.5$). The shaded area indicates the 95% CI of the mean. The red bar marks the limit between horizontal and vertical shape ratios. *Map documents are historically oriented horizontally, a trend that tends to disappear over time.*

Both horizontal shape ratio and axiality are particular to maps and stand in contrast to most Western textual documents, especially books, whose orientation tends to be vertical. Figure 8b suggests that this distinction has its own history. In the 16$^{th}$ and 17$^{th}$ centuries, cartography and landscape iconography were closely related (Gehring & Weibel, 2015). This is evident, for example, in Matthaeus Merian's city maps (Fig. 9). Oblique projections such as these were more common than aerial perspectives; most city maps at that time were simply drawn from a high vantage point. Under this approach, the landscape extended horizontally on the page, resembling a panorama of the environment it depicted. A strikingly delayed consequence can be observed later, in the 18$^{th}$ and 19$^{th}$ centuries. While aerial perspective became overwhelmingly dominant during that period, horizontal orientation *still largely prevailed.* This persistent trend suggests a cultural association between maps and horizontal orientation. From this standpoint, document horizontality may have been perpetuated through the reproduction of conventions instantiated by earlier maps, and the related replication of printing devices, long after the original presumed cause had disappeared.

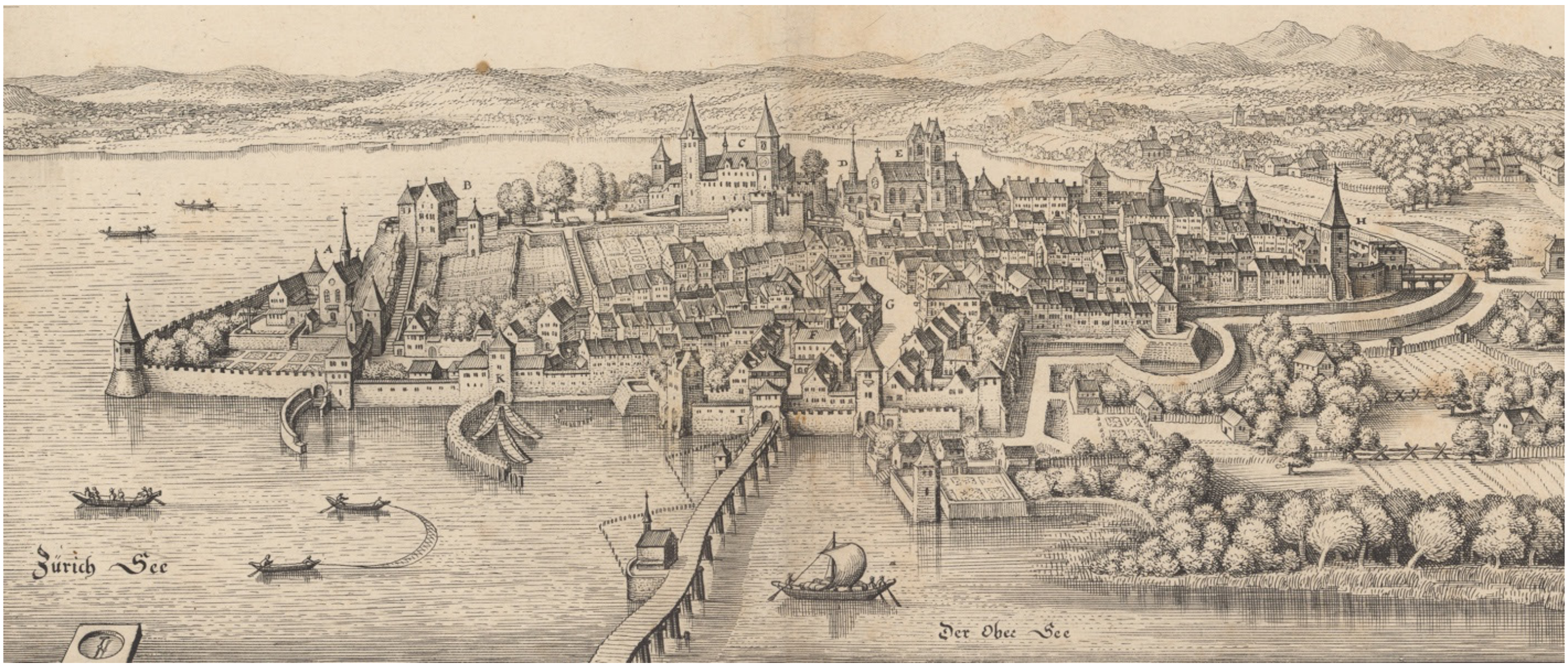

**Figure 9 | Map of Rapperswil, Switzerland, around 1642.** Matthaeus Merian. *Rapperswÿl*, around 1642. Frankfurt am Main. Copperplate engraving. 20 × 32 cm. Zentralbibliothek Zürich, STF XVII 83. doi: 10.3931/e-rara-57516. *Early chorographic maps were closely related to landscape iconography, which explains their frequent horizontal orientation.*

## 5.3 The types of cartographic composition

Whereas the first part of this chapter focused on the overall analysis of compositional relationships, this section investigates how map composition may instead fall into distinct *types*. The derivation of such types, based on the ADHOC Images database, could inform the history of cartography, and more specifically, contribute to the debate on the division of cartography into distinct cartographic "modes", or themes.

The operationalization of cartographic types involves two steps. First, modeling the semantic composition of each map into a single vector representation. Second, the analysis of the resulting space of representation for the manifestation of clusters. The model of semantic composition $\boldsymbol{\Phi}$ is based for one part on overall semantic distribution, and for the other part on the model of semantic relationships. Because a separate feature representation is here computed for each individual map, correlations cannot be estimated directly. Instead, the semantic ratio $\rho_q$ of each quadrant is used:

$$\boldsymbol{\Phi} = \begin{bmatrix} \boldsymbol{\varphi}_{ratios} \\ \boldsymbol{\varphi}_{center} \\ \boldsymbol{\varphi}_{vertical} \\ \boldsymbol{\varphi}_{horizontal} \end{bmatrix} = \begin{bmatrix} \boldsymbol{\rho} \\ log_{10}\left(\dfrac{3 \cdot \vec{\rho}_4}{\vec{\rho}_7 + \vec{\rho}_8 + \vec{\rho}_9}\right) \\ log_{10}\left(\dfrac{\vec{\rho}_1 + \vec{\rho}_4 + \vec{\rho}_7}{\vec{\rho}_3 + \vec{\rho}_6 + \vec{\rho}_9}\right) \\ log_{10}\left(\dfrac{\vec{\rho}_1 + \vec{\rho}_2 + \vec{\rho}_3}{\vec{\rho}_7 + \vec{\rho}_8 + \vec{\rho}_9}\right) \end{bmatrix}$$

The two principal components of $\boldsymbol{\Phi}$ are visualized in Figure A1, in Appendix; color intensity shows distinct dimensions of $\boldsymbol{\varphi}_{center}$. Separable clusters seem to manifest in the dimensionally reduced space. Consequently, two clustering methods were shortlisted for the identification of semantic types: spectral clustering and Gaussian mixture. The silhouette score (Rousseeuw, 1987) was used to evaluate clustering quality, select the most appropriate method and determine the optimal number of clusters. The results of the experiment, considering a subset of n = 50,000 maps, are reported in Fig. A2, in the Appendix.

The two methods tend to disagree on whether the division of the space into two clusters is appropriate. Thus, the second peak was considered, and spectral clustering was used to identify eight clusters. The types were then propagated to the entire dataset through a k-nearest neighbors algorithm (k = 7). The Silhouette score, reaching 0.18, is low, indicating that the data only loosely follow the classification into types. Thus, the composition of many maps lies in-between modes or follows an entirely distinct pattern. Besides, the clustering largely hinges on the composition of the ADHOC Images corpus itself. Nevertheless, considering the large diversity of possible geographic configurations, the clustering remains informative of map topics and their relative importance.

The eight resulting types are depicted in Figure 10. In this figure, each type is illustrated as a composition tiling of nine tiles, corresponding to the nine quadrants. For each quadrant, the average semantic ratio is represented as a tile in which the most prevalent semantic class occupies the center and the least prevalent the periphery. The frequency of each semantic class is calculated as the average frequency within the corresponding type and quadrant. To facilitate the interpretation of each type, Figure 11 presents the relative frequency of each semantic type by year of publication, map scale, publication country, and coverage country.

The first type may be named *#1 Sea-centric maritime*. It is characterized by abundant water, framed by land on some sides. It seems to be the least common type, representing only about 2.4% of the corpus. An example of this type is provided in Figure A3 in the Appendix, a chart of Antarctica (*Terra Australis*), whose landmass remains undefined and therefore appears as ocean. Figure 8b shows that world maps drawn at a scale of 1:20,000,000 or smaller are indeed the most common map scale for this type. The type likewise appears typical of Dutch cartography and of the Dutch East Indies (present-day Indonesia).

The second type, *#2 Chorographic*, is the most frequent in the corpus. Here, chorographic is used in its modern meaning, i.e. pertaining to regional or country maps. Figure 8b shows that this type occurs most often at scales ranging from 1:50,000 to 1:1,000,000. It appears especially prevalent in Swiss and German cartography. The map of Switzerland portrayed in Figure 3, for instance, exemplifies this type.

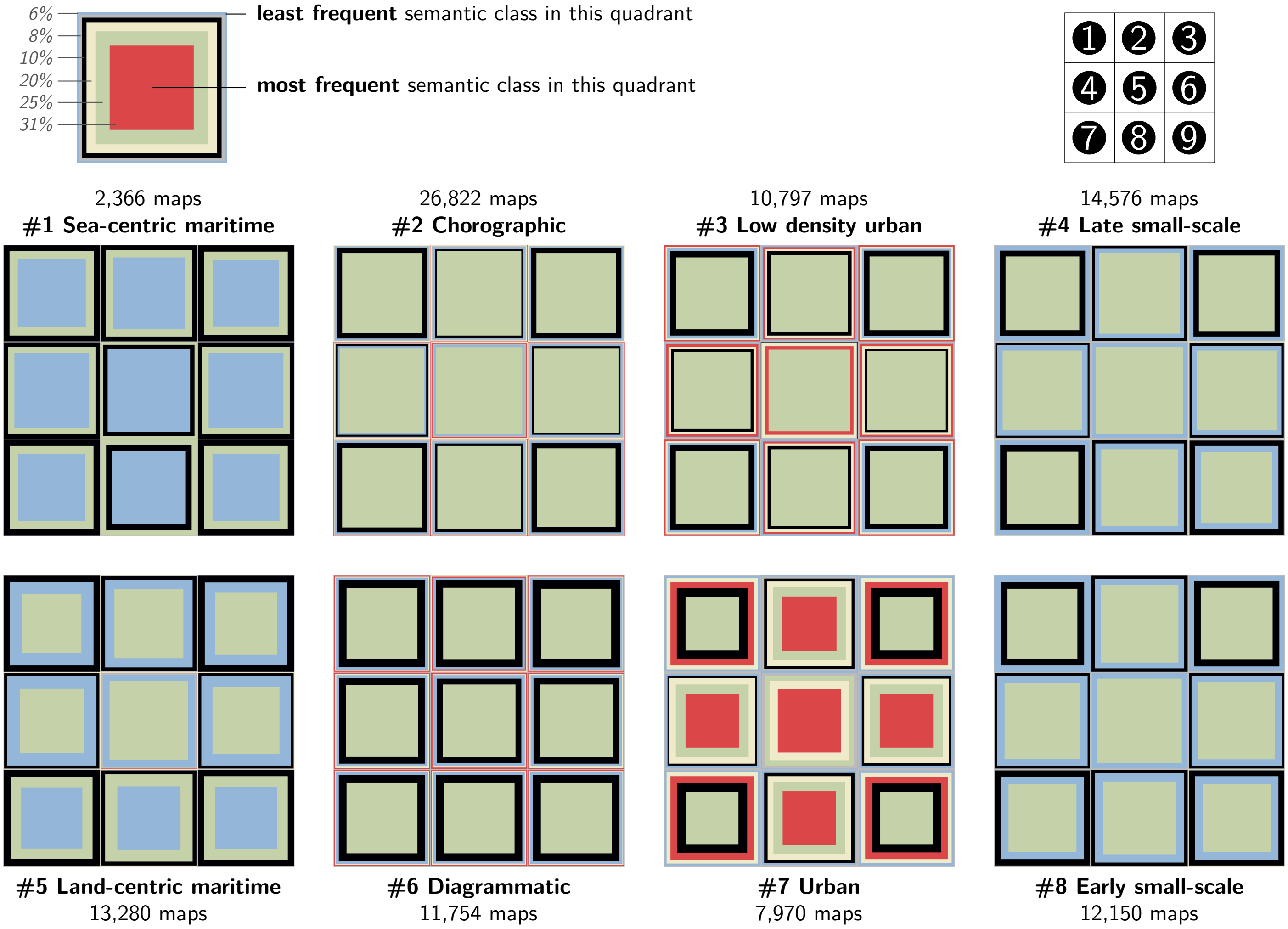

**Figure 10 | Composition tiling of the eight semantic types (1–8)** represented as the relative area of each semantic class within each map quadrant. In each quadrant, the type's relative semantic coverage is visualized as concentric squares; the semantic class most represented appears in the central square, whereas the one least represented appears in the outermost square. The surface area assigned to each semantic class is proportional to the type's average coverage. Every map in the ADHOC Images corpus is classified into one of the eight semantic types, and the number of maps (n) attributed to each type is reported in the figure. *E.g. urban type is characterized by the centering of urban and road network classes and typically includes background elements in the corners.*

The third type comprises *#3 Low-density urban* maps, typically large scale plans (>1:10,000) depicting peri-urban landscapes. It is most common in American cartography and emerged in the 19th century. Sanborn's fire insurance plans exemplify this type (see Fig. A4). The low content density is reflected in the rarity of the contours class (Fig. 10). The urban focus is visible in the slightly higher centrality of the built class, which is the second least frequent in quadrants 1, 3, 7, and 9, yet the second-most common in the central quadrant.

Semantic types #4 and #8 are relatively similar. Both focus on the depiction of non-urban, predominantly continental landscapes. The prevalence of contours and road networks appears slightly higher in type #4, whereas water is more common in type #8. Map scale is generally below 1:20,000 for type #4 and below 1:50,000 for type #8. They may be labelled *#4 Late small-scale* and *#8 Early small-scale* because the principal distinction between these types is chronological. Type #8 dominated from the late 16th through the 18th century, whereas type #4 emerged in the mid-18th century and peaked in the 19th century. Its steadier production in later periods renders #4 the second most common type overall. Representative examples of these types include topographic or regional maps (Fig. A5).

The fifth type constitutes a second variant of maritime cartography, which may be described as *#5 Land-centric maritime*. It generally replicates the composition statistically highlighted in the previous section, in which water was tendentially more prevalent at the bottom of the map. Nevertheless, the semantic distribution indicates that water can occupy any of the peripheral tiles; the central tile occasionally contains a settlement or dense road network. This land-centric type is much more frequent than the maritime type #1, which is sea-centric. As in type #1, most of the maps assigned to this type are medium or small scale (< 1:500,000).

The cluster *#6 Diagrammatic* is a peculiar type because the background class is distributed across all image quadrants. An illustrative exemplar of this type is provided in Figure A6. As in this example, the diagrammatic type may comprise technical drawings forming a series of diagrams or submaps. Such materials often include technical plans produced at a large map scale (> 1:2,000). In addition to cartographic elements, these documents may also incorporate statistical information or terrain cross-sections.

The final semantic type, *#7 Urban*, has already been illustrated by several examples (Figs. 6, 7, and 9). Although it is arguably among the most examined types in the historiography of cartography (M. H. Edney & Sponberg Pedley, 2020, pp. 1529–1624), it nonetheless appears to be the second least frequent in the corpus. Both the scale range and the chronological profile of this type are tightly delimited: it comprises almost exclusively maps at scales larger than 1:20,000. This type seems that have emerged during the second half of the 19th century. American publishers seem to have been the most active contributors to this semantic type.

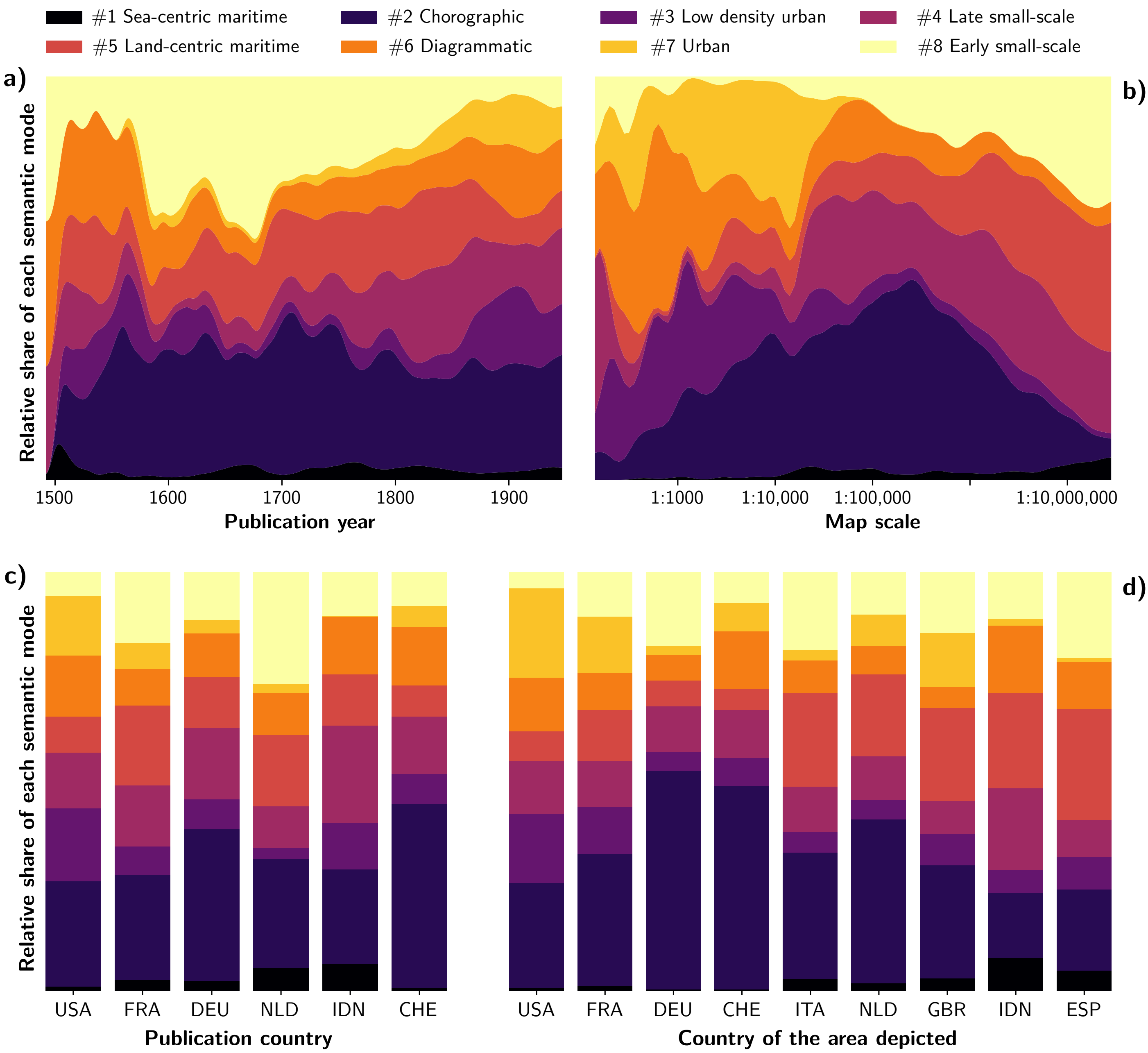

**Figure 11 | Relative share of each semantic type by publication year, map scale, publication country, and country of the area depicted.** For publication year **(a)**, values are averaged over 20-year bins, and a Gaussian filter ($\sigma = 4$) is applied to reduce aleatoric noise. For map scale **(b)**, the scale is logarithmic; values are averaged over a range of $\pm 10^{0.1}$, and a Gaussian filter ($\sigma = 1.5$) is applied to remove aleatoric noise. For publication countries **(c)** and countries of the area depicted **(d)**, the legend corresponds to ISO-3 country codes (USA = United States, FRA = France, DEU = Germany, NLD = Netherlands, IDN = Indonesia, CHE = Switzerland, ITA = Italy, GBR = United Kingdom, ESP = Spain). *E.g. type #8 dominates in the 17th century and at maps made at smaller scales; it is comparatively frequent in Dutch maps.*

The semantic types of cartography exhibit complex temporal dynamics. Certain types, such as the #2 Chorographic type, maintain a persistent presence, whereas others, such as #8 Early small-scale and #4 Late small-scale, appear primarily in specific periods. This case may be construed as the substitution of one semantic type for another. In other cases, however, change does not result in outright replacement but is manifested as shifts in the relative prevalence of specific types. These shifts may signal concurrent cartographic trends, which may affect both cartographic content and form.

Some results from the categorization of maps into semantic types are surprising—for instance, the low prevalence of #1 sea-centric maritime charts and #7 urban maps. Although maritime and urban semantic types are among the most studied in map scholarship, they apparently account for only a small proportion of published maps.

The analysis also highlights that map scale is a key differentiating factor: certain map types occur mainly at smaller scales (e.g., #4, #8) and others at larger scales (e.g., #7). Place of publication may also be quite informative. For instance, #1 sea-centric maritime charts are almost always produced by Dutch mapmakers; #2 chorographic mapping is far more prevalent in German and Swiss publications, whereas urban-focused cartography (#3, #7) is predominantly American. These patterns reflect specific cultural foci, resulting in different objectives, functions, and, consequently, semantic types.

## 5.4 A look at the road

The first part of this chapter focused on the analysis of maps as composed pictures. However, it would be equally reductive to negate the geographic dimension of maps. This last section will attempt to qualify the respective balance between representation and geographic, referential function.

As highlighted in the study of spatial semantic relationships, the influence of distance on spatial semantic correlations is strong ($r = 82.7$). This finding recognizes that a substantial portion of representation serves to convey geographic information. Roads constitute a class of cartographic features that can offer insight into the ambivalent interplay between map representation and the physical landscape. Figure 12 presents the relationship between map scale and average road width. If the map scale did exactly reflect the scale of the environment, road width would diminish as the map scale decreased, until it became practically invisible—as observed, for instance, on satellite imagery. Mathematically, the relationship between depicted road width $W$ and the denominator of the map scale $S$ should be linear on a log-log plot—assuming that the physical variation of road width is negligible.

As shown in Figure 12, the observed relationship is less straightforward. If we compute the regression:

$$\psi = log_{10}(S^{-1}) = \beta_0 + \beta_1 \cdot log_2(W) = \beta_0 + \beta_1 \cdot \omega \tag{1}$$

whose mean absolute error (MAE) is defined regularly as:

$$MAE = \frac{1}{n}\sum_{\forall i}|\psi_i - \widehat{\psi_i}|$$

we observe that the mean absolute error (MAE) equals 0.817 [0.805–0.828][3], a value significantly lower than the chance level of 1.351 [1.343–1.357][4]. If the dependent and independent variables in Equation 1 are reversed, that is, if we attempt to predict road width from map scale, the MAE decreases to 0.671 [0.659–0.683], whereas chance lies at 1.080 [1.075–1.085]. Put differently, it means that around 25% of the depicted road width can be explained by map scale[5]; the remaining 75% appears to stem from graphic choices or, arguably, from the physical variation of road widths on the ground. Conversely, if road width were used as a proxy to infer map scale, the latter would explain about 71% of the variation[6], while the residual 29% would remain unexplained by this simple mathematical model[7].

The role of physical variation in ground road width is assessed through a short experiment. If that impact were substantial, we would expect road width to increase over the study period. Indeed, from the middle of the 19th century, road-widening projects were undertaken in many cities and on interurban roads. In cities, this allowed more light into dwellings, made traffic more fluid, facilitated the installation of tramways, and later, the spread of automobiles (Akerman, 2006). A short statistical analysis confirms this expected increase. For $H_1 \colon \boldsymbol{\omega}_{\text{before 1850}} < \boldsymbol{\omega}_{\text{after 1850}}$, we obtain $p_{val} = 1 \cdot 10^{-45} \ll .05$. The mean width thus seems to increase by about 24% over the period, a moderate yet tangible change that supports the hypothesis that physical road width is, to some degree, and on average, mirrored in the width depicted on maps.

---

[3] The MAE and its 95% CI is computed by applying 100 times the Eq. (1) on the data (n=43,799), each time with a different random partition into train and test set (80/20 %).
[4] Calculated by applying random permutations.
[5] $2^{0.671}/2^{1.080} = 0.25$
[6] $10^{0.817}/10^{1.351} = 0.71$
[7] Note that we also evaluate the bias introduced by the use of pixel values as proxy of the actual road width on the map document. Focusing on the subset of n=18,971 map records for which both the map scale and physical dimensions of the map document are known, based on catalog metadata, we introduce a resolution factor $f = \bar{d}\ [cm] / \bar{d}\ [px]$ where $\bar{d}$ is the average of both horizontal and vertical dimensions. The variable $\omega$ in Eq. (1) then becomes $\omega = log_2(f \cdot W)$. The results of that alternate experiment are very similar to the ones reported. We get MAE=0.85, with chance unchanged, when setting map scale as regression target, indicating that 68% of the variation is explained by the model. Conversely, when trying to predict road width, MAE=0.99, with chance at 1.57, suggesting that 33% of the variation is explained.

This short case study illustrates the inherent ambiguity of cartography. On the one hand, graphical choices affect maps, making roads, for instance, visible on small-scale representations. On the other hand, they do not eclipse the influence of geographic features—i.e., the actual road width on the ground—on the information depicted. Similarly, although the very notion of map scale rests on an idealized conception of cartographic rationality (M. H. Edney, 2019), scale remains a relevant framework for analyzing and interpreting historical maps. This ontological dissonance enables maps to function simultaneously as repositories of empirical knowledge and as cultural artifacts, products of deliberate graphic choices and composition.

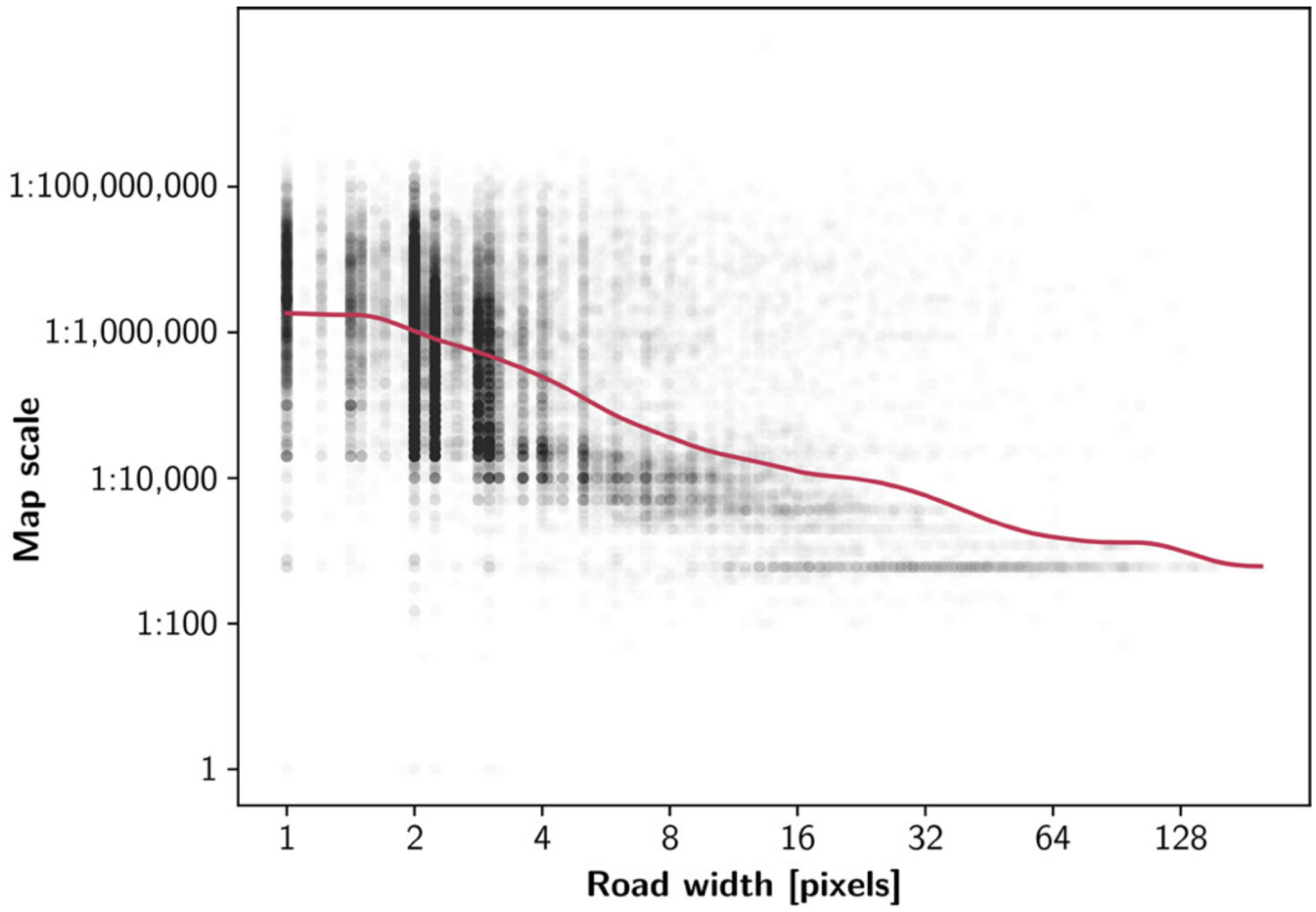

**Figure 12 | Road width as a function of map scale.** Both axes are shown on logarithmic scales. The red line represents the mean, averaged over a range of $\pm 10^{0.5}$, and smoothed with a Gaussian filter ($\sigma = 25$). *The relationship between road width and map scale is not straightforward.*

## 5.5 Conclusion

In this chapter, a quadrant-based framework was introduced to analyze maps of whole images. This approach made it possible to identify spatial patterns, such as the central cross, circumjacent consistency, and horizontal axiality, that characterize the composition of maps as images. Image composition may also be examined through specific types, like chorographic, diagrammatic, urban, or sea-centric maritime maps, which tendentially structure maps into broad semantic modes. The final section of the chapter took a step back by discussing the ambivalence of cartographic representation, between figurative choices and geographic function.

By exploring the grammar of cartographic images, the study showed that classes of geographic objects may be emphasized through mapping processes such as framing. Concurrently, the experiment on the horizontality of maps documents indicated that cultural conventions, such as the adoption of a landscape orientation, can be instantiated and perpetuated through replication. The analysis of semantic types suggested that cultural categories diffuse unevenly across time and space, affected by cultural and historical contexts. From the perspective that mapping is a resource-intensive process, these variations can be interpreted as a form of competition between distinct cartographic modes. Indeed, the relative frequencies in Figure 11 also reflect market shares. In this regard, not only do map types compete for public visibility but so do mapped features. Cases in which certain geographic features are emphasized at the center of the image, whereas others are considered dispensable—whether occluded by layout elements or excluded from the map frame altogether—offer another example of selection process, influenced by intentional choices and cultural norms.

The dimensions on which this chapter focused, such as composition and framing, pertain to cartographic form. Beyond these, however, *form* is also constituted by signs, like icons and symbols, as well as motifs, textures, lines, and colors. The next two chapters will concentrate on cartographic signs. In particular, they will examinate the hypothesis that the cultural diffusion of signs can be understood and modeled as a selective process.

# Appendix A – Supplementary materials

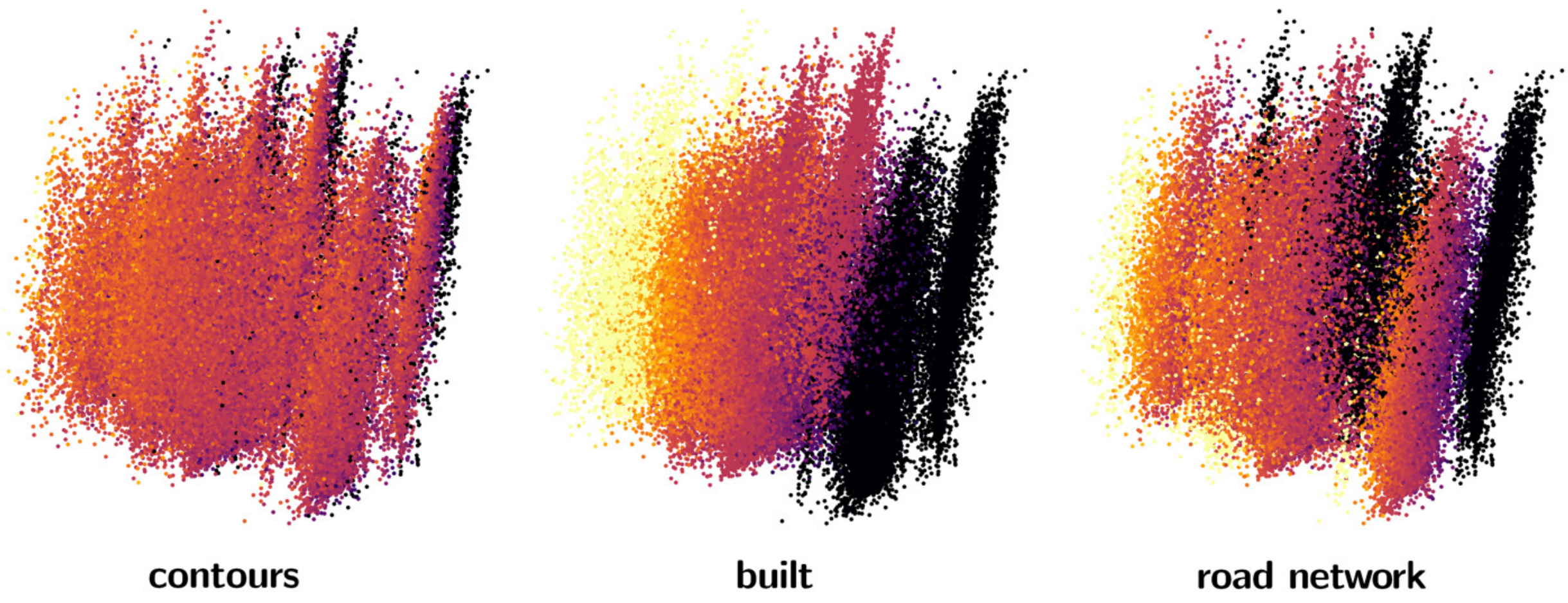

**Figure A1 | Projection of the two principal components of the composition features $\boldsymbol{\Phi}$.** Highlighting $\boldsymbol{\varphi}_{center}$ for contours, built, and road network components. *Distinct spectral clusters appear, encoding different semantic classes.*

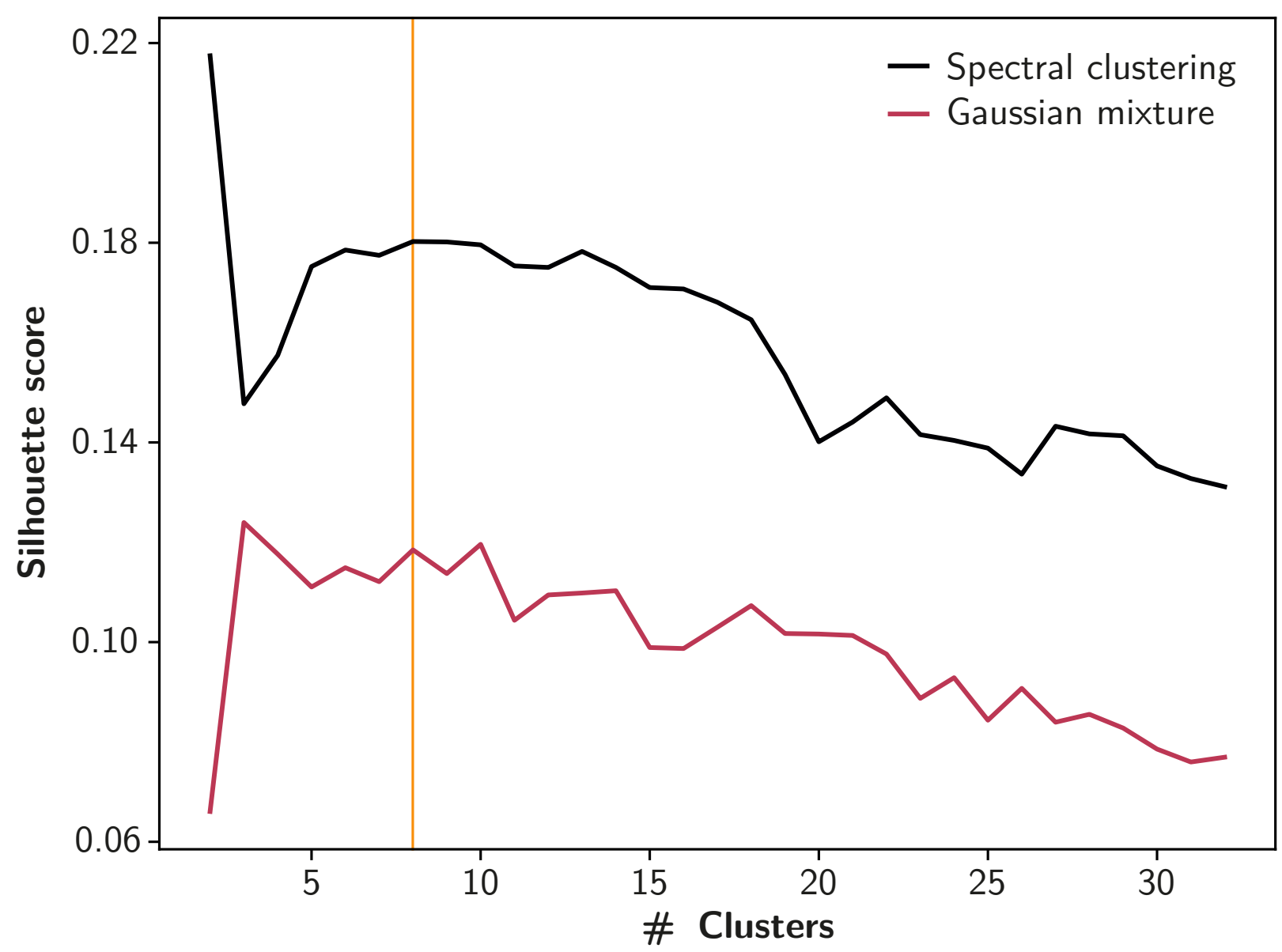

**Figure A2 | Variation of the silhouette score according to the clustering algorithm and number of clusters.** Each experiment is repeated 5 times with a different training set ($n_{train}$=10,000), sampled randomly. The value plotted is the average silhouette score computed on the separate test set ($n_{test}$=10,000). The yellow horizontal bar marks the selected number of clusters (k=8).

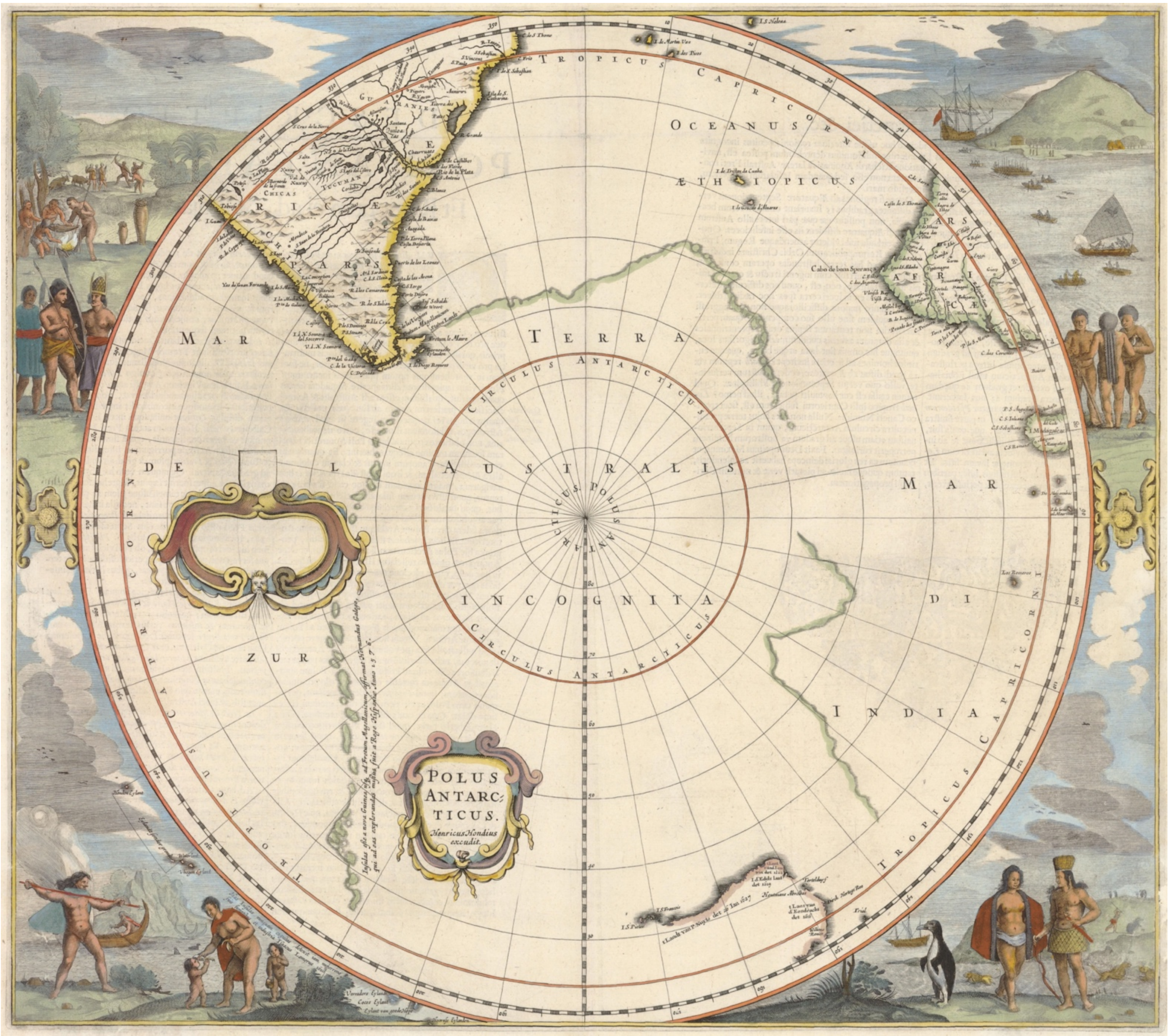

**Figure A3** | **Map of Antarctica, 1638.** Hondius Hendrik, and Jan Jansson. *Polus Antarcticus*. 1638. Amsterdam. Hand colored. 44 x 50 cm. Princeton library, HMC01.3748 D Alc. 14 Draw. 15. URL: catalog.princeton.edu/catalog/9952841883506421. The map is an exemplar of the semantic mode #1 (sea-centric maritime).

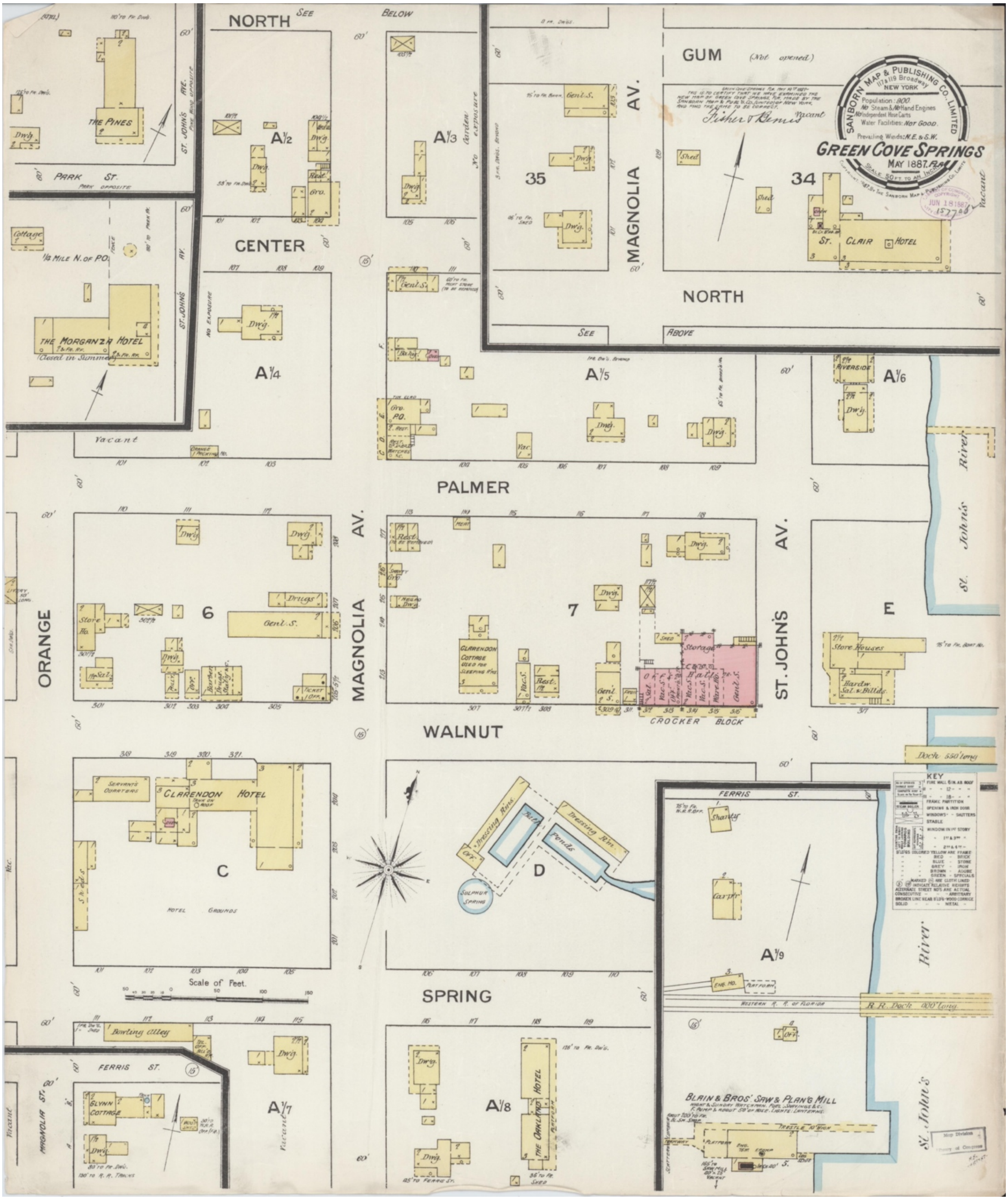

**Figure A4 | Fire insurance plan of Clay County, Florida, 1887.** *Green Cove Springs*. 1887. Published by Sanborn Map Company, New York. Hand colored. 44 x 50 cm. Library of Congress. URL: www.loc.gov/item/sanborn01275_002/. The map is an exemplar of the semantic mode #3 (low density urban).

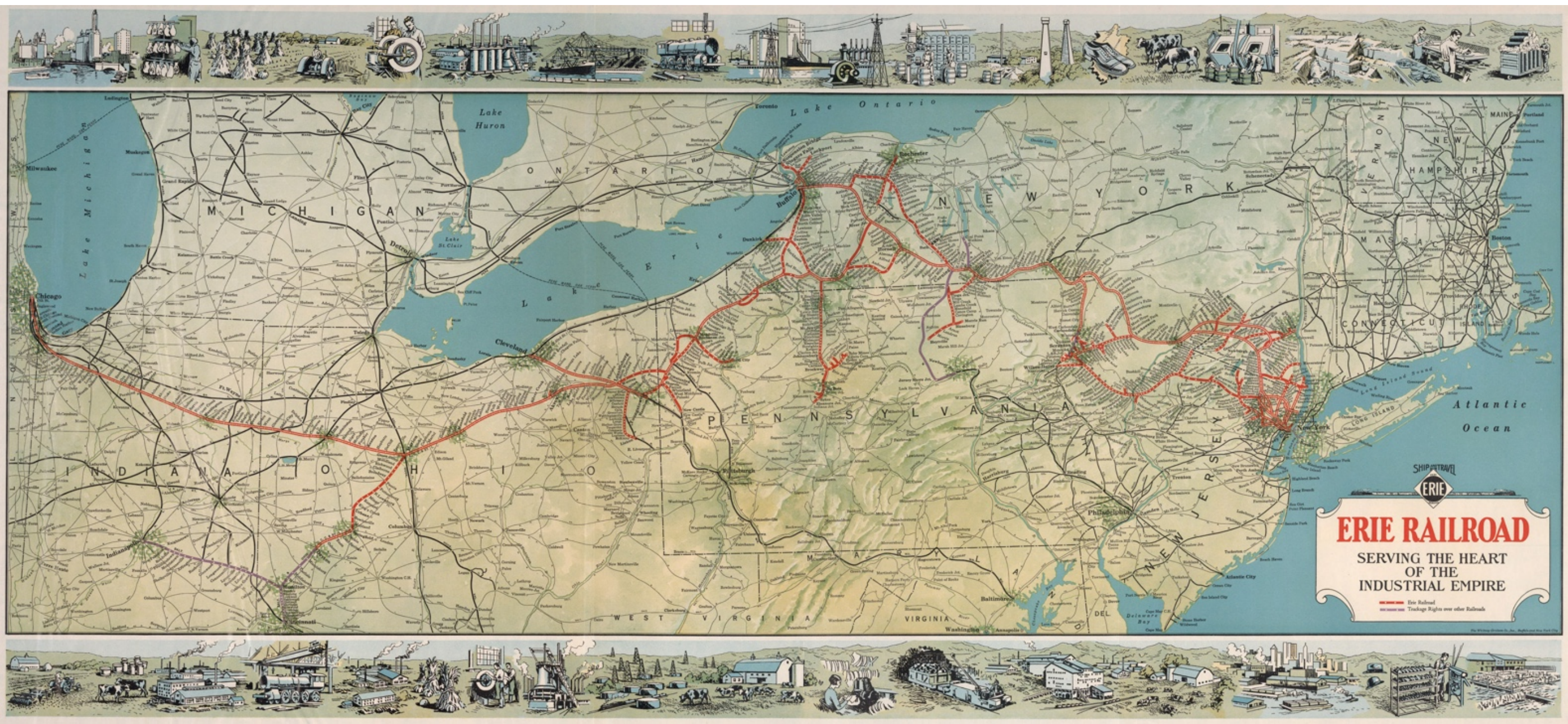

**Figure A5 | Erie Railroad Map, around 1930.** *Erie Railroad: Serving the heart of the industrial empire*. between 1927 and 1930. Published by Erie Railroad Company, and Whitney-Graham Company, New York. 46 x 101 cm. David Rumsey Map Collection, 11489.022. URL: www.davidrumsey.com/luna/servlet/detail/RUMSEY~8~1~297222~90068813. The map is an exemplar of the semantic mode #4 (late small-scale).

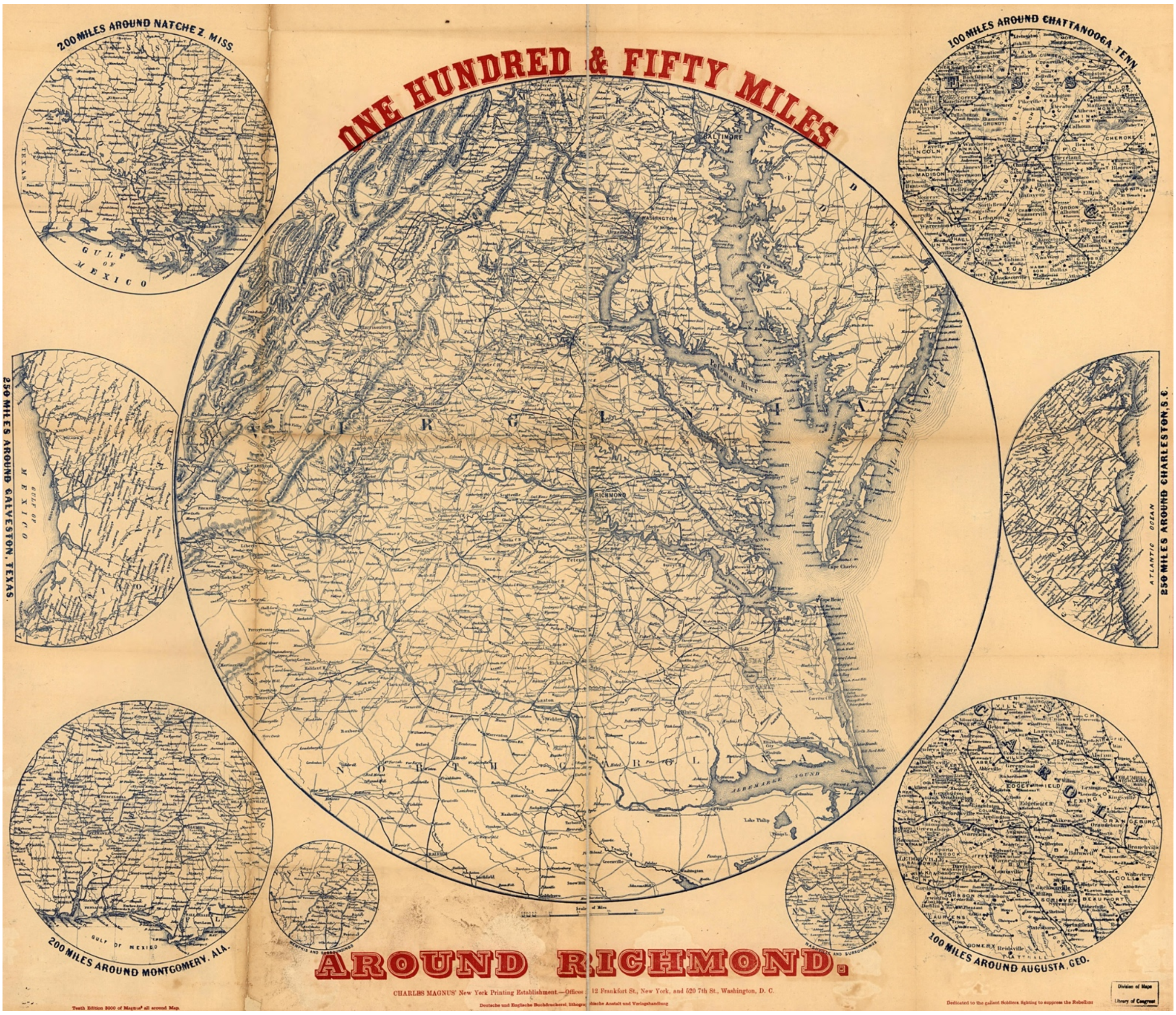

**Figure A6 | Map of Richmond region, around 1863.** Charles Magnus. *One hundred & fifty miles around Richmond*. Between 1861 and 1865. Published by Charles Magnus, New York. 66 x 76 cm. Library of Congress, G3884.R5S5 186-M32. URL: hdl.loc.gov/loc.gmd/g3884r.cw06324001. The map is an exemplar of the semantic mode #5 (land-centric maritime).

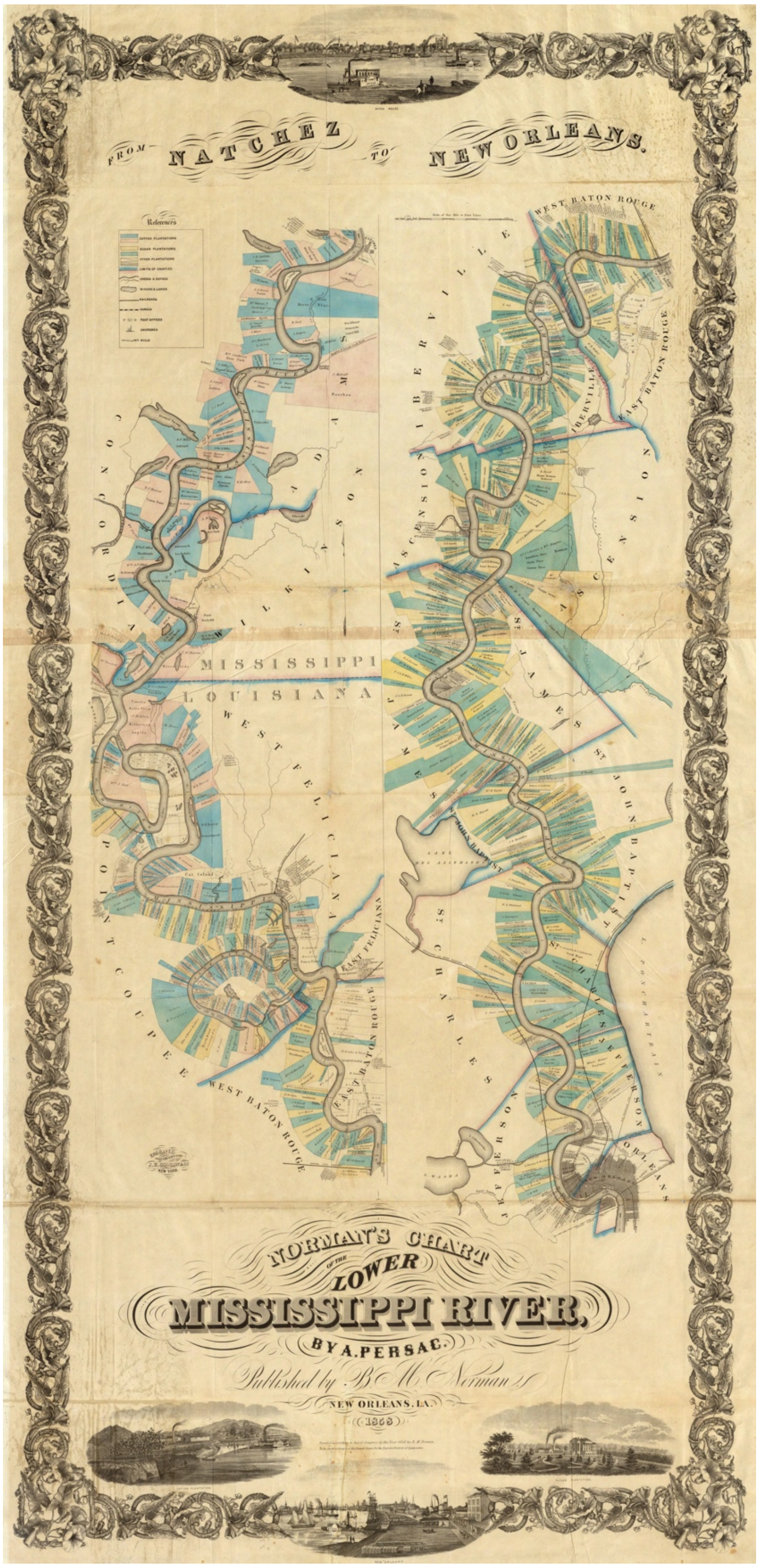

**Figure A7 | Map of the course of the Mississippi River, 1858.** Adrien Persac, Joseph H. Colton, and Benjamin M. Norman. *Norman's chart of the Lower Mississippi River, from Natchez to New Orleans*. 1858. Published by Benjamin M. Norman, New Orleans. Engraved and printed by J. H. Colton & Co, New York. 155 x 73 cm. David Rumsey Map Collection, 2752.000. URL: www.davidrumsey.com/luna/servlet/detail/RUMSEY~8~1~2131~180009. The map is an exemplar of the semantic mode #6 (diagrammatic).

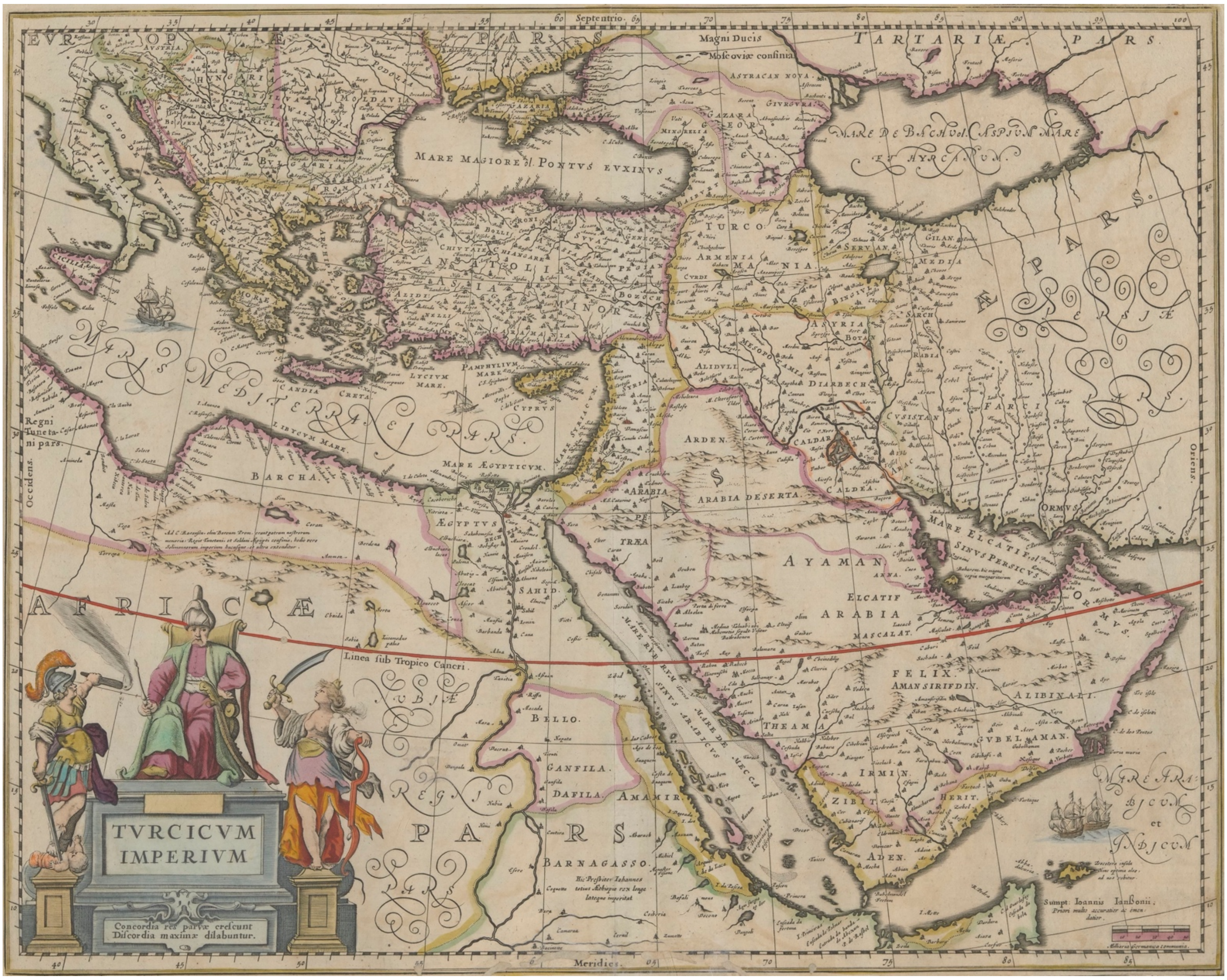

**Figure A8 | Map of the Turkish Empire, around 1630.** Jan Jansson. *Turcicum Imperium*. Around 1630. Amsterdam. Copperplate engraving, Hand colored. 50 x 39 cm. Universitäts- und Landesbibliothek Sachsen-Anhalt, BII 7[6]6. doi: 10.25673/39516. The map is an exemplar of the semantic mode #8 (early small-scale).

# Chapter 6

## *The Map Code: Investigating the Evolution of Cartographic Signs*

In this Chapter, the dissertation moves to a more granular level of observation: that of cartographic signs. Signs, like icons and symbols, pervade historical cartography. In this chapter, the term “sign” designates what semiotic theory specifically names *point signs* i.e., discrete symbols and icons, in contrast to dimensional signs, like lines or areas, which will be discussed in the next chapter. The present Chapter thus focuses on discrete figurative elements, like the icon of a tree, or the use of a circle to represent a town.

Studying signs at scale involves three preparatory steps. First, recognizing them and extracting them from whole map images. Second, encoding them in a descriptive and culturally meaningful representation space. Third, defining a methodological framework to operate them—e.g., identify representative samples through assimilation or, on the contrary, compare them and highlight dissimilarities. The methodological challenges lie with the relative exploratory character of map signs recognition, the large number of extracted signs (63 million), and the important diversity of those signs, which form more of a continuum than a discrete visual lexicon. Thus, the development of a meaningful space of representation and its validation against interpretable cultural dimensions will occupy the first half of this chapter.

The second half will concentrate on studying the historical evolution of map signs. It also constitutes a new development in the interpretative framework, which until now primarily interpreted the history of cartography through the lens of control, power, and intentionality. With respect to map signs, however, this perspective is supplemented with another mechanism, that can be described as evolutionary. From that standpoint, map semiotics is regarded as an adapting system, composed of sign variants, whose equilibrium is continuously shifted by historical and cultural contexts.

## 6.1 Automated detection of map signs

### *Related work*

Despite the high information density represented by cartographic signs, few works have attempted to extract icons and symbols from maps. By contrast, a growing number of publications focus on the reputedly more complex task of text detection (Chiang, 2015; Kim et al., 2023; Z. Li et al., 2024; Weinman et al., 2019; Zou et al., 2025a). In parallel, Uhl et al. (2018, 2020) developed pioneering patch-based classification approaches for extracting geographical features, such as urbanized spaces. These methods were subsequently developed and adapted to the extraction of other features, like wetlands (Ståhl & Weimann, 2022), coastlines, and archaeological features (Kersapati & Grau-Bové, 2023). Hosseini et al. (2022) further generalized and disseminated this technique in the MapReader tool.

However, to the best of my knowledge, the first article describing the application of object detection to map icons was published by Smith et al. (2025), who extracted urban trees from the Ordnance Survey map of Leeds. Smith et al. selected Mask-RCNN (He et al., 2018), a region-based convolutional neural network, over YOLO (Redmon et al., 2016), primarily on the basis of qualitative assessments and documented shortcomings of early versions of YOLO, such as YOLOv3 (Redmon & Farhadi, 2018). A distinct project focused on extracting icons from the Cassini map of France (Sutty & Duménieu, 2024). In that work, Sutty and Duménieu adopted YOLOv10-X (A. Wang et al., 2024) after comparing its performance against DETR (Carion et al., 2020) and Detectron2 (Y. Wu et al., 2019).

In the present study, YOLOv10-X (A. Wang et al., 2024) will be used to detect icons and symbols. Indeed, recent YOLO models have reportedly outperformed Mask-RCNN across varied benchmarks (Sapkota et al., 2024; C.-Y. Wang et al., 2023). YOLO—*You Only Look Once*—refers to a family of object detection models offering computationally efficient implementations, while maintaining competitive accuracy (Redmon et al., 2016). YOLOv10 is among the most recent iterations of this family. Compared to its predecessors, it offers greater efficiency with comparable accuracy. It relies on a strategy known as dual assignment consistency to resolve multiple superimposed detections, thereby obviating the need for non-maximum suppression (NMS) in post-processing, which constituted hitherto a major computational bottleneck.

### *Data*

For the purpose of this research, the Cartographic Sign Detection Dataset (CaSiDD), comprising 796 map patches, whose dimensions ranged from $768 \times 768$ to $1000 \times 1000$ pixels, was labeled manually and used for training and validation. The dataset is published along with this dissertation (Petitpierre & Jiang, 2025). A representative example is provided in Figure 1. In total, approximately 18,750 icons and symbols were annotated. Annotation was performed in a class-agnostic fashion; that is, the process was not constrained by an ontology defined a priori but the latter was instead adapted throughout the work. Although certain classes of icons—for instance those signifying trees or cities—appear very frequently across different maps, other classes are much rarer, such as *embassy*, *battlefield*, *lighthouse*, or *post office*. A few hapaxes are also observed, the meaning of which may be, by definition, difficult to infer without additional context. The statistics of annotation data indicate that icon frequencies overall follow a power–law distribution[1]. Thus, between hapaxes and utterly frequent categories, it is difficult to establish the optimal level of ontologic granularity. In the end, an ontology of 24 classes was adopted.

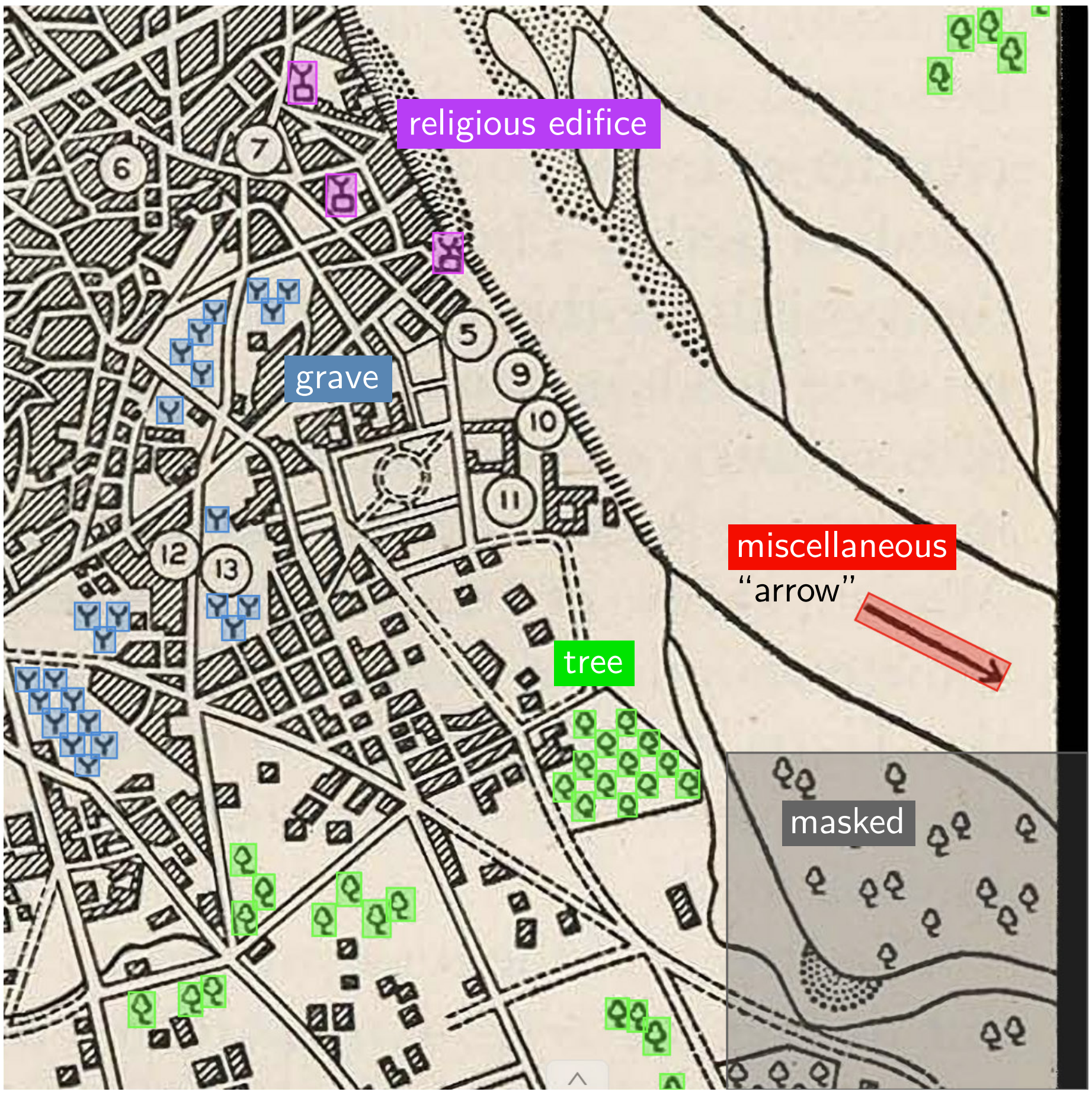

**Figure 1 | Example of a labeled image.** Each bounding box corresponds to a single icon instance. Colors indicate distinct classes (e.g., grave, religious edifice, tree). The class *miscellaneous* is reserved for hapax or relatively rare icon classes, specified with a text attribute (e.g. "arrow"). The class *masked* designates image areas that were not labeled due to high symbolic redundancy with other labeled areas.

[1] The likelihood ratio of *power law* distribution against *exponential* is 137 ($p_{val} = 3 \cdot 10^{-7}$).

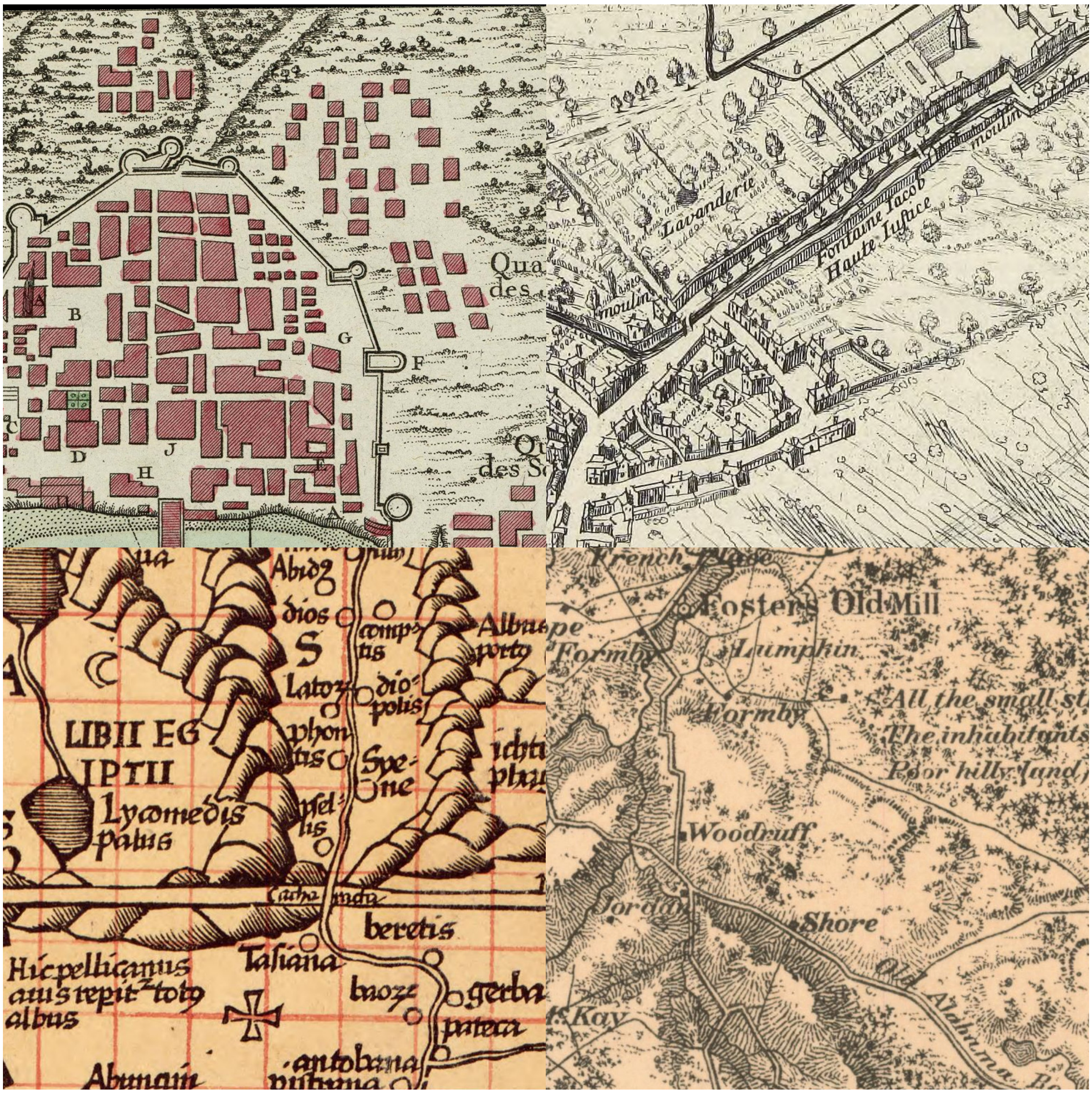

**Figure 2 | Variations on iconicity and icon's discreteness.** *It can be difficult to identify and distinguish entangled signs from one another.*

A difficulty of the annotation task itself is that the form of icons, or symbols, is not as well defined as could be expected. Although automatic detection methods prescribe strict bounding boxes, signs do not always conform to them forthrightly. While Figure 1 fits the paradigm, Figure 2 presents more challenging examples, suggesting that the discreteness of signs is a gradient, rather than a binary property. As illustrated in Figure 2a, the ontological distinction among vegetation types (trees, bushes, grasses) and the evaluation of background texture remains inherently subjective. The same phenomenon appears in Figures 2b, or 2d, where some trees—symbolized in the latter by a star icon—are clearly delineated, while others coalesce into an indistinct tangle. In Figure 2b, buildings likewise tend to overlap. Orography—the representation of relief—also varies considerably. Whereas contemporary cartography usually treats elevation as a continuous variable, maps prior to the 18th century tended to discretize relief into separate mounds, as shown in Figure 2c. These examples expose the limitations of point sign annotation. Examples of labeled data for each class are shown in Figures B1–B7.

### Training

The YOLOv10-X model, pretrained on the COCO[2] dataset, was fine-tuned to convergence using the AdamW optimizer and a degressive learning-rate schedule initialized at $4 \cdot 10^{-4}$. The training data were augmented through random cropping to 640 × 640 pixels, horizontal flipping, and mild jittering of color value (±8 %) and saturation (±14 %). Mosaic augmentation, which has been standard for YOLO models since its introduction in YOLOv4, was not considered appropriate for the present task and was thus disabled[3]. The entire training phase required just 35 minutes on an RTX3090 GPU. In this first step, classification considers a single class.

### Results & Assessment of detection

Performance was validated quantitatively by computing precision and recall (see Fig. 3). Detection was considered correct when the intersection-over-union (IoU)[4] exceeded a predefined threshold. Figure 3 shows that performance degrades rapidly as this threshold increases. At the baseline level ($IoU \geq 0.3$), precision reached 81.1 % and recall 41.5 %. When the requirement was raised to $IoU \geq 0.5$, average precision (AP50) degrades to 75.7 % and recall to 38.9 %. This early sensitivity to the IoU threshold suggests that the predicted bounding boxes are not particularly tight. To provide a comparison to these quantitative results, the AP50 of YOLOv10-X on COCO (Lin et al., 2014), a multi-class reference object detection benchmark, is 71.3%, which is slightly lower than the value observed on ADHOC Images (A. Wang et al., 2024). Yet, COCO comprises 330'000 image samples for training and validation.

[2] COCO is a classic computer vision dataset, comprising some 200,000 photographs annotated according to 80 semantic classes, including distinct animals, everyday objects, furniture, vehicles and people. It contains 1.5 million annotated objects in total, i.e., nearly 80 times more than the present dataset.

[3] Mosaic augmentation assembles composite images from a number of unrelated examples, in a manner similar to patchwork. A potential consequence the reduction of the sign's relative size relatively to the image, or the loss of visual and spatial-geographical contextual information; neither of these effects being desirable in the present case.

[4] Calculated as the proportion of true positives (TP) divided by the total of true positives, false negatives (FN) and false positives (FP): $TP/(TP + FN + FP)$

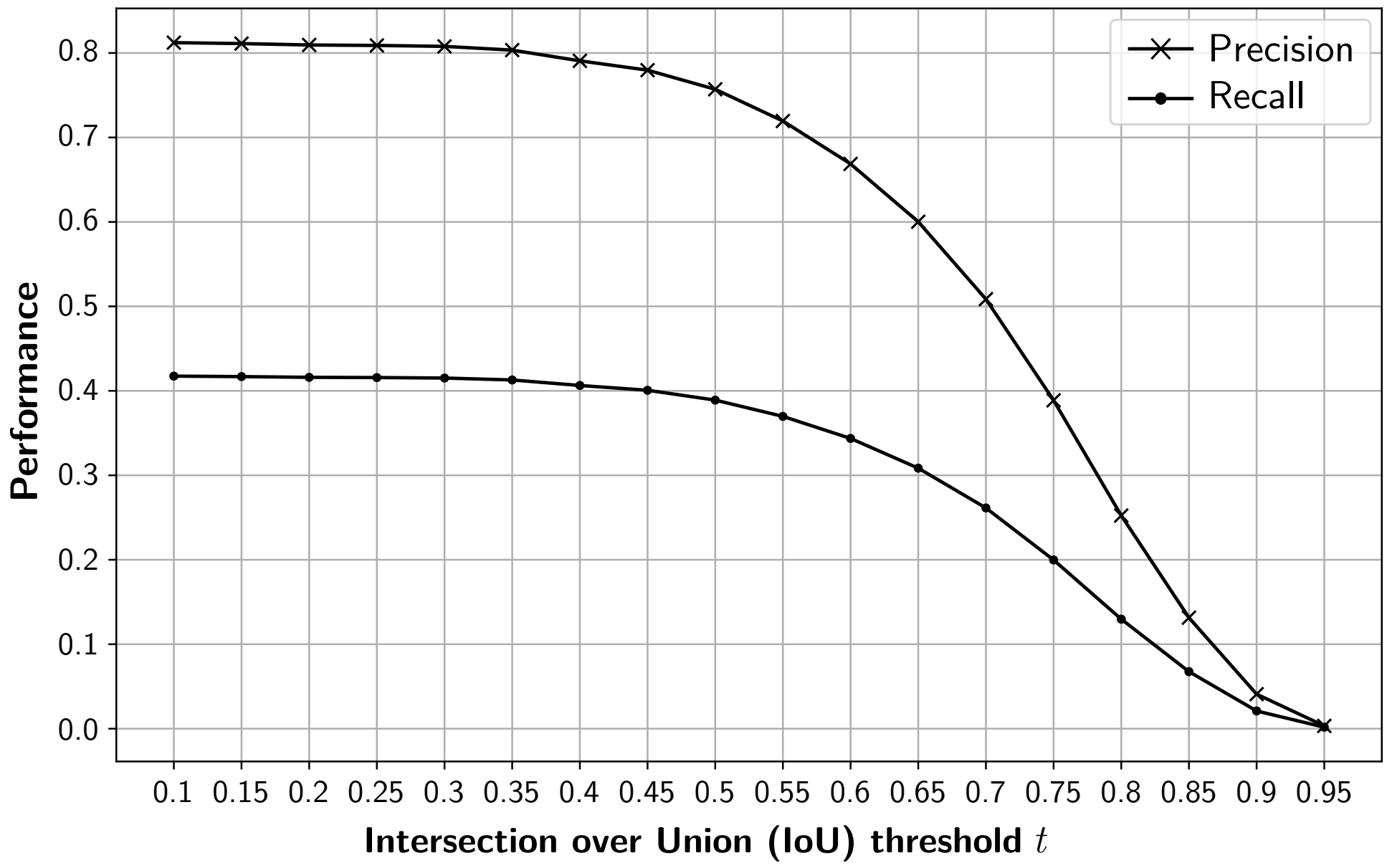

**Figure 3 | Precision and recall of sign detection** as a function of the IoU threshold applied, on the validation set. *Precision scores substantially higher than recall. Performance decreases for higher IoU thresholds, indicating loose bounding boxes.*

The multiple qualitative examples which will be presented in Sections 6.7, 6.9 and 6.10 further suggest that actual false positive predictions are less frequent than the precision metric indicates, a discrepancy that may be attributed to interpretive divergences arising from the difficulty of discretizing and identifying signs, as discussed earlier. Here, the limited recall is not critical; it is indeed not necessary to extract every single sign to study map semiotics. Accordingly, high precision is preferable to high recall. Finally, the bounding-box tightness issue—or the graphic noise it might incur—will be addressed in a subsequent post-processing step through a foreground segmentation strategy.

Inference on the 99,715 maps in the ADHOC Images corpus required 15 days and 9 hours on an RTX 3090 GPU, or approximately 13 seconds per map. Overall, 63,180,211 signs were detected. One in four maps comprised more than 953 signs, one in four had less than 32 signs detected. After adjusting for precision, one can estimate with a confidence level above 99 %[5] that 94.2 % of the maps in the corpus contain at least one cartographic sign.

[5] This is a conservative estimate. Since the sign count does not appear to be distributed according to a normal law, we must approximate the probability based on precision. For 94.2% of the maps, at least 3 icons are detected. The probability for all these 3 icons to be false positives is $(1 - precision)^3 = 0.189^3 = 0.7\% < 1\%$. The actual probability would be even lower if recall could be accounted for.

## 6.2 Representation spaces: technical and theoretical context

Whereas the first section focused on the retrieval of 63 million icons and symbols from the map corpus, the present section addresses the ways of transforming these signs into operable units. Up to this stage, the extracted icons and symbols are merely arrays of image pixels. The *embedding* process aims to represent—i.e., encode—the form and semantics of the retrieved map signs, making them assimilable and comparable within a mathematical space.

**Classical feature extraction approaches.** Until a few years ago, visual features had to be extracted using classical computer vision methods. These included color-based histograms (e.g., hue, saturation, value) and texture-based approaches such as Local Binary Patterns (LBP, Ojala et al., 2002). Generic embedding methods, like the Histogram of Oriented Gradients (HOG, McConnell, 1986) and Scale-Invariant Feature Transform (SIFT, Lowe, 1999), which locate visual keypoints, were also widely employed in digital humanities and digital art history (Chung et al., 2014; Ruf et al., 2010; Seguin et al., 2016). These methods typically aggregated local descriptors into global representations, as with visual Bag-of-Words (Csurka et al., 2004; Sivic & Zisserman, 2003), Vectors of Locally Aggregated Descriptors (VLAD, Arandjelovic & Zisserman, 2013; Jégou et al., 2010), or Fisher Vector coding (Perronnin & Dance, 2007).

**Deep autoencoders.** More recently, deep convolutional autoencoders, trained via unsupervised learning, have become prominent (Mienye & Swart, 2025). Autoencoder models are structured like an hourglass: they first compress (*encode*) an input image into a compact latent representation located at the narrowest point, or “bottleneck”. They then reconstruct (*decode*) the original image from this latent representation. Variational Autoencoders (VAEs) extended this idea by adopting a Bayesian approach, resulting in probabilistic latent spaces (Kingma & Welling, 2022). A variant, Vector-Quantized Variational Autoencoder (VQ-VAE, Oord et al., 2017), encodes images into discrete, learned vocabularies, effectively reducing image regions to visual *prototypes*. A limitation, however, is that computational constraints compel VQ-VAE to apply this prototyping to very small image fragments, often just a few pixels wide. Consequently, the resulting descriptors tend to represent overly simplistic geometric patterns such as line fragments, or points, rather than interpretable image components.

**Debates on component-based versus holistic processing in human vision**. The issue of granularity resonates with longstanding discussions in cognitive science regarding object recognition. On the one hand, theories such as Biederman’s recognition-by-components model (Biederman, 1987) advanced that objects are recognized by decomposition into component parts and subsequent assessment of the spatial relationships among parts (Cave & Kosslyn, 1993). From this perspective, the fragment-based approach of VQ-VAEs appears tenable. Nevertheless, behavioral and neuroimaging evidence has reported that humans also benefit from holistic processing, recognizing certain objects—particularly faces and words—as integrated wholes rather than discrete parts (Ventura et al., 2019; Yovel & Kanwisher, 2005; Zhang et al., 2012). Recent empirical work further suggest that holistic perception might actually emerge from the integrated processing of localized, spatially distributed features (Chang & Tsao, 2017). This indicates that effective representation methods emulating cognitive processing and human perception must not only capture local details but simultaneously integrate them into global structures.

**Vision Transformers function as integrative digital models.** Vision Transformers (ViT, Dosovitskiy et al., 2021) align closely with this concept. Unlike convolutional neural networks—which generate representations through sequential local operations—ViTs explicitly integrate multiple local features into a unified global representation. ViTs thus constitute a compromise between computational constraints and the informativeness of the latent representation. Global integration ensures that the embedding is not only informative about local resemblances but also provides a holistic representation of the image sign.

In this respect, the intuition underlying ViTs resonates with the requirement to produce integrated feature representations of map signs. The remaining question, therefore, is whether the model ought to be retrained—either contrastively or as an autoencoder—or whether a pretrained ViT, that is, a foundation model, is suitable. This choice arguably hinges on two core properties of map signs: (1) iconicity and (2) symbolic simplicity.

**Iconicity.** Iconicity refers to the degree of visual resemblance between a sign utterance and the object it represents. For example, the character **ϙ** can signify a tree because its shape visually evokes one. Likewise, the cross ☨ used to mark a tomb directly reflects the tomb's physical shape[6]. By contrast, purely symbolic signs—such as the letter "A"—exhibit no evident visual relation to their referent, as the glyph bears no direct link to the audible realization of the phoneme [ɑ]. Iconicity and symbolism are best conceived as *gradients* rather than exclusive categories. For illustration, any of the signs [🌳, **ϙ**, **o**, **o**] can represent a tree, albeit with progressively decreasing iconicity.

**Symbolic simplicity**. Map symbols tend to be visually minimalist compared with other semantic-symbolic systems. While certain cartographic icons may be sophisticated, most map symbols (e.g., ◉, ☆, ⊗, →, ×) remain relatively simple. They are typically limited to fewer than five curves or lines, a degree of reduction that surpasses the concision of many writing systems, such as the Chinese script.

**Why foundational models can encode maps signs.** The two properties I enunciated—iconicity and symbolic simplicity—position map signs within a domain that, although specialized, is closely related to the physical environment as it can be captured in photographs and iconographic documents. Icons refer to forms from the natural world (e.g., trees, bushes, grass) or the built environment (e.g., houses, churches, tombs, windmills), thereby aligning with these forms. Because of their simplicity, symbols are also likely to appear in other document types. Therefore, the hypothesis that map icons and symbols can be recognized by a foundational vision model whose knowledge derives from images of the environment and common iconographic documents appears plausible. This proposition justifies the adoption a generic solution to the embedding problem, such as a vision foundation model. Such an approach offers several advantages. First, it capitalizes on the vast diversity of online images. The representation space into which map icons are projected directly references images of the natural or anthropized environment, situating embeddings within a broader structure of geo-cultural knowledge representation. Second, relying on pretrained vision models founded on diverse datasets mitigates the risk of mode collapse[7], which is a notable risk under the imbalanced, power-law distribution of cartographic signs, as any de novo training attempt on such data would demand severe arbitration and trade-offs. Finally, employing a foundational model is energy-efficient, as it entails minimal additional training.

[6] Of course, the physical tomb itself is a sign, since its cross shape descends from the idea of crucifixion.

**DINOv2.** In agreement with the requirements outlined above, the DINO model family was selected. The DINO architecture (Caron et al., 2021) is based on ViT. The model relies on a dataset of 142 million images from the web including, for instance, photographs of both natural and anthropized environment. The images are selected to augment existing curated datasets, such as ImageNet-22k (Deng et al., 2009). DINOv2 (Oquab et al., 2024) demonstrated effective zero-shot capabilities on complex image understanding tasks, including semantic segmentation, keypoints description, monocular depth estimation, and image classification, superseding earlier state-of-the-art supervised approaches on several of these tasks. It is noteworthy that DINOv2 performs equally well on non-photographic, pictorial images, a capability essential to the present use case. The model relies on discriminative self-supervised training based on a student-teacher *self-distillation* mechanism: the student model is presented intentionally noisy training samples (e.g., partly masked, Zhou et al., 2022), and its embeddings are then compared with those produced generated by a teacher model. The teacher model is presented with a non-noisy version corresponding to different crops of the same images. The student thus learns to infer missing or distorted information.

[7] Mode collapse implies that the model overfits on the most frequent signs, learning a latent space that optimally represents them, to the expense of other less common signs.

## 6.3 Embedding adaptation methodology

One characteristic of DINOv2 is that it was trained on large, high-resolution images while map icons and symbols tend to appear at a much smaller observation scale. To adjust for that aspect, the patch-embedding convolution kernel (*patcher*) was downsized to $7 \times 7$ instead of the original $14 \times 14$, by interpolating pretrained weights to the new kernel dimensions (cf. Fig. 4). As shown in Figure 6, this smaller-scale DINOv2 model already appears to produce high-quality embeddings in a *zero-shot*[8] setting.

In Figure 6, all icons extracted from the iconographic map of the Canton of Zürich (Fig. 5) are projected into a two-dimensional space by t-distributed stochastic neighbor embedding (t-SNE, Maaten & Hinton, 2008). Although this step substantially reduces the dimensionality of the original representation space (from 384 to 2), t-SNE generally tends to preserve local relationships, as observed in this instance. The representation space distinctly delineates six large clusters and one smaller cluster.

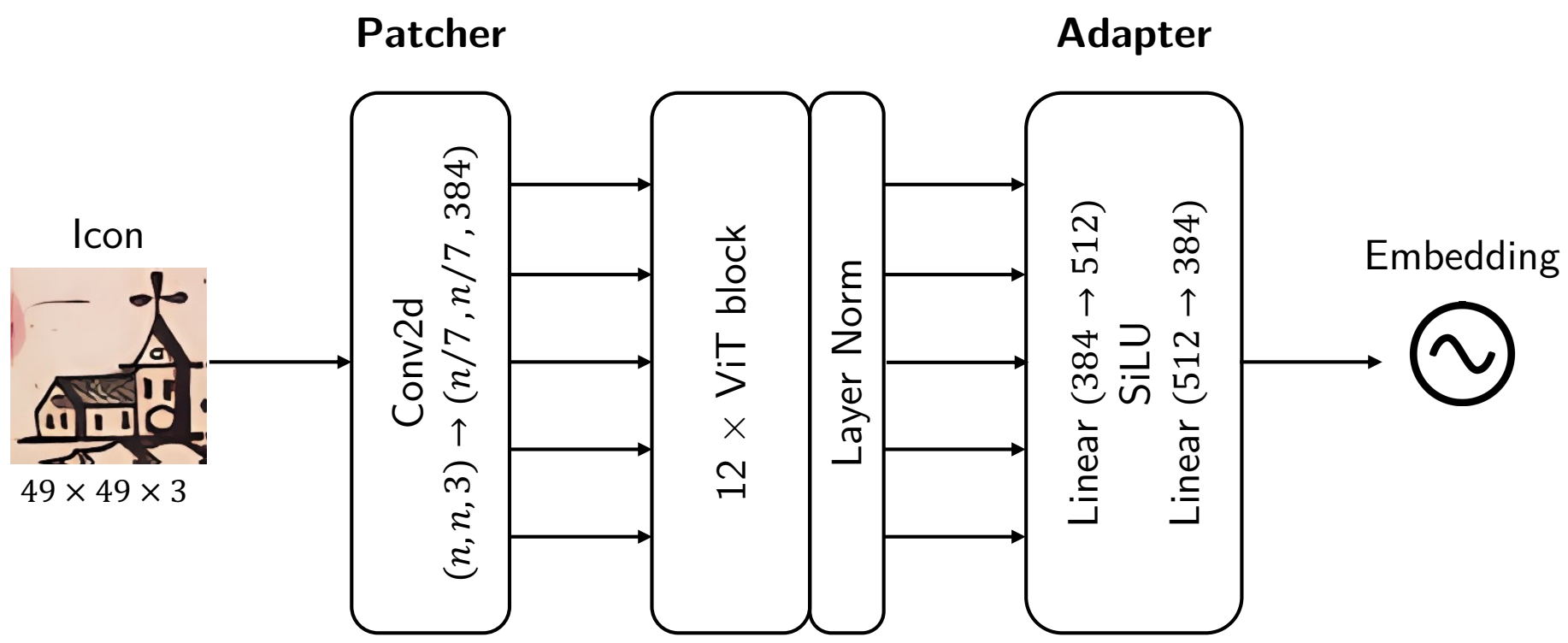

**Figure 4 | The adapted DINOv2 architecture**. The patcher kernel was resized to $7 \times 7$, and an adapter module was appended after the ViT blocks.

[8] i.e. without any complementary training on the particular task or dataset.

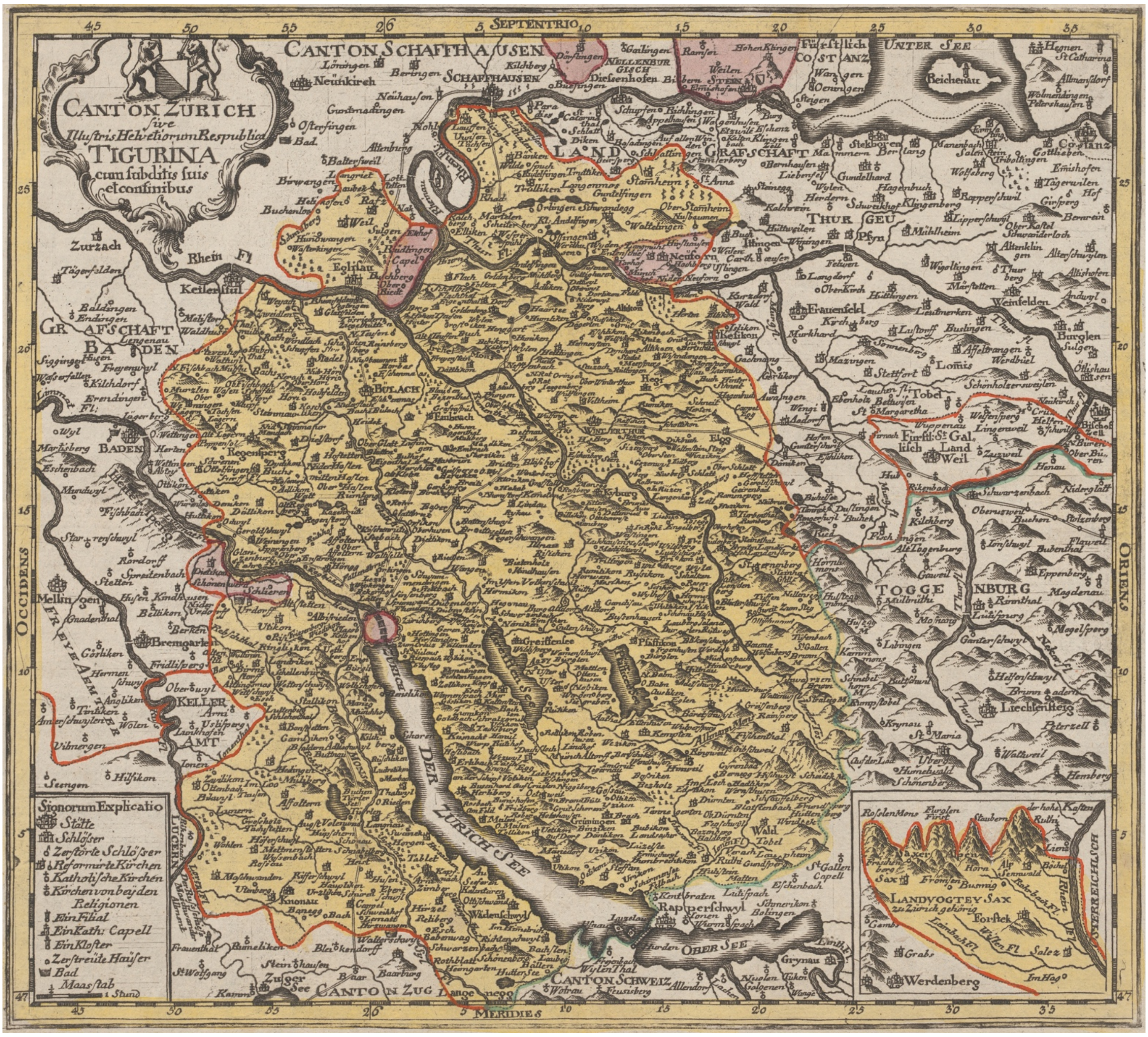

**Figure 5 | Iconographic map of the Canton of Zürich, circa 1770.** Attributed to Gabriel Walser (uncertain). *Canton Zürich sive illustris Helvetiorum respublica Tigurina cum subditis suis et confinibus*. Possibly printed in Nürnberg, Germany, by the heirs of Johann Baptist Homann. ca. 1770. Zentralbibliothek Zürich, 3 Kb 04: 8. 23 × 26 cm. doi: 10.3931/e-rara-27868.

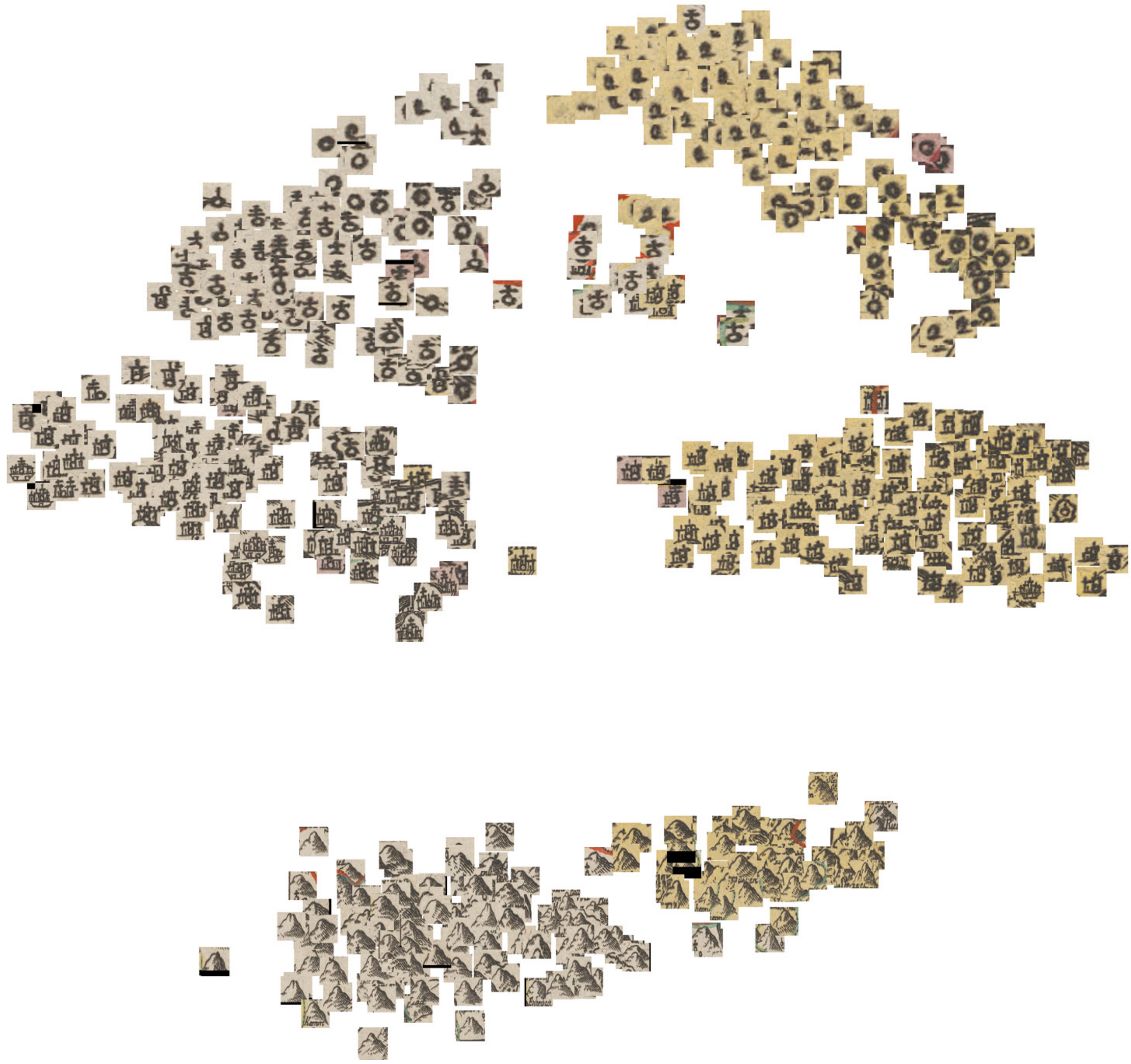

**Figure 6 | Two-dimensional, constrained t-SNE visualization of the raw embeddings**, derived from the icons extracted from Fig. 5. *The icons are undesirably separated by background color. Overlapping objects (e.g. red lines) also affect their spatialization.*

Starting from the top right, the first cluster in Figure 6 consists primarily of trees, depicted as ovals with projected shadows, and small settlements, rendered as circles. Immediately below lies a second cluster containing larger settlements, illustrated by two or three buildings—sometimes arranged around a circle or accompanied by stylized, diamond-shaped fortifications at the bottom. The third cluster, further down, comprises mountains or hills. The three clusters on the left half of the visualization are almost identical to those on the right; the sole difference is the paper color, which is yellow on the right, and white on the left hemisphere. This distinction is due to another apparent graphical choice, where the Canton of Zürich itself is highlighted with a yellow hue (Fig. 5). This causes icon clusters to be duplicated, suggesting that color constitutes a predominant feature within the embedding generated by DINOv2. A smaller cluster further appears in the upper central area of the visualization. It is constituted of icons which, at first sight, do not resemble each other; their only common attribute is the red boundary line demarcating the canton's border.

Thus, the model accurately describes the salient graphical characteristics of the input signs. However, its output may not fully meet our expectations. For instance, encoding the background color, or the graphical elements that overlap with or encroach upon the icon's patch (e.g., the red boundary) is undesirable. Moreover, the portion of the representation that addresses color seems excessive, compared to other aspects, like shape, which are essential to the study of cartographic signs. Besides, the excessive focus on color may later affect the ability to assimilate signs from distinct maps, as colors and contrasts are particularly dependent on digitization conditions. Therefore, it seems necessary to find a way to ensure that the representation focuses primarily on encoding the shape of the sign itself, while ignoring encroaching objects, and moderating the impact of background color.

A possible approach to this issue is to apply color normalization. In addition, a foreground-segmentation convolutional neural network (CNN) can be employed to concentrate the embedding on the center of the image, particularly on the sign itself. In the present research, for instance, a ResNet18 (He et al., 2015) UNet model was trained to predict foreground region from sign crops. Two datasets, one containing synthetic training samples, and one leveraging pseudolabels, were created for this purpose (Fig. 7).

On the one hand, the synthetic samples were generated by superimposing digital icon sprites onto real map crops. In total, 679 unique digital icon sprites created by digital artists for historicized map-art creation, were used. These sprites were overlaid on map crops extracted from the corpus (cf. Chapter 7), with random variation in size, placement, transparency and, where applicable, color. The resulting synthetic images and their corresponding ground-truth masks are illustrated in Figure 7. The pseudolabeled dataset, on the other hand, was created by applying a set of computer vision heuristics to real icon samples. The approximate foreground segmentation method used for pseudolabeling comprised three steps: binarizing the image, center cropping the mask, and retaining the largest connected component—i.e., the largest graphical feature after binarization.

Both synthetic and pseudolabeled datasets exhibit limitations. For one, the synthetic data might not represent the full diversity of historical icons. Moreover, digital overlay may not accurately emulate the physical printing process. Conversely, the pseudolabeled data merely rely on approximate segmentation masks; they may not adequately remove overlapping components or, conversely, may omit portions of icons located near the crop border. Nevertheless, the limitations of each dataset are partially offset by the strengths of the other.

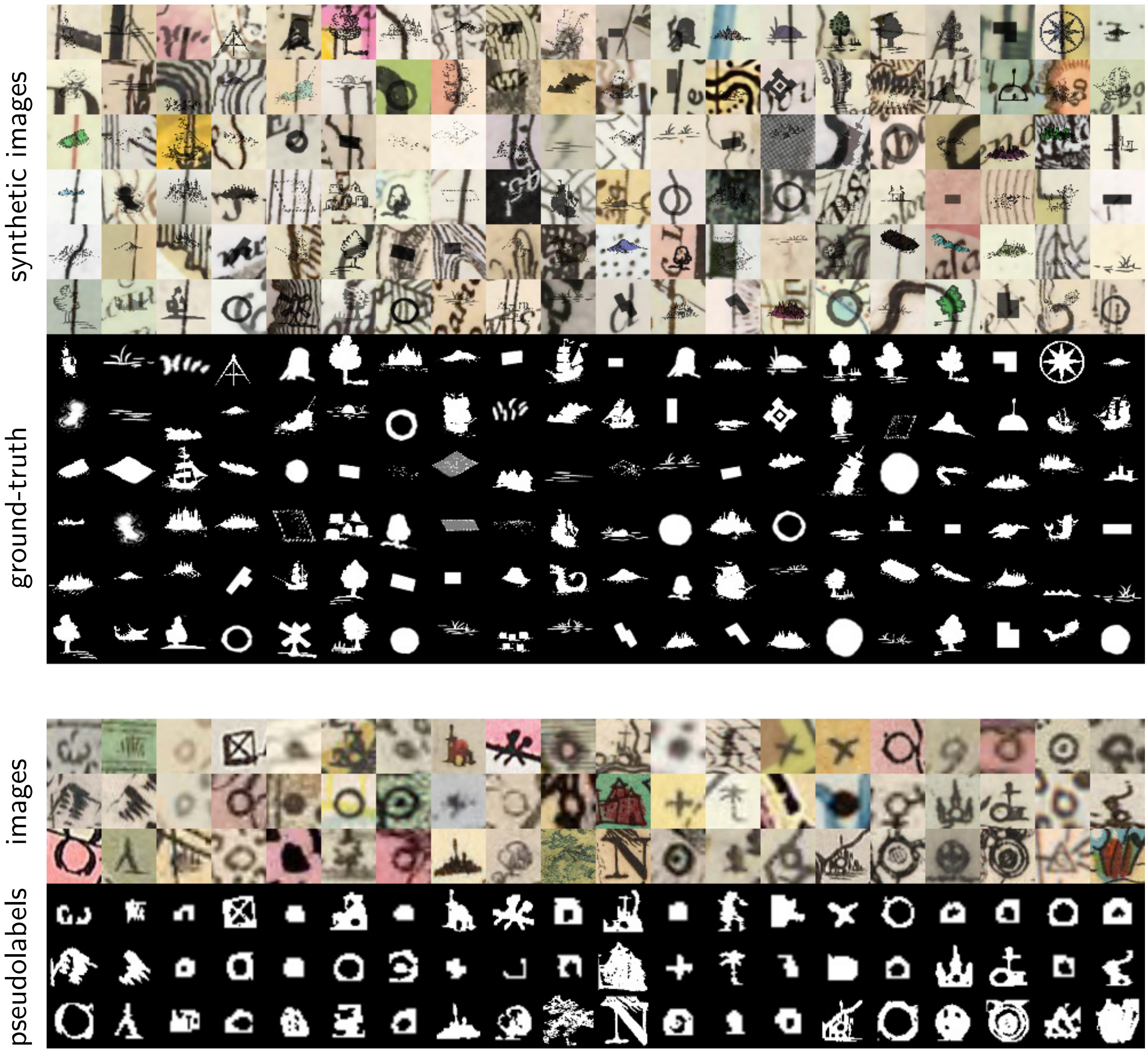

**Figure 7 | Synthetic and pseudolabeled data used for training the background-foreground segmentation model.** (top) Synthetic data are produced by overlaying digital cartographic signs onto map fragments. (bottom) Pseudolabels are created from real sign samples using classical computer vision techniques. *Synthetic data are associated with perfect masks, while pseudolabels help the model generalize on more realistic sign samples.*

Overall, the dataset used for training the foreground segmentation module comprises 3,800 samples, consisting of 2,716 synthetically generated samples (71 %) and 1,084 pseudolabeled samples (29 %). Of these, 3,350 samples are used for training and 450 for validation. The ResNet-UNet model is then trained for 35 epochs with a batch size of 50, employing the Adam optimizer (Kingma & Ba, 2014), and mean squared error (MSE) as the loss function. The model attains a final MSE of $3 \cdot 10^{-2}$ on the validation set. Representative foreground predictions are shown in Figure 8.

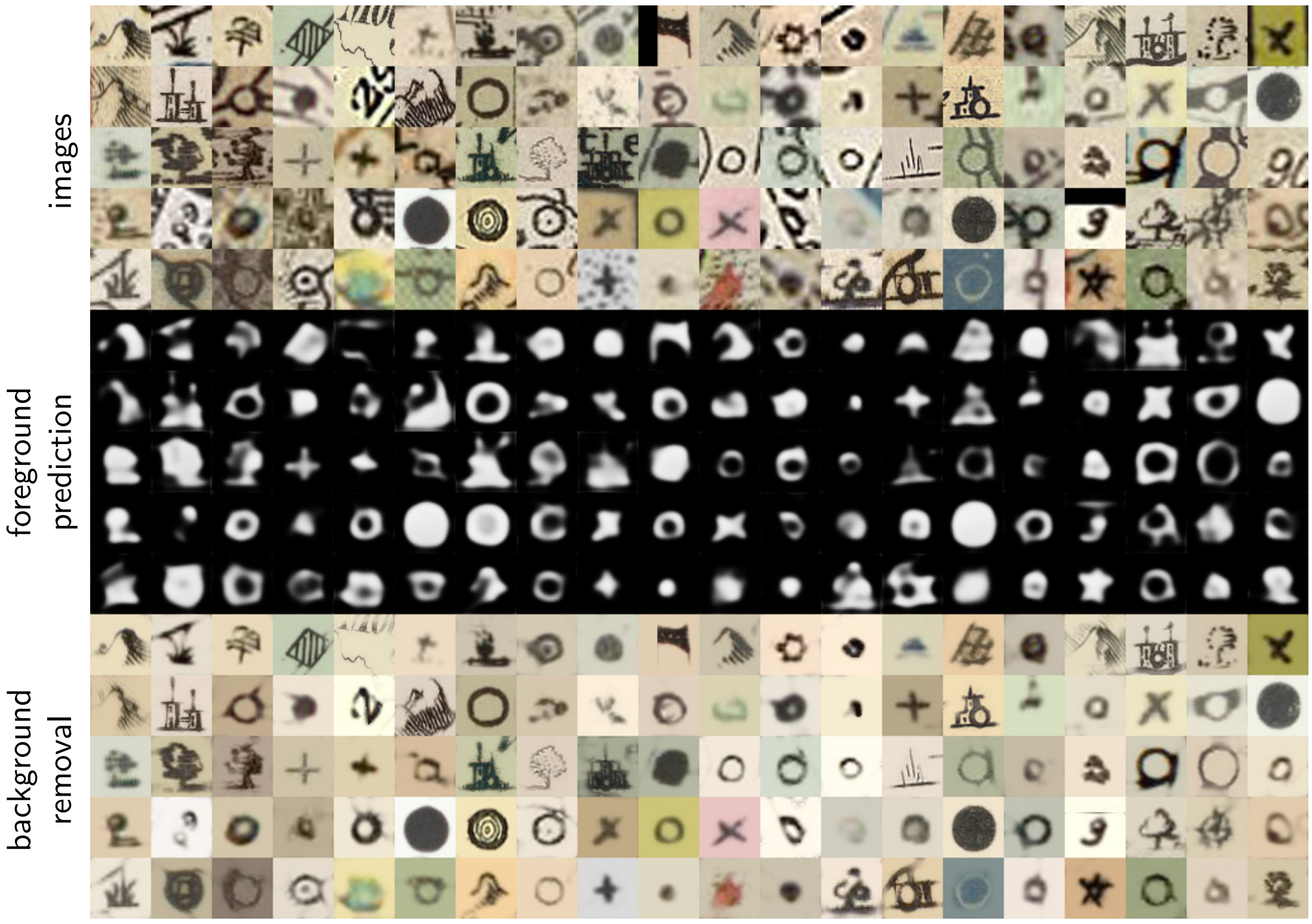

**Figure 8 | Preprocessing strategy.** A smooth foreground mask content is predicted by a neural network, and background content is removed based on the smooth mask. *The foreground segmentation model fades undesirable background content quite effectively.*

Figure 9 presents the updated t-SNE visualization of the sign embeddings after grayscale conversion, min-max normalization, and background removal. Several improvements in the spatialization can be observed. First, the hill icons remain isolated from the other clusters; however, they now form a single cluster rather than two groups separated by background paper color. Conversely, the trees and smaller settlements are now correctly distinguished into two distinct clusters. Small settlements, represented as circles, are grouped in the lower right portion of the visualization. The center of the figure comprises three distinct lobes corresponding to cities and larger settlements on the left, religious edifices (e.g., chapels) in the middle, and trees on the right. The slight connection between these lobes is not problematic; their visible detachment suggests a clear distinction in the higher-dimensional embedding space.

At first glance, the preprocessing step—consisting of background removal and color normalization—appears successful. However, the quantitative tests presented in Sections 6.4 and 6.5 sections will nuance this qualitative assessment. As shown in Figure 9, preprocessing is severe and may occasionally obliterate useful visual cues, including certain sign details. In addition, preprocessing may arguably render the signs more distant from the images on which DINOv2 was trained. Ultimately, even if one wishes the model to concentrate less on color-related features—which may be misleading because of divergent digitization settings—it is not desirable to remove them from the input entirely either.

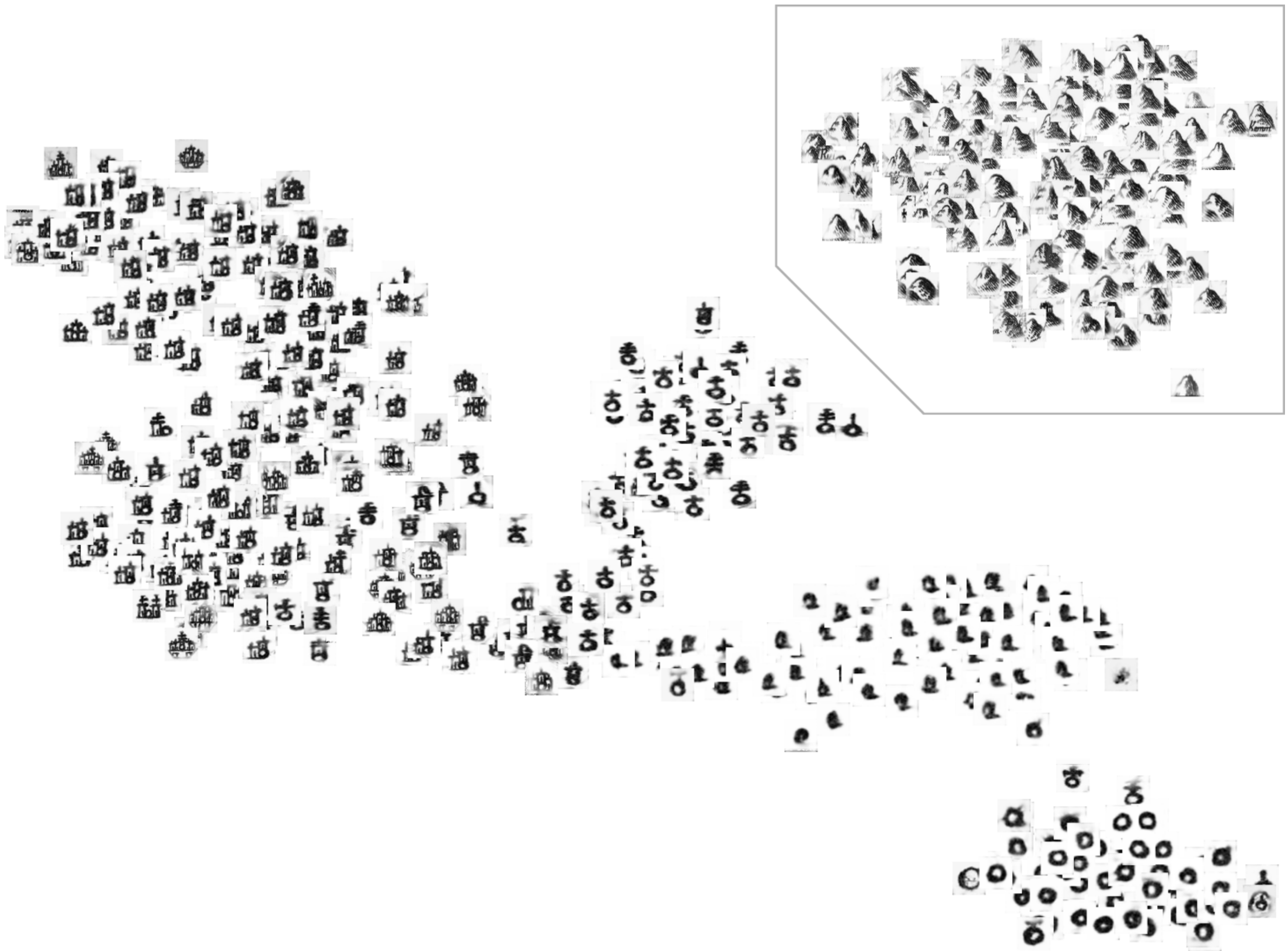

**Figure 9 | Two-dimensional, constrained t-SNE visualization of the embeddings generated after background removal and min–max normalization**, derived from the signs extracted from Fig. 5. The signs were preprocessed by min–max color normalization, background removal, and grayscale conversion. The hills cluster was slightly translated for visualization purposes. *The icons are effectively grouped by shape.*

Consequently, and considering the results that precede, the final embedding methodology is a compromise between the zero-shot and the preprocessing approach. Rather than completely erasing the background and canceling colors and contrasts, the model is modified to focus *primarily* on the cues enhanced by preprocessing—such as the shape of the central icon. Concretely, an adapter module is appended to the DINOv2 model, as was illustrated in Figure 4. The adapter module comprises two fully connected linear layers, the first followed by a Sigmoid Linear Unit (SiLU) activation function. At training time, the DINOv2 backbone was "frozen", and the adapter head was trained to *imitate* the DINOv2 embeddings obtained after preprocessing while receiving the corresponding original, unprocessed images as inputs. Precisely, the model was optimized to maximize the cosine similarity between the embeddings produced by (1) the base DINOv2 fed with preprocessed samples and (2) the DINOv2 augmented with the adapter head, fed with the original samples. It took 117 minutes to train the adapter, using 128-samples batches and AdamW optimized. Optimizing only the adapter head is computationally efficient compared to fine-tuning the entire DINOv2 model. It also prevents the risk of catastrophic forgetting. Replacing the computationally heavy ResNet-based preprocessing with this lightweight adapter also substantially reduces the inference load: the average feature extraction time decreases from 850 ms to 300 ms per batch.

The effectiveness of adaptation was tested on a distinct dataset partition comprising 500 batches. Specifically, the mean cosine similarity between the target embeddings (i.e., those obtained after min–max normalization and foreground segmentation) and two alternative treatments: (1) the base DINOv2 model and (2) DINOv2 with adapter head, was computed. The base embeddings exhibited a similarity of $0.789 \pm 0.008$ with the target, whereas the adapted model achieved $0.961 \pm 0.002$. For reference, the similarity between the adapted and base embeddings was $0.825 \pm 0.006$. These results indicate that the adapter fulfills its intended function.

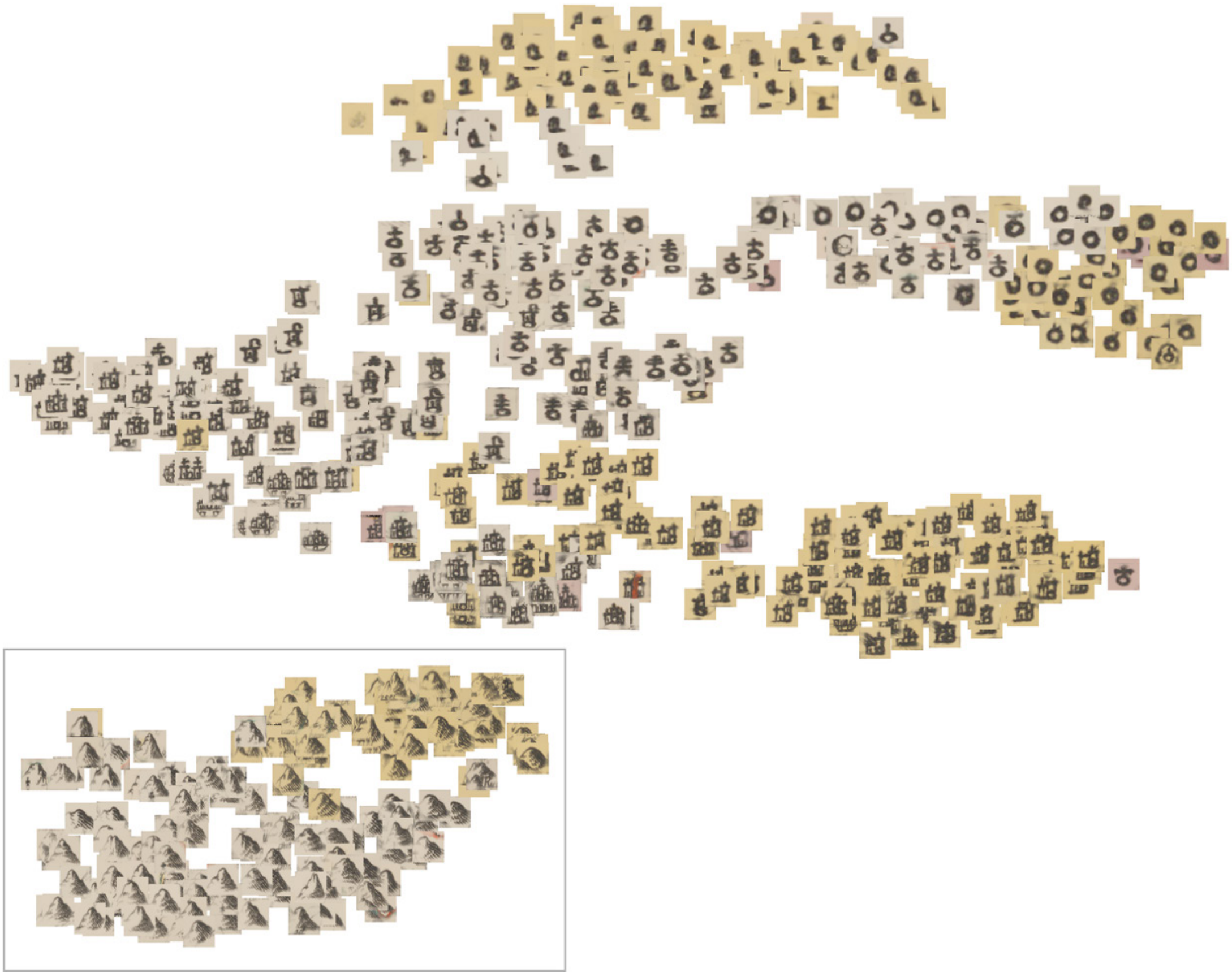

**Figure 10 | (a) Two–dimensional, constrained t-SNE visualization of the embeddings generated by the adapted network**, derived from the signs extracted from Fig. 5. The hills cluster was translated for visualization purposes. *Shape now seems to be the primary determinant of spatial proximity.*

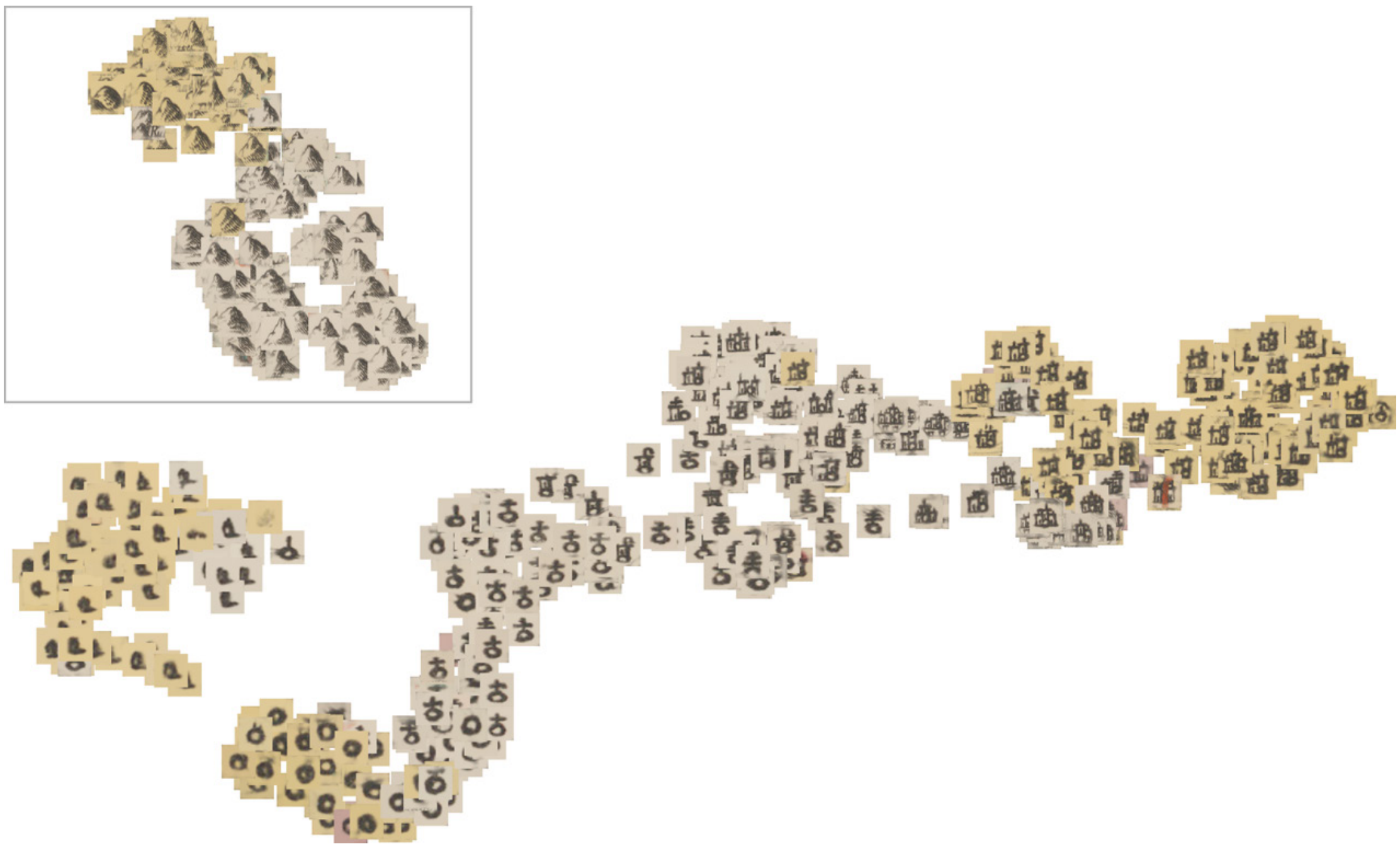

**Figure 10 | (b) Two-dimensional, constrained UMAP visualization of the embeddings generated by the adapted network**, derived from the signs extracted from Fig. 5. The hills cluster was slightly translated for visualization purposes. *Shape now seems to be the primary determinant of spatial proximity.*

The final embedding results are visualized in Figure 10. Subfigure 10a adopts t-SNE spatialization, while Subfigure 10b uses UMAP (McInnes et al., 2020) as an alternative method. In both cases, the hills are correctly separated from the other icons. Here, background color moderately influences spatialization, no longer dividing the clusters as it previously did (cf. Fig. 6). Trees are also correctly distinguished. Small settlements constitute a continuous cluster subdivided into four, or arguably five, lobes: one containing religious places (depicted as a circle topped by a cross) and two to three more indefinite lobes corresponding to larger settlements. Settlements are again spatialized according to color, yet they arguably exhibit a more global and distinctive spatial coherence that primarily reflects sign form variants. This tendency is even more apparent when this outcome is compared with zero-shot DINOv2 embedding (Fig. 6). At last, the “orphan” clusters, formerly caused by encroaching graphical objects (e.g. red lines), have disappeared, indicating that the impact of peripheric visual cues was effectively mitigated. The current outcome therefore appears satisfactory, as it alleviates most of the limitations initially identified.

In the next three sections, the descriptive power of these embeddings is tried through three independent experiments, related to form and content dimension of cartographic signs.

## 6.4 Test 1: Communities and the replication of icons

### *Methodology*

A first way to test the descriptive power of the modified DINOv2 embeddings is to verify their ability to identify replicated signs, and code for cultural or stylistic similarities. One approach to this problem is to consider set of maps produced within homogenous communities of map makers. Presumably, maps created by the same group of makers, or closely related creators, should contain similar signs. In the present case, communities were verified manually, based on the normalized metadata extracted in Chapter 1. Specifically, two maps are considered kindred if and only if they were created by the same individual or, in cases of collective authorship, if at least half of the contributors participated in the creation of both maps.

Relying on larger communities rather than individual creators also mitigates the potential impact of weak predictors—like paper color, grain, or digitization artifacts—which can constitute effective yet presently irrelevant stylistic cues (see Appendix A).

Let $\mathcal{D}$ be the map dataset. These maps were produced by cartographers collaborating within a finite set of communities $\mathcal{C}$. Let $m$ be an individual map, and $\mathcal{M}_c$ the set of maps produced by the community $c \in \mathcal{C}$, i.e. $m \in \mathcal{M}_c \in \mathcal{D}$, or

$$\mathcal{D} = \bigcup_{c \in \mathcal{C}} \mathcal{M}_c$$

Let us also write $c(m) \in \mathcal{C}$ the community that created the map $m$. Each map $m$ is itself considered a collection of icon signs.

$$\mathcal{I}_m = \{i_m^1, i_m^2, \dots, i_m^{N_m}\}$$

where $N_m$ is the number of icons extracted from the map $m$. Each icon $i$ is described by a feature vector $\phi(i)$

$$\phi : \mathcal{I} \to \mathbb{R}^{384}$$

For any two icons $i$ and $i'$ we define the cosine distance, which quantifies the angular discrepancy between their feature vectors.

$$d_{icon}(i, i') = 1 - \frac{\langle \phi(i), \phi(i') \rangle}{\|\phi(i)\| \|\phi(i')\|} \tag{1}$$

The distance between two maps is computed by comparing their respective sets of icons. Intuitively, each icon that matches *at least one* counterpart in the other map contributes to the proximity between the two maps; it is a shared icon. Given two maps $m$ and $m'$ with icon sets $\mathcal{I}_m$ and $\mathcal{I}_{m'}$, we first construct the matrix of pairwise icon distances:

$$\Omega_{j,k}^{(m,m')} = d_{icon}\left(i_m^j, i_{m'}^k\right) \text{ for } j = 1, \dots, N_m \text{ and } k = 1, \dots, N_{m'}$$

The map–level distance is computed as

$$d_{map}(m,m') = \frac{1}{N_m}\sum_{j=1}^{N_m} tanh\left(\min_{\forall k}\Omega_{j,k}^{(m,m\prime)}\right) + \frac{1}{N_{m\prime}}\sum_{k=1}^{N_{m\prime}} tanh\left(\min_{\forall j}\Omega_{k,j}^{(m,m\prime)}\right)$$

The non-linearity $tanh$ functions as a smooth thresholding operation, facilitating the identification of matching icon pairs, located *close* to one another (e.g. $d_{icon}(i,i') < 0.5$), and considered replicas[9].

For the two maps $m_a$ and $m_b$, the quality of belonging to the same community is noted

$$\delta_{map}(m_a,m_b) = \begin{cases} 0, & \text{if } \mathrm{c}(m_a) = \mathrm{c}(m_b) \\ 1, & \text{if } \mathrm{c}(m_a) \neq \mathrm{c}(m_b) \end{cases}$$

Now consider a batch $\mathcal{B} = \{m_1, m_2, \dots, m_n\}$ of maps sampled from $\mathcal{D}$. These maps belong to a subset of communities $\mathcal{C}_\mathcal{B} \in \mathcal{C}$. The set of pairwise computed map distances is

$$D_\mathcal{B} = \left\{d_{map}(m_a,m_b) : 1 \leq a < b \leq n\right\}$$

and the corresponding ground–truth, expressing whether map pairs belong to the same community

$$\Delta_\mathcal{B} = \left\{\delta_{map}(m_a,m_b) : 1 \leq a < b \leq n\right\}$$

Note that the set of pairwise map distances $D$ correspond to the lower triangle of the distance matrix, owing to the term $1 \leq a < b \leq n$. Likewise, the shared-community encoding $\Delta$ matches the lower triangle of the shared-community matrix. The Pearson correlation coefficient between $D$ and $\Delta$ can be computed as

$$\rho_\mathcal{B} = \frac{\sum_{a<b}\left(d_{map}(m_a,m_b) - \overline{D_\mathcal{B}}\right)\left(\delta_{map}(m_a,m_b) - \overline{\Delta_\mathcal{B}}\right)}{\sqrt{\sum_{a<b}\left(d_{map}(m_a,m_b) - \overline{D_\mathcal{B}}\right)^2}\sqrt{\sum_{a<b}\left(\delta_{map}(m_a,m_b) - \overline{\Delta_\mathcal{B}}\right)^2}} \tag{2}$$

The experiment is repeated 10 times, each trial using 17 batches $\mathcal{B}$ of 50 maps each, sampled from a subset $\mathcal{C}_\mathcal{B}$ of 10 communities $\mathcal{C}_\mathcal{B} \subset \mathcal{C}$. For each map, a maximum of $N_m \leq 128$ icons were considered.

[9] Here, the value 0.5 is given as an example. Defining the threshold under which two icons can be considered replicas—provided that such a threshold exists at all—is a challenging problem that will be discussed later.

### *Results & Discussion*

The results of the experiment are detailed in Table 1. The min–max normalization preprocessing does not degrade the ability of the embedding to represent stylometric proximity between maps. This outcome is encouraging, as it partially alleviates the concern that the representation space overfits to weak graphical cues such as paper color or digitization artifacts. Accordingly, it supports the hypothesis that community relationships constitute an adequate context for identifying replicated sign shapes.

**Table 1 | Performance of the stylometric experiment** in each experiment. The computation of $\rho_{\mathcal{B}}$ is detailed in Eq. 2.

| Experiment | $\rho_{\mathcal{B}} \pm$ 95% CI |
|---|---|
| base | $0.321 \pm 0.06$ |
| min-max normalization | $0.324 \pm 0.06$ |
| min-max + foreground segmentation | $0.264 \pm 0.06$ |
| adapter | $0.316 \pm 0.06$ |

Foreground segmentation, however, does affect community prediction. This observation indicates that the preprocessing step erases information that could support style representation. Indeed, the examination of Figure 9 suggested that certain visual details in the linework or sign shapes were erased or faded by the segmentation step. Presumably, this could more strongly affect infrequent icons or details that may be better predictors of style[10]. This result corroborates the qualitative intuition that the impact of preprocessing is overly aggressive and should be attenuated.

This is, precisely, the purpose of the adapter. As the results in Table 1 show, it fulfills its goal: the explanatory power is almost completely recovered. This result indicates that the adapter preserves essential stylistic cues, while attenuating the effect of irrelevant factors, as previously demonstrated.

[10] Indeed, the foreground segmentation model was primarily trained on synthetic data generated from a limited set of icon sprites. Consequently, foreground prediction may work optimally on frequently occurring sign parts, but poorly on rare details.

## 6.5 Test 2: Sign Semantics

### *Aim & Methods*

The aim of the second test is to validate the descriptive power of the generated embeddings to encode content. It consists of a supervised classification task in which each image sample is assigned one of the 24 semantic classes (e.g. tree, mill, or hill), determined during the annotation process described in Section 6.1. Hapaxes are excluded, reducing the number of sign samples from 18,750 initially to 18,202. Classification is performed via linear probing. First, the samples are randomly partitioned into training (80 %) and validation (20 %) sets; the features are standardized relative to the training set. Then, a multinomial logistic regression classifier is trained on 24 categories. Performance is computed as the class-weighted averages of precision and recall, across 100 repetitions.

### *Results & Discussion*

The test results are shown in Table 2. Contrary to the previous experiment (Section 6.4), min–max normalization does appear to reduce performance, relative to the base condition. This effect is possibly indirectly caused by the lower diversity of document sources in the second test compared with the first (Section 6.4). Indeed, in the present case, the samples derive from only 796 documents, and the classes are unevenly distributed across them. Consequently, the model may concentrate on *weak* yet highly predictive document features related to digitization or paper characteristics. A complementary experiment detailed in Appendix A, corroborates this interpretation, demonstrating the capacity of these *weak* features to introduce a classification bias.

Foreground segmentation weighs even more on both precision and recall, confirming the excessive intensity of this preprocessing intervention. Again, the decrease in performance could be attributed to the fading of small, albeit significant, details.

**Table 2 | Performance metrics for the classification of icons into 24 semantic classes**. In each experiment, classification is conducted via linear probing on the embeddings. For both precision and recall, the reported intervals denote the 95% confidence interval of the mean. Detailed per-class performance and the corresponding confusion matrix are provided in Fig. 11 and Fig. 12, respectively

| Experiment | Precision [95% CI] | Recall [95% CI] |
|---|---|---|
| base | 73.8% [72.7–74.8] | 75.1% [74.1–75.9] |
| min-max normalization | 72.6% [71.8–73.8] | 74.0% [73.0–75.0] |
| min-max + foreground seg. | 67.2% [65.8–68.2] | 69.3% [68.1–70.1] |
| adapter | 72.7% [71.0–74.1] | 74.4% [73.3–75.4] |

The adapter module helps restore performance, scoring only marginally lower than the base condition. This behavior is consistent with the results of the first test. With simple linear probing, the model attains a precision of 72.7% and a recall of 74.4%, which suggests that DINOv2 encodes map signs not only as shapes but also, to some extent, semantically. As discussed in Section 6.2, this may be fostered by the relatively high iconicity of cartographic signs and the simplicity of map symbols.

Figure 11 presents the F1-score of classification, computed as the average of precision and recall, for each class. Figure 12 provides an alternative depiction of the same results, showing the confusion matrix of the classification experiment.

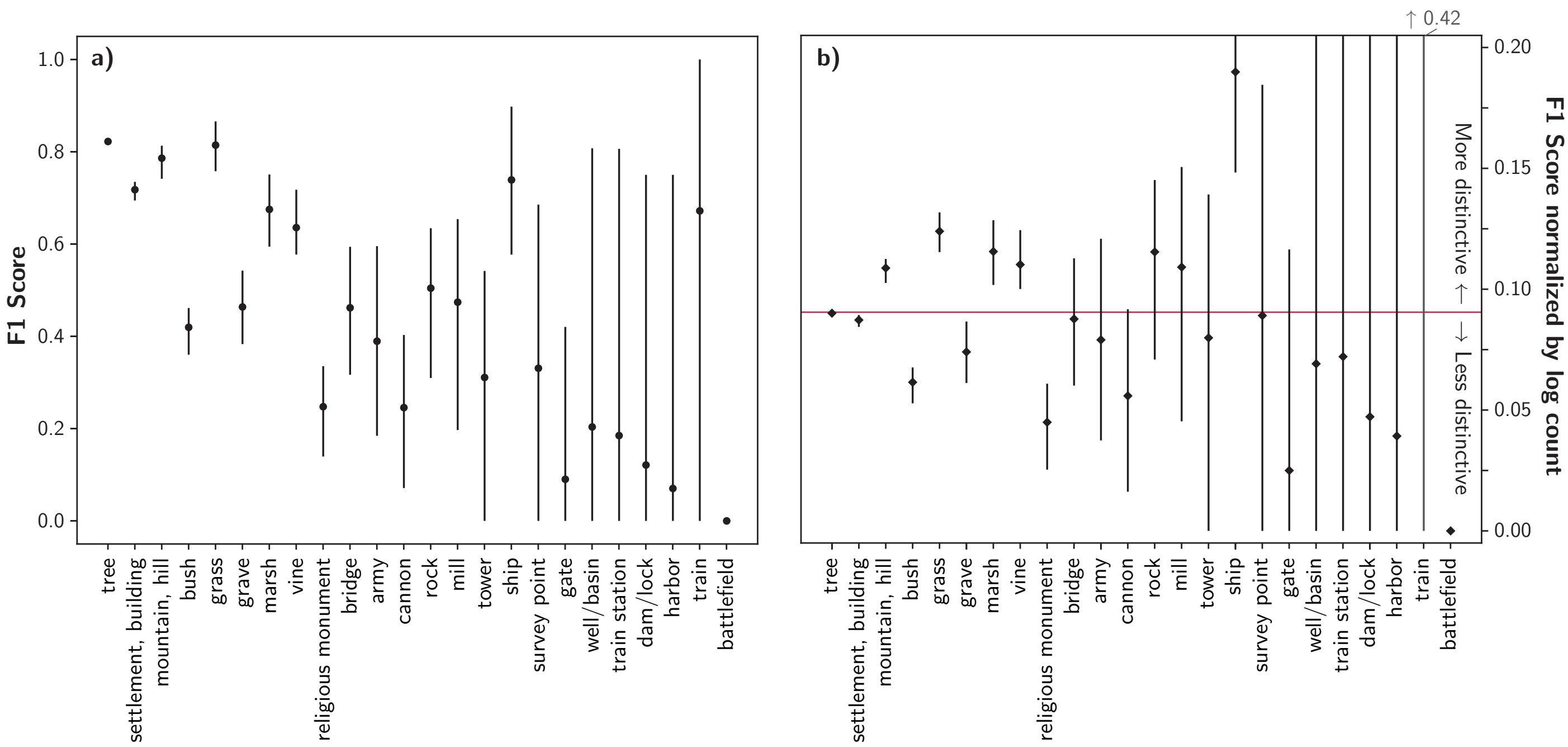

**Figure 11 | F1-Score of the sign classification task, for each semantic class** computed from the embeddings generated by the adapted DINOv2. Using a multinomial logistic regression classifier and 24 classes. The reported results correspond to the validation set ($n_{val}$ = 3,640). The vertical bars denote the 95% confidence interval of the mean across 100 trials. Labels are ordered by their representation in the training set (from $n_1$=6,679 for trees to $n_{24}$=4 for battlefield). The mean precision and recall values are reported in Tab. 2. Subfigure **(a)** reports the original F1–Score for each class, where F1–Score = (Precision + Recall) / 2, whereas subfigure **(b)** presents the F1–Score after log normalization based on the class counts in the training set. *E.g. grass is well detected, especially considering its representation in the training data.*

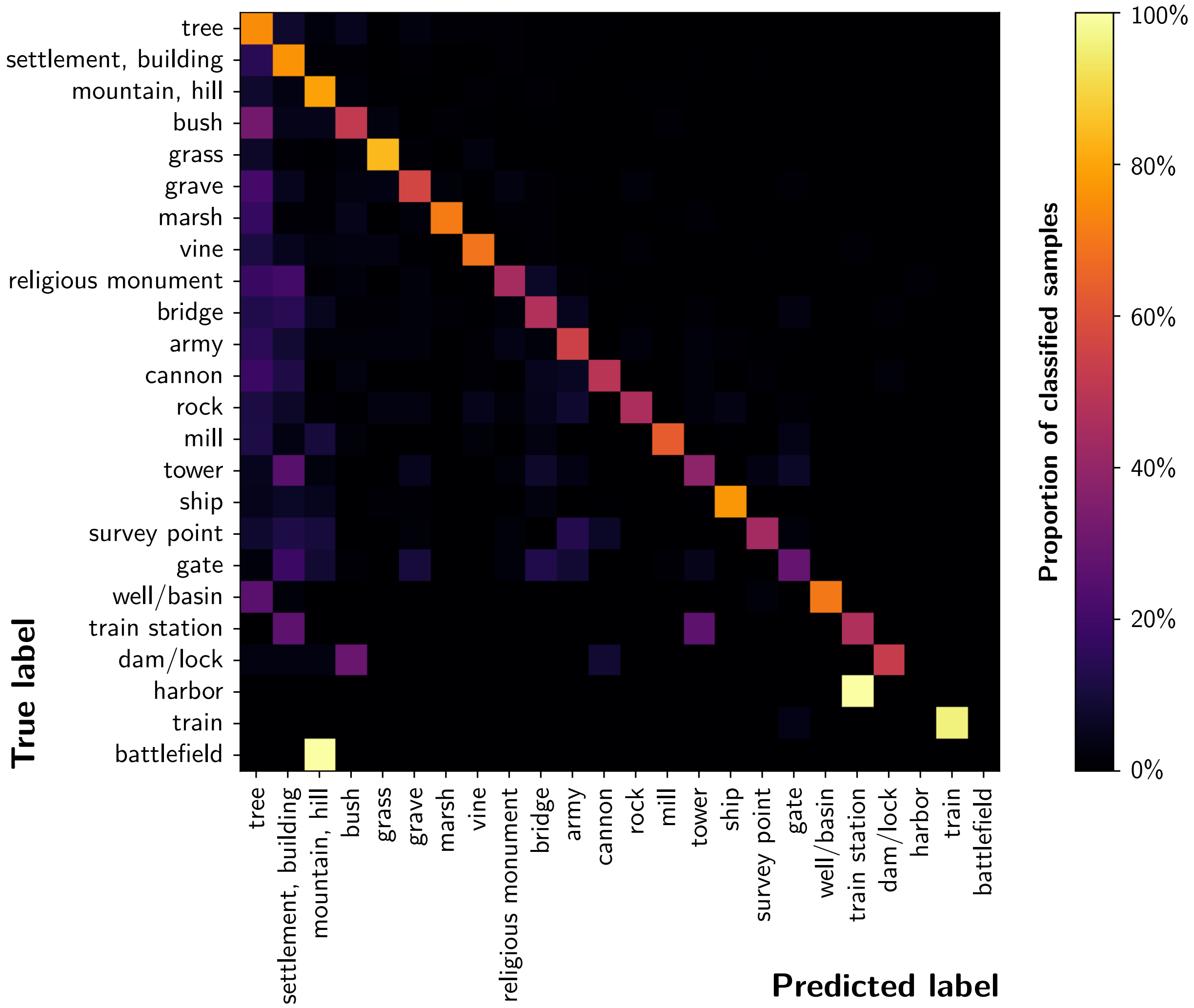

**Figure 12 | Confusion matrix for the sign classification task**, computed from the embeddings generated by the adapted DINOv2. Using a multinomial logistic regression classifier and 24 classes. The reported results correspond to the validation set ($n_{val}$ = 3,640). Color intensity denotes the proportion of samples assigned to each predicted label, normalized for each row. The diagonal entries therefore represent recall. Labels are ordered by their representation in the training set (from $n_1$=6,679 for trees to $n_{24}$=4 for battlefield). The mean precision and recall values are reported in Tab. 2. *E.g. bushes are frequently mistaken for trees.*

Figure 11a indicates that four classes—grass, hill, settlement, and tree—are detected with comparatively high F1 accuracy, ranging from 72 % to 82 %. Marsh (68%) and vine (64%) classes seem to be likewise well detected, whereas the bush class (42 %) seems markedly more difficult to predict. Figure 11b complements these findings by log-normalizing the F1–Score based on the class counts. This adjustment controls for the effect of training sample size and enables an assessment of whether each class performs in accordance with its representation in the training data. From this viewpoint, grass stands out for its good performance, along with marsh, vine, and hill. Ship icons are especially well recognized, despite the comparatively small number of labeled examples. In contrast, religious monument, bush, and grave classes remain hard to distinguish, as does the infrequently occurring battlefield class.

The relatively good F1-scores of the grass and marsh classes can be illustrated by the samples shown in Figures B4 and B5d of Appendix B. Both classes exhibit a measure of symbolic simplicity. Grass is often represented with a few vertical lines, whereas marsh icons typically comprise both vertical lines and horizontal ones symbolizing water. While they could be confused, especially with each other, this simplicity confers them a strong visual identity and distinctive embeddings. They also differ markedly from the most common classes (tree, settlement, hill, and bush), which further reduces the confusion likelihood. Figures B3 and B5b indicate that hill and vine classes are also generally represented with distinctive iconographic signs. The use of the barred S icon $ to represent vine on historical maps is well documented since the 16th century (Dainville & Grivot, 1964, p. 210; Mourey, 2022). The good performance of the model at recognizing ships may be supported by the high level of iconicity of those signs and their occurrence in the original DINOv2 training data.

The confusion matrix (Fig. 12) suggests that the difficulty in classifying bushes arises from their proximity to trees (Figs. B1, B4a). This outcome could be anticipated from the examples in Figure 2, where trees and bushes can hardly be distinguished. Likewise, graves (Fig. B5a) tend to be confused with trees. Indeed, graves, like trees, are typically located in “green” areas. Whether represented as a Latin cross (✝), a forked cross (Y), or even a simple cross (×), these symbols may resemble tree icons. This phenomenon may also affect religious monuments, which are sometimes depicted as a simple Latin cross or incorporate one. Additionally, religious monuments (Fig. B5c) can be confused with other buildings or settlements. This may be in part due to the ambiguous conceptual distinction between settlements—towns and villages—and parishes. As observed in Figure B2, towns are often figured by the icon of a church, sometimes flanked by two other buildings.

## 6.6 Test 3: Features explanation

### *Motivation*

The results of the above two tests converge on three points. First, the basic model accounts for more variance than the model that employs foreground segmentation and min–max normalization, indicating the severity of this preprocessing approach. This difference may arise from the fading of certain details, which arguably alter the signs morphology. The comparatively lower performance, however, presumably also results from a diminished influence of highly predictive, yet uninformative, *weak* features such as paper color or grain. Second, in both tests, adaptation successfully narrows the performance gap observed between the base condition and preprocessed features. This outcome could signal the partial recovery of information lost through the severe preprocessing. Third, in both cases, the adapter model tends to perform slightly lower than the base model. However, this decline may not be detrimental as the adapted model could simply be attending to more explanatory features, at the expense of *weak*, uninformative cues. A third short experiment is here conducted to validate this interpretation.

### *Methodology*

The main objective of the adapter strategy was to enhance the ability of the generated features to inform about aspects like sign shape or color, while attenuating the effect of background color and the morphology of encroaching objects, considered irrelevant. To estimate the extent to which the representation spaces satisfied these expectations, the pairwise distances within each embedding space were compared to the distances obtained from using classical computer vision descriptors. The underlying strategy is to assess the ability of the representation spaces to represent meaningful visual information. Baseline descriptors were computed using two distinct techniques targeting, the color, and the morphology of cartographic signs, respectively.

**Color.** The color space is approximated as the mean color, derived from the red, green, and blue (RGB) channels of the digital image. Specifically, the mean weighted color is computed for both the background and the foreground of each sign sample, using the logits as weights for the computation of the average, resulting in two distinct values, representing the average paper and sign color, for each sample

**Shape.** The histogram of oriented gradient (HOG) is used as a baseline method to compute morphological descriptors (McConnell, 1986). The key idea behind the HOG algorithm is that shapes can be effectively captured by the distribution of edge orientations, analogously to a rough sketch. Prior to the advent of convolutional neural networks, HOG descriptors were commonly used for object recognition and image classification tasks (Cao et al., 2011; Dalal & Triggs, 2005; Freeman & Roth, 1994; Liang & Juang, 2015). In the present experiment, gradients are computed on $9 \times 9$ pixel cells. A sliding window iterates over cells, both vertically and horizontally, storing orientations in 9-bin histograms, resulting in 729-dimensional shape descriptors.

**Comparison approach.** In this experiment, a sample of 8,000 signs is randomly drawn from distinct maps. Icons are embedded using DINOv2, with or without adaptation or preprocessing. In parallel, RGB and HOG descriptors are computed on the same sign samples. Distances between all 32 million unique sign pairs are then calculated, using cosine distance for features derived from DINOv2 or HOG and Euclidean distance for RGB descriptors. Structural similarity between the representation spaces is defined as the linear relationship between paired distances, quantified by the Pearson coefficient.

### *Results*

Table 3 reports the results of the feature explanation experiment. As anticipated, color normalization degrades the descriptive power of the embedding for both background and foreground colors. Increasing contrast, however, appears to slightly enhance shape saliency. Foreground segmentation improves the representation of sign color as well as shape description. Nevertheless, and despite these improvements in the interpretability of target visual features, foreground segmentation significantly degrades the ability of the embeddings to capture style (Tab. 1) and semantics (Tab. 2).

The introduction of the adapter, however, brings substantial improvements. Relative to foreground segmentation, the adapter strategy produces features that represent shape and sign color significantly better, while attenuating the impact of background color. The contrast with the base method is even more striking. Whereas the representations generated by the original DINOv2 encoder focused essentially on background color and the color of encroaching objects, the adapted model seems essentially representative of the sign's color and shape. These results therefore confirm the descriptive improvements brought about by the introduction of the adapter module, and comforts previous interpretations on the pertinence of the embedding strategy.

**Table 3 | Ability of the embeddings to describe aspects of sign color and shape**. Each value indicates the linear association between distances in the representation spaces, measured on 31,992,000 sample pairs using Pearson's coefficient. The confidence interval is negligible. The best scores are highlighted in bold; an asterisk (*) indicates that lower values are preferable.

| Experiment | Background color (RGB) | Foreground color (RGB) | Shape (HOG) |
|---|---|---|---|
| base | 36.4% | 19.5% | 18.4% |
| min-max normalization | 25.6% | 18.7% | 19.7% |
| min-max + foreground seg. | 25.4% | 22.2% | 21.2% |
| adapter | **25.0%*** | **30.7%** | **28.2%** |

## 6.7 Exemplars

The first half of this chapter involved the conception and validation of a meaningful space of representation for map signs. This helped coding the 63 million extracted signs in a mathematical space, within which distances and similarities are consistent with cultural expectations. The next step involves developing a methodological framework that enables the investigation of signs at scale and the study of their broader organization as semiotic systems. The next sections will also develop visualization approaches to ensure the interpretability of the results. The adopted framework relies primarily on the concept of *exemplar*, defined hereafter.

### *Theoretical framework*

The first step aims to make the embedding space more intelligible. This process entails reducing icons to a small number of prototypes, or representative *exemplars*, which also involves the identification of categories. The concepts of category and *prototype* are pervasive in deep learning and computational humanities research (Battleday et al., 2021; Kozlowski et al., 2019; Snell et al., 2017). A prototype is defined as an abstract model that exhibits the average features of the category it stands for. It may also be conceptualized as the *mean* of all entities within that category. The prototype is not merely a concept of the computational sciences. Rather, it is an empirical concept originating in cognitive psychology (Hampton, 2006; Posner & Keele, 1968; Rosch, 1975). Identifying prototypes or cultural categories using unsupervised approaches, like clustering, is a notably difficult task (e.g. Uglanova & Gius, 2020). Representation spaces are imperfect and, therefore, subspaces corresponding to distinct object categories are often entangled. Although mixture approaches exist, the normality they assume rarely accommodates real data. Because finding prototypes implies selecting exactly one model per category, the empirical lack of unimodality often complicates the task.

A more appropriate concept is that of *exemplar*. Unlike a prototype, an exemplar is a concrete entity, rather than an abstract model. Furthermore, each category may be represented by multiple exemplars. The operationalization of the concept of exemplar is facilitated by the adaptability of the construct to multipolar or hierarchical categorization. The theoretical underpinning is just as rigorous as that of prototype; contemporary cognitive psychology recognizes both concepts, which are not necessarily considered antinomic, and both rest on robust evidence (Nosofsky & Zaki, 2002). Therefore, both prototypes and exemplars coexist within the current scientific consensus (Ashby & Maddox, 2005; Bowman et al., 2020).

An example may help clarify the distinction between the concepts of prototype and exemplar. When the concept of *city* is invoked in a discussion, for instance, you are able to cognize the concept by forming an abstract mental image of the idea *city*, without having to concentrate intentionally on forming this image. That image may borrow indistinctly from many cities you know. As such, it constitutes an *average*, intangible prototype. This mental model probably integrates the most common features of a city, and can help someone follow a discussion, or read a text, without having to think about specific, concrete, instances of cities. Conversely, one may mobilize the image of specific *exemplars* of cities, like London, Paris, or New York; typically, cities they have visited, or culturally significant examples. Recalling such *exemplars* may facilitate reasoning about concrete aspects entailed by the concept of a city, for instance when comparing distinct urban centers. Likewise, in the present work, exemplars may help represent a discrete portion of the sign space.

### *Methodology employed for the identification of exemplars*

Certain maps feature thousands of signs; here, the total sign count is capped to 256 instances per map to balance the impact of each individual document on the analysis. After subsampling, the overall sign count is 16,132,327, against 63,180,211 initially. Signs are then grouped into $2^{14}$=16,384 distinct clusters using a mini-batch k-means (Arthur & Vassilvitskii, 2007) algorithm, with a mini-batch size of $2^{17}$=131,072 samples. The algorithm converges after 123 iterations. In k-means, samples are assigned to the nearest cluster center irrespective of relative densities. To address this limitation, a density-aware reclustering step is implemented. The 16,384 initial cluster centers are upsampled proportionally to the estimated cluster densities, with a minimum of one sample per initial cluster, resulting in a total of 131,072 data points, representing densities. A Gaussian mixture model is then fitted to the resampled data, using k-means cluster centers for initialization. The resulting component centers coincide with the k-means centers. However, the Gaussian mixture model weights components by their relative densities, so that larger clusters exert proportionally greater influence on attribution compared to smaller ones. In k-means, a sample located halfway between two cluster centers would have an equal likelihood of being assigned to either cluster, regardless of the surrounding densities. With the reclustering approach, the same sample is more likely to be assigned to the nearest center with higher density. At inference time, only the Gaussian mixture model is employed. For each cluster, the sample closest to the center is selected as the exemplar. The distribution of final cluster sizes is reported in Figure B8 in the Appendix.

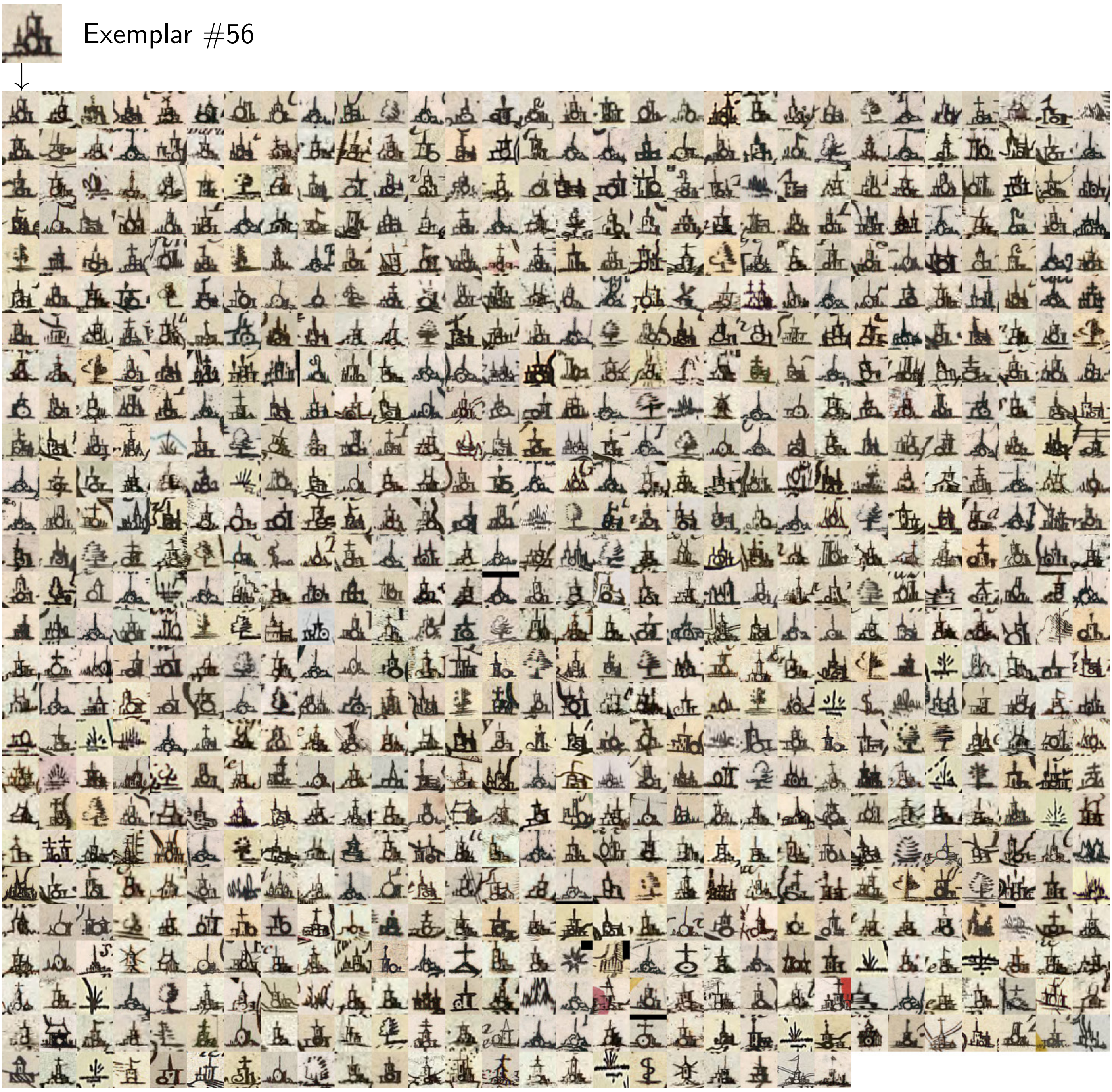

**Figure 13 | Visualization of all signs attributed to cluster #56.** Signs are arranged by their probability of belonging to cluster #56, row by row. The first sign is selected as the exemplar. *Clusters comprise a diversity of signs; each exemplar primarily represents the signs most confidently attributed to this cluster—located at the top of the Figure.*

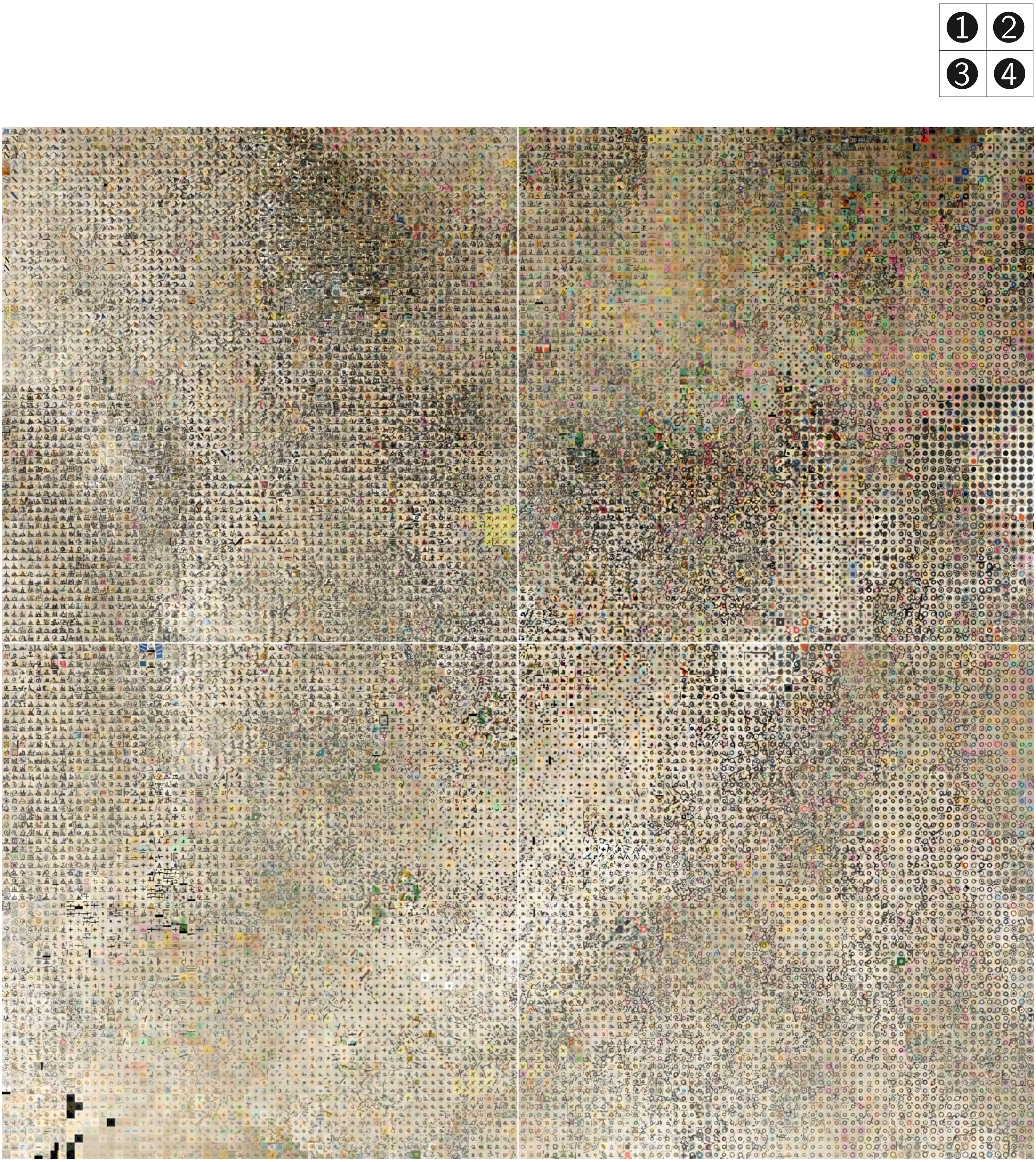

**Figure 14 | Sign mosaic.** Representing the exemplars of each of the 16,384 sign clusters. The exemplars are spatialized by t-distributed stochastic neighbor embedding (t-SNE). The mosaic is subdivided into four quadrants to facilitate analysis and discussion. Enlargements are provided in Figs. B13–B16, in the Appendix. *The sign mosaic represents the embedding space in a more interpretable form. The spatialization presented here will serve as the basis for the analysis and discussion.*

To provide a clearer sense of what such a cluster might contain, Figure 13 presents a visual example of cluster #56 and its exemplar. The exemplar is followed by all signs in cluster #56, ordered by their probability of belonging to that particular cluster. Clustering is not restricted to visual similarity only; the embedding space visibly encodes both semantic and stylistic information. Consequently, the results can appear noisy, especially for complex signs like this one. This example also underscores the difficulty of identifying prototypes. Nevertheless, considering the effective diversity, the exemplar adequately represents the first few lines, corresponding to reliably classified samples.

Additional examples of sign clusters and their exemplars are presented in Figures B9–B12 in the Appendix, which show 256 randomly selected sign clusters, ordered chronologically by average publication date.

*Sign mosaic*

Figure 14 displays the sign mosaic, a visual device spatializing the exemplars of each of the 16,384 sign clusters. The mosaic was computed by t-distributed stochastic neighbor embedding (t-SNE). The visualization seeks to render the much larger space of signs more intelligible; it is subdivided into four quadrants to facilitate analysis and discussion. Enlargements of quadrant are provided in the Appendix (Figs. B13–B16).

## 6.8 Phylogeny and variation

The mosaic (Fig. 14) illustrates the sheer diversity of map signs. However, it constitutes a severely constrained and dimensionally reduced representation of the initial 384-dimensional space. The color hues and values of the phylogenic mosaic (Fig. 15) illustrate another reduction of the same space, depicted with three color dimensions. The spatialization is identical to that shown in Figure 14. The dimensionality reduction used for colorization, however, is achieved through a distinct algorithm, namely Uniform Manifold Approximation and Projection (UMAP). Both t-SNE and UMAP algorithms are foremost intended for visualization and tend to preserve local proximities and relationships. However, UMAP tends to preserve the global structure more effectively (McInnes et al., 2020).

It is noteworthy that the phylogenic mosaic does not display the complete extent of color shades. Instead, it is divided in a few "color nebulae", or phylogenic regions, corresponding to supercluster structures. For instance, the magenta region (■■ Quadrant 1) is predominantly populated by hill icons. The lavender region (■■ Q1,3) contains settlement icons, whereas the purple region (■■ Q1) corresponds to buildings. Within the second quadrant, two olive-green regions (■■ Q2) comprise full-circle symbols, while hollow circles are found in the orange regions (■■■ Q4). The mint-green region (■■ Q3) is occupied by various pastel-colored tree icons.

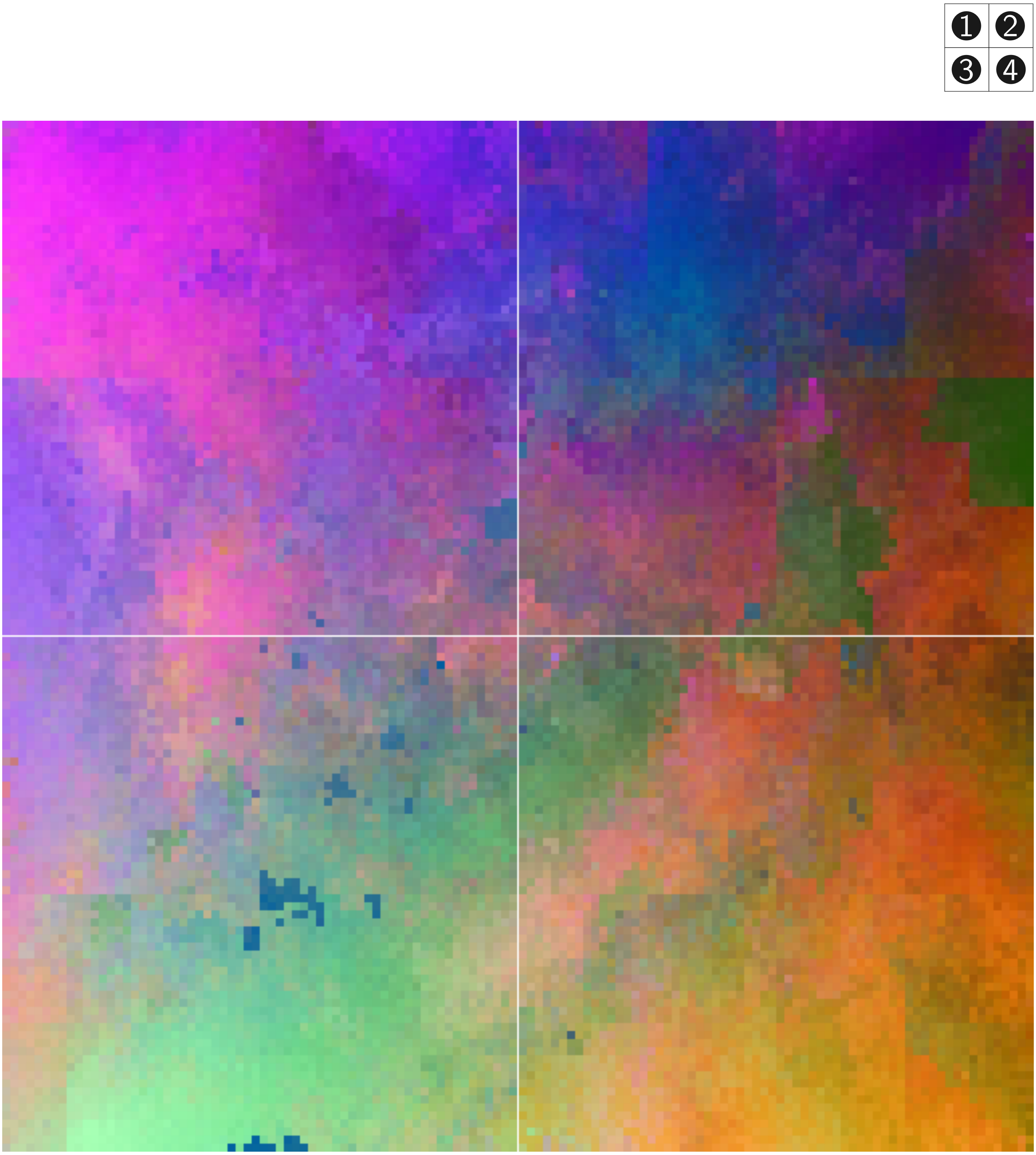

**Figure 15 | Phylogenic mosaic** representing the exemplars of each of the 16,384 sign clusters. Exemplars are spatialized by t-SNE, as in Fig. 14. The values of the red, green, blue (RGB) color channels reflect an independent UMAP dimensionality reduction. The mosaic is subdivided into four quadrants to facilitate analysis and discussion (see Figs. B13–B16 in the Appendix). A color-accessible version of this figure with clear distinction of the regions mentioned in the text, is provided in Fig. B20, in the Appendix. *The continuity of color hues indicates that the space of signs is structured hierarchically.*

This result illustrates the effect of *variation*. The exemplars depicted in Figure 16 show that kindred branches correspond to related, yet slightly different, sign *variants*. Variation is one of the core principles of cultural evolution, often summarized as a process of "variation and selective retention" (D. T. Campbell, 1960, 1965). Whether it stems from deliberate intervention or from unintentional divergence, variation can shift the entire system slightly (D. T. Campbell, 1974). A misfitted icon, for instance, may incite the mapmaker to adjust the corresponding punch, thereby introducing variation. Alternatively, printing errors may arise, persist, and gradually lead readers to accept an increasing amount of sign variants. Each variation is also likely to prompt further adaptations and, ultimately, generate additional variants. This feedback loop constitutes the main mechanism that drives cultural evolution.

The absence of strict, clearly defined boundaries and form categories in the phylogenic also indicates that the space is structured *hierarchically*. The phylogenic tree of map signs (Fig. 16) visualizes this hierarchical structure as a circular dendrogram. The tree was obtained through agglomerative hierarchical clustering of the exemplars, using Ward's variance-minimization method (Ward, 1963). Figure 16 employs the same color codes as Figure 15. In the dendrogram, each exemplar corresponds to a leaf. Branch thickness is indicative of the number of downstream exemplars, i.e., clusters. Moreover, the fifth and sixth hierarchical levels are illustrated with representative exemplars.

For the orange (■■■ Q4) and purple (■■ Q1) regions, branches appear thicker compared the mint green (■■ Q3) and olive green (■■ Q2) regions, for instance. Thicker branches indicate that the leaves—i.e., the exemplars—are located farther from the core, indicating that the tree is deeper on that side. At each iteration, Ward's algorithm merges branch pairs whose fusion results in the smallest possible increase in overall within-cluster variance. Thus, a deeper subtree also indicates that the corresponding region is more *heterogeneous*. Concretely, this entails that the regions corresponding to hollow circles (orange ■■■ Q4) and dwellings (purple ■■ Q1) are visually more heterogeneous than those corresponding to full circles (olive green ■■ Q2) or hills (magenta ■■ Q1).

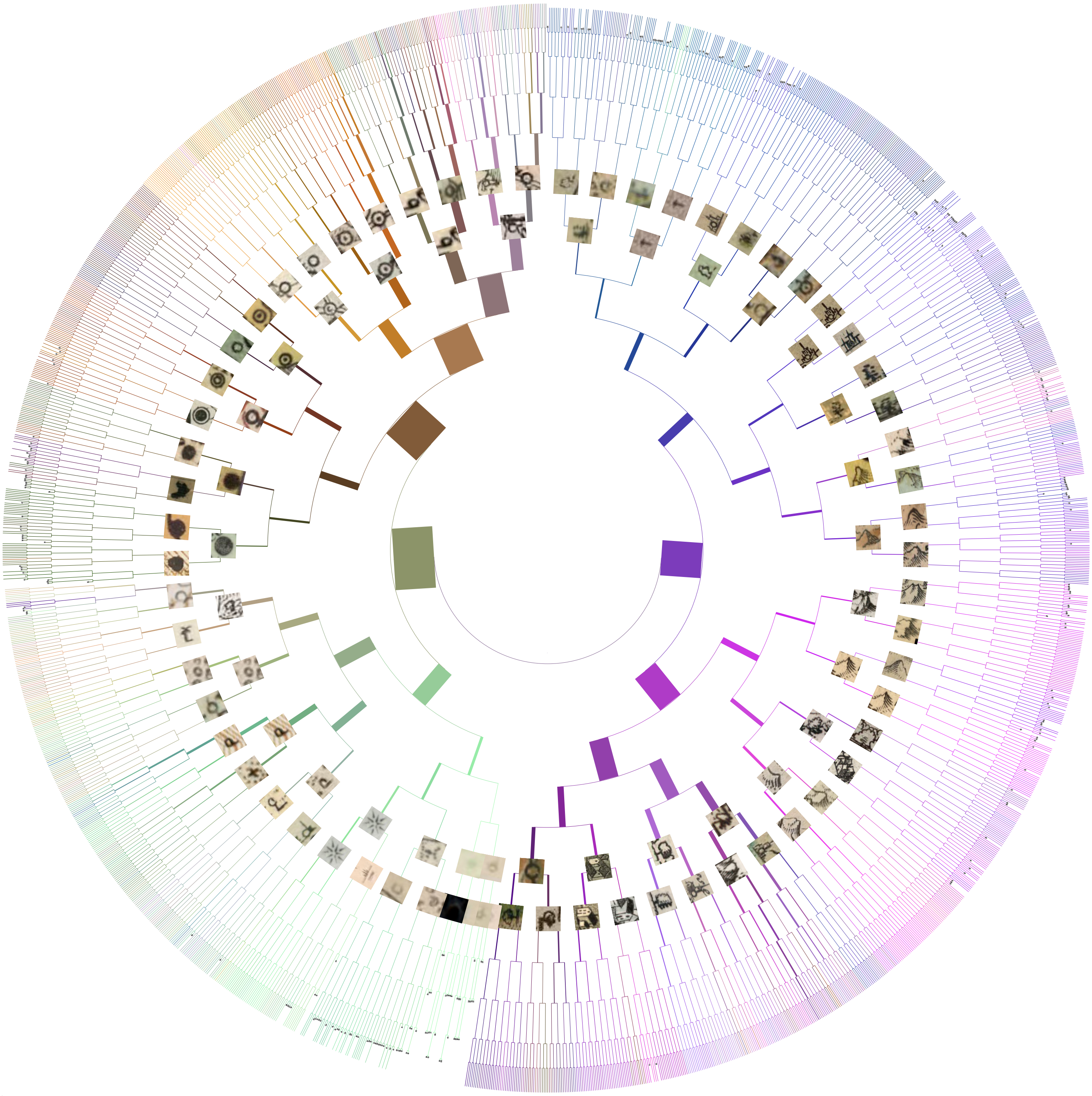

**Figure 16 | Circular dendrogram, or phylogenic tree of map signs.** Result of the hierarchical agglomerative clustering of the 16,384 sign cluster using Ward variance minimization method. The figure only displays the first 11 levels of the resulting tree, in which each leaf represents a single cluster. Exemplars are provided for hierarchical levels 5 and 6. The RGB color values of the leaves match those of the phylogenic mosaic (Fig. 14). They are obtained by applying UMAP dimensionality reduction to the cluster centers in a three-dimensional space. The color of each branch is computed as the average color of all downstream leaves. Branch width is proportional to the number of downstream leaves (i.e., exemplars), whereas branch length is proportional to the distances in the embedding space. *Sign clusters may be considered lineages and can themselves be grouped into a phylogenic tree.*

### *Diversification*

One of the main consequences of variation, is *diversification*. Like other cultural traits (Newson et al., 2007), map signs tend to *diversify* over time. Figure 17 reports the historical change in diversity, measured as the macro diversity, defined as the number of distinct sign clusters active per year; a cluster is considered active when at least three sign instances are assigned to this particular cluster in the year. Macro diversity indicates the overall diversity of cartography at a time, computed as the total number of simultaneously active sign clusters. Micro diversity, instead, reflects the average contribution of each map to the overall diversity; it is computed as the proportion of active sign clusters in a year, normalized by the number of maps published that year.

As expected, overall diversity increased consistently between the 1570s and 1900. A Mann–Kendall trend test confirms this pattern ($p_{val} < 1 \cdot 10^{-16}$, slope = 0.18%). The rise was particularly steady in the 17th century ($p_{val} < 2 \cdot 10^{-7}$, slope = 0.24%), and in the 19th century ($p_{val} < 1 \cdot 10^{-16}$, slope = 0.39%). By contrast, micro diversity soared between 1500 and 1650 ($p_{val} < 1 \cdot 10^{-16}$, slope = 0.012‰) but decreased again after 1775 ($p_{val} < 1 \cdot 10^{-16}$, slope = –0.013‰).

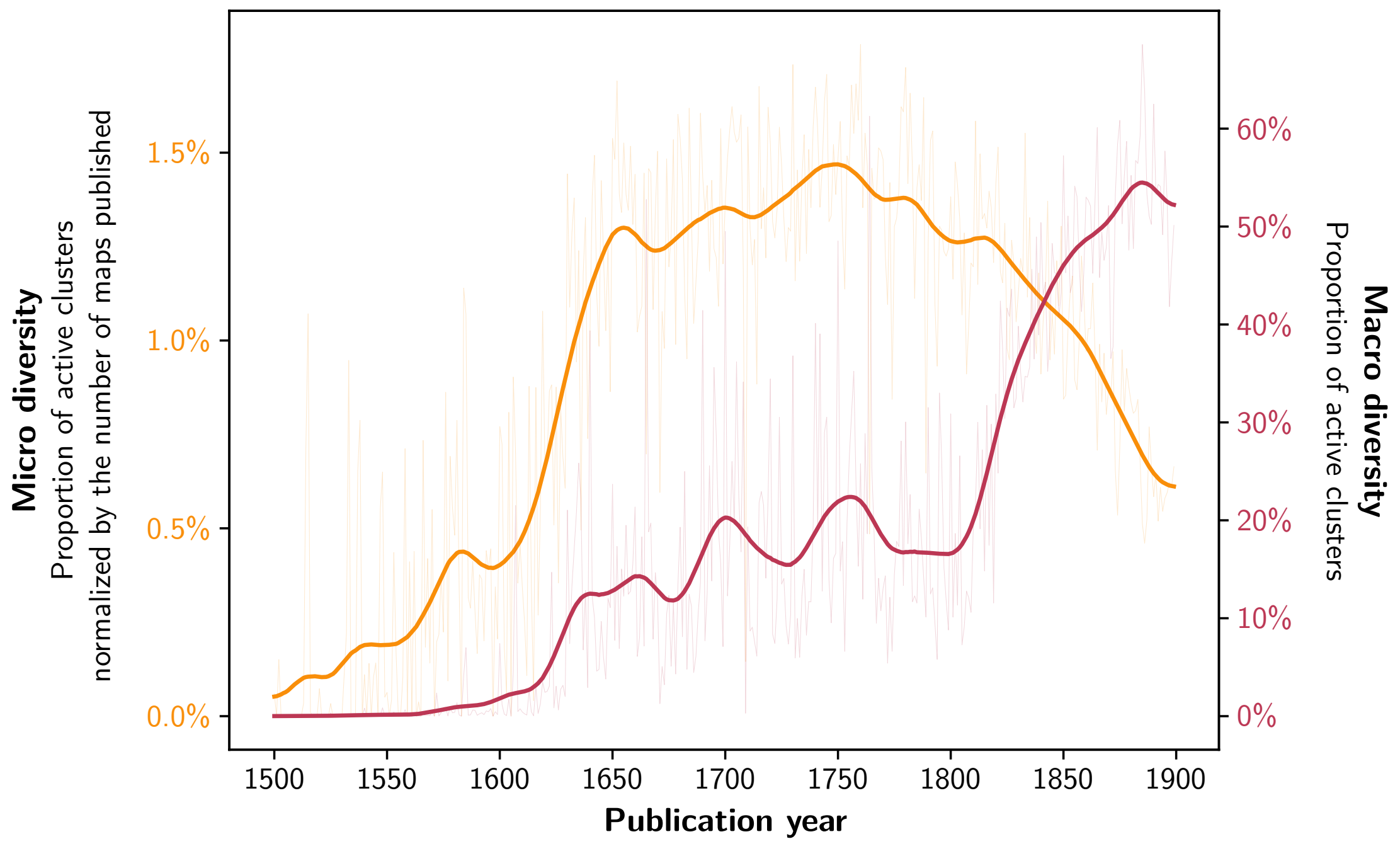

**Figure 17 | Change in sign diversity over time.** For each publication year, macro diversity is computed as the proportion of active cluster per year, while micro diversity is computed as the proportion normalized by the number of maps published. Both time series are smoothed with a low-pass Gaussian filter (σ = 8). The unfiltered time series are depicted in transparency. Years before 1500 and after 1900 are excluded due to data scarcity (before 1500) and the uncertainty in publication volumes caused by copyright restrictions (after 1900). *Overall sign diversity, i.e. macro diversity, increases over time, first owing to the increase of between-map diversity, i.e. micro diversity, (ca. 1600–1650) then, due to the increase in publication volumes (ca 1810–1880).*

Presumably, periods of diversification, such as the first half of the 17th century and the first half of the 19th century, played a key role in the emergence of modern semiotic systems in cartography. At the beginning of the 19th century, the number of maps published increased markedly, owing to the arrival of new printing technologies, notably lithography and cerography (wax engraving). These techniques facilitated the efficient replication of map signs. For instance, with wax engraving, signs could be applied directly to the wax plate using a stamp[11]. The industrialization of map production also contributed to a certain graphical monotony (Stooke, 2015). Globalization, the centralization of nation-states, and intensified cultural circulation are also possible explanations for the decline in micro-diversity. At the same time however, the increase in the number of maps published was such that macro-diversity increased, implying a raise in the total number of signs in circulation. In other words, while each individual map contributed less to the global diversity, the greater volume of maps resulted in a broader sign palette overall.

### *Simplification*

It would be a mistake to assume that adaptation entails complexification. Just as evolutionary processes can produce very simple life forms, cultural evolution does not necessarily lead to greater complexity (Lombard, 2016; Vaesen et al., 2016; Vaesen & Houkes, 2021). From a semiotic perspective, a simple heuristic to study the complexity of a document is to measure the number of distinct sign variants it contains. This idea is akin to the concept of type-token ratio, which uses the proportion of unique words in a text to measure its lexical diversity but also its complexity (Feldkamp et al., 2025; Kettunen, 2014). In this perspective, a map that employs a lexicon of only 2 or 3 distinct signs, like a map of the metro, would be considered semiotically less complex than a topographic map that relies on 40 or 50 different signs. Moreover, if these signs are reproduced with regularity, that is if they can be assigned to the same exemplar, the map would also arguably appear simpler than one in which signs that are supposed to look similar display substantial visual discrepancy.

Figure 18 depicts the trajectory of complexity, operationalized as the average count of distinct sign clusters per map. The trend before 1650 is not statistically significant, due to variation and limited data ($p_{val} = 0.09 > 0.05$). Hence, unlike diversification, one cannot assume that semiotic complexity increased in the late 16th or early 17th century. By contrast, the decline that begins in the 1730s is highly significant ($p_{val} < 1 \cdot 10^{-16}$, slope = –0.17). These results indicate that complexity, together with micro-diversity, diminished during the 18th and 19th centuries, leading in average to more streamlined and simpler, or at least more regular symbolic systems. This hypothesis will be further explored and exemplified in the next two sections.

[11] Note that the wax plate was then not directly used as printing plate. The engraving first had to be transferred to a more enduring metallic printing plate by electrotyping.

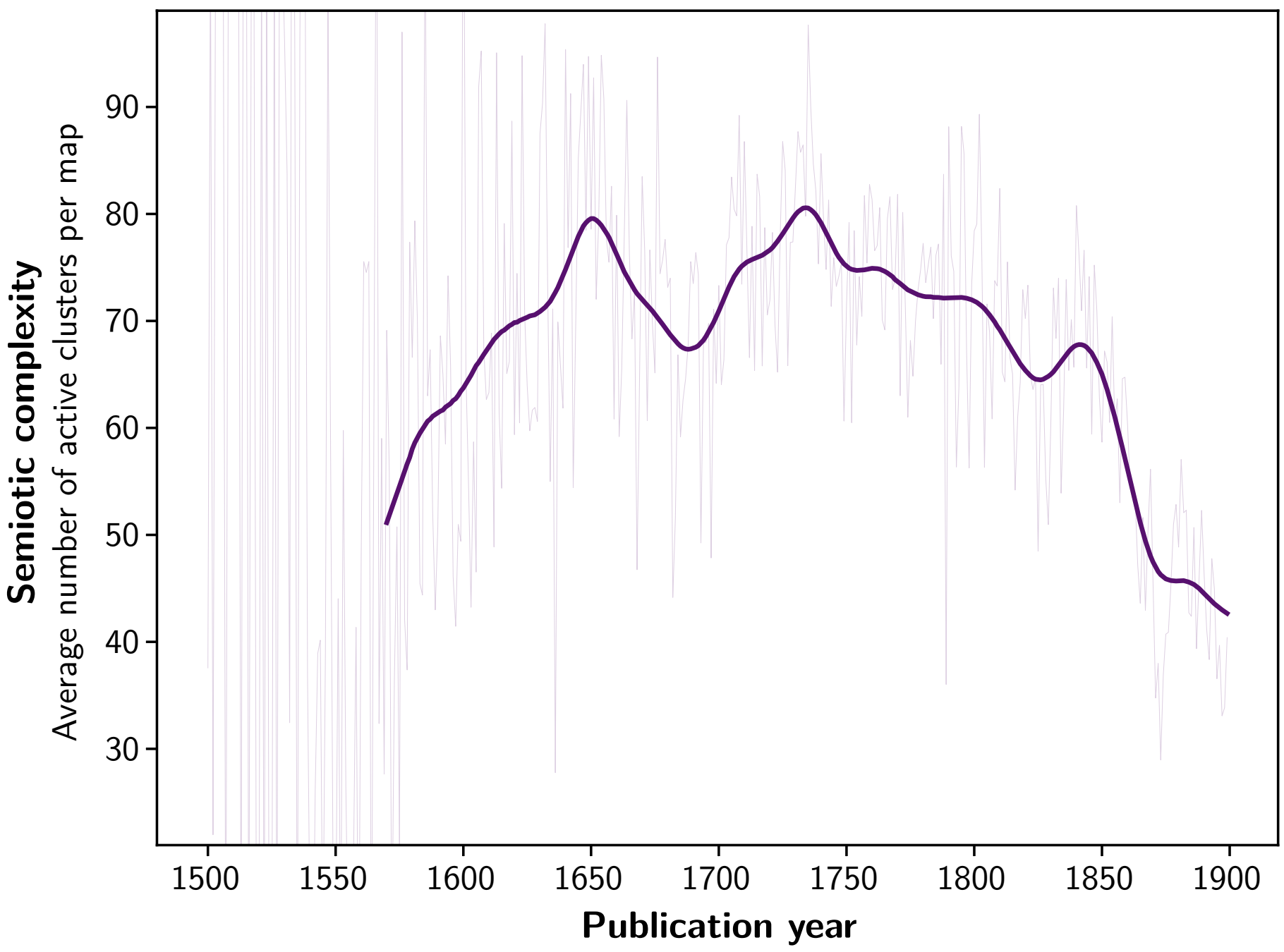

**Figure 18 | Change in semiotic complexity over time.** For each publication year, semiotic complexity is computed as the average number of active clusters per map. The resulting time series is smoothed with a low-pass Gaussian filter (σ = 8). The unfiltered time series is depicted in transparency. Before 1570, the average complexity value remains uncertain due to data scarcity. Years before 1500 and after 1900 are excluded to align with the temporal window of Fig. 17. *The complexity, calculated as the number of distinct signs per map, does not appear to increase over time.*

## 6.9 Coadaptation and sign complexes

### *Visualizing sign surges*

On average, each sign cluster is detected in 693 distinct maps (median = 305). These maps tend to be published within a similar period. Consequently, the distribution of sign clusters is modal: sign frequency typically surges for a few dozen years before regressing. This phenomenon is illustrated in Figures 19a and 19b. The figures display the frequency trajectory of 512 randomly drawn clusters. The time series are normalized and ordered chronologically by their peak frequency.

The quantitative examination of frequency trajectories indicates that sign variants tend to be distributed around a period of 30 to 100 years. Although the temporal dispersion fluctuates, most instances are concentrated around a single large mode. Smaller antecedent or subsequent surges could be construed as forerunners or sequels, though I suspect that they may more likely be manifestations of a noisy clustering. These minor peaks might also correspond to closely related variants. Except from these slight abnormalities, the distribution of sign variants within frequency trajectories exhibits archetypical qualities of diffusion processes. Figures B9–B12 in the Appendix, provide a sense of how the underlying variants usually appear. Specifically, they display 256 randomly selected sign clusters, ordered chronologically according to the same procedure applied to Figure 19.

Observation of the time series also highlights that surges are not evenly spaced in time. For example, several peaks occur around the same periods, ca. 1660, 1700, and the 1760s. The distribution appears more uniform after 1840, although additional modes are discernible around 1850 and 1890. This phenomenon suggests the existence of periods of sudden change, during which many new sign variants get incorporated in the canons of cartographic representation, akin to evolutionary sweeps. Moreover, it indicates the statistical association of distributions or, equivalently, the non-independence of cluster frequencies.

### *Definition: coadaptation & complex*

Because they form coherent semiotic systems, signs cannot evolve independently from one another. If they did, it would impede their ability to differentiate and distinguish the concepts they denote and, ultimately, to function as a semantic-symbolic system. Consider the example of a circle: in some maps, the circle represents a city, whereas in others it symbolizes a tree. However, there cannot be a map in which the same circle stands for both a tree and a city or that semiotic system would fail in its function to distinguish the two signified concepts. Instead, if a map depicted both cities and trees with circles, the two symbols would not be quite identical; their size or color, for instance, would differ sufficiently to prevent ambiguity.

Recognizing that signs constitute a system also entails that their evolution is interdependent. Any variation from a prior form must remain congruent with the other signs with which the sign appears. If the variation arises from a deliberate choice, the other signs must potentially be *adapted* to the new variant. This chapter does not yet address textures and color fills; these will be examined in Chapter 7. Nevertheless, I will rely on an example involving them to illustrate the point. Blue is routinely used to signify water. Consider, however, a map in which the mapmaker decided instead to color the streets blue. Such a choice would prevent the water itself from being represented in blue. This particular case is exemplified in Figure 20.

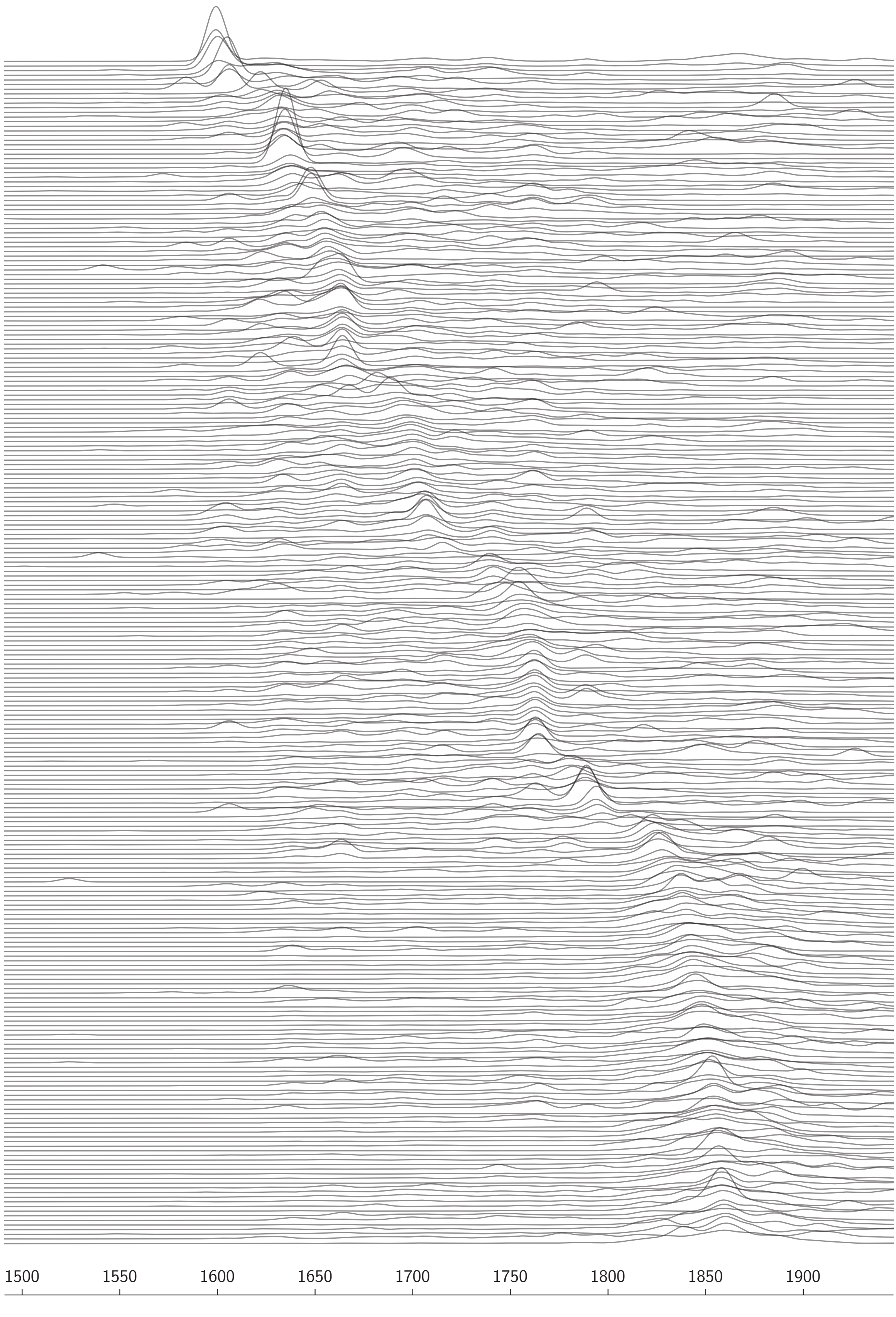

**Figure 19 | (a) Visualization of the frequency trajectory of 256 sign clusters.** Each time series, ordered by peak frequency, represents one cluster. The series are normalized and subsequently smoothed with a Gaussian filter ($\sigma = 5$).

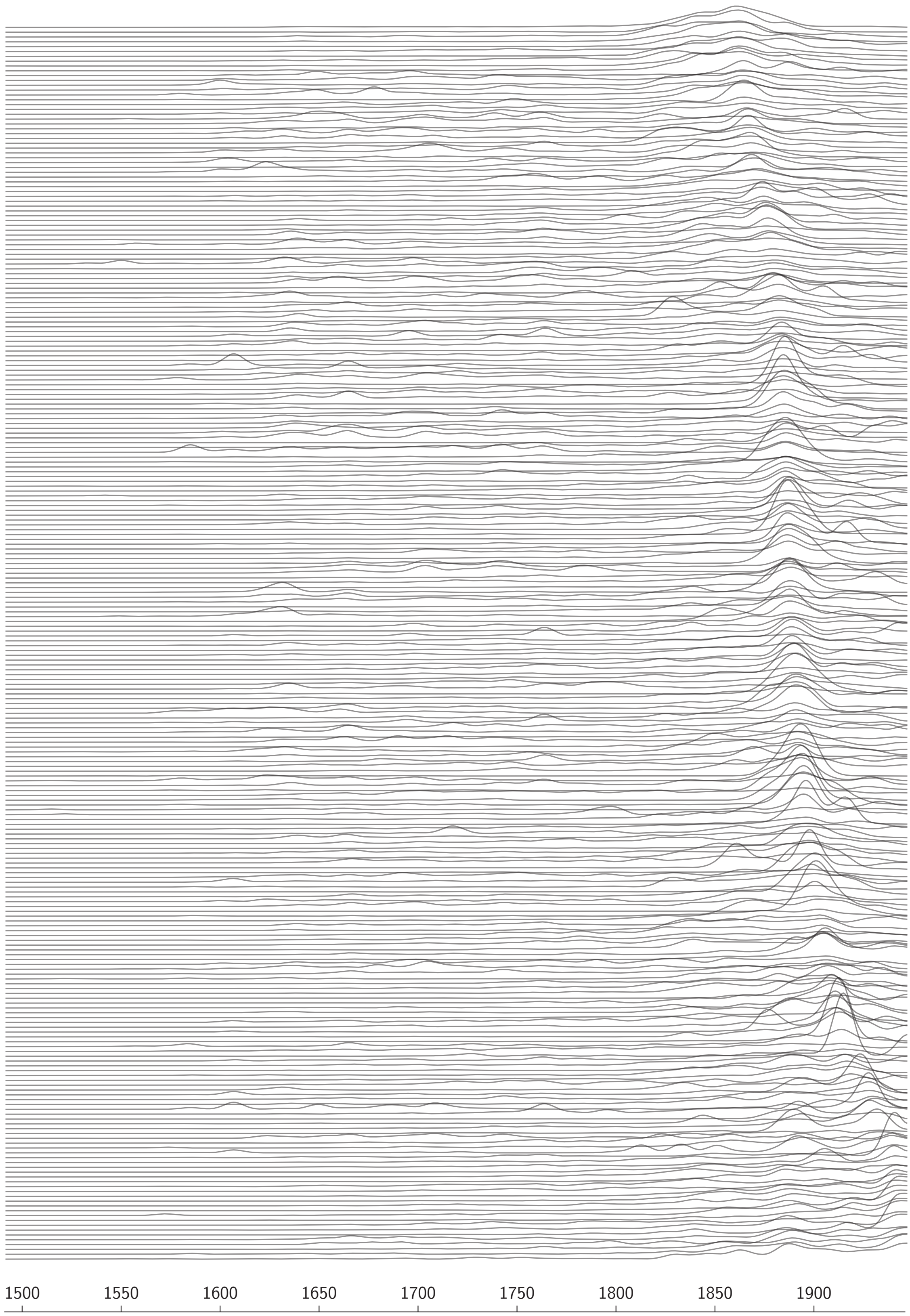

**Figure 19 | (b) Visualization of the frequency trajectory of 256 sign clusters (continued).** Each time series, ordered by peak frequency, represents one cluster. The series are normalized and smoothed with a Gaussian filter ($\sigma = 5$). *Signs grouped in the same cluster are distributed around a single temporal peak. Clusters replace one another through time.*

a) b)

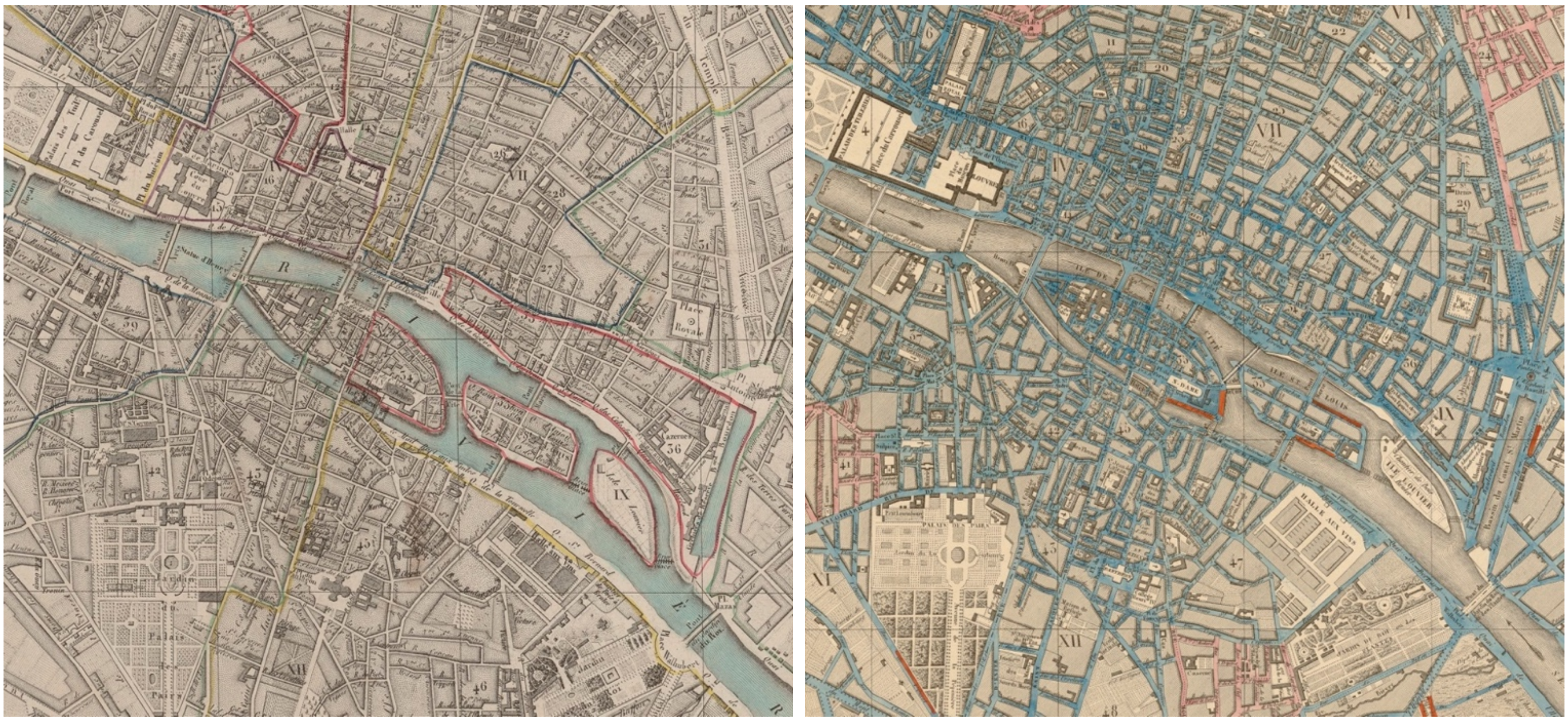

**Figure 20** | **The semiotics of blue. (a)** Map of Paris with the water signified in blue. **(b)** Map of Paris with transportation network signified in blue. Images source (a) L. Hennequin. *Plan de Paris*, 1829. Basset, Paris. 80 x 55 cm. BnF, GE C-3692. (b) Eustache Hérisson. *Plan de la ville de Paris*, 1834. Jean, Paris. 92 x 61 cm. BnF, GE C-2437. *Deviation from conventions of representation, as in b, may prompt the need for further figurative adjustments. It can also render the map harder to read.*

Coloring the road network in blue is presumably intentional; by default, mapmakers are more likely to *replicate* customary conventions of representation. Introducing variations that disrupt these conventions is costly because such changes often necessitate further adjustments. Reproducing existing cultural conventions, instead, is usually convenient and easy because it requires little thinking, or effort. It also simplifies communication, since readers are more likely to share the same cultural interpretation of the signs. Moreover, conventions of representation constitute stable semiotic systems that usually exhibit relatively few inconsistencies. As such, they can readily be replicated.

The relative rarity of inconsistencies, that causes conventional, or customary sign systems to be coherent implies that the signs that compose them are *coadapted*, i.e. they appear unequivocal with respect to one another. Coadapted signs tend to be reproduced together, as parts of balanced subsystems, forming what we may call *sign complexes*. A complex, in this definition, is a group of cultural traits that tends to be replicated jointly and whose frequency trajectories are, therefore, statistically associated.

## *Methodology employed for the identification of complexes*

To identify complexes, we first compute the presence matrix $X = (x_{i,j}) \in \{0,1\}^{N \times M}$ with $N$ rows, corresponding to the number of maps that contain at least one icon, and $M$ columns, one for each sign cluster. Each entry in the matrix is set to 1 if the map $i$ contains at least three instances of the sign cluster $j$. Let us write $A_j = \sum_{i=1}^{N} x_{i,j}$, the number of maps in which the sign cluster $j$ is present, and $B_{jk} = \sum_{i=1}^{N} x_{i,j} x_{i,k}$, the number of maps in which both $j$ and $k$ are present.

For each possible cluster pairs, we compute the contingency table $C_{jk}$ reporting the observed co-occurrences[12] between sign clusters $j$ and $k$:

$$C_{jk} = \begin{pmatrix} B_{jk} & A_j - B_{jk} \\ A_k - B_{jk} & N - (A_j + A_k - B_{jk}) \end{pmatrix}$$

Provided the denominator is non-zero, the conditional odds ratio for the pair is:

$$\hat{\vartheta}_{jk} = \frac{B_{jk}N - B_{jk}(A_j + A_k - B_{jk})}{\left(A_j - B_{jk}\right) \cdot \left(A_k - B_{jk}\right)}$$

Under the null hypothesis that *j and k are independent* versus the alternative that *the presence of k is more likely if j is also present*, the p-value is obtained from Fisher's exact test.

Then, we construct the graph $G \in \{0,1\}^{M \times M}$, in which the nodes $j$ and $k$ are connected by an edge if and only if $p_{val} < .01$. The complexes are retrieved by partitioning the graph $G$ into communities, using the Louvain method (Blondel et al., 2008).

## *Results & Discussion*

This approach yields eight distinct sign complexes. Figure 21 presents their distribution by publication year, map scale, publication country, and coverage country. Figure 22 displays representative exemplars of these complexes. Each complex combines a set of signs that can function together, as part of the same semiotic system. Accordingly, they comprise various icons and symbols with distinct significations that complement one another to produce a comprehensive semantic-symbolic system. The outcome, in this case, is largely influenced by the composition of the ADHOC Images corpus itself and its representativeness. The qualitative analysis of the exemplars primarily focuses on identifying the distinctive figurative characteristics of each complex.

---

[12] Note that the contingency matrix can also be written $C_{jk} = \begin{pmatrix} \{j=1; k=1\}_{\#} & \{j=1; k=0\}_{\#} \\ \{j=0; k=1\}_{\#} & \{j=0; k=0\}_{\#} \end{pmatrix}$, where $\{\cdot\}_{\#}$ is the count operator.

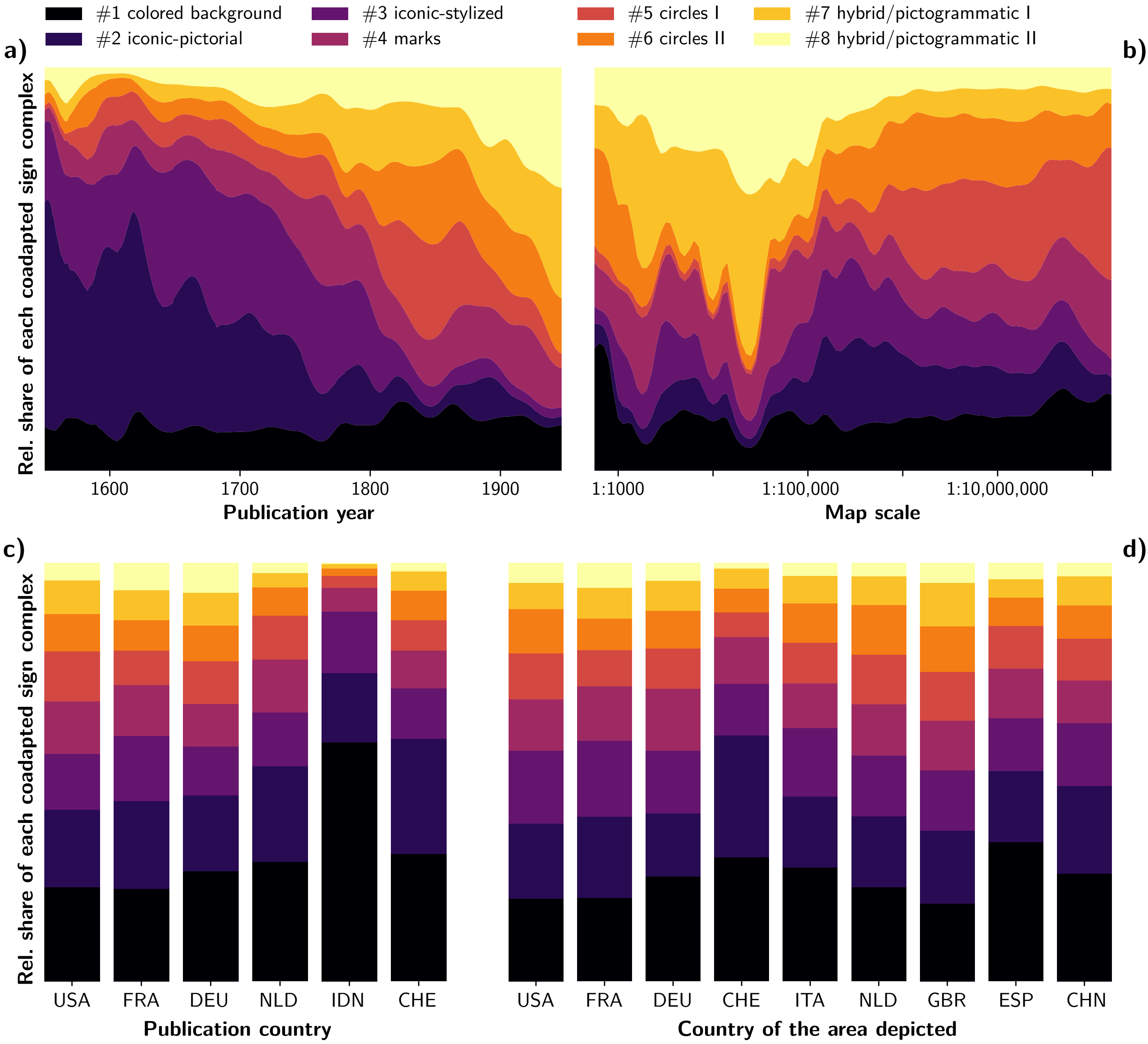

**Figure 21 | Relative share of each coadapted sign complex by publication year, map scale, publication country, and country of the area depicted.** Continuous relative distributions, like publication year and map scale, are smoothed with a low-pass Gaussian filter. *E.g. sign complex #2 is most common in maps before 1750, with scale < 1:100,000.*

For instance, the first complex (#1) may be designated the *colored background* complex, as it primarily comprises signs found in maps that incidentally implement visual surface differentiation through color. The frequency of this complex is remarkably stable (ca. 10–15% of the total) over the study period.

The second complex (#2) reflects a semantic system primarily based on icons displaying sustained iconicity. Its signs most closely resemble landscape iconography. The third complex (#3) is also dominated by icons. However, the latter tend to be more stylized. The shapes of hills are more geometric and angular, almost triangular, whereas trees and buildings, which generally denote settlements, are less detailed, reduced to the few essential strokes. Because these signs comprise fewer lines, their graphic load also seems reduced compared with that of the second complex. Complexes #2 and #3 together account for approximately 45–60% of the maps published up to the second half of the 18th century. Complex #2 encompasses nearly half of the documents produced from the 16th to the early 17th century, whereas #3 is proportionally more prevalent from the mid-17th to the mid-18th century. These two complexes correspond mainly to small-scale (< 1:100,000) or large-scale (1:2,000–1:10,000) maps.

Complex #4 comprises simple geometric marks, such as crosses (×+), dots, triangles, and circles. It exhibits a relatively stable distribution over time, corresponding to nearly 10% of published maps.

The fifth (#5) and sixth complexes (#6) both mainly comprise circles of varying types. The symbols appear more regularly drawn in #6, whereas the circles in #5 are more highly contrasted and slightly less regular. The discrepancy may potentially be attributable to differing printing techniques. For instance, the use of stamps can result in more regular shapes compared to engraving or hand drawing.

Together, the two complexes represent the relative majority (30–40%) of maps published from the beginning of the 19th century to the 1870s. Regular circles occur chiefly in plans drawn at scales smaller than 1:2,000 (e.g., fire insurance maps). In addition, both complexes form the most common sign systems in small-scale maps (< 1:500,000), where circles are typically employed to symbolize settlements.

Both the seventh (#7) and eighth (#8) complexes rely on a combination of pictograms and symbols. As such, they constitute the most hybrid semiotic coadapted systems, combining icons and symbols in roughly equal proportions. By pictograms, I refer to icons that are substantially stylized; for instance, trees reduced to a simple circle with a projected shadow, as observed in both complexes. Each set also contains color-printed icons, for instance blue wetlands in #7 or green trees in #8. In addition, both complexes comprise marks and simple geometric shapes, like crosses, dots, triangles, circles, and so on. The seventh complex emerges earlier and maintains a stable frequency, corresponding to 15–20% of the maps from the beginning of the 19th century to the end of the period under study. By contrast, the relative frequency of #8 triples between the middle of the 19th century—when it represents only 10% of map icons—until the end of the period, when it reaches 30%. The eighth complex seems to follow a harmonic distribution across map scales, with four marked peaks at scales 1:1,800, 1:5,000, 1:10,000, and 1:25,000. This pattern suggests that this complex is prevalent in map series, for example city atlases, rural land registers, or large-scale topographic maps. Together, the two complexes represent the majority of medium-scale maps.

#1 colored background

#2 iconic-pictorial

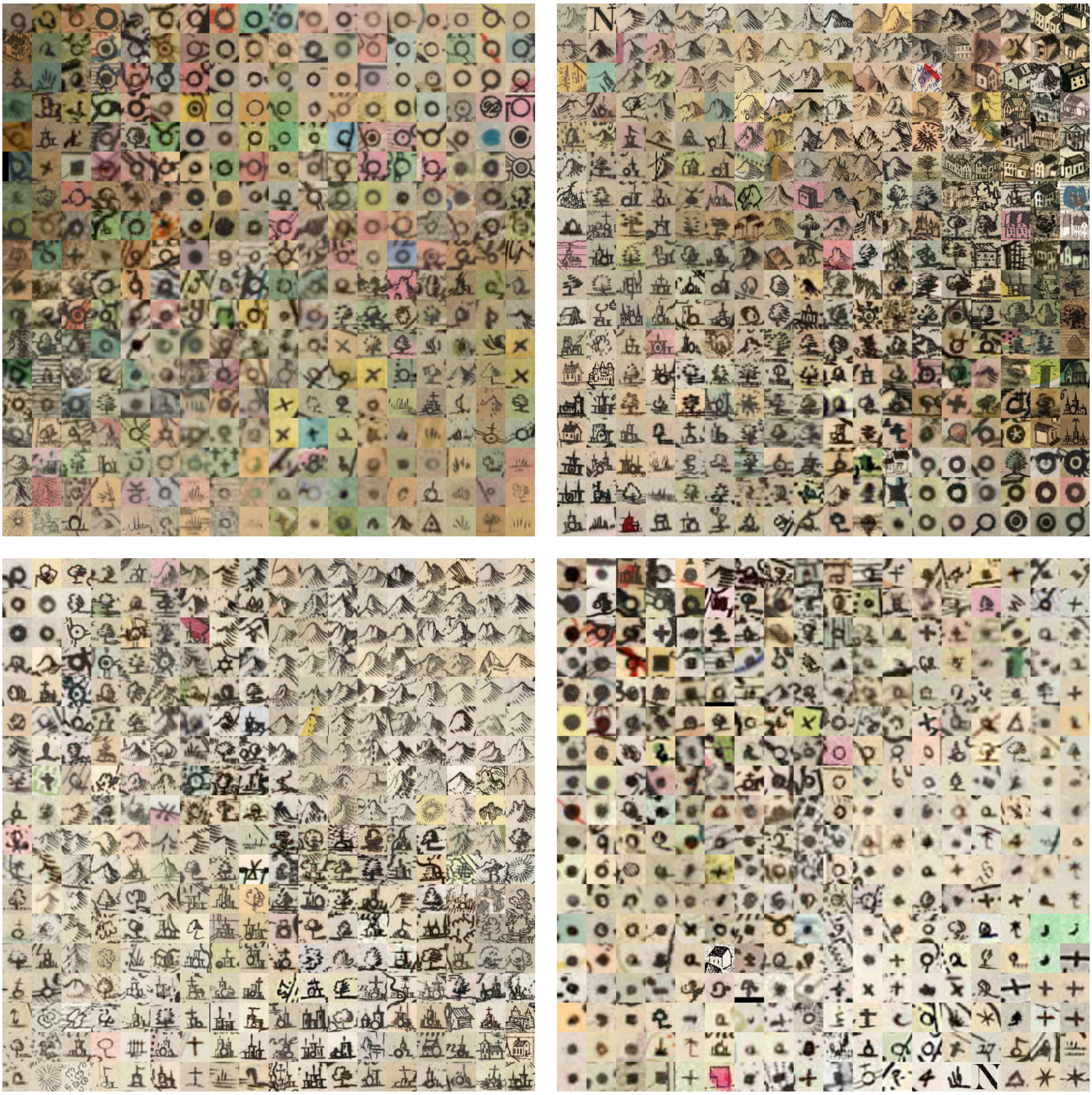

#3 iconic-stylized

#4 marks

**Figure 22 | (a) Exemplar signs from complexes #1 to #4.** Each block depicts a sample of 324 exemplars, representative of the coadapted complex. *E.g. sign complex #2 contains chiefly icons, such as hills, settlements, and trees, while complex #4 comprises primarily symbolic marks, such as crosses and dots.*

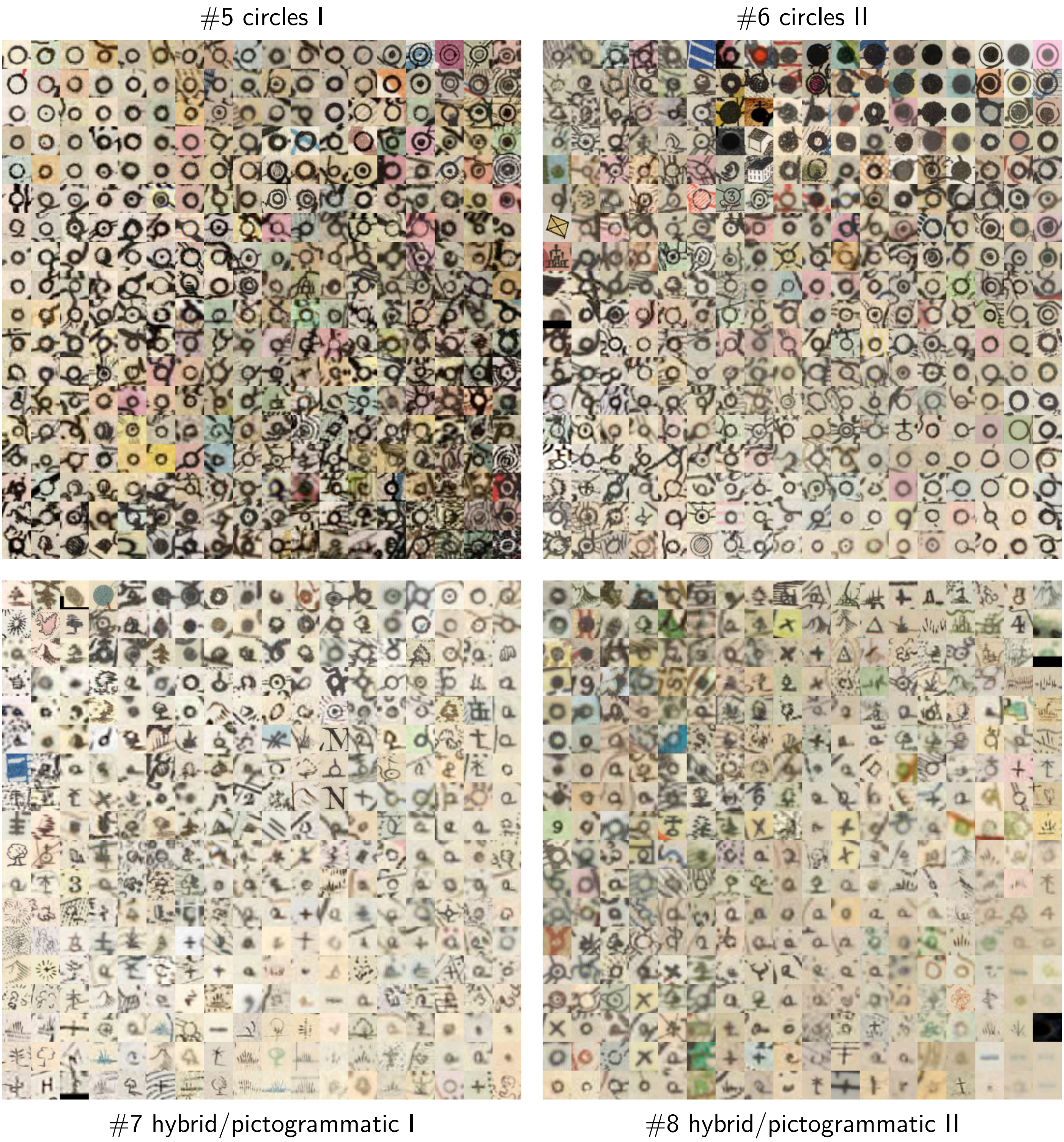

**Figure 22 | (b) Exemplar signs from complexes #5 to #8 (continued).** Each block depicts a sample of 324 exemplars, representative of the coadapted complex. *E.g. sign complexes #5 and #6 contain chiefly circles, while complexes #7 and #8 comprises a broader variety of pictograms, including different types of trees printed in pastel tones.*

Apart from the greater output of #1 colored-background maps in Indonesia, cross-national variations in the relative representation of map complexes appear minimal. Iconic-pictorial signs (#2) are slightly more prevalent in Swiss and Dutch cartography; in the Swiss case, this preference is also visible by the higher incidence of iconic-pictorial icons on maps depicting Swiss territories.

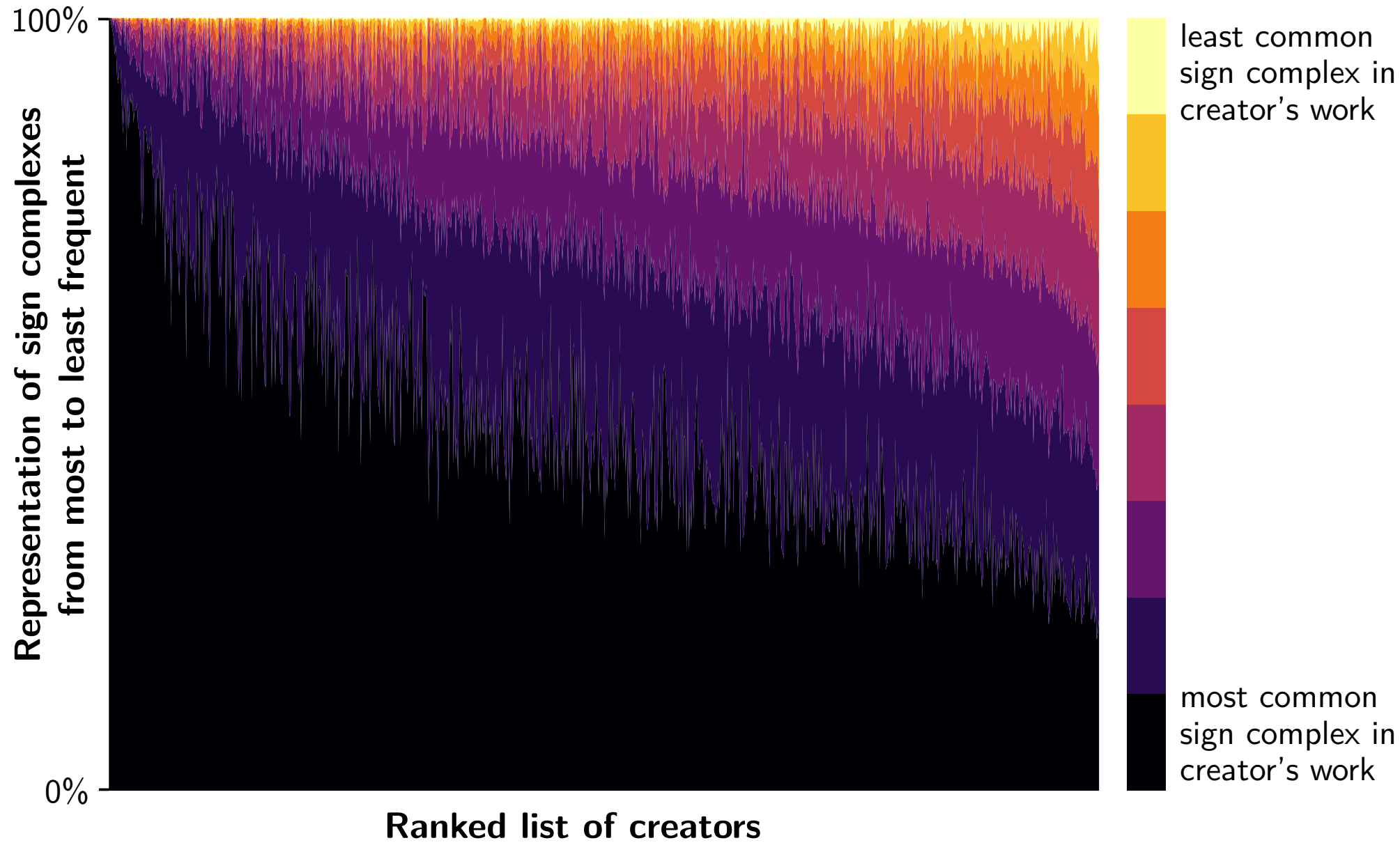

**Figure 23 | Ranked relative frequency of sign complexes employed by map creators.** The color gradient ranges from the most frequent complex (darker) to the least frequent complex (brighter), for each creator. The colors are not related to Fig 8; they indicate the relative frequency rank of each coadapted sign complex within the creator's work. *On average, about half of the signs produced by a creator are attributable to a single sign complex.*

Figure 23 shows that the variation between sign complexes within a creator's work is low. In this representation, the relative frequency of each sign complex is calculated and ranked, for each mapmaker, from the most frequent (darker) to the least frequent (brighter). Thus, on average, the most common complex accounts for 49% of all signs produced by a particular mapmaker, while the second complex constitutes an additional 21%. Thus, in spite of the misattributions due to the imperfection of sign representation, complexes constitute effective stylometric markers.

## 6.10 Identifying ruptures

The previous sections have highlighted the hierarchical structure of the embedding space, as well as nested structures of contingencies i.e., *complexes*. They have also quantified the diversity of cartographic signs. However, they did not yet discuss how the observed differences in sign frequencies may help identify distinct historical periods or cultural moments of cartographic figuration. This section attempts to characterize such *ruptures* more directly.

Here, ruptures are conceptualized as cultural breaks, translated into *structural changes in the occupation of the semiotic space*. Breaks distinguish discrete patterns and paradigms of representation. Here, the concept of rupture is not limited to its chronological dimension; spatial or geographical differences are likewise regarded as cultural ruptures. Additionally, moments of rupture may also be observed as a function of map scale.

Because the study of ruptures only makes sense insofar as the nature of the changes can be explained, this section pairs the statistical identification of ruptures with a more qualitative examination of shifts in representation. Accordingly, the following paragraphs demonstrate how time, map scale, and country of publication affect the representation of signs and illustrate these differences with *characteristic exemplars*.

Here, *characteristicity* reflects not merely the highest raw frequencies but the typicality of a sign cluster in a specific context, indicated by its *deviation from its average frequency*. Continuous variables, such as time and scale, are partitioned into several distinct or overlapping *strata*. Let $S$ denote the number of strata for continuous variables (or the number of categories for categorical variables such as publication country) and let $M$ denote the number of sign clusters. Then, let us define $C \in \mathbb{N}^{M \times S}$, the strata table, where $C_{m,s}$ represents the count of signs from cluster $m$ in stratum $s$. To mitigate the influence of extreme values, the strata table is first normalized and then saturated at the 95th percentile for each stratum, i.e.,

$$\widetilde{C_{m,s}} = min\left(\frac{C_{m,s}}{\mathrm{P}_s^{95}(C_{m,s})}, 1.0\right)$$

Then, we can write $\mu_s$ the mean of the normalized cluster frequencies within the stratum $s$. Ultimately, the index of characteristicity $\chi_{m,s}$ is calculated as the log-ratio transformation of the normalized cluster frequencies for each stratum:

$$\chi_{m,s} = \log\left(\frac{\widetilde{C_{m,s}}}{\mu_s}\right)$$

The coefficient of rupture, noted $\rho$, quantifies the difference between two strata as the mean absolute difference between their normalized frequency tables. A smoothed estimate of the rupture coefficient over time and across map scales is computed by employing sliding, overlapping, strata.

### *Historical ruptures*

Figure 24 reports the evolution of the coefficient of rupture $\rho$ over the study period. A relatively higher coefficient of rupture indicates a *contrast* between the signs used before that year and those that follow. Thus, local peaks denote periods of relatively rapid change compared to the years that precede or follow them. Statistical analysis highlights a major rupture, peaking between 1785 and 1792. Three secondary ruptures also emerge, around 1633, 1737, and 1876. A minor rupture is detected near 1895. In contrast, a period of stasis develops, reaching a minimum around 1849 before rebounding rapidly. The temporal granularity of the rupture coefficient depends directly on the density of available data for a given period. Consequently, the level of temporal detail is higher for the 19th century compared to earlier periods.

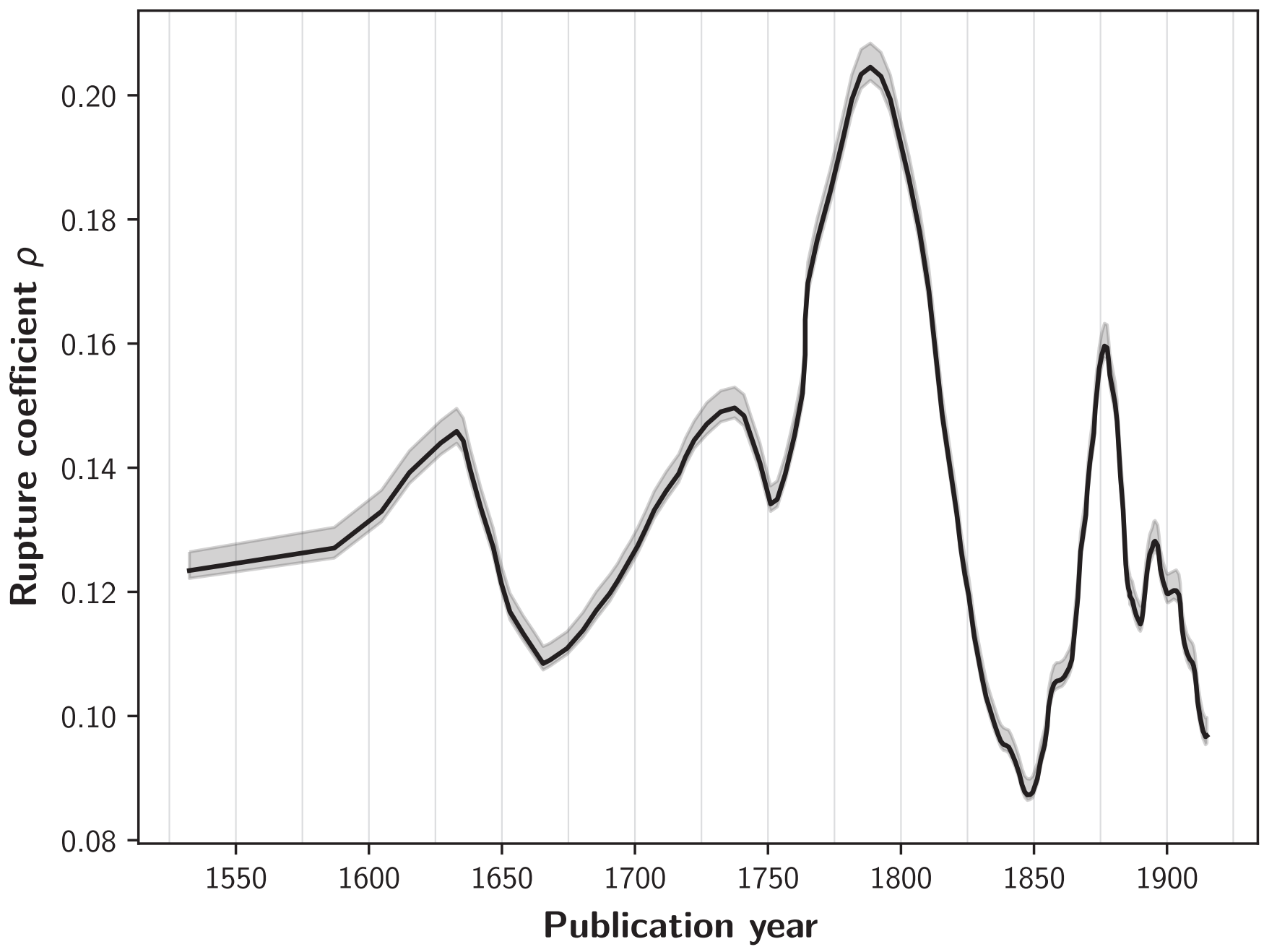

**Figure 24 | Evolution of the coefficient of rupture $\rho$ by year of publication**. The coefficient is computed using a sliding window (200 steps) and a pair of time strata, each corresponding to 5% of the dataset and overlapping by 50%. The shaded area denotes the 95% confidence interval of the mean. *The most marked shift in the signs used occurred around the year 1789.*

Identifying breakpoints offers a *periodization* of map signs, based on the identification of distinct patterns in the occupation of sign space. Most transitions are notably gradual and relative; periods of stasis alternate with periods of change. This dynamic is visible in Figure B17 in the Appendix, in which the occupation of the sign space, defined by uniform stratification of the study period, is represented.

By contrast, Figure 25 shows the relative occupancy of the sign space for each time stratum; here, time-strata extents are based on the periodization derived from Figure 24. For each stratum, characteristic exemplars are presented in Figure 26.

As visible from Figure 25, Quadrant 1, is strongly represented in the first two strata (1492–1633, 1634–1737). It corresponds to icons of hills and cities, coinciding with the lavender ■■ and the magenta ■■ regions of the phylogenic mosaic (Fig. 15), respectively. The most visible difference between the first two strata is the increased stylization of trees and settlements, visible in Figure 26. This process involves an increasingly bidimensional representation of objects. Conversely, before 1633, many icons depicting trees and buildings were represented in perspective, using shading, bringing maps closer to landscape iconography. From the mid-17th century onward, shading became lighter and more stylized, apart from hills which still retained a denser treatment.

The period 1738–1788 continued these changes. During this interval, hills also became sharper and more stylized. The first instances of terrain hachuring are also visible. This technique of depicting elevation became increasingly common throughout the 18th century; it was systematized by the German geologist Johann G. Lehmann in 1799 (Horn, 1981; Lehmann, 1799). As the morphology of icons was stylized and simplified, their size decreased as well[13].

The most significant temporal rupture occurs in 1789. Interestingly, this rupture also coincides with the French historical periodization, in which 1789 marks the transition from the Modern to the Contemporary Era. In the sign mosaic, this transition is characterized by a shift away from Q1 (Fig. 14) and by the greater representation of the orange region (■■■ Q4, cf. Fig. 15), corresponding to circle symbols, which indeed appear characteristic of this period (Fig. 25). They epitomize the process of settlement symbolization, whereby the former town and city icons nearly disappear and are replaced by more standard circle symbols.

The time stratum 1877–1895 is less clearly delineated by ruptures. It is characterized by the activation of the purple region (■■ Q1, cf. Fig. 15), which mainly corresponds to iconographic perspective maps, like O. H. Bailey's View of Thomaston, published in 1879 and discussed in Chapter 4 (Fig. 11). Another prominent pattern, shared with the sixth and final time stratum (1896–1947) is the increased representation of the mint-green region (■■ Q3), which corresponds to varied tree pictograms. Marks, such as simple crosses and full circles, are also observed. This period is further characterized by highly stylized depictions of trees, sometimes reduced to simple circles with cast shadows, green circles, or five-pointed stars for palms. At last, the development of color printing is evident, especially in the representation of vegetation.

[13] This reduction in sign size implies that each sign occupies a smaller area of the image and is therefore composed of fewer pixels. This diminution is visible in the representative exemplars shown in Fig. 26 and does not seem to be caused by corpus effects, e.g., a decrease in digitization quality for the maps published during this period.

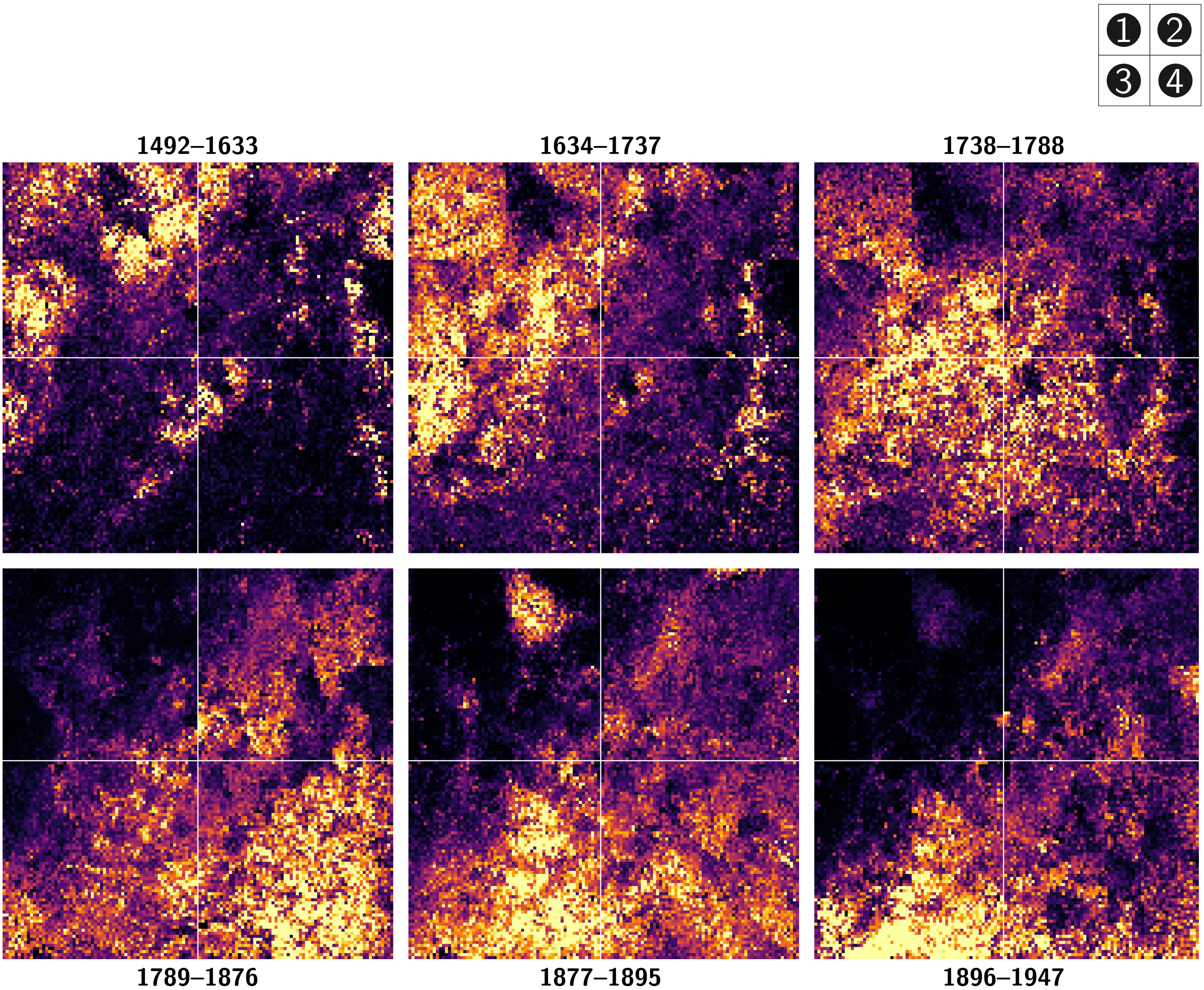

**Figure 25 | Evolution of the relative distribution of map signs by time stratum.** Each block depicts the relative frequency of sign clusters for a specific time stratum, based on publication year. Sign clusters are spatialized as in the sign mosaic (Fig. 14). Brighter areas correspond to higher frequencies. Frequencies are normalized, with saturation at the 95$^{th}$ percentile of each stratum. The graticule in the upper right corner indicates the four quadrants reported in Fig. 14, and enlarged in Figs. B13–B16, in the Appendix. *The occupancy of the sign space changes progressively and consistently over time; each temporal rupture reflects a shift in the signs used.*

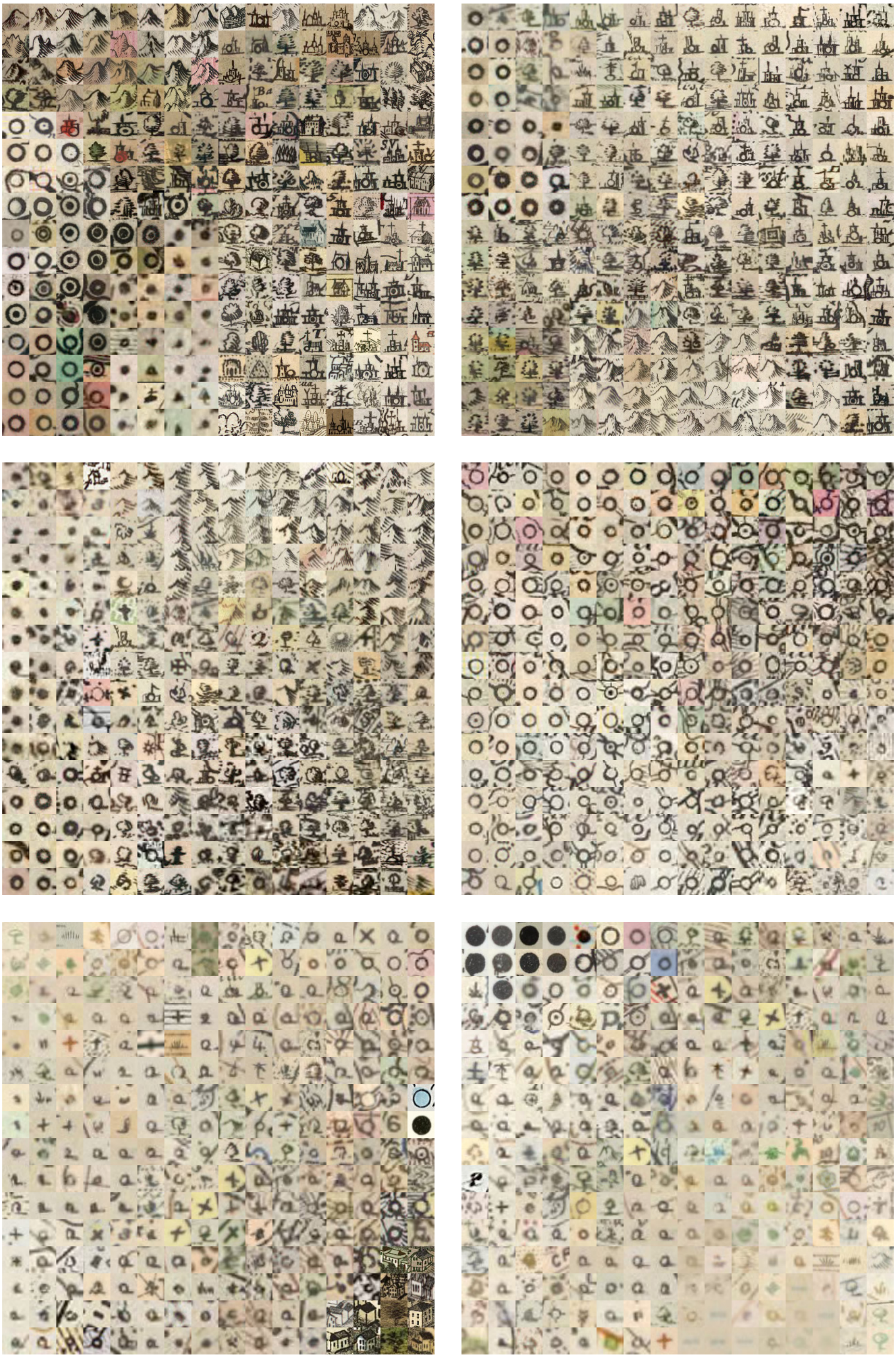

**Figure 26 | Exemplars of characteristic signs for each time stratum.** Each block depicts the 256 most characteristic exemplars, accordingly to the time strata from Fig. 25. *Before 1633, signs are chiefly iconic. Circles dominate in the period 1789–1876, whereas pictograms and color-printed signs prevail from 1877 onward.*

### *How signs differentiate map scales*

Marked differences in sign representation can also be observed across map scales (Fig. 27). The most pronounced rupture occurs between 1:82,300 and 1:101,000; the coefficient $\rho$ peaks around the scale 1:93,200. A secondary rupture appears at the scale 1:4,750 [1:7,550–1:3,650]. Finally, a third, minor rupture is observed around the scale 1:671,000 [1:781,000–1:550,000].

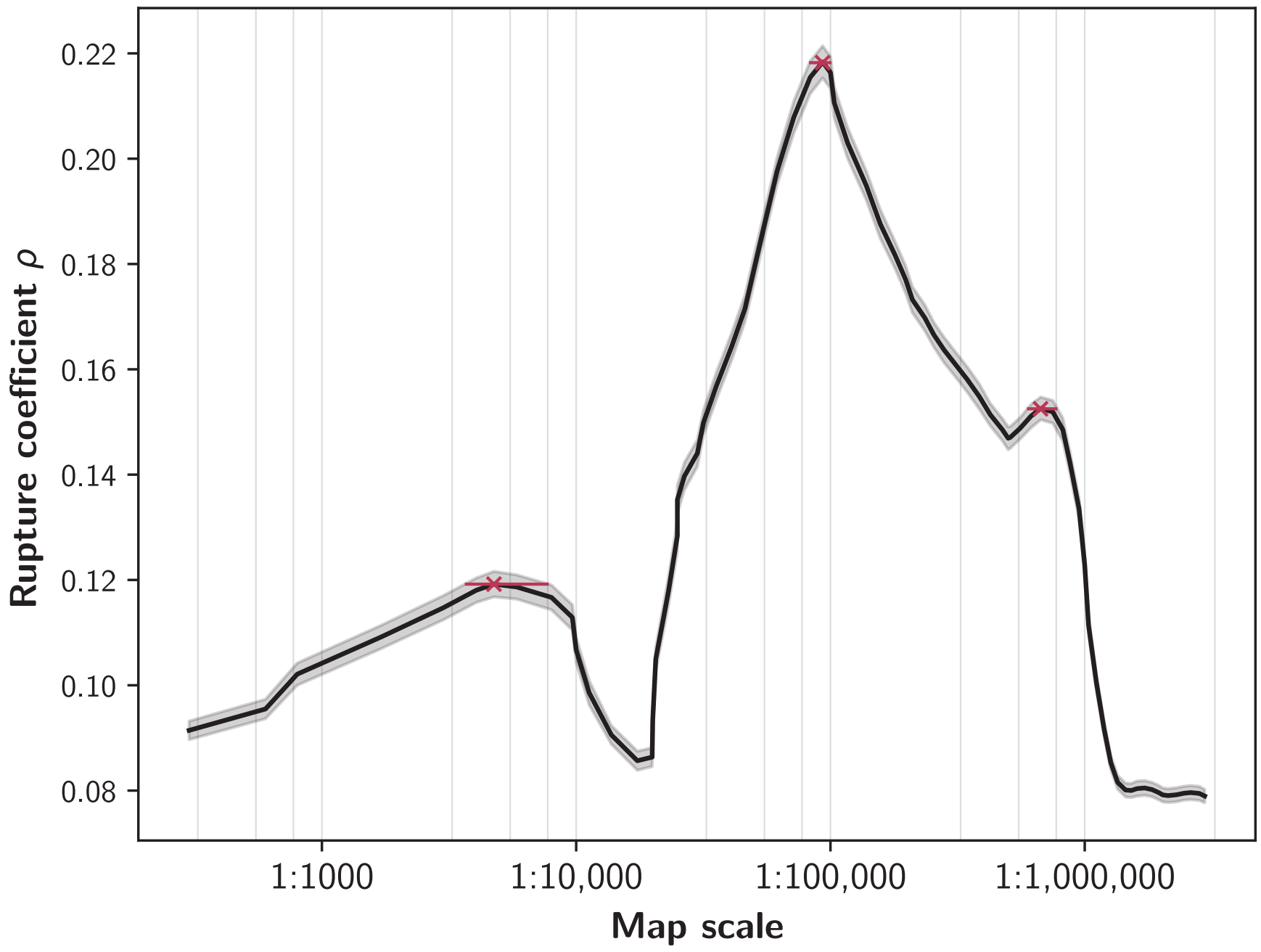

**Figure 27 | Coefficient of rupture $\rho$ by map scale**. The coefficient is computed using a sliding window (200 steps) and a pair of time strata, each corresponding to 10% of the dataset and overlapping by 50%. The shaded area denotes the 95% confidence interval of the mean. Red crosses and horizontal bars mark the position of the local maxima within this interval. *The most marked shift in the signs used occurs around the scale 1:93,200.*

Figure 28 shows the changes in the relative frequency of map signs by scale stratum. Two distinct distribution patterns emerge, characteristic of medium map scales (1:4,750–1:93,200) and small map scales (< 1:671,000), respectively, reflecting the higher rupture coefficient around the scale 1:93,200. Maps whose scale falls between 1:93,200 and 1:671,000 seem to exhibit features from both groups, along with their own distinctive features. The distribution pattern for large-scale maps (> 1:4,750) resembles that of medium-scale maps (1:4,750–1:93,200), while also containing additional characteristic areas—such as the olive-green region (Fig. 15, ■■ Q2).

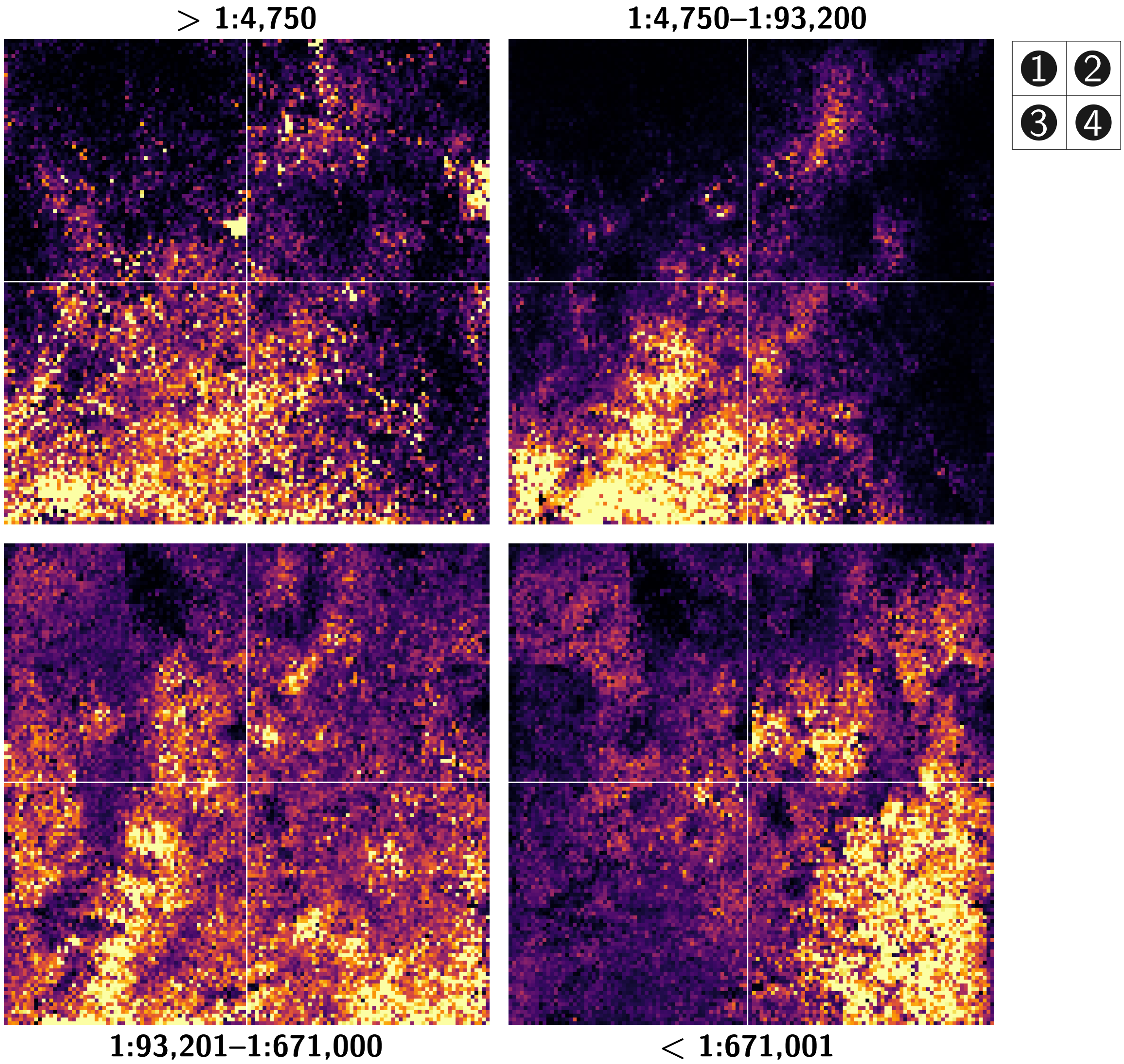

**Figure 28 | Relative distribution of map signs by scale stratum.** Each block depicts the relative frequency of sign clusters within a specific map scale stratum. Sign clusters are spatialized as depicted in the sign mosaic (Fig. 14). Brighter areas correspond to higher frequencies. Frequencies are normalized with saturation at the 95th percentile of each stratum. The graticule in the upper right corner indicates the four quadrants reported in Fig. 14 and enlarged in Figs. B13–B16, in Appendix. *Distinct signs are used as a function of map scale. Signs from the 3rd quadrant—corresponding to diverse icons—prevail in larger-scale maps, whereas the 4th quadrant–—containing mainly circles—seems more represented in smaller-scale maps, cf. Figs. B15, B16 respectively.*

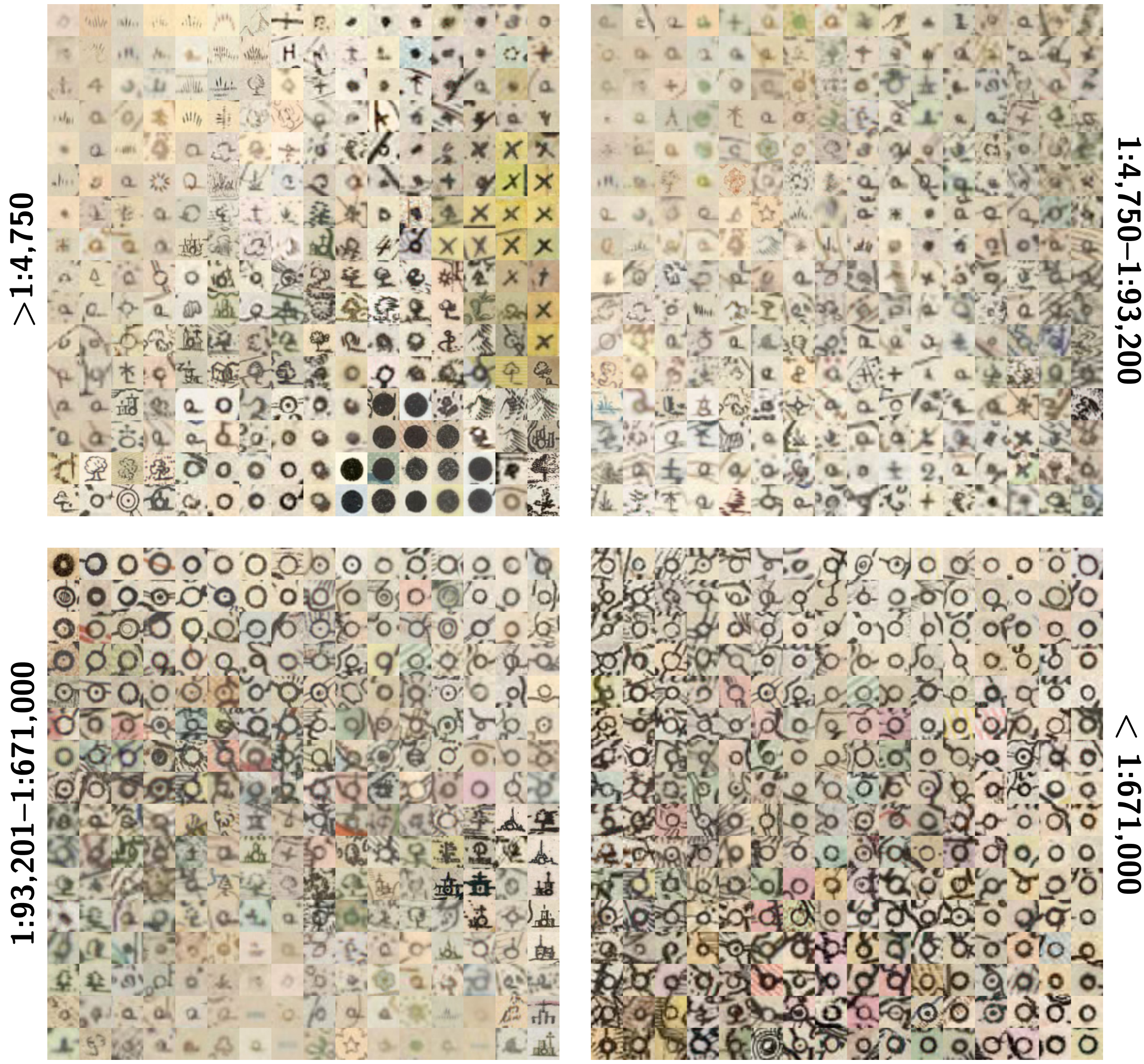

**Figure 29 | Exemplars of characteristic icons for each scale stratum.** Each block depicts the 256 most characteristic exemplars, according to the scale strata from Fig. 28. *Marks like crosses, and grass icons are more characteristic of larger-scale maps, while circle dominate smaller-scale maps. Tree and settlement icons are still used on rather small-scale maps (up to the stratum 1:93,201–1:671,000).*

An examination of characteristic map-scale exemplars (Fig. 29) shows that the first three strata (> 1:671,000) are characterized by tree icons (e.g., pine trees, palm trees) and other vegetation (e.g., bushes, grass, marshes, cultures). A few buildings, such as dwellings, religious monuments (e.g., chapels, crosses, graves), and mills, are also visible. Distinctive black circles and crosses derived from Sanborn fire insurance maps appear in the first stratum (> 1:4,750); they correspond to the olive-green region in Figure 15 (■■ Q2) and to a small denim-blue ■■ region located in the first quadrant (see Fig. B20). Full circles symbolize hydrants, whereas crosses indicate frame structures, such as wooden walls (Stoner, n.d.). Some exemplars of hills are mistakenly attributed to the characteristic exemplars of the larger map-scale stratum (Fig. 29), likely because of the relatively low representation of that stratum in the data, or due to the imperfection of the space of representation. Medium- and large-scale (> 1:93,200) map signs, particularly those depicting vegetation, tend to be rendered in pastel color tones and display a lower graphic map load.

Maps with scales between 1:93,200 and 1:671,000 exhibit signs characteristic of both smaller- and larger-scale maps. For instance, one can observe trees among the characteristic exemplars of that scale, as well as bushes, marshes, crosses, graves, and mills. Circles and other signs symbolizing settlements are also visible. Settlement icons (lavender ■■ Q1, Q3), tend to be underrepresented in Figure 29. Nevertheless, the relative distribution indicates that these icons occur most frequently on maps with scales between 1:93,200 and 1:671,000, which historically correspond, for instance, to chorographic maps.

The last stratum, corresponding to map scales smaller than 1:671,000, consists chiefly of circles and dotted circles that symbolize settlements of varying importance, often connected by roads. Larger circles and inner dots generally denote larger cities. At the smallest scales, islands are iconized as circles surrounded by several additional circles (cf. Fig. 29). Figure 28 further suggests that hills, corresponding to the magenta ■■ region (Q1), are considerably more common in small-scale maps.

### *National trends*

Whereas the temporal and scalar dimensions are continuous variables that can be divided based on moments of rupture, the ADHOC Images dataset encodes space as a categorical variable. Chapter 2 suggested that both contemporary borders and urban nuclei constitute meaningful divisions regarding social transmission. This section first considers countries, which will provide a broad overview of geographic cultural differences. It will also use the index of characterity to identify characteristic *creators*. Then, a more distant and quantitative approach, focusing on cities, will be adopted to study the finer interplay between sign similarity and geographic-cultural factors. Figure 30 presents the relative frequency of map signs by country of publication. Figure 31 displays characteristic exemplars for each country. Characteristic exemplars from the United States include Sanborn fire insurance maps and iconographic perspective maps, as anticipated. These exemplars also feature settlement symbols, typically rendered as circles, occasionally printed on a colored background. These signs are typical of American atlases, like those created by Robert DeSilver (active 1822–1827), Henry S. Tanner (1831–1918), Samuel A. Mitchell (1790–1868), Sidney E. Morse (1794–1871), or Fielding Lucas (1781–1854). Analogous symbols also appear on road and railroad maps, such as those issued by J. H. Colton (1800–1893) or Rand McNally Co. (est. 1856).

Figure 30 shows that French maps employ a wide variety of cartographic signs. Alternatively, this pattern may be interpreted as evidence of cultural intermediacy among the other countries under study. Only the purple region (■■ Q1, cf. Fig. 15) and the second quadrant exhibit low frequencies—these two regions are foremost characteristic of American cartography. The French diversity is reflected in the exemplar signs (Fig. 31), which contain varied marks, circles, and icons (e.g., mountains, trees, settlements). This heterogeneity probably also reflects the relative stability of French map publication over the long term (see Chapter 2).

German signs seem to be characterized by the abundant use of pictograms typical of topographical or military cartography produced by the Generalstab (*General Staff*) or the Reichsamt für Landesaufnahme (*Imperial Office for Land Survey*). Icons representing trees and vegetation were also frequently color-printed in green or bistre tones. Map makers or publishers, including Homann Heirs (1730–1848), but also Heinrich Kiepert (1818–1899), Friedrich W. Streit (1772–1839), and Joseph Meyer (1796–1856), seem emblematic of the late 18$^{th}$ and 19$^{th}$ century German cartography, that typically included small-scale atlases with settlements symbolized as circles.

Like their German counterparts, Swiss maps appear to be characterized by pictogrammatic signs typical of topographic mapping. Emblematic examples include the national map series (1845–1865) prepared under Guillaume H. Dufour (1787–1875) and its forerunners, the Atlas de la Suisse (1813) by Heinrich Keller (1778–1862) and Samuel J. Scheuermann (1770–1844), as well as the Atlas Suisse (ca. 1800) by Johann H. Weiss (1758–1826). A particularly *avant* example is the 1:25,000 map of the canton of Zürich, published 1852–1865 (see Fig. B18 in the Appendix).

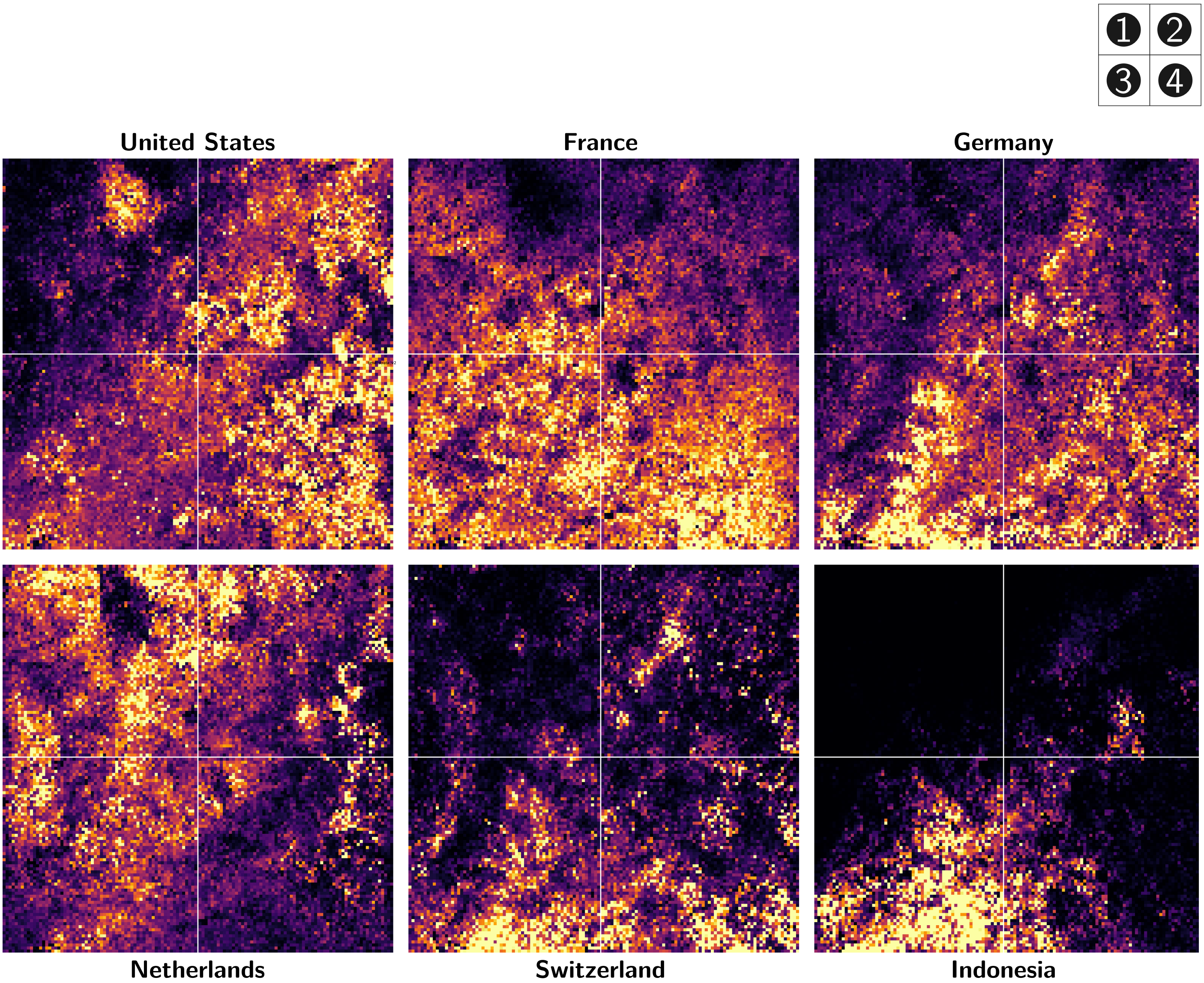

**Figure 30 | Relative distribution of map signs by publication country.** Each block depicts the relative frequency of sign clusters for a specific publication country. Sign clusters are spatialized as depicted in the sign mosaic (Fig. 14). Brighter areas correspond to higher frequencies. Frequencies are normalized with saturation at the 95th percentile of each stratum. The graticule in the upper right corner indicates the four quadrants reported in Fig. 14, and enlarged in Figs. B13–B16, in the Appendix. *National mapping practices tend to correspond to distinct—though overlapping—regions of the sign space.*

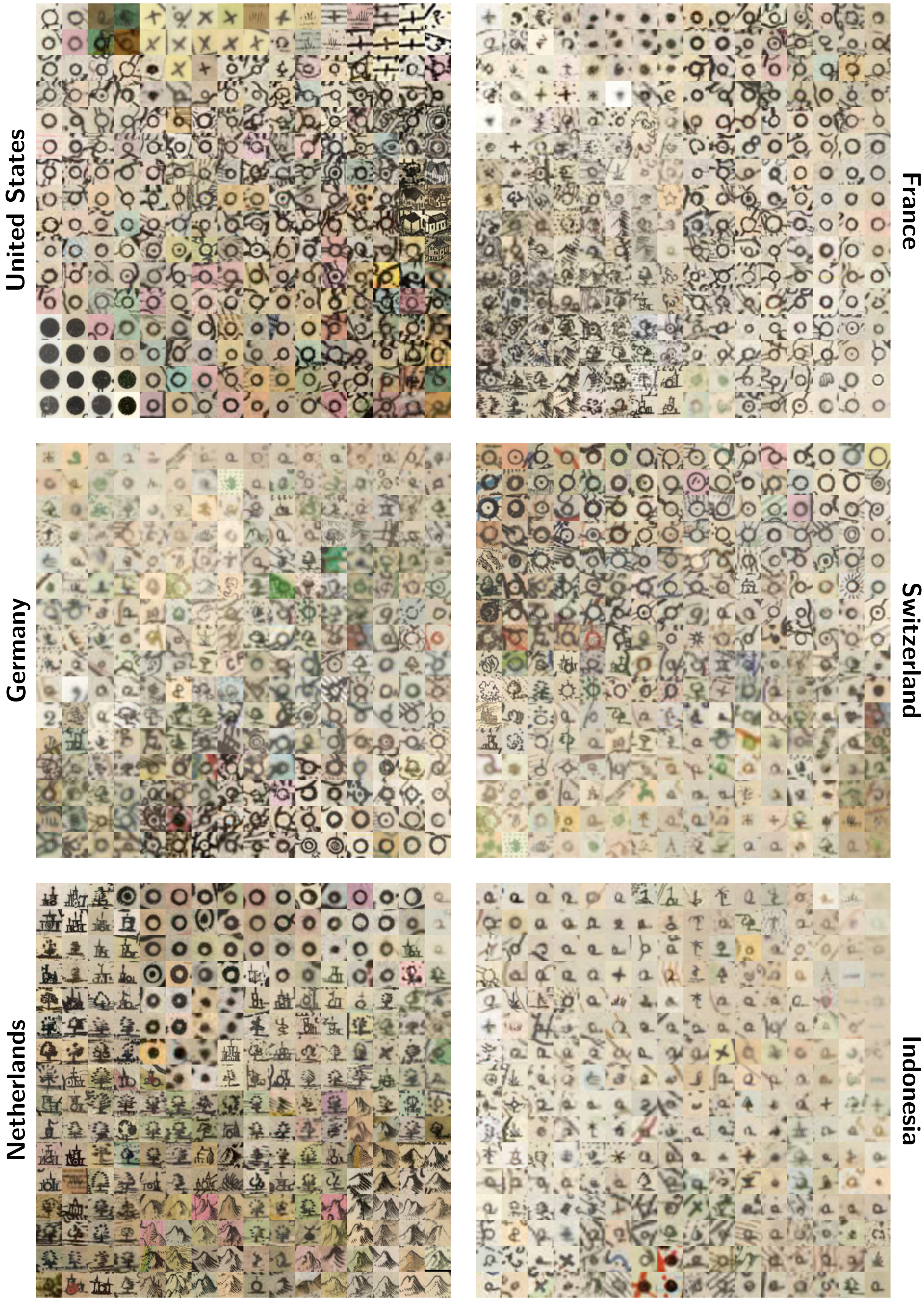

**Figure 31 | Exemplars of characteristic icons for each publication country.** Each block depicts the 256 most characteristic exemplars for each publication country (see Fig. 30). *E.g. full circles, cross marks, detached house iconographies, and circles on chromatic background are characteristic of American cartography.*

Dutch maps stand out by the prominence of icons representing trees, settlements, hills, vines, and marshes. Thick circle symbols are also common. These signs are characteristic of the Golden Age of Dutch cartography and are associated with creators such as Joan Blaeu (1596–1673), Pieter van der Aa (1659–1733), Frederick de Wit (1629–1706), Nicolas Visscher (1618–1709), Cornelius Cruys (1655–1727), and Gerard Valck (1652–1726). The relative distribution of these cartographic signs also seems to parallel the 1634–1737 temporal stratum, a finding that is not surprising provided that most Dutch maps were published during this period (cf. Chapter 2). By contrast, in the other national cases, older signs are largely outnumbered by comparatively recent pictograms or symbols.

More recent examples of "Dutch" signs appear in the maps published in Indonesia. Although these maps were usually published in Batavia (current Jakarta), the most characteristic publishers were Dutch colonial offices, such as the Topographisch Bureau. Figure 31 indicates that the most characteristic signs were stylized icons. Common motifs included trees denoted by a circle with projected shadow (like a reverted $\sigma$), palm trees signified as stars with six radial branches or as a vertical bar topped with a five-branched star. Stylized versions of grass, pine tree, and vine icons are also visible.

### *Cultural–geographic landscape*

While the comparison of map signs by country of publication reveals some differences, these distinctions remain rather broad. They do not account for subtle cultural specificities within countries, nor for the specificities of production centers and cartographic workshops. To enable a more granular analysis, we compute $\Gamma_\rho$, the geographic rupture matrix. In this matrix, the value of each cell $\Gamma_\rho^{a,b}$ is computed as the rupture coefficient $\rho$ between the two normalized frequency tables corresponding to the sets of maps published in location $a$, respectively location $b$. Put plainly, the coefficient $\Gamma_\rho^{a,b}$ reflects the distance, or dissimilarity, between the two sign sets. Although the matrix is indexed by geographic entities, its values are derived exclusively from semiotic distances, not from geodesic distances. We retain as location indexes all geographic entities (cities, provinces, or countries) corresponding to at least 125 map records in ADHOC Images, that is, 80 locations in total. An excerpt of the geographic rupture matrix, corresponding to the 35 principal urban publication centers, is reported in Figure B19, in the Appendix. Ultimately, the geographic rupture matrix is projected to two dimensions, using UMAP, for visualization. The result of the dimensionality reduction is depicted in Figure 32.

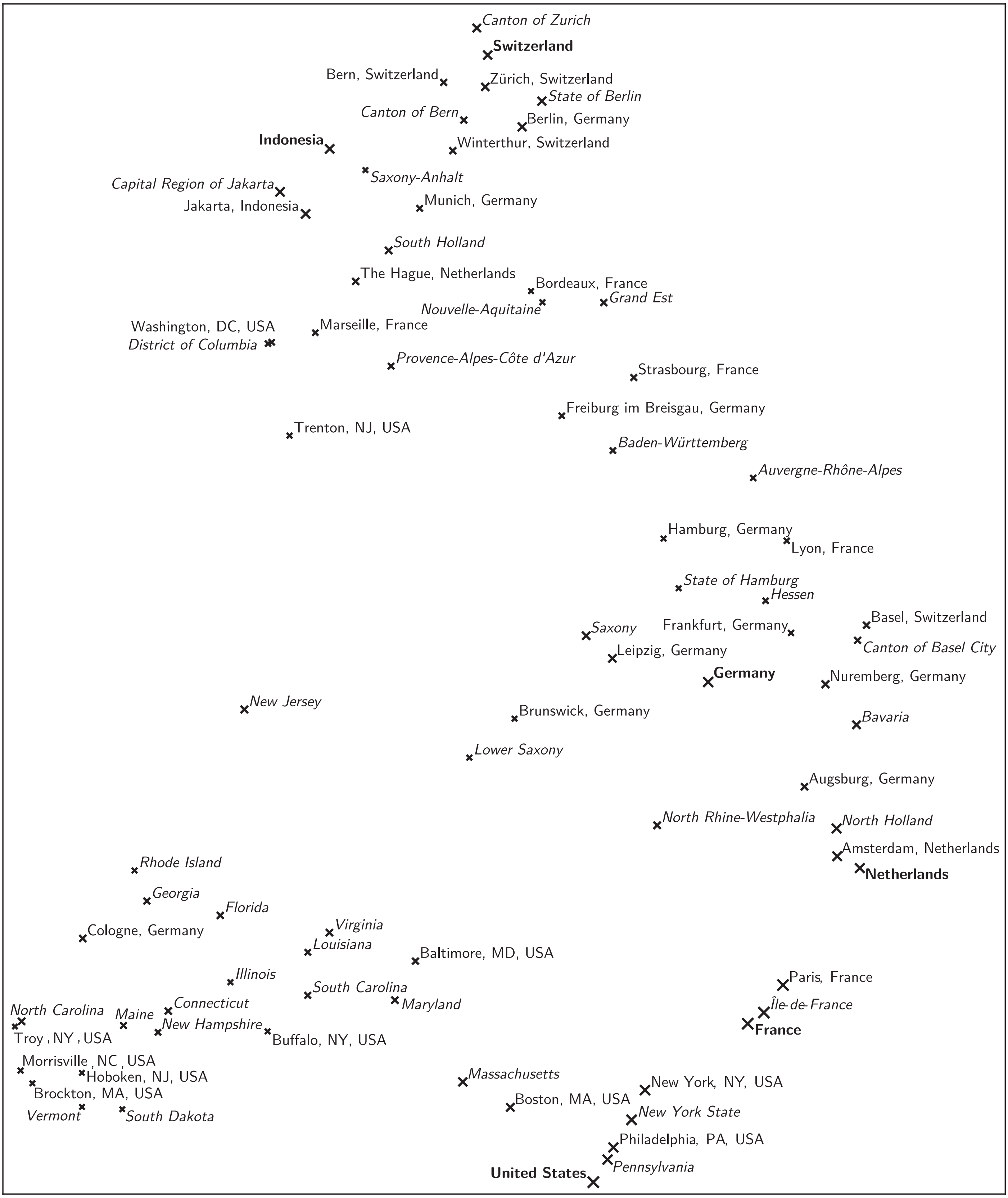

**Figure 32 | Semiotic distance between signs by publication place.** UMAP projection of the rupture graph $\Gamma_\rho$. The axes are dimensionless. Formatting differentiates three location levels: Cities, *States/Provinces* (italics), and **Countries** (bold). *Proximity in the sign space seems to reflect geographic vicinity or cultural similarities between publication places.*

Although the relative projected positions of publication centers do not faithfully reproduce geography, they arguably reflect geographic and cultural proximities. For instance, signs appear to be quite similar in maps published in Zürich, Bern, and Winterthur[14]. The signs produced in the German publication centers of Berlin and Münich also resemble the Swiss cluster. Close enough are the Hague and Indonesia, a historical colony of the Netherlands. Apart from Paris, French cities and provinces form a sparse yet apparent cluster comprising Marseille, Bordeaux, Strasbourg, and Lyon. The German city of Freiburg im Breisgau seems to share semiotic similarities with Strasbourg, which is located approximately 80 km farther north, on the other side of the Rhine.

With the exception of Berlin, Munich, and Cologne, the sign productions of German cities such as Frankfurt, Leipzig, Nuremberg, Brunswick, Augsburg, and Freiburg im Breisgau are closely related. Although they are located farther apart in the UMAP visualization, Berlin and Munich remain very similar to each other (Fig. B19). Basel appears to be idiosyncratic; its rupture coefficient vis-à-vis other Swiss and nearby German cities is comparatively low (Fig. B19). The major map-making center of Amsterdam is related to several German cities, especially Augsburg and Nuremberg. Paris, on the other hand, presents the most distinctive sign production and maintains rather weak relationships with other geographic areas. This observation partly qualifies the earlier interpretation that the peculiarity of the French frequency table could be interpreted as an indication of the country's cultural intermediacy with respect to the other countries under study. Indeed, as discussed in Chapter 2, Paris accounts for the large majority of French maps.

American cities appear well detached from their European counterparts. They form a separate cluster, with the exceptions of Washington D.C. and Trenton, NJ (near Philadelphia), as anticipated. However, this result should be interpreted with caution, since maps published solely in Washington are not included in ADHOC Images, as pointed out in Chapter 1. Consequently, the presence of Washington in this sample is foremost explained by cases of co-publication. American cities are spatialized following discernible substructures, one of the most notable grouping New York, Boston, and Philadelphia together. The remaining American cities are grouped in another distinct cluster. It is worth noting that suburban centers such as Troy, NY (near Albany), Morrisville, NC (near Raleigh), Hoboken, NJ (near New York), and Brockton, MA (near Boston) are projected farther from Europe, a pattern that may reflect the variability introduced by smaller sample sizes, if not cultural distinction. Mid-sized cities, such as Baltimore and Buffalo, NY, appear comparatively closer to both European and American large city centers.

Interestingly, the German city of Cologne, a historical cradle of printed cartography, as outlined in Chapter 2, appears close to the American secondary publication centers in Figure 32. Figure B19, in the Appendix, nuances this observation, indicating that the two cities whose sign productions are most similar to Cologne are, in fact, Marseille and Augsburg. Nevertheless, Cologne exhibits a marked idiosyncrasy—as reflected by its relatively high rupture coefficient with all other cities—which could in part explain its unexpected spatialization.

[14] Winterthur is a mid-sized city located about 20 km northeast of Zürich, in Switzerland.

### *Diffusion model of map signs*

To assess the determinants of observed similarities among semiotic systems, a city-centric linear model is defined. The initial choice of predictors is inspired from a gravitational intuition, hypothesizing that cartographic conventions flow primarily from smaller urban centers to larger ones and among larger centers. The model also assesses the effects of geographic proximity, shared language, and political-administrative national structure. Historical-chronological distance between periods of activity is incorporated through a simple heuristic.

Specifically, let us consider the following predictors:

- $d_{ij}^{geo}$ the geodesic distance between city pairs
- $d_{ij}^{time}$ the absolute difference between average years of publication
- $d_{ij}^{pop}$ the absolute difference between urban population sizes, as of 1850 (Tab. B1, in Appendix)
- $min_{ij}^{pop}$ the population of the smaller of the two cities as of 1850
- $\mathbf{1}_{ij}^{country}$ a Boolean indicating whether cities are located in different countries
- $\mathbf{1}_{ij}^{continent}$ a Boolean indicating whether cities are located on different continents
- $\mathbf{1}_{ij}^{language}$ a Boolean indicating, for each city pair, whether the majority language is different

The response is $\rho_{ij}$, the dyadic coefficient of rupture. Each predictor is normalized to take values between 0.0 and 1.0. Then, we define a linear model with fixed city effects:

$$\tilde{\rho}_{ij} = \beta_0 + \beta_1 d_{ij}^{geo} + \beta_2 d_{ij}^{time} + \beta_3 d_{ij}^{pop} + \beta_4 min_{ij}^{pop} + \beta_5 \mathbf{1}_{ij}^{country} + \beta_6 \mathbf{1}_{ij}^{continent} + \beta_7 \mathbf{1}_{ij}^{language} + \iota_i + \gamma_j + \varepsilon_{ij}$$

Because empirical values are measured for each city pair $(i, j)$, the observations are not independent but cross-correlated. Explicitly defining $\iota_i$ and $\gamma_j$ (i.e. city 1 and city 2) as fixed effects helps cancelling out city-specific factors in the estimation of the coefficients $\beta$. Additionally, the model is estimated with two-way cluster-robust variance, which accounts for residual interdependence and preserves the validity of the estimated variances and p-values despite the dyadic relationships between observations.

Here, only cities with at least 200 published maps are included in the model. The results, summarized in Table 4, indicate the relative effects of the significant predictors. $\mathbf{1}_{ij}^{language}$ and $\mathbf{1}_{ij}^{country}$ were not found to be statistically significant and were therefore not retained in the final model.

**Table 4 | Determinants of semiotic rupture between urban centers**. The response variable is the observed semiotic rupture between city pairs. Only significant predictors were retained in the final model. *i* and *j* account for city-specific determinants.

| predictor | $\beta$ | Sum sq. | df | p-value |
|---|---|---|---|---|
| $d^{time}$ | 0.31 | 1.35 | 1 | $< 0.001$ |
| $min^{pop}$ | 0.07 | 0.62 | 1 | $< 0.001$ |
| $d^{geo}$ | 0.71 | 0.37 | 1 | $< 0.001$ |
| $d^{pop}$ | 0.27 | 0.28 | 1 | $< 0.001$ |
| $\mathbf{1}^{continent}$ | –0.17 | 0.17 | 1 | $< 0.001$ |
| *i, j* | | 6.71 | 46 | |
| Residual | | 7.16 | 501 | |

Table 4 shows that cultural similarity between urban centers is conditioned by the historical period of activity, urban population size, and geographic proximity. First, and quite intuitively, cities active during the same period tend to be semiotically more similar. The second most important determinant is the size of urban production centers ($min_{ij}^{pop}$). In this respect, the coefficient suggests that *semiotic rupture is greater between large urban centers*. This effect is tempered by the comparatively lower impact of population size discrepancies ($d_{ij}^{pop}$), which implies that sign systems remain more consistent among cities of similar sizes. Given the former result, this effect applies foremost to smaller production centers. Third, conformally to expectations, geographical distance engenders semiotic rupture. Let us note, however, that the fitted model imposes a negative coefficient on the $continent_{ij}$ variable, indicating that intercontinental (e.g., transatlantic) distances contribute *less* compared to intracontinental land distances.

Modeling the semiotic landscape suggests that larger urban centers, such as Paris, New York, or Berlin, tend to develop more contrasted sign systems. Smaller centers tend to converge, particularly when they are located close to one another. While this finding deviates somewhat from the initial hypothesis, it remains consistent with the relationships observed in the graph of geographical ruptures (Fig. B19). The observation that larger cities employ distinctive cartographic signs further suggests that they might play an essential role in the process of diversification.

The influence of geographic distance confirms the qualitative interpretation of Figure 32. Moreover, it highlights a process of spatial diffusion of cartographic signs, akin to linguistic diffusion phenomena (Séguy, 1971; Wieling & Nerbonne, 2015). The importance of geographical proximity has also been documented in the broader literature on the diffusion of ideas (Abramo et al., 2020). The reduced spatial effect of intercontinental divides further suggests the globalized character of cartographic signs across the Western world and its colonies.

As evidenced in Table 4, a substantial share of the explained variance is attributable to city-specific effects. City-specific effects excepted, global effects only explain 26% of the remaining variance. The next chapter will be dedicated to narrow this gap by extending the model of map semiotics outlined here, through the introduction of more generic operable units.

## 6.11 Conclusion

This chapter focused on the encoding and study of cartographic signs. First, it involved the extraction of 63 million cartographic signs, such as icons and symbols. The resulting object detection model and training data are published along with this dissertation. Then, it presented and validated the methodology for encoding signs by adapting DINOv2, a contrastive ViT. The resulting space of representation is concentrated on salient shapes and colors attached to the sign itself, discriminating background content. It is readily able to solve semantic classification tasks, and estimate sociocultural distinctions and similarities, as demonstrated in the experiment based on communities of mapmakers.

To handle the large quantity of data, the analysis of cartographic signs relies on the grouping of signs into clusters, each represented by a representative exemplar. This also facilitates the visual interpretation of the sign space.

The structure of the operationalized semantic–symbolic space appears hierarchical; thus, the space can be studied as a phylogeny of related signs. The diversification of cartographic signs seems to have occurred in two waves, first, in the middle of the 17$^{th}$ century, and then at the end of the 18$^{th}$ century, with the progressive mechanization of reproduction processes. Individually, Figure 19 shows that the frequency trajectory of sign clusters parallels an evolutionary diffusion process. Periods of synchronic, rapid, change are visible, indicating, first, the relatedness of cluster frequencies, and second, the existence of moments of rupture. The first finding led to the conclusion that signs might form coadapted complexes, corresponding to jointly replicated, consistent, semiotic systems. Eight such complexes were identified, based on the co-occurrence of sign variants. The second line of inquiry, corresponding to the search for moments of ruptures, was also pursued, leading to the periodization of cartographic signs in six semiotic moments: 1492–1633, 1634–1737, 1738–1788, 1789–1876, 1877–1895, and 1896–1947. The same approach also permitted the identification of three scale ruptures: 1:4,750, 1:93,200, and 1:671,000. Finally, the national differences and distinctions among urban publication centers were examined. The statistical modeling of similarities in sign frequencies indicated that signs tend to diffuse foremost to similar-sized, geographically close, active publication centers. By contrast, large publication centers tend to develop into distinct, contrasted schools of representation.

This chapter marks a shift in the analysis. First, it moves to the observation to small, discrete, elements of cartographic figuration, by contrast to the observation of whole map images in the previous chapter. Second, it adopts an analytical and interpretative framework in which the replication of sign variants is hypothesized a cultural evolution process. In this perspective, the variations in sign frequencies are considered the manifestation of slight shifts in the balance of the cultural and semiotic system of cartography.

This perspective offers a coherent frame of interpretation of the results. However, insofar as this chapter was focused on the analysis of very specific aspects of cartographic figuration—pertaining to icons and symbols—it is difficult to generalize these results and their theoretical implications to the broader context of cartographic semiotics. Furthermore, the supervised methodology employed for detecting signs, based on manual annotation, implies that the composition of the resulting dataset is partly influenced by the training data itself and its representativeness. While that dependency was inevitable in the specific case of signs, less directed, unsupervised methods may also be considered to model complementary aspects of cartographic figuration. The next and final chapter will attempt to generalize the approach in this prospect and pursue the hypothesis that the dynamics in map figuration can be interpreted as cultural evolutionary processes.

# Appendix A – Testing the impact of *weak* features on classification

## *Aim & Method*

The aim of this complementary experiment is to demonstrate the existence of *weak* features, related to the grain and color of the map material (e.g. paper type), and to the specifics of digitization (e.g., scanner). The objective is to prove the opportunity for such features to bias classification. For this purpose, the setting of the second test (Section 6.5) was replicated. Instead of DINOv2–derived features, however, color descriptors that specifically target the sample background were used. Each sample was represented by a three-dimensional vector, corresponding to the red-green-blue (RGB) color channels of the sample background. The background is segmented using the foreground segmentation model. The background color is computed as the average pixel color $\vec{c_i}$, weighted by the soft mask of the background prediction: $w = (1 - \sqrt{y})$, where $y$ denotes prediction logits, as in the third test (Section 6.6).

$$background_color = \frac{1}{\sum^{\forall i} w_i} \sum^{\forall i} \vec{c_i} \cdot w_i$$

A multinomial logistic regression is then used to classify the samples in 24 classes. The experiment is repeated 100 times, with a different partition into training and validation sets (80%/20%) for each trial.

## *Results*

Table A1 shows the results of the experiment. As expected, the RGB color descriptors are less discriminative compared to the features derived from the DINOv2 model (Table 2). However, their explanatory power is significantly above chance level. This indicates that the multinomial logistic model can learn from these *weak* features to achieve a better-than-random classification performance. In other words, the background color can effectively bias the classification, and I argue that it does. Min-max normalization marginally reduces the effect on precision.

**Table A1 | Impact of weak features on sign classification**. In each experiment, the classification is performed by linear probing. For both precision and recall, the provided range corresponds to the 95% confidence interval of the mean. The last row indicates the level of chance.

| Experiment | Precision [95% CI] | Recall [95% CI] |
|---|---|---|
| color background (base) | 38.8% [36.2–40.1] | 51.6% [51.1–52.0] |
| – after min-max norm. | 35.8% [32.5–38.8] | 50.7% [50.6–50.9] |
| *chance* | *25.6% [24.9–26.2]* | *50.6% [50.6–50.6]* |

# Appendix B – Supplementary Materials

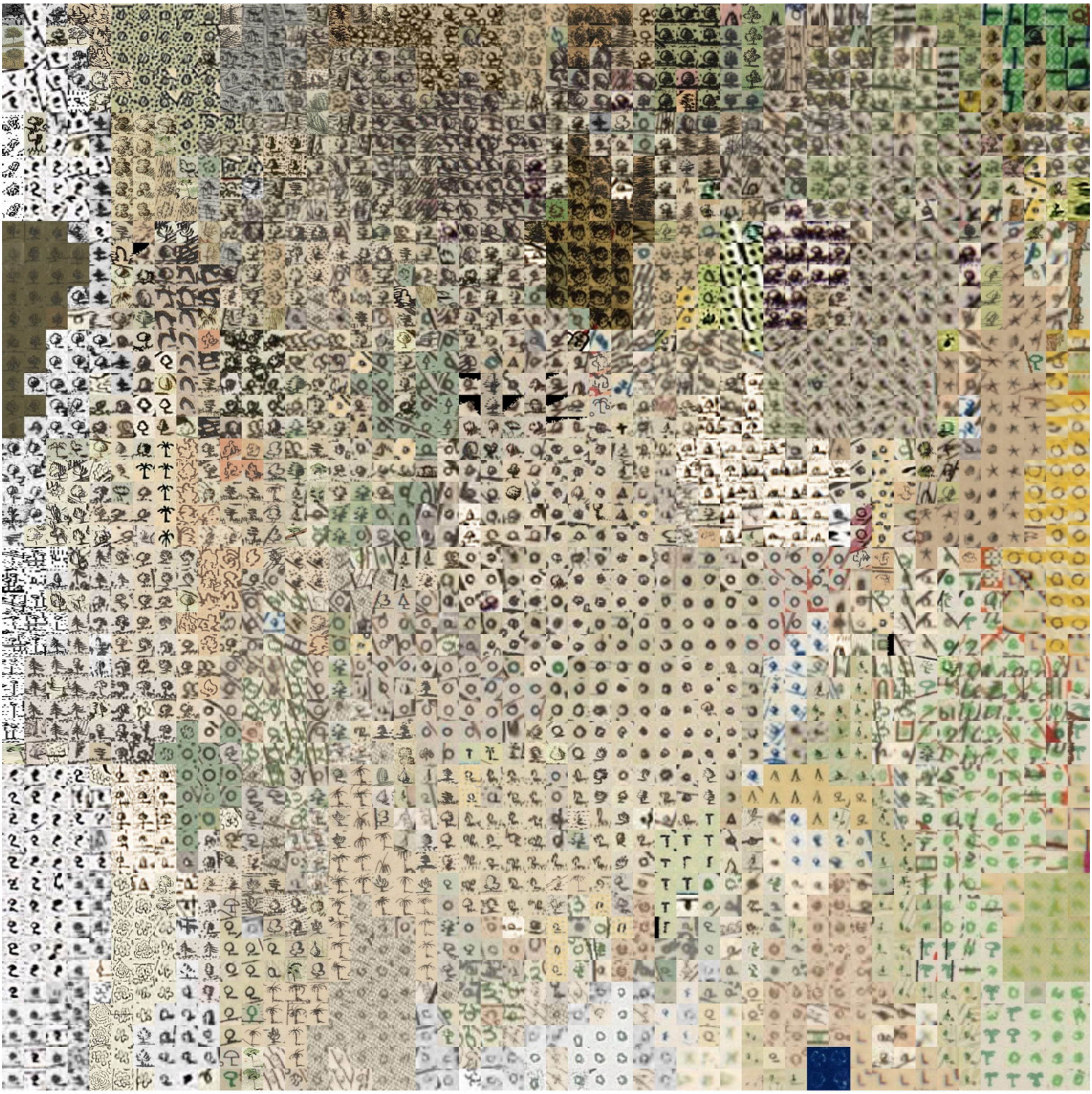

**Figure B1 | Tree icons.** Subset of 2500.

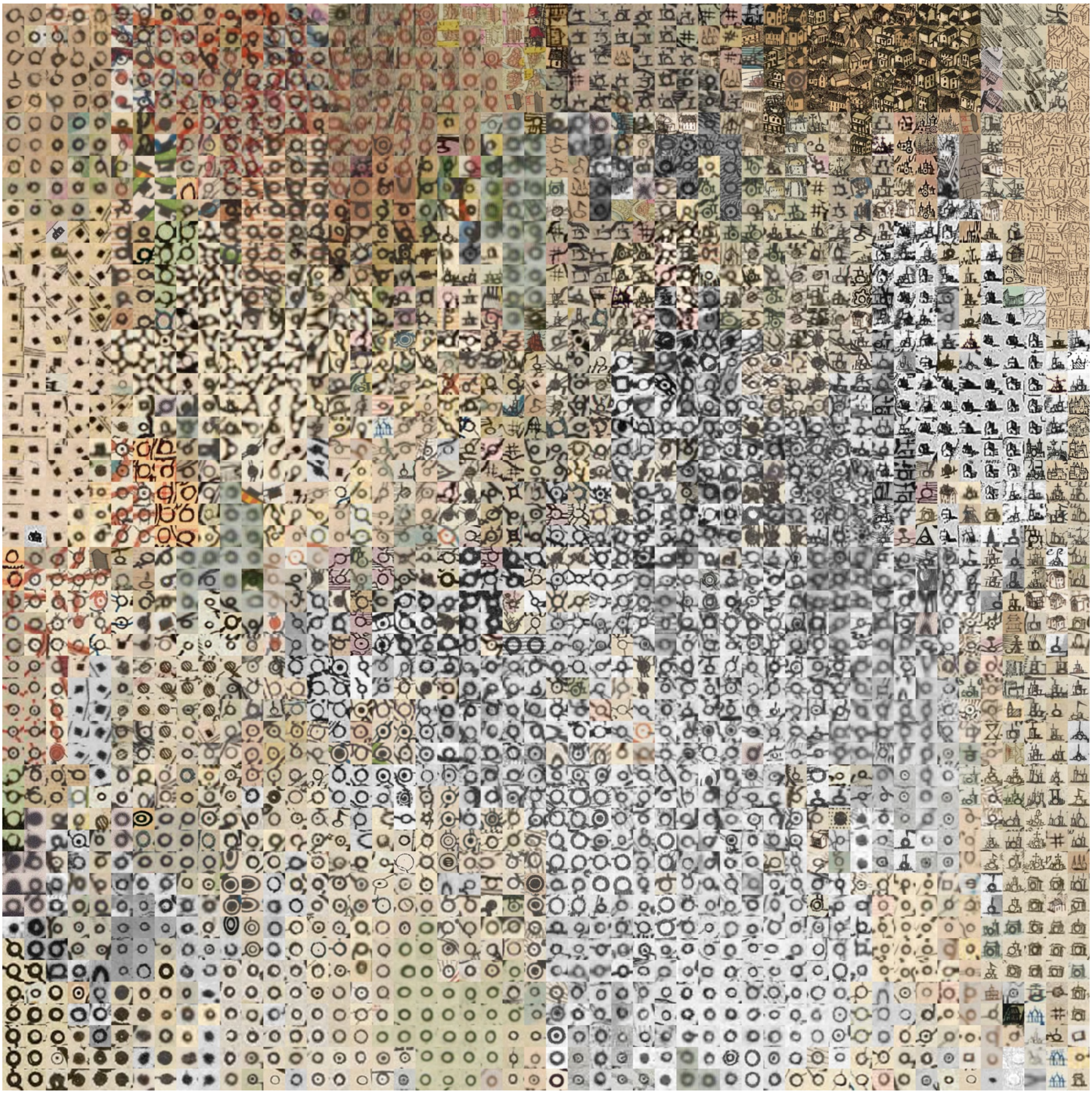

**Figure B2 | Settlements and buildings icons.** Subset of 2500 icons.

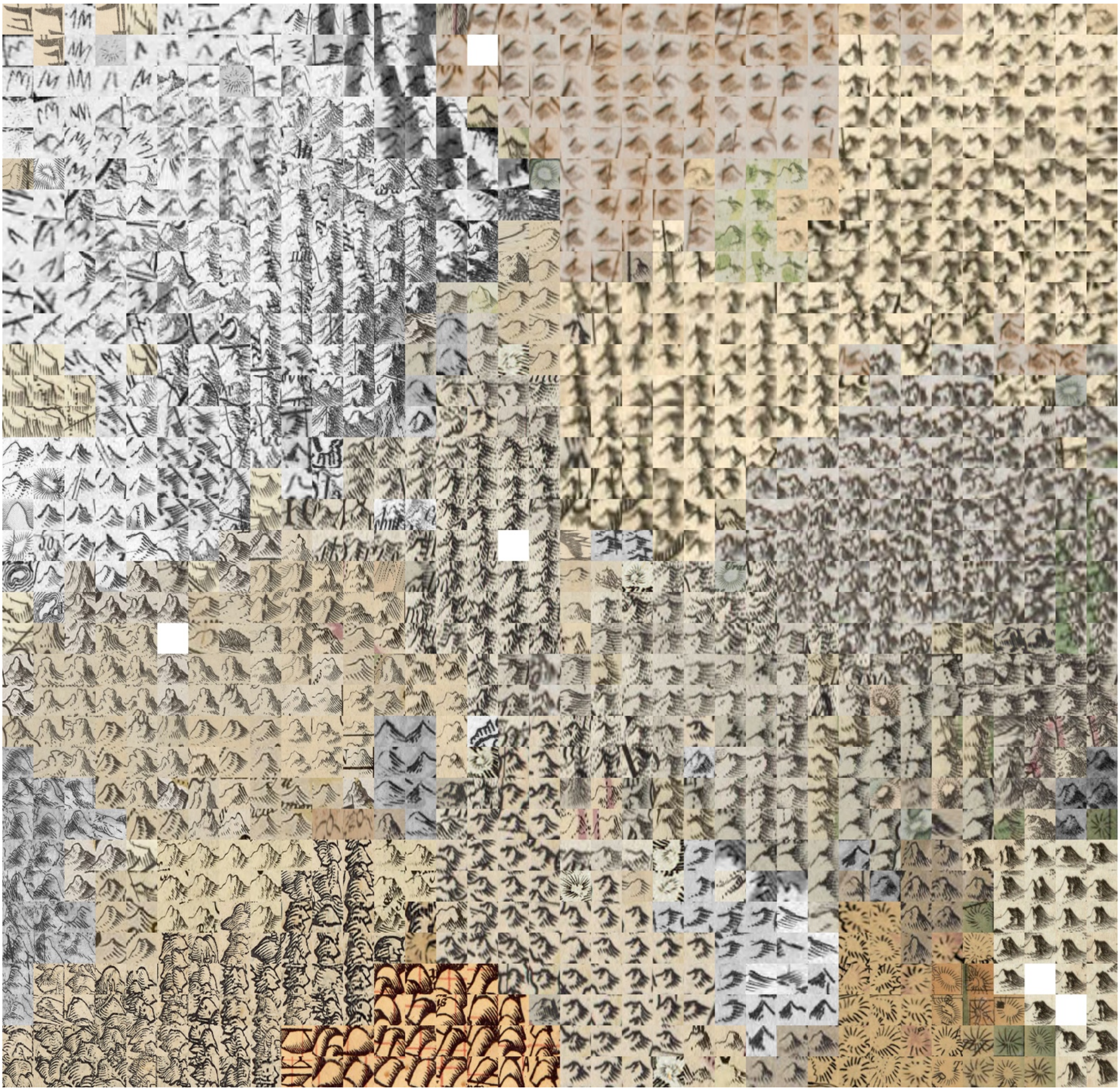

**Figure B3 | Mountains and hills icons.**

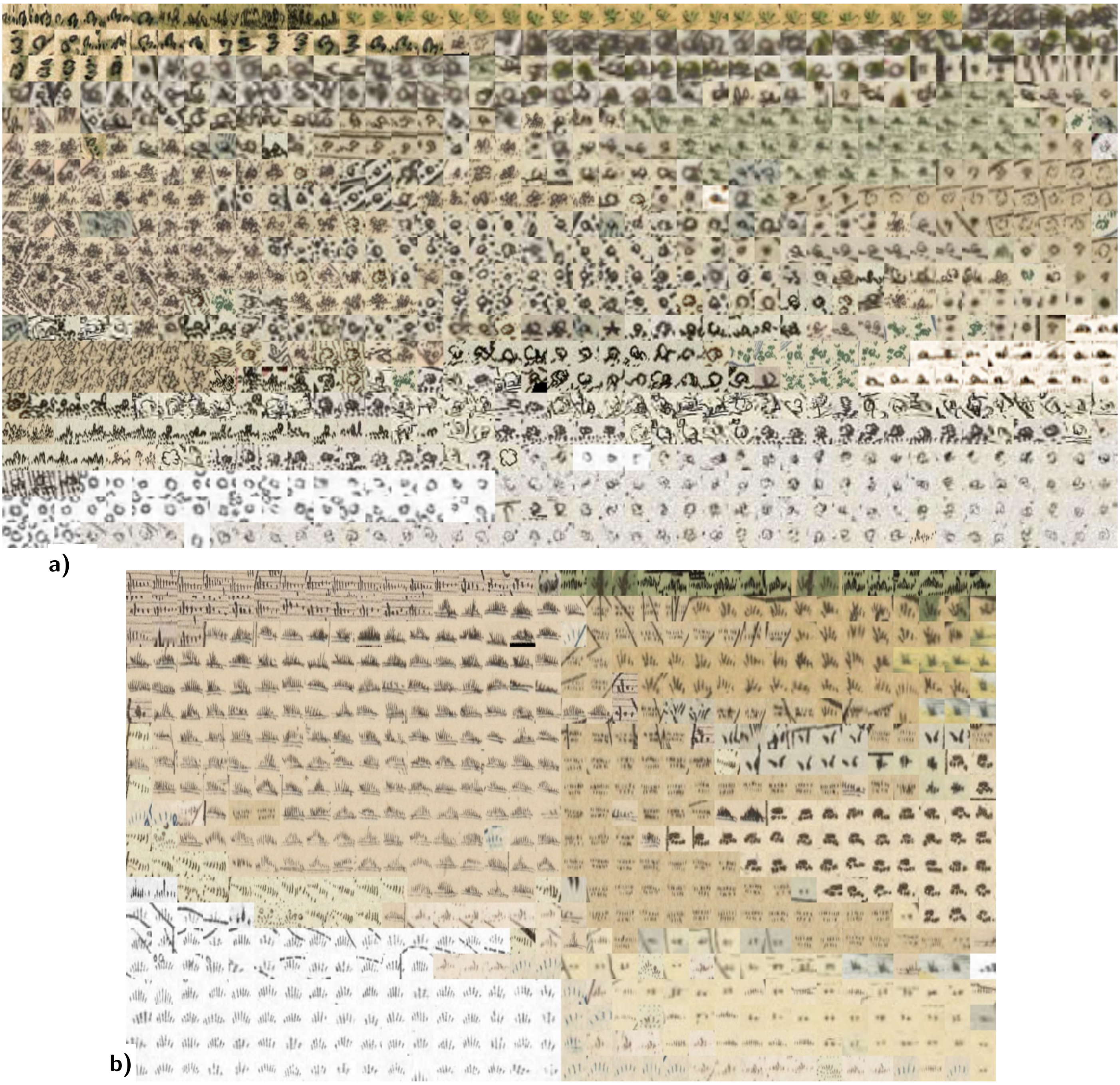

Figure B4 | Icons (a) bushes, (b) grass.

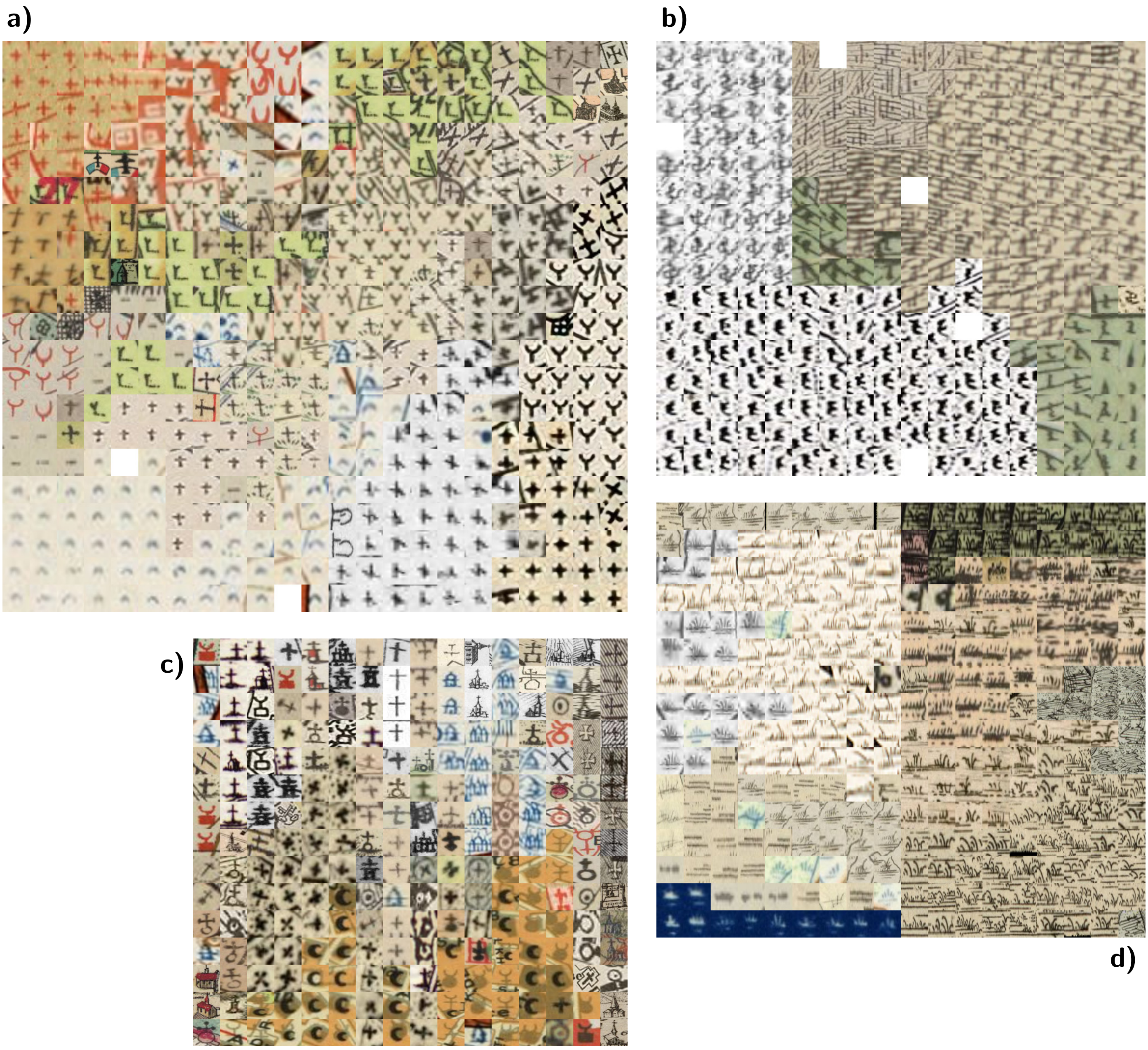

**Figure B5 | Icons (a) graves, (b) vine, (c) religious monument, and (d) marsh.**

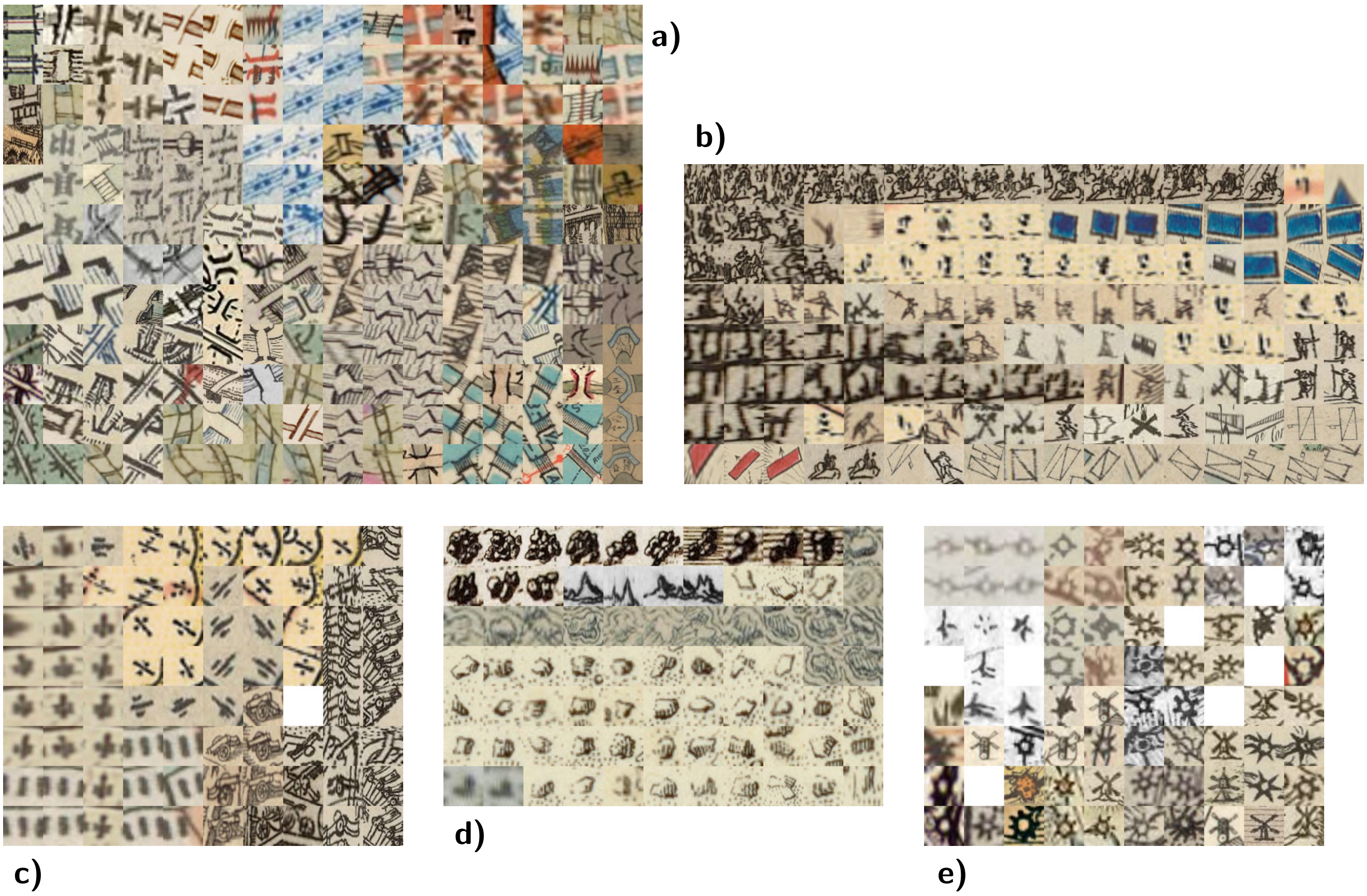

**Figure B6** | **Icons (a) bridges, (b) armies, (c) cannons, (d) rocks, and (e) mills.**

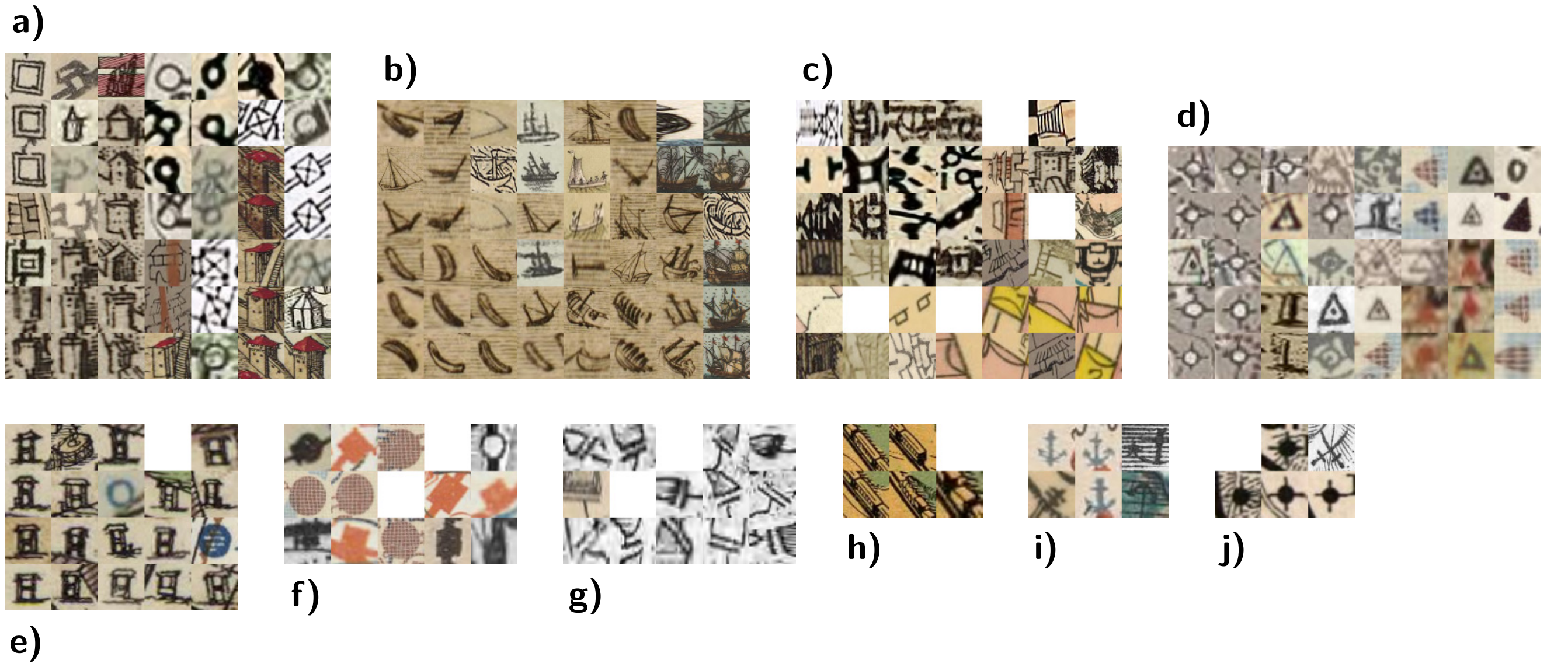

**Figure B7** | **Icons (a) towers, (b) ships, (c) gate, (d) survey point, e) well or bassin, f) train or metro station, g) dam or lock, h) train, i) harbor, and j) battlefield.**

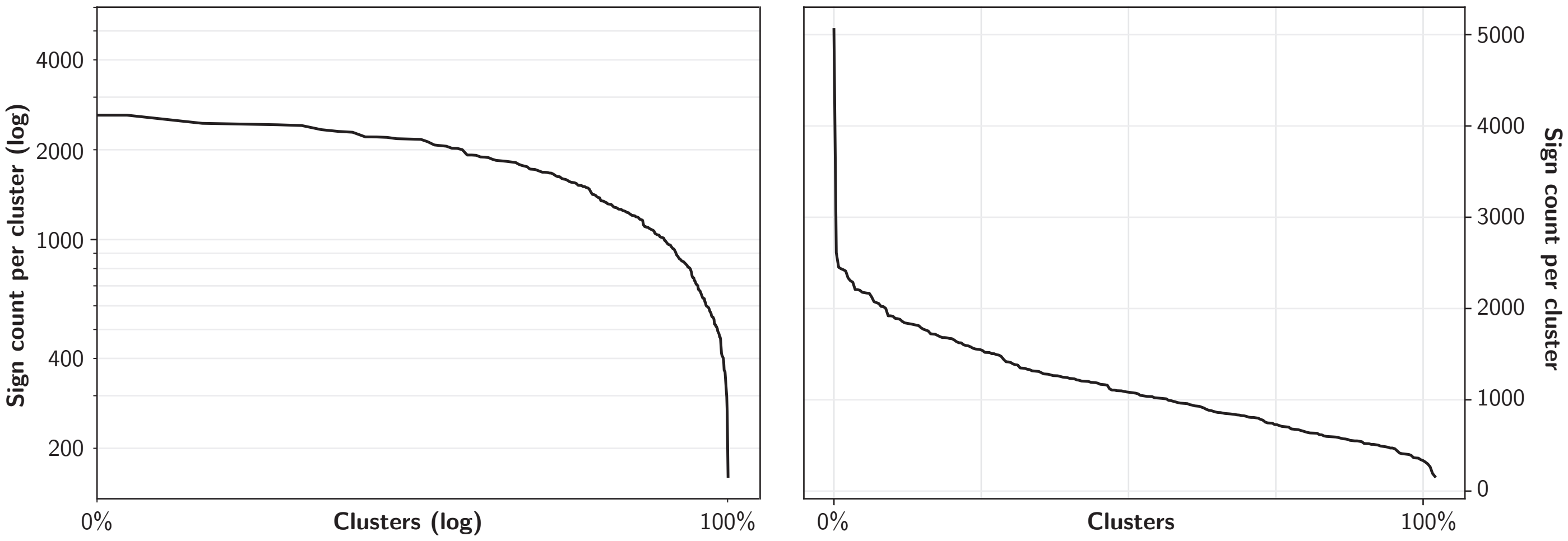

**Figure B8 | Distribution of cluster sizes.** Computed on a subsample of 256 clusters. The left subfigure reports the distribution in a log-log plot, while the right one plots it linearly. *The distribution approximately follows a power law.*

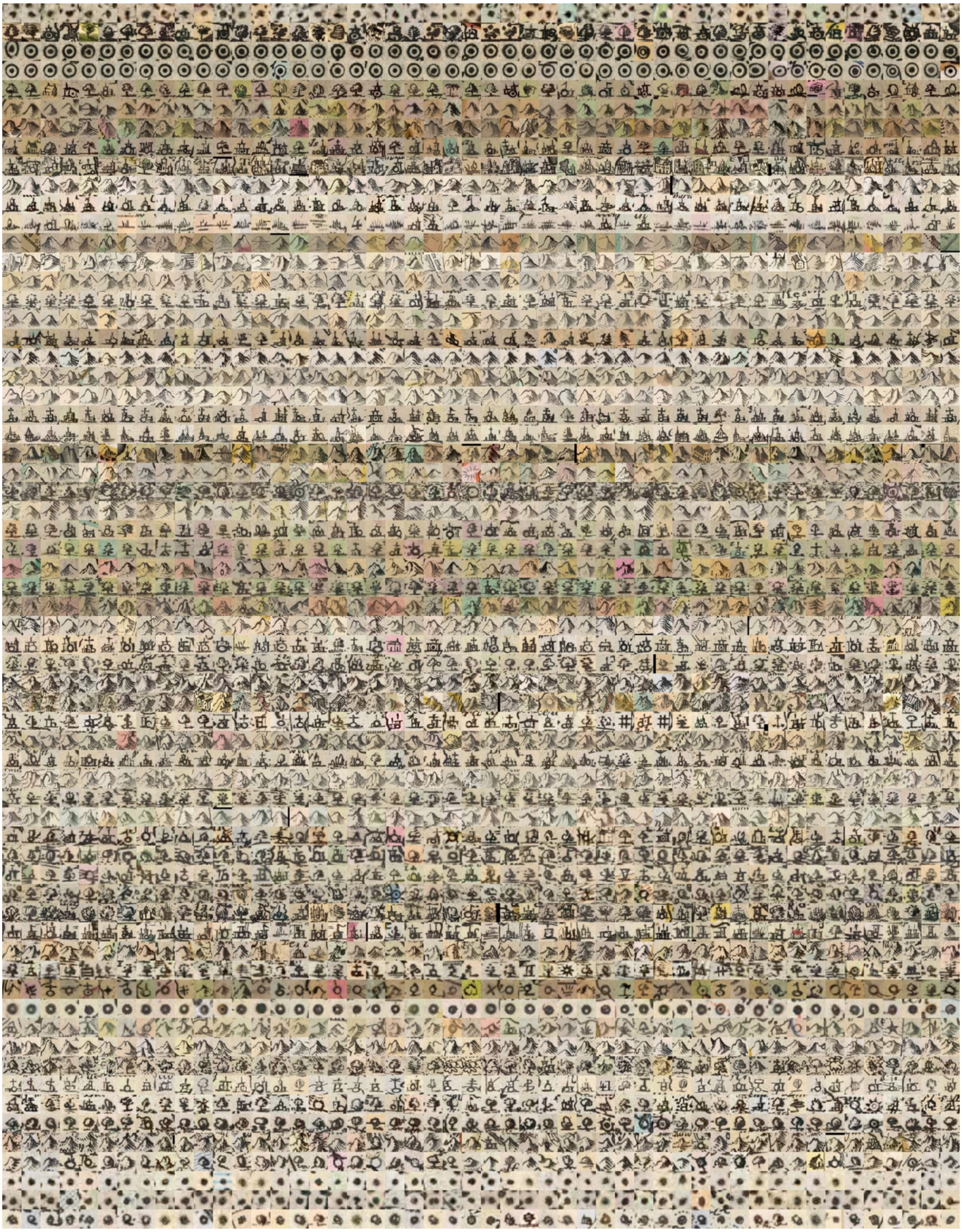

**Figure B9 | Examples of 64 sign clusters.** One cluster per row, arranged in a chronological order (from 1492 to ca. 1770). For each row, the signs are horizontally arranged by their probability of belonging to this particular cluster. The first sign is the exemplar. Continued in Fig. B10.

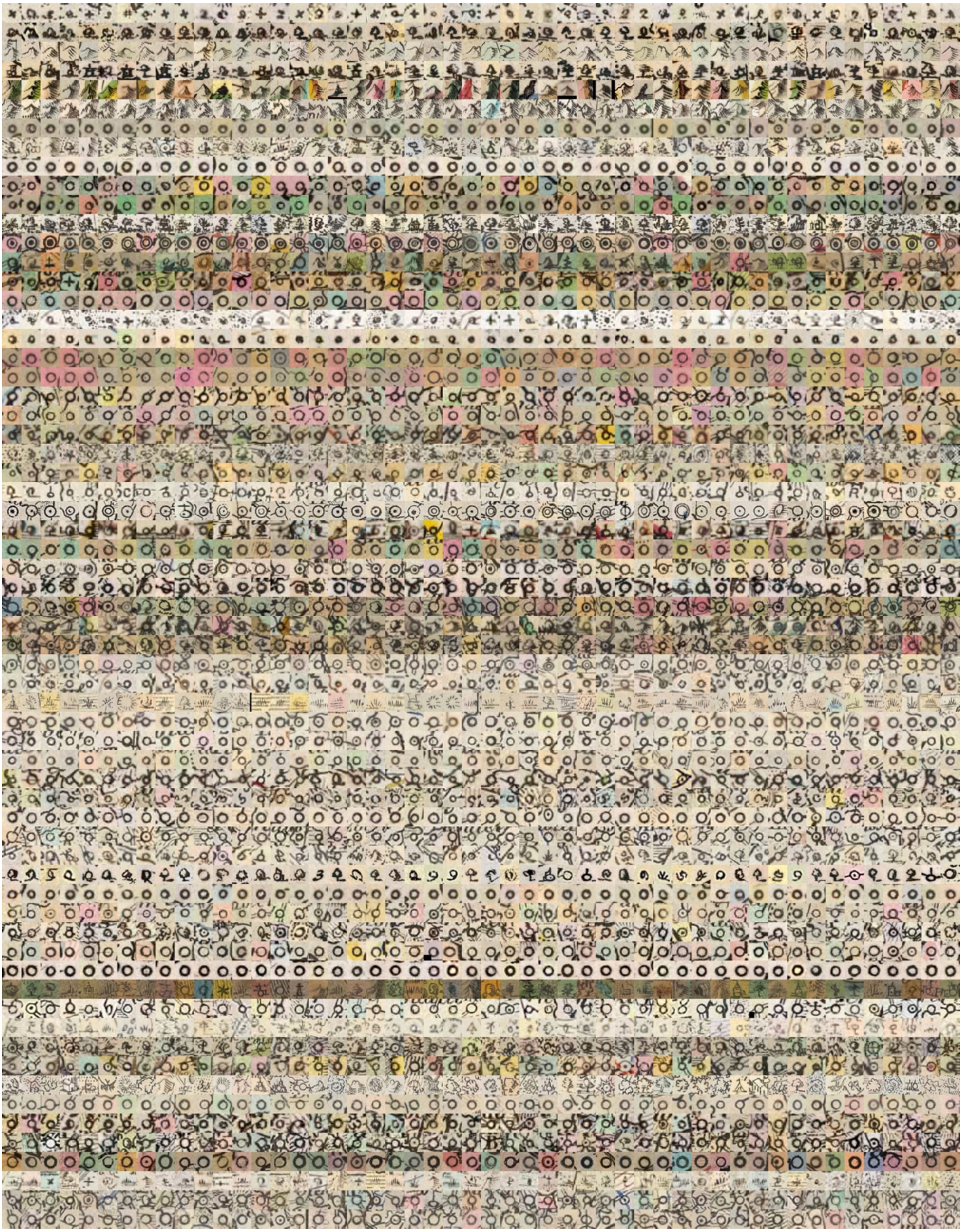

**Figure B10 | Examples of 64 sign clusters.** One cluster per row, arranged in a chronological order (ca. 1750–1870). For each row, the signs are horizontally arranged by their probability of belonging to this particular cluster. The first sign is the exemplar. Continued in Fig. B11. *Hill, tree, and city icons, prevalent until the early 18th century (Fig. B9) were progressively replaced by more symbolic signs at the turn of the 19th century (Fig. B10).*

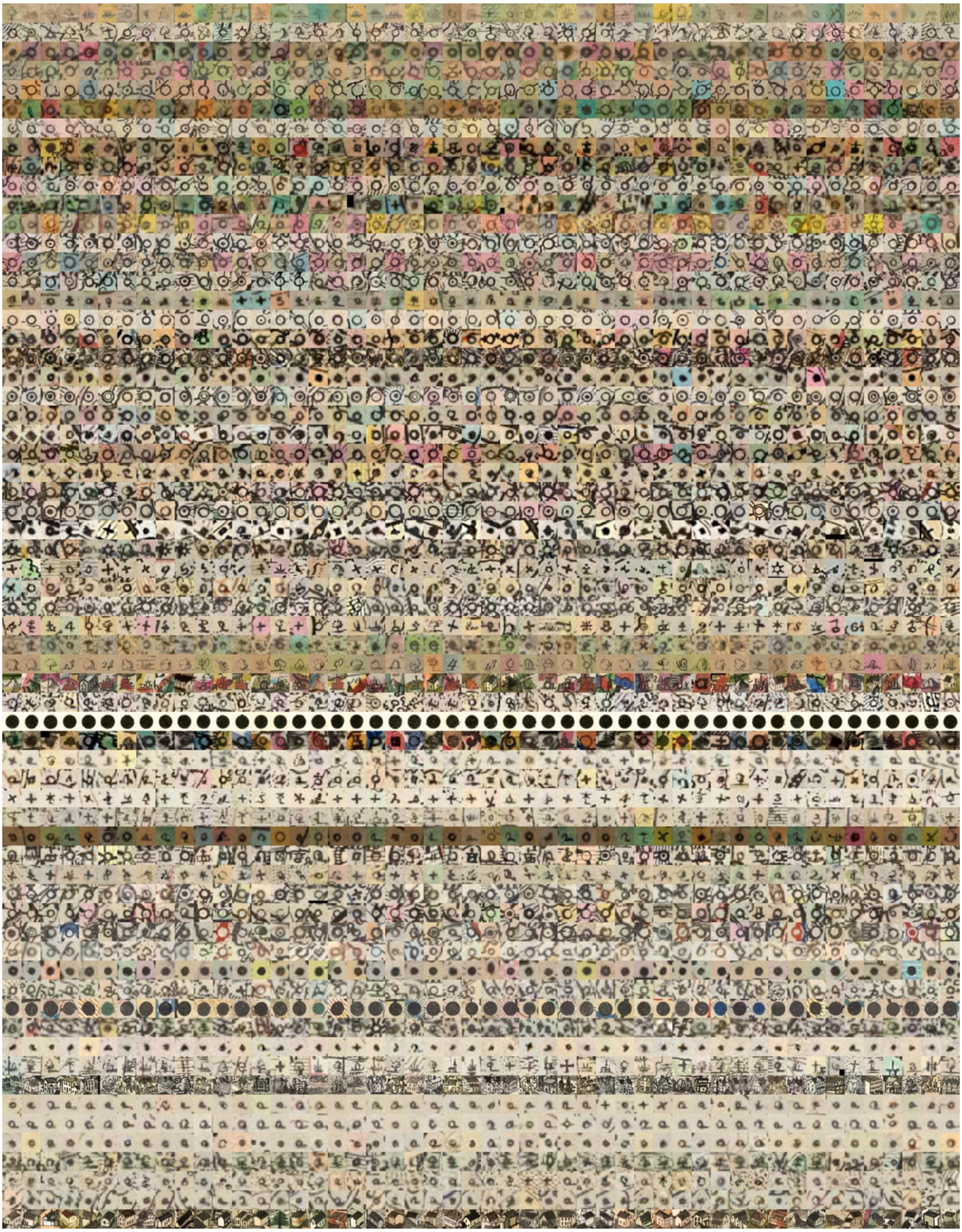

**Figure B11 | Examples of 64 sign clusters.** One cluster per row, arranged in a chronological order (ca. 1850–1900). For each row, the signs are horizontally arranged by their probability of belonging to this particular cluster. The first sign is the exemplar. Continued in Fig. B12.

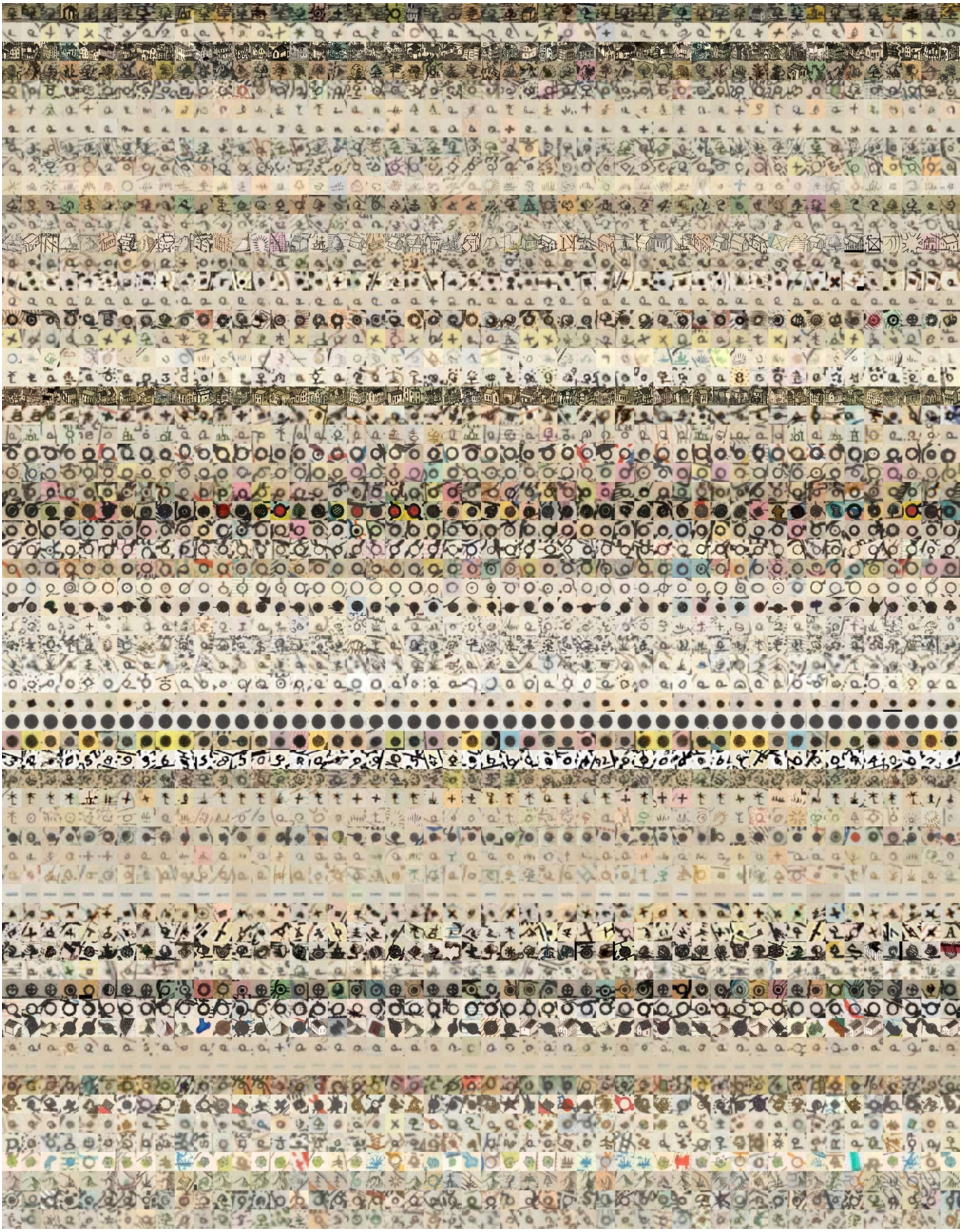

**Figure B12 | Examples of 64 sign clusters.** One cluster per row, arranged in a chronological order (from ca. 1880 to 1947). For each row, the signs are horizontally arranged by their probability of belonging to this particular cluster. The first sign is the exemplar.

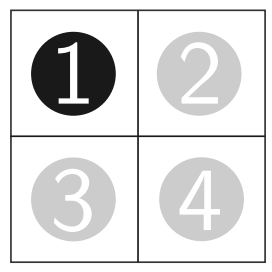

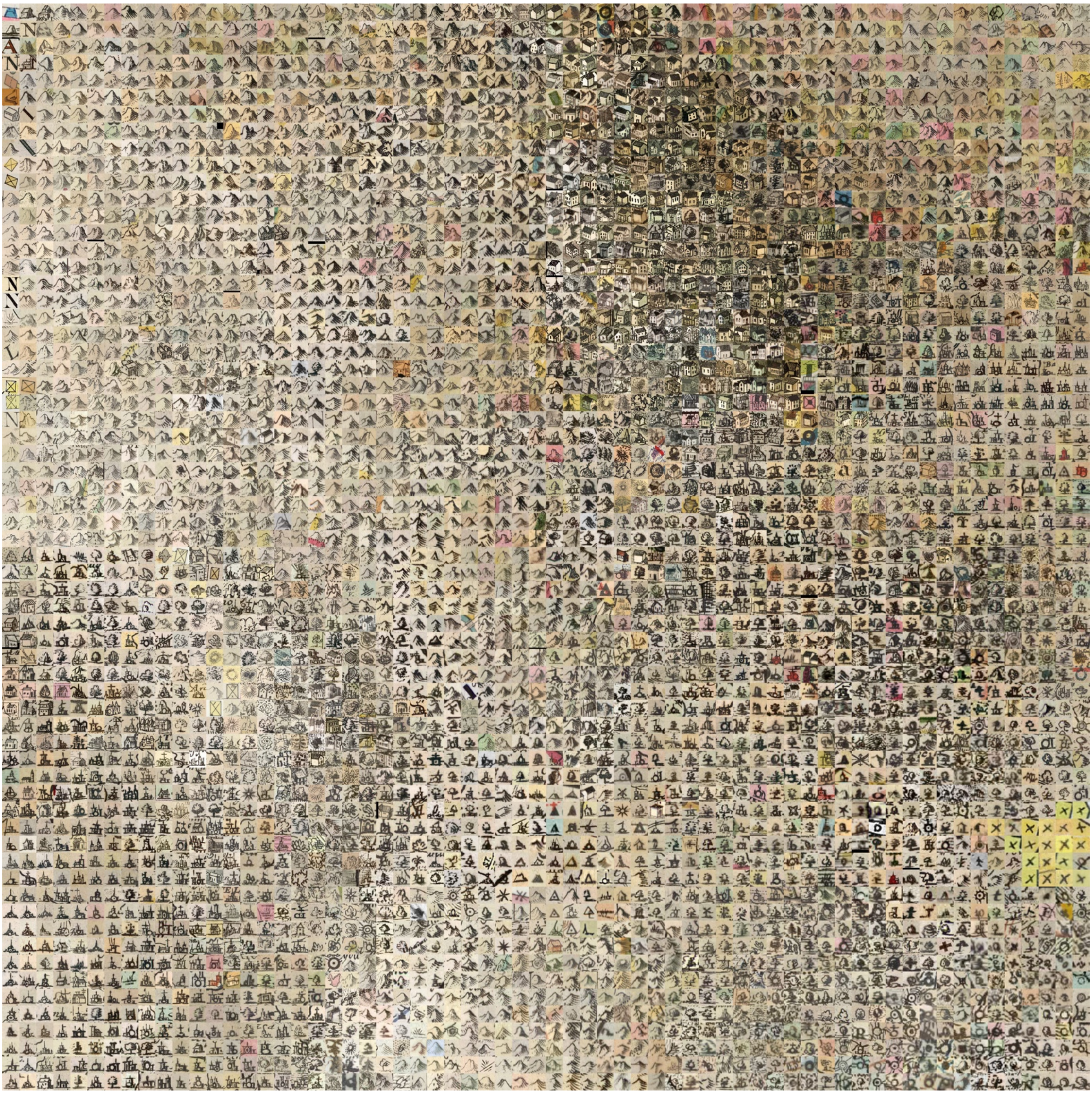

**Figure B13 | Enlarged excerpt from the sign mosaic.** Quadrant 1 (top left). The exemplars are spatialized, using t-stochastic neighbor embedding (t-SNE). The mosaic (Fig. 13) is subdivided in four quadrants for the facilitating its analysis and discussion.

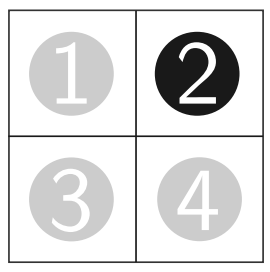

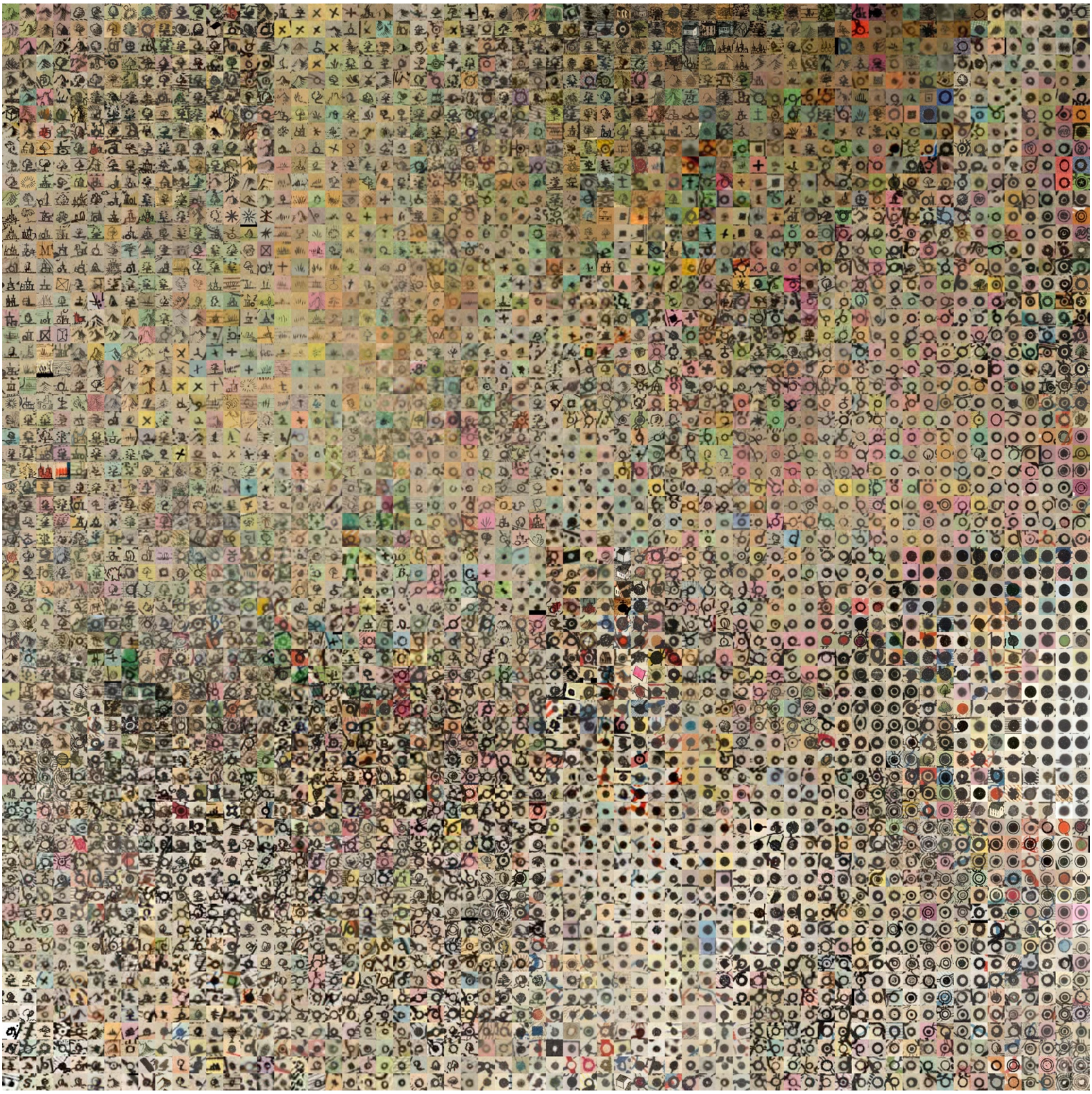

**Figure B14 | Enlarged excerpt from the sign mosaic.** Quadrant 2 (top right). The exemplars are spatialized, using t-stochastic neighbor embedding (t-SNE). The mosaic (Fig. 13) is subdivided in four quadrants for the facilitating its analysis and discussion.

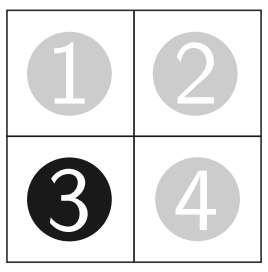

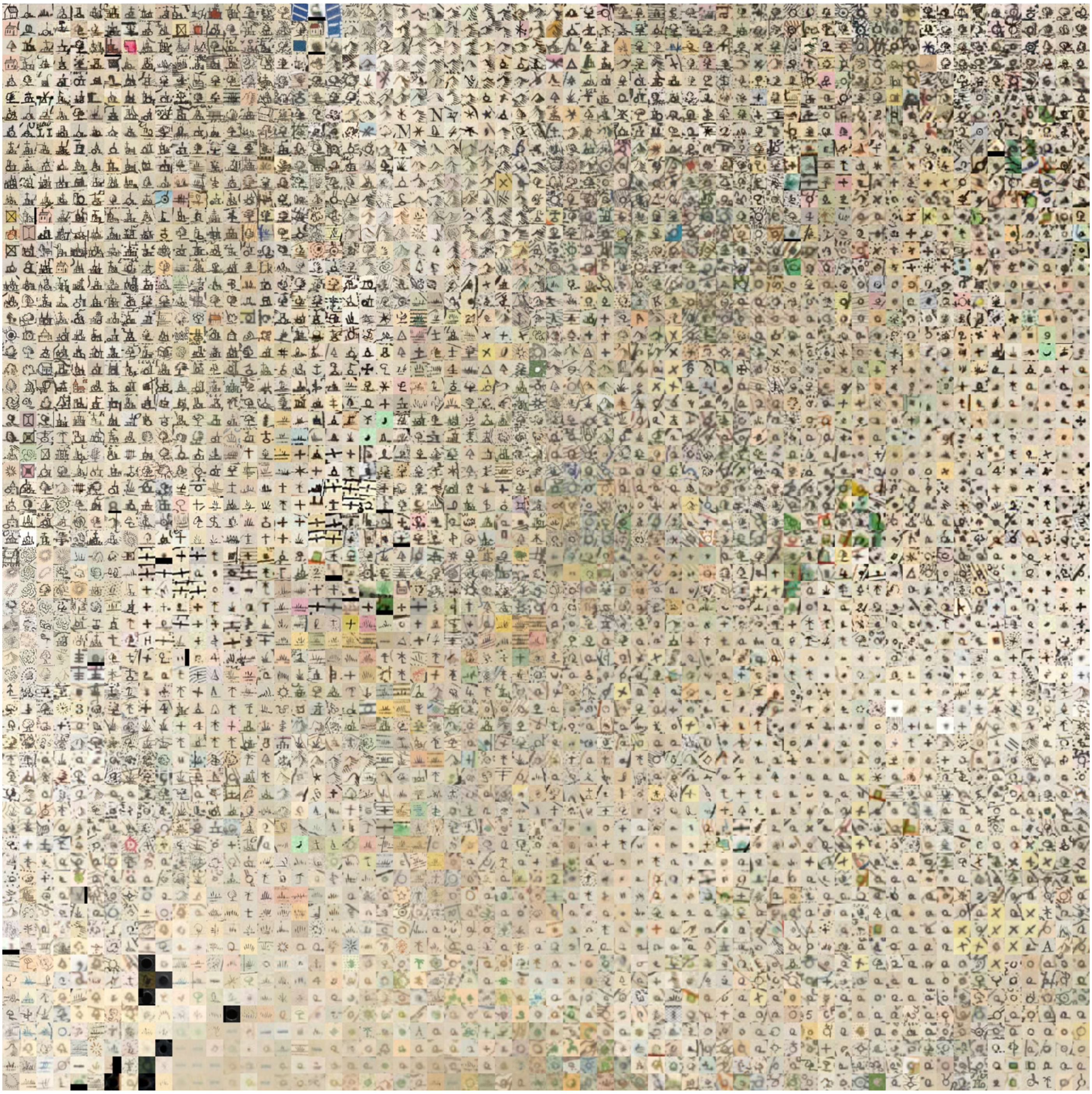

**Figure B15 | Enlarged excerpt from the sign mosaic.** Quadrant 3 (bottom left). The exemplars are spatialized, using t-stochastic neighbor embedding (t-SNE). The mosaic (Fig. 13) is subdivided in four quadrants for the facilitating its analysis and discussion.

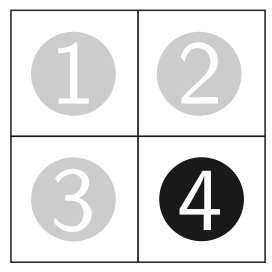

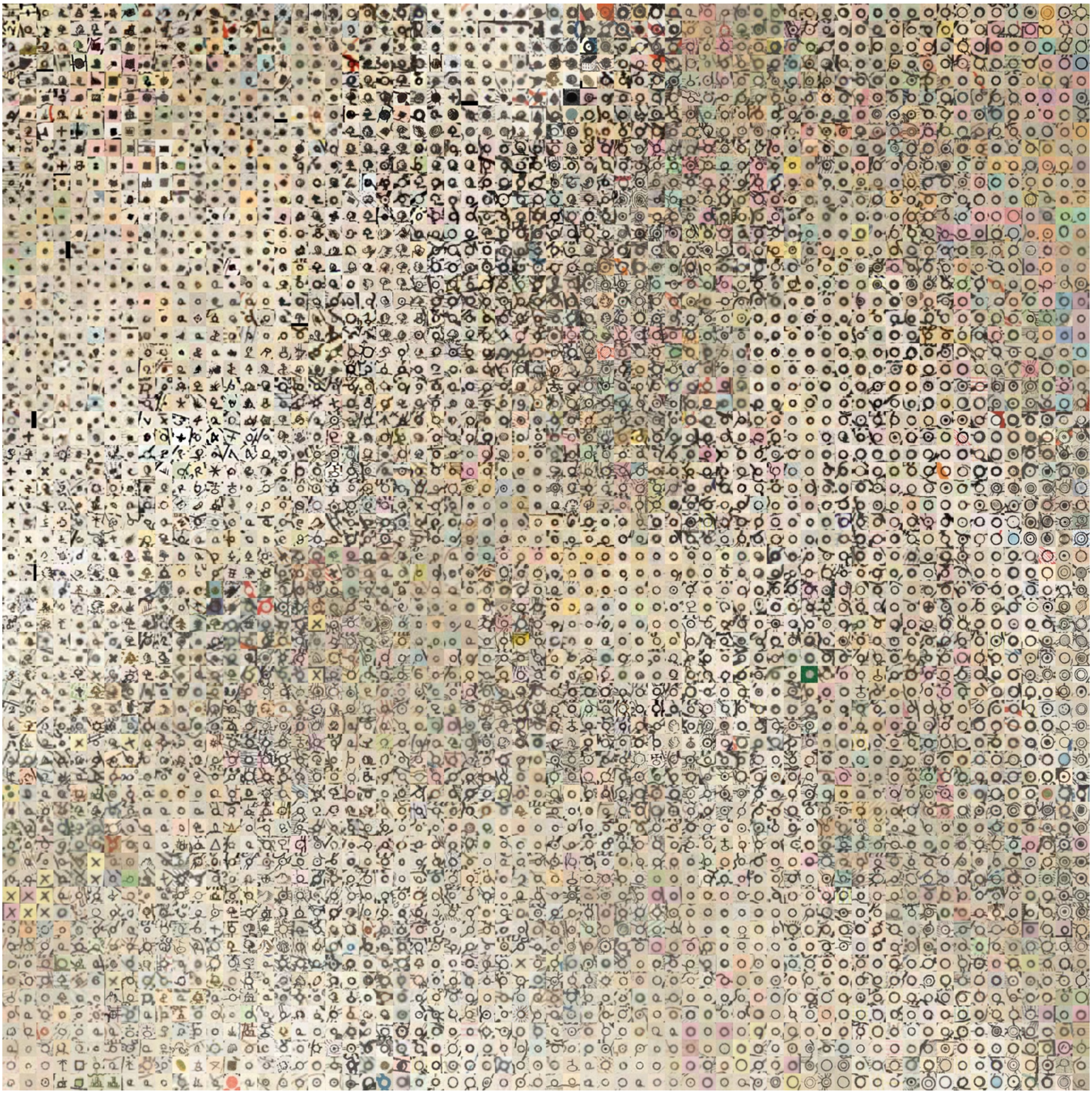

**Figure B16 | Enlarged excerpt from the sign mosaic.** Quadrant 4 (bottom right). The exemplars are spatialized, using t-stochastic neighbor embedding (t-SNE). The mosaic (Fig. 13) is subdivided in four quadrants for the facilitating its analysis and discussion.

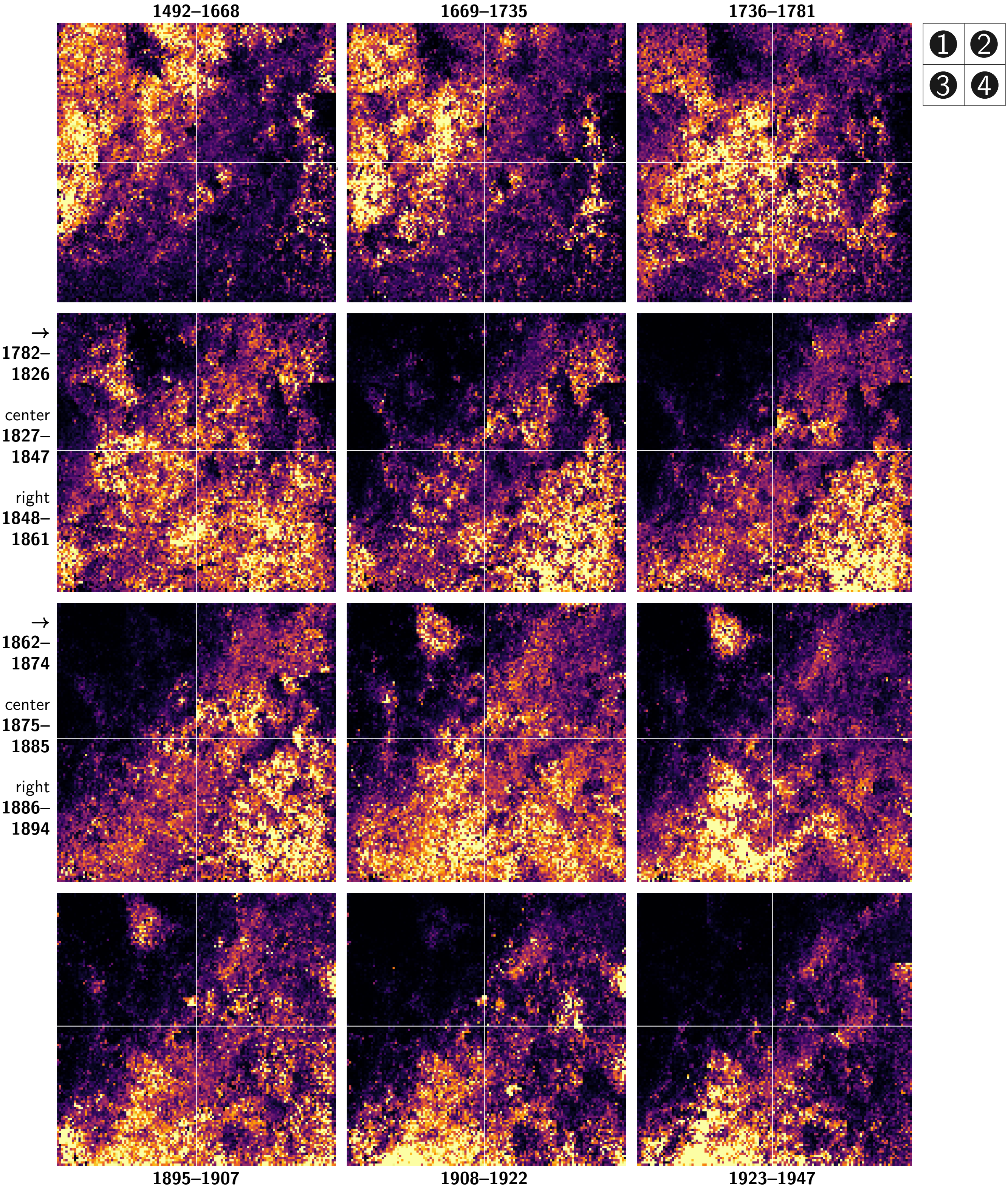

**Figure B17 | Evolution of the relative distribution of map signs by equal time stratum.** Each block depicts the relative frequency of sign clusters for a specific time stratum, based on the publication year. Sign clusters are spatialized as depicted in the sign mosaic (Fig. 14). Brighter areas correspond to higher frequencies. Frequencies are normalized with saturation at the 95$^{th}$ percentile of each stratum. The 12 time strata are computed by uniform stratification of the data. The graticule in the upper right corner indicates the four quadrants reported in Fig. 14. *Equal stratification does not capture shifts in the occupation of the sign space as consistently as the periods defined by moments of rupture (e.g. Fig. 25).*

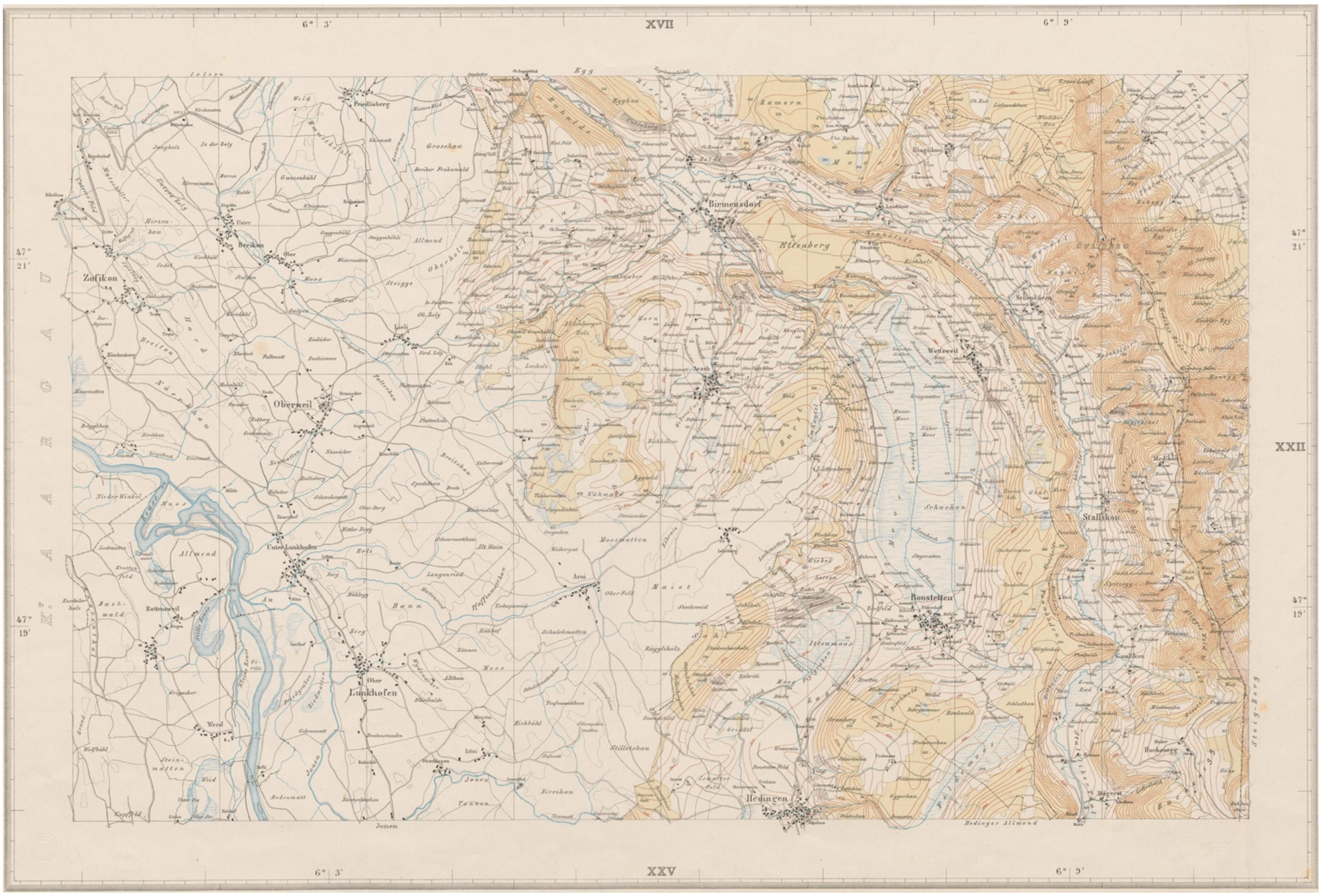

**Figure B18 | Birmensdorf sheet in the topographical map series of the canton Zürich at the scale of 1:25,000.** H. Enderli, Johannes Wild, J. Graf, J. Brack. *N° XXI Birmensdorf*, 1855. Topographisches Bureau, Zürich. Chromolithography. 35 x 52 cm. ETH Bibliothek, Rar KA 156:21. doi: 10.3931/e-rara-23700

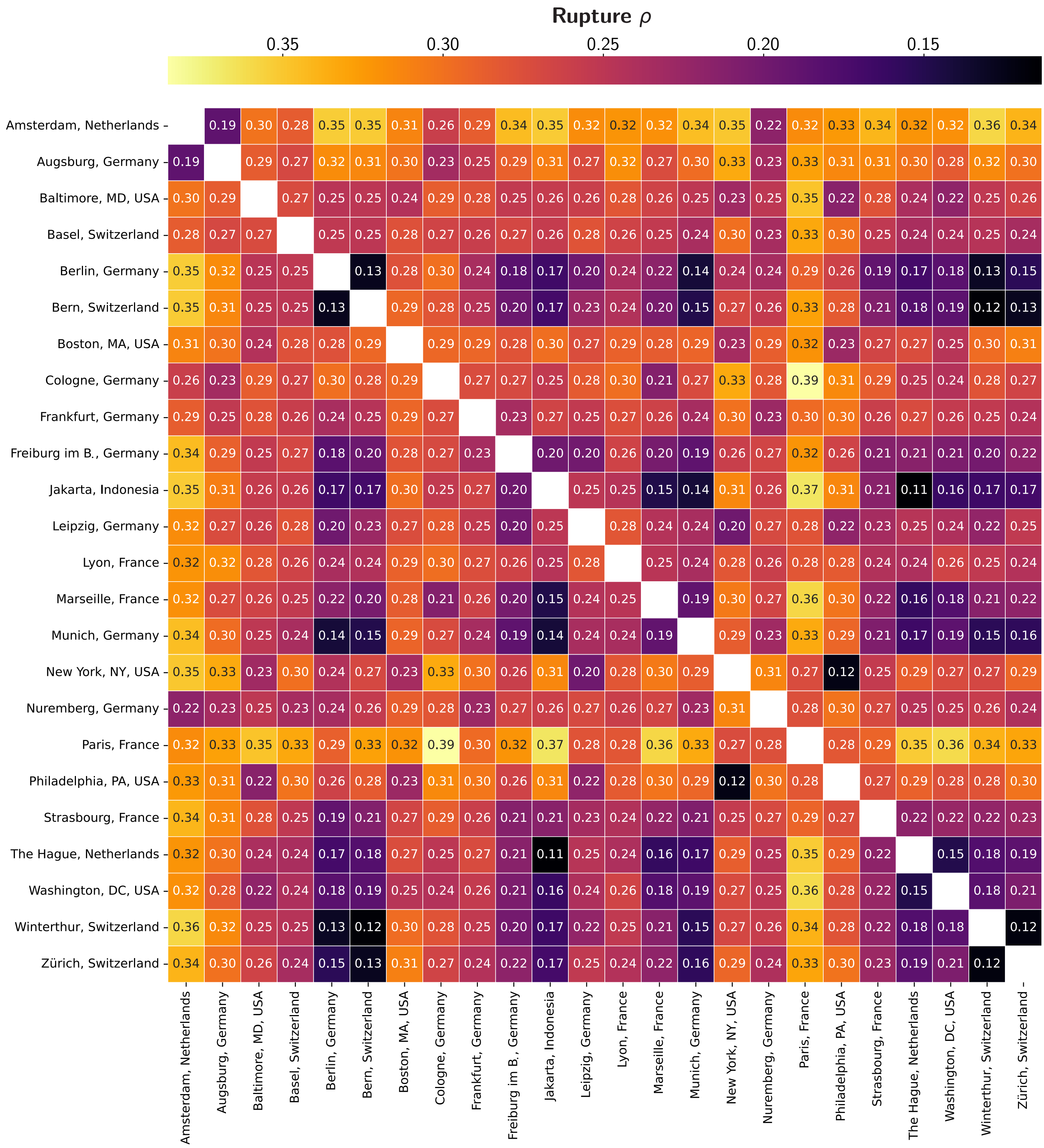

**Figure B19 | Excerpt of the geographic rupture matrix $\Gamma_\rho$.** The matrix reports semiotic rupture between pairs of map production centers. The present excerpt corresponds to the 35 main publication cities. Lower values indicate higher similarity. *E.g. maps published in Bern tend to use signs similar to those produced in Berlin, whereas Paris and Cologne differ.*

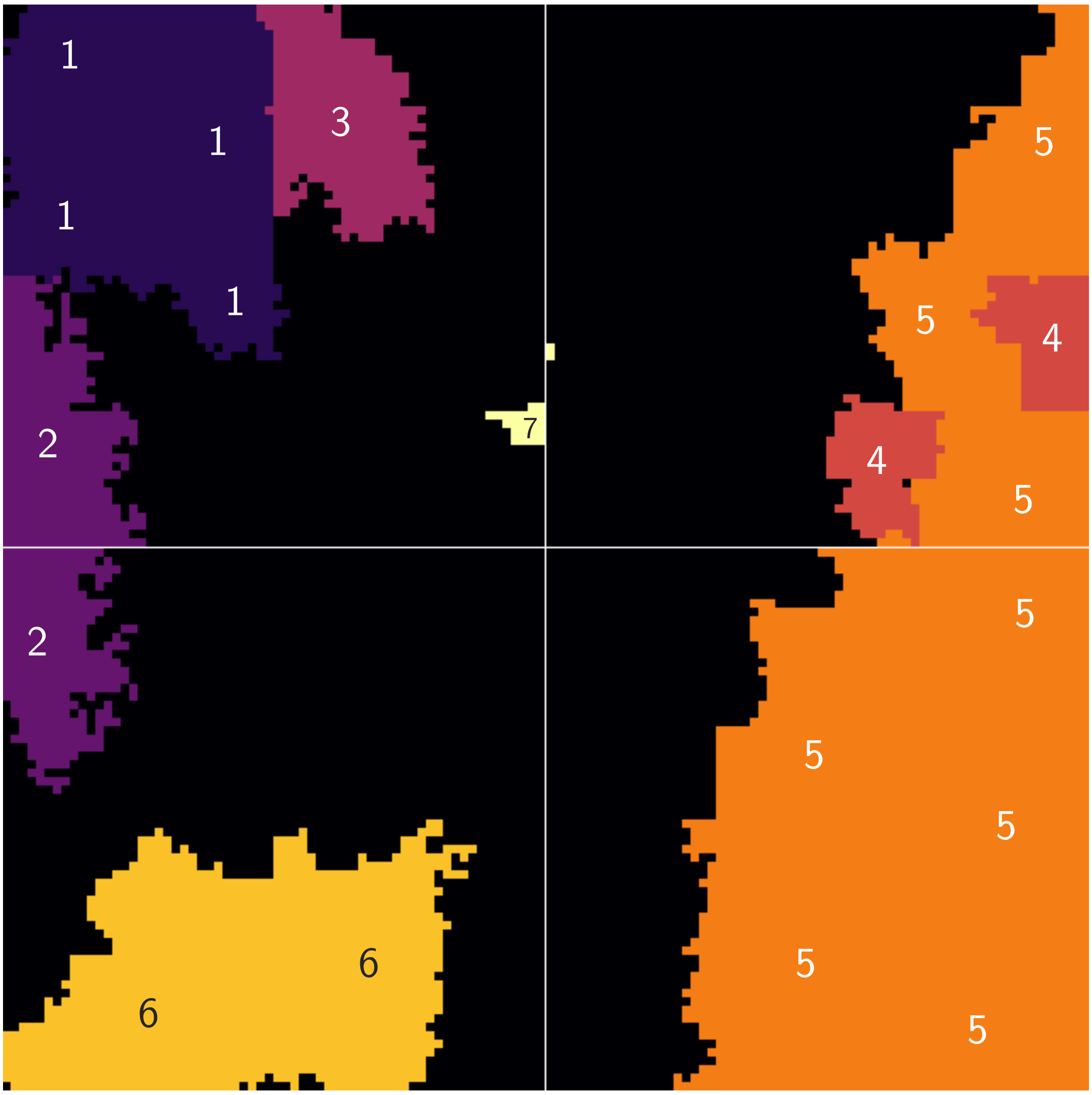

1. magenta 2. lavender 3. purple 4. olive green
5. orange 6. mint green 7. denim blue

**Figure B20 | Accessible version of Figure 15.** With delineation of the color regions referenced in the text.

**Table B1 | City population estimates around 1850.**

| City | Year | Population est. | Source |
|---|---|---|---|
| Amsterdam | 1850 | 223,700 | Wikipedia |
| Augsburg | 1849 | 37,986 | Wikipedia/1849 Census |
| Baltimore | 1850 | 210,646 | (De Bow, 1853) |
| Basel | 1850 | 27,844 | (De Bow, 1853) |
| Berlin | 1849 | 419,000 | Wikipedia/1849 Census |
| Bern | 1850 | 29,670 | (Ritzmann-Blickenstorfer et al., 2012) |
| Boston | 1850 | 136,881 | (De Bow, 1853) |
| Cologne | 1849 | 94,789 | Wikipedia/1849 Census |
| Frankfurt am M. | 1849 | 59,316 | Wikipedia/1849 Census |
| Freiburg im B. | 1849 | 15,298 | Wikipedia/1849 Census |
| Jakarta/Batavia | 1850 | 40,000 | Estimate based on population in 1900 |
| Leipzig | 1849 | 62,374 | Wikipedia/1849 Census |
| Lyon | 1851 | 177,190 | (Le Mée, 1989) |
| Marseille | 1851 | 195,257 | (Le Mée, 1989) |
| Munich | 1849 | 96,396 | Wikipedia/1849 Census |
| New York City | 1850 | 515,547 | (De Bow, 1853) |
| Nuremberg | 1849 | 50,828 | Wikipedia/1849 Census |
| Paris | 1851 | 1,053,262 | (Le Mée, 1989) |
| Philadelphia | 1850 | 340,045 | (De Bow, 1853) |
| Strasbourg | 1851 | 75,565 | (Le Mée, 1989) |
| The Hague | 1850 | 72,000 | Wikipedia |
| Washington D.C. | 1850 | 51,687 | (De Bow, 1853) |
| Winterthur | 1850 | 13,651 | (Ritzmann-Blickenstorfer et al., 2012) |
| Zürich | 1850 | 41,585 | (Ritzmann-Blickenstorfer et al., 2012) |

# Chapter 7

## *The Semiotic Microscope*

While the previous chapter investigated *signs*, it concentrated in fact on a specific case that semiotic theory generally refers to as *point signs*. Point signs, such as pictograms, reduce three-dimensional physical features to a "zero dimension"[1]. In contrast, most geographic features portrayed on maps, like rivers, roads, seas, or continents, are typically represented as *lines* or *polygonal areas* that, contrary to points, signify spatial extension. Theoretically and etymologically, lines and areas can also be considered signs, insofar as they convey meaning. Like points, the representation of lines and areas may be more iconic (e.g., a river colored in blue) or, conversely, more symbolic (e.g., a path depicted as a dotted black line).

Chapter 6 showed that the latent embedding of map signs can effectively represent visual-formal characteristics, such as stylistic proximity across maps produced by the same mapmakers, and semantic content, such as icon categories (e.g. trees, hills, settlements, etc.). However, it did not directly address the relationship *between* form and content, which would have required disentangling both dimensions within the latent vector representation. Map segmentation, discussed in Chapters 4 and 5, precisely provides a *semantic* representation of linear and areal features. The resulting semantic code is independent of feature extraction itself and is not merely derived from the graphical form of signs but rather leverages spatial context.

Thus, the present chapter addresses two gaps left by the previous one. First, the analysis of map signs is extended beyond the specific case of point signs to the figuration of lines and areas, which effectively constitute the most ubiquitous symbols. Second, I complement the analysis of visual representation based on latent embeddings with the integration of semantics. These two steps results in a holistic representation of the cartographic semantic-symbolic system that can account, for instance, for semantic shifts and the emergence of semantics from visual contrasts.

[1] In practice, obviously, the graphical space occupied by point signs on the image is bidimensional. But what makes point signs "zero-dimensional" is that their extension is neglected as part of the sign model. The object they represent is modelled as discrete and extensionless.

## 7.1 The *Map Element* (mapel) as generic operable unit

The first challenge for operationalization lies in defining relevant units for describing map lines and areas. Even more than point signs, map lines and areas can take diverse shapes and dimensions. They may exhibit implicit boundaries, holes, branching, and are often visually overlapping.

The implemented approach is founded on the fragmentation of the map image into small, operable visual *tokens*. Recent advances in computational linguistics have profoundly challenged earlier attempts to model grammar and syntax, owing to the adoption of new methodological paradigms such as the fragmentation of texts into tokens. In computational linguistics, tokens are not necessarily words but rather short operable units of about 4 characters. The process of tokenization enables the digital representation and computational processing of words in neural models. The intuition of tokenization has also been extended to images (Dosovitskiy et al., 2021). Interestingly—and perhaps counterintuitively—fragmentation does not alter the ability for the machine to learn visual syntax or grammar, quite the contrary.

Drawing inspiration from tokenization, the methodology we develop is based on the *fragmentation* of the digitized map image into tiles, or elementary map units (mapels). In our perspective, a *mapel* is an etic unit, analogous to a *phone* for instance[2]. Mapels are considered physical events, specific instances, cartographic utterances, and the realization of a mapping process (Petitpierre et al., 2024b).

[2] Here we understand etic unit, in the sense of semiotics, as defined for instance by Nöth in his *Handbook of semiotics*: "The etic unit, the phone or *morph*, consists of individual and contextual variants of this abstract unit [NB: the emic unit]" (Nöth, 1995, p. 183).

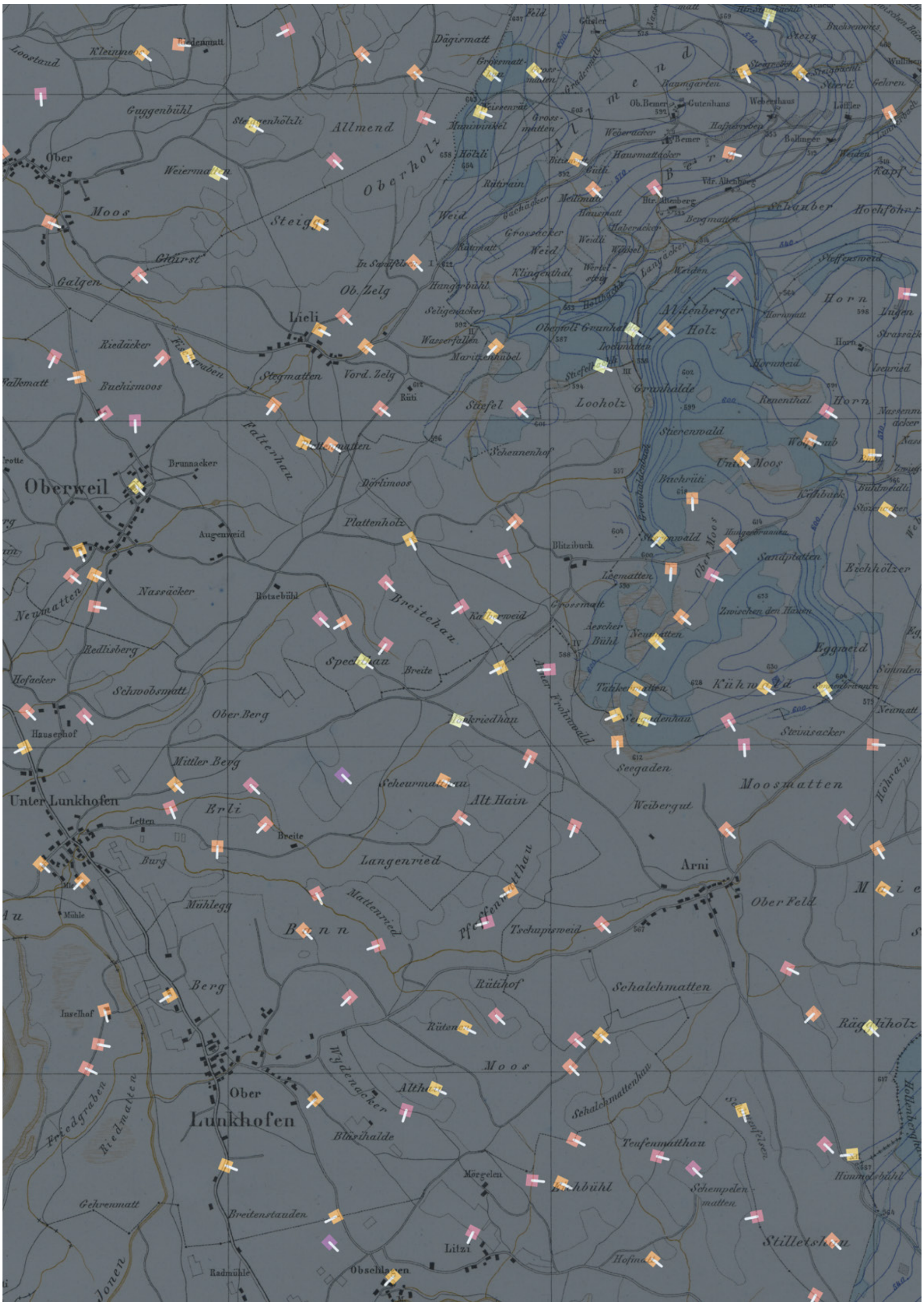

**Figure 1 | Extraction of the mapels.** Excerpt from the map of Birmensdorf, part of the topographical series of the canton of Zürich at a scale of 1:25,000 (1855). Brighter tiles indicate the mapels selected for extraction after sampling. Mapels are positioned at local graphic-load maxima (e.g. centered on lines or other graphical features). Color scale intensity denote greater graphic load. White segments indicate the orientation of each mapel. Image source: H. Enderli, Johannes Wild, J. Graf, and J. Brack. *N° XXI Birmensdorf*, 1855. Topographisches Bureau, Zürich. Chromolithography. 35 × 52 cm. ETH Bibliothek, Rar KA 156:21. doi: 10.3931/e-rara-23700. *Mapels are sampled across the entire map image.*

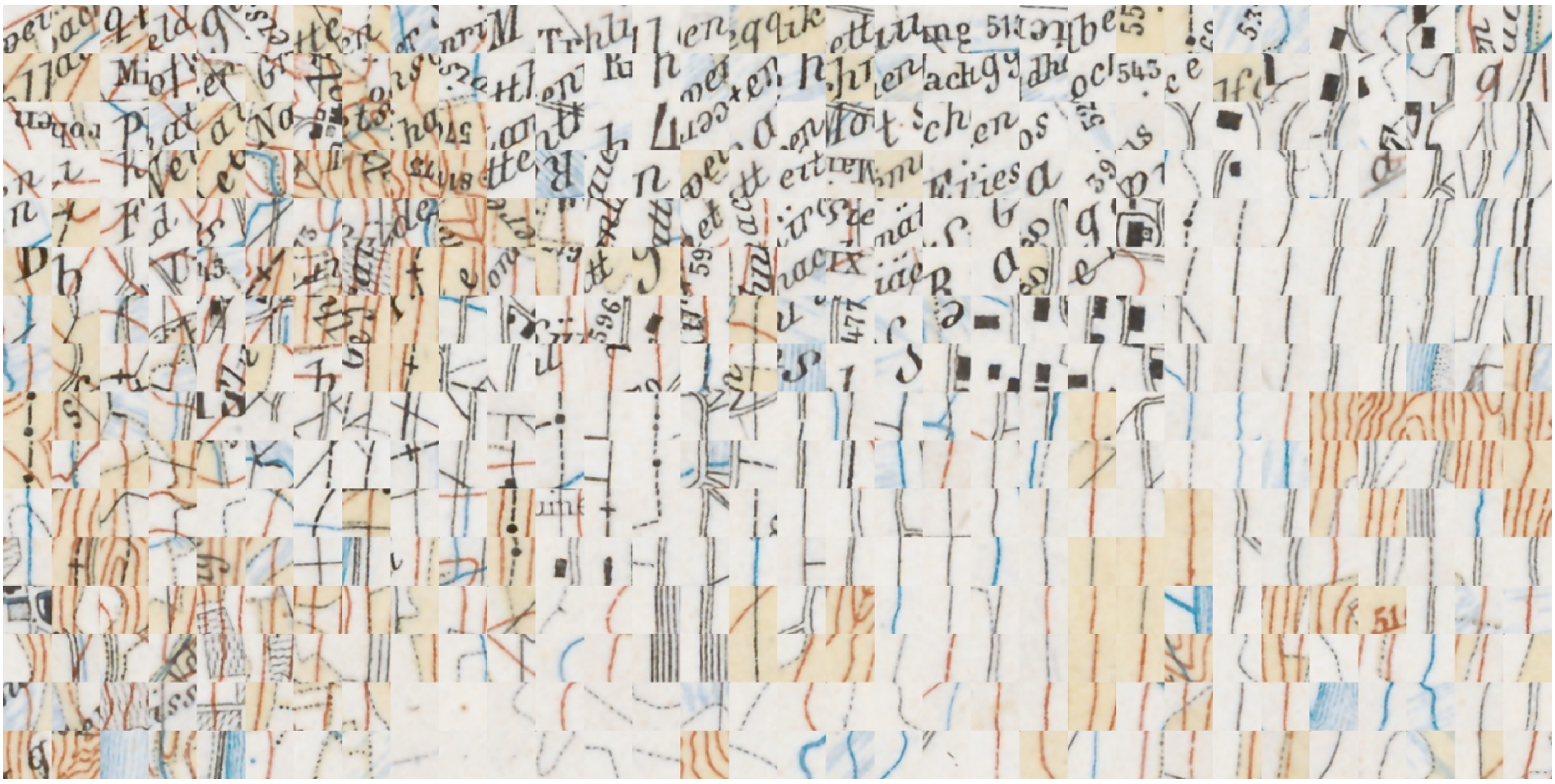

**Figure 2 | Result of the map fragmentation.** 512 mapels extracted from the topographical map of Birmensdorf (Fig. 1). *The sampled subset of mapels provides an operable summary of the map's figuration.*

Concretely, 256 mapels were cut out automatically from each map in the corpus[3]. Each unit has a size of $49 \times 49$, $70 \times 70$, or $98 \times 98$ pixels, which permits to cover a minute area of the map. Together, mapels constitute a summary of map figuration. First, a mild bilateral filter was applied to smooth paper irregularities. The map background, as well as blank areas that are devoid of graphical content are then masked[4]. The presence of graphical content was measured based on graphic map load, an edge density ED2 metric proposed and validated against a human panel by Barvir and Vozenilek (2020).

The initial positions of the 256 mapels were randomly drawn from the two-dimensional maxima of the graphic map load, subject to a buffer distance from the map background, and a minimal distance between positions of 100 pixels (Fig. 1). The principal orientation was then computed, using the Histogram of Oriented Gradient (HOG, Dalal & Triggs, 2005). The resulting angle was stored for information, and the corresponding inverse rotation was applied to each mapel to neutralize its initial orientation.

---

[3] This number of 256 is defined arbitrarily. It is considered sufficient to adequately cover multiple areas of each map image, and account for the overall figuration. Taking the same number of 256 samples on each map results in an equal representation of the figurative content of each map, regardless of the dimensions of the document.

[4] Blank areas all resemble each other, except maybe for some peculiarities of the support. While the presence of blank areas is informative, there is no need to extract them and compute an embedding to study this aspect. Instead, the information on their presence is stored explicitly in a dedicated vector.

In addition to the mapel orientation angle, a set of interpretable features is computed for each mapel, analog to the procedure implemented by Petitpierre et al. (2024b). These features, obtained through conventional computer-vision techniques, include the color histogram; the first three moments of the HLS and CMYK color spaces; the histogram of local binary patterns (LBP), considered a powerful texture descriptor (Ojala et al., 2002); the number of connected components in the binarized image; the average line thickness; the average graphic load; and the maximum response of the Harris corner detection map (Harris et al., 1988). Figure 19, in Section 7.4, demonstrates the visual interpretability of four of these features on a random subset of mapels.

The process results in a manageable representation of map signs. Figure 2 shows a set of 512 mapels extracted from the iconographic map of the Canton of Zürich (Fig. 1). The figure demonstrates that mapel extraction can effectively complement point-sign detection by encoding distinct and complementary visual map features.

## 7.2 Latent representation

### *Methodology*

The process of embedding mapels is similar to the method employed to embed signs in the previous chapter. A specific challenge, however, is that the scale at which mapels lie is not as clearly defined as that of point signs. While point signs are, by definition, of limited extent, lines and areas might cover large portions of a map image. The representation strategy based on mapels aims to capture the figuration of these elements through fragments. Although relatively small mapel sizes presumably capture more elementary and essential visual features, excessively small sizes may fail to capture textural motifs adequately and may produce embeddings that are not descriptive due to graphic noise. Previous studies that focused primarily on the semantic classification of map patches considered patch sizes ranging from 48 × 48 px to 100 × 100 px (Hosseini et al., 2022; Uhl et al., 2018, 2020). Within this range, Petitpierre (2023; 2024b) decided on 50 × 50 px mapels to represent cartographic style, hypothesizing that style resides at the lower end of the range, whereas semantic classification might benefit from a broader spatial context.

One of the first objectives is thus to identify the mapel scale that most effectively captures figurative form while also its relation to semantics. By contrast, the semantic mask is extracted from broader 768 × 768 px contextual patches (cf. Chapter 4). In the following experiments, the protocols will evaluate three candidate sizes: 49 × 49, 70 × 70, and 98 × 98 pixels; all multiples of 7, as required by the modified DINOv2 architecture[5]. Similar to Chapter 6, the informative value of the latent embeddings will be assessed with respect to (1) their ability to discriminate maps produced by distinct communities of map makers and (2) their capacity to differentiate semantics. For the latter task, the target corresponds to the compositional semantic ratio comprising the six classes established in Chapter 4 (built, non-built, water, road network, contours, and background).

[5] N.B. As explained in Chapter 6, the original DINOv2 architecture requires inputs to be multiples of 14. In this research, the weights of the Patcher layer are interpolated to support multiples of 7.

Beyond mapel size, the experiments evaluate the benefits of min–max color normalization, similar to the validation tests implemented in Chapter 6. Min–max normalization aims to attenuate the impact of physical support, or digitization artifacts, on color. Because this operation does not appear to introduce a marked degradative effect in the present case, and as the target extent corresponds to the entire mapel, the inclusion of an adapter module appears unnecessary.

The descriptive power of DINOv2 features is also benchmarked against a set of interpretable features derived via traditional computer vision methods: color moments, local binary pattern descriptors of texture, line thickness, connected components, graphic load, and Harris corner detection response. For this trial, the smaller patch size considered is $50 \times 50$, since the Harris corner detector requires the image patch to be a multiple of 2.

### *Test 1: Communities*

As anticipated in Chapter 6 (Section 6.4), one way to evaluate the descriptive power of latent embeddings is to assess their ability to identify similar patterns, given the expectation that such patterns are replicated within a community of mapmakers, defined by collaboration relationships. The aptitude to retrieve such relationships is expressed by $\rho_{\mathcal{B}}$, the correlation between pairwise map distance and the graph of community relationships (Chap. 6, Eq. 2). The pairwise map distance between two maps is computed as the average minimal distance between each mapel and its closest neighbor in the other map.

The experiment is repeated 10 times, with 17 batches $\mathcal{B}$ of 50 maps each, sampled from a subset $\mathcal{C}_{\mathcal{B}}$ of 10 communities $\mathcal{C}_{\mathcal{B}} \subset \mathcal{C}$. For each map, we consider a maximum of $N_m \leq 256$ mapels.

**Table 1 | Performance of the stylometric experiment**. The computation of $\rho_{\mathcal{B}}$ is detailed in Eq. 2 of Chapter 6. The asterisk (*) indicates that the shape has been adjusted to be a multiple of 7, a requirement of the adapted DINOv2 architecture.

| Experiment | Mapel size | $\rho_{\mathcal{B}} \pm$ 95% CI |
|---|---|---|
| DINOv2 | $49 \times 49$* | $0.345 \pm 0.02$ |
| DINOv2 + min-max norm. | $49 \times 49$* | **$0.445 \pm 0.02$** |
| CV features | $50 \times 50$ | $0.398 \pm 0.02$ |
| DINOv2 | $70 \times 70$ | $0.331 \pm 0.02$ |
| DINOv2 + min-max norm. | $70 \times 70$ | $0.417 \pm 0.02$ |
| CV features | $70 \times 70$ | $0.391 \pm 0.02$ |
| DINOv2 | $98 \times 98$ | $0.317 \pm 0.02$ |
| DINOv2 + min-max norm. | $98 \times 98$ | $0.423 \pm 0.02$ |
| CV features | $98 \times 98$ | $0.389 \pm 0.02$ |

The results of the experiment are presented in Table 1. Trials consistently achieved higher performance when using the smallest mapel size (49 × 49 or 50 × 50) compared to larger configurations. DINOv2 features appear significantly more descriptive than benchmark features derived via classical computer vision (CV) methods, particularly after min–max normalization. It should be noted that the same normalization is also used as a pre-processing step to the computation of CV features. The results of Table 1 are directly comparable to those reported in Chapter 6, Table 1, as both employ identical metrics. Whereas the highest value of $\rho_{\mathcal{B}}$ for point signs was 0.324 ± 0.06, mapels emerge as more powerful descriptors of authorship and style, attaining a maximum value of 0.445 ± 0.02.

### *Test 2: Semantic classification*

The second experiment evaluates the descriptive power of mapels in a semantic regression task. Although spatially accurate recognition of map semantics usually requires the consideration of a wider image context, map patches akin to mapels are also sometimes used for semantic classification (Hosseini et al., 2022; Uhl et al., 2017). Even on heterogeneous sets of maps, one might expect dominant graphical patterns to recur in similar semantic contexts. For instance, waterlines or the color blue typically signify water, while the color green usually denotes non-built areas and vegetation. Counterexample exists. Nevertheless, we should be able to predict semantics better than chance from mapel embeddings.

One peculiarity of the present setting is that semantics are extracted from semantic segmentation masks, which, they do not quite correspond to discrete categories but rather to pixel masks whose size matches that of the extracted mapels. Although accounting for semantic shapes within each patch would require defining complex target features and might interfere with the form plane, collapsing the masks into single categorical labels would neglect most of their informational potential. A compromise between these two options is the reduction of semantic fragments to compositional semantic ratios that record the relative frequency of each semantic class within a mapel.

Compositional data are expressed as proportions that sum to one. It means that they do not reside in a standard Euclidean space but in the *simplex*. This constraint implies that the value of each ratio in a composition is not independent from the values of the other ratios. Equivalently, compositional data are redundant: if five of six ratios are known, the sixth can be inferred. To eliminate this effect known as *multicollinearity*, one pertinent approach is to transform compositional data $x$ to lower-dimensional, orthonormal, vectors using isometric log-ratio transformation ($ilr$):

$$z = ilr(x) = [\langle x, e_1 \rangle, \dots, \langle x, e_{D-1} \rangle] \in \mathbb{R}^5$$

The dataset was randomly divided into training and validation sets. An ordinary least-squares regression was fitted on $z$, with 10 repetitions. The validation error was quantified by the Aitchison distance $d_{\mathcal{A}}$, corresponding to the Euclidean distance between the ilr-transformed vector $z$ and the prediction of linear regression $\hat{z}$: $d_{\mathcal{A}} = \|z - \hat{z}\|$.

The results of the experiment are reported in Table 2. As in the previous experiment, performance was consistently better when employing smaller mapels ($49 \times 49/50 \times 50$) compared to larger ones. The descriptive power of DINOv2 appears substantial, relative to CV features, irrespective of min–max normalization preprocessing. To provide a more interpretable estimate of performance, the mean absolute error, obtained after reverse transformation of the prediction estimates in the simplex is 0.108. This value is only marginally lower than that of the dummy solution (raw class averages) which would yield a mean absolute error of 0.121.

**Table 2 | Performance of the semantic classification experiment**. The semantic classification was performed by fitting a linear regression model to a 20% training sample (ca. 5,048,277 signs). The target was the semantic composition, transformed using Aitchison's isometric log ratio. Each experiment was repeated 10 times, and performance was measured on a distinct 10% validation sample (ca. 2,524,138 signs); 95% CI denotes the confidence interval of the mean. $d_{\mathcal{A}}$ is the Aitchison distance (lower is better). The asterisk (*) indicates that the shape was adjusted to be a multiple of 7, as required by the adapted DINOv2 architecture.

| Experiment | Mapel shape | $d_{\mathcal{A}} \pm$ 95% CI |
|---|---|---|
| DINOv2 | 49 × 49* | **4.87 ± 0.01** |
| DINOv2 + min-max norm. | 49 × 49* | **4.88 ± 0.01** |
| CV features | 50 × 50 | 5.08 ± 0.01 |
| DINOv2 | 70 × 70 | 5.19 ± 0.01 |
| DINOv2 + min-max norm. | 70 × 70 | 5.17 ± 0.01 |
| CV features | 70 × 70 | 5.42 ± 0.01 |
| DINOv2 | 98 × 98 | 5.49 ± 0.01 |
| DINOv2 + min-max norm. | 98 × 98 | 5.46 ± 0.01 |
| CV features | 98 × 98 | 5.70 ± 0.01 |

### *Synthesis*

Given that $49 \times 49$ mapels appear more informative in both experiments, this fragment size is selected for all subsequent analyses. Furthermore, the DINOv2 encoder, paired with min–max normalization, is adopted, since it performs best in both experiments and even better when combined with preprocessing in the first experiment.

In both experiments, but more prominently in the second, mapel size appears to be a key variable. Although larger mapels practically contain more visual information and are additionally closer in size to the images on which DINOv2 was originally trained on [6], smaller mapels appear systematically more informative. Because neither model architecture nor raw information count can explain the higher performance of smaller mapel sizes, this suggests that the scale is significant from the viewpoint of the document, cartographic figuration, or the average scale at which essential visual patterns are effectively replicated.

In this respect, it is worth underlining that baseline CV features exhibited high performance in both tests. This outcome suggests that, within mapels, comparatively simple, interpretable features, such as color moments, graphic load, and texture descriptors are pertinent to the analysis of cartographic figuration. Section 7.4 will leverage this interpretability to characterize observable graphical evolutions, based on CV features.

## 7.3 The mosaic of map elements

### *Representation as a semantic-symbolic system*

As anticipated, one limitation of relying solely on visual embeddings to analyze map semiotics is the difficulty of disentangling form from content. Consequently, this chapter introduces the semantic composition of each mapel as a conditional parameter for analysis. Compared with an approach that considers embeddings alone, the present strategy may detect complex semiotic evolutions, such as the semantic shifts that arise when a sign form changes meaning, for instance.

In this perspective, it is not only fitting to examine the frequency of map signs but also their relative occurrence as a function of map semantics. The aim, for instance, is to differentiate cases in which a set of wavy black lines denotes coastlines, signifying the presence of the sea, from cases in which the same set of lines depict a high-relief, non-built area, e.g., hill contours. Specifically, the relative frequency of signs should be analyzed conditional on fixed semantic states. The approach also ought to account for graphic delimiters and semantic combinations. For example, the visual form of a mapel representing the boundary between a continent and the sea is not necessarily halfway between a mapel covered entirely by non-built land and one covered entirely by water: limits adopt their own, distinctive visual forms. This dimension is particularly relevant in the present context, because mapels focus on areas of high graphic load, which frequently coincide with boundaries, due to the visual saliency of delimiters.

[6] DINOv2 model was trained on $518 \times 518$ image samples during pre-training and $224 \times 224$ during finetuning. Considering that the Patcher layer was downsampled by a factor 2, as explained in Chapter 6, the present architecture would be expected to perform better on image sizes around $112 \times 112$, which is not confirmed.

As such, the mapel mosaic is considered under eight distinct compositional semantic modes:

1. Primarily built (> 90%)
2. Primarily non-built (> 90%)
3. Primarily water (> 90%)
4. Primarily road network (> 90%)
5. Part built, part non-built (> 30% each)
6. Part water, part non-built (> 30% each)
7. Part road network, part non-built (> 30% each)
8. Presence of a boundary (contours > 4%)

The boundaries with non-built are selected because of their comparatively higher frequency (the non-built class is the most frequent class, as outlined in Chapter 4). Adding rarer modes would result in higher variability among strata. It should also be noted that the modes are not mutually exclusive. For instance, the 8$^{th}$ compositional mode typically overlaps with the 5$^{th}$, 6$^{th}$, and 7$^{th}$ compositional modes. Furthermore, it encompasses boundaries within the same semantic class.

Figure 3 shows characteristic exemplars for each compositional semantic mode.

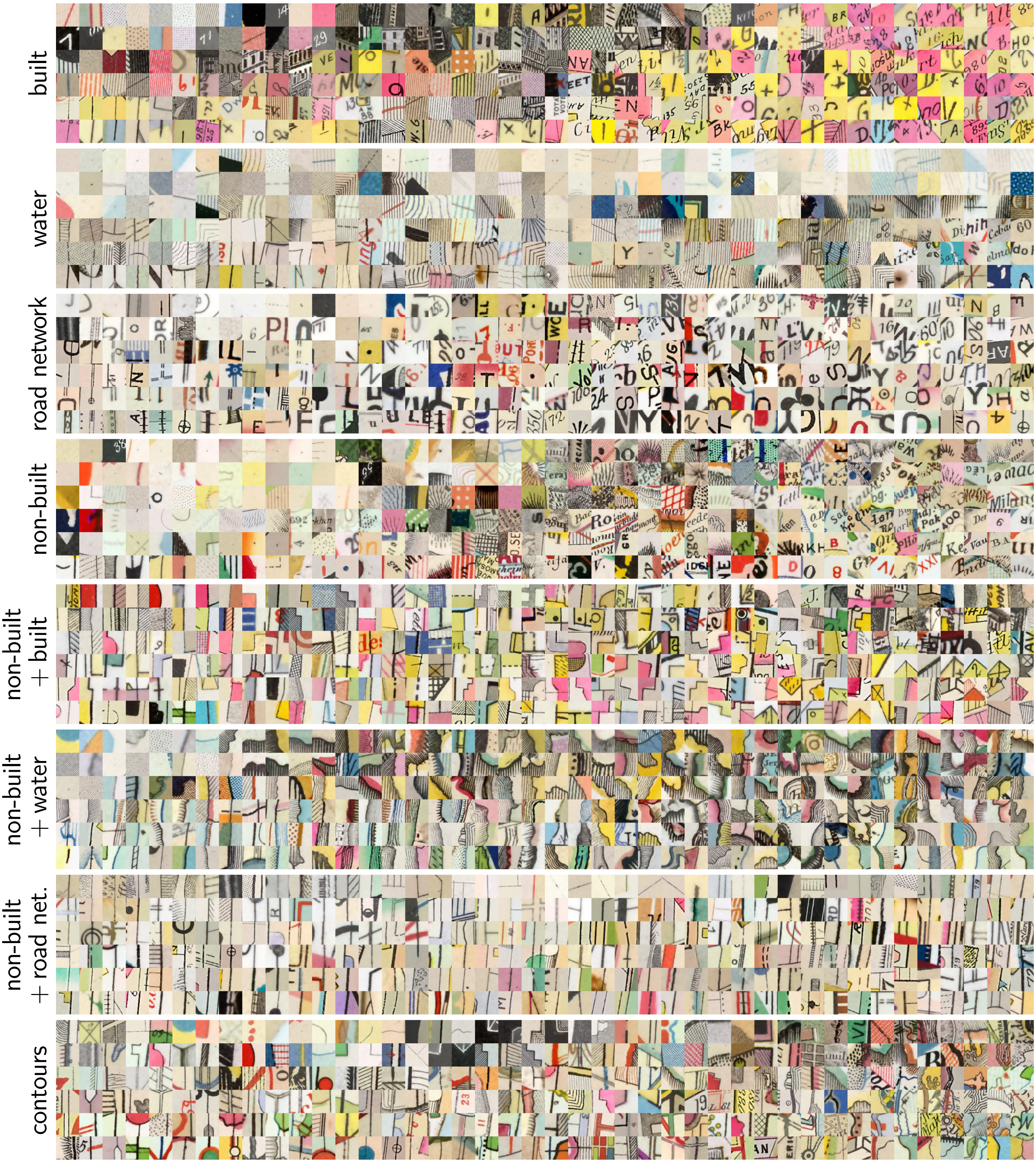

**Figure 3 | Representative mapels for each semantic composition**. Only the 8 semantic compositions considered are represented. *The eight modes are visually distinctive, indicating their potential to discriminate semantic contexts.*

### *Mapel mosaic*

The mapel mosaic was constructed using the same approach as the sign mosaic in Chapter 6. The 25,241,385 mapels extracted with the methodology described in Section 7.1 were first pre-clustered into 4,096 clusters using mini-batch k-means. Subsequently, a Gaussian mixture model reclustering step was applied to integrate density awareness into the clustering process. For each cluster, the mapel closest to the cluster center was designated as the exemplar. The reduced number of clusters is motivated by the intention to maintain an informational density comparable to that observed in the previous chapter: although the number of mapels is more than one and a half times the number of point signs considered for clustering, the mapel mosaic is analyzed under eight distinct compositional semantic modes. Thus, dividing the number of clusters by four maintains the same informational density, that is, approximately the same number of map elements or signs per cluster. In addition, greater symbolic simplicity is expected for mapels, compared to point signs, potentially leading to more consistent clustering. Figure A2 in the Appendix displays representative mapel clusters, whose internal consistency seems particularly high, compared to point-sign clusters. The resulting mapel mosaic is presented in Figure 4. It will be used for subsequent analyses and discussion.

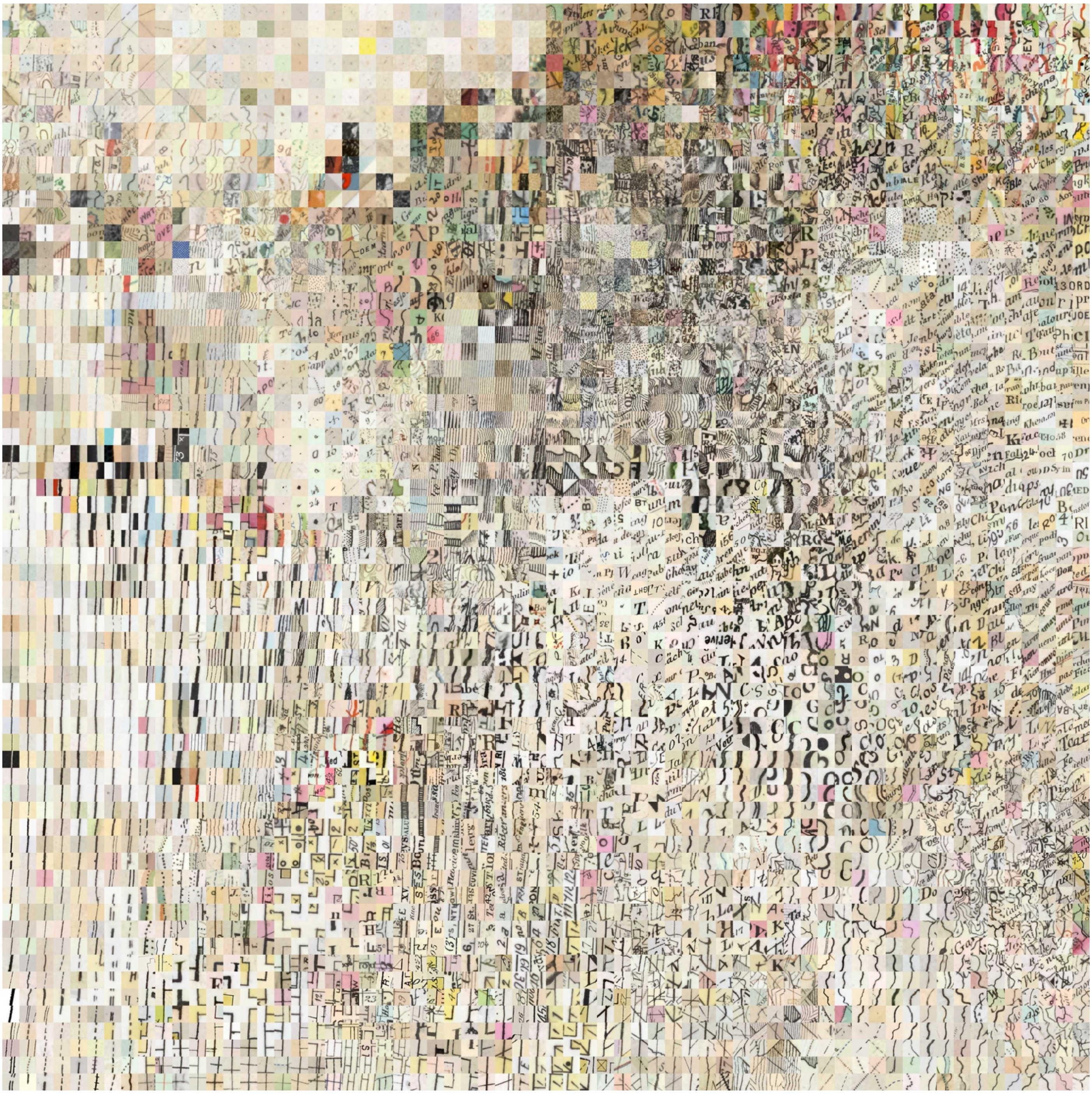

**Figure 4 | Mapel mosaic.** Representing the exemplars of each of the 4,096 mapel clusters. Exemplars are spatialized by t-distributed stochastic neighbor embedding (t-SNE). *The mosaic represents the diversity of map figuration.*

### *Phylogenic mosaic*

The phylogenic mosaic (Fig. 5) provides a complementary visualization of the embedding space, represented in the mapel mosaic. A color-accessible version of this figure is available in the Appendix (Fig. A3). The semiotic space should again be construed as a hierarchical rather than a horizontal structure. Starting from the top of the mosaic and proceeding clockwise, the blue region contains small dots. Some of these are actual dots—slightly faded by the bilateral filtering. Others derive from paper grain and correspond de facto to blank areas. The lavender region chiefly comprises maps with colored backgrounds or high graphic-load motifs. Just below it, and not quite distinct, the deep-purple regions encompass more textured mapels, typically straight hatchings, curved hatchings (e.g., waterlines), and dotted patterns.

Proceeding to the top-right corner, the light purple region contains boundaries signified as dotted lines with color emphasis, dotted lines, track lines with perpendicular tick marks, wavy parallel lines denoting riverways, colored single- or double-lines signifying roads, and diverse multicolor printed maps. Just below, dotted lines and dotted patterns also appear in the tiny turquoise region.

The right part of the mosaic comprises numerous text mapels that adopt primarily slanted typographies with serifs. These mapels are grouped within a large mint green region. Mapels from that area are visually similar to those spatialized in the fuchsia region located nearer the center. By contrast, the intervening pale pink region, in between, displays larger uppercase typographies that are often only partially contained within the mapel extents and intermingle with wide curved lines. Other wavy black lines—generally signifying riverways—appear in the sage green, salmon, and flax regions, extending toward the bottom-right corner of the mosaic. The small moss green region contains dashed lines.

Proceeding clockwise toward the bottom of the mosaic, the terracotta region displays text bounded by one or two lines. These mapels may typically correspond to street labels. They employ diverse typographies, with or without serifs, sometimes slanted but not as often as observed in the fuchsia and mint green regions. The blood orange and cherry regions exhibit various corner configurations, right-angled or oblique intersections, and brick-like motifs.

The entire left side of the mosaic is filled with distinct line types, such as dashed lines (brown ), dotted lines (muddy cherry ), double lines (kaki ), and thicker (burgundy ) or thinner lines (teal ). The mosaic also features color boundaries and spaced hatching patterns, as well as waterlines, nearer the center. Other patterns, clearly visible yet occupying areas not delineated by distinctive hue in the phylogenic mosaic include terrain contours, hachures, double lines with hatched intervals, and negative linework (e.g., blueprints).

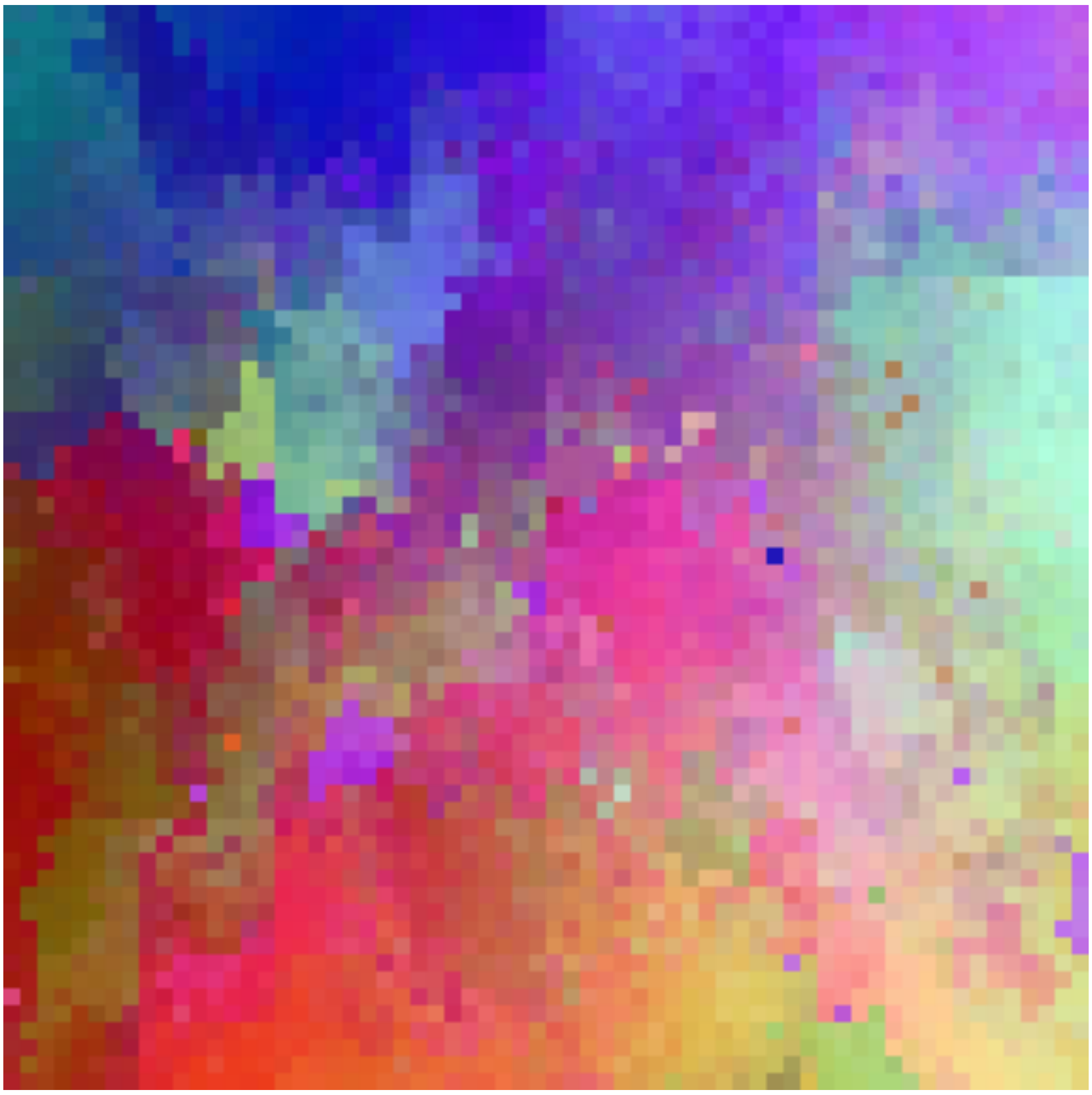

**Figure 5 | Phylogenic mosaic** representing the exemplars of each of the 4,096 sign clusters. Exemplars are spatialized with t-SNE. The values of the red, green, blue (RGB) color channels reflect an independent dimensionality reduction based on UMAP. A color-accessible version of this Figure with clear distinction of the regions mentioned in the text, is provided in Fig. A3, in the Appendix. *The continuity of color hues indicates that the space of signs is structured hierarchically.*

## *Form classes*

The phylogenic mosaic highlights the hierarchical structure of visual form. By exposing the most salient distinctions among sign forms, it enables the derivation of higher-level mapel form classes, such as hatched and dotted textures, waterlines, terrain contours, and thin, thick, dotted, or dashed lines. These *form classes* function as higher-level aggregates within the more detailed mapel mosaic. Examining these classes complements the study of the evolution of individual mapel clusters by providing less granular yet more intelligible results.

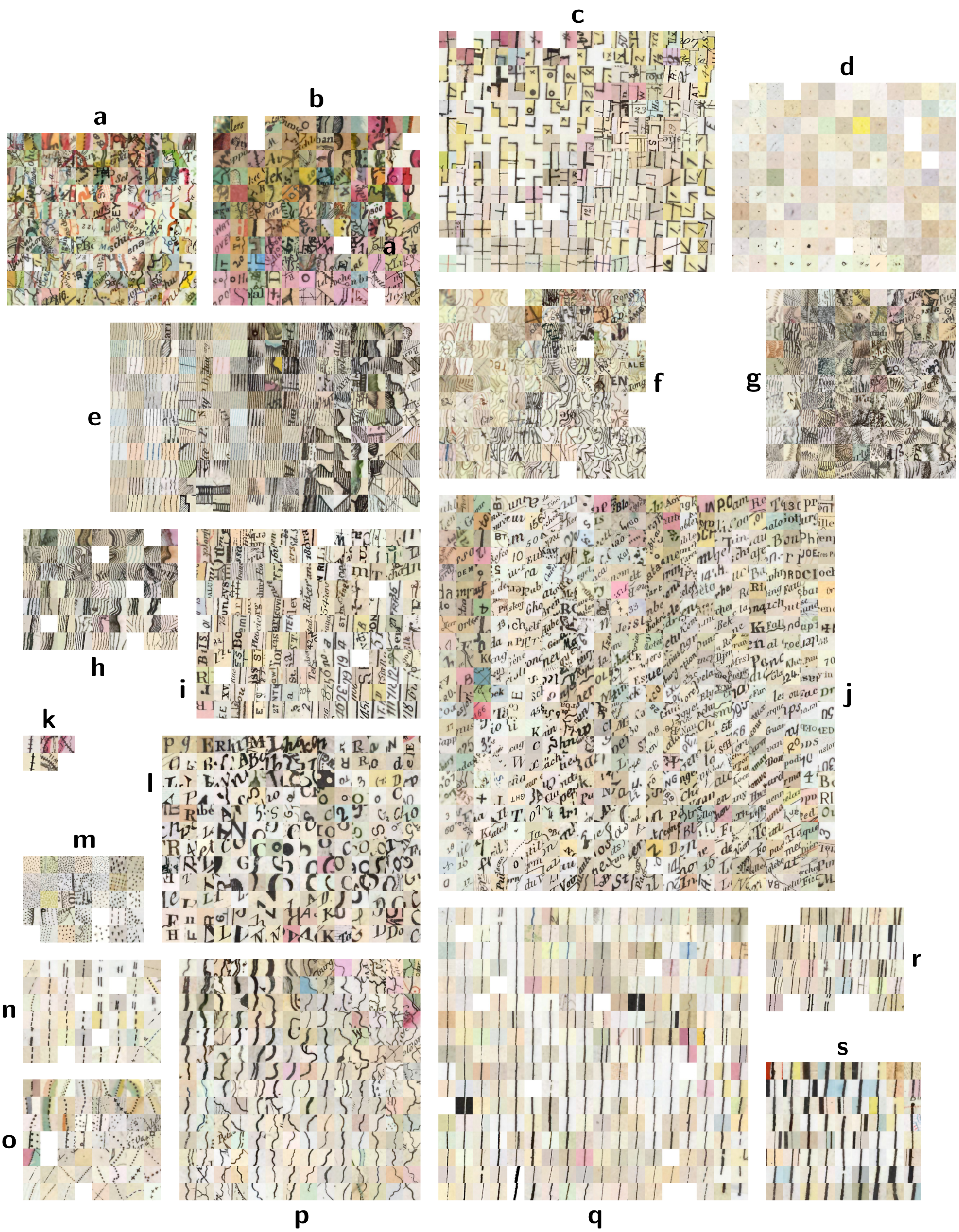

**Figure 6 | Form class exemplars. a)** multicolor, **b)** colored background, **c)** corners, **d)** blank or dots, **e)** hatchings, **f)** terrain contours, **g)** hachures, **h)** waterlines, **i)** directional text, **j)** text, **k)** track lines or ticked lines, **l)** large texts, **m)** dotted pattern, **n)** dashed lines, **o)** dotted lines, **p)** wavy lines, **q)** regular lines, **r)** double lines, **s)** thick lines. *Distinct spaces of the phylogenic mosaic (Fig. 5) correspond to separate visual forms.*

In total, 19 form classes were identified, corresponding to 2,681 mapel clusters out of 4,096. Form classes are closely connected to the phylogenic regions described in the previous section. They might, however, regroup several regions, such as the mint green and fuchsia regions, both corresponding to visually kindred text mapels. Conversely, a single region may also be subdivided. For instance, the deep purple region regroups visually distinct forms, such as regular hatchings, waterlines, and hachures. The annotation of mapel form classes is semi-automated, leveraging the embedding space and its phylogenic dimensionality reduction. Qualitative assessment of visual coherence permits to identify the 19 form classes, illustrated by their exemplars in Figure 6.

## 7.4 Conditional analysis of the semantic-symbolic system

The previous sections introduced four fundamental methodological building blocks that enable the analysis of the semantic-symbolic system of maps. First, *mapels*, defined as empirical and operable cartographic instances. Second, exemplars, which help make the semiotic space graspable and intelligible. Third, form classes, which constitute higher-level categories of visual form and, by extension, of the semiotic space. Fourth, compositional semantic modes, defined as discrete semantic ratios under which the differential expression of mapel clusters is examined. In this section, these four building blocks are used to investigate the conditional expression of the semantic-symbolic system of cartography. Specifically, two conditional variables are retained for analysis: chronology and cartographic scale.

### *Temporal ruptures & semiotic adaptation to the cultural system*

The coefficient of rupture $\rho$ was introduced in Chapter 6 to quantify the discrepancy between two sign frequency tables. It can be generalized to the semantic–symbolic case by treating each conditional semantic mode as a parallel framework of expression. If the rupture coefficient between two frequency tables in the discrete semantic mode $k$ is noted $\rho_k$, then the overall coefficient of rupture $\rho$ between two strata is equal to the mean of the semantic coefficients $\rho_k$.

Figure 7 represents the differential change in the coefficient of rupture $\rho$ by year of publication. The chronological profile of rupture resembles that observed in the analysis of point signs presented in Chapter 6, whose peaks are indicated by asterisks (*) in Figure 7. The former profile was characterized by a primary peak, around 1789, and three secondary peaks, the last of which was further subdivided. The present profile, by contrast, exhibits four to six peaks of comparable magnitude. The first peak occurs between 1615 and 1634. The second swell comprises two distinct peaks, one around 1732–1743 and, more prominently, one between 1778 and 1788. The third swell consists of three closely spaced peaks, around 1865, 1877, and 1887. This last peak is less marked than the preceding ones and is thus not considered for the subsequent analyses. Between the three major swells, two periods of slower change—characterized by a lower coefficient of rupture—are observed: 1661–1705 and 1837–1852.

When semantic modes are disregarded, i.e., when the coefficient of rupture is computed solely from relative mapel frequencies, the curve—now represented by a dashed red line—barely changes. This outcome suggests that semantic shifts are rarely found in successive time strata. Such stability is not unexpected, for unlike languages, the maps examined exist solely in written form. This provides them a potentially long "lifespan", during which they continue to influence and structure cartographic conventions of representation[7]. While changes and innovations can be introduced, semantic *inversions* would constitute inconsistencies within the semiotic system. From an evolutionary perspective, *conflicting variations*, such as semantic shifts, tend to be selected against because they are *unadapted to their cultural environment.* This environment is constituted by the semiotic system itself, which, in order to function, must maintain *distinctive* figurations for representing different meanings. Semiotic distinction is not only required within each individual map: it is *structurally* enforced by cartographic conventions and the broader social system of map making. As such, semantic shifts involve displacing the equilibrium of the entire semiotic system, an intervention that demands considerable effort or influence.

An analogy with language might help illustrate this requirement. While a speaker can introduce a slight semantic variation in the use of a familiar word, employing it with a radically different signification would likely impair the effectiveness of the message. Other speakers would be unlikely to replicate the novel word–sign, and the originator would eventually abandon it as well. By contrast, the meaning of an archaic or infrequently used term can sometimes be altered with minimal repercussions. For example, a modified form of the ancient Greek word δημοκρατία (*demokratía*) came to designate the modern political system —*démocratie*— in which power is exercised by elected representatives, even though the original signification of the term was very different[8]. Reversing the meaning of widely used English word would likely prove more difficult. If achievable at all, such a shift would require significant resources and influence.

---

[7] As discussed in Chapter 2, the influence of some maps can last for more than a century, as in the case of the chorography of Muscovy by Sigmund von Herberstein, reused 180 years later by Guillaume Delisle. An even more compelling example, of course, is the *Geography* of Ptolemy (2nd century), which was still considered a reference at the end of the 15th century.

[8] In Ancient Greece, the concept of δημοκρατία (*demokratía*), literally "the rule of the people" was defined in contrast to αριστοκρατία (*artistokratía*), "the rule of the best", and ὀλιγαρχία (*oligarchía*), "the rule of few". While *democratía* involved direct participation of the citizens, or eventually the random selection of representatives with rapid turnover of mandates, *artistokratía* or *oligarchía* entailed the selection of the few "best" citizens, typically by election. Aristotle, for instance, opposed the two concepts: "the appointment of magistrates by lot is thought to be democratical, and the election of them oligarchical" (Aristotle, 330 C.E., p. 1294b; Jowett, 1943). At the beginning of the 18th century, Montesquieu still uses the term *democratie* in this sense (Cervera-Marzal & Dubigeon, 2013). The meaning of the root was inverted shortly after, during the French Revolution, when it took its modern and opposite signification—a political system characterized by the rule of a few *elected* representatives (Sintomer, 2007).

Arguably, semiotic variation in cartography is primarily characterized by the introduction of *new sign forms* into the system, rather than semantic shifts. New forms may emerge through figurative innovation and/or the introduction of new printing technologies (e.g., color printing, mechanical rulers). Altering the meaning of existing, actively used signs is difficult, and therefore rare or delayed, because semiotic evolution is constrained by the *cultural system* in which it occurs and by prevailing *conventions*.

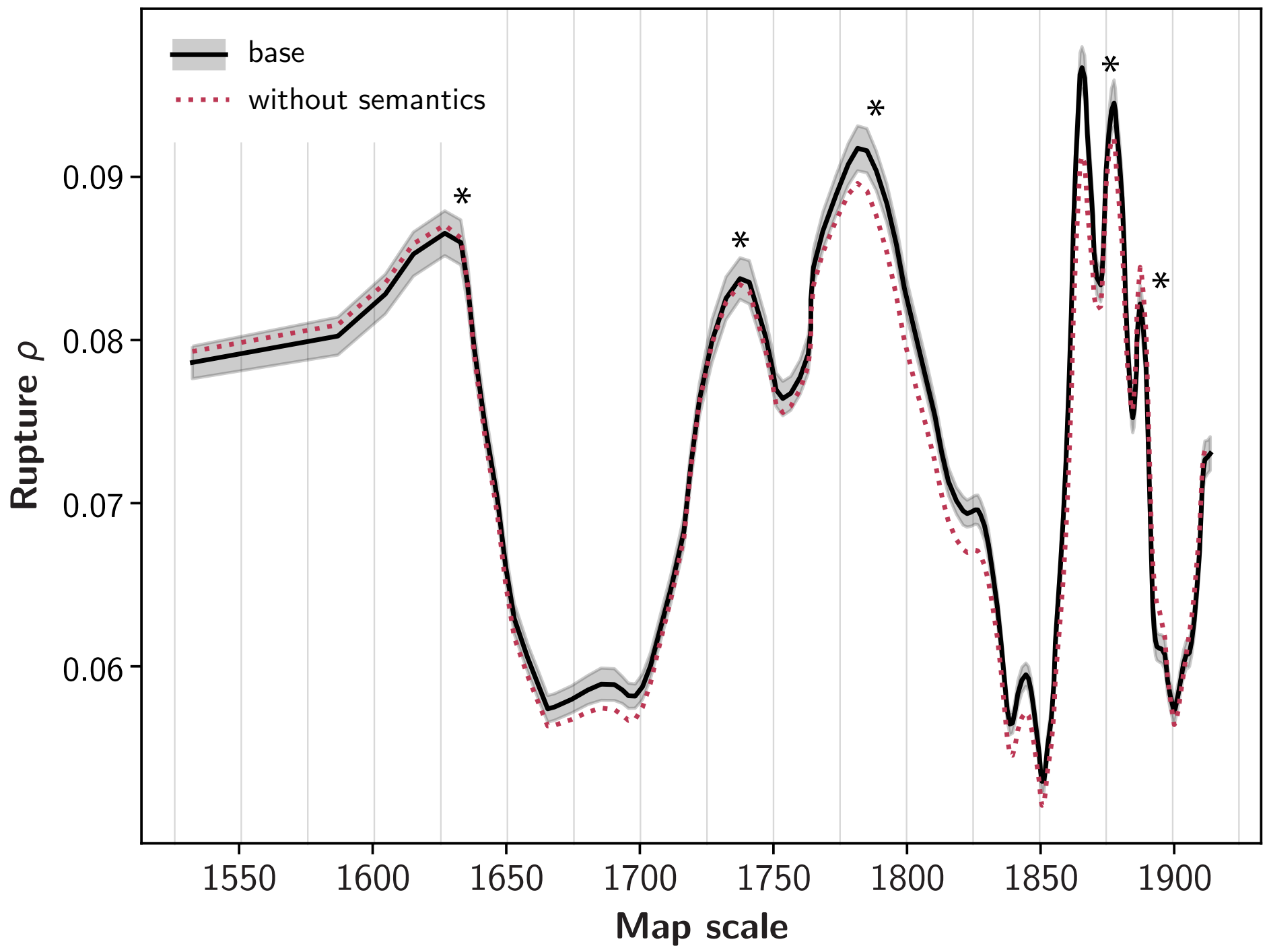

**Figure 7 | Evolution of the coefficient of rupture $\rho$ by year of publication**. The coefficient is computed using a sliding window (200 steps) and two time strata, each corresponding to 5% of the dataset and overlapping by 50%. The shaded area denotes the 95% confidence interval of the mean. The red dotted line depicts the coefficient of rupture when conditional semantic modes are disregarded (i.e., when it derives only from mapel cluster frequencies). Asterisks (*) indicate the peaks identified in the analysis of point sign ruptures (Chap. 6, Fig. 22). *The most marked shifts in the signs used occurred around ca. 1625, 1785, and 1865.*

From this point of view, it is not surprising either to see rupture peaks—corresponding to *evolutionary events*—align with those observed in the analysis of point signs in Chapter 6. When point signs, like icons or symbols change, the entire semiotic system is affected and eventually transitions to a new equilibrium. The magnitude of the changes involved, however, can differ between point signs and mapels, with some events impacting one or the other aspect of cartographic representation more significantly.

### *Distinguishing map scales*

The profile of rupture by map scale (Fig. 8) differs markedly from that reported in Chapter 6. Whereas the principal turning point for icons and symbols was located slightly before 1:100,000, around 1:93,500, the corresponding peak is barely significant in the analysis of mapels. Contrastively, rupture seems to be particularly high at larger, more detailed, map scales. The most substantial change is detected at scales close to 1:8,000; the associated peak exhibits comparatively low kurtosis, resulting in a wide confidence interval spanning from 1:3,000 to 1:10,000. A second, sharper rupture peak is observed just before the 1:25,000 scale. Lastly, a less prominent peak is observed near 1:750,000.

These empirical moments of rupture, grounded in map figuration and cartographic semiotics highlight three distinct scale classes. First, large-scale maps correspond to what are commonly called *plans*. They typically encompass plats, cadasters, and detailed town plans. Second, the medium map scale covers the ranges employed in topographic mapping, such as 1:25,000, 1:50,000, 1:100,000, and 1:250,000. These scales are, for instance frequently used by public agencies for land surveys or military cartography. Larger map scales include regional or country maps, as well as world maps, which are habitually drawn at scales ranging from approximately 1:1,000,000 to 1:50,000,000. For these documents, however, the concept of scale is only partially pertinent; map distances can diverge considerably within the same image owing to Earth curvature and projection systems. The semiotic contrast between medium- and large-scale maps appears less marked than the difference between small- and medium-scale maps. This result probably constitutes the most empirical attempt to infer classes of maps by scale. Indeed, the terms *small*-, *medium*-, and *large-scale* are often used in the literature without any clear consensus on the ranges they denote.

In contrast to the chronological analysis, the rupture coefficient per map scale differs significantly when semantic composition is omitted. This finding implies that the meaning of sign-forms can diverge markedly between maps whose scales differ only marginally—particularly for scales between 1:500 and 1:20,000. An example may illustrate this result. For instance, plats or cadastral maps, typically drawn at a scale between 1:1,000 to 1:4,000 commonly signify land plot boundaries with simple black lines; building polygons may be colored or textured, whereas non-built areas are generally left blank. By contrast, city maps at scales between 1:10,000 to 1:15,000 seldom delineate land plots. Thus, the same black line on a blank background can instead signify a narrow footpath, a tramway, a telephone line, or even the map graticule. Similarly, a black line separating two identically colored areas can indicate the boundary between building blocks and streets—or between parks and streets. These cases illustrate conditions under which a substantial discrepancy would emerge between the base model and the model that omits semantic composition.

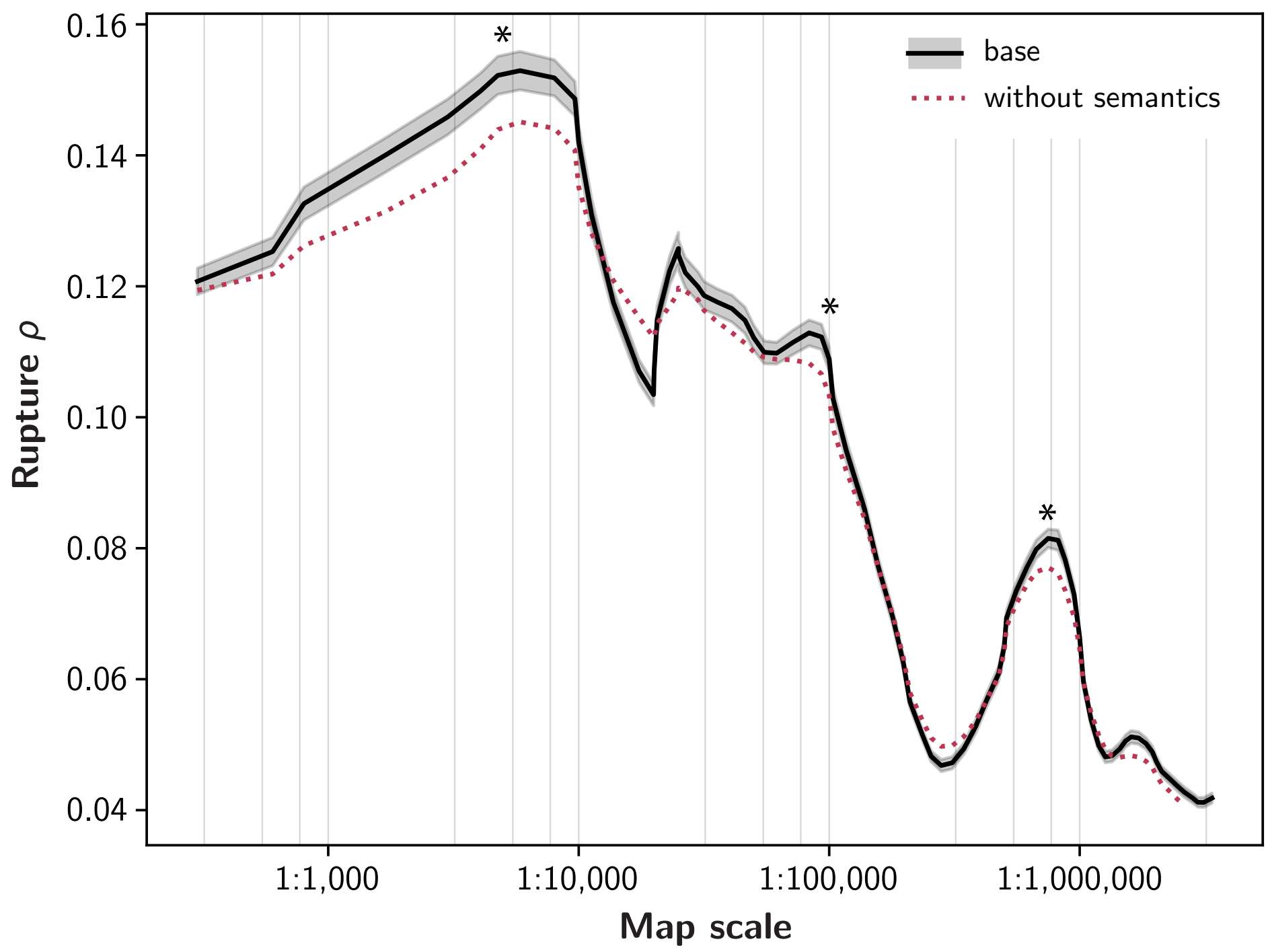

**Figure 8 | Coefficient of rupture ρ by map scale**. The coefficient is computed using a sliding window (200 steps) and two time strata, each corresponding to 10% of the dataset and overlapping by 50%. The shaded area denotes the 95% confidence interval of the mean. The red dotted line depicts the coefficient of rupture when conditional semantic modes are disregarded (i.e., when it derives only from mapel cluster frequencies). Asterisks (*) indicate the peaks identified in the analysis of point sign ruptures (Chap. 6, Fig. 25). *The largest semiotic distinction occurs between the maps whose scale is above ca. 1:3,000, and those below 1:10,000.*

This relatively high variability also contributes to the overall elevated coefficient of rupture at larger, more detailed map scales, particularly when semantics are considered. An illustration of this result is the variability of figurative forms used to represent built areas on large-scale maps[9]. A possible explanation for the higher semiotic variation observed is the increased familiarity with the geographic objects depicted. One characteristic of large-scale maps is indeed that the spaces they represent can be *experienced* directly and thus recognized by association with similar known spaces (Edney, 2019, p. 33). Concretely, this entails that most Westerners with elementary cartographic literacy would be able to distinguish the principal objects (building blocks and streets) on Alphand's blank Atlas of Paris (1880–1888), even if they have not been there. By contrast, it might be difficult for a reader to distinguish an island from a lake on a blank[10] small-scale map of an unfamiliar region. From this perspective, cultural knowledge of physical places and spaces constitutes the *cultural system* within which cartographic semiotics evolve. The same dynamic was also apparent in the evolution of map icons discussed in the previous chapter, where highly iconic signs, like trees, were employed to signify larger geographic objects, like forests, even at relatively small map scales, well beyond the scale at which individual trees would be visible on an aerial image.

[9] For instance, the French renovated cadastres depicted buildings in carmine pink; the Sanborn fire insurance maps rendered them in yellow or pink, depending on the construction material; the Bavarian cadastre employed hatchings while the British Ordnance Survey adopted dotted areas. The Atlas of Paris by Adolphe Alphand (1817–1891) simply left urban blocks blank. This set of examples illustrate the limits of acceptable semiotic variation, especially given that these maps were all published around the same period, that is, toward the end of the 19th century.

[10] In this paragraph, *blank* refers to a map figured in black and white, where only the shape contours are marked with a black line.

### *The evolution of mapels through time*

Figure 9 shows the evolution of the relative frequency of mapels over time by compositional semantic class. The division into six time strata is based on salient rupture peaks (Fig. 7). For each semantic–symbolic time stratum, the most characteristic[11] exemplars are depicted in Figure 10. Figure 11 combines the identified form classes with the relative mapel frequencies reported in Figure 9 to provide aggregate average mapel frequencies for each form class, by semantic compositional mode and time stratum; it offers a statistical overview of how mapel forms change over time and according to semantic composition.

**Buildings and built-up areas.** The representation of buildings and built-up areas underwent substantial changes over the study period. Until 1737, buildings areas were commonly filled with dense cross-hatching. Hatching is a drawing technique originating from print making, where it was used for shading and conveying perspective (Griffiths, 1996). While its presence in oblique perspective iconographic city maps is expected, observation of characteristic exemplars indicates that similar hatched textures were later adapted to the depiction of built-up areas on planimetric maps. Dotted patterns were also frequent, particularly from the mid-18th century to the end of the 19th century. During this interval, rocker mechanisms and mechanical rulings became increasingly common, facilitating the engraving of such textures, and rendering them more even and regular (Pearson, 1983; Robinson, 1975). Hand-painting of built-up areas, prevalent in the 15th and 16th centuries, appears to have declined in the 18th and the first half of the 19th century, mirroring the expansion of textures. From the second half of the 19th century, plain color fills became more frequent again, driven by the spread of color printing (Pearson, 1980; Ristow, 1975). Figure 11 also indicates the increasing frequency of text overlays. As visible in the characteristic exemplars (Fig. 10b), texts include additional information layers like house numbers, building names and function, or owners, as well as technical details. The incidence of linework also increased markedly around the turn of the 19th century, which seems to reflect the increasing subdivision of built-up areas, e.g. through the delineation of individual building footprints.

[11] Characteristicity $\chi_{m,s}$ was defined in Chapter 6. Here, we additionally consider semantic conditional modes for each stratum $s$.

**Built + non-built.** While the previous semantic compositional mode primarily comprised large-scale urban maps, due to geographic constraints inherent to the representation of built areas, the mode that integrates both built and non-built features also includes medium-scale maps in which city icons and symbols appear. This is most evident in the characteristic exemplars of the first three time strata (Fig. 10a). It illustrates the generic character of mapels, which also encompass point signs, although their treatment may not be as precise or specific as permitted by the approach adopted in Chapter 6. Beyond pictorial city icons and circle symbols, one can also observe hybrid signs, such as the symbolization of fortified cities based on their actual urban shape. The progression of textures—especially hatchings—and an increased use of color fills become apparent from the end of the 19th century onward. A notable semiotic phenomenon is the persistence of building shading in an alternative graphical form observed in 19th century maps. Although shading hachures disappeared—presumably to avoid confusion with relief hachuring and hatched building patterns—a thicker line was often still drawn on one or two sides of building geometries[12] to signify the shadow. This practice is visible in both Figure 10b and Figure 11, which document the growing frequency of thick-line visual forms.

**Water.** In the first time stratum, water was commonly depicted with thick, dense, and slightly curved hachures, as well as with dotted patterns. The latter were often employed in marine charting to differentiate navigable sea from shallow waters, typically rendered with dotted textures—perhaps to signify sand. During the 17th and 18th centuries, water was increasingly represented as a blank area. Hatchings or waterlines became less frequent, whereas isolated lines became more common. Indeed, regular or dashed lines were sometimes used to represent azimuths or the map graticule. A resurgence of hatchings, waterlines, and dotted patterns was observed between 1782 and 1887. At that time, however, the spacing between lines or dots appeared to be very narrow and regular, likely attributable to the mechanization of texture-engraving processes. From the mid-19th century, the graphical processes for drawing waterlines evolved as well, with the introduction of a central broken line. Moreover, waterlines were no longer confined to coasts but could occasionally fill entire bodies of water with concentric contours. From the end of the 19th century, color became prevalent, especially blue and green hues. Color was not only applied as a plain fill; printing textures, such as hatchings or waterlines, with colored ink also became a recurrent practice.

[12] This graphic choice is also well-known in the context of early 19th century European cadastres. For instance, the French *Recueil méthodique* states that “the line is reinforced on the shadow side” (Hennet, 1811, p. 79).

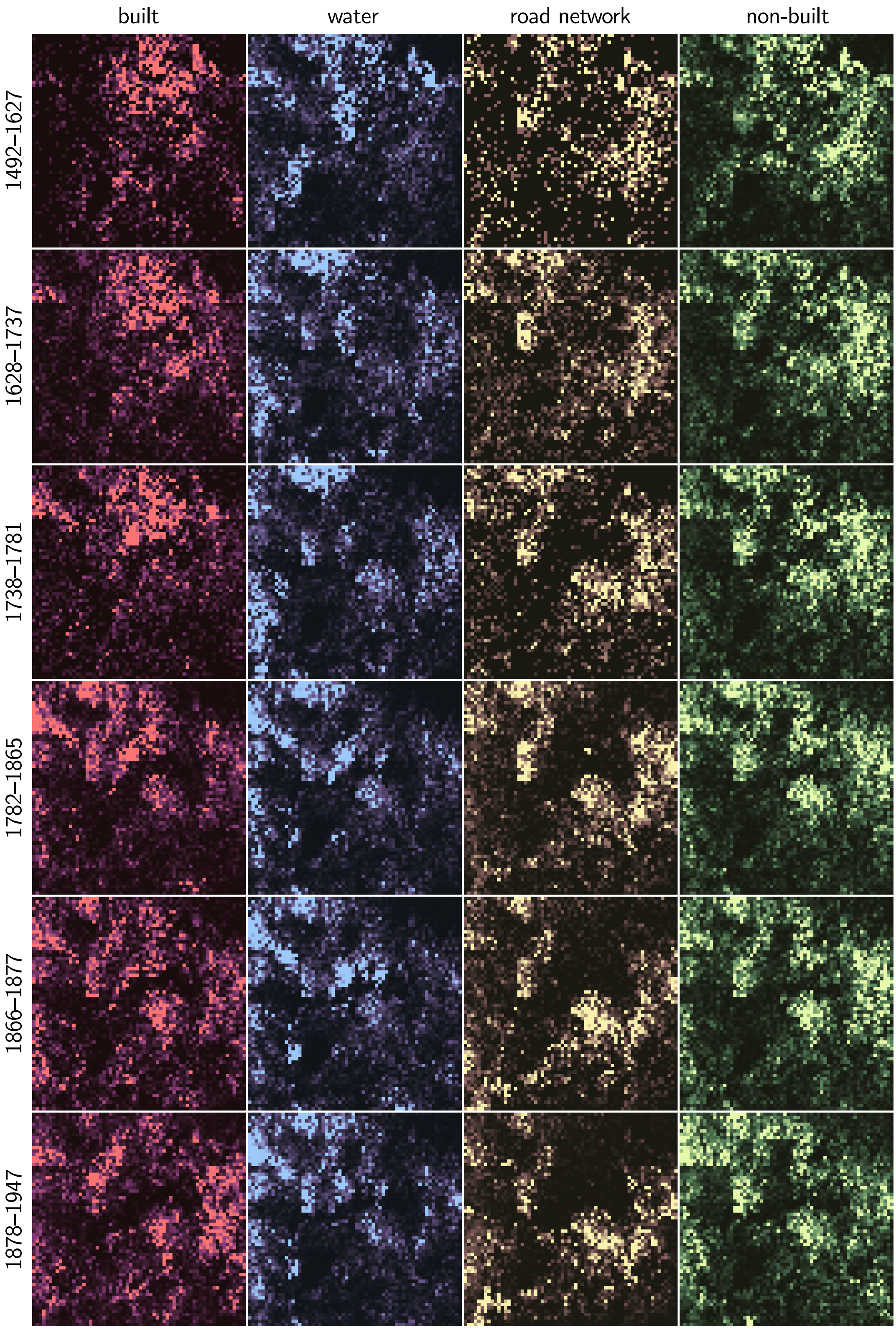

**Figure 9 | (a) Evolution of the relative distribution of mapels across time strata.** Each block represents the relative frequency of mapel clusters within a specific semantic–symbolic time stratum. Blocks are ordered by publication year (vertically) and compositional semantic mode (horizontally). Mapel clusters are spatialized following the mapel mosaic (Fig. 4). Brighter areas indicate higher frequencies (corresponding color scales are provided in Fig. A1). Frequencies are normalized, with saturation at the 95th percentile for each semantic–symbolic stratum. Continued in Fig. 9b.

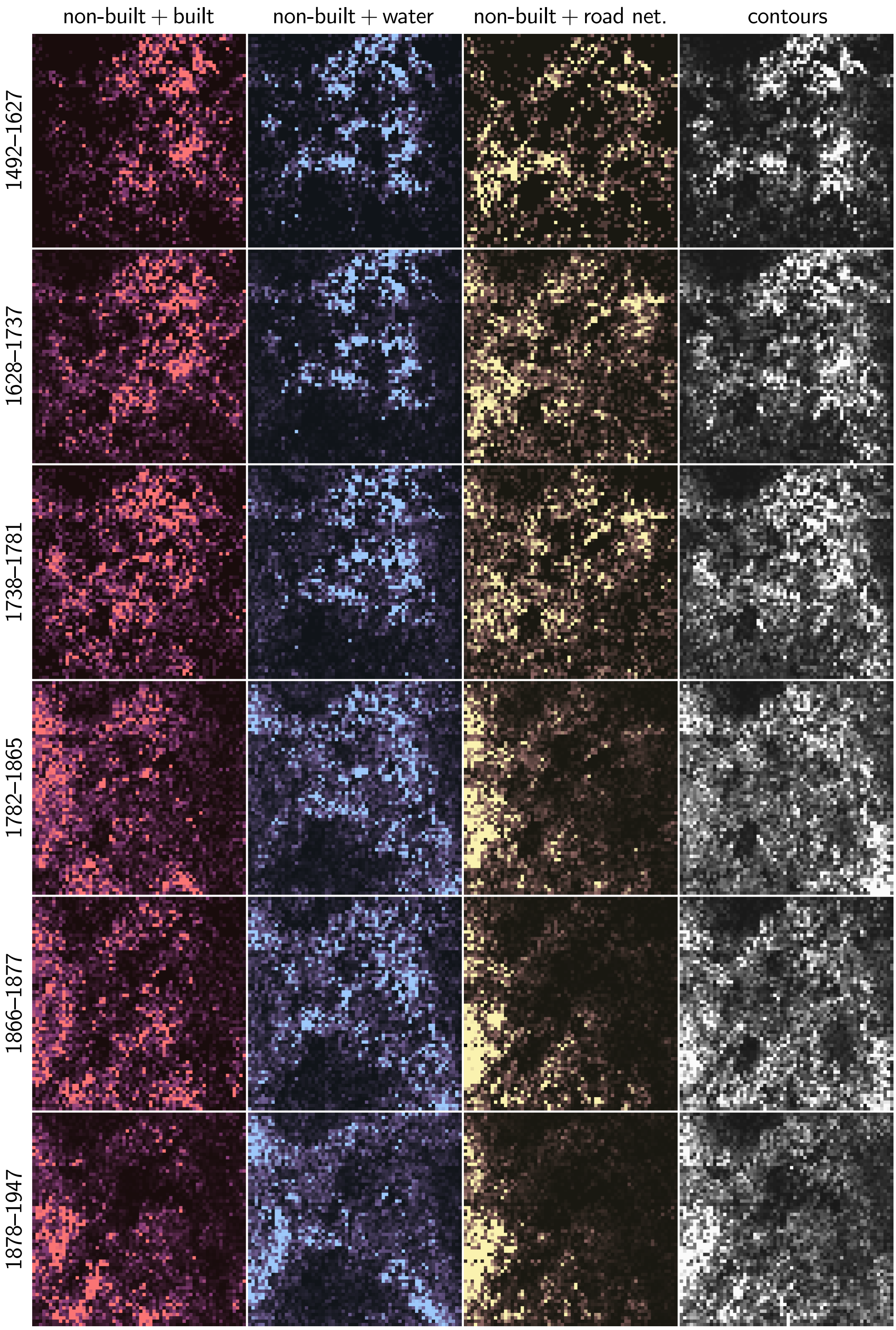

**Figure 9 | (b) Evolution of the relative distribution of mapels across time strata.** Each block represents the relative frequency of mapel clusters within a specific semantic–symbolic time stratum. Blocks are ordered by publication year (vertically) and compositional semantic mode (horizontally). Mapel clusters are spatialized following the mapel mosaic (Fig. 4). Brighter areas indicate higher frequencies (color scales are provided in Fig. A1). Frequencies are normalized, with saturation at the 95$^{th}$ percentile for each semantic–symbolic stratum. *E.g. the space of figuration corresponding to road network mapels changes only marginally over the study period, whereas built and contours evolve substantially.*

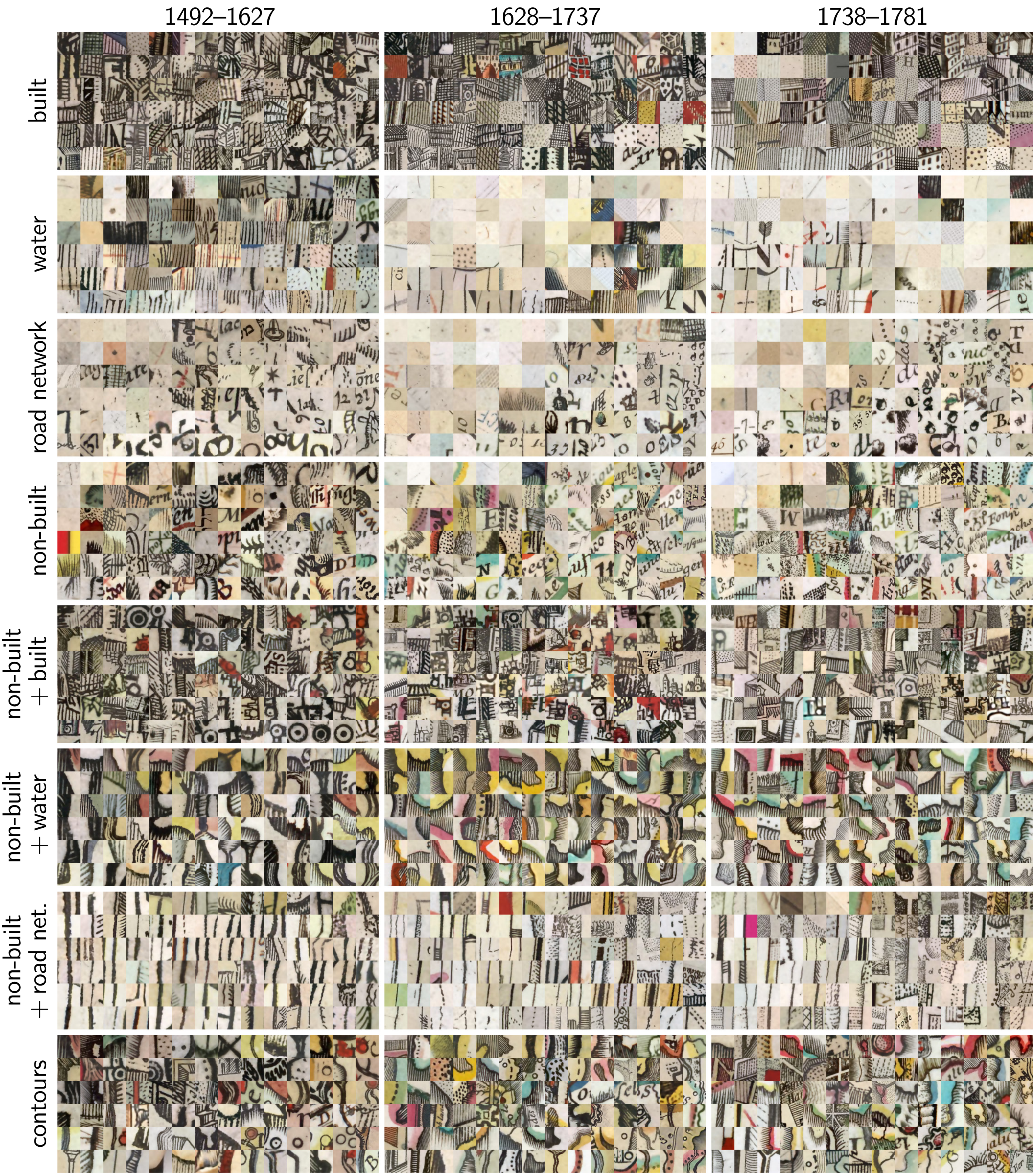

**Figure 10 | (a) Exemplars of characteristic mapels for each time stratum.** Each block depicts the 84 most characteristic exemplars, for the corresponding semantic-symbolic time stratum highlighted in Fig. 9. The blocks are arranged by publication year (horizontally) and compositional semantic mode (vertically). Continued in Fig. 10b.

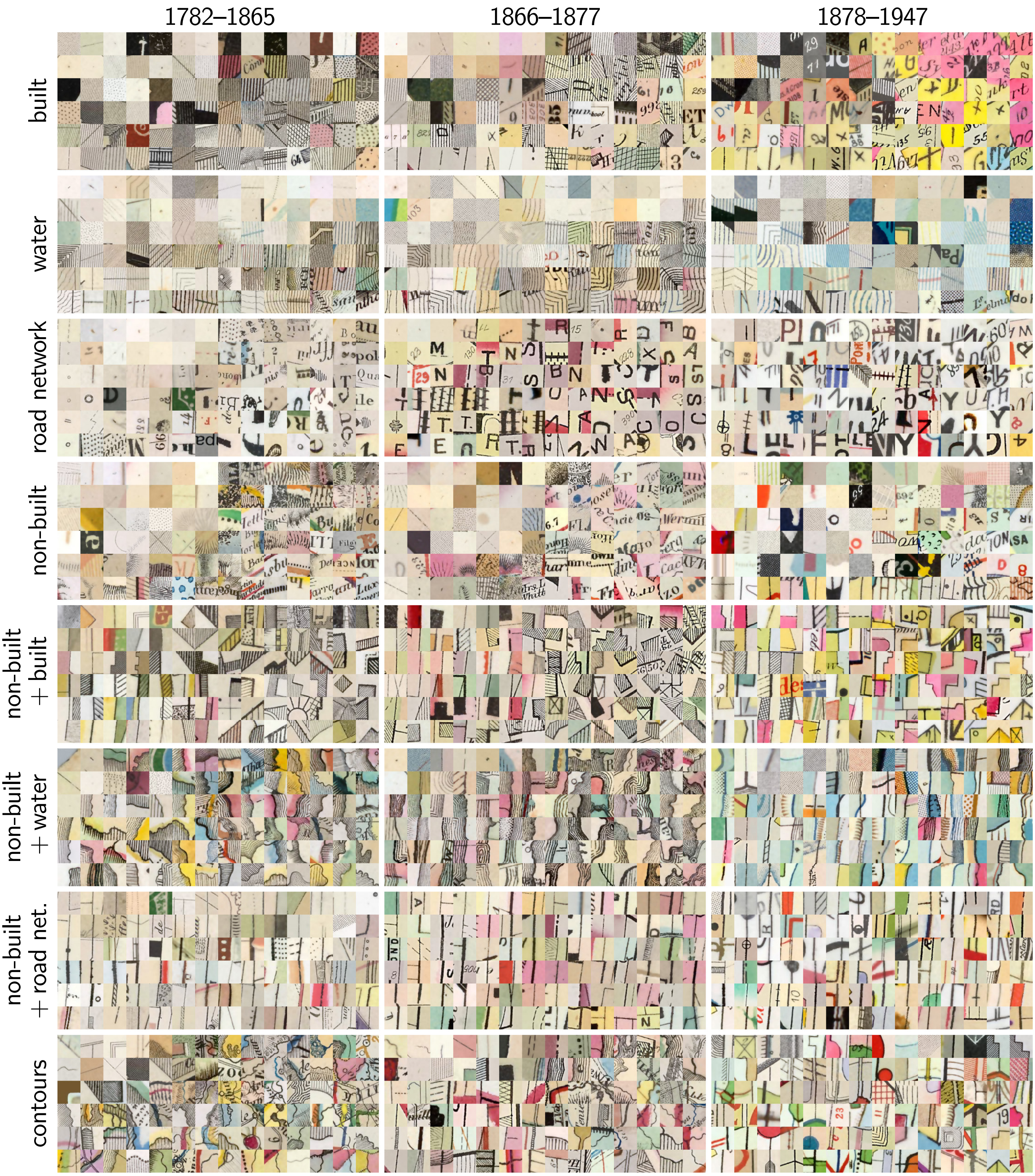

**Figure 10 | (b) Exemplars of characteristic mapels for each time stratum.** Each block depicts the 84 most characteristic exemplars, for the corresponding semantic-symbolic time stratum highlighted in Fig. 9. The blocks are arranged by publication year (horizontally) and compositional semantic mode (vertically). *E.g. shading and cross–hatching denoting buildings were superseded by dotted and hatched textures between 1738–1865, then by colored fills from 1878 onwards.*

**Coasts and rivers (water + non-built).** The representation of coastlines is coherent with previous observations. Initially, coastlines were generally marked with a thick, dark line. Short hatchings drawn perpendicular to the coast were frequent, and the shores were often colored in yellow, carmine, or green hues. These short perpendicular hatches left high waters blank, as stated above. From the 18th century until the middle of the 19th century, relief hachuring was commonly used to emphasize shores. The occurrence of marked coastlines rose continually until the 1870s. At that time, delimitation between land and sea with a single line became the most prevalent form; water and/or the shore tended to be further differentiated with colored fills or textures. Dotted patterns or stylized embankment hachuring are also characteristic of the more recent maps. Few rivers are visible in the characteristic mapels of this semantic compositional mode, except perhaps in the first time stratum, where larger courses appear, denoted with waterlines. This relative absence is attributable to the fact that mapels figuring thin rivers fall below the 30% water threshold imposed on the base semantic composition for this mode. The *wavy line* is nonetheless accounted for by visual forms (Fig. 11), especially from the end of the 18th century.

**Streets and road network (wide).** The depiction of roads differs from that of buildings and water in that it is less distinctive, particularly in the first three time strata. This is evident in the scattered initial distribution of mapels (Fig. 9a), which also reflects the comparatively small area covered by roads on early maps. During that period, road areas were mostly left blank. Relief hachures occasionally appeared, as did tree icons and, later, circle symbols indicating tree-lined alleys. Text remained prevalent throughout the study period. From the mid-19th century, large, capital, and sans-serif fonts appear particularly prevalent in the results (Figs. 10b and 11).

**Roads (thin, i.e., road-network + non-built).** Thin roads, often associated with smaller map scales, are predominantly depicted as simple lines. Whereas wide or double lines were more common in the 17th century, thinner lines were preferred from the end of the 18th century onward. Text seems comparatively rare relative to the frequency observed for wider roads. In both semantic compositional modes related to the road network, Figure 11 indicates the sudden appearance, around 1866, of a new visual form corresponding to track lines. Related exemplars are also visible in Figure 10.

| Compositional semantic modes | Publication date | regular lines | dotted lines | dashed lines | double lines | thick lines | shape corners | directional text | hachures | terrain contours | waterlines | wavy lines | large texts | ticked track lines | dotted patterns | text | hatchings | blank or dots | multicolor | col. background | . |
|---|---|---|---|---|---|---|---|---|---|---|---|---|---|---|---|---|---|---|---|---|---|
| built | 1492–1627 | .01 | .06 | .01 | .04 | .02 | .06 | .16 | .25 | .18 | .13 | .03 | .17 | .00 | .27 | .11 | .26 | .05 | .04 | .30 | .19 |
| built | 1628–1737 | .05 | .11 | .04 | .04 | .04 | .05 | .16 | .27 | .19 | .11 | .04 | .23 | .04 | .66 | .25 | .28 | .21 | .03 | .26 | .20 |
| built | 1738–1781 | .07 | .14 | .05 | .05 | .05 | .06 | .09 | .29 | .22 | .08 | .03 | .13 | .00 | .74 | .18 | .37 | .27 | .03 | .22 | .19 |
| built | 1782–1865 | .24 | .18 | .14 | .05 | .07 | .06 | .11 | .22 | .16 | .07 | .04 | .16 | .00 | .62 | .29 | .37 | .57 | .01 | .16 | .17 |
| built | 1866–1877 | .31 | .10 | .15 | .05 | .05 | .07 | .22 | .17 | .14 | .07 | .07 | .23 | .03 | .30 | .38 | .28 | .40 | .03 | .20 | .18 |
| built | 1878–1947 | .28 | .09 | .24 | .05 | .07 | .10 | .25 | .08 | .10 | .02 | .08 | .28 | .07 | .13 | .50 | .11 | .21 | .05 | .29 | .22 |
| water | 1492–1627 | .18 | .15 | .18 | .48 | .14 | .04 | .03 | .17 | .15 | .29 | .09 | .20 | .00 | .47 | .16 | .42 | .49 | .01 | .24 | .19 |
| water | 1628–1737 | .40 | .16 | .20 | .12 | .19 | .09 | .08 | .10 | .11 | .16 | .12 | .27 | .01 | .35 | .29 | .31 | .74 | .01 | .11 | .17 |
| water | 1738–1781 | .45 | .20 | .45 | .06 | .25 | .07 | .09 | .08 | .09 | .10 | .07 | .21 | .01 | .24 | .32 | .18 | .76 | .01 | .06 | .13 |
| water | 1782–1865 | .36 | .14 | .19 | .08 | .08 | .07 | .12 | .16 | .19 | .27 | .06 | .16 | .01 | .19 | .25 | .43 | .66 | .01 | .06 | .14 |
| water | 1866–1877 | .31 | .14 | .16 | .18 | .07 | .06 | .10 | .23 | .29 | .44 | .07 | .17 | .02 | .13 | .21 | .47 | .54 | .01 | .08 | .16 |
| water | 1878–1947 | .40 | .11 | .22 | .14 | .10 | .05 | .08 | .15 | .23 | .25 | .07 | .20 | .03 | .18 | .25 | .34 | .64 | .02 | .13 | .16 |
| road network | 1492–1627 | .10 | .25 | .30 | .13 | .14 | .05 | .08 | .11 | .08 | .10 | .11 | .39 | .00 | .08 | .34 | .09 | .34 | .01 | .19 | .16 |
| road network | 1628–1737 | .10 | .18 | .18 | .12 | .15 | .06 | .19 | .10 | .10 | .08 | .11 | .38 | .07 | .10 | .35 | .09 | .58 | .01 | .16 | .19 |
| road network | 1738–1781 | .09 | .20 | .24 | .02 | .12 | .02 | .15 | .06 | .06 | .03 | .09 | .37 | .00 | .16 | .40 | .03 | .55 | .01 | .15 | .16 |
| road network | 1782–1865 | .18 | .27 | .26 | .05 | .11 | .05 | .17 | .04 | .07 | .01 | .12 | .41 | .03 | .40 | .46 | .03 | .60 | .02 | .15 | .15 |
| road network | 1866–1877 | .22 | .15 | .41 | .15 | .14 | .09 | .21 | .03 | .04 | .01 | .10 | .48 | .40 | .07 | .37 | .03 | .38 | .01 | .12 | .12 |
| road network | 1878–1947 | .29 | .05 | .48 | .15 | .17 | .07 | .15 | .02 | .05 | .01 | .12 | .48 | .40 | .02 | .32 | .02 | .28 | .01 | .07 | .12 |
| non-built | 1492–1627 | .06 | .14 | .14 | .16 | .11 | .02 | .06 | .17 | .14 | .17 | .15 | .39 | .04 | .20 | .34 | .18 | .41 | .02 | .28 | .20 |
| non-built | 1628–1737 | .09 | .21 | .13 | .04 | .07 | .03 | .09 | .18 | .13 | .09 | .14 | .33 | .05 | .24 | .51 | .09 | .48 | .06 | .25 | .20 |
| non-built | 1738–1781 | .15 | .22 | .19 | .02 | .10 | .05 | .10 | .28 | .16 | .06 | .10 | .24 | .11 | .50 | .49 | .09 | .57 | .05 | .17 | .20 |
| non-built | 1782–1865 | .22 | .20 | .19 | .03 | .07 | .05 | .11 | .34 | .21 | .06 | .08 | .22 | .06 | .38 | .49 | .11 | .62 | .07 | .22 | .21 |
| non-built | 1866–1877 | .24 | .19 | .26 | .04 | .07 | .07 | .14 | .28 | .16 | .07 | .08 | .25 | .07 | .17 | .52 | .10 | .50 | .05 | .22 | .20 |
| non-built | 1878–1947 | .36 | .13 | .25 | .06 | .11 | .08 | .11 | .22 | .41 | .08 | .14 | .25 | .08 | .18 | .39 | .08 | .56 | .07 | .20 | .23 |
| non-built + built | 1492–1627 | .02 | .03 | .01 | .07 | .10 | .05 | .20 | .18 | .12 | .09 | .05 | .36 | .05 | .06 | .17 | .13 | .01 | .08 | .36 | .19 |
| non-built + built | 1628–1737 | .10 | .13 | .04 | .20 | .30 | .12 | .29 | .22 | .14 | .13 | .08 | .30 | .13 | .27 | .23 | .18 | .03 | .10 | .32 | .24 |
| non-built + built | 1738–1781 | .15 | .13 | .07 | .29 | .43 | .18 | .25 | .24 | .19 | .16 | .10 | .24 | .06 | .45 | .15 | .27 | .03 | .05 | .27 | .27 |
| non-built + built | 1782–1865 | .49 | .20 | .18 | .27 | .55 | .36 | .19 | .17 | .18 | .09 | .11 | .17 | .03 | .27 | .12 | .24 | .10 | .03 | .21 | .27 |
| non-built + built | 1866–1877 | .46 | .15 | .22 | .15 | .44 | .34 | .26 | .10 | .13 | .10 | .15 | .18 | .09 | .14 | .17 | .18 | .08 | .03 | .20 | .24 |
| non-built + built | 1878–1947 | .45 | .10 | .34 | .17 | .54 | .54 | .21 | .05 | .11 | .02 | .12 | .13 | .05 | .03 | .12 | .05 | .04 | .06 | .14 | .19 |
| non-built + water | 1492–1627 | .04 | .02 | .04 | .18 | .17 | .02 | .05 | .15 | .09 | .37 | .16 | .28 | .14 | .11 | .09 | .27 | .03 | .03 | .35 | .18 |
| non-built + water | 1628–1737 | .06 | .06 | .04 | .11 | .16 | .03 | .04 | .18 | .10 | .45 | .17 | .27 | .10 | .16 | .09 | .31 | .07 | .10 | .38 | .18 |
| non-built + water | 1738–1781 | .12 | .14 | .11 | .08 | .21 | .03 | .08 | .28 | .15 | .44 | .21 | .26 | .31 | .34 | .17 | .31 | .12 | .11 | .33 | .21 |
| non-built + water | 1782–1865 | .21 | .18 | .17 | .13 | .23 | .06 | .10 | .35 | .32 | .51 | .48 | .22 | .34 | .13 | .22 | .33 | .12 | .22 | .36 | .31 |
| non-built + water | 1866–1877 | .25 | .21 | .27 | .29 | .29 | .06 | .11 | .39 | .33 | .64 | .46 | .18 | .44 | .10 | .20 | .24 | .10 | .17 | .31 | .31 |
| non-built + water | 1878–1947 | .52 | .20 | .30 | .25 | .64 | .09 | .10 | .16 | .29 | .38 | .59 | .21 | .16 | .10 | .18 | .13 | .12 | .24 | .31 | .27 |
| non-built + road net. | 1492–1627 | .23 | .18 | .24 | .49 | .50 | .08 | .21 | .06 | .09 | .12 | .16 | .15 | .10 | .03 | .18 | .08 | .01 | .01 | .20 | .20 |
| non-built + road net. | 1628–1737 | .43 | .45 | .45 | .55 | .55 | .14 | .25 | .15 | .15 | .13 | .17 | .10 | .15 | .50 | .23 | .14 | .10 | .04 | .22 | .27 |
| non-built + road net. | 1738–1781 | .34 | .37 | .21 | .47 | .50 | .15 | .32 | .18 | .14 | .09 | .11 | .10 | .08 | .46 | .22 | .12 | .06 | .01 | .19 | .26 |
| non-built + road net. | 1782–1865 | .69 | .25 | .35 | .46 | .49 | .17 | .25 | .07 | .11 | .04 | .10 | .05 | .05 | .20 | .10 | .06 | .08 | .02 | .12 | .19 |
| non-built + road net. | 1866–1877 | .66 | .13 | .36 | .40 | .46 | .22 | .27 | .03 | .05 | .02 | .09 | .06 | .22 | .04 | .06 | .03 | .05 | .01 | .06 | .14 |
| non-built + road net. | 1878–1947 | .71 | .07 | .37 | .46 | .49 | .19 | .19 | .03 | .06 | .02 | .08 | .05 | .21 | .02 | .05 | .03 | .03 | .02 | .05 | .12 |
| contours | 1492–1627 | .08 | .07 | .07 | .29 | .32 | .06 | .14 | .14 | .11 | .21 | .25 | .40 | .06 | .07 | .15 | .19 | .01 | .05 | .37 | .21 |
| contours | 1628–1737 | .13 | .38 | .12 | .26 | .39 | .09 | .18 | .18 | .14 | .30 | .39 | .41 | .12 | .18 | .25 | .26 | .03 | .18 | .45 | .26 |
| contours | 1738–1781 | .20 | .34 | .16 | .29 | .51 | .12 | .24 | .34 | .23 | .28 | .41 | .32 | .30 | .42 | .26 | .30 | .03 | .18 | .37 | .32 |
| contours | 1782–1865 | .40 | .31 | .32 | .33 | .45 | .21 | .31 | .33 | .33 | .26 | .54 | .23 | .22 | .19 | .33 | .22 | .05 | .21 | .39 | .38 |
| contours | 1866–1877 | .54 | .25 | .47 | .43 | .44 | .34 | .45 | .25 | .26 | .34 | .44 | .20 | .39 | .11 | .29 | .16 | .05 | .11 | .30 | .33 |
| contours | 1878–1947 | .65 | .18 | .40 | .43 | .62 | .46 | .36 | .14 | .28 | .11 | .27 | .17 | .19 | .08 | .20 | .08 | .06 | .23 | .26 | .29 |

(Visual form)

**Figure 11 | Evolution of the average distribution of form classes across time strata.** Each table cell reports the average frequency of mapel clusters attributed to a certain form class (defined in Fig. 6), within a particular time stratum and compositional semantic mode, calculated from the mapel cluster frequencies reported in Fig. 9. E.g. *Hachures dominate non-built relief depiction ca. 1738–1877, after which they are replaced by terrain contours.*

**Non-built.** The "non-built" semantic class encompasses a variety of land uses, including forests, mountains, farmland, parks, and grassland. This ontological heterogeneity is manifested in the diversity of mapels. In the first two strata, the representation of non-built areas did not differ notably from that of the road network, as evidenced by their distribution in Figure 9a. Until the end of the 18th century, non-built areas were mainly characterized by the presence of tree or mountain icons, relief hachures, and dotted patterns. The latter motif, which may, depending on context, depict forests, orchards, crops, or wasteland, appears to have peaked in the mid–18th century. Texts, especially slanted serif typographies from the mint green and fuchsia regions (cf. Fig. 5, Fig. A3) are also recurrent. Blank regions were frequent throughout the entire study period, and became particularly common between 1738 and 1865. While the frequency of relief hachures peaked between 1782 and 1865, terrain contours became the predominant technique for representing elevation from 1878 onward (Fig. 11).

**Contours.** The "contours" semantic compositional mode encompasses diverse cartographic objects, such as coasts, roads, rivers, and regional boundaries. Despite this diversity, general tendencies can be discerned, particularly in the relative evolution of visual forms shown in Figure 11. This includes, for instance, the progressive adoption of regular thin lines, as revealed by the relative distribution of mapels in Figure 9. Dotted lines became prevalent in the 17th century and were gradually supplanted by dashed lines during the 18th and 19th centuries (Fig. 11). The contours class is also strongly associated with the presence of rivers, as apparent from the characteristic examples in Figure 10 and in the comparatively high occupation of the flax region (Figs. 5, 9b). Wavy lines appear throughout the entire period, peaking between 1782 and 1877. This visual form may subsequently have been replaced by blue lines, a hypothesis supported by the increased frequency of the *multicolor* visual form in both the *non-built+water* and *contours* semantic compositional modes. Alternatively, the heightened activation of the salmon region for the *non-built+water* mode in Figure 9b indicates that rivers may simply have been rendered with thicker wavy lines.

### *Map scale*

Figure 12 reports the variation in mapel frequencies across map scales and semantic compositional mode. Figure 13 presents representative exemplars for each case, and Figure 14 displays the aggregate variation by form class. The description of the results will be primarily focused on elements that complement those identified in the historical analysis of map figuration.

Figure 12 shows that the relative distribution of mapels in the built, and road network semantic compositional modes exhibits only limited variation across scales. However, at smaller map scales, densities seem reduced and therefore less evenly distributed. This observation is consistent with the results of Chapter 4, Figure 8, which established that the two corresponding semantic classes predominantly occur at scales larger than 1:50,000. Below that threshold, point and line signs tend to replace area signs; consequently, areal and linear features predominate in the *non-built+built* and *non-built+road network* compositional modes. In the water mode, the main difference is a slight increase in text visual forms at smaller scales. A comparable change is observed for the non-built mode, where the mint-green and fuchsia regions, consisting mainly of slanted texts, appear more frequently on small-scale maps. Conversely, the blue region, corresponding to dots or blank areas, is comparatively more prevalent at larger scales. These observations suggest that place naming is a predominant way, at smaller scales, to signify provinces, water bodies, seas, or landmasses. Within the non-built class, one can also note the predominance of elevation representation devices, like terrain contours, and relief hachuring at intermediate scales (1:8,000–1:750,000), while they are relatively rare at both lower and higher scales.

The representation of boundaries is slightly more differentiated across scales, compared to semantically uniform areas. The *non-built+built* mode undergoes significant changes, marked, as anticipated, by a transition from area to point signs. This is shift manifested in the progressive abandonment of isolated lines at smaller scales(e.g., building boundaries), which may also reflect the rarer occurrence of building subdivisions into individual footprints. At medium scales, dotted patterns and hatching also appear quite common, whereas building coloring predominates at larger scales.

The map scale at which the *non-built+road network* mode seems most distinctive lies between 1:8,000 and 1:25,000. At this scale, various road types are differentiated by distinct line forms, such as dashed, dotted, or double lines (Fig. 14). At smaller scales, by contrast, these line forms occur less frequently than thick or regular lines. Road naming is likewise most common in this second scale stratum, as indicated by the higher prevalence of directional text forms.

Figure 12 indicates that the two remaining semantic compositional modes, namely *non-built+water* and *contours* undergo comparable changes across map scales. An examination of characteristic exemplars reveals that many mapels assigned to the *contours* mode represent coastlines or rivers, particularly at smaller scales (Fig. 13b). At larger scales, the principal boundaries are depicted with regular or thick lines, whereas at smaller scales both modes are characterized by wavy lines and by a *multicolor* visual form, which typically correspond to dense mapels with substantial informational overlay (cf. Fig. 6).

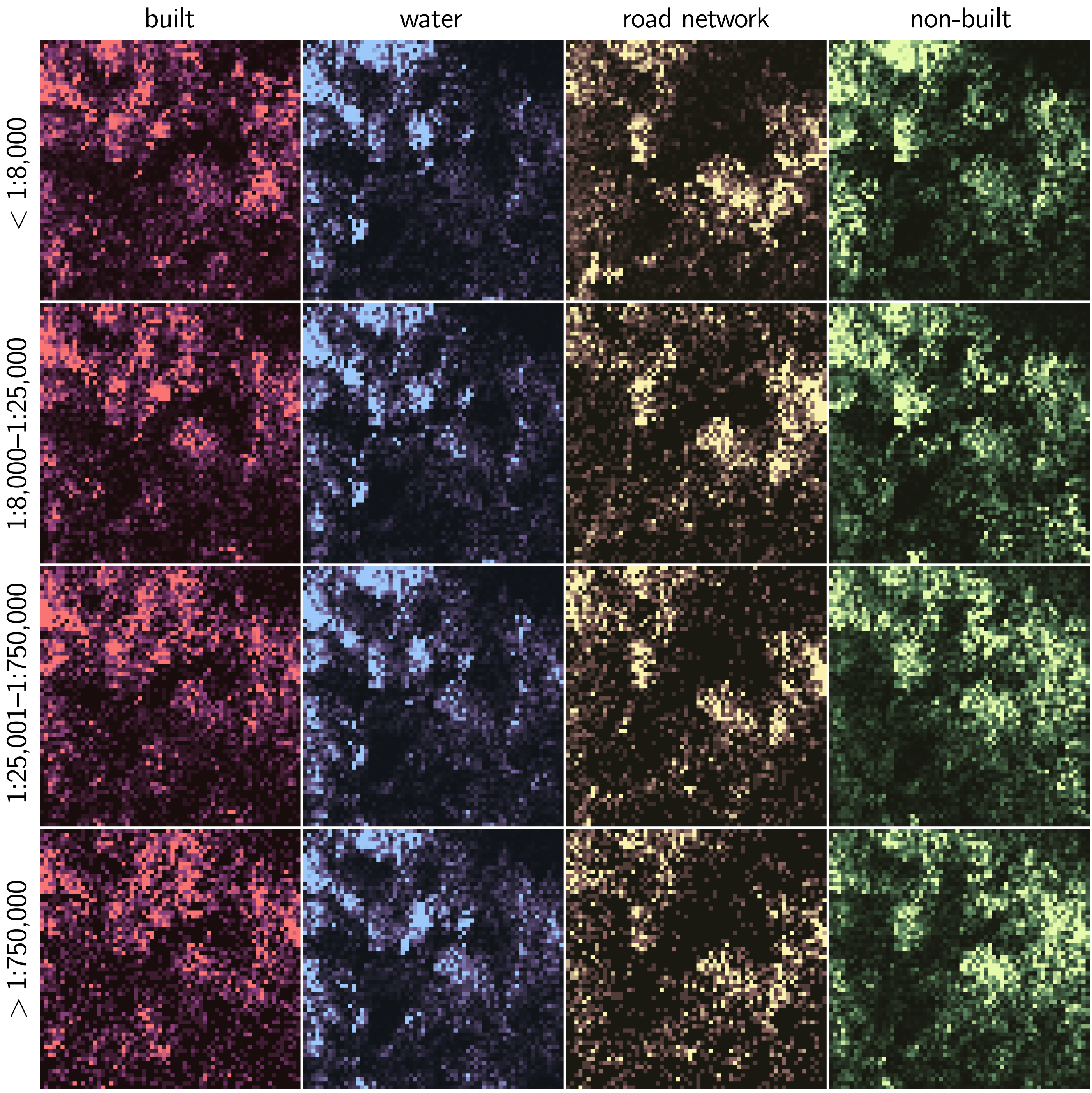

**Figure 12 | (a) Relative distribution of mapels across scale strata.** Each block represents the relative frequency of mapel clusters within a specific semantic–symbolic scale stratum. Blocks are ordered by map scale (vertically) and compositional semantic mode (horizontally). Mapel clusters are spatialized following the mapel mosaic (Fig. 4). Brighter areas indicate higher frequencies (corresponding color scales are provided in Fig. A1). Frequencies are normalized, with saturation at the 95$^{th}$ percentile for each semantic–symbolic stratum. Continued in Fig. 12b.

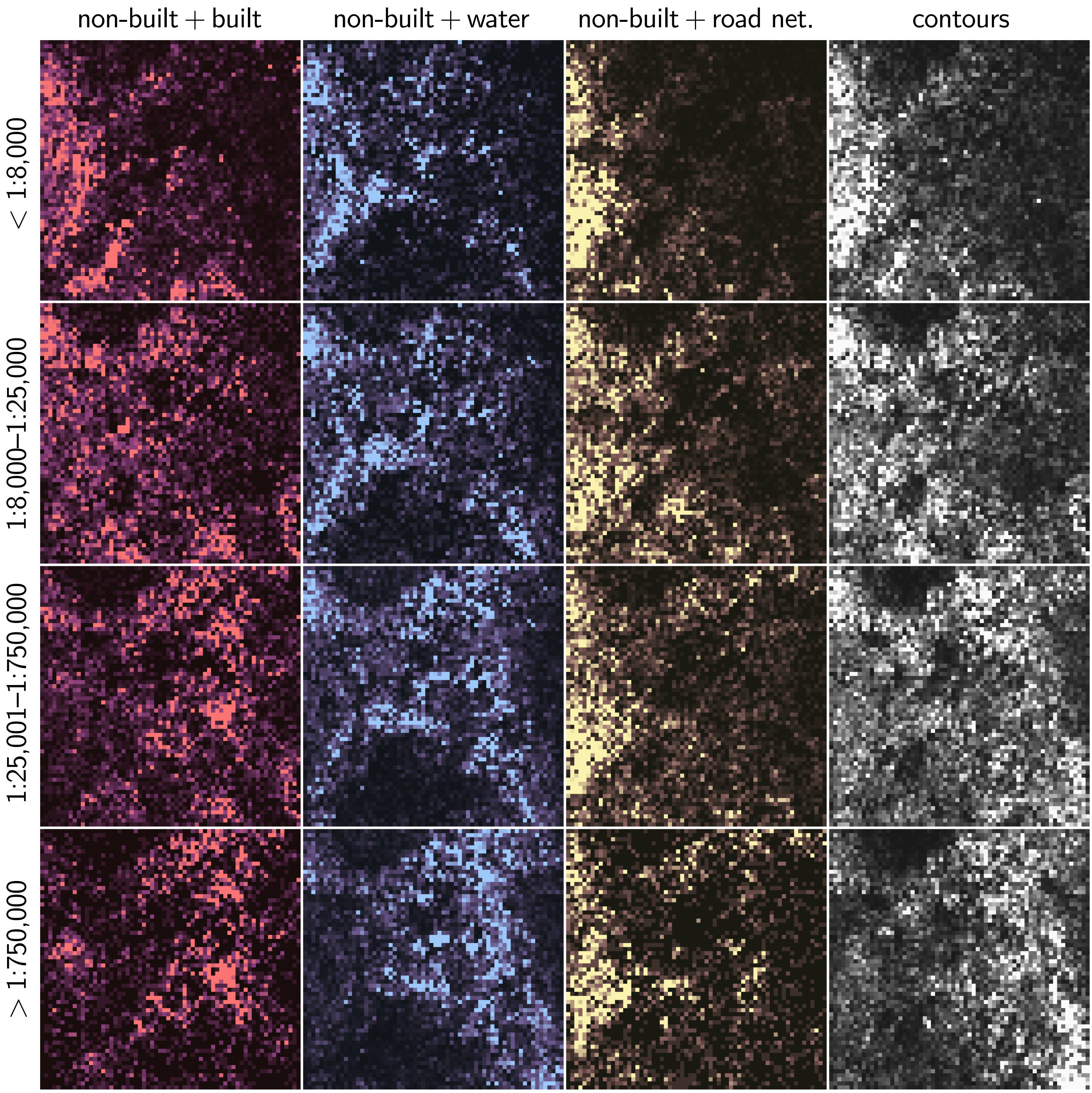

**Figure 12 | (b) Relative distribution of mapels across scale strata.** Each block represents the relative frequency of mapel clusters within a specific semantic–symbolic scale stratum. Blocks are ordered by map scale (vertically) and compositional semantic mode (horizontally). Mapel clusters are spatialized following the mapel mosaic (Fig. 4). Brighter areas indicate higher frequencies (corresponding color scales are provided in Fig. A1). Frequencies are normalized, with saturation at the 95$^{th}$ percentile for each semantic–symbolic stratum. *Areal features (Fig. 12a) appear to change only marginally as a function of scale—except for the non-built class—, whereas contours and mapels comprising multiple semantic classes differ more substantially.*

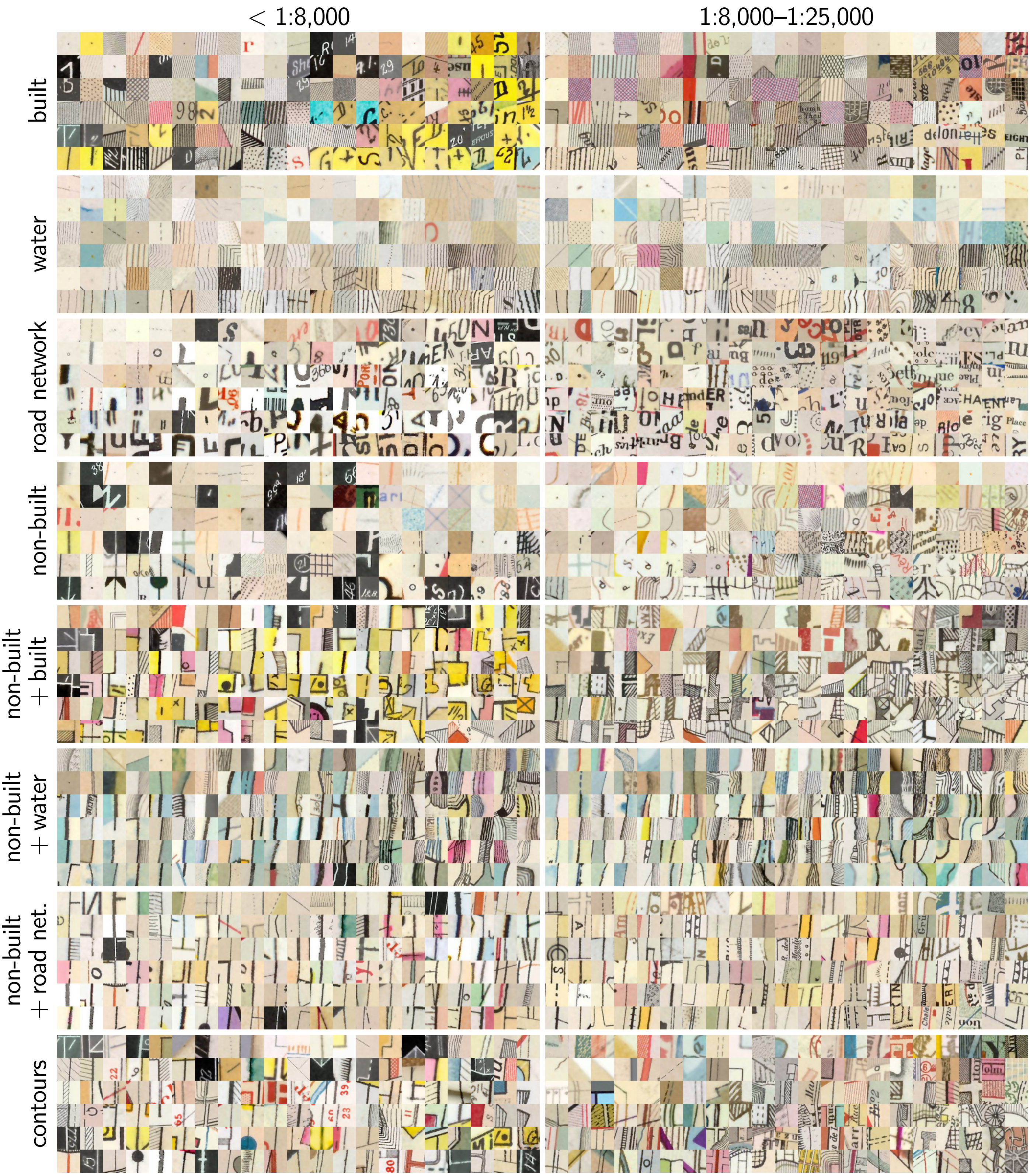

**Figure 13 | (a) Exemplars of characteristic mapels for each scale stratum.** Each block depicts the 126 most characteristic exemplars for the corresponding semantic-symbolic scale stratum highlighted in Fig. 12. The blocks are arranged according to map scale (horizontally) and compositional semantic mode (vertically). Continued in Fig. 13b.

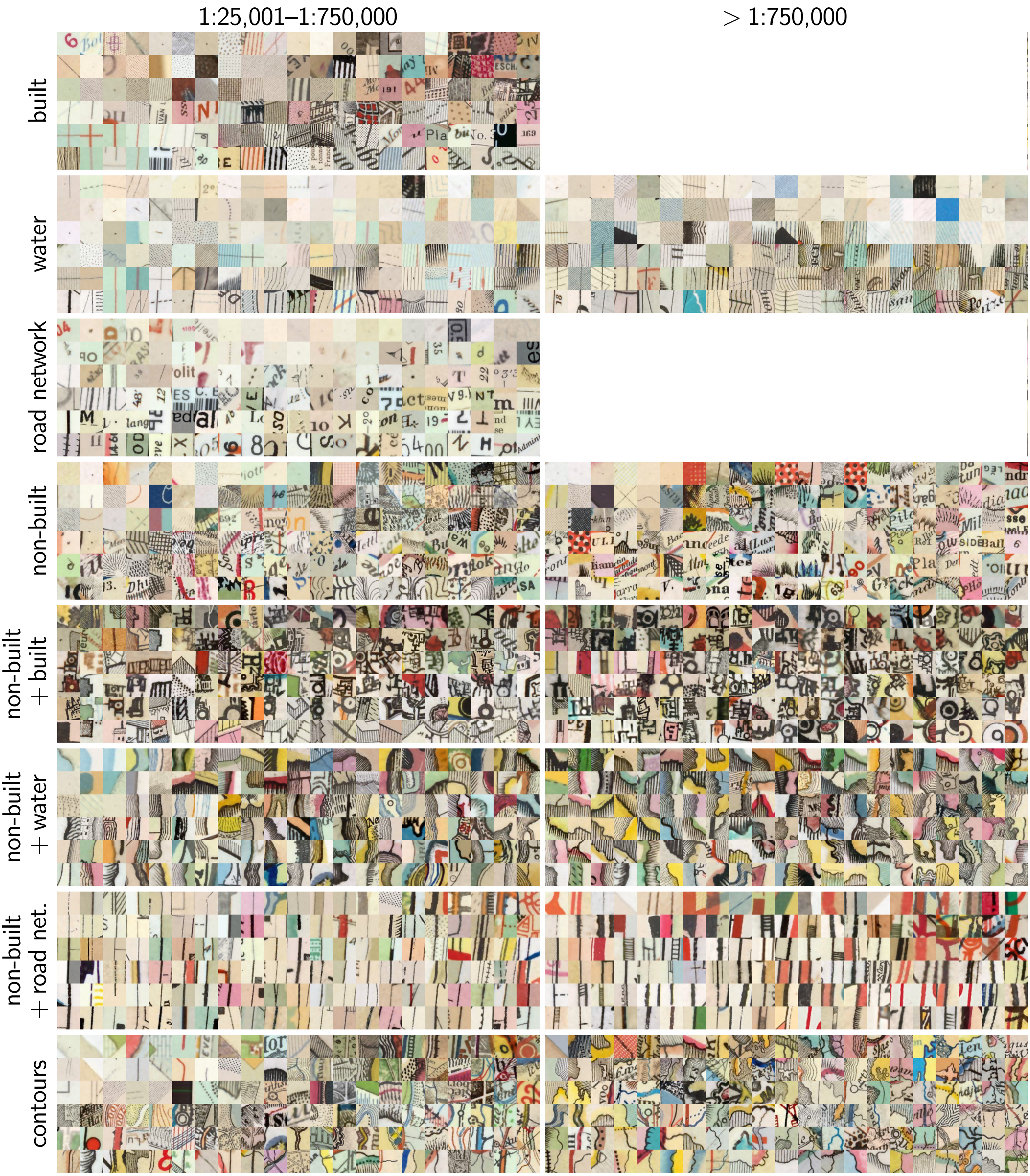

**Figure 13 | (b) Exemplars of characteristic mapels for each scale stratum.** Each block depicts the 126 most characteristic exemplars for the corresponding semantic-symbolic scale stratum highlighted in Fig. 12. The blocks are arranged according to map scale (horizontally) and compositional semantic mode (vertically). *E.g. the rendering of non-built mapels becomes increasingly dense at lower scales, due to the higher prevalence of terrain contours, hachures, and texts.*

**Visual form**

| Compositional semantic modes | Scale stratum | regular lines | dotted lines | dashed lines | double lines | thick lines | shape corners | directional text | hachures | terrain contours | waterlines | wavy lines | large texts | ticked track lines | dotted patterns | text | hatchings | blank or dots | multicolor | col. background | . |
|---|---|---|---|---|---|---|---|---|---|---|---|---|---|---|---|---|---|---|---|---|---|
| built | i | .29 | .12 | .24 | .06 | .08 | .08 | .17 | .11 | .13 | .04 | .07 | .29 | .02 | .40 | .41 | .29 | .49 | .06 | .30 | .21 |
| | ii | .27 | .17 | .15 | .08 | .07 | .06 | .14 | .19 | .18 | .06 | .04 | .18 | .02 | .60 | .35 | .35 | .41 | .05 | .23 | .18 |
| | iii | .31 | .17 | .14 | .07 | .06 | .07 | .17 | .20 | .21 | .05 | .05 | .18 | .00 | .48 | .36 | .27 | .36 | .03 | .30 | .22 |
| water | i | .33 | .10 | .24 | .20 | .08 | .03 | .04 | .13 | .16 | .27 | .08 | .13 | .02 | .14 | .15 | .42 | .74 | .01 | .08 | .15 |
| | ii | .31 | .10 | .24 | .17 | .08 | .03 | .04 | .16 | .27 | .31 | .07 | .15 | .01 | .13 | .18 | .39 | .67 | .01 | .07 | .14 |
| | iii | .40 | .14 | .28 | .12 | .12 | .04 | .06 | .15 | .22 | .28 | .07 | .17 | .01 | .38 | .23 | .40 | .78 | .01 | .09 | .15 |
| | iv | 40 | 14 | 21 | 08 | 13 | 08 | 14 | 18 | 20 | 33 | 07 | 22 | 01 | 22 | 29 | 46 | 65 | 01 | 09 | 15 |
| road net. | i | .23 | .12 | .43 | .14 | .19 | .04 | .12 | .02 | .07 | .01 | .13 | .49 | .21 | .10 | .38 | .03 | .56 | .01 | .07 | .15 |
| | ii | .09 | .23 | .27 | .07 | .06 | .04 | .17 | .05 | .10 | .02 | .05 | .34 | .03 | .29 | .49 | .06 | .30 | .12 | .21 | .15 |
| | iii | .17 | .09 | .24 | .11 | .12 | .04 | .16 | .05 | .09 | .01 | .07 | .31 | .40 | .03 | .36 | .04 | .52 | .05 | .11 | .14 |
| non-built | i | .45 | .16 | .25 | .06 | .16 | .03 | .08 | .15 | .13 | .04 | .08 | .24 | .02 | .38 | .31 | .11 | .80 | .01 | .10 | .18 |
| | ii | .22 | .20 | .23 | .11 | .05 | .08 | .07 | .24 | .43 | .09 | .17 | .16 | .02 | .42 | .31 | .16 | .61 | .02 | .13 | .23 |
| | iii | .14 | .17 | .16 | .05 | .06 | .06 | .08 | .35 | .41 | .12 | .12 | .26 | .07 | .26 | .50 | .12 | .54 | .09 | .25 | .24 |
| | iv | .22 | .20 | .19 | .03 | .09 | .07 | .15 | .27 | .17 | .05 | .12 | .35 | .09 | .20 | .58 | .07 | .51 | .11 | .27 | .22 |
| non-built + built | i | .48 | .16 | .29 | .18 | .64 | .51 | .15 | .05 | .11 | .04 | .12 | .14 | .00 | .12 | .11 | .14 | .07 | .07 | .16 | .22 |
| | ii | .32 | .18 | .23 | .33 | .37 | .35 | .23 | .18 | .35 | .10 | .15 | .15 | .03 | .28 | .15 | .23 | .05 | .07 | .23 | .32 |
| | iii | .14 | .11 | .11 | .09 | .23 | .18 | .20 | .17 | .30 | .08 | .12 | .38 | .15 | .10 | .22 | .16 | .04 | .13 | .40 | .27 |
| | iv | .11 | .10 | .06 | .11 | .40 | .13 | .18 | .11 | .13 | .07 | .10 | .41 | .00 | .11 | .18 | .10 | .03 | .13 | .37 | .20 |
| non-built + water | i | .47 | .10 | .24 | .42 | .66 | .07 | .06 | .13 | .15 | .49 | .24 | .10 | .05 | .11 | .08 | .25 | .13 | .02 | .14 | .20 |
| | ii | .41 | .24 | .27 | .37 | .43 | .07 | .12 | .25 | .37 | .61 | .41 | .14 | .08 | .20 | .13 | .24 | .10 | .06 | .17 | .26 |
| | iii | .25 | .18 | .30 | .16 | .29 | .05 | .08 | .30 | .33 | .57 | .45 | .26 | .24 | .24 | .16 | .29 | .13 | .20 | .38 | .30 |
| | iv | .12 | .14 | .12 | .08 | .18 | .06 | .09 | .33 | .28 | .45 | .49 | .29 | .38 | .10 | .22 | .33 | .09 | .32 | .45 | .29 |
| non-built + road net. | i | .73 | .12 | .36 | .39 | .48 | .19 | .20 | .04 | .07 | .02 | .07 | .04 | .08 | .06 | .07 | .05 | .06 | .01 | .04 | .13 |
| | ii | .61 | .26 | .38 | .70 | .37 | .27 | .41 | .07 | .17 | .06 | .13 | .06 | .04 | .16 | .14 | .08 | .04 | .06 | .17 | .25 |
| | iii | .66 | .16 | .35 | .57 | .58 | .21 | .20 | .08 | .18 | .07 | .18 | .13 | .16 | .13 | .09 | .07 | .05 | .18 | .21 | .23 |
| | iv | .52 | .07 | .25 | .46 | .66 | .13 | .13 | .05 | .12 | .07 | .13 | .20 | .13 | .07 | .07 | .06 | .05 | .21 | .23 | .21 |
| contours | i | .69 | .15 | .32 | .46 | .62 | .43 | .30 | .08 | .13 | .12 | .09 | .12 | .08 | .13 | .11 | .16 | .06 | .05 | .15 | .22 |
| | ii | .51 | .31 | .39 | .66 | .41 | .39 | .47 | .21 | .40 | .23 | .22 | .14 | .12 | .25 | .20 | .23 | .06 | .13 | .26 | .37 |
| | iii | .37 | .31 | .40 | .35 | .40 | .22 | .28 | .34 | .47 | .26 | .53 | .30 | .25 | .16 | .30 | .16 | .07 | .38 | .45 | .41 |
| | iv | .18 | .30 | .24 | .15 | .34 | .11 | .20 | .29 | .27 | .24 | .58 | .31 | .27 | .09 | .34 | .18 | .04 | .35 | .44 | .31 |

**Scale strata ↑**

| | |
|---|---|
| > 1:8,000 | i. |
| 1:8.000–1:25,000 | ii. |
| 1:25,001–1:750,000 | iii. |
| < 1:750,000 | iv. |

**Figure 14 | Average distribution of form classes by scale stratum.** Each table cell reports the average frequency of mapel clusters attributed to a certain form class (defined in Fig. 6), for a particular combination of scale stratum and compositional semantic mode, calculated from the mapel cluster frequencies highlighted in Figure 12. *E.g. when denoting non-built, blank areas are more prevalent at larger scales, whereas water tend to appear blank at any scale.*

### *The evolution of visual variables*

Figures 15–18 illustrate selected variable analyses afforded by the mapel approach. Figure 15 depicts the overall evolution of visual-form frequencies over time, whereas Figure 16 reports the distribution of visual variables across map scales. Figure 17 traces the evolution of six interpretable CV features related to: mapel density (number of connected components, graphic load), color (ink darkness, color saturation), linework (average line width), and the proportion of blank areas. Figure 18 presents the distribution of density-related features, line width, and blank areas as a function of map scale. These CV features are visually demonstrated in Figure 19. To preserve the concision of the argument, not every trend is described or discussed here. Rather, the following paragraphs will focus on the discussion of four particularly notable aspects, related to relief representation, text, coloring, and blank areas.

**Hachures and terrain contours.** While, hatching was employed during the Renaissance to convey perspective on engravings and iconographic maps, Figure 15 indicates that mapels containing *hachures* were comparatively more frequent from the mid-18$^{th}$ century to the latter half of the 19$^{th}$ century. During that period hachuring[13] became the dominant technique for representing orography, paralleling the decline of hill icons in the second half of the 18$^{th}$ century (cf. Chapter 6). Notable examples of topographic maps that used hachuring to depict relief at that time include Cassini's maps of France (1750–1815), the État-Major series (1832–1880), and the Dufour map of Switzerland (1845–1865). Although contour lines were invented in the mid-18$^{th}$ century already (Hansen, 2020; Rann & and Johnson, 2019), they did not outcompete hachures until the end of the 19$^{th}$ century, when the use of the latter declined sharply. Figure 16 shows that hachures predominated at map scales ranging between 1:50,000 and ca. 1:110,000, whereas terrain contours were most common at scales around 1:15,000–1:50,000. On the one hand, this pattern may partly reflect the progressive enlargement of national topographic series over time. For example, the Dufour map of Switzerland was drawn at 1:100,000, whereas its successor, the Siegfried maps, were engraved at 1:50,000, or even 1:25,000 for the Central Plateau (Feldman, 2015). On the other hand, the data reveal a distinction that prevailed in the first half of the 19$^{th}$ century, when hachures and contour lines coexisted as competing methods. During the early years of the État-Major surveying campaign, engineers commonly preferred contour lines for 1:40,000 maps and hachures for 1:80,000 maps.

[13] Note that in this text, I am referring to *hachures* in the broad sense, including the systematic methods developed by Johann G. Lehman at the end of the 18$^{th}$ century (Koch, 2013), but also anterior, related iconographic shading processes.

Contour lines were deemed more "rigorous," whereas hachures were valued for their aesthetic qualities and greater interpretability (Arnaud, 2022b; Zentai, 2018). In this regard, a possible explanation to the evolutionary success of contour lines, at the end of the 19th century is related to the development of hill shading at the same time. Hill shading could indeed be combined with contours to achieve a graphic representation considered both rigorous and aesthetically pleasing. In parallel, the expansion of color lithography at the same time offered new ways of differentiating lines, for instance with brown shades, which facilitated the superimposition of terrain contours onto other informational layers while limiting the impact on readability.

**Text.** Figure 16 indicates that *directional texts*—interpreted here as primarily denoting street names and house numbers—were concentrated in detailed, large-scale maps above 1:20,000. They were most prominent in the second half of the 19th century, a period regarded as particularly productive for urban atlases (Picon, 2003). Some emblematic examples include Julius Staube's *Übersichtsplan von Berlin* in 44 sheets (1866), G. W. Bromley's Atlas of the City of Boston (1883–1933), and J. C. Loman's *Buurtatlas* of Amsterdam (1876). Besides the specific subset of *directional text*s, textual elements in general seem more prevalent in small-scale maps (< 1:250,000). This implies that place naming played a key role in the semanticization of territories and of macroscopic geographic objects that lie beyond the scope of the experienced environment. Yet, the precise contribution of place names—as texts—to the cultural construction of the world model remains to be elucidated. That question would probably benefit from the digital analysis of map text datasets.

**Coloring.** Color saturation can be interpreted as an indicator of vivid hues (cf. Fig. 19). While maps were commonly colored by hand during the Renaissance, color printing remained impractical until the advent of color lithography in the 1860s (Ristow, 1975; Verdier & Besse, 2020; Woodward, 2007). However, while the overall trend is consistent with this chronology, Figure 17 indicates that the rise in color saturation over time is gradual. The development of color printing, indeed, was arguably affected by multiple technical, cultural, and economic factors, including semiotic evolution, aesthetic preferences, printing technologies, chemical knowledge, global pigment-trading networks, and printing costs. Consequently, Figure 17 appears to capture not a single rapid shift attributable to a single technical innovation but the gradual layout of associated cultural and economic contingencies. *A contrario*, the same figure registers a sharp decline in ink *darkness* during the second half of the 19th century. This decrease probably reflects the growing use of colored inks at the time, as an increasing proportion of map features—such as vegetation icons, waterlines, and contour lines—were rendered in color rather than in dark, black ink.

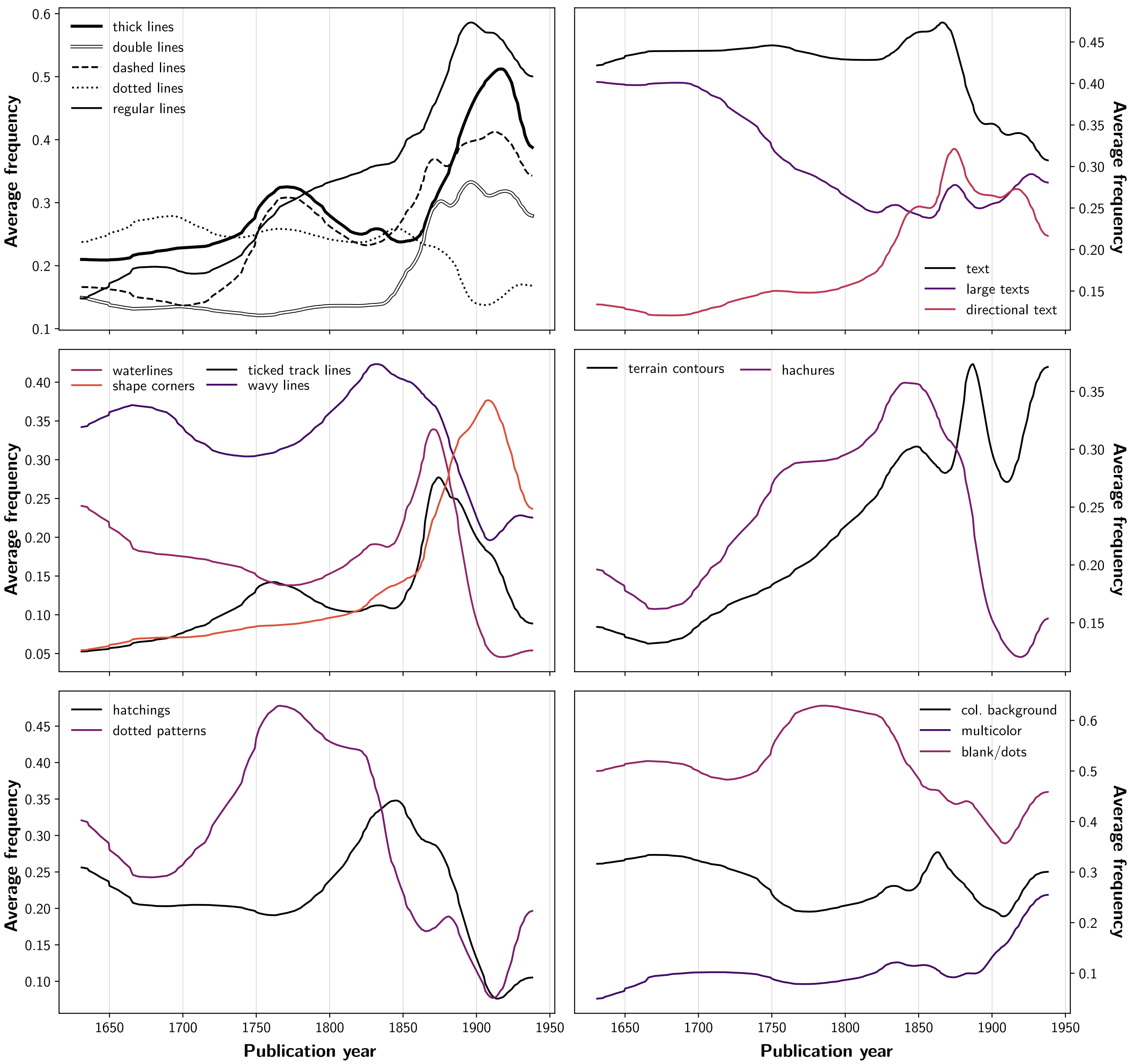

**Figure 15 | Relative evolution of visual forms over time.** Each curve represents the average frequency of mapel clusters assigned to a given form class (cf. Fig. 6) by publication date. Mean values were computed using sliding strata (164 steps) encompassing ca. 5% of the dataset each. Average frequencies were then normalized to the 95$^{th}$ percentile within each stratum. *E.g. the use of hachures and related visual forms to denote relief peaked around 1845; they were superseded by terrain contours in the 1870s.*

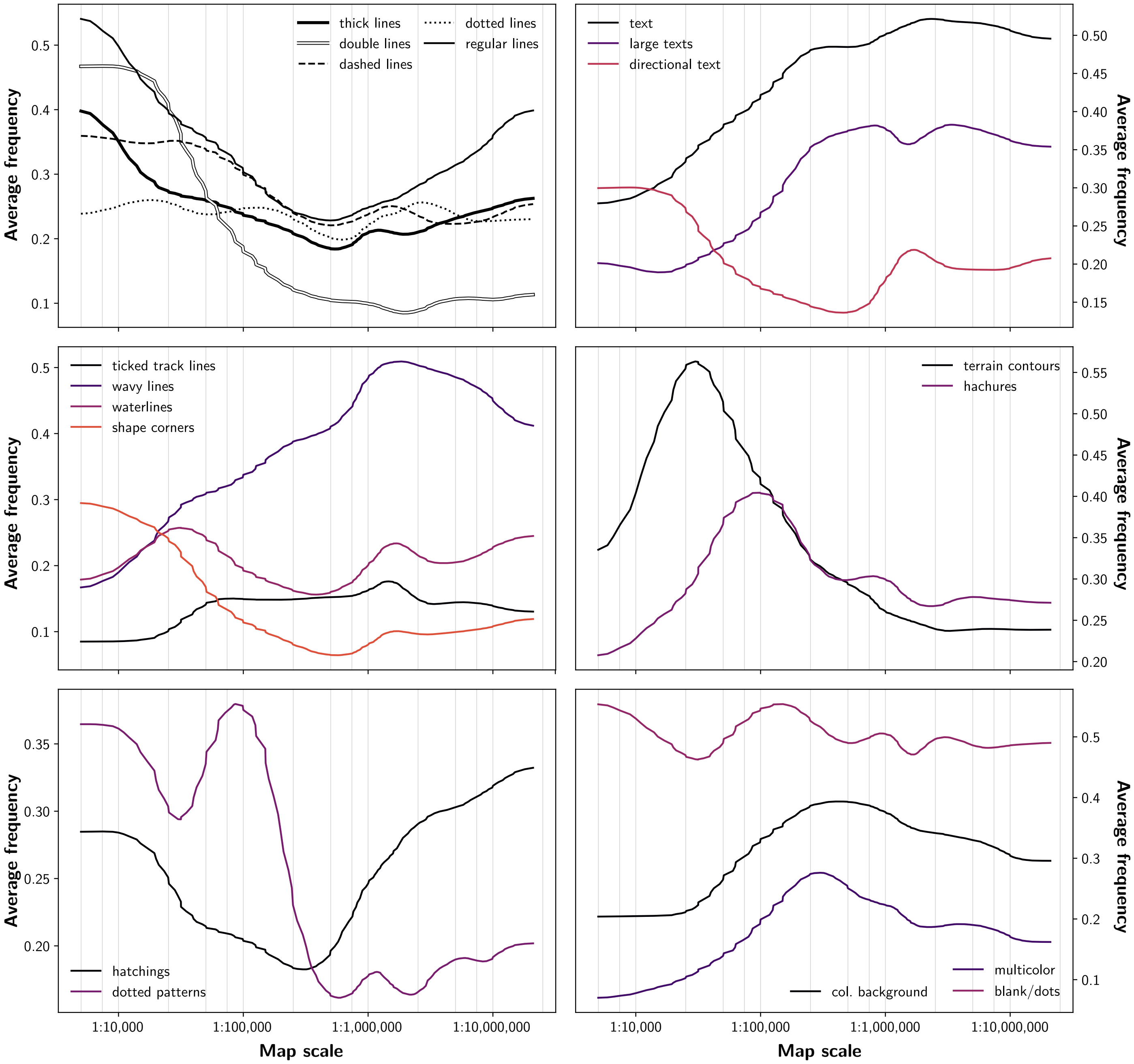

**Figure 16 | Distribution of visual forms by map scale.** Each curve represents the average frequency of mapel clusters assigned to a given form class (cf. Fig. 6) as a function of map scale. Mean values were computed using sliding strata (141 steps) encompassing ca. 10% of the dataset each. Average frequencies were then normalized to the 95th percentile within each stratum. *E.g. directional texts were more common at larger scales while other text forms (regular and large texts) prevailed at smaller scales.*

**Blank area**. Figure 17 shows a progressive increase in the proportion of map surfaces devoid of graphical content, from about 65% in the 17th century to more than 77% at the beginning of the 20th century. In contrast, and perhaps counterintuitively, the amplitude of the variation attributable to map scale is smaller, ranging from ca. 72% for large–scale maps to roughly 67% at its minimum for map scales of approximately 1:500,000 (Fig. 18). Beyond map scale, the historical change in the share of blank area can be related to distinct factors. First, the declining costs occasioned by the mechanization and industrialization of paper production in the 18th and 19th centuries (McGaw, 2019; Norman, n.d., b). Cheaper support presumably entailed fewer incentives to optimize the use of space on the document, a trend manifested in the diminishing graphic map load, along with the increase of blank areas. A second explanatory factor pertains to the philosophy of cartographic representation. Until the end of the 18th century, it was quite common to fill regions that were unexplored, or relatively unknown to Westerners, with speculative elements derived from biblical or fictional sources (Delaney, 2007; Van Duzer, 2013). Because little information was available, these speculative elements tended to be replicated across maps (Smith, 2019; Vaienti et al., 2025a). This practice of embellishing the extent of geographic knowledge resulted in visually richer and denser maps, that were arguably more marketable. This practice was increasingly challenged at the end of the 18th century, when the prevailing episteme required that geographical uncertainties be acknowledged and graphically reflected by leaving unexplored territories blank (Genevois et al., 2024).

### *Synthesis*

This section focused on highlighting selected macroscopic evolutions of cartographic figuration. The semiotic dynamics observed suggest a complex model of propagation of the innovations. Several results can be consistently explained by adaptive, competitive, and selective mechanisms. Signs can replace one another within the semantic-symbolic space. For instance, hatching was diverted to texture buildings on planimetric maps, whereas building shading became a simple reinforced line on one side. Later, following the widespread adoption of color printing, color fills outcompeted hatching. These changes can also be considered selective, hinging on aesthetic and epistemic considerations. For instance, terrain contouring did not outcompete hachuring—considered aesthetically more satisfactory—until it was combined with hill shading. Overall, the adopted perspective of cultural evolution acknowledges the influence of multiple historical developments, and the complexity of social transmission, offering a nuanced interpretative framework. However, its concrete implications for the constitution of the semiotic system, or its potential for elucidating the broader dynamics of socio-cultural transmission among map makers remains to be demonstrated. The next section undertakes a modest attempt to discuss these issues.

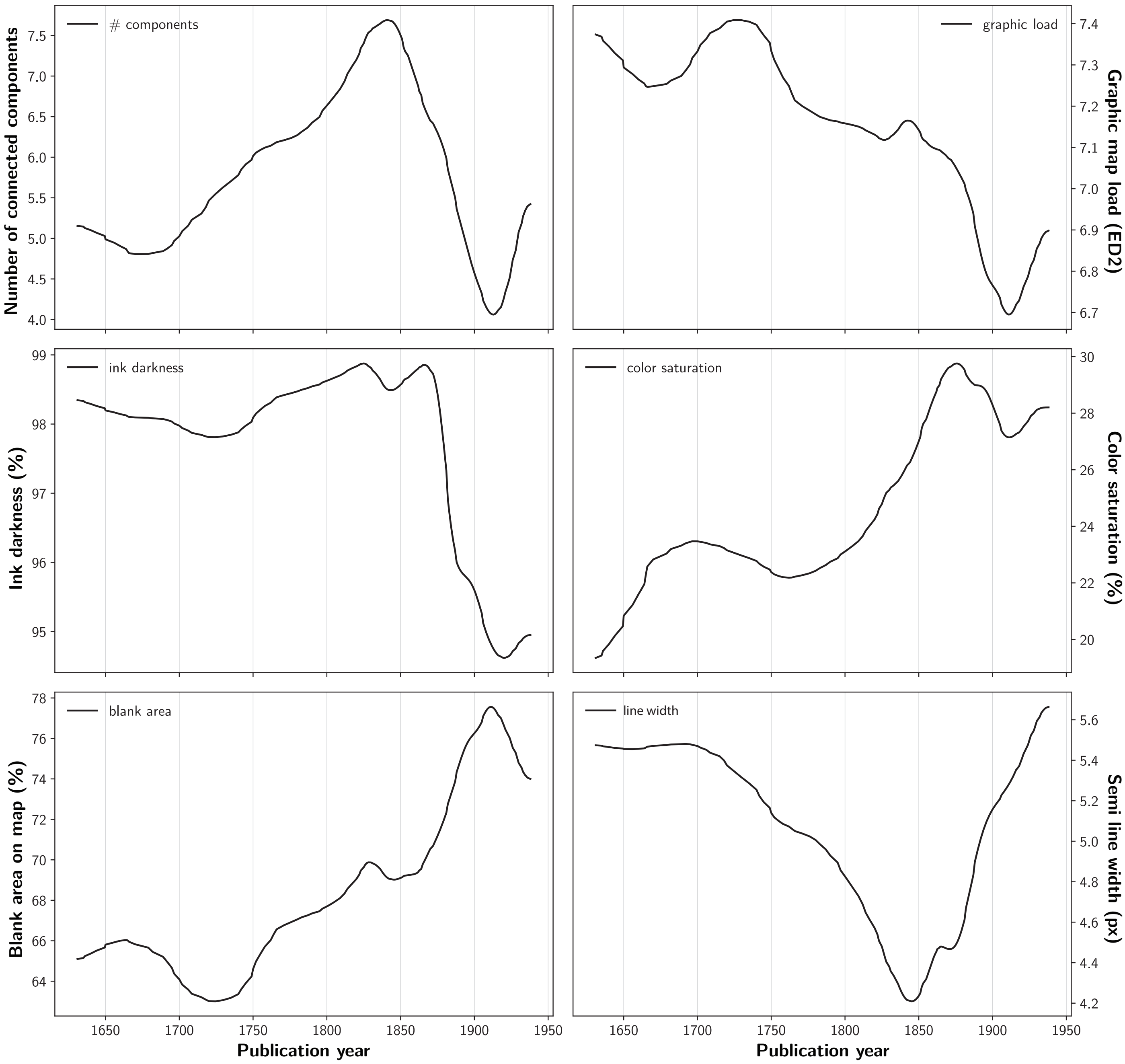

**Figure 17 | Relative evolution of CV-derived features over time.** Each curve represents the mean value of a computer vision derived feature, as a function of publication date. Mean values were computed using sliding strata (164 steps) encompassing ca. 10% of the dataset each. For # components, graphic load, color saturation, and line width, the average was calculated over all mapels within the corresponding stratum, whereas the share of blank areas and ink darkness were computed at the map level. A visualization of mapel-level CV features is provided in Figure 19. *E.g. color saturation and the proportion of blank areas increased over the study period, whereas graphic load (cf. Fig 19) decreased.*

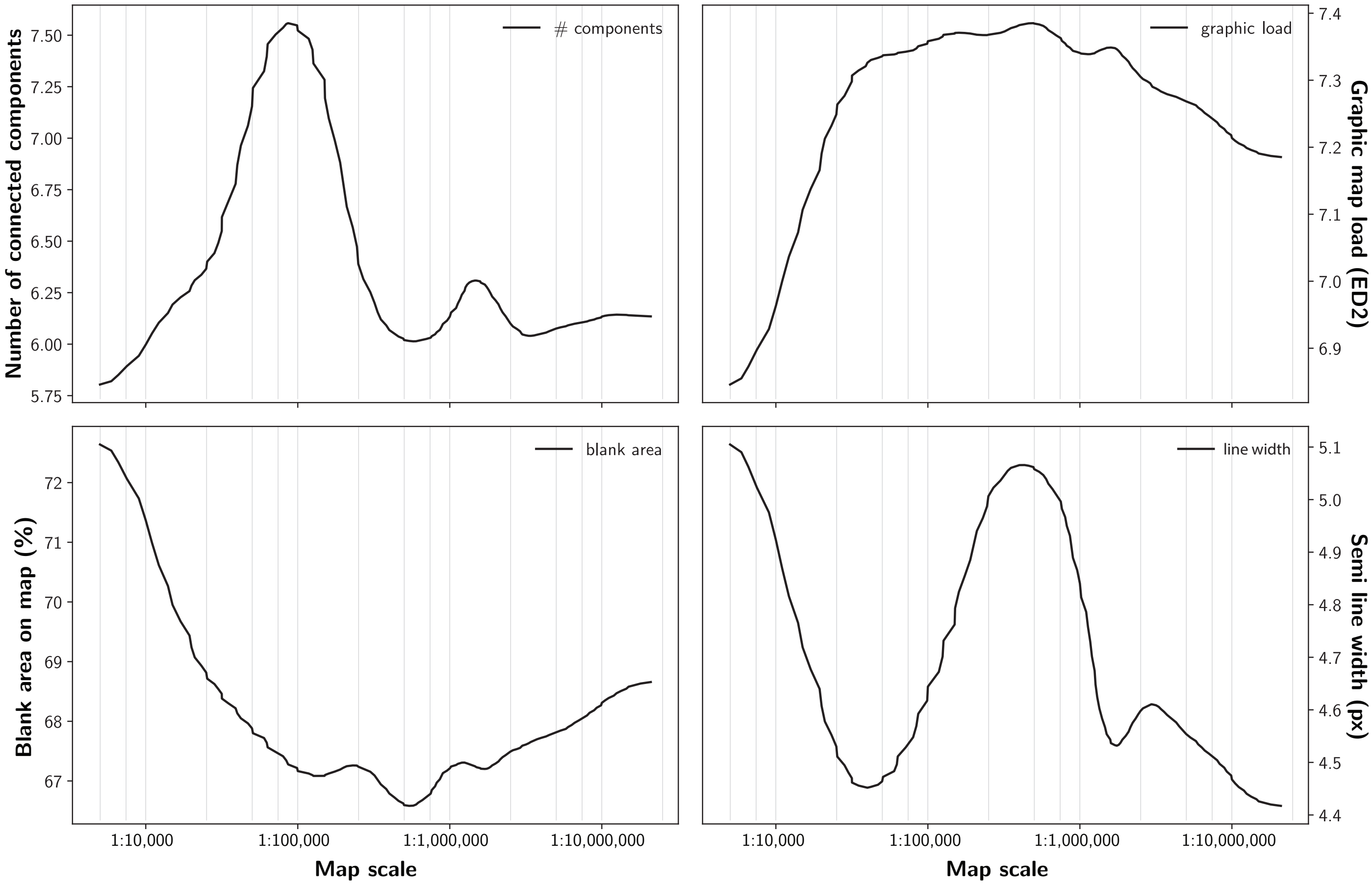

**Figure 18 | Distribution of CV-derived features by map scale.** Each curve represents the mean value of a computer vision derived feature, as a function of map scale. Mean values were computed using sliding strata (141 steps) encompassing ca. 5% of the dataset each. For # components, graphic load, and line width, the average was calculated over all mapels within the corresponding stratum, whereas the share of blank areas was computed at the map level. A visualization of mapel-level CV features is provided in Figure 19. *E.g. blank areas and lower graphic load are characteristic of larger map scales (< ca. 1:10,000).*

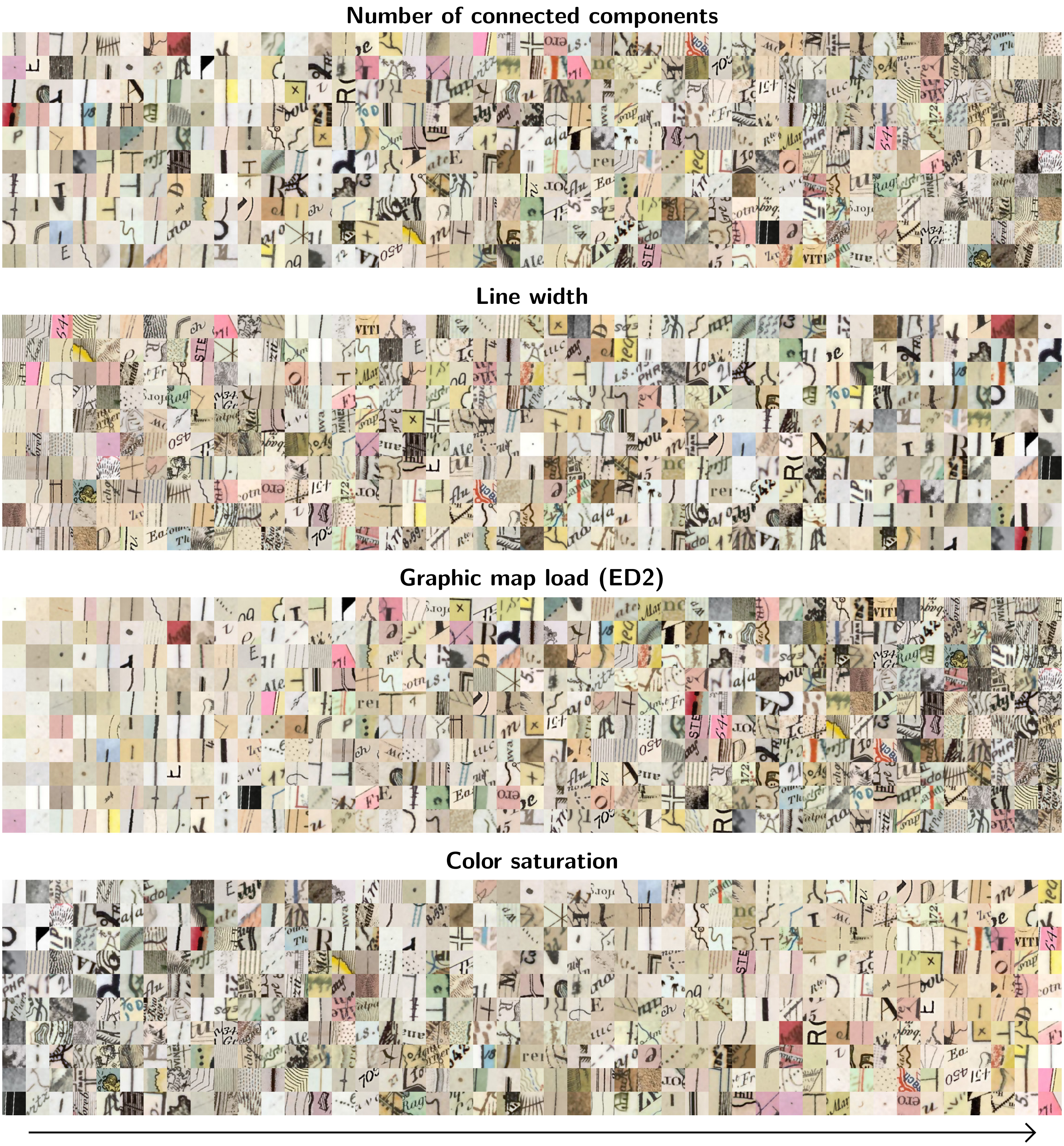

**Figure 19 | Visual interpretation of 4 CV features.** On a random sample of 450 mapels. Each subplot orders the sample according to one CV feature, progressively from left to right. For color saturation, the histogram peak is used. *The effect of ordering mapels according to CV features is visible. The Figure renders these features more interpretable.*

## 7.5 Modeling the circulation of map elements

### *Univocity & the consistency of semiotic systems*

The previous section hypothesized that the structural coherence of the semiotic system is one of the cultural determinants that condition the evolution of map signs. This intuitively applies, for instance, to individual maps. Within each map, the semiotic system should be minimally equivocal, and each sign form should ideally correspond to a distinct meaning. Furthermore, coherence is also expected to be maintained structurally through the existence of social conventions of representations. The consistency of the semiotic system can thus be comprehended as a *selective pressure* that discourages conflicting variations. New signs are expected to be sufficiently distinct from existing, commonly used signs to express a different meaning, and avoid confusion or ambiguity. As such, the cultural structure—or the system that shapes how signs evolve—is partly formed by previous sign utterances, which collectively establish cultural conventions.

A corollary of this hypothesis is that semiotic consistency is expected to be higher at the local level—that is among maps published in the same place, produced by the same community of creators, or during the same period. In order to test this proposition, we define $\Upsilon$, the index of *univocity*, expressing the share of signs which, within a map or a set of maps, are used in similar semantic contexts.

Let us define $X = (x_{n,m,k}) \in \mathbb{N}^{N\times M\times K}$, the *semantic–symbolic count matrix*, with $N$ rows, corresponding to the number of maps, and $M = 4{,}096$ columns, one for each mapel cluster. In this matrix, each of the 25,241,385 extracted mapel is assigned to one of $K = 7$ compositional semantic modes. Here, the *contours* semantic mode is excluded, as it semantically overlaps with other modes.

For each map $n$ and mapel cluster $m$, the *univocity ratio* $u_{n,m}$ is computed as the ratio between the most common semantic mode $k$, and the total number of occurrences within that mapel cluster:

$$u_{n,m} = \frac{\max_{1\le k\le 7} x_{n,m,k}}{\sum_{k=1}^{7} x_{n,m,k}}$$

In other words, the univocity ratio expresses the percentage of occurrences that adopt the most common "meaning", or the operative equivalent thereof, within a given mapel cluster. By extension, the *univocity index* $\Upsilon$ computes the average mapel univocity across a defined set of maps $\mathcal{S}$. Note that only mapel clusters with at least one occurrence in the set are included in the computation of the average.

$$\Upsilon = \frac{1}{M}\sum_{\forall m \mid (\sum_{k=0}^{7} m) \ge 1} \frac{\max_{1\le k\le 7} \sum_{\forall n\in\mathcal{S}} x_{n,m,k}}{\sum_{k=1}^{7}\sum_{\forall n\in\mathcal{S}} x_{n,m,k}}; \qquad (1)$$

Thus, a high univocity index indicates that sign forms are almost always used with the same sense. Conversely, low univocity suggests semiotic ambiguity or the presence of polysemic signs. Table 3 reports the empirical univocity index, computed for distinct subsets of maps, each defined by publication place or period, map scale, or community of mapmakers. Bootstrapping is used to control for the statistical effect of set size on the computed univocity index. The resulting values are therefore comparable, irrespective of the initial set sizes.

The first row shows that univocity within each map is very high (97.44%). This finding suggests that, within a single map, signs are employed with the same signification—determined by the semantic compositional mode—in 97.44% of instances. In contrast, across the entire dataset and period of study, the univocity index is only 74.02%, implying that signs are used with a sense different from their most frequent semantic signification in nearly 26% of occurrences.

The other trials demonstrate that univocity—that is, semantic consistency—is significantly higher within related map sets. For instance, map makers or publishers operating within the same professional community are more inclined to employ identical sign forms to convey an equivalent meaning ($\Upsilon = 81.96\% > 74.02\%$). It is also more probable for maps of similar scale to use signs in the same sense ($\Upsilon = 81.45\%$).

Publication place emerges as the third most important factor ($\Upsilon = 81.16\%$). Table 3 also details the univocity index for the five largest cities, by population in 1850 (see Tab. B1, Chap. 6). With the notable exception of Berlin (82.29%), the index for each of these cities is significantly lower than the publication place average. For New York (72.14%) and Philadelphia (71.64%), the coefficient is even lower than the average computed across the entire dataset. This pattern suggests that semantic–symbolic divergences are more frequent within large urban centers, compared to smaller cities.

Ultimately, maps published in the same country (78.49%), or during the same period (77.40%) are also more likely to share the same conventions of representation. However, these aspects tend to be less predictive than shared community relationships, specific publication place, or map scale.

These results are consistent with the hypothesis that local sociocultural contexts favor greater semiotic coherence. Within individual maps, univocity is minimal, indicating that signs employed to express distinct meanings are nearly separable. The remaining ambiguities, about 2.56%, are surprisingly low, considering that no contextual information is even considered in the case of mapels.

**Table 3 | Index of univocity** $\Upsilon$ computed for distinct sets of maps $\mathcal{S}$. The Table reports the index of univocity $\Upsilon$ as defined in Equation 1. The values correspond to the mean computed over 120 repetitions, together with the 90% Confidence Interval (CI) of the mean. Except for the first row, each repetition is based on a random sample $s \subset \mathcal{S}$ of 200 maps. The first row presents the per-map average, computed over $N = 99{,}693$ maps, and therefore has a negligible CI. For the experiments on communities and publication places, only sets comprising at least 200 maps are considered. For publication place, additional detail is provided for the five most populous cities in 1850. An asterisk (*) indicates that the index is significantly larger than the overall value, computed on the entire dataset (second row).

| Experiment / set | $\Upsilon$ univocity (%) | [$CI_5$–$CI_{95}$] |
|---|---|---|
| Single map | 97.44* | |
| Entire dataset | 74.02 | [72.61–75.54] |
| Community | 81.96* | [73.94–89.99] |
| Scale percentile | 81.45* | [73.46–88.66] |
| Publication country | 78.49* | [70.46–85.05] |
| Publication date percentile | 77.40* | [72.80–81.97] |
| Publication place | 81.16* | [71.79–89.37] |
| – *Paris* | 76.20* | [75.00–77.60] |
| – *New York* | 72.14 | [70.82–73.45] |
| – *Berlin* | 82.29* | [80.78–83.97] |
| – *Philadelphia* | 71.64 | [70.25–73.01] |
| – *Amsterdam* | 77.94* | [76.77–78.97] |

The overall index $\Upsilon$ is lower than the context-specific univocity indices yet remains substantially higher than would be expected by chance. This finding indicates that *most cartographic semiotic conventions were shared globally across the study period*. Nevertheless, the results also imply that map semiotics were *significantly more consistent within the local geographical, social, or historical contexts of their production*.

## *Social transmission*

The previous experiment established the influence of communities on the emergence of shared, consistent, and univocal semiotic systems. Building on this finding, the present experiment evaluates the effect of the social collaboration on micro-transmission. To this end, and to account for the systemic nature of social relationships, the approach is not confined to the annotated communities of map makers but relies on the larger social graph of work relationships constructed in Chapter 2.

Since ADHOC Images is not entirely included in ADHOC Records, as outlined in Chapter 1, the creators, including both map makers and publishers, must first be realigned on the normalized list of creators. The method for doing so is identical to the one described in Chapter 1. First, unique variant names were embedded in a 1,536-dimensional space using GPT text-embedding-3-small API. Variant names were then collapsed into a single normalized name, considering for each variant, the neighboring names located within a cosine distance $\mathrm{d} \leq 0.17$[14]. Names belonging to the same connected component were merged. The resulting reduced and aggregated graph $G$, composed of both ADHOC and ADHOC Images, comprises 238,603 unique entities[15].

The aim of this experiment is to predict the semiotic distance between map creators, based on the social graph $G$, and the historical context of production. The target response of the model is $\delta_{ij}$, the dyadic difference between the semantic–symbolic distributions of signs in the works of map makers $i$ and $j$[16]. Only map creators present in the ADHOC Images dataset who participated in the publication of at least 10 maps were considered when forming dyadic pairs (n = 1,435).

Formally, the analysis considers the following candidate variables:

- $\mathbf{1}_{ij}^{connected}$ a boolean variable indicating whether the two creators have a direct collaborative relationship, that is whether they have worked together on the creation of a map.
- $d_{ij}^{shortest\ path}$ the number of intermediate nodes separating the two creators on the graph $G$
- $d_{ij}^{weigh.\ shortest\ path}$ the shortest weighted path separating the two creators in the graph $G$. Here, weights between node pairs are inversely proportional to the number of maps on which the creators have collaborated.
- $d_{ij}^{production}$ the absolute difference between the number of maps produced by creators $i$ and $j$
- $min_{ij}^{production}$ the lowest number of maps produced, between the creators $i$ and $j$
- $max_{ij}^{production}$ the largest number of maps produced, between the creators $i$ and $j$
- $d_{ij}^{time}$ the absolute difference between the average publication date
- $min_{ij}^{time}$ the shortest time difference between publication dates (e.g., the variable equals 0 if the two creators published at least one map in the same year)
- $d_{ij}^{geo}$ the geodesic distance between average publication places
- $min_{ij}^{geo}$ the shortest geodesic distance between publication places
- $\mathbf{1}_{ij}^{long\ dist}$ a boolean indicating if average publication places are more than 3200 km apart

[14] This threshold was defined experimentally in Chapter 1.
[15] Versus 236,925 for ADHOC Records alone.
[16] The reason why we do not use $\rho_{ij}$, the dyadic coefficient of semiotic rupture in this experiment is due to sample size. Indeed, normalization by the 95th percentile tends to be unstable with small sample sizes, e.g. with the semantic–symbolic distributions of less prolific creators. The computation of $\delta_{ij}$ is very similar to that of $\rho_{ij}$ but adopts the standard approach to compute distributions, normalizing sign counts by the total number of occurrences for each semantic mode.

Each predictor variable was normalized to range between 0.0 and 1.0, after which the following linear model were defined:

$$\tilde{\delta}_{ij} = \beta_0 + \beta_1 \mathbf{1}_{ij}^{connected} + \beta_2 d_{ij}^{shortest\ path} + \beta_3\, rank\left(d_{ij}^{weigh.\ shortest\ path}\right) + \beta_4\, rank\left(d_{ij}^{production}\right) + \beta_5\, rank\left(min_{ij}^{production}\right) + \beta_6\, rank\left(max_{ij}^{production}\right) + \beta_7 rank\left(d_{ij}^{time}\right) + \beta_8 min_{ij}^{time} + \beta_9 rank(d_{ij}^{geo}) + \beta_{10} min_{ij}^{geo} + \beta_{11} \mathbf{1}_{ij}^{long\ dist} + \varepsilon_{ij} \quad (2)$$

The model was fitted to a sample of 1,435 dyadic pairs to avoid redundancy and to ensure independence between observations. The results are reported in Table 4. Only statistically significant predictors were retained in the final model. Compared with the null model ($SS_{res} = 717.0$), the full model explains 48 percent of the variation in semiotic distance. About half of the variance thus remains unaccounted for.

The most influential predictor appears to be the degree of map producers. The negative $\beta_6$ coefficient indicates that *when at least one of the two creators is a large map producer*, the semiotic distance between the two creators *tends to be reduced*. This finding may be interpreted as evidence of the cultural influence exerted by large producers over smaller ones[17].

Four other predictors are also significant and exhibit smaller effects. First, degree differences between map creators tend to result in more distinct semiotics, whereas producers of comparable size tend to be more alike. Second, map makers whose periods of activity nearly overlap appear to employ more similar signs. Third, map makers that have either directly collaborated on the publication of several maps, or are indirectly connected through strong intermediate colaborations, seem to produce maps that are semiotically more similar. Finally, the fourth predictor supports the intuition that maps produced in geographically remote locations are more likely to differ in semiotic terms.

[17] It could be counterargued that this statistical effect is due to the higher intermediacy of distributions computed from larger sample sizes. However, this interpretation can be dismissed with a simple verification step. Indeed, even when subsampling creator map subsets to 10—i.e. the minimal considered sample size—the predictor $rank\left(max^{production}\right)$ remains statistically significant and prominent. Furthermore, the value of $max^{production}$ is not entirely based on image counts, since it also integrates published map records referenced in the ADHOC Records dataset.

**Table 4 | Determinants of semiotic distance among map creators**. The response variable is the observed difference between semiotic distributions. Only predictors with p-values < .025 are retained in the final model.

| predictor | $\beta$ | Sum sq. | df | p-value |
|---|---|---|---|---|
| $rank(max^{production})$ | –0.004 | 51.0 | 1 | < 0.001 |
| $rank(d^{production})$ | 0.001 | 6.2 | 1 | < 0.001 |
| $\mathbf{1}^{long\ dist}$ | 0.243 | 5.3 | 1 | 0.001 |
| $rank(d^{weigh.\ shortest\ path})$ | 0.001 | 4.2 | 1 | 0.005 |
| $min^{time}$ | 0.069 | 3.4 | 1 | 0.011 |
| Residual | | 371.6 | 711 | |

The results of this experiment on social transmission suggest that, in cartography, the influence of large map producers on their peers is the most substantial determinant of the adoption of semiotic conventions. The impact of legitimacy and marketing was already qualitatively highlighted in Chapter 2; the present findings provide evidence of the effect of commercial scale. Larger actors, who were able to invest in ambitious cartographic campaigns, productive printing infrastructure, and extensive distribution networks also appear to have set semiotic conventions that were then replicated by smaller workshops. Furthermore, the previous results on univocity indicate that such replication was probably necessary for smaller map producers to remain legible.

The other core results, which highlight the importance of collaborative relationships, comparable actor size, temporal overlap, and the negative impact of geographical remoteness appear consistent with the expectations and literature on the diffusion of ideas (Comin et al., 2012; Rogers, 1995; Singh, 2005). Ultimately, it is worth mentioning that temporal and geographic variables may account more for indirect transmission, e.g. through the medium of maps, that direct social transmission.

### *Extended diffusion model of cartography & Macro–transmission*

In this final section, the diffusion model established in Chapter 6 will be adapted to the case of mapels, augmented through the integration of semantic compositional modes, and extended diachronically to account for the temporal macro–dynamics of cultural transmission.

The diffusion experiment conducted in the previous chapter aimed to measure and model the determinants of semiotic rupture between map production centers. The model included seven candidate predictors: the geographic location and distance between centers ($d_{ij}^{geo}$, $\mathbf{1}_{ij}^{country}$, $\mathbf{1}_{ij}^{continent}$), the period of activity ($d_{ij}^{time}$), the relative population sizes of urban centers[18] ($d_{ij}^{pop}$, $min_{ij}^{pop}$), and whether the two centers share the same majority language ($\mathbf{1}_{ij}^{language}$). The response variable was defined as $\rho_{ij}$, the dyadic coefficient of rupture. Only cities represented by at least 200 maps in the ADHOC Images corpus were considered. Extending the approach to semantic–symbolic representation mainly requires adapting the computation of $\rho_{ij}$ so that it reflects additionally the relative sign frequencies for each semantic mode.

[18] Documented estimates for the year 1850.

The resulting model is reported in Table 5. Most of the variance in the resulting model is explained by city-specific effects rather than global variables. Nevertheless, the period of activity, country, and urban center size were retained as statistically significant predictors. Together, these indicators accounted for 37% of the variance not captured by dyadic city effects. Globally, cities that were active map-publication centers during a similar period tended to produce maps that are semiotically more similar. The same effect is observed for cities located within the same modern country. These two variables therefore constitute the most salient determinants of semiotic similarity when the model is applied to the general case of mapels. Finally, the population size of urban centers ($min^{pop}$) is also significant, although its effect is weaker; the positive coefficient suggests that *map semiotics tend to diverge more between large cities.*

These results are consistent with those obtained in the previous chapter. Although they do not rely on the same variables, both models suggest that time is the primary determinant of semiotic rupture. Both also indicate that greater geographic distance or, in the present model, the presence of national borders tends to increase semiotic divergence. Finally, both models highlight a pattern of cultural distinction among large urban centers.

**Table 5 | Determinants of semiotic rupture between urban centers**. The response variable is the empirical semiotic rupture observed between city pairs. Only predictors with p-values < .025 are retained in the final model. The indices *i* and *j* account for city-specific determinants.

| predictor | $\beta$ | Sum sq. | df | p-value |
|---|---|---|---|---|
| $d^{time}$ | 0.19 | 0.56 | 1 | < 0.001 |
| $\mathbf{1}^{country}$ | 0.08 | 0.45 | 1 | < 0.001 |
| $min^{pop}$ | 0.10 | 0.04 | 1 | 0.005 |
| $i, j$ | | 11.41 | 46 | |
| Residual | | 1.91 | 502 | |

This diffusion model adopts a synchronic view of map semiotics. However, because time emerges as the most distinctive marker, a diachronic perspective could enrich the analysis by elucidating the dynamics of cultural transmission. Figure 20 presents a diachronic flow chart that depicts the 24 map-production centers considered in the preceding analysis across six time strata. The strata were derived through equal stratification, so that each stratum contains the same number of maps. In this visualization, the radius of each node corresponds to the number of maps published by a given production center during the corresponding period. The width of each edge is inversely proportional to the coefficient of semiotic rupture $\rho$, and can thus be interpreted as a measure of cultural similarity, or, within a diachronic framework, *cultural transmission* and the semiotic influence of a former cartographic school. Figure 20 reports, for each time step and city, the centers whose map production in the preceding and subsequent strata most closely resemble the cartographic figuration of that city.

Examining the first two time strata, one can observe that the semiotics of maps produced in Nuremberg before 1736 appear to have influenced the semiotics of maps published in Paris between 1736 and 1827. The influence of the school of Nuremberg also locally persisted in Nuremberg itself, and furthermore affected the cartographic conventions in Amsterdam and Augsburg, although not as decisively as for Paris.

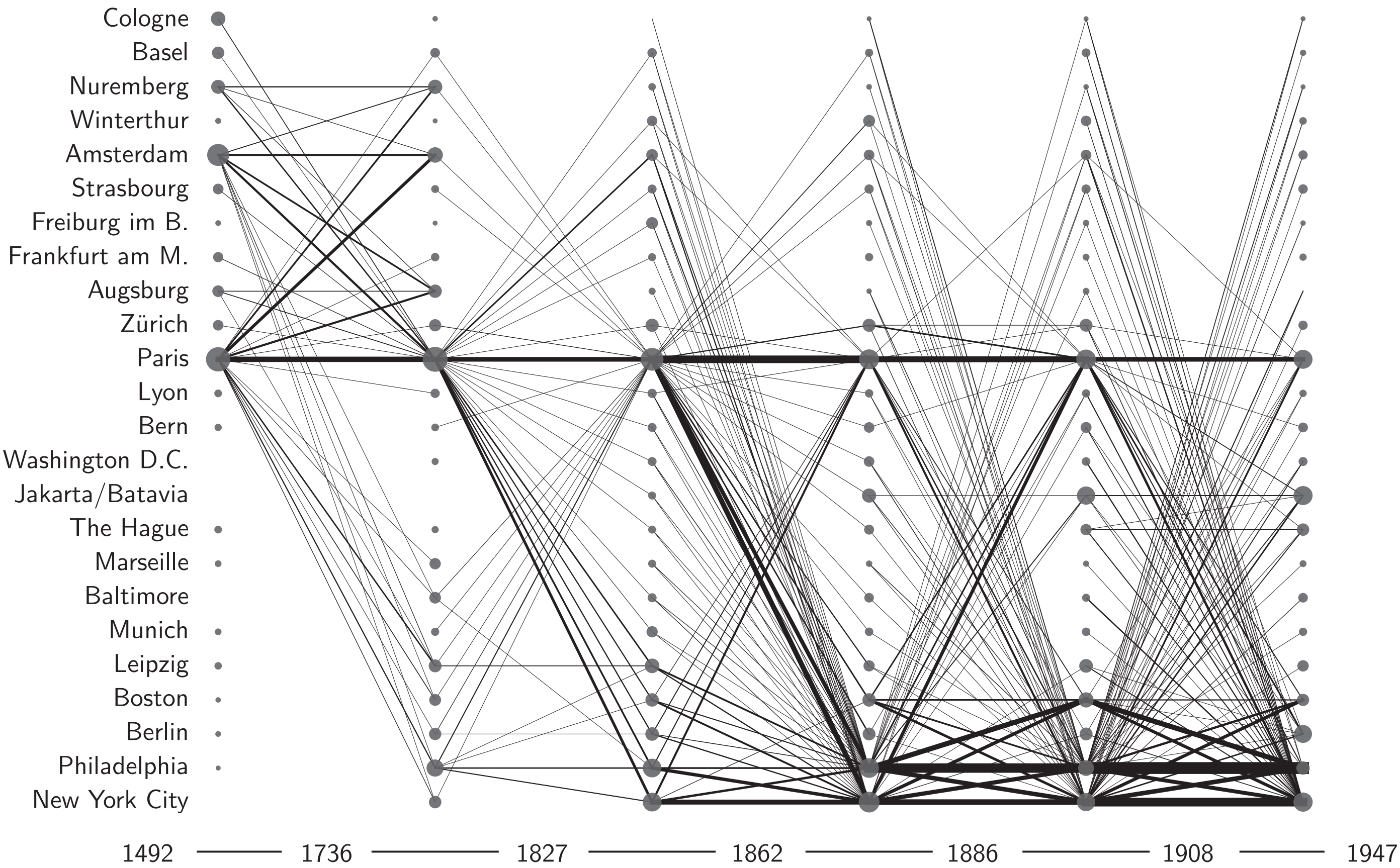

**Figure 20 | Diachronic chart of semiotic flow between urban centers.** Each row corresponds to an urban center represented by at least 200 maps. The horizontal axis delineates six equal temporal strata derived from the overall map distribution. The area of each circle is proportional to the number of map records in ADHOC Images for the corresponding city and time stratum. Edge width is inversely proportional to $\rho$, the observed coefficient of rupture between the two semiotic distributions. Edges are shown only when the corresponding coefficient $\rho$ is significantly lower than the average rupture coefficient computed across all possible pairs. *Paris, New York and later, Philadelphia emerge as hubs. They exhibit strong self-transmission patterns while also broadcasting signs from and to other publication centers.*

Overall, and particularly from the early 18th century onward, a small number of large cities functioned as *hubs*. Paris, followed by New York from the mid-19th century and Philadelphia during the latter half of the 19th century, were connected to nearly all other cities. Transmissions are evident both from smaller cities to larger ones and from larger cities to smaller ones. This indicates that, in most instances, signs were relayed through these hubs before being broadcast to smaller map-publication centers. Horizontal connections, in contrast, can be interpreted as the indication of self-sustained semiotic culture. This property is hereafter designated as *autopoiesis*[19]. In biology, *autopoiesis* indicates the capacity for an entity to replicate itself, to continuously renew and sustain itself, in interaction with its environment, to maintain its own structure and boundaries, despite constantly changing fundamental components (Maturana & Varela, 1980). In the present context, in which the components are ideas and signs rather than biochemical materials, *autopoietic hubs* appear to exhibit analogous qualities. The following paragraphs are dedicated to their description.

In the first time stratum, the cities exhibiting horizontal self-transmission are Paris and Amsterdam, with Nuremberg and Augsburg, although less markedly. Paris is the only city in the study that maintains a strong semiotic identity throughout the entire period, as evidenced by continuous horizontal self-transmission. Two other cities that emerge as strong autopoietic hubs are New York and Philadelphia. In Philadelphia, self-transmission is observed from the 1736–1827 stratum onward. At that time, the city appears to have been influenced by European cartographers, especially the schools of Paris and Amsterdam. The influence of Philadelphia on other cities only became substantial from 1862–1886 onward, implying that its cartographic output acquired a stronger identity, rivaling both Paris and New York. New York cartographic culture seems to originate between 1827 and 1862, initially under the influence of Paris. From that moment on, New York seemed to have developed as an independent cartographic school. Sustained self-transmission appears to constitute a prerequisite for global influence. A common characteristic of autopoietic hubs is that self-transmission—i.e., self-making and self-replication—is stronger than transmission to or from any other city. For instance, transmission between New York and Philadelphia during 1862–1947 ranks among the strongest relationships observed, yet the influence each city receives from the other remains lower than its own self-transmission over the same period.

In this regard, several cities—foremost Boston, but also The Hague, Jakarta/Batavia, Berlin, and Zürich—exhibit incipient autopoietic cartographic cultures. Nevertheless, the influence they receive from larger hubs is consistently equal to or greater than their own horizontal self-transmission. An apparent consequence is that (1) self-transmission tends to remain negligible or even declines, and (2) these cities do not evolve into hubs, i.e., they fail to acquire significant influence over smaller production centers. For example, between 1736 and 1827, one can observe significant semiotic flows from Paris to Amsterdam, Nuremberg, and Augsburg. These flows exceeded the respective self-transmission observed during the same period in these cities, suggesting that Parisian cartographic conventions may have *superseded* local semiotic practices. Indeed, the corresponding edges indicate that the maps produced in these former hubs during the second time stratum are more closely related to the maps produced in Paris during the first stratum than to those produced locally at the same period.

[19] Literally “self–making”.

While signs appear to transit through *hubs*, they do not always seem to originate there. Indeed, the cities in Figure 20 are ranked according to the ratio of total posterior similarity to total anterior similarity. The top-ranked city, for instance, is Cologne. The school of Cologne regularly influenced Paris, New York, and Philadelphia; however, the relative influence it received from these three major hubs—or from any other city—is comparatively low. In this respect, one may deduce that Cologne *generated more new signs than it assimilated*. Using the same criterion, Basel, Nuremberg, Winterthur, and Amsterdam may also be considered innovators. In contrast, Leipzig, Boston, Berlin, Philadelphia, and New York, which occupy the bottom of the ranking, are distinguished by their capacity to efficiently *integrate* new signs. Paris is situated near the middle, combining effective integration with strong retransmission—including to American cities—and regular innovation (e.g., flow mapping, aerial photography, scientific cartography). It is noteworthy that the cost of early adoption apparently exceeds that of innovation, as illustrated by the decline of pioneering map-printing centers such as Nuremberg. This finding is consistent with the literature (Golder & Tellis, 1993; Tellis & Golder, 1996).

Cities ranked near the middle tend to function as "dialogue partners" to larger hubs, i.e. they influence each other's mapping practices. One manifestation of this pattern is the relationship between Zürich and Paris. Large hubs can maintain numerous such dialogue partners, with whom they exchange signs. Moreover, larger hubs often function as dialogue partners to one another.

To synthesize, some secondary centers generate new signs but are able to disseminate them only through larger hubs, whereas others contribute little to innovation but are able to integrate signs replicated by major hubs and retransmit them later to other—or the same—hubs. As such the results do not describe a transmission model in which signs invariably flow from larger source nodes (e.g., influencers) to smaller sink nodes (e.g., followers); in such a model, signs would not return from secondary nodes to the hubs. The present findings rather suggest a system in which smaller cities act as peripheric centers that help make and shape the semiotic system of larger cities, in a decentralized manner.

When interpreting this finding, one should recall that the making process is inherently dynamic. Horizontal self-transmission, therefore, does not entail the perpetuation of a stable, unchanging, idiosyncratic semiotic system. Similarly, signs are not merely reverberated on peripheric centers; they undergo cultural integration, and thus modification, before being propagated again. Each time step consequently involves variation and semiotic evolution. The implications of these results become clearer when related to the univocity of semiotic systems. Among the five largest cities in the dataset, Berlin is the only one that fails to develop as an *autopoietic hub*; it also exhibits by far the highest univocity. Lower univocity indicates more variable semiotic systems. The existence of more diverse semiotic systems, in turn, implies that the associated conventions are comparatively less prescriptive and definitive, and accordingly more adaptable. The present experiment thus indicates that the centers exhibiting higher semiotic variability are also the ones demonstrating greater effectiveness at integrating new signs.

A most notable—and arguably puzzling—result is that autopoietic centers, while centralizing signs from many other centers, and being themselves influenced by other major hubs can nonetheless maintain their own cultural idiosyncrasy and cultural specificities. This observation derives not only from the self-transmission patterns illustrated in Figure 20, but also from the results reported in Table 5 and Chapter 6, all of which reinforce the idea that larger urban centers tend to develop significantly more distinctive semiotic systems. How? The answer may be related to their ability to integrate new signs efficiently. Indeed, cultural evolutionary models show that acculturation rate constitutes one of the most significant factors for the perpetuation of between-group cultural variation (Mesoudi, 2018). In this perspective, the capacity of autopoietic hubs to accommodate and adapt exogenous signs within their own semiotic systems may help them maintain their local cultural structure. The introduction and local adaptation of such sign variants might even concur to the cumulative making of their own distinctive and recognizable cultural identity.

In contrast, cities with lower cultural cumulation rates do not appear to differentiate themselves sufficiently and thus seem unable to position themselves as audible dialogue partners from the perspective of smaller centers. One might argue that limited transmission, on the contrary, denotes outstanding independence. This interpretation, however, is undermined by evidence that these centers even fail at being audible to themselves and hence to perpetuate their own semiotic idiosyncrasy.

# Appendix A – Supplementary Materials

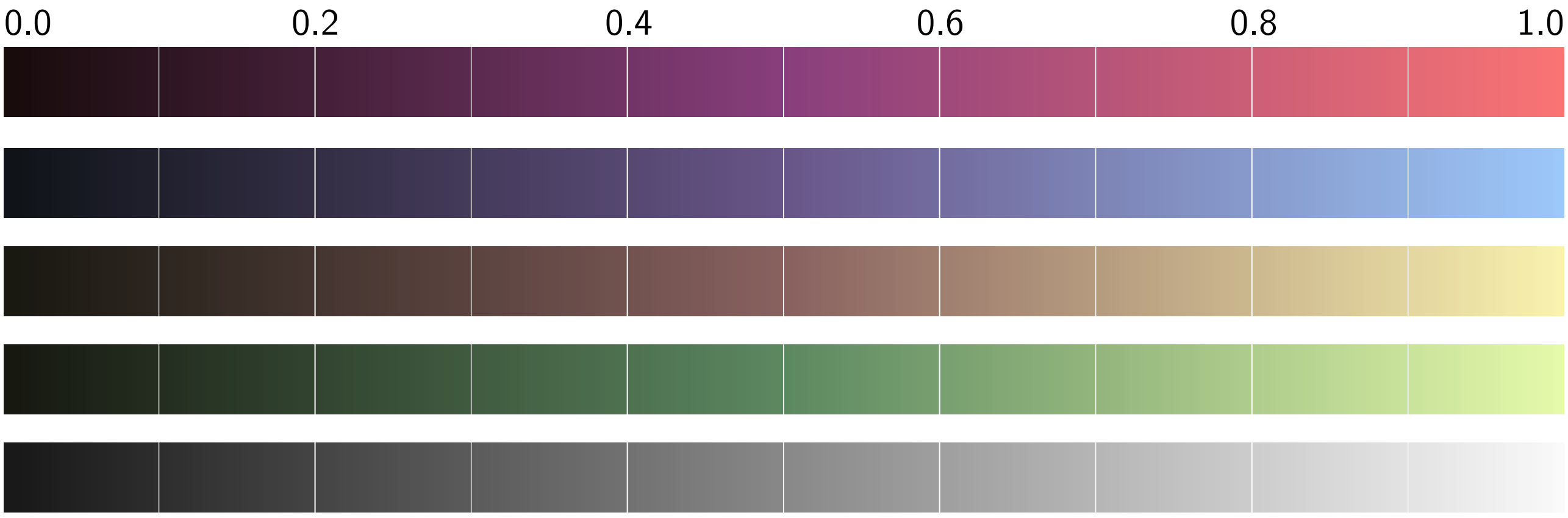

**Figure A1 | Color scales used for the semantic-symbolic frequency maps.**

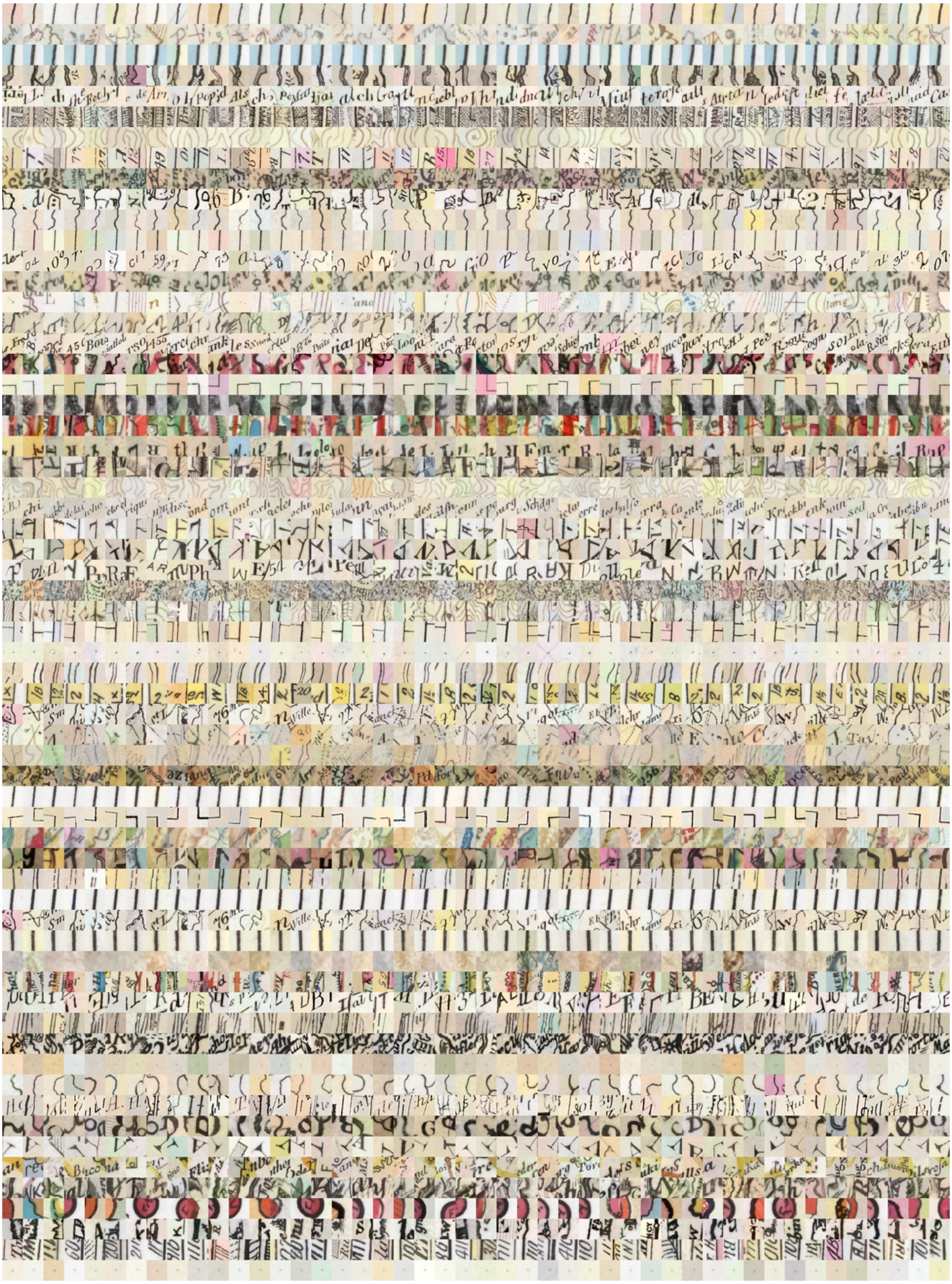

**Figure A2 | Examples of 62 mapel clusters.** One cluster per row. For each row, mapels are horizontally arranged from left to right by their probability of belonging to this particular cluster. The first mapel is the exemplar. *Mapels clusters exhibit a relatively high consistency; the exemplar is representative of each cluster.*

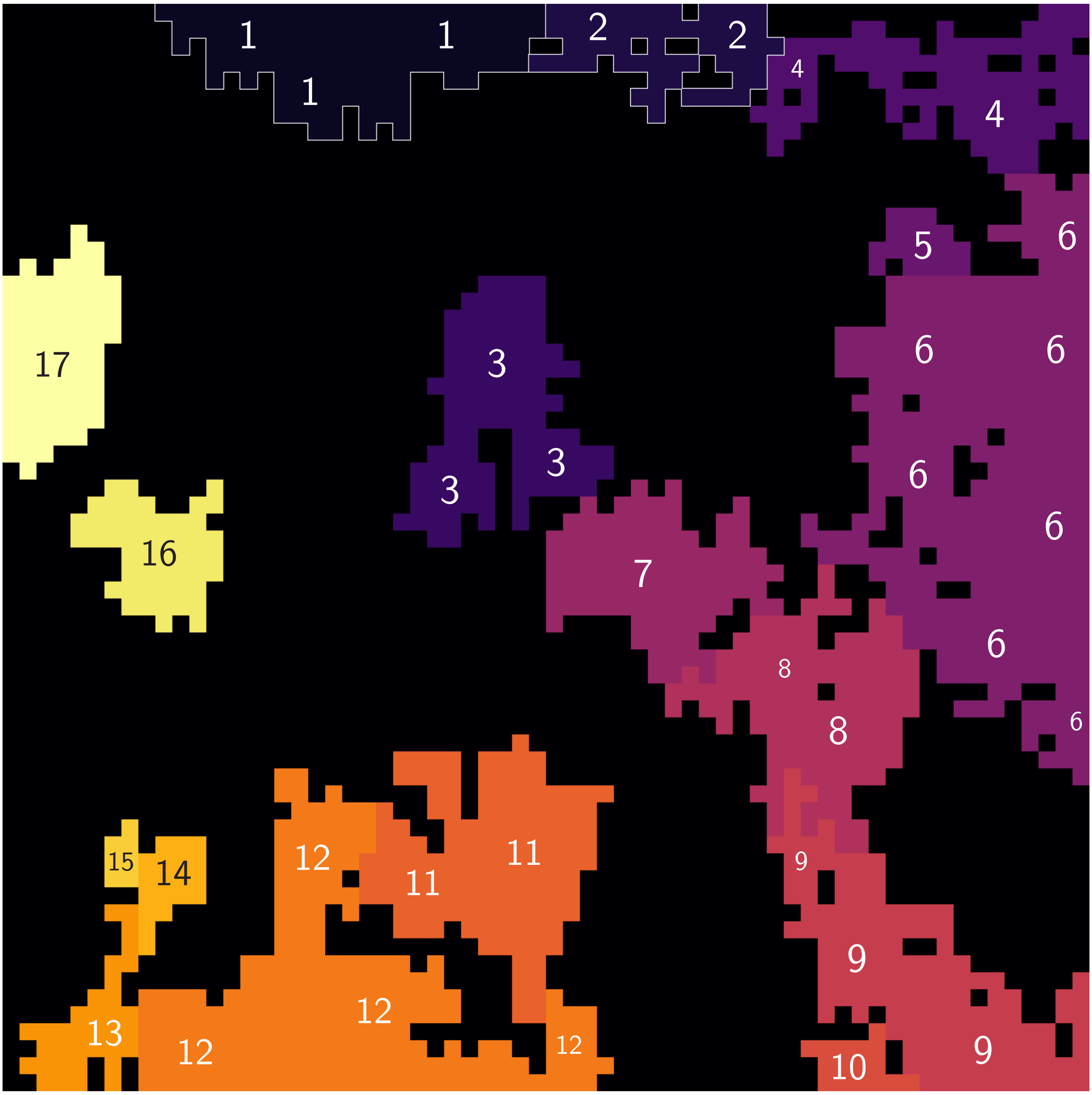

1. blue
2. lavender
3. deep purple
4. light purple
5. turquoise
6. mint green
7. fuchsia
8. pale pink
9. sage green, salmon, flax
10. green moss
11. terracotta
12. blood orange, cherry
13. brown
14. muddy cherry
15. kaki
16. burgundy
17. teal

**Figure A3** | **Accessible version of Figure 5.** With delineation of the color regions referenced in the text.

# General discussion & Conclusion

The purpose of this final chapter is to summarize the contributions and trace links between the results and ideas discussed in the preceding chapters. The present conclusion will also consider these achievements in light of the overarching aim of the dissertation and the research objectives (ROs) stated in the introduction. Finally, it will provide closure to the thesis, discuss its theoretical implications, define its limitations, and outline future perspectives.

As stated in the introduction, the overarching aim of this thesis was to *create datasets and develop methodologies to investigate cartographic heritage on a large scale from a cultural perspective.* This aim entailed three distinct dimensions: first, the creation of datasets for historical cartographic scholarship; second, the development of approaches to model and analyze historical maps; and third, the demonstration of the discovery potential of the latter two contributions through the interpretation of results and the discussion of hypotheses.

## C.1 Contributions to the Research Objectives

### *Datasets & techniques*

RO1 Chapter 1 offers the most significant contribution to the creation of aggregated historical map databases. The first output is a set of 771,561 map records, representing effectively 1.4 million maps cataloged in 38 heritage institutions across 11 countries; the second set comprises 99,715 digital images with associated metadata. Together, these databases span 457 years of cartographic history, from 1492 to 1948. In addition to publication year, they include structured information on authorship, place of publication, geographic coverage, and map scale (Petitpierre, 2025a).

Three associated datasets are derived from the digital images: (1) 99,715 semantic masks encoding classes of geographic objects; (2) 63,180,211 cartographic signs, i.e, icons and symbols, and their attendant embeddings; and (3) 25,241,381 map elements (mapels), conceived as generic, operable units for the study of cartographic figuration.

RO2 By developing methods for the extraction and recognition of map images, this thesis also contributes to the creation of two significant training datasets. The first, Semap, comprises 1,439 manually annotated map samples for the segmentation of six cartographic classes: built, non-built, water, road network, contours, and background. It is released in open access together with the thesis, constituting the largest and most diverse open dataset for map segmentation (Petitpierre et al., 2025). These manually annotated data are further complemented by 12,122 synthetically generated samples (Section 4.2).

The second dataset, the Cartographic Sign Detection Dataset (Petitpierre & Jiang, 2025), comprises 18,750 manually annotated icons and symbols located in 796 distinct map samples (Section 6.1). It can serve as a benchmark for the emerging task of map-sign detection and may also be employed as pre-training data for specialized sign-detection models.

Accordingly, the work also delivers two trained models for the detection of map sign and the semantic segmentation of historical maps, respectively. On the HCMSSD–Paris and HCMSSD–World benchmarks, the Mask2Former-based segmentation method surpasses the previous state of the art by 9.8 and 30.9 percentage points, respectively, owing to a pretraining strategy leveraging both Semap and synthetically generated data, as well as the implementation of a multiscale integration mechanism at inference time (4.4).

RO3 The semantic segmentation is instrumental for the conditional analysis of cartographic figuration. This aspect becomes more prominent in Chapter 7, which introduces a generic methodological framework for the analysis of figurative cues, including linework, textures, and colors (7.1). This approach relies on the fragmentation of images into discrete map elements, or mapels. As a set of graphically salient image samples, mapels collectively provide an operable representation of map figuration. An experimental setting demonstrates the ability of mapels to encode culturally relevant variables, like style (7.2). When associated with semantic ratios, mapels permit the analysis of map semiotics.

### *Analysis of publication volumes*

RO4 The first practical demonstration of the value of assembling large datasets for the study of historical cartography was provided in Chapter 2, through the systematic mapping of publication centers (2.2). The analysis uncovered distinctive production patterns, either polycentric, i.e. broadly distributed across multiple centers, as in the case of the United States, Germany, Italy, and Switzerland or, conversely, heavily concentrated in the capital, as for France, Spain, and Japan. It also revealed larger geographic structures, most notably the Rhine Valley, the Northeast Corridor in the United States—extending westward into the present-day Rust Belt—and the Venice–Milan axis in Italy.

The examination of chronologies (2.3) suggested a relationship between strong, centralized, sovereign political structures and sustained volumes of cartographic publication. In France, Spain, and Denmark, periods of absolute monarchy appeared to coincide with high cartographic activity. American output increased only after recognition of the nation's independence, whereas Australian publication expanded following the grant of legislative autonomy. In Japan, cartographic production rose during the Meiji era.

RO5 Chapter 3 introduced the construct of *spatial attention*, recognizing mapping as a directed effort requiring sustained resources (3.1). The visualization approach highlighted two distinct forms of attention: *close* and *distant*. Close attention, corresponding to detailed maps, affords direct control over the territory, facilitating its organization, exploitation, and defense. Distant attention—corresponding for instance to country and world maps—primarily participates to the diffusion of world models and spatial ideologies.

The visualization of spatial attention through the geographic foci of maps (3.2) indicates that all countries primarily direct their cartographic production toward their own national territories and political or economic capitals. In Europe, they also tend to depict neighboring regions. Nations such as France, the Netherlands, Spain, Portugal, and Italy also concentrate on historical Western colonies, chiefly in the Americas, Southeast Asia, and insular regions. Beyond its current territory, the United States exhibits a sustained—albeit distant—attention to South America and a closer attention to Western Europe. Australia, and even more importantly the United States, allocate a notable share of maps production to major Western cities—like Paris, London, and New York—underlining the role of these centers as cultural references. Notably, Germany and Japan distinguish themselves by dedicating substantial distant attention to Europe and the Asia-Pacific region, respectively, a pattern that may be linked to their expansionist ambitions in the early 20th century, when their cartographic production peaked.

A distinct, complementary approach to the examination of geographic focus, based on semantic masks, was presented at the end of Chapter 4. The analysis evidenced the disproportionate cartographic representation of landmasses relative to seas, even at very small map scales. It also revealed a predominance of urban over rural or natural landscapes, particularly from the early 18th century onward and within American cartography.

RO6 The intuition that maps participate in the cultural construction of worldviews and can reinforce political control over a territory motivated further investigation of the quantitative relationship among map publication, geographic foci, and historical-political developments such as colonization, wars, and the construction of modern nation-states. In this respect, Section 3.4 presented significant evidence of a conjunctural relationship between Atlantic charting and the triangular trade. Along with the observed close attention to the colonies (3.2), this finding suggests a mutually reinforcing feedback loop whereby maps facilitated colonial expansion, which in turn fueled additional map production.

With respect to the construction of modern nation-states, besides the noteworthy emergence of capitals, like Berlin, Bern, Tokyo, and Washington D.C., as publication centers, from the second half of the 19th century (2.2), the results also showed the substantial progression of self-attention, corresponding to a domestic geographic focus, between 1600 and ca. 1880 (3.3). Domestic maps offered increased control over the territories they depicted and participated in the social and political construction of the nation.

The strategic control of space offered by maps is also instantiated in the strong correlation between the intensity of military conflicts and the volume of map publication during the first half of the 20th century (3.5). Indeed, cartography constituted a strategic military technology, and as such, most of the largest map producers during this period were related to the armed forces or government agencies.

RO7 The prominence of military-related mapmakers is also visible in the analysis of map creators presented in Chapter 2 (Section 2.4). The textual analysis of semantic markers highlighted the surge of private mapping companies in the latter half of the 19th centuries, chiefly in the United States. The statistical inquiry into the social graph of map makers, however, suggested that creator typology (e.g., private company, military, scientific, administration) do not substantially influence the structure of the network of collaborations. Instead, geographic factors, like the city and the country of publication, appear to be stronger determinants of connectivity. Creators active within the same period of approximately 50 years similarly exhibit a higher probability of collaboration. A more informed understanding of the social structures of mapmaking may help delineate more potent research scopes within historical studies.

### *Analysis of map images & design*

RO8 An alternative approach to studying historical maps is to group them by topic or cartographic *mode*. Accordingly, Chapter 5 examined the possibility of categorizing maps into distinct types that reflect their semantic composition, i.e., the arrangement of the different classes of geographic objects within the image. The clustering approach yielded eight distinct types, including urban, maritime, chorographic, and diagrammatic maps. Although these types were found to be neither entirely distinctive nor exhaustive, considering the diversity of maps, the analysis revealed differential expression patterns as a function of historical context of production, map scale, and area covered.

RO9 Bulding on the premise that maps are cultural products, Section 5.2 introduced the viewpoint that they are also designed images whose framing and composition reflect purposeful choices and cultural conventions. Dividing each map into nine *quadrants* enabled a systematic analysis of semantic distribution and spatial relationships within the image. The results indicated that built—e.g., urban—areas and the road network tend to occupy the center of the image. Contours—which denote dense instantiation or subdivision of space—and non-built classes also appear prevalent at the center. By contrast, water and background elements are more likely to be located at the periphery of the representation.

The study of spatial relationships confirmed the presence of compositional patterns, like centering, and circumjacent (as opposed to radial) semantic regularities. The statistical inquiry also demonstrated a central cross pattern, and horizontal symmetries that cannot be accounted for by geographic factors alone. The presence of horizontal symmetries is particularly striking, given that a majority of maps were historically oriented horizontally, implying that horizontal long-range relationships outweigh the effect of additional distance caused by document orientation. The horizontal orientation of cartographic documents may be attributable to the early processual relationships between mapping and landscape iconography. However, the imbalance persisted well into the modern era, up to the early 20th century, suggesting the perpetuation of horizontality through cultural conventions of representation. This mechanism of normative reproduction probably also explains part of the semantic dependencies discussed above, suggesting that map composition is as much a reflection of intentional design choices as it is a manifestation of cultural norm.

### *Cartographic signs*

RO10 The final two chapters focused on the investigation of cartographic signs, like icons and symbols, as well as texture, linework, color, and other figurative features, through the study of mapels. The digital study of cartographic signs presents two principal challenges: first, the definition of a culturally pertinent space of representation; and second, designing approaches that enable analysis at scale while preserving interpretability.

The embedding methodology was based on the modification of a contrastive vision model, namely DINOv2. The decision to adapt a pretrained model was justified by the relative iconicity of map signs—that is their similarity to the physical environment—and their symbolic simplicity (6.2). For the embedding of cartographic icons and symbols, i.e., point signs, the model was additionally affixed an adapter module that emphasizes the icon's foreground and regulates the impact of background color (6.3). The cultural pertinence of signs was assessed from multiple, complementary perspectives, demonstrating that the embedding can model the social structures of work collaborations among mapmakers, as well as semantic categories (6.4–6.5, 7.2). The latter property was verified in two distinct ways: using manually labeled categories in the case of cartographic signs and employing segmentation masks for mapels. Furthermore, for point sign, the latent representation demonstrated the ability to encode interpretable dimensions, including color and icon shape (6.6).

The approach for analysis relies on the identification of multiple *exemplars* (6.7). This strategy hinges on the discretization of the space into clusters, each represented by an exemplar. Clustering facilitates the interpretation of the temporal and spatial distribution of signs within the latent space, through t-SNE-constrained visualizations, called *mosaics* (6.7, 7.3).

The analysis of distributions enables the identification of *moments of rupture*, which indicate transitions in cartographic figuration (6.10, 7.4). For icons and symbols, the most significant rupture occurs around 1789, whereas the analysis of mapels highlights three pronounced inflection points, near 1625, 1785, and 1865. Secondary ruptures also appear around 1877 and 1887. Apart from the 1865 shift, the two independent experiments agree on the timing of transition periods.

Beginning in the European Renaissance, the research evidenced a progressive diminution of iconic forms in favor of symbols, culminating at the end of the 18th century with the rise of scientific cartography. This transition was manifest, for instance, in the spread of new relief-depiction techniques, such as hachures, and the corresponding decline of hill icons. The trend appeared to shift again at the end of the 19th century with the rise of pictorial maps, fostered by the expansion of the commercial map market and the arrival of color printing, which facilitated the production of colored pictograms.

A similar development was observed in the broader figurative context of mapels, with the progressive transition from iconographic hatching—widely employed to convey perspective and shading until the late 17th century—to hatched textures in the 18th century and, ultimately, to color fills at the close of the 19th century. The analysis also highlighted, for example, the gradual replacement of water hatchings with blank areas during the 17th century; the latter were subsequently supplanted by regular hatchings and waterlines at the turn of the 18th century and, from the 1880s onward, by colored fills and textures. The quantitative assessment of visual variables and annotated map forms also evidenced the overall augmentation of blank areas from the 17th century, reflecting an epistemic shift in the cartographic representation of the unknown.

These examples, along with others not revisited here, illustrate the influence of historical and cultural developments—including the advent of new printing or surveying technologies—as well as epistemic shifts, exemplified by the theorization of scientific cartography. Other factors include geographic transformations, economic dynamics, aesthetic considerations, and the need to preserve a coherent semiotic system both within individual maps and in the broader cultural context.

For instance, the advent of color printing, or bathymetry, led to the introduction of new *variant* signs. Likewise, economic dynamics tied to international trade and industrialization affected the map market and the cost of producing specific signs. Changes in the overall system imply that the relative cultural *fitness* of signs fluctuates. Furthermore, because space on the document is limited, and as a geographic feature can be rendered with only so many visual variables, and provided that signs can be accurately read and reproduced only if they remain sufficiently distinctive and do not stray excessively from semiotic conventions, the selection of a sign over another can be understood as a form of *competition*.

Processes of change which involve the selection of the most adapted variants are best described as *evolutionary*. Section 6.8 supports this perspective by showing that the frequency trajectories of signs mirror evolutionary sweeps, i.e., moments of transition in the system whose temporality coincide with the ruptures identified earlier. The recognition that signs vary under the joint pressure of external conditions and internal constraints of consistency, along with the finding that frequency trajectories are not independent from one another led to the hypothesis that signs can form *coadapted complexes*. To operationalize this construct, statistical contingencies among signs were computed, which led to the identification of eight complexes, some predominantly iconic, others consisting mainly of marks or circles, and still others of pictograms. These complexes exhibited differential expression with respect to publication year and map scale; they also appeared to be indicative of author style.

Analysis of sign distributions revealed distinctions across the geographic contexts of publication. These differences were significant at the national level, as highlighted in the case of icons and symbols. A more granular approach was also implemented, focusing on the similarities among individual urban centers (6.10, 7.5). This methodology indicated that semiotic similarity tends to reflect geographic proximity. Although the period of activity emerged as the principal determinant of resemblance, statistical modeling demonstrated the substantial effect of geographic proximity on semiotic convergence. Moreover, the size of a urban center is influential: in both cases—icons and mapels—the models predicted greater divergence among larger centers, indicating that the latter are more prone to developing distinctive, idiosyncratic figurative cultures.

RO11 The analysis of map figuration as a function of scale provides a deeper understanding of the sign space (6.10, 7.4). Following the same methodology as before, moments of rupture are identified, corresponding to points of transition in the figuration, according to scale. In the case of icons and symbols, the main rupture was observed just before 1:100,000 (ca. 1:93,200); two secondary pivots were visible around 1:4,750 and 1:671,000, respectively. For mapels, the most prominent ruptures seem to occur around 1:8,000, immediately before 1:25,000, and around 1:750,000. It is not necessarily surprising that scale affects icons and mapels—i.e. textures, lines, colors, etc.—in different ways. The main distinction for icons lies in the greater incidence of marks and vegetation at larger scales and, conversely, the higher prevalence of settlement and hill signs at smaller scales. The dynamics observed in the analysis of mapels are more complex; an example is the differential use of text forms, manifested in the greater incidence of text at smaller scales to signify landmasses. The results also showed that orographic forms—like hill icons, hachures, and terrain contours—historically spanned overlapping yet not identical map scales. This indicates that semiotic shifts can also affect—of be affected by—the evolution of cartographic functions.

The final experiment in Chapter 5 demonstrated the ambivalence of the relationship between map scale, cartographic figuration, and geographic, referential function, through a case study on the depiction of road width as a function of scale (5.4). The results indicated that, although map scale appears to influence road width only to a limited extent, historical developments, like road-widening projects initiated in the mid-19th century, are nevertheless reflected in the figuration.

### *Semiotic system & transmission*

RO12 Chapter 7 also investigates the prevalence of mapel frequencies with respect to semantic ratios. This makes it possible to study the relative evolution of semiotic consistency as a function of map scale or publication period. Specifically, the experiment on semantic univocity provided evidence of increased consistency within local semiotic systems (7.5). In other words, similar signs tend to converge locally toward a single signification; they appear locally univocal. Here, locality is considered in a broad sense, including both geographic and chronological overlap. This result supports the hypothesis that semiotic consistency functions as a selective pressure: cartographic signs are selected to limit ambiguity within individual maps, but also with respect to maps published in the same social, geographic, historical, or functional context. Thus, the production of signs is affected by the requirement for consistency with previously generated instances, and in turn modulates the reproduction of future sign variants. This finding implies that semiotic similarities, e.g. among publication centers, can be interpreted as manifestations of transmission dynamics. In this regard, is it also noteworthy that large urban centers—particularly New York, Philadelphia, Paris, and Amsterdam—are characterized by markedly lower univocity compared to smaller ones.

One can examine how the pressure for semiotic consistency affects the broader system by discussing the results on macro-diversity, micro-diversity, and overall complexity of cartographic signs (6.8). As cartography spread to new regions, including France and the Netherlands, at the beginning of the 17th century, between-map diversity—i.e., micro-diversity—increased markedly. Concretely, this implies that the diversity of signs across maps rose. This finding does not contradict the earlier results which emphasized the greater semiotic consistency at the local level: it supplements them by indicating that the implantation of cartography in new geographic areas leads to *differentiation*. The pressure for differentiation arguably balances the drive for local consistency, preventing the system from collapsing into strictly univocal semiotic modes, or globally uniform figuration.

The second increase in the proportion of active signs—i.e., macro-diversity—was observed in the 19th century, when the mechanization of map production led to larger publication output and, in parallel, to more regular and reliable sign replication. Consequently, micro-diversity and semiotic complexity within individual maps decreased, whereas macro-diversity augmented following the expansion of publication volumes.

RO13 A closer look at both specific and global transmission patterns might partly elucidate the processes underlying semiotic similarities or differences among local contexts. The mechanisms of transmission were examined in three distinct instances throughout the manuscript: twice focusing on social transmission among individuals or organizations (micro-transmission), and once through modeling macro-transmission between cities.

First, the model of semiotic diffusion throughout the social graph of map makers indicated that map creators tend to adopt semiotic conventions established by larger actors (7.5). Connectedness through collaboration—potentially via intermediaries—, geographic proximity, and overlapping periods of activity also induce semiotic similarity among map creators. These results are consistent with the findings of Chapter 2, which indicated that geography and period of activity are the principal determinants of social connectivity. The statistical redundancy possibly accounts for unobserved transmissions, e.g., through map objects or undocumented social relationships.

The qualitative inquiry into the career paths of seven early adopters, presented in Section 2.4, complements and explicates the results of the statistical model. Whereas actor size appears to be the principal determinant of similarity in the model, the examination of early adopters highlights the importance of legitimacy—rooted in social rank, authority conferred by the State, or epistemic ideals—in career success. The capacity to activate dependable distribution networks also seems an important factor, as is the ability for mapmakers to capture imagination. Therefore, the results of the two approaches reinforce one another: node size can be considered a quantitative proxy for social legitimacy, while distribution networks are reflected in node connectivity.

At a different scale, larger production centers appear to have played a particular role in the cultural transmission of signs (7.5). As anticipated, major cities exhibit a lower semantic univocity; they also appear to develop differentiated, distinctive semiotic systems. The diachronic chart of semiotic flow further indicates that these centers function as *hubs*, both capable of assimilating signs from other centers and broadcasting them. Paris, New York, and, later, Philadelphia emerge as the principal hubs over the study period, connecting almost all publication centers. These centers are designated *autopoietic hubs* because they also exhibit strong self-transmission patterns, which implies that they are able to maintain their semiotic idiosyncrasy over time. By contrast, secondary, peripheric publication centers operate as dialogue partners to larger hubs, concurrently creating signs that will be broadcast through hubs and adopting semiotic trends relayed by them. However, peripheric centers do not exhibit significant self-transmission patterns, suggesting that they fail to sustain distinctive and idiomatic semiotic systems. Overall, the capacity to emerge as a major transmission center appears to rest on a sustained self-transmission pattern, combined with efficient integration and adaptation of external signs—possibly facilitated by a lower semantic univocity. These centers are characterized by their cultural distinctiveness, with respect to the global system of cartographic figuration.

## C.2 Interpretive & theoretical standpoint

Ever since the publication of J. B. Harley's seminal essay *Deconstructing the map* (Harley, 1989), the Foucauldian notion that maps constitute a form of power-knowledge has pervaded map studies (Crampton, 2001; Crampton & Krygier, 2005; Wood & Fels, 1992). The realization that maps are political products has opened the field to the critical discussion of the role of cartography in geo-historical processes. Illustrative topics include the partitioning of land along racial boundaries (Crampton, 2007), the social construction of natural spaces (Brosnan & Akerman, 2018), and colonial place-naming practices (Williamson, 2023). Scholars have likewise examined the role of maps in the construction of modern nation-states (Batuman, 2010; Gibson, 2022; Schulten, 2012).

In many respects, the first part of this thesis engages with this perspective, underlining, for instance, the relationship between strong, centralized, and sovereign political structures, on the one hand, and sustained volumes of cartographic publication (2.3), on the other hand. Concurrently, the conceptualization of cartographic coverage as an attention mechanism is founded on the view that maps reflect political interests. Close attention, manifested through detailed mapping, can be interpreted as a marker of control over a territory, whereas distant attention primarily conveys broader spatial ideologies and conceptions of the world (3.1–3.2). This approach highlights the substantial resources mobilized by colonial powers to map their overseas territories and the relationship between dense cartographic coverage and prolonged historical control. The progressive augmentation of national self-attention—concomitant with the development of administration and the modern state—and the demonstrated dependency between Atlantic charting and the slave trade (3.3) further endorse the perspective that the history of cartography is political. Using a distinct methodology, I also highlighted the greater emphasis accorded to urban spaces compared with natural landscapes (5.2).

By acknowledging the role of agency in the construction of geographic knowledge, the socio-constructivist lens offers an interpretive framework that informs contemporary debates on space and control. This dissertation, however, argues that a critical reading of maps and cartography is essential but not sufficient to explain all the dynamics involved. This argument is primarily substantiated in Chapters 6–7, which concentrate on the observation of signs. The diffusion of millions of cartographic signs can hardly be explained by purposeful political processes alone. Rather, a large part of the visual information contained in maps can be described as the outcome of replication and the reproduction of conventions of representation. This view does not contradict the premise that political, economic, artistic, and environmental developments affect maps. It conceptualizes cartography as *a continuously changing* ensemble of practices and products, situated in, and adapted to, particular places, periods, social circles, and cultural contexts. As such, adaptation can be motivated and intentional. However, cultural replication can preserve a trait beyond its initial context of production without explicit, deliberate intervention. An eloquent example of this phenomenon is the perpetuation of horizontal document orientation, through norms and conventions, long after the inceptive processual relationship between cartography and landscape iconography had receded (5.2).

Insofar as maps function as semantic-symbolic systems, they are subject to internal constraints of space and consistency. Limited document space implies that cartographic signs, as well as geographic features, compete with one another. Here, competition should not be understood as an active quality, but as the consequence of existing limitations. Together with the imperative of semiotic consistency, these constraints imply that sign replication is not an independent and self-regulated but rather the outcome of a cultural selection process. This proposition was corroborated by the analysis of sign-frequency trajectories (6.9) and local univocity (7.5), which evidenced statistical dependencies between sign occurrences, and semiotic constraints. Selection, or prioritization, progressively contributes to the formation of cultural norms, which tend to be replicated unless other constraints, or choices, intervene. The study of map composition similarly suggested an internal competition for emphasis and centrality (5.2).

This perspective, in which signs and feature frequencies vary as the result of selective replication processes, is best described as *evolutionary*. Specifically, cultural evolution is the theoretical framework that describes change in human culture as a cumulative process of variation, selection, and transmission across generations (Henrich & McElreath, 2003; Mesoudi, 2016). Insofar as works on cultural evolution are primarily concerned with issues like dual inheritance (Rowthorn, 2011), cumulative cultural evolution (Dean et al., 2012), and the emergence of language structure (Kirby et al., 2008), employing methods ranging from simulation to ethnographic fieldwork, this dissertation constitutes more of an outsider contribution. Nevertheless, it observes—with its own methods—the same mechanisms of variation, selection, and transmission as they manifest in the context of historical cartography.

Specifically, the present research independently highlights mechanisms of competitive selection, selective sweeps—inducing moments of rupture—and patterns of social transmission. The results further suggest the existence of co-adapted sign complexes. They also evidence stabilizing mechanisms in micro-level cultural transmission, such as the pressure for univocity and normative reproduction, and the antagonist drive for differentiation, observed primarily in macro-level transmission among urban centers. The modalities of introduction of new variants, through innovation or stochastic replication drift, were also mentioned. The process of differentiation is apparent in the geographic distribution of signs, which reveals locally idiomatic systems. Consistently, the spread of mapmaking to new regions during the 17$^{th}$ century stimulated semiotic diversity. By contrast, the mechanization of reproduction processes throughout the 19$^{th}$ century induced a degree of standardization. In the latter case, overall sign diversity still increased, owing to the sharp rise in publication volumes.

Albeit graphical density tends to increase over time, there is no evidence indicating that the semiotic complexity of individual maps augments. Assuming that evolution follows a deterministic trajectory toward greater individual complexity would be a fundamental misunderstanding of its operation: although evolution *can* lead to more complex objects, it does not inevitably do so. In the particular case of maps, complexity is modulated by the tension between informational density, and readability.

The view that evolution is *undirected* aligns with the proposition that cartography evolves without "progress" (Edney, 1993). Indeed, the history of cartography cannot be understood as a mere trajectory of improvement. Expectations regarding maps, or the interpretation thereof, continuously change. While the scientific-rationalist episteme may treat geometric accuracy as a primary objective of cartographic practice, this criterion has its own history. Moreover, it must continually contend with cultural, political, economic, and artistic expectations and constraints. The very criteria for establishing scientific knowledge also shift over time. Consequently, the evolution of cartography is best understood as a dynamic process of adaptation to changing historical and cultural contexts.

Therefore, this dissertation adopts two complementary interpretive frameworks. The first construes maps as power-knowledge devices that originate from purposeful choices, within prevailing relations of power and social norms, which they ultimately contribute to reproduce and reinforce. The second treats maps as culturally adapted artifacts whose traits are socially transmitted and selectively replicated under political, economic, epistemic, cognitive, artistic, and semiotic constraints. Although the structure of the dissertation tends to separate these frameworks for the sake of argumentation, the two perspectives are not considered mutually exclusive, nor are they confined to any single level of observation. The political implications of cartography are manifest in the examination of geographic coverage. However, they are also involved in the depiction of border lines or in the choice of using terrain contours to facilitate ballistic calculations. Conversely, cultural transmission and conventions of representation operate across multiple scales of replication: from the reproduction of a single sign to the copying of a specific cosmographic view, or the representation of a particular geographic subject in a distinct way.

## C.3 Limitations

Although the research was conducted diligently and to the best of my abilities and possibilities, I acknowledge several limitations, the principal of which are discussed below.

**Data representativity.** While the research pursued an integrative strategy, the 38 digital portals examined still represent only a portion of existing data providers. Some platforms were unintentionally overlooked, while others were excluded because of access limitations, or a smaller document count. Consequently, the analysis focused primarily on large national portals, which may constitute a collection bias. Additionally, the dataset covers only a limited set of countries, most of them high-income Western nations endowed with adequate resources for digitization, cataloging, and online dissemination. Historical representativeness also depends on past collection and archiving practices for the conservation of cartographic heritage. Furthermore, digitization priorities and workflows, and ultimately the digitization and accessibility to data, hinges on curatorial choices. The corpus further bears the imprint of the individual perspectives of catalogers and their cultural-historical sensibilities. A substantial portion of Chapter 1, and the opening of Chapter 2, are devoted to the quantitative assessment and discussion of corpus biases.

**Propagation of statistical error.** Tracking statistical error across multiple processing steps is a well-documented challenge in document-processing research. Each task—recognition, embedding, clustering, and statistical analysis under variables that are themselves derived from distinct recognition, normalization, categorization, or mathematical transformation procedures—contributes to the complexity of computing the cumulated statistical error, which ultimately becomes intractable. In the present study, the analyses are therefore confined to estimating the uncertainty of the final processing stage. The consistency of treatments, the large data volumes, and the pronounced statistical effects—frequently significant by several $\sigma$—support the results viability. Smaller observed differences, however, may still constitute statistical artifacts arising from processing errors. A critical assessment of the plausibility of the findings, weighted against qualitative observations, thus remains essential to attenuate the limitations imposed by cumulated statistical uncertainty.

**Theory-ladenness, Research and discovery bias.** The corpus's size, width, and complexity preclude a systematic analysis from every perspective and across all parameters. Consequently, the choice of analyses is guided by hypotheses formulated in the literature, or derived from the theoretical framework. This situation is more likely to confirm existing hypotheses and knowledge than lead to new discoveries. Under these constraints, the potential for new findings depends largely on serendipity. Moreover, the interpretation of results is itself vulnerable to confirmation bias aligned with theoretical expectations. Employing distinct and diversified analytical approaches partially mitigates the risk that misinterpretation of a single experiment will unduly influence the final conclusions.

**Document materiality and paracontext.** Whereas historical methodologies generally rely on acquaintance with contextual sources—legal texts, epistolary correspondence, books of measurements, and so forth—the present research focuses exclusively on cartographic documents. Accordingly, the specific context of production, function, or the individual aims of the mapmaker are overlooked. Moreover, the study of digital images necessarily entails a loss of sensitivity to the document's materiality. While quantitative and digital approaches enable researchers to observe phenomena beyond the reach of manual inquiry, they cannot replace the need for direct, contextualized historical research and can only serve as a complementary line of analysis.

**Implications for contemporary and digital cartography.** Due to copyright reasons, the study ends in 1948; therefore, recent developments in cartographic figuration, or intrinsic to evolving map uses, lie beyond its scope. The transition to digital mapping and digital maps has induced significant alterations not only to the materiality of the document but also to its affordances. In dynamic web maps, for instance, because users can seamlessly zoom in and out, the very notion of scale has assumed a different sense. Similarly, the distribution and reproduction of digital maps bears little resemblance to historical map trade. Accordingly, the scope of the analysis remains confined to the historical understanding of cartography and its processes of transmission.

## C.4 Future perspectives

This dissertation has developed datasets and digital methodologies for extracting, visualizing and analyzing cartographic information, thereby opening new avenues for studying maps as cultural objects. This section outlines potential directions for future research.

**Contextual and generic map processing.** Beyond the advances already reported, the present work introduced several building blocks that can accelerate the development of context-informed, generic map-processing technologies. Contextual awareness, for instance, could be achieved through the injection of metadata about production context (e.g., year and place of publication, workshop), content (e.g., scale, geographic area depicted), and figurative style (e.g., descriptors derived from mapel-based modeling). Such approach would leverage the creation of generic training annotations, the normalization and aggregation of diverse map collections, and the operationalization of cartographic figuration itself. Complementary strategies include fine-tuning contrastive image-representation models based on metadata or building open-vocabulary segmentation models. Such techniques would, in turn, improve the latent representation of cartographic semantics, which could benefit the study of cartographic semantic-symbolic system.

**Icon detection.** The Cartographic Sign Detection Dataset is the first training set dedicated to the recognition of cartographic signs in generic contexts. Although the constraints of this work favored a relatively lightweight, fast detection model, the same training data could readily be employed to train slower yet more accurate architectures. Such models could support additional applications, including the location of archaeological remains (e.g., churches, mills, cemeteries), the analysis of land functions, and the dating of urban tree heritage.

**Stylometry and authorship attribution.** Mathematical descriptors of cartographic figuration could complement human expertise in the attribution of cartographic works. When authorship is uncertain, stylometric cues could potentially corroborate or challenge curatorial hypotheses. Furthermore, the systematic examination of departures from the social transmission model may reveal undocumented collaborations or even highlight the omission of mapmakers from library catalogs.

**Tracing the diffusion of technologies.** Several printing techniques, such as wood engraving, copper engraving, etching, lithography, and more recent processes such as Ben Day dots, are discernible in digital images of sensible quality. Although catalog metadata rarely document the printing process, targeted annotation, combined with mapel descriptors, could permit the inference of the employed technique. This approach would make it possible to trace the historical spread of printing technologies, through the concrete case of cartography. Likewise, analysis of visual map forms has demonstrated that orographic methods—such as hillshading, hachuring, and terrain contours—can be detected automatically; targeted annotation would enable a comprehensive case study of how these methods historically spread between production centers.

**Digital tools for the history of cartography.** This thesis developed methodologies for engaging with cartographic heritage across multiple scales of observation, ranging from publication volumes to individual map images and down to specific motifs or signs within those images. The creation of representation vectors effectively enables searching map heritage through visual queries. Because the data are connected across scales, it becomes possible, for example, to instantly visualize the chronological and geographic distribution of a specific map sign. The same technologies enable searching within maps by semantic similarity; with few adjustments, they could also support dynamic document segmentation. Finally, the aggregated database extracted from multiple catalogs could serve for the creation of a meta-portal, thereby facilitating cross-database searches.

**Fragment-based representation as a methodological framework.** The fragment-based approach employed to operationalize cartographic figuration is not technically restricted to map images. It could, in fact, be transposed to other contexts concerned with the study of cultural traces. Potential examples include the study of architectural details, the investigation of motifs from the decorative arts, or the analysis of landscape patterns.

**Interdisciplinary perspectives.** Finally, within a broader research framework, modeling sign distribution can complement geolinguistics studies by facilitating the analysis of semiotic systems that appear relatively independent of usual dialect boundaries. In this respect, cartography provides a practical proxy for exploring Western cultural spaces beyond linguistic borders. At the same time, the diffusion of cartographic figuration is also an indicator of technological dissemination. Consequently, historical cartography can function as a chronological bridge between anthropological investigations into the spread of early tools and technologies, and sociological research on large-scale transmission patterns in the contemporary era.

# Remi Petitpierre

Scientific assistant and PhD student at EPFL, Swiss Federal Institute of Technology in Lausanne. My research interests include the computational analysis of cartography, cultural evolution and the diffusion of ideas, quantitative and spatial approaches in the humanities, digital urban history, and historical geography.

EPFL, Lausanne, Switzerland    remi.petitpierre@alumni.epfl.ch

## EDUCATION

**PhD student in Digital Humanities** 09.2021 – 08.2025
EPFL. Thesis director: prof. Frederic Kaplan. Co-director: dr. Isabella di Lenardo

**Master of Science in Digital Humanities** 07.2020
EPFL. Class delegate. Student committee. DH Teaching committee. GPA 5.62/6.00

**Bachelor of Science in Life Sciences and Technology** 07.2018
EPFL.

**Bilingual French-German high school diploma** 07.2014
Gymnase de Morges, Switzerland. Scientific track. Award of Excellence in French and History

## EXPERIENCE

**Doctoral assistant** (100%) from 09.2021
Digital Humanities Institute, EPFL.
Research on computer vision and quantitative approaches to historical cartography. Publication and dissemination. Teaching and TA. Support to grant writing.

**Scientific assistant** (100%) 08.2020 – 08.2021
Lausanne Time Machine Initiative, EPFL.
Research on historical document processing (cadasters, census records), and digital urban history. Teaching and TA.

**Communal counsellor** 06.2016 – 06.2021
City of Morges, Switzerland.
Political work on municipal projects. From 2018, president of the finance dicastery subcommittee (auditing c.a. 82 mio CHF/y).

**Master project** (100%) 03 – 07.2020
Digital Humanities Laboratory, EPFL. Grade 6.00/6.00

**Trainee engineer** (100%) 09.2019 – 03.2020
Bibliothèque nationale de France (BnF), Paris. Dept. of Maps and Plans.
AI, computer vision, databases. Development of an experimental algorithm for vectorizing and georeferencing historical city maps.

**Academic tutor** 09 – 12.2019
University of Fribourg, Switzerland. Dept. of special education.
Computer vision and data science support to scientific projects.

**Bachelor project** 09 – 12.2017
Laboratory of Molecular & Chemical Biology of Neuro-degeneration. EPFL. Grade 6.00/6.00

**Substitute teacher** 03 – 06.2017
Public education system. Canton of Vaud.

**TEACHING, TA**

**Digital urban history** (3 hours weekly / full year) — from 09.2020
EPFL–Unil Master course in Humanities and Social Sciences. c.a. 30 students.

**Digital humanities** (2 hours weekly / half year) — from 02.2023
EPFL Bachelor course in Humanities and Social Sciences. c.a. 30 students.

**Master project and thesis** — from 02.2021
EPFL MSc in Digital Humanities (x4). MSc in Computer Science (x3).
UNIL MA in Digital Humanities (x1)

**Bachelor project** — 2022-2023
EPFL BSc in Computer Science (x2).

**PEER-REVIEWED PUBLICATIONS**

1. Petitpierre R., Uhl J. H., di Lenardo I., Kaplan F. (2024). A fragment-based approach for computing the long-term visual evolution of historical maps. *Humanities and Social Sciences Communications*. 11(363). Nature portfolio. doi: 10.1057/s41599-024-02840-w
2. Vaienti B., Petitpierre R., di Lenardo I., Kaplan F. (2023). Machine-Learning-Enhanced Procedural Modeling for 4D Historical Cities Reconstruction. *Remote sensing*. 15(13), 3352. MDPI. doi: 10.3390/rs15133352
3. Petitpierre R., Kramer M., Rappo L. (2023). An end-to-end pipeline for historical censuses processing. *International Journal on Document Analysis and Recognition (IJDAR)*. vol. 26, 419–432. Springer. doi: 10.1007/s10032-023-00428-9
4. Petitpierre R., Guhennec P. (2023). Effective annotation for the automatic vectorization of cadastral maps. *Digital Scholarship in the Humanities*. 38(3):1227–1237. Oxford University Press. doi: 10.1093/llc/fqad006
5. Petitpierre R., Kaplan F., di Lenardo I. (2021). Generic Semantic Segmentation of Historical Maps. In CHR 2021: *Computational Humanities Research Conference*. CEUR. Amsterdam, NL.17-19 Nov. 2021. 2989:228–48. ceur-ws.org/Vol-2989/long_paper27.pdf

**ACADEMIC WORKS**

1. Petitpierre R. (2025). *Studying Maps at Scale: A Digital Investigation of Cartography and the Evolution of Figuration*. Tentative PhD thesis, EPFL.
2. Petitpierre R. (2020). *Neural networks for semantic segmentation of historical city maps: Cross-cultural performance and the impact of figurative diversity*. Master's thesis, EPFL. doi: 10.13140/RG.2.2.10973.64484

**BOOK**

1. di Lenardo I., Petitpierre R. (2025). *Lausanne 1831*. Silvana Ed. 358 pages.

**CHAPTERS, PROCEEDINGS, PREPRINTS**

1. Lin Y. et al. (2025). ICDAR 2025 Competition on Historical Map Text Detection, Recognition, and Linking. In ICDAR 2025: *19th International Conference on Document Analysis and Recognition*. Wuhan, CN. 16-21 Sep. 2025. doi: 10.1007/978-3-032-04630-7_33
2. Zou M., Dai T., Petitpierre R., Vaienti B., Kaplan F., di Lenardo I. (2025). Recognizing and Sequencing Multi-word Texts in Maps Using an Attentive Pointer. *In Review*. doi: 10.21203/rs.3.rs-6330456/v1
3. Petitpierre R. (2025). La machine qui lisait les cadastres. In di Lenardo I., Petitpierre R. (Eds.) *Lausanne 1831*. 17–19. Silvana Ed.
4. Petitpierre R. (2025). Le conte des deux cités. In di Lenardo I., Petitpierre R. (Eds.) *Lausanne 1831*. 39–41. Silvana Ed.
5. di Lenardo I., Petitpierre R. (2025). L'agencement multifonctionnel de la ville et de la campagne. In di Lenardo I., Petitpierre R. (Eds.) *Lausanne 1831*. 42–48. Silvana Ed.
6. Rappo L., Petitpierre R. (2025). Le plan cadastral Berney dans le contexte vaudois, suisse et européen. In di Lenardo I., Petitpierre R. (Eds.) *Lausanne 1831*. 12–16. Silvana Ed.
7. Petitpierre R., di Lenardo I., Rappo L. (2024). Revealing the Structure of Land Ownership through the Automatic Vectorisation of Swiss Cadastral Plans. In Baudry J., Burkhart L., Joyeux-Prunel B. et al. (Eds.) *DigiHistCH24 Book of Abstracts*. Basel, CH. 12-13 Sep. 2024. digihistch24.github.io/book-of-abstracts/submissions/456/

## WORKING PAPERS

1. (working paper) Petitpierre R. et al. (2025). A Layer of Late Victorian London: Extracting the Building Footprints from the 1:1,056 Ordnance Survey Map (1891-1896).

## CONFERENCE AND WORKSHOP PRESENTATIONS

1. *A computational and spatial exploration of historical cartography*. Conférence finale du trimestre IA et Humanités Numériques. CNRS. Paris. Dec. 10-12, 2024. Invited speaker.
2. *Revealing the Structure of Land Ownership through the Automatic Vectorisation of Swiss Cadastral Plans*. DigiHistCH24 – Digital History Switzerland 2024. University of Basel, CH. Sep. 12-13, 2024. Speaker.
3. *Studying the Spatial and Temporal Evolution of Occupations in Lausanne*. RELEX Workshop: Registry of Firms' Life and Exit. Università della Svizzera italiana, Lugano, CH. Sep. 3, 2024. Invited speaker.
4. *Exploring the evolution of cartographic representation at scale through visual computing*. ICHC 2024 – 30th International Conference on the History of Cartography. Université Jean Moulin Lyon 3 and Imago Mundi. Jul. 1-5, 2024. Speaker.
5. *Mapping 200 years of Paris: an algorithmically-enhanced approach to urban history*. ICHC 2024 – 30th International Conference on the History of Cartography. Université Jean Moulin Lyon 3 and Imago Mundi. Jul. 1-5, 2024. Speaker.
6. *Qui travaille dans ma rue? Les métiers lausannois en 1832*. Festival Histoire et Cité. Université de Genève, Université de Lausanne, Université de Neuchâtel, and Musée cantonal d'archéologie et d'histoire de Lausanne. Apr. 15-21, 2024. Speaker, with Lucas Rappo and Isabella di Lenardo.
7. *Des cartes au modèle 4D. Des sources pour catalyser la reconstruction historique des villes.* Humanité Numérique et IA. Study day. Association française pour l'Intelligence Artificielle. Paris. May. 3, 2024. Invited speaker.
8. *Machine Reading for Demographic Time Series: the evolution of occupations in Lausanne Records 1805-1898*. 5th Conference of the European Society of Historical Demography (ESHD). ESHD and Radboud University, Nijmegen, NL. Aug. 30 – Sep. 2, 2023. Speaker, with Lucas Rappo.
9. *Protocartographie et approches sans carte : réimaginer la géographie incertaine du passé.* Approches numériques de l'histoire urbaine médiévale : problèmes et méthodes. Study days. Université de Lausanne. Nov. 9-10, 2023. Invited speaker, with Isabella di Lenardo.
10. *Mapping Memes in the "Napoleonic" Cadastre: Expanding Frontiers in Cartographic Stylometry*. Digital Humanities Conference 2023. Alliance of Digital Humanities Organizations and University of Graz. Jul. 10-14, 2023. Speaker.
11. *From Automated Bootstrapping to Collaborative Editing: A Framework for 4D City Reconstruction*. Digital Humanities Conference 2023. Alliance of Digital Humanities Organizations and University of Graz, AT. Jul. 10-14, 2023. Speaker, with Beatrice Vaienti, Paul Guhennec, and Didier Dupertuis.
12. *Recartographier l'espace napoléonien : Une lecture computationnelle du cadastre historique de Lausanne.* Colloque Humanistica 2023. University of Geneva. Jun. 26-28, 2023. Speaker, with Lucas Rappo and Isabella di Lenardo.
13. *Du document à l'information. La segmentation des documents historiques avec dhSegment et nouvelles perspectives sur d'autres méthodes.* Digital Methods in the Humanities. Summer School. Université de Lausanne. June 12-15, 2023. Invited lecturer, with Isabella di Lenardo.
14. *SuperMapRealigner, A Generic Method for Segmented Map Realignment*. MapReader Launch. Alan Turing Institute, London. Jun. 7-8, 2023. Invited speaker.
15. *Modelling Past Cities: Forward the Time Machine.* 3D Models in Digital Cultural Heritage: Current Trends and Experimental Applications. Master's course. University of Basel, CH. May 23, 2023. Invited lecturer, with Beatrice Vaienti.

16. *The Memetics of Cartography: Computing Long-term Cultural Evolution through the Lens of Historical Maps*. dhX22 – 2nd Swiss Digital Humanities Exchange. Université de Lausanne. Nov. 26-27, 2022. Speaker.
17. *Vocabulaires contrôlés pour l'annotation des images : de l'objet au thème*. Journée campus Richelieu – Segmenter et annoter les images : déconstruire pour reconstruire. Institut national d'histoire de l'art, Paris. Nov. 15, 2022. Invited speaker, with Federico Nurra.
18. *Lausanne Time Machine : IA & traitement des sources historiques*. Research Seminar in Modern History. Université de Lausanne. Oct. 10, 2022. Invited lecturer, with Lucas Rappo and Isabella di Lenardo.
19. *Effective Annotation: How to Leverage Manual Work for the Automatic Vectorization of Cadastral Maps*. EAUH2022 – 15th Conference of the European Association for Urban History. University of Antwerp, NL. Aug. 31 – Sep. 3, 2022. Speaker, with Paul Guhennec.
20. *Recartographier le passé: Une infrastructure pour la Time Machine*. 1st Seminar BnF-SoDuCo – Social Dynamics in Urban Context. Bibliothèque nationale de France, Paris. Apr. 4, 2022. Invited speaker.
21. *dhSegment: a Generic Deep-learning Approach for Document Segmentation*. Time Machine Academies. Time Machine Organization, remote. May 20, 27, and June 3, 2021. Invited speaker, with Frederic Kaplan.
22. *Generic Semantic Segmentation of Historical Maps*. Computational Humanities Research Conference 2021. Amsterdam and remote. Nov. 17-19, 2021. Speaker.
23. *JADIS : Se promener dans les géodonnées du passé*. L'intelligence artificielle à la BnF. Bibliothèque nationale de France, Paris and remote. Jun. 14 2021. Invited speaker.
24. *Lausanne Time Machine Initiative*. Pint of Science Switzerland. Lecture. May 10, 2021. Invited lecturer, with Irene Bianchi.
25. *Workshop on dhSegment data extraction from texts*. ARCHiVE Online Academy. Fondazione Giorgio Cini, Venice and remote. Nov. 27, Dec. 4, and 11, 2020. Invited speaker, with Frederic Kaplan and Isabella di Lenardo.

---

## CONFERENCE AND WORKSHOP ORGANISATION

*DHMAPS: Digitizing Historical Maps. Apr. 25-26, 2024. Lausanne. archive.ph/mWEkj.*
*dhX22 – 2nd Swiss Digital Humanities Exchange. Nov. 26-27, 2022. Lausanne.*

---

## OTHER CONFERENCE AND WORKSHOP ATTENDANCE

*100 years of the Institute of Cartography and Geoinformation, Sep. 4-5 2025. Zurich.*
*2nd Seminar BnF-SoDuCo – Social Dynamics in Urban Context, Nov. 10, 2022. Paris.*
*Time Machine Conference 2021, Nov. 22-23, 2021. Remote.*
*Séminaire "Des sources aux SIG", 2021. Paris and remote.*
*dhX19 – 1st Swiss Digital Humanities Exchange. Feb. 22-23, 2019. Basel.*
*Time Machine Conference 2018, Oct. 30-31, 2018. Lausanne.*

---

## FUNDING ACQUISITION

- Selected for a **two-year fully funded PhD scholarship (129,850 CHF)** by the College of Humanities at EPFL based on the quality of the doctoral project proposal.

---

## AWARDS AND NOMINATIONS

- PhD thesis nominated by the jury for the **EPFL Doctorate Award**.
- **Winner (ex-aequo) of the Competition on Historical Map Text Detection**, Recognition, and Linking organized by the 19th International Conference on Document Analysis and Recognition (ICDAR 2025).
- Finalist for the **Paul Fortier Prize 2023** of the Alliance of Digital Humanities Organizations, rewarding the best paper and presentation by an emerging scholar.

**OPEN CODE REPOSITORIES AND DATASETS**

1. Petitpierre R. (2025). *Aggregated Database on the History of Cartography (ADHOC)*. v1.0.0. doi: 10.5281/zenodo.16277852 Zou M., Petitpierre R., di Lenardo I. (2025). *London 1890s Ordnance Survey Text Layer*. v0.1.0 Pre-release. doi: 10.5281/zenodo.14982946
2. Petitpierre R., Gomez Donoso D., Krisel B. (2025). *Semantic Segmentation Map Dataset (Semap)*. v1.0.0. doi: 10.5281/zenodo.16164781
3. Petitpierre R., Jiang J. (2025). *Cartographic Sign Detection Dataset (CaSiDD)*. v1.0.0. doi: 10.5281/zenodo.16278380
4. Zou M., Petitpierre R., di Lenardo I. (2025). *London 1890s Ordnance Survey Text Layer*. v0.1.0 Pre-release. doi: 10.5281/zenodo.14982946
5. Dai T., Johnson K., Petitpierre R., Vaienti B., di Lenardo I. (2025). *Paris and Jerusalem City Maps Text Dataset*. v1.0.0 Initial release. doi: 10.5281/zenodo.14982662
6. Petitpierre R., Jan M., Guhennec P. (2023). *SuperMapRealigner: A Generic Method for Segmented Map Realignment using Local Features and Graph Neural Network*. v0.1 Initial release. doi: 10.5281/zenodo.638211649
7. Rappo L., Petitpierre R., Kramer M. (2023). *Lausanne Historical Censuses Dataset HTR 35k*. Zenodo. doi: 10.5281/zenodo.7711178
8. Petitpierre R., Rappo L., Kramer M., di Lenardo I. (2023). *1805-1898 Census Records of Lausanne: a Long Digital Dataset for Demographic History*. Zenodo. doi: 10.5281/zenodo.7711640
9. Petitpierre R. (2021). *Historical City Maps Semantic Segmentation Dataset*. Zenodo. doi: 10.5281/zenodo.5497934
10. Petitpierre R. (2020). *Projet JADIS*. v1.0-beta. Github. doi: 10.5281/zenodo.6594483

**PUBLIC DISSEMINATION AND POPULARIZATION**

1. (2022). *Lausanne 1830 - Histoires de Registre*. Educational Video Game. lausanne1830.ch, with Saara Jones, Andrew Dobis, Sarah Jallon, Yannick Rochat, Selim Krichane, Lucas Rappo, Isabella di Lenardo and Simon Dumas-Primbault.
2. Petitpierre R. (2020). *Projet JADIS*. v1.0-beta. Github. doi: 10.5281/zenodo.6594483
3. Barman R., Guhennec P., Petitpierre R., Kaplan F. (2020). *Vectorisation automatique des cadastres du Canton de Neuchatel*.. Collaboration report. Swiss Cadastre and DHLAB EPFL.
4. Petitpierre R. (2020). *Automatic Extraction of Lausanne Cadastres Using Artificial Intelligence*. Lausanne Time Machine EPFL-Unil. www.epfl.ch/schools/cdh/lausanne-time-machine/automatic-extraction-of-lausanne-cadastres-using-artificial-intelligence
5. Petitpierre R. (2020). *4000 cartes par jour: des réseaux de neurones artificiels pour récupérer les géodonnées du passé*. Hypothèses, Carnet de la recherche de la Bibliothèque nationale de France. bnf.hypotheses.org/9676

**SUPERVISION OF STUDENT PROJECTS**

1. Jiang, Jiamin (2024). *Cartonomics*. EPFL, MSc in Digital Humanities, Semester project.
2. Dai, Tianhao (2024). *Toponym-assisted Georeferencing for Historical Maps*. EPFL, MSc in Digital Humanities, Semester project.
3. Zou, Mengjie (2024). *Automated Toponym Extraction Pipeline for Historical Maps*. EPFL, MSc in Digital Humanities, Semester project.
4. Kauffmann, Titaÿna (2023). *La reconnaissance automatique de l'écriture manuscrite et les cahiers de Jean-Henri Polier de Vernand (1715–1791)*. Université de Lausanne, MA in Digital Humanities, Master's thesis.
5. Rieupouilh, Louise (2023). *The Cultural Microscope*. EPFL, BSc in Computer Science, Bachelor project.
6. Jan, Maxime (2022). *A generic method for cartographic realignment using local feature matching: towards a computational urban history*. EPFL, MSc in Digital Humanities, Master's thesis.
7. Wang, Danyang (2022). *Automatic vectorization of the renovated cadastre of Lausanne*. EPFL, MSc in Computer Science, Semester project.
8. Kang, Tianqu (2022). *Automatic vectorization and quantitative analysis of XIXth century maps of Paris*. EPFL, BSc in Computer Science, Bachelor project.
9. Damiani, Théo (2022). *Automatic Georeferencing of Historical Cadastres*. EPFL, MSc in Computer Science, Semester project.
10. Camara, Boubacar (2021). *Icono Lausanne: Reorientation of the photographic archives of the Historical Museum*. EPFL, MSc in Computer Science, Semester project.

## SUPERVISION OF STUDENT ASSISTANTS

1. Zou, Mengjie. *Automated extraction of toponyms from historical city maps*. 10.2024 – 12.2024
2. Dai, Tianhao. *Automated extraction of toponyms from historical city maps, data labeling*. 07.2024 – 10.2024
3. Johnson, Kaede. *Data labeling for text extraction from historical maps*. 11.2023 – 07.2024
4. Gomez Donoso, Damien. *Data curation of the renovated cadastre of Lausanne, labeling for historical maps semantic segmentation*. 11.2023 – 05.2024
5. Kriesel, Ben. *Data labeling for historical maps semantic segmentation*. 11.2023 – 01.2024
6. Kauffmann, Titaÿna. *Data curation of the historical censuses of Lausanne 1804–1898*. 03.2022 – 06.2023

## PEER REVIEWING

**2025**. Digital Humanities Conference (x4); npj Cultural Heritage

**2024**. Scientific Reports; Built Heritage; Geo-Spatial Information Science; Transactions in GIS; Digital Scholarship in the Humanities; Digital Humanities Conference (x4); Digital History Switzerland (x2)

**2023**. Humanities and Social Science Communications; Digital Scholarship in the Humanities

**2022**. IWEG2023 The Sixth International Workshop on Environment and Geoscience

## ASSOCIATIVE EXPERIENCE

**Vice-president**, dhelta Unil-EPFL — 12.2021– 03.2025
Digital Humanities Association of EPFL and the University of Lausanne.

**Scout Leader** — 09.2012 – 12.2017
Responsible from a team of 5 instructors (from 2013). Weekly organisation and animation of activities for 30+ children. Training young people (15-17 y.o.) to the role of instructor.

## LANGUAGE SKILLS

| | |
|---|---|
| **French** | Native |
| **English** | C1+. Proficient |
| **German** | B2+. Intermediate to Fluent |
| **Italian** | A2. Passive knowledge / Beginner |

## PROGRAMMING

| | |
|---|---|
| **Python** | Professional |
| **JavaScript/HTML** | Intermediate |
| **C++** | Intermediate |
| **MATLAB** | Intermediate |

## PHYSICAL MOBILITY

| | |
|---|---|
| **Lausanne, Switzerland**. EPFL | from 09.2014 |
| **Paris, France**. Bibliothèque National de France (BnF) | 09.2019 – 03.2020 |
| **Fribourg, Switzerland**. University of Fribourg | 09 – 12.2019 |
| **Bensheim, Germany**. Goethe Gymnasium | 04 – 07.2012 |

## DISCIPLINARY MOBILITY

| | |
|---|---|
| **Cultural analytics, Cultural evolution, Social transmission** | from 2022 |
| **Cartography, Historical geography, Urban studies** | from 2019 |
| **Digital humanities, Computational humanities** | from 2018 |
| **Life Sciences** | 2014 – 2018 |

## EDUCATION – COURSES OVERVIEW

| | |
|---|---|
| **Digital Humanities** (108 ECTS) | Digital humanities, Quantification of UX, Cultural data sculpting, Digital musicology Measuring literature, Critical digital humanities |
| **Maths** (34 ECTS) | Analysis, Probabilities and statistics, Linear algebra |
| **Computer Science** (33 ECTS) | Virtual reality, Mathematical and computational modelling, Computer vision, Programming, Numerical analysis, Information calculation communication |
| **Data Science** (23 ECTS) | Deep learning, Machine Learning, Applied data analysis, Computational social media |
| **Humanities** (22 ECTS) | Approaches in linguistics, Cognitive psychology, Social ethics, Epistemology |
| **Others** | Life Sciences (42 ECTS), Physics and mechanics (34 ECTS), Engineering (22 ECTS), Chemistry (21 ECTS) |